\documentclass[10pt,twocolumn,letterpaper]{article}
\usepackage[accsupp]{axessibility}  % Improves PDF readability for those with disabilities.

\usepackage{wacv}              % To produce the CAMERA-READY version
\usepackage{tabularray}
\usepackage{array}
\usepackage{algorithm}
\usepackage{algpseudocode}  %
\usepackage{amsmath}
\usepackage{capt-of}

\usepackage{microtype}
\renewcommand{\paragraph}[1]{\vspace{.0em}\noindent\textbf{#1}}
\usepackage{soul}
\setuldepth{foobar}
\usepackage[table]{xcolor}

\usepackage{booktabs}
\usepackage{multirow}
\usepackage{array}
\usepackage{graphicx}
\usepackage{enumitem}
\definecolor{wacvblue}{rgb}{0.21,0.49,0.74}
\usepackage[pagebackref,breaklinks,colorlinks,allcolors=wacvblue]{hyperref}
\newcommand{\dataname}{\textsc{GigaBench}}

\newcommand{\imageone}{Venice} 
\newcommand{\imagetwo}{Statuario} 
\newcommand{\imagethree}{Calacatta} 
\newcommand{\imagefour}{Br. Can.} 
\newcommand{\imagefive}{Kidney} 
\newcommand{\imagesix}{Munich} 
\newcommand{\imageseven}{Androm.} 
\newcommand{\imageeight}{Italy} 
\newcommand{\imagenine}{Lynx} 
\newcommand{\imageten}{Chores} 

\newcommand{\rothead}[1]{\rotatebox{0}{#1}}

\definecolor{implicit}{HTML}{0072B2}
\definecolor{hybrid}{HTML}{009E73}
\definecolor{explicit}{HTML}{D55E00}
\definecolor{classic}{HTML}{7F7F7F}
\newcommand{\Nparticipants}{170}

\def\wacvPaperID{106} % *** Enter the WACV Paper ID here
\def\confName{WACV}
\def\confYear{2027}

\title{Benchmarking Learnable Representations for Gigapixel Image Compression and Super-Resolution}

\author{Matteo Fusconi$^1$, Alex Costanzino$^1$, Pierluigi Zama Ramirez$^2$, Luigi Di Stefano$^1$\\
$^1$ CVLab, University of Bologna \\
$^2$ Ca’ Foscari University of Venice %\\
}

\begin{document}
    \maketitle
    \begin{abstract}
    Learnable Representations (LRs) can fit parametric families of functions to diverse kinds of signals. 
    These emergent neural-native data formats are gathering interest in the scientific community due to their compactness, resolution-agnosticism and amenability to direct neural processing.
    On the other hand, gigapixel images are becoming increasingly widespread across a variety of scientific, cultural, and industrial domains.
    In this paper, we present \dataname{}, the first benchmark and experimental evaluation focused on deploying LRs with gigapixel imagery. 
    \dataname{} includes images collected, acquired, and synthetically rendered across diverse domains, and addresses image compression and super-resolution.
    We adapt a representative set of LRs, spanning Implicit, Hybrid, and Explicit methods, so as to operate reliably at the gigapixel scale on consumer hardware.
    Through systematic evaluation across reconstruction quality, model size, compression and inference times, we identify consistent trends across methods and suggest actionable guidance for future investigation. The project page is available \href{https://matteofusconi.github.io/GigaBench}{here}.
\end{abstract}

    \section{Introduction}
\label{sec:introduction}
    The emergence of Learnable Representations (LRs) has opened up a different paradigm for image storage, reconstruction and processing.
    Rather than representing an image as a discrete grid of pixel values, LRs enable encoding it as a continuous function, defined by a set of optimizable parameters, that maps spatial coordinates to color values.
    Prominent variants of LRs include Implicit Representations \cite{sitzmann2019siren, liu2024finer, costanzino2026fdho}, which encode an image in the weights of an MLP; Hybrid Representations \cite{mueller2022instantngp, wang2025metricgrids, chen2023neurbf}, which additionally incorporate spatially localized learnable features; and Explicit Representations \cite{zhang2024gaussianimage, li2025smartsplat}, which represent images using Gaussian primitives.
    LRs have attracted significant attention for their compactness, resolution-agnosticism and amenability to direct neural processing.
    As a matter of fact, several recent works \cite{dupont2022functa, ramirez2023nf2vec, gielisse2025e2eclass, ballerini2026weightspace} have shown how the learned parameters can be fed into specifically designed meta-networks to perform downstream tasks like classification and retrieval, without requiring reconstruction of the original discrete signal.
    Such property positions LRs as a pathway to a new generation of neural-native data formats.
    
    On the other hand, digital imaging is undergoing a quiet resolution revolution, as gigapixel images are becoming increasingly common across a variety of scientific, cultural, and industrial domains, from whole-slide pathology imaging \cite{brancati2022bracs, deng2025kpis2024challengeadvancing} and satellite remote sensing \cite{copernicus_browser2026} to high-fidelity museum digitisation and panoramic photography \cite{martin}.
    Yet, the paradigms and tools for handling images at this scale remain largely anchored to classical pipelines, designed in an era when such resolutions were inconceivable.
    Hence, we pose a timely research question concerning whether and how LRs can be effectively deployed for gigapixel-scale images, and which of their variants show the greatest promise for enabling adoption within this increasingly relevant regime.
    However, despite the favourable properties of LRs and the increasing adoption of gigapixel imagery, the current literature lacks a comprehensive and rigorous evaluation of LR methods on gigapixel images.
    Indeed, the reported experimental findings primarily concern datasets with much lower and more conventional resolutions, typically ranging from less than one megapixel to a few megapixels, where the computational cost of fitting LRs remains easily tractable.
    The gigapixel regime, however, introduces fundamentally different challenges, as fitting a single image may require hours of training, naive partitioning strategies introduce boundary artifacts, and memory constraints force architectural compromises that do not arise at standard resolutions.
    Adapting existing LRs to this setting is therefore a non-trivial task.
    
    In this paper, we introduce \dataname{}, a benchmark specifically designed to address the aforementioned scientific questions and foster research into the design of LR approaches suitable for gigapixel imagery.
    We curate a collection of gigapixel images across diverse domains and evaluate LRs on compression and super-resolution under controlled settings.
    Our evaluation spans the three main families of approaches: Implicit, Hybrid and Explicit Gaussian-based Representations.  
    Our analysis reveals several consistent trends across the three families.
    Hybrid representations achieve the highest reconstruction fidelity but are structurally limited to moderate compression ratios, as their grid-based components impose a memory floor that prevents compression at extreme rates.
    On the other hand, they neatly outperform other LR approaches when it comes to super-resolution, although a human evaluation suggests that the merits of LR at the gigapixel scales are somewhat uncertain compared to classical methodologies, such as bicubic interpolation.
    Implicit representations, by contrast, can provide high to extreme compression ratios, trading fidelity for compression.
    Explicit representations exhibit the most variable intra-family behaviours in both compression and super-resolution, with performance depending heavily on the chosen parametrisation.
    Across all families, no single approach dominates uniformly as the optimal choice depends on the target compression or super-resolution ratio, as well as on the quality-speed trade-off. 
    
    Our contributions can be summarised as follows: (i) we introduce a curated benchmark of gigapixel images spanning 10 scenes and different domains, released publicly to foster future research; (ii) we provide a systematic comparison of representative Implicit, Hybrid, and Explicit representations to gigapixel image compression and super-resolution; (iii) we characterise the practical trade-offs among the specific methods evaluated across fidelity, speed, and downstream usability, offering concrete guidance for researchers and practitioners working in this emerging regime.

    \section{Related Work}
\label{sec:related_work}
    Learnable Representations are a continuous signal representation paradigm in which a function parametrized by a set of weights is optimized to map coordinates to the values of a target signal. 
    This abstraction is broad enough to cover image ~\cite{chen2021learning}, video~\cite{chen2021nerv}, and audio reconstruction ~\cite{sitzmann2019siren}, neural radiance fields ~\cite{mildenhall2020nerf}, SDF fitting~\cite{park2019deepsdf}, inverse problem solving, and more. 
    In this work, we focus on images. 
    Their main advantages include resolution agnosticism, a parameter count decoupled from sampling density, and full differentiability of the input domain. 
    We divide LRs into three main categories, described in the following sections: Implicit Representations (IRs), Hybrid Representations (HRs), and Explicit Representations (ERs).

    \paragraph{Implicit Representations.}
        Implicit representations~\cite{xie2022neural} parametrize the function as a Multi-Layer Perceptron.
        Their principal drawback is the spectral bias of MLP-based approximators, which favors low frequencies and impedes fitting fine-grained signal content~\cite{rahaman2019spectral}.
        Following the taxonomy of~\cite{essakine2025wheredowestand}, IRs differ along two axes: coordinate encoding and activation design, and network architecture.
        On the first, random Fourier feature mappings lift input coordinates into a high-dimensional sinusoidal basis before a standard MLP~\cite{tancik2020fourfeat}, while a complementary line replaces ReLU with intrinsically high-frequency activations: periodic sinusoids~\cite{sitzmann2019siren}, complex Gabor wavelets~\cite{saragadam2023wire}, Gaussians~\cite{ramasinghe2022gauss}, variable-period sinusoids~\cite{liu2024finer}, and oscillator-based designs~\cite{costanzino2026fdho}. 
        More recent methods select activations adaptively from a learned dictionary~\cite{jayasundara2025mire}.
        On the second, the inductive bias is moved into the architecture: multiplicative filter networks~\cite{fathony2021multiplicative} replace additive layer composition with element-wise products of Fourier and Gabor filters, BACON~\cite{lindell2021bacon} adds explicit per-layer bandwidth limits for multi-scale representation, and INCODE~\cite{Kazerouni2024incode} couples the MLP with a conditioning sub-network that adapts activation parameters to the input.
        Beyond signal fitting, IRs are increasingly used as a data representation in their own right, with the network weights as the object of downstream learning.
        Functa~\cite{dupont2022functa} treats each datum as a function, learning a shared prior via per-sample modulations of a meta-learned backbone for classification and generation in function space.
        ~\cite{deluigi2023inr2vec} and~\cite{ramirez2023nf2vec} instead embed the weights of independently fitted IRs into a compact latent space, enabling retrieval, segmentation, and generation without reconstructing the signal.
        Recently,~\cite{gielisse2025e2eclass} fits IRs end-to-end with direct classifier feedback, addressing the structural inconsistency of weights optimized without downstream supervision.
    
    \paragraph{Hybrid Representations.}
        As the target signal grows in size and bandwidth, the spectral bias of pure IRs becomes increasingly difficult to overcome through activation and encoding design alone.
        This motivates hybrid representations, which pair a lightweight MLP decoder with a learnable explicit structure, such as grids, hash tables, or basis functions, that injects spatial diversity and localizes capacity where the signal demands it.
        ACORN~\cite{martel2021acorn} adaptively partitions the input domain into blocks, each assigned a small MLP, refining the partition to match local signal complexity.
        MINER~\cite{saragadam2022miner} represents the signal through a Laplacian pyramid of per-block coordinate networks, yielding a coarse-to-fine multi-scale decomposition.
        Instant-NGP~\cite{mueller2022instantngp} couples a compact MLP with a multi-resolution hash grid of trainable feature vectors, achieving orders-of-magnitude speed-ups at the cost of hash collisions.
        NeuRBF~\cite{chen2023neurbf} generalizes the regular-grid prior to adaptive radial basis functions whose centers and shapes follow the signal, further composed with multi-frequency sinusoids to recover fine details.
        MetricGrids~\cite{wang2025metricgrids} extends the grid paradigm with elementary grids defined in distinct metric spaces, acting as higher-order Taylor terms that approximate strongly non-linear signals more faithfully than linear interpolation alone.
            
    \paragraph{Explicit  Representations.}
        Explicit representations parametrize a signal as a collection of graphic primitives placed directly in the signal domain, with their location and shape exposed as optimization variables.
        They have been investigated under a variety of primitives, including Gabor wavelets~\cite{wurster2024gabor}, triangles~\cite{held2025triangle, held20243d}, and anisotropic Gaussians~\cite{kerbl20233dgs}.
        Among these, Gaussians adapt most naturally to image fitting: GaussianImage~\cite{zhang2024gaussianimage} reduces 3D Gaussian Splattings~\cite{kerbl20233dgs} to eight parameters per primitive and a tile-based rasterizer operating at $10^3$ FPS.
        Subsequent work scales the paradigm to ultra-high-resolution inputs via feature-aware sampling~\cite{li2025smartsplat} and multi-scale seeded parametrizations~\cite{pan2026sgi}, with further variants refining initialization and density control~\cite{zeng2025instantgi}.
    
    \paragraph{Positioning.}
        Despite the breadth and maturity of these methods, existing evaluations most commonly rely on Kodak~\cite{kodak} and DIV2K~\cite{div2k}, datasets originally designed for classical compression benchmarking rather than for assessing neural representations.
        Yet even these are not used consistently, as different works evaluate on different subsets, at different resolutions, under different compression ratio definitions, and with different metrics, making cross-paper comparisons unreliable even when the data is the same.
        No prior work has systematically evaluated these methods at a gigapixel scale, a regime that introduces qualitatively distinct challenges in fitting cost, memory, and rate-distortion behaviour.
        \dataname{} addresses that gap.

    \section{Benchmark}
\label{sec:benchmark}
    
    \subsection{Data Collection}
\begin{figure*}[ht]
    \centering
    \includegraphics[width=\linewidth]{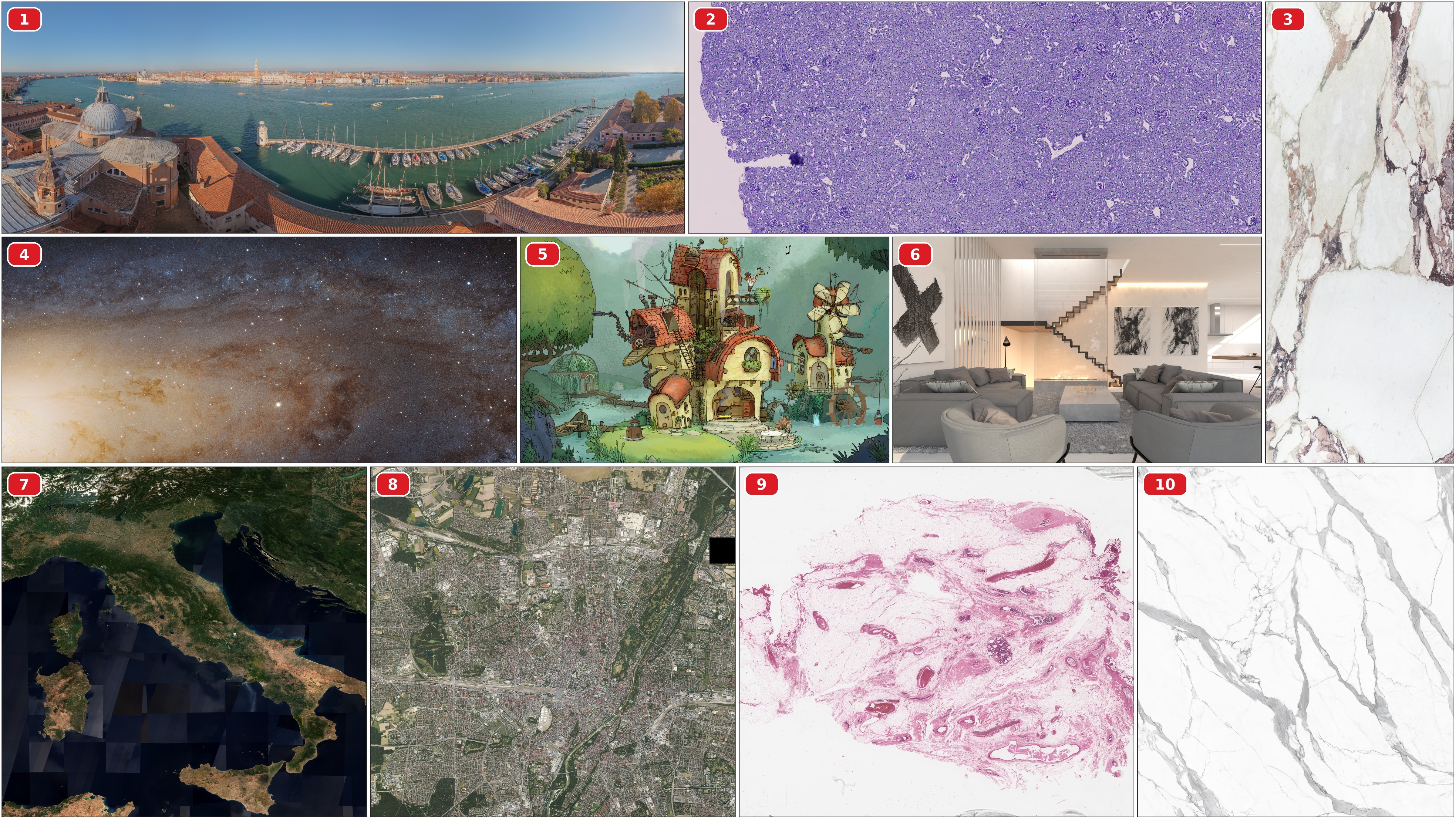}
    \caption{
        \textbf{The images in \textsc{GigaBench}.} 
        (1) \textit{Venice}, (2) \textit{Kidney}, (3) \textit{Calacatta Viola}, (4) \textit{Andromeda}, (5) \textit{House Of Chores}, (6) \textit{Lynxsdesign}, (7) \textit{Italy}, (8) \textit{Munich}, (9) \textit{Breast Cancer}, (10) \textit{Statuario}.
        }
    \label{fig:gallery}
\end{figure*}

        \dataname{} comprises $10$ images, curated from publicly available sources according to four design principles:
        \begin{itemize}[noitemsep, nolistsep]
            \item \textbf{Gigapixel Resolution.} 
            Each image meets or marginally exceeds the one-gigapixel threshold (in practice, $1.0$--$1.19$ GPx). 
            We enforce this floor so the benchmark deliberately clears the symbolic gigapixel mark, and caps it both for uniformity across the set and to keep fitting feasible on consumer hardware;
            \item \textbf{Domain Diversity.} 
            The images span panoramas, industrial scans, scientific histopathology slides, satellite imagery at varying scales, and synthetic images. 
            This range addresses several domains where gigapixel images are becoming increasingly relevant; %
            \item \textbf{Lossless Appearance.} 
            Many gigapixel images distributed online are stored as JPEG files and thus carry compression artifacts. 
            We exclude these images: since each representation method introduces artifacts of its own, pre-existing ones would add up and be difficult to disentangle from reconstruction errors;
            \item \textbf{Licensing.} 
            We retain only images whose original license permits research use and redistribution. 
        \end{itemize}
        The dataset is released as a bundle under the Creative Commons Attribution-NonCommercial 4.0 license, with attribution to this work, plus the individual attributions to the original authors of the images.
        \cref{fig:gallery} depicts the images comprising our benchmark, while \cref{tab:gigabench} reports their metadata, with more details provided in the Supplementary. 
    
    \begin{table}[ht]
    \centering
    \resizebox{\linewidth}{!}{
        \begin{tabular}{ccccccc}
    
            \toprule
        
            \textbf{Name} & \textbf{Domain} & \textbf{Resolution (H,W)} & \textbf{GPx} & \textbf{Size [GB]} & \textbf{Sub-Licence} & \textbf{Attribution} \\
        
            \midrule
            
            \textit{Venice}              & Panorama   & $(20000, 64000)$ & $1.19$ & $3.58$ & \href{https://creativecommons.org/licenses/by-nc/4.0/deed.it}{CC BY-NC 4.0} & Jeffrey Martin \cite{martin}                      \\
            \textit{Statuario}           & Industrial & $(34015, 33750)$ & $1.07$ & $3.21$ & \href{https://creativecommons.org/publicdomain/zero/1.0/deed.it}{CC0 1.0}      & Anonymous                                         \\
            \textit{Calacatta Viola}     & Industrial & $(50040, 22677)$ & $1.06$ & $3.17$ & \href{https://creativecommons.org/publicdomain/zero/1.0/deed.it}{CC0 1.0}      & Anonymous                                         \\
            \textit{Breast Cancer}       & Scientific & $(29474, 36430)$ & $1.00$ & $3.00$ & \href{https://creativecommons.org/licenses/by-nc/4.0/deed.it}{CC BY-NC 4.0} & BRACS \cite{brancati2022bracs}                    \\
            \textit{Kidney}       & Scientific & $(55967, 20800)$ & $1.08$ & $3.25$ & \href{https://creativecommons.org/licenses/by/4.0/deed.it}{CC BY 4.0} & KPIs \cite{deng2025kpis2024challengeadvancing, tang2026holohisto}    \\
            \textit{Munich}              & Satellite  & $(37454, 32765)$ & $1.14$ & $3.43$ & \href{https://creativecommons.org/licenses/by/4.0/deed.it}{CC BY 4.0}    & BVV \cite{bvv}                                    \\
            \textit{Andromeda}           & Satellite  & $(22230, 55000)$ & $1.14$ & $3.42$ & \href{https://creativecommons.org/licenses/by/4.0/deed.it}{CC BY 4.0}    & ESA Hubble Telescope \cite{esahubble2026}         \\
            \textit{Italy}               & Satellite  & $(30654, 35040)$ & $1.00$ & $3.00$ & Copernicus open data      & Copernicus Sentinel \cite{copernicus_browser2026} \\
            \textit{Lynxsdesign}       & Synthetic  & $(24588, 43712)$ & $1.00$ & $3.00$ & \href{https://creativecommons.org/licenses/by/4.0/deed.it}{CC BY 4.0}    & Lynxsdesign \cite{lynxsdesign2026}                \\
            \textit{House of Chores}     & Synthetic  & $(24588, 43712)$ & $1.00$ & $3.00$ & \href{https://creativecommons.org/licenses/by/4.0/deed.it}{CC BY 4.0}    & Blender Studio \cite{houseofchores2024}           \\
        
            \bottomrule
            
        \end{tabular}
    }
    \caption{
        \textbf{\dataname{} Overview.} 
        List of the gigapixel images, together with their typology, resolution, raw size in GigaBytes, and original license and attribution. 
        More in the Supplementary.
    }
    \label{tab:gigabench}
\end{table}

    \subsection{Methods Selection}
        We evaluate compression and super-resolution performance across several LRs drawn from Implicit, Hybrid and  Explicit Representations. As for super-resolution, we also compare LRs to classical interpolation methods, such as nearest-neighbour, bicubic, and Lanczos.
        We select, within each family, at least one seminal method alongside one representing the current state of the art.        
        Moreover, the methods are used in different configurations, spanning three orders of magnitude of compression ratio ($10^1$, $10^2$, $10^3$).
        Within the family of Implicit Representations, we choose SIREN~\cite{sitzmann2019siren} as the seminal model, FINER~\cite{liu2024finer} as a major advancement, especially in fitting sharper details, and FDHO~\cite{costanzino2026fdho} as a very recent improvement, yet it was tested at a very small scale for Image Compression. 
        All three models come in two architectures, as defined by the number of hidden layers and hidden features, and denoted by the pair (L, F): a $(5,256)$ architecture, as proposed in~\cite{sitzmann2019siren}, along with a $(8,1024)$ variant, matching respectively the $\sim100$ and $\sim1000$ scales (skipping $\sim10$ as it would have been computationally unfeasible).
        As for Hybrid Representations, we consider the seminal Instant-NGP~ \cite{mueller2022instantngp}, and the two main follow-ups NeuRBF~\cite{chen2023neurbf} and MetricGrids~\cite{wang2025metricgrids}, each at two hash-table sizes, $2^{24}$ and $2^{22}$, denoted as ($T=24$) and ($T=22$), respectively. 
        The first is the value recommended by~\cite{mueller2022instantngp, wang2025metricgrids} for gigapixel fitting, while the second is recommended to fit images in general. 
        These configurations address the $\sim10$ scale, landing at $\sim5$ and $\sim 15$ respectively.
        Concerning Explicit Representations, we choose the seminal GaussianImage~ \cite{zhang2024gaussianimage} method and the more recent SmartSplat~\cite{li2025smartsplat}, as it has been trained and evaluated at the highest resolution (DIV16K dataset), while also being robust enough to converge at very high compression ratios.
        Given the possibility to choose the number of Gaussians, we could select configurations that match $15, 100$, and $3000$ compression ratios.
        However, as the forward step of Gaussian splatting scales linearly with the number of pixels, it is unfeasible to compute it for gigapixel images on a consumer GPU. 
        For this reason, we split the image into tiles and leverage gradient accumulation to reduce the memory footprint while maintaining the same theoretical accuracy. 
        Additional details in \cref{sec:impl} and \cref{alg:tiled-gs-training} of the Supplementary.
        
    \subsection{Metrics}
    
        \paragraph{Compression.}
            We evaluate reconstruction quality via Peak Signal-to-Noise Ratio (PSNR) and Structural Similarity Index (SSIM, reported in the Supplementary)~\cite{wang2004ssim}, which together capture both pixel-level fidelity and perceptual structure preservation.
            Compression efficiency is measured by the Compression Ratio (CR), defined as the ratio between the size in bytes of the uncompressed image and that of the stored Learned Representation:%
            \begin{equation}
                \text{CR} = \frac{H \cdot W \cdot 3 \cdot \text{sizeof}(\text{uint8})}{\#\text{Params} \cdot \text{sizeof}(\text{float32})}
            \end{equation}
            To characterize computational cost, we report Compression Time (CT), the time required to fit the representation to the given image, and Inference Time (IT), the time required to decode the full image from the fitted representation. 
            CT is particularly critical in the gigapixel regime, where it reaches several hours, and IT affects the practical adoption of the representation whenever the image must be fully reconstructed for downstream use. 
            While PSNR and CR are intrinsic to the representation and thus directly comparable across methods regardless of codebase, CT and IT are runtime measurements of each method's reference implementation.
        
        \paragraph{Super-Resolution.}
            In the LRs setting, Super-Resolution consists of querying the fitted model on a coordinate grid finer than the one used during training,  with the expectation that the reconstructed image preserves edges and fine-grained patterns more faithfully than classical interpolation methods.
            To address this standard setting, we downsample the \dataname{} images by $\times4$ and $\times8$, then fit LRs on the downsampled versions and assess the reconstruction quality of images sampled on a finer grid relative to the corresponding original. 
            Yet, it has been observed that the popular distortion metrics used to evaluate reconstruction methods, like  PSNR and SSIM, may correlate poorly with human perceptual judgement \cite{blau2018perception}. This is particularly the case in super-resolution, where perceptually relevant cues relate to sharpness, texture fidelity, and artifact structure rather than pixel-level distortion.  
            Therefore, to complement our experimental assessment of LRs for super-resolution, we designed and carried out a human evaluation study.
            Participants were recruited through an open call, with the study link publicly released and further spread by word of mouth. Each was shown the outputs of four methods: the best-performing model from each of the Implicit, Hybrid, and Explicit LR families, plus the best classical algorithm, and asked to rank them by perceived quality, i.e., sharpness of detail, absence of artifacts, and overall visual fidelity. 
            Stimuli were $2048^2$ crops, manually selected to be visually engaging and to span a diverse spectrum of low- and high-frequency content. 
            All participants saw the full set, with crop order and method arrangement randomized per participant to avoid ordering and position biases, and could freely zoom to inspect fine texture and artifacts.
            From the collected rankings, we derive an Elo score via the Plackett--Luce model \cite{plackett1975analysis, luce1959individual}, following the protocol of established human preference benchmarks for image generation such as Arena AI~\cite{arena2026}, mapping the estimated strengths $\theta_i$ onto the Elo scale as $E_i=k\cdot\theta_i+A$, with anchor $A$ fixed to 1000 and scaling factor $k=400/\ln(10)$.

    \section{Experiments}
\label{sec:experiments}

    \begin{table}[t]
    \centering
    \resizebox{\linewidth}{!}{
       \begin{tabular}{ccc cccccccccc c}
    
            \toprule

            &
            &
            &
    \rothead{\imageone}
    & \rothead{\imagetwo}
    & \rothead{\imagethree}
    & \rothead{\imagefour}
    & \rothead{\imagefive}
    & \rothead{\imagesix}
    & \rothead{\imageseven}
    & \rothead{\imageeight}
    & \rothead{\imagenine}
    & \rothead{\imageten}
    & \textbf{Avg} \\

            \midrule

\multirow{26}{*}{\rotatebox{90}{\textbf{Implicit Representations}}}
		& \multirow{4}{*}{\rotatebox{0}{\shortstack{SIREN \\ (5,256)}}} 
			& PSNR & 21.55 & 31.26 & 22.39 & 16.91 & 16.98 & 14.12 & 13.16 & 22.55 & 20.66 & 23.09 & 20.27 \\
			& & CR & \underline{2904.70} & \underline{2605.17} & \underline{2575.10} & \underline{2436.63} & \underline{2641.72} & \underline{2784.84} & \underline{2774.55} & \underline{2437.49} & \underline{2439.02} & \underline{2439.02} & \underline{2603.82} \\
			& & CT & 17.27 & 20.59 & 19.21 & 18.09 & 15.72 & 16.75 & 20.83 & 18.07 & 18.06 & 18.17 & 18.28 \\
			& & IT & 44.06 & 40.85 & 38.81 & 37.46 & 40.20 & 42.81 & 42.85 & 36.54 & 36.40 & 36.58 & 39.66 \\

			\cmidrule(lr){3-14}

		& \multirow{4}{*}{\rotatebox{0}{\shortstack{FINER \\ (5,256)}}} 
			& PSNR & 14.17 & 32.01 & 22.86 & 17.28 & 14.88 & 13.00 & 13.19 & 22.20 & 20.73 & 21.48 & 19.18 \\
			& & CR & \underline{2904.70} & \underline{2605.17} & \underline{2575.10} & \underline{2436.63} & \underline{2641.72} & \underline{2784.84} & \underline{2774.55} & \underline{2437.49} & \underline{2439.02} & \underline{2439.02} & \underline{2603.82} \\
			& & CT & 23.37 & 26.31 & 24.42 & 24.14 & 21.43 & 22.50 & 27.82 & 23.08 & 24.02 & 23.08 & 24.02 \\
			& & IT & 80.69 & 75.50 & 67.61 & 67.22 & 75.36 & 75.39 & 77.19 & 65.14 & 67.56 & 65.03 & 71.67 \\

			\cmidrule(lr){3-14}

		& \multirow{4}{*}{\rotatebox{0}{\shortstack{FDHO \\ (5,256)}}} 
			& PSNR & 15.61 & 27.16 & 26.98 & 15.88 & 14.17 & 12.60 & 13.11 & 21.91 & 19.36 & 21.30 & 18.81 \\
			& & CR & 2904.49 & 2604.98 & 2574.91 & 2436.45 & 2641.53 & 2784.63 & 2774.35 & 2437.31 & 2438.84 & 2438.84 & 2603.63 \\
			& & CT & 24.41 & 27.23 & 25.12 & 23.68 & 22.78 & 23.90 & 27.26 & 23.66 & 23.60 & 23.70 & 24.53 \\
			& & IT & 31.15 & 31.21 & 30.49 & 29.11 & 30.12 & 32.77 & 33.15 & 28.15 & 28.11 & 28.15 & 30.24 \\

			\cmidrule(lr){3-14}

		& \multirow{4}{*}{\rotatebox{0}{\shortstack{SIREN \\ (8,1024)}}} 
			& PSNR & 23.83 & \cellcolor{implicit!80}32.29 & 27.39 & 18.01 & 23.84 & \cellcolor{implicit!80}17.17 & 13.21 & 22.82 & \cellcolor{implicit!80}21.18 & \cellcolor{implicit!80}27.35 & \cellcolor{implicit!80}22.71 \\
			& & CR & 114.25 & 102.46 & 101.28 & 95.84 & 103.90 & 109.53 & 109.13 & 95.87 & 95.93 & 95.93 & 102.41 \\
			& & CT & 192.38 & 175.10 & 172.95 & 165.09 & 174.15 & 192.88 & 188.92 & 163.28 & 165.48 & 165.51 & 175.57 \\
			& & IT & 482.18 & 431.01 & 425.52 & 402.91 & 439.50 & 465.96 & 455.06 & 401.50 & 404.77 & 405.06 & 431.35 \\

			\cmidrule(lr){3-14}

		& \multirow{4}{*}{\rotatebox{0}{\shortstack{FINER \\ (8,1024)}}} 
			& PSNR & \cellcolor{implicit!80}24.00 & 31.66 & \cellcolor{implicit!80}28.45 & \cellcolor{implicit!80}18.74 & 24.22 & 16.30 & \cellcolor{implicit!80}13.28 & \cellcolor{implicit!80}23.34 & 20.87 & 25.78 & 22.67 \\
			& & CR & 114.25 & 102.46 & 101.28 & 95.84 & 103.90 & 109.53 & 109.13 & 95.87 & 95.93 & 95.93 & 102.41 \\
			& & CT & 229.20 & 202.47 & 204.00 & 189.15 & 204.67 & 216.85 & 215.60 & 189.31 & 192.59 & 192.24 & 203.61 \\
			& & IT & 698.06 & 613.59 & 615.98 & 574.99 & 632.03 & 658.29 & 663.93 & 575.01 & 592.25 & 578.52 & 620.27 \\

			\cmidrule(lr){3-14}

		& \multirow{4}{*}{\rotatebox{0}{\shortstack{FDHO \\ (8,1024)}}} 
			& PSNR & 23.87 & 30.45 & 25.76 & 17.20 & \cellcolor{implicit!80}24.26 & 14.80 & 13.21 & 22.83 & 20.67 & 24.84 & 21.79 \\
			& & CR & 114.25 & 102.46 & 101.28 & 95.84 & 103.90 & 109.53 & 109.13 & 95.87 & 95.93 & 95.93 & 102.41 \\
			& & CT & 229.19 & 205.17 & 202.71 & 193.43 & 208.29 & 225.06 & 217.92 & 191.57 & 194.48 & 194.37 & 206.22 \\
			& & IT & 440.56 & 388.29 & 384.59 & 363.54 & 404.65 & 413.65 & 421.07 & 362.90 & 365.26 & 364.76 & 390.93 \\

			\cmidrule(lr){2-14}

		\multirow{26}{*}{\rotatebox{90}{\textbf{Hybrid Representations}}}
		& \multirow{4}{*}{\rotatebox{0}{\shortstack{I-NGP \\ (T=22)}}} 
			& PSNR & 29.15 & 45.24 & 40.59 & 29.93 & 34.82 & 24.66 & 16.77 & 27.26 & 29.15 & 33.76 & 31.13 \\
			& & CR & 16.64 & 16.84 & 15.22 & 15.50 & 15.41 & 17.60 & 16.22 & 15.65 & 14.79 & 14.79 & 15.86 \\
			& & CT & \textbf{1.08} & \textbf{0.85} & \textbf{0.96} & \textbf{0.97} & \textbf{1.07} & \textbf{1.09} & \textbf{1.11} & \textbf{0.95} & \textbf{1.06} & \textbf{0.97} & \textbf{1.01} \\
			& & IT & 12.73 & 11.26 & 11.49 & 10.62 & 11.24 & 12.38 & 11.88 & 10.77 & 10.83 & 10.80 & 11.40 \\

			\cmidrule(lr){3-14}

		& \multirow{4}{*}{\rotatebox{0}{\shortstack{I-NGP \\ (T=24)}}} 
			& PSNR & \cellcolor{hybrid!80}\underline{31.58} & 46.27 & \underline{42.11} & \underline{32.04} & 39.41 & \cellcolor{hybrid!80}\underline{27.60} & 19.46 & \cellcolor{hybrid!80}\underline{29.08} & \cellcolor{hybrid!80}\textbf{30.37} & \underline{34.02} & \underline{33.19} \\
			& & CR & 5.11 & 5.47 & 4.89 & 5.04 & 4.84 & 5.72 & 5.11 & 5.08 & 4.82 & 4.82 & 5.09 \\
			& & CT & \underline{1.90} & \underline{1.26} & \underline{1.52} & \underline{1.58} & \underline{1.85} & \underline{1.84} & \underline{2.02} & \underline{1.52} & \underline{1.78} & \underline{1.56} & \underline{1.68} \\
			& & IT & 12.79 & 11.36 & 11.92 & 10.73 & 11.41 & 12.32 & 11.92 & 10.81 & 10.91 & 10.82 & 11.50 \\

			\cmidrule(lr){3-14}

		& \multirow{4}{*}{\rotatebox{0}{\shortstack{NeuRBF \\ (T=22)}}} 
			& PSNR & 27.10 & 44.21 & 38.12 & 27.58 & 31.97 & 22.17 & 15.56 & 25.57 & 27.22 & 31.45 & 29.10 \\
			& & CR & 16.64 & 16.84 & 15.22 & 15.50 & 15.41 & 17.60 & 16.22 & 15.65 & 14.79 & 14.79 & 15.86 \\
			& & CT & 11.61 & 11.00 & 11.26 & 10.87 & 11.30 & 11.84 & 12.06 & 10.67 & 10.69 & 10.73 & 11.20 \\
			& & IT & 45.46 & 46.56 & 42.13 & 43.51 & 45.51 & 47.21 & 49.37 & 41.24 & 45.84 & 42.73 & 44.96 \\

			\cmidrule(lr){3-14}

		& \multirow{4}{*}{\rotatebox{0}{\shortstack{NeuRBF \\ (T=24)}}} 
			& PSNR & 28.44 & 46.22 & 40.50 & 29.79 & 35.57 & 24.66 & 17.38 & 27.11 & 28.68 & 32.41 & 31.08 \\
			& & CR & 5.11 & 5.47 & 4.89 & 5.04 & 4.84 & 5.72 & 5.11 & 5.08 & 4.82 & 4.82 & 5.09 \\
			& & CT & 15.51 & 14.41 & 14.96 & 14.36 & 15.36 & 14.74 & 15.43 & 14.09 & 14.34 & 14.37 & 14.76 \\
			& & IT & 47.48 & 46.53 & 45.71 & 42.90 & 43.57 & 47.75 & 46.71 & 39.86 & 43.62 & 45.62 & 44.97 \\

			\cmidrule(lr){3-14}

		& \multirow{4}{*}{\rotatebox{0}{\shortstack{MetricGrids \\ (T=22)}}} 
			& PSNR & 29.23 & \underline{47.06} & 41.19 & 30.04 & 36.26 & 24.60 & 17.06 & 27.20 & 29.29 & 33.88 & 31.58 \\
			& & CR & 16.63 & 16.83 & 15.21 & 15.49 & 15.40 & 17.59 & 16.21 & 15.64 & 14.78 & 14.78 & 15.86 \\
			& & CT & 13.46 & 12.22 & 12.45 & 12.39 & 13.08 & 12.70 & 12.87 & 12.60 & 12.23 & 12.14 & 12.61 \\
			& & IT & 46.66 & 40.68 & 38.35 & 36.54 & 42.55 & 42.44 & 43.27 & 39.38 & 38.55 & 33.29 & 40.17 \\

			\cmidrule(lr){3-14}

		& \multirow{4}{*}{\rotatebox{0}{\shortstack{MetricGrids \\ (T=24)}}} 
			& PSNR & 31.29 & \cellcolor{hybrid!80}\textbf{48.58} & \cellcolor{hybrid!80}\textbf{43.65} & \cellcolor{hybrid!80}\textbf{32.12} & \cellcolor{hybrid!80}\textbf{40.57} & 27.25 & \cellcolor{hybrid!80}\underline{19.54} & 28.46 & \underline{30.34} & \cellcolor{hybrid!80}\textbf{34.13} & \cellcolor{hybrid!80}\textbf{33.59} \\
			& & CR & 5.10 & 5.47 & 4.89 & 5.04 & 4.84 & 5.72 & 5.11 & 5.08 & 4.82 & 4.82 & 5.09 \\
			& & CT & 20.80 & 18.08 & 18.67 & 18.10 & 20.07 & 19.31 & 19.74 & 18.43 & 18.40 & 18.15 & 18.98 \\
			& & IT & 47.49 & 42.03 & 36.97 & 37.96 & 42.66 & 43.60 & 45.17 & 38.90 & 37.37 & 34.07 & 40.62 \\

			\cmidrule(lr){2-14}

		\multirow{26}{*}{\rotatebox{90}{\textbf{Explicit Representations}}}
		& \multirow{4}{*}{\rotatebox{0}{\shortstack{GaussianImage \\ (CR=3000)}}} 
			& PSNR & 6.52 & 1.24 & 2.19 & 1.82 & 4.98 & 6.87 & 7.04 & 13.92 & 4.55 & 6.69 & 5.58 \\
			& & CR & \textbf{3000.00} & \textbf{3000.00} & \textbf{3000.00} & \textbf{3000.00} & \textbf{3000.00} & \textbf{3000.00} & \textbf{3000.00} & \textbf{3000.00} & \textbf{3000.00} & \textbf{3000.00} & \textbf{3000.00} \\
			& & CT & 36.01 & 33.95 & 31.84 & 30.12 & 32.65 & 34.22 & 34.22 & 30.03 & 30.68 & 31.90 & 32.56 \\
			& & IT & 4.57 & \textbf{0.11} & \textbf{0.10} & \textbf{0.09} & \textbf{0.10} & \textbf{0.11} & \textbf{0.11} & \textbf{0.09} & \textbf{0.10} & \textbf{0.10} & \textbf{0.55} \\

			\cmidrule(lr){3-14}

		& \multirow{4}{*}{\rotatebox{0}{\shortstack{GaussianImage \\ (CR=100)}}} 
			& PSNR & 16.58 & 20.68 & 17.22 & 18.41 & 15.28 & 16.62 & 13.70 & 19.75 & 15.76 & 17.84 & 17.18 \\
			& & CR & 100.00 & 100.00 & 100.00 & 100.00 & 100.00 & 100.00 & 100.00 & 100.00 & 100.00 & 100.00 & 100.00 \\
			& & CT & 41.70 & 38.62 & 36.55 & 34.58 & 37.38 & 38.72 & 38.47 & 32.90 & 35.02 & 35.11 & 36.90 \\
			& & IT & \textbf{3.53} & \underline{0.12} & \underline{0.11} & \underline{0.10} & \underline{0.11} & 3.94 & 4.43 & \underline{0.10} & \underline{0.11} & \underline{0.11} & 1.27 \\

			\cmidrule(lr){3-14}

		& \multirow{4}{*}{\rotatebox{0}{\shortstack{GaussianImage \\ (CR=15)}}} 
			& PSNR & 17.20 & 17.03 & 15.28 & 16.27 & 16.67 & 19.50 & 15.77 & 22.55 & 17.33 & 19.76 & 17.74 \\
			& & CR & 15.00 & 15.00 & 15.00 & 15.00 & 15.00 & 15.00 & 15.00 & 15.00 & 15.00 & 15.00 & 15.00 \\
			& & CT & 74.88 & 71.24 & 67.07 & 62.46 & 67.31 & 68.16 & 68.67 & 55.10 & 66.79 & 63.76 & 66.54 \\
			& & IT & 7.05 & 7.49 & 5.13 & 4.92 & 5.87 & 6.59 & 6.77 & 4.87 & 4.91 & 4.88 & 5.85 \\

			\cmidrule(lr){3-14}

		& \multirow{4}{*}{\rotatebox{0}{\shortstack{SmartSplat \\ (CR=3000)}}} 
			& PSNR & 24.27 & \cellcolor{explicit!80}38.39 & 31.40 & 22.37 & 21.59 & 17.60 & 13.16 & 22.44 & 24.29 & 24.98 & 24.05 \\
			& & CR & \textbf{3000.00} & \textbf{3000.00} & \textbf{3000.00} & \textbf{3000.00} & \textbf{3000.00} & \textbf{3000.00} & \textbf{3000.00} & \textbf{3000.00} & \textbf{3000.00} & \textbf{3000.00} & \textbf{3000.00} \\
			& & CT & 77.03 & 69.23 & 72.52 & 68.72 & 69.38 & 71.71 & 70.71 & 63.20 & 64.74 & 65.34 & 69.26 \\
			& & IT & 4.90 & 0.13 & 0.13 & 0.13 & 0.15 & \underline{0.16} & \underline{0.14} & 0.13 & 0.13 & 0.15 & \underline{0.62} \\

			\cmidrule(lr){3-14}

		& \multirow{4}{*}{\rotatebox{0}{\shortstack{SmartSplat \\ (CR=100)}}} 
			& PSNR & 28.18 & 38.07 & \cellcolor{explicit!80}34.97 & 26.39 & 35.16 & 23.36 & 15.05 & 25.47 & 27.10 & \cellcolor{explicit!80}33.65 & 28.74 \\
			& & CR & 100.00 & 100.00 & 100.00 & 100.00 & 100.00 & 100.00 & 100.00 & 100.00 & 100.00 & 100.00 & 100.00 \\
			& & CT & 85.50 & 79.41 & 78.63 & 73.29 & 81.54 & 84.59 & 75.32 & 68.32 & 71.29 & 71.78 & 76.97 \\
			& & IT & \underline{4.51} & 0.17 & 0.16 & 0.17 & 0.19 & 3.59 & 3.83 & 0.15 & 0.18 & 0.16 & 1.31 \\

			\cmidrule(lr){3-14}

		& \multirow{4}{*}{\rotatebox{0}{\shortstack{SmartSplat \\ (CR=15)}}} 
			& PSNR & \cellcolor{explicit!80}\textbf{31.82} & 33.32 & 27.51 & \cellcolor{explicit!80}26.54 & \cellcolor{explicit!80}\underline{40.43} & \cellcolor{explicit!80}\textbf{30.04} & \cellcolor{explicit!80}\textbf{21.73} & \cellcolor{explicit!80}\textbf{29.39} & \cellcolor{explicit!80}29.92 & 33.32 & \cellcolor{explicit!80}30.40 \\
			& & CR & 15.00 & 15.00 & 15.00 & 15.00 & 15.00 & 15.00 & 15.00 & 15.00 & 15.00 & 15.00 & 15.00 \\
			& & CT & 110.57 & 98.14 & 102.66 & 94.11 & 100.59 & 107.43 & 95.90 & 90.42 & 91.34 & 93.65 & 98.48 \\
			& & IT & 5.94 & 5.12 & 5.17 & 0.26 & 5.31 & 5.69 & 5.36 & 0.25 & 0.27 & 0.26 & 3.36 \\

        \bottomrule
                
        \end{tabular}
    }
    \caption{
        \textbf{Compression Performance on \dataname{}} according to the
        PSNR (dB), CT (min), IT (s) and CR metrics.
        Best results in \textbf{bold}, runner-ups \underline{underlined}.
        Best PSNRs within each family are highlighted by different colours. 
    }
    \label{tab:compression}
\end{table}

\begin{figure*}[t]
    \centering
    \begin{minipage}[b]{0.33\linewidth}
        \centering
        \includegraphics[width=\linewidth]{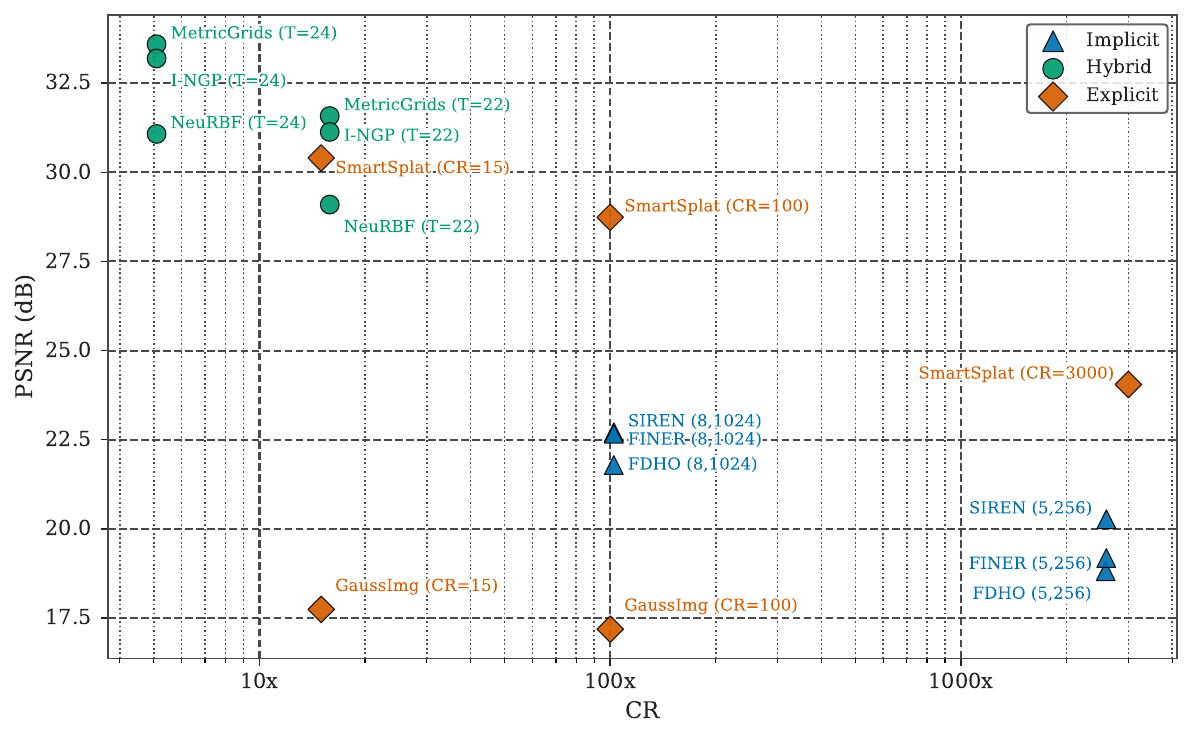}
        \captionof{figure}{\textbf{PSNR vs. Compression Ratio.}}
        \label{fig:timings_cr}
    \end{minipage}
    \hfill
    \begin{minipage}[b]{0.66\linewidth}
        \centering
        \includegraphics[width=0.49\linewidth]{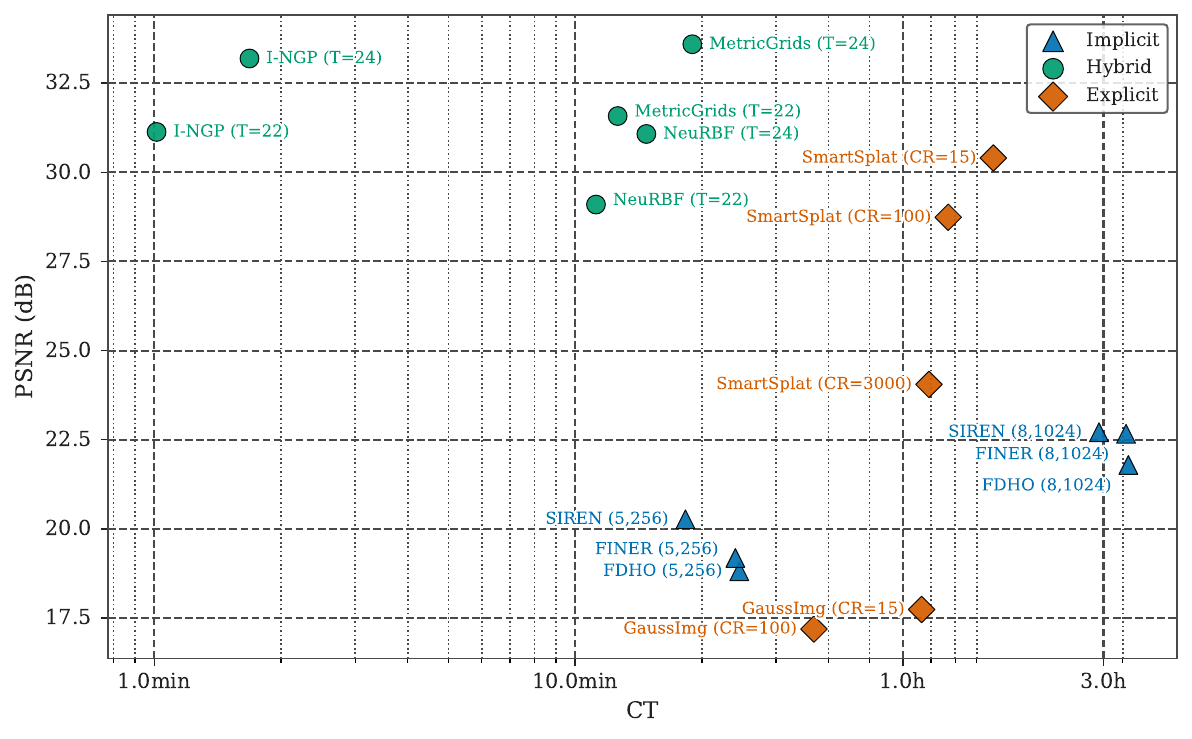}
        \includegraphics[width=0.49\linewidth]{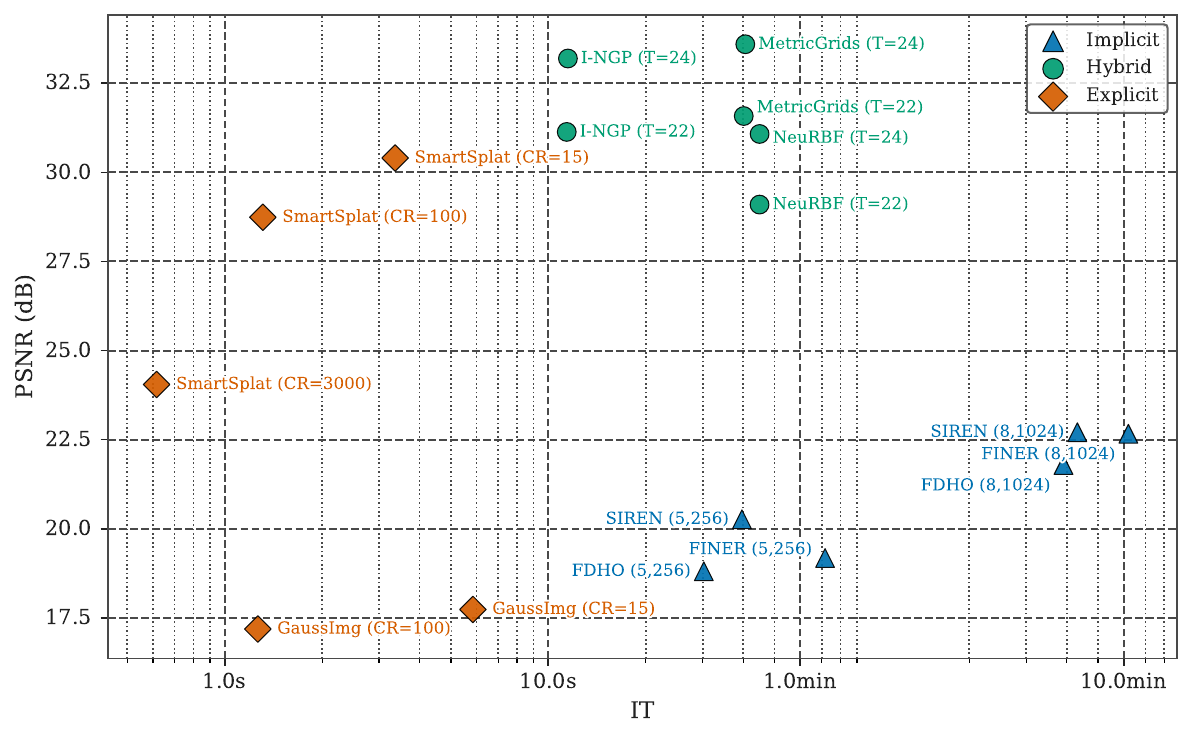}
        \captionof{figure}{\textbf{PSNR vs. Compression Time and Inference Time.}}
        \label{fig:timings_ct_it}
    \end{minipage}
\end{figure*}

    \subsection{Compression}
        \cref{tab:compression} reports PSNR (dB), Compression Time (CT, minutes), Inference Time (IT, seconds) and Compression Ratio (CR) for all images in  \dataname{} and for all considered methods.        
        Hybrid representations achieve the highest reconstruction fidelity across the board, with MetricGrids ($\text{T}=24$) and I-NGP ($\text{T}=24$) reaching $33.59$ dB and $33.19$ dB, respectively, but are structurally confined to lower compression ratios ($\text{CR} \approx 5$--$16\times$), as their hash-table capacity sets a hard floor on model size.
        Within this family and configuration, I-NGP stands out for its fitting efficiency, completing in under $2$ minutes on average, while MetricGrids requires up to almost $19$ minutes for a marginal quality gain, suggesting I-NGP offers a more practical operating point for most use cases.
        Such an advantage is not only due to the lighter decoding of I-NGP, but especially to the original implementation, as NVIDIA's I-NGP code relies on custom CUDA kernels while MetricGrids is fully developed using PyTorch.
        Both configurations of NeuRBF lag behind by approximately $2$ dB, indicating that adaptive radial basis functions transfer less effectively to the gigapixel setting.
        Implicit representations occupy the opposite end of the compression spectrum.
        The smaller variants operate at $\text{CR}~\approx~ 2600\times$, achieving $18$--$20$ dB PSNR at compression times close to the slowest hybrid methods ($\sim$ $18$--$25$ minutes).
        The bigger variants trade compression ratio for quality ($\text{CR}~\approx~102\times$, $\sim$~$22$--$23~\text{dB}$) at the cost of dramatically longer compression times, up to 206 minutes.
        Quality differences across activation functions (SIREN, FINER, FDHO) are modest within each architecture, suggesting that at the gigapixel scale, the bottleneck lies in the fitting budget rather than the activation design.
        Explicit Representations exhibit the largest intra-family variance.
        SmartSplat maintains competitive quality across all tested compression ratios, reaching $30.40$ dB at $\text{CR}=15$ and degrading gracefully to $24.05$ dB at $\text{CR}=3000$.
        GaussianImage, by contrast, collapses at high compression: while it reaches $17.73$ dB at $\text{CR}=15$, its PSNR drops catastrophically to $5.58$ dB at $\text{CR}=3000$, well below acceptable perceptual quality.
        This gap underscores that the choice of Gaussian parametrisation is critical at extreme compression ratios, and that results obtained with one Gaussian-based method should not be generalised to the family as a whole.
        Yet, both Explicit methods offer the fastest inference times among the considered LRs.
        Finally, in \cref{fig:timings_cr} we report the quality (PSNR)  vs. Compression Ratio (CR) plot.  We note that implicit methods can achieve extreme compression consistently exceeding $1000\times$, albeit with the lowest overall PSNR. Conversely, Hybrid methods deliver the highest reconstruction quality but are largely constrained to the low-compression regime ($10\times$--$30\times$). Explicit representations bridge this gap, spanning the widest range of CRs. Notably, SmartSplat at $\text{CR}=3000$ maintains a clear PSNR advantage over Implicit methods at similar extreme CRs.

        \paragraph{Computational Efficiency}
            \cref{fig:timings_ct_it} illustrates the Quality vs. Efficiency landscape, plotting the PSNR along with the Compression Time (CT) and Inference Time (IT).
            The CT plot reveals a separation between families. 
            I-NGP is the outlier in the Hybrid group, fitting an image in under $1.7$ minutes on average at any tested ratio, an order of magnitude faster than MetricGrids and several times faster than NeuRBF, at equivalent CR, and faster than any other LRs overall. 
            Explicit representations are more expensive to train, with SmartSplat consistently exceeding GaussianImage and both exhibiting a counter-intuitive inverse scaling. 
            Indeed, fitting an image at $\text{CR}=15$ costs substantially \emph{more} than at $\text{CR}=3000$, as allocating more primitives demands longer optimisation. 
            The larger (8,1024) Implicit architectures are the slowest overall, approaching $175$--$206$ minutes per image, with a modest quality advantage over the smaller (5,256) variants.
            Such separation is also observable in the IT plot.
            Explicit representations, despite their high CT, are the quickest to decode, with SmartSplat decoding in 0.62, 1.31, and 3.36 seconds for CR 3000, 100, and 15, respectively, making them the only family compatible with latency-sensitive downstream pipelines.
            The larger implicit variants also pay the steepest inference penalty, approaching $7$--$11$ minutes to decode a single gigapixel image, which largely precludes their use in interactive applications.
            Hybrid methods occupy a stable middle ground, decoding in $10$--$40$ seconds regardless of their hash-table size, offering a predictable and practical operating point. 
            These considerations highlight that practitioners face a multidimensional trade-off when choosing an LR to handle gigapixel images.
            Training and inference costs are largely inversely correlated across families, as the methods that yield the fastest decoding times are the slowest at compression time, and vice versa, while compression ratio is heavily constrained by the chosen representation's capacity.
    
    \begin{table}[t]
    \centering
    \resizebox{\linewidth}{!}{
       \begin{tabular}{ccc cccccccccc c}
    
            \toprule
            &
            &
            &
            \rothead{\imageone}
            & \rothead{\imagetwo}
            & \rothead{\imagethree}
            & \rothead{\imagefour}
            & \rothead{\imagefive}
            & \rothead{\imagesix}
            & \rothead{\imageseven}
            & \rothead{\imageeight}
            & \rothead{\imagenine}
            & \rothead{\imageten}
            & \textbf{Avg} \\

            \midrule
            
\multirow{30}{*}{\rotatebox{90}{Upsample $\mathbf{\times4}$}}
		& \multirow{8}{*}{\rotatebox{90}{\textbf{Implicit}}}
			& {\rotatebox{0}{\shortstack{SIREN (5,256)}}} & 21.72 & 32.12 & 26.95 & 17.35 & 16.88 & 14.13 & 13.19 & 22.71 & 20.84 & 24.05 & 20.99 \\
			\cmidrule(lr){3-14}

			& & {\rotatebox{0}{\shortstack{FINER (5,256)}}} & 20.55 & 27.68 & 24.10 & 16.39 & 16.00 & 13.63 & 13.13 & 22.06 & 19.61 & 21.25 & 19.44 \\
			\cmidrule(lr){3-14}

			& & {\rotatebox{0}{\shortstack{FDHO (5,256)}}} & 20.24 & 27.30 & 22.79 & 15.85 & 14.89 & 12.98 & 13.10 & 21.56 & 19.28 & 21.18 & 18.92 \\
			\cmidrule(lr){3-14}

			& & {\rotatebox{0}{\shortstack{SIREN (8,1024)}}} & 23.00 & 32.27 & 27.58 & 18.43 & 20.83 & 16.41 & 13.22 & 22.87 & 21.09 & 27.37 & 22.31 \\
			\cmidrule(lr){3-14}

			& & {\rotatebox{0}{\shortstack{FINER (8,1024)}}} & \cellcolor{implicit!80}23.45 & \cellcolor{implicit!80}33.26 & \cellcolor{implicit!80}28.44 & \cellcolor{implicit!80}18.86 & \cellcolor{implicit!80}22.42 & \cellcolor{implicit!80}17.03 & \cellcolor{implicit!80}13.24 & \cellcolor{implicit!80}\textbf{23.23} & \cellcolor{implicit!80}21.20 & \cellcolor{implicit!80}27.74 & \cellcolor{implicit!80}22.89 \\
			\cmidrule(lr){3-14}

			& & {\rotatebox{0}{\shortstack{FDHO (8,1024)}}} & 20.79 & 25.56 & 23.43 & 16.42 & 16.48 & 13.36 & 13.14 & 22.02 & 19.76 & 22.38 & 19.34 \\
			\cmidrule(lr){2-14}

		& \multirow{8}{*}{\rotatebox{90}{\textbf{Hybrid}}}
			& {\rotatebox{0}{\shortstack{I-NGP (T=22)}}} & \cellcolor{hybrid!80}\textbf{25.52} & \underline{41.13} & \underline{35.56} & \underline{25.08} & 31.18 & \underline{21.05} & \cellcolor{hybrid!80}\textbf{14.43} & \cellcolor{hybrid!80}\underline{23.11} & \cellcolor{hybrid!80}\textbf{25.24} & \underline{28.56} & \underline{27.09} \\
			\cmidrule(lr){3-14}

			& & {\rotatebox{0}{\shortstack{I-NGP (T=24)}}} & \underline{25.47} & \cellcolor{hybrid!80}\textbf{41.15} & \cellcolor{hybrid!80}\textbf{35.56} & 25.06 & 31.83 & \cellcolor{hybrid!80}\textbf{21.14} & \underline{14.43} & 23.08 & \underline{25.04} & 28.32 & \cellcolor{hybrid!80}\textbf{27.11} \\
			\cmidrule(lr){3-14}

			& & {\rotatebox{0}{\shortstack{NeuRBF (T=22)}}} & 22.48 & 38.89 & 33.25 & 22.96 & 26.07 & 17.48 & 11.33 & 21.24 & 22.65 & 26.04 & 24.24 \\
			\cmidrule(lr){3-14}

			& & {\rotatebox{0}{\shortstack{NeuRBF (T=24)}}} & 24.01 & 39.33 & 34.49 & 23.78 & 28.15 & 18.18 & 11.74 & 22.02 & 23.15 & 26.28 & 25.11 \\
			\cmidrule(lr){3-14}

			& & {\rotatebox{0}{\shortstack{MetricGrids (T=22)}}} & 23.86 & 40.75 & 35.01 & 24.61 & \underline{32.67} & 20.51 & 12.89 & 22.62 & 24.34 & \cellcolor{hybrid!80}\textbf{29.12} & 26.64 \\
			\cmidrule(lr){3-14}

			& & {\rotatebox{0}{\shortstack{MetricGrids (T=24)}}} & 24.61 & 41.00 & 35.53 & \cellcolor{hybrid!80}\textbf{25.09} & \cellcolor{hybrid!80}\textbf{32.67} & 20.66 & 13.09 & 22.98 & 24.45 & 28.53 & 26.86 \\
			\cmidrule(lr){2-14}

		& \multirow{8}{*}{\rotatebox{90}{\textbf{Explicit}}}
			& {\rotatebox{0}{\shortstack{GaussianImage (CR=3000)}}} & 14.48 & 22.87 & 17.40 & 17.57 & 12.90 & 12.76 & 11.88 & 18.02 & 13.97 & 14.65 & 15.65 \\
			\cmidrule(lr){3-14}

			& & {\rotatebox{0}{\shortstack{GaussianImage (CR=100)}}} & 23.24 & 25.96 & 21.12 & 21.00 & 24.33 & \cellcolor{explicit!80}19.80 & \cellcolor{explicit!80}13.97 & 21.94 & 22.96 & 24.96 & 21.93 \\
			\cmidrule(lr){3-14}

			& & {\rotatebox{0}{\shortstack{GaussianImage (CR=15)}}} & 16.55 & 23.10 & 19.17 & 19.49 & 16.13 & 14.52 & 12.55 & 15.65 & 16.96 & 16.59 & 17.07 \\
			\cmidrule(lr){3-14}

			& & {\rotatebox{0}{\shortstack{SmartSplat (CR=3000)}}} & \cellcolor{explicit!80}23.96 & \cellcolor{explicit!80}38.83 & \cellcolor{explicit!80}31.61 & \cellcolor{explicit!80}22.26 & 21.64 & 17.41 & 13.03 & \cellcolor{explicit!80}22.21 & \cellcolor{explicit!80}23.93 & 25.53 & \cellcolor{explicit!80}24.04 \\
			\cmidrule(lr){3-14}

			& & {\rotatebox{0}{\shortstack{SmartSplat (CR=100)}}} & 23.74 & 30.45 & 27.48 & 21.90 & \cellcolor{explicit!80}28.93 & 19.67 & 12.83 & 22.02 & 23.18 & \cellcolor{explicit!80}26.64 & 23.68 \\
			\cmidrule(lr){3-14}

			& & {\rotatebox{0}{\shortstack{SmartSplat (CR=15)}}} & 19.75 & 23.67 & 16.83 & 19.51 & 18.71 & 17.73 & 12.91 & 21.33 & 18.67 & 18.40 & 18.75 \\
			\cmidrule(lr){2-14}

		& \multirow{4}{*}{\rotatebox{90}{\textbf{Classic}}}
			& {\rotatebox{0}{\shortstack{Nearest Neighbour}}} & 24.17 & 38.36 & 32.40 & 22.07 & 27.27 & 17.55 & 12.14 & 21.33 & 23.79 & 26.72 & 24.58 \\
			\cmidrule(lr){3-14}

			& & {\rotatebox{0}{\shortstack{Bicubic}}} & \cellcolor{classic!80}25.05 & \cellcolor{classic!80}39.89 & \cellcolor{classic!80}33.87 & \cellcolor{classic!80}23.59 & \cellcolor{classic!80}29.32 & \cellcolor{classic!80}19.26 & \cellcolor{classic!80}13.75 & \cellcolor{classic!80}22.71 & \cellcolor{classic!80}24.70 & \cellcolor{classic!80}27.83 & \cellcolor{classic!80}26.00 \\
			\cmidrule(lr){3-14}

			& & {\rotatebox{0}{\shortstack{Lanczos}}} & 24.84 & 39.67 & 33.70 & 23.40 & 29.30 & 19.09 & 13.54 & 22.50 & 24.46 & 27.67 & 25.82 \\
			\midrule

		\multirow{30}{*}{\rotatebox{90}{Upsample $\mathbf{\times8}$}}
		& \multirow{8}{*}{\rotatebox{90}{\textbf{Implicit}}}
			& {\rotatebox{0}{\shortstack{SIREN (5,256)}}} & 21.59 & 32.07 & 26.54 & 17.20 & 16.76 & 14.06 & 13.19 & 22.65 & 20.62 & 23.77 & 20.85 \\
			\cmidrule(lr){3-14}

			& & {\rotatebox{0}{\shortstack{FINER (5,256)}}} & 20.82 & 28.95 & 25.43 & 16.60 & 16.45 & 14.02 & 13.17 & 22.38 & 20.13 & 22.72 & 20.07 \\
			\cmidrule(lr){3-14}

			& & {\rotatebox{0}{\shortstack{FDHO (5,256)}}} & 20.30 & 29.69 & 24.69 & 16.40 & 15.30 & 13.45 & 13.12 & 22.25 & 19.82 & 22.26 & 19.73 \\
			\cmidrule(lr){3-14}

			& & {\rotatebox{0}{\shortstack{SIREN (8,1024)}}} & \cellcolor{implicit!80}22.21 & \cellcolor{implicit!80}32.43 & \cellcolor{implicit!80}27.65 & \cellcolor{implicit!80}17.56 & \cellcolor{implicit!80}19.63 & \cellcolor{implicit!80}15.18 & \cellcolor{implicit!80}\underline{13.21} & \underline{22.77} & \cellcolor{implicit!80}20.82 & \cellcolor{implicit!80}\textbf{26.13} & \cellcolor{implicit!80}21.76 \\
			\cmidrule(lr){3-14}

			& & {\rotatebox{0}{\shortstack{FINER (8,1024)}}} & 21.79 & 29.71 & 26.97 & 17.55 & 19.61 & 15.13 & 13.20 & \cellcolor{implicit!80}\textbf{22.91} & 20.22 & 24.90 & 21.20 \\
			\cmidrule(lr){3-14}

			& & {\rotatebox{0}{\shortstack{FDHO (8,1024)}}} & 21.22 & 29.63 & 25.25 & 16.58 & 16.36 & 14.86 & 13.18 & 22.76 & 19.73 & 24.59 & 20.42 \\
			\cmidrule(lr){2-14}

		& \multirow{8}{*}{\rotatebox{90}{\textbf{Hybrid}}}
			& {\rotatebox{0}{\shortstack{I-NGP (T=22)}}} & \underline{23.40} & 38.81 & 31.09 & 22.66 & 27.00 & \cellcolor{hybrid!80}\textbf{18.27} & \cellcolor{hybrid!80}11.70 & 21.41 & 22.86 & 25.48 & \underline{24.27} \\
			\cmidrule(lr){3-14}

			& & {\rotatebox{0}{\shortstack{I-NGP (T=24)}}} & 23.39 & 38.71 & 31.09 & 22.66 & \cellcolor{hybrid!80}\textbf{27.05} & \underline{18.27} & 11.62 & \cellcolor{hybrid!80}21.42 & 22.84 & 25.43 & 24.25 \\
			\cmidrule(lr){3-14}

			& & {\rotatebox{0}{\shortstack{NeuRBF (T=22)}}} & 22.05 & 38.04 & 31.96 & 21.93 & 22.23 & 15.65 & 10.87 & 20.67 & 21.96 & 24.18 & 22.95 \\
			\cmidrule(lr){3-14}

			& & {\rotatebox{0}{\shortstack{NeuRBF (T=24)}}} & 23.39 & 38.06 & 32.17 & 22.06 & 25.65 & 16.08 & 11.21 & 20.68 & 22.30 & 24.74 & 23.63 \\
			\cmidrule(lr){3-14}

			& & {\rotatebox{0}{\shortstack{MetricGrids (T=22)}}} & 22.72 & \underline{39.09} & \underline{32.80} & \underline{22.75} & 25.50 & 17.62 & 11.42 & 21.37 & 22.86 & \cellcolor{hybrid!80}\underline{25.92} & 24.21 \\
			\cmidrule(lr){3-14}

			& & {\rotatebox{0}{\shortstack{MetricGrids (T=24)}}} & \cellcolor{hybrid!80}\textbf{23.54} & \cellcolor{hybrid!80}\textbf{39.16} & \cellcolor{hybrid!80}\textbf{32.97} & \cellcolor{hybrid!80}\textbf{22.83} & \underline{27.01} & 17.73 & 11.68 & 21.38 & \cellcolor{hybrid!80}\underline{22.91} & 25.84 & \cellcolor{hybrid!80}\textbf{24.51} \\
			\cmidrule(lr){2-14}

		& \multirow{8}{*}{\rotatebox{90}{\textbf{Explicit}}}
			& {\rotatebox{0}{\shortstack{GaussianImage (CR=3000)}}} & 21.46 & 24.63 & 20.38 & 19.29 & 19.41 & 16.67 & \cellcolor{explicit!80}\textbf{13.29} & \cellcolor{explicit!80}21.25 & 19.74 & 21.11 & 19.72 \\
			\cmidrule(lr){3-14}

			& & {\rotatebox{0}{\shortstack{GaussianImage (CR=100)}}} & 21.04 & 28.29 & 23.31 & 20.20 & 21.85 & 16.63 & 12.00 & 19.86 & 21.25 & 21.95 & 20.64 \\
			\cmidrule(lr){3-14}

			& & {\rotatebox{0}{\shortstack{GaussianImage (CR=15)}}} & 6.28 & 22.35 & 15.15 & 15.54 & 7.79 & 5.31 & 3.75 & 2.65 & 8.02 & 5.45 & 9.23 \\
			\cmidrule(lr){3-14}

			& & {\rotatebox{0}{\shortstack{SmartSplat (CR=3000)}}} & \cellcolor{explicit!80}22.81 & \cellcolor{explicit!80}35.32 & \cellcolor{explicit!80}30.86 & \cellcolor{explicit!80}20.99 & 21.28 & \cellcolor{explicit!80}16.68 & 12.33 & 21.18 & \cellcolor{explicit!80}\textbf{22.99} & \cellcolor{explicit!80}23.84 & \cellcolor{explicit!80}22.83 \\
			\cmidrule(lr){3-14}

			& & {\rotatebox{0}{\shortstack{SmartSplat (CR=100)}}} & 20.45 & 23.92 & 20.39 & 19.49 & \cellcolor{explicit!80}22.59 & 16.37 & 11.43 & 19.90 & 20.55 & 21.78 & 19.69 \\
			\cmidrule(lr){3-14}

			& & {\rotatebox{0}{\shortstack{SmartSplat (CR=15)}}} & 14.21 & 19.48 & 17.47 & 15.50 & 15.01 & 12.37 & 11.30 & 18.58 & 17.26 & 15.64 & 15.68 \\
			\cmidrule(lr){2-14}

		& \multirow{4}{*}{\rotatebox{90}{\textbf{Classic}}}
			& {\rotatebox{0}{\shortstack{Nearest Neighbour}}} & 22.26 & 36.92 & 30.02 & 20.69 & 22.70 & 15.73 & 11.04 & 20.28 & 21.77 & 23.07 & 22.45 \\
			\cmidrule(lr){3-14}

			& & {\rotatebox{0}{\shortstack{Bicubic}}} & \cellcolor{classic!80}23.05 & \cellcolor{classic!80}38.13 & \cellcolor{classic!80}31.17 & \cellcolor{classic!80}21.82 & \cellcolor{classic!80}24.17 & \cellcolor{classic!80}16.96 & \cellcolor{classic!80}11.99 & \cellcolor{classic!80}21.25 & \cellcolor{classic!80}22.75 & \cellcolor{classic!80}24.42 & \cellcolor{classic!80}23.57 \\
			\cmidrule(lr){3-14}

			& & {\rotatebox{0}{\shortstack{Lanczos}}} & 22.84 & 37.93 & 31.00 & 21.62 & 24.08 & 16.78 & 11.80 & 21.04 & 22.50 & 24.30 & 23.39 \\
			\bottomrule                
        \end{tabular}
}
    \caption{
        \textbf{Super-Resolution Performance on \dataname{}.} 
        Best results in \textbf{bold}, runner-ups \underline{underlined}. Best PSNRs within each family are highlighted by different colours. 
    }
    \label{tab:superres}
\end{table}

    \begin{figure*}
    \centering
    \includegraphics[width=0.96\linewidth]{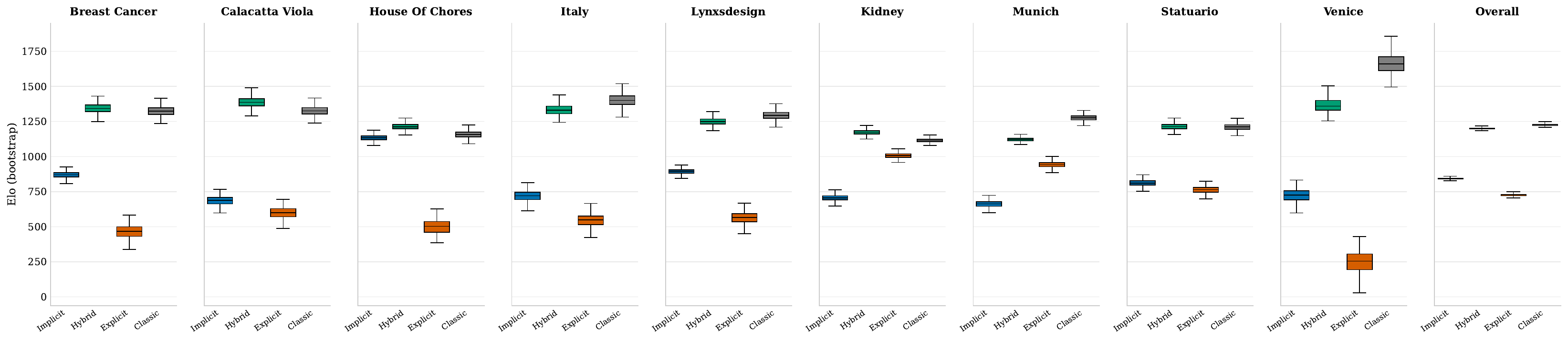}
    \includegraphics[width=0.96\linewidth]{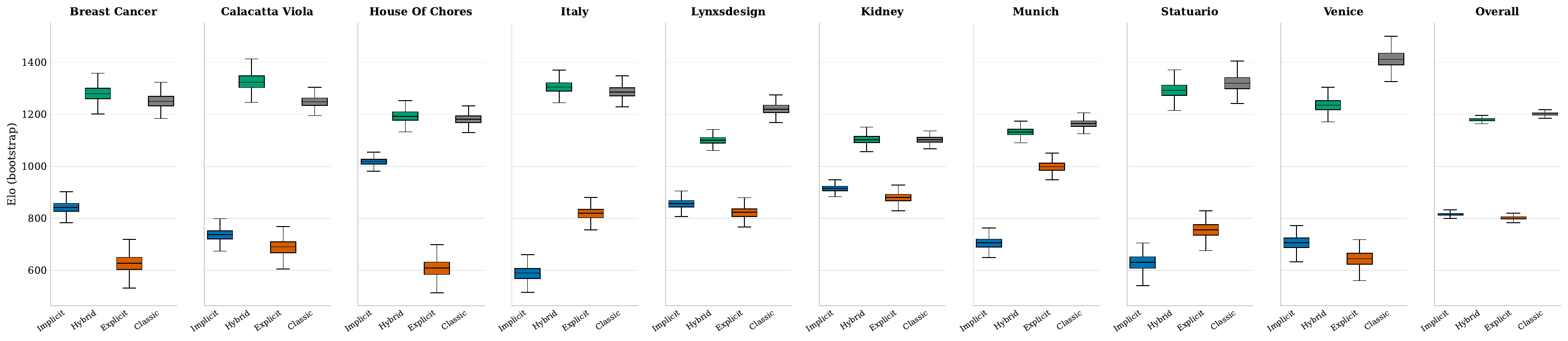}
    \caption{
    \textbf{Elo Rating of the Human Evaluation Study.} Top: $\times4$, Bottom: $\times8$}
    \label{fig:elo_boxplot}
\end{figure*}

    \subsection{Super-Resolution}
        \cref{tab:superres} reports PSNR at $\times4$ and $\times8$ upsampling factors for all images in \dataname{}. 
        The super-resolution performance unveils a different picture from the compression task, highlighting several findings.
        The most salient observation is that classical methods are highly competitive in the gigapixel regime. 
        For example, bicubic interpolation outperforms all Implicit Representations at both upsampling scales. 
        Indeed, the (8, 1024) variants approach bicubic performance at $\times4$ ($22.31$--$22.89$ dB), and the (5, 256) variants fall $3$--$5$ dB below. 
        However, Hybrid Representations are the only family to consistently surpass classical baselines.
        I-NGP ($\text{T}=24$) leads at $\times4$ with $27.11$ dB ($+1.11$ dB over bicubic interpolation), while MetricGrids ($\text{T}=24$) achieves $26.86$ dB. 
        Notably, the advantage over classical methods narrows at $\times8$, where I-NGP ($\text{T}=24$) leads by only $0.68$ dB, suggesting that the representational advantage of grid-based hybrids diminishes as the upsampling demand increases. 
        NeuRBF again falls behind both I-NGP and MetricGrids by approximately $\sim2$ dB at $\times4$, narrowing to less than $1$ dB at $\times8$.
        Explicit Representations exhibit once again the most variable behaviour.
        SmartSplat ($\text{CR}=3000$) achieves $24.04$ dB at $\times4$ and degrades gracefully to $22.83$ dB at $\times8$, remaining competitive with classical methods. 
        GaussianImage, however, displays severe instability. 
        While GaussianImage ($\text{CR}=100$) reaches $21.93$ dB at $\times4$, GaussianImage ($\text{CR}=15$) collapses at $\times8$ ($9.23$ dB), suggesting that the low-CR regime over-allocates Gaussian primitives to low-frequency content at the expense of generalisation to unseen scales.
        At gigapixel resolution, the expectation that LRs confer an advantage in super-resolution does not hold uniformly, and practitioners should not discount classical baselines.

        \begin{figure}[t]
    \centering
    \resizebox{\linewidth}{!}{
    \setlength{\tabcolsep}{1px}
    \begin{tabular}{l cccc}
    
        & \textbf{GT} & \textbf{Implicit} & \textbf{Hybrid} & \textbf{Explicit} \\

        \rotatebox{90}{ \small \hspace{0.75em} \textit{Venice}} &
        \includegraphics[width=0.2\linewidth]{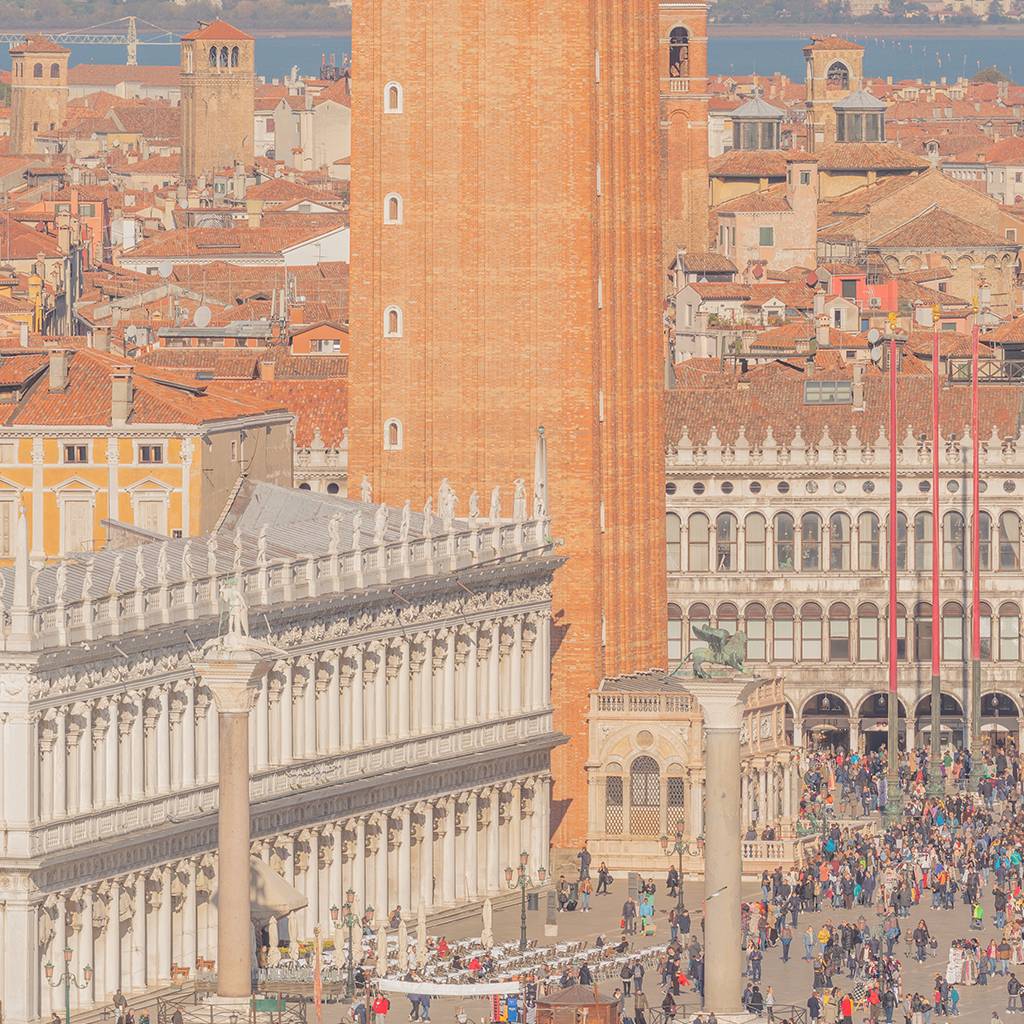} & 
        \includegraphics[width=0.2\linewidth]{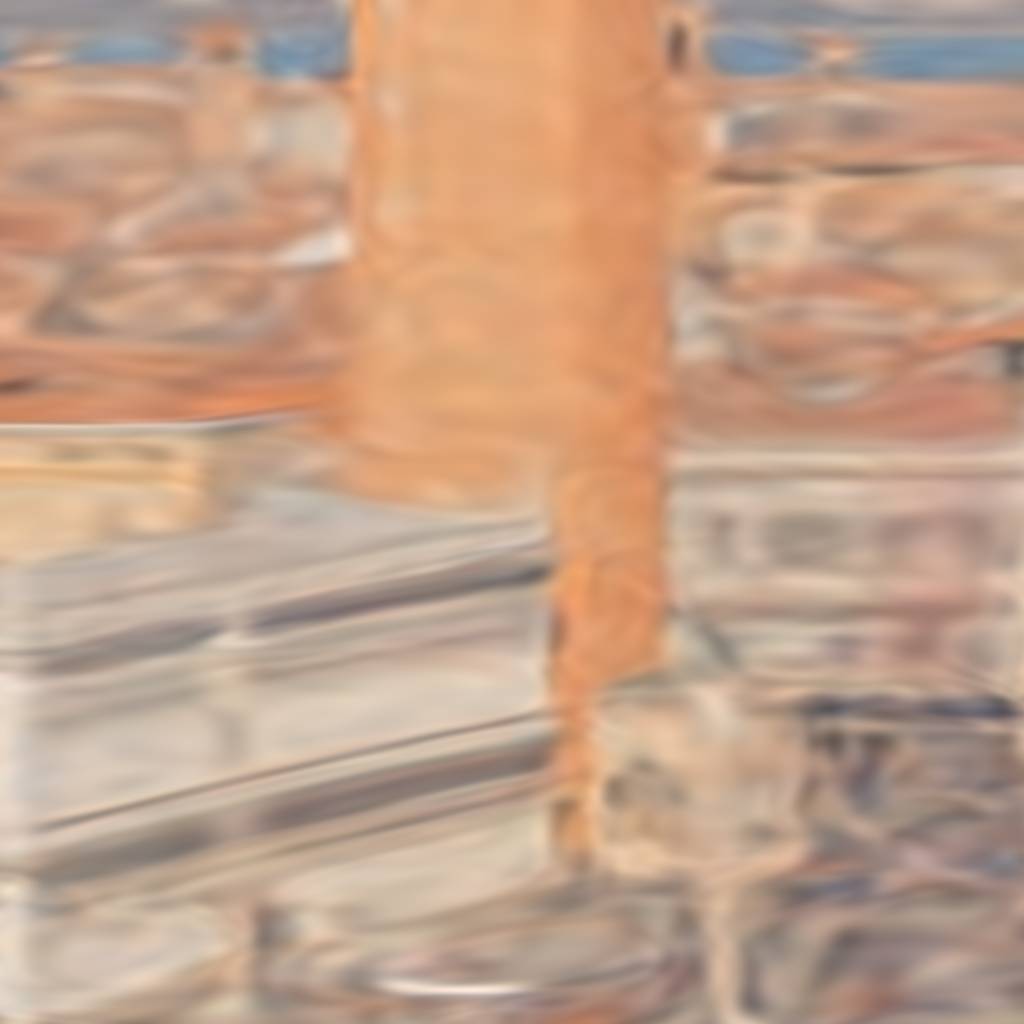} & 
        \includegraphics[width=0.2\linewidth]{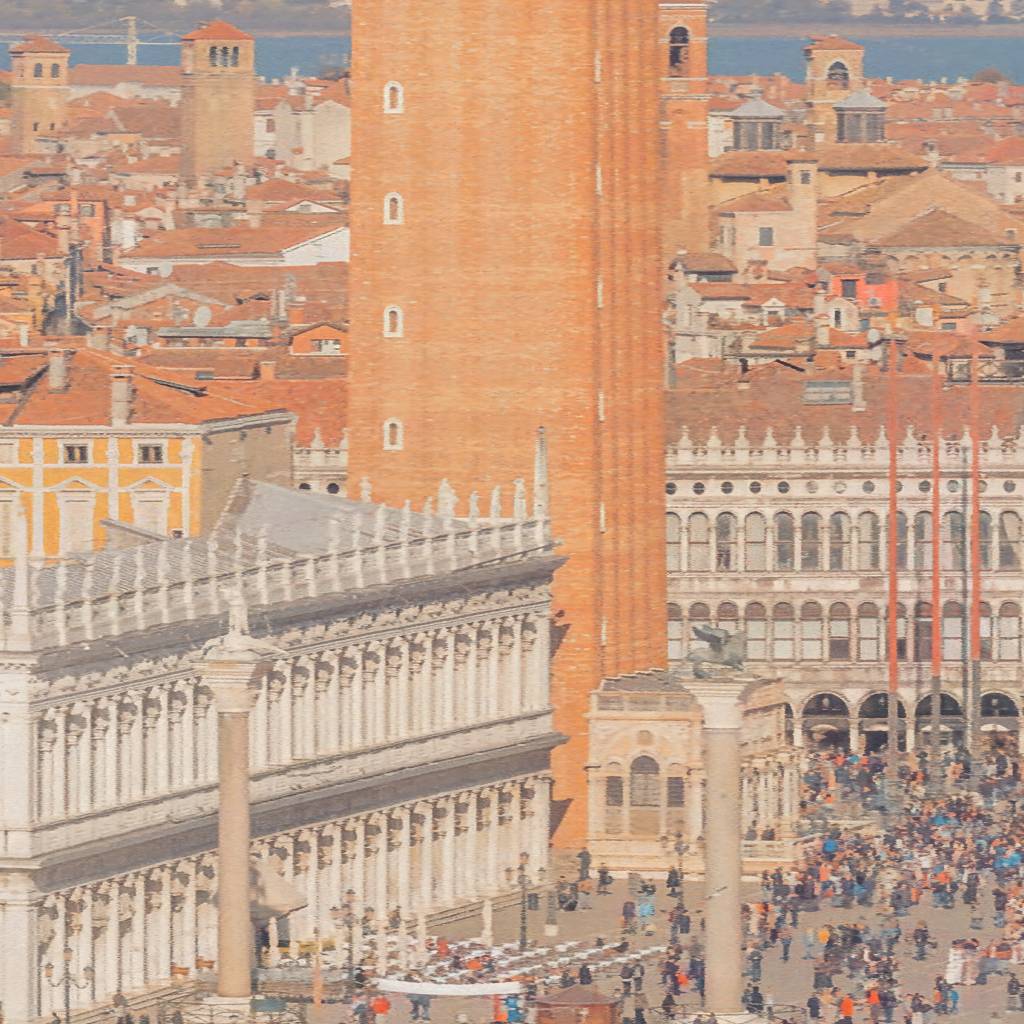} & 
        \includegraphics[width=0.2\linewidth]{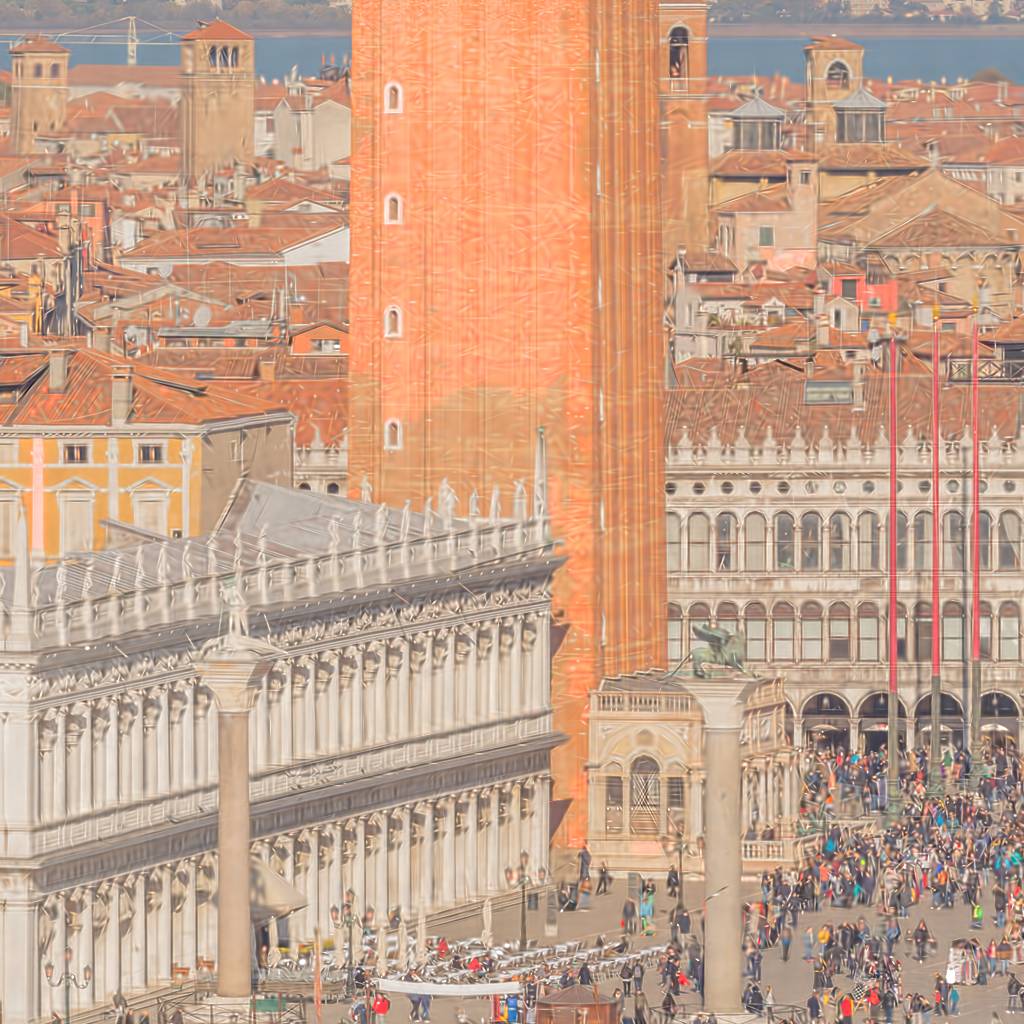} \\

        \rotatebox{90}{ \small \hspace{0.5em} \textit{Munich}} &
        \includegraphics[width=0.2\linewidth]{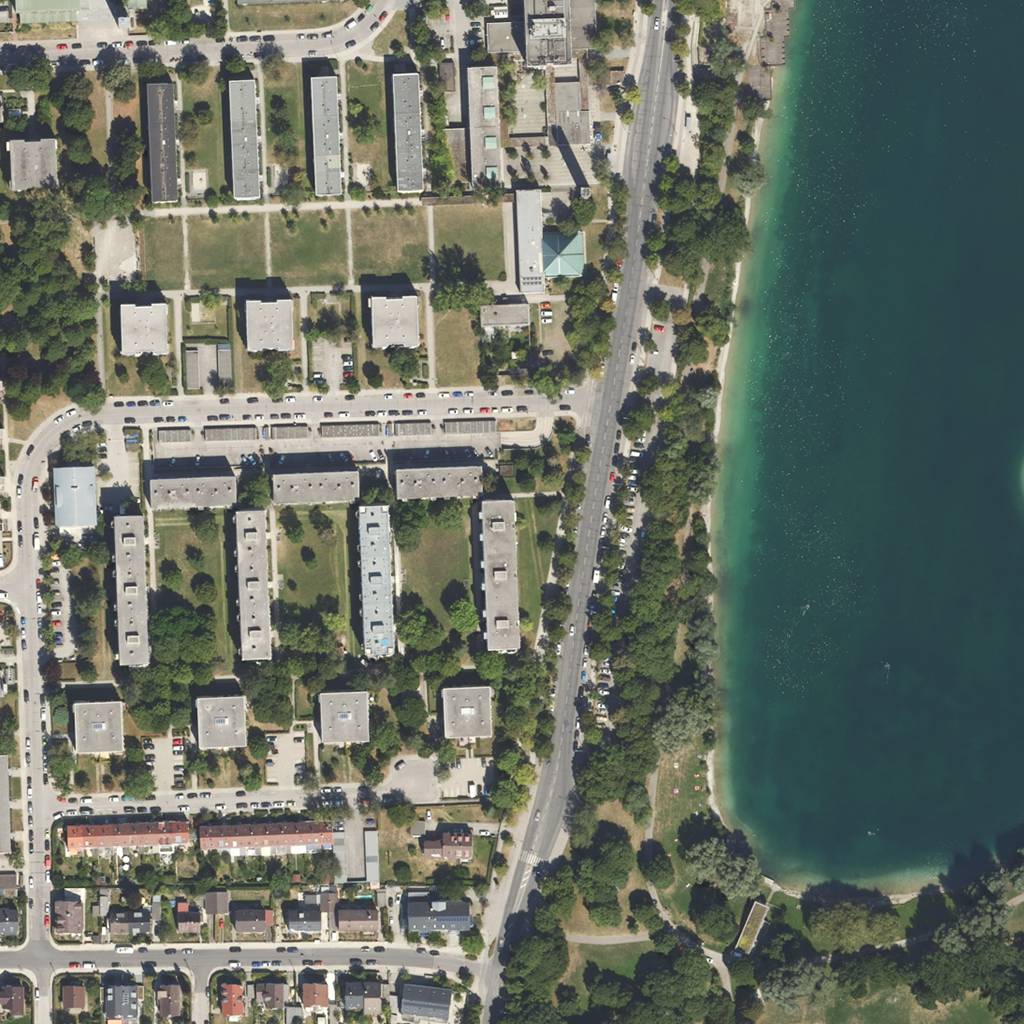} & 
        \includegraphics[width=0.2\linewidth]{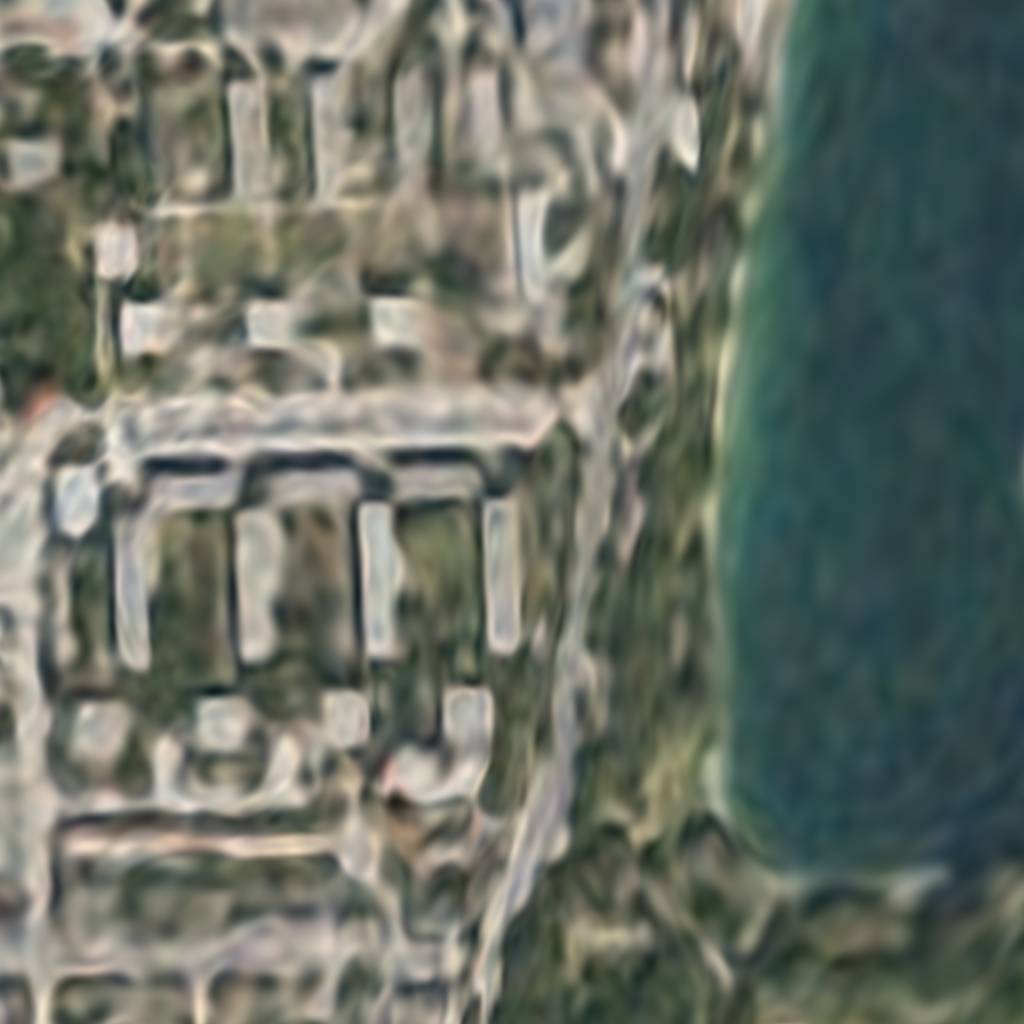} & 
        \includegraphics[width=0.2\linewidth]{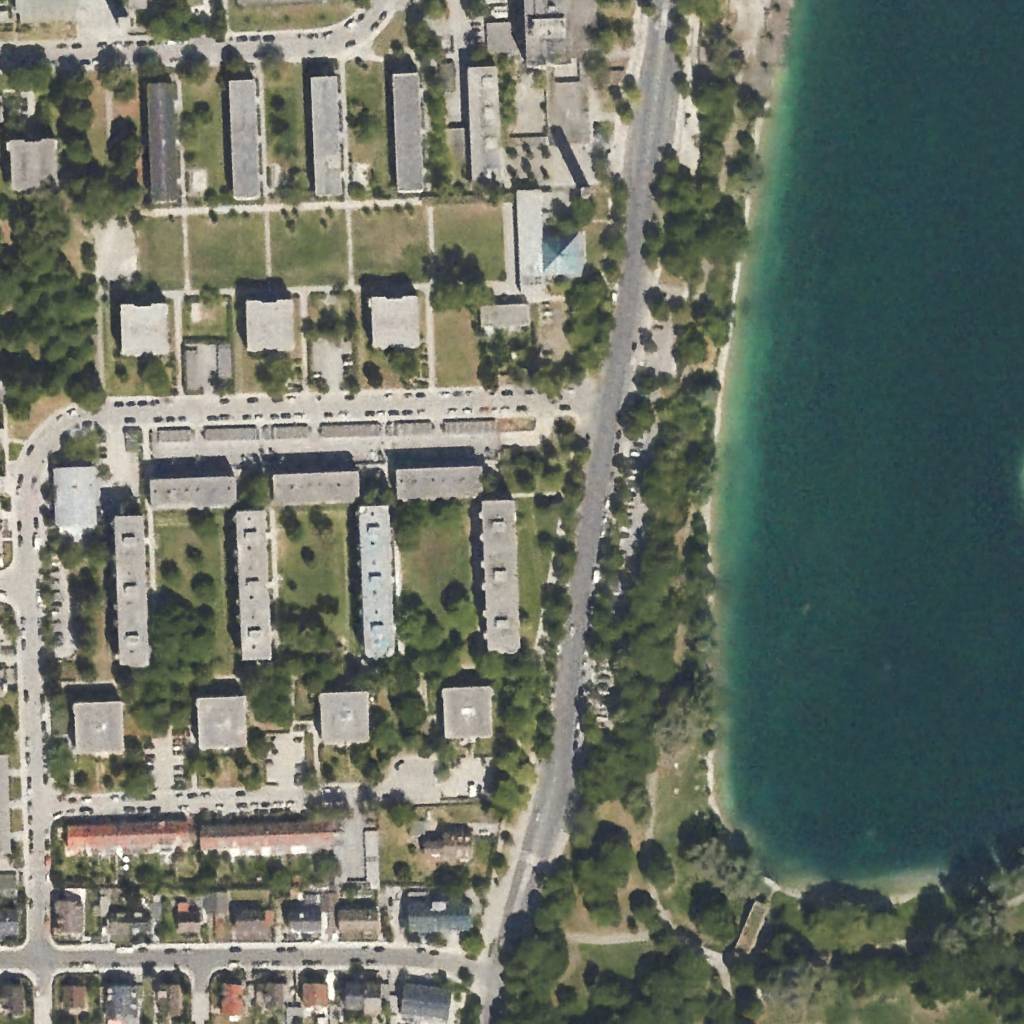} & 
        \includegraphics[width=0.2\linewidth]{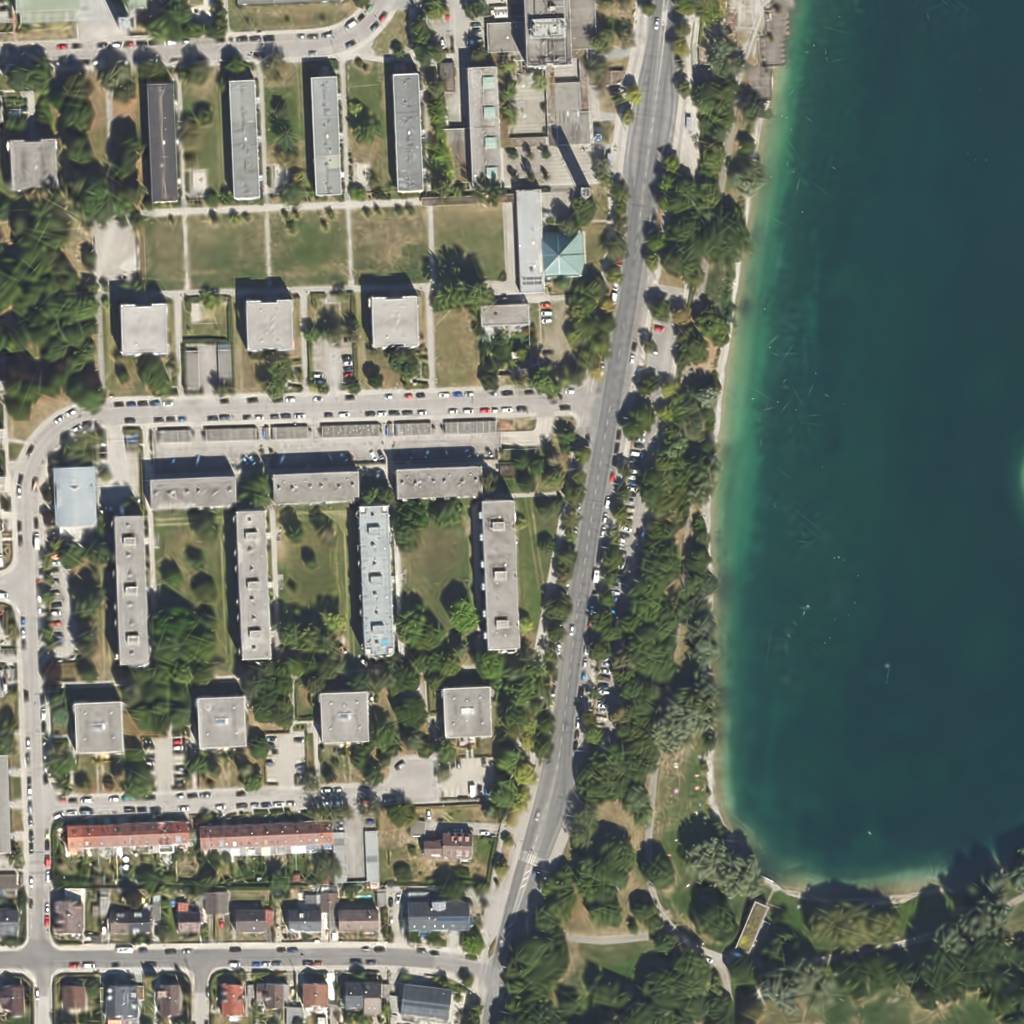} \\

        \rotatebox{90}{ \small \hspace{0.4em} \textit{Br. Can.}} &
        \includegraphics[width=0.2\linewidth]{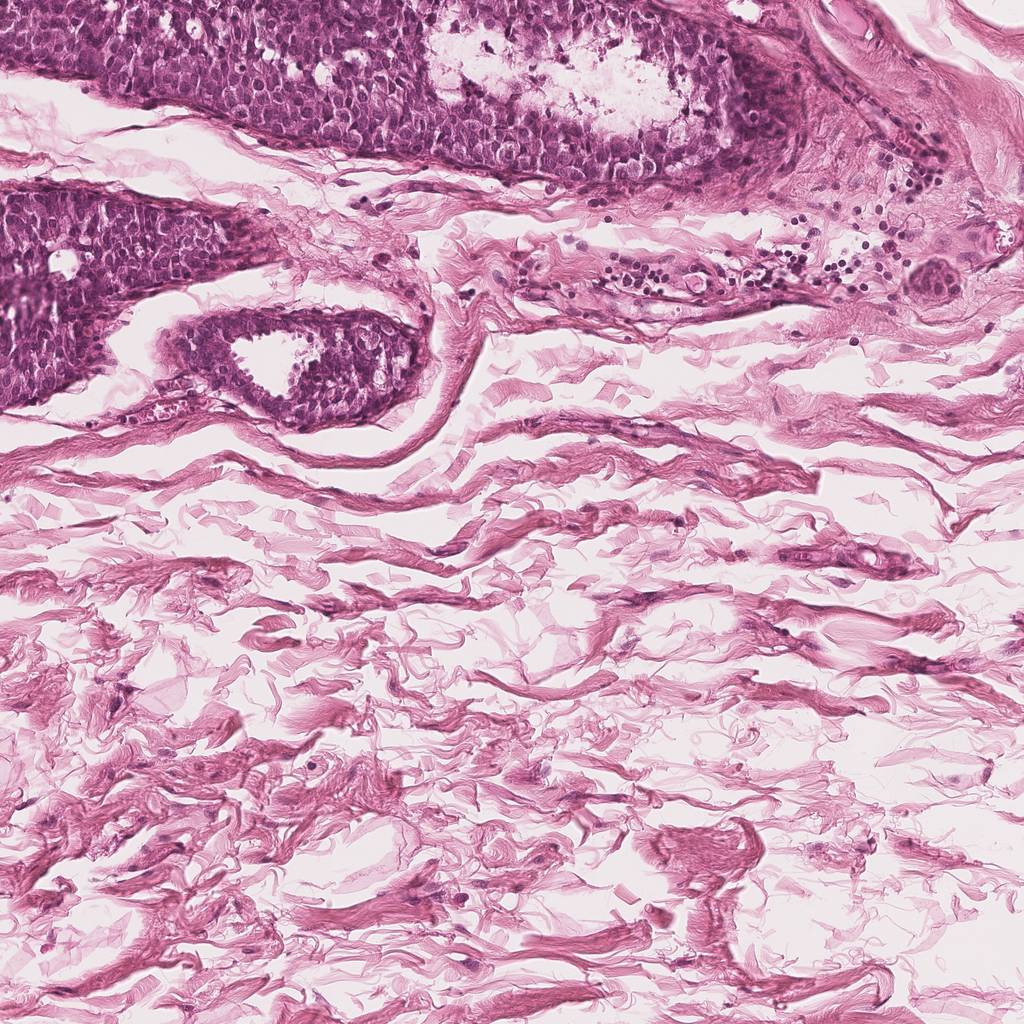} & 
        \includegraphics[width=0.2\linewidth]{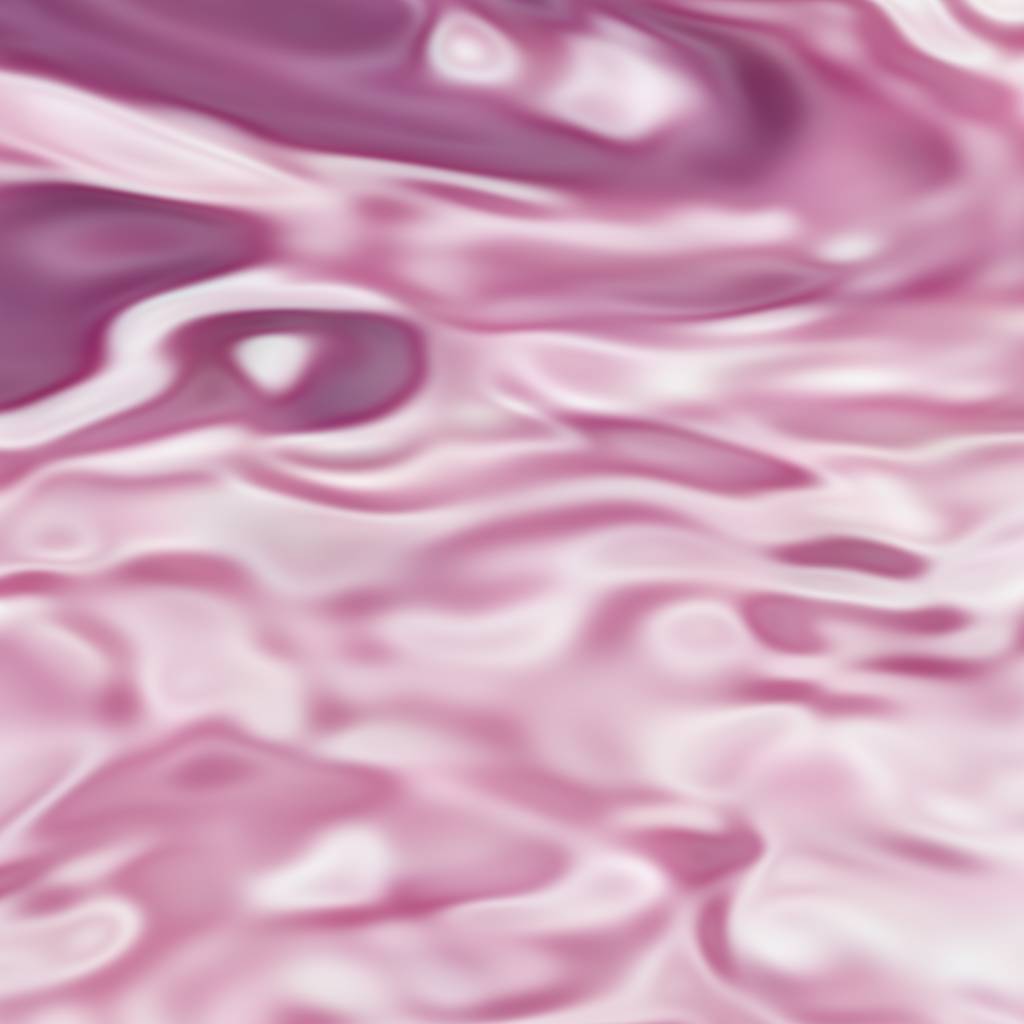} & 
        \includegraphics[width=0.2\linewidth]{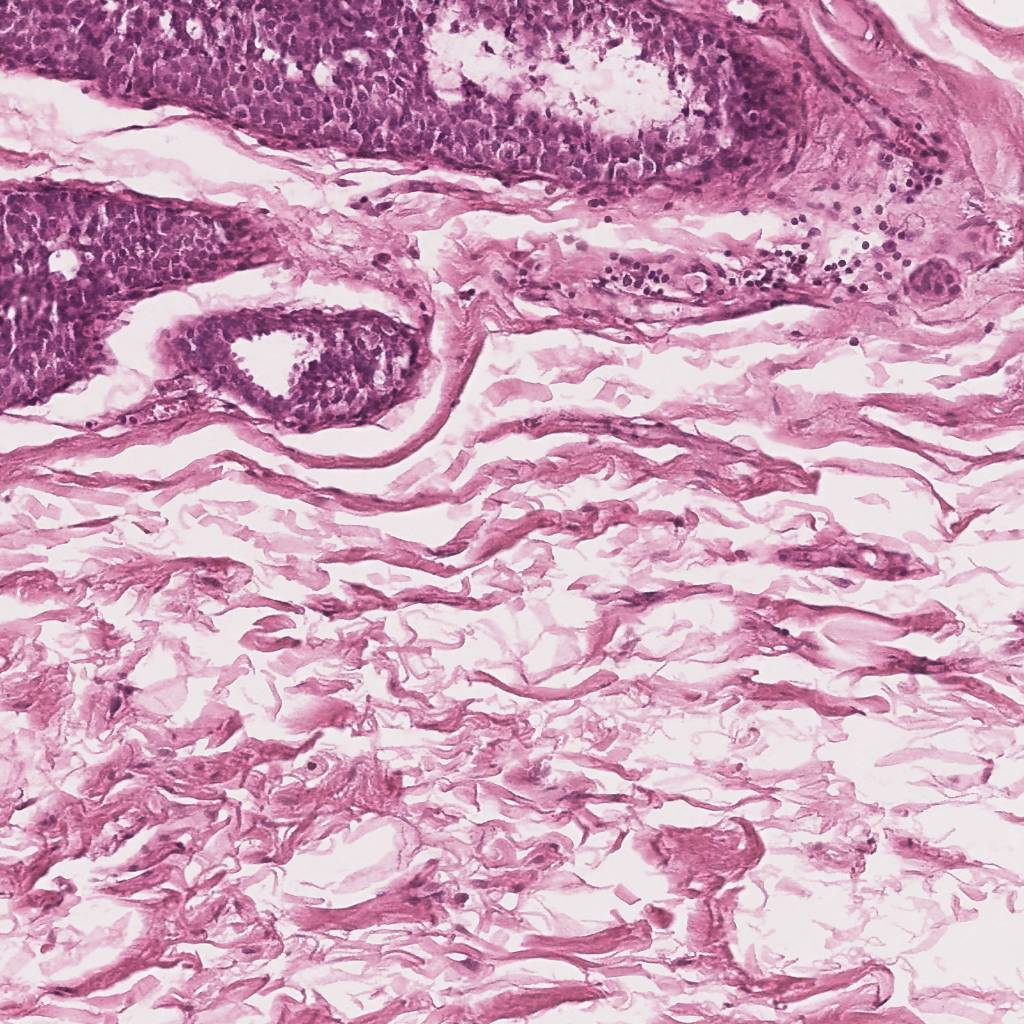} & 
        \includegraphics[width=0.2\linewidth]{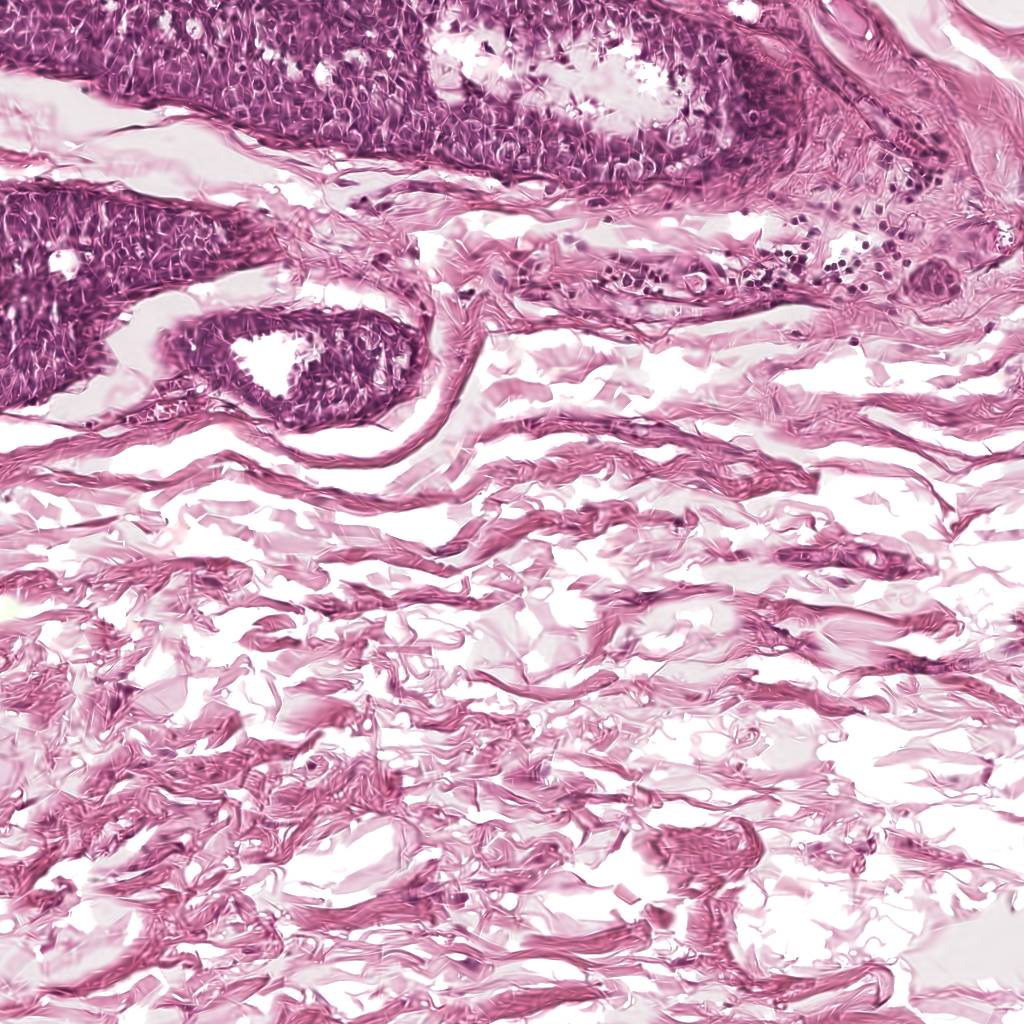} \\

    \end{tabular}}
    \caption{
    \textbf{Compression on \dataname{}: qualitative results.}
    We show $1024 \times 1024$ crops from full-resolution images and consider the best method for each family, as highlighted by the different colours along the columns of \cref{tab:compression}.
    }
    \label{fig:qualitatives_compression}
\end{figure}

        \begin{figure*}[ht]
    \centering
    \resizebox{\linewidth}{!}{
    \setlength{\tabcolsep}{1px}
    \begin{tabular}{l ccc @{\hskip 0.5em} cc @{\hskip 0.5em} cc @{\hskip 0.5em} cc @{\hskip 0.5em} cc}
    
        & \multicolumn{3}{c}{\textbf{Ground-truth}} & \multicolumn{2}{c}{\textbf{Implicit}} & \multicolumn{2}{c}{\textbf{Hybrid}} & \multicolumn{2}{c}{\textbf{Explicit}} & \multicolumn{2}{c}{\textbf{Classic}} \\
        & Full & $\div4$ & $\div8$ & $\times4$ & $\times8$ & $\times4$ & $\times8$ & $\times4$ & $\times8$ & $\times4$ & $\times8$ \\

        \rotatebox{90}{ \small \hspace{0.75em} \textit{Venice}} &
        \includegraphics[width=0.1\linewidth]{images/qualitatives/Venice-gt.jpg} & 
        \includegraphics[width=0.1\linewidth]{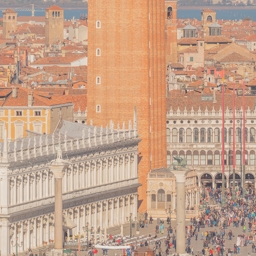} & 
        \includegraphics[width=0.1\linewidth]{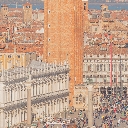} & 
        \includegraphics[width=0.1\linewidth]{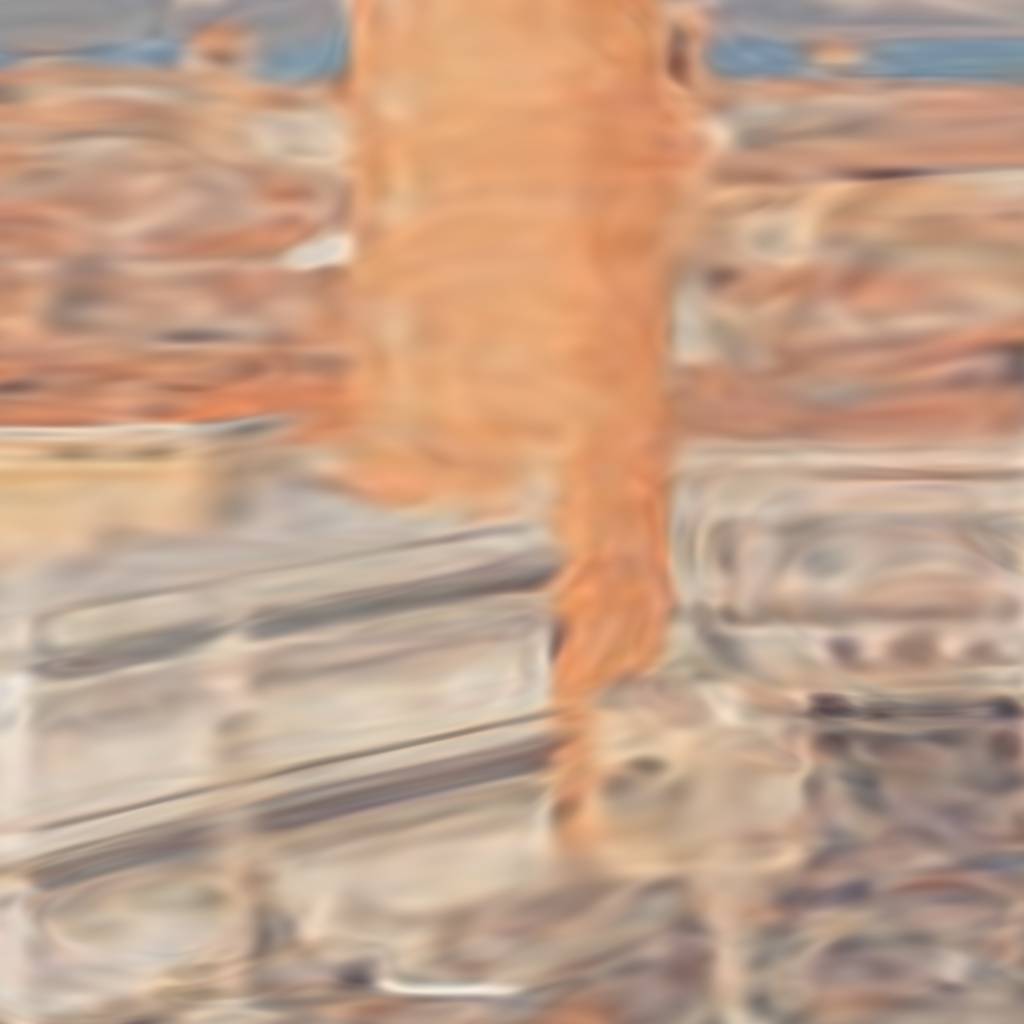} & 
        \includegraphics[width=0.1\linewidth]{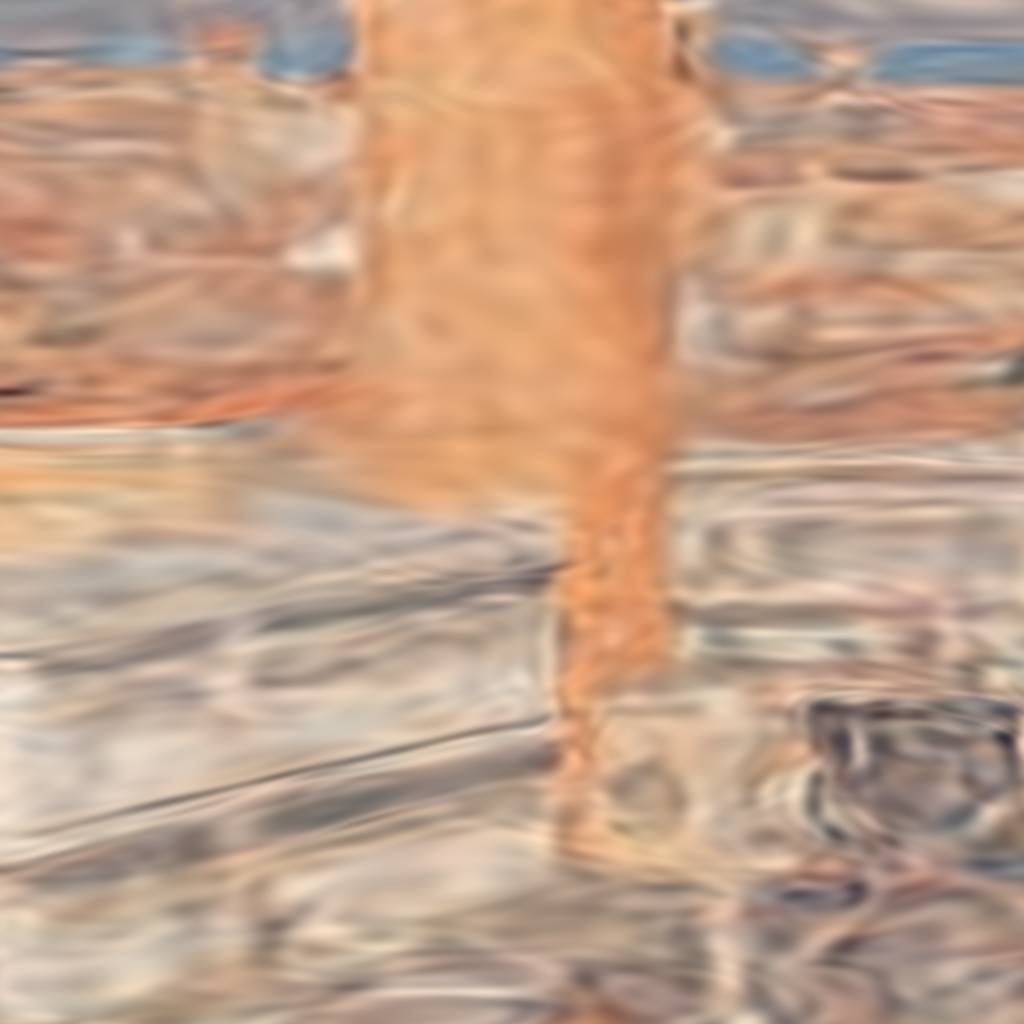} & 
        \includegraphics[width=0.1\linewidth]{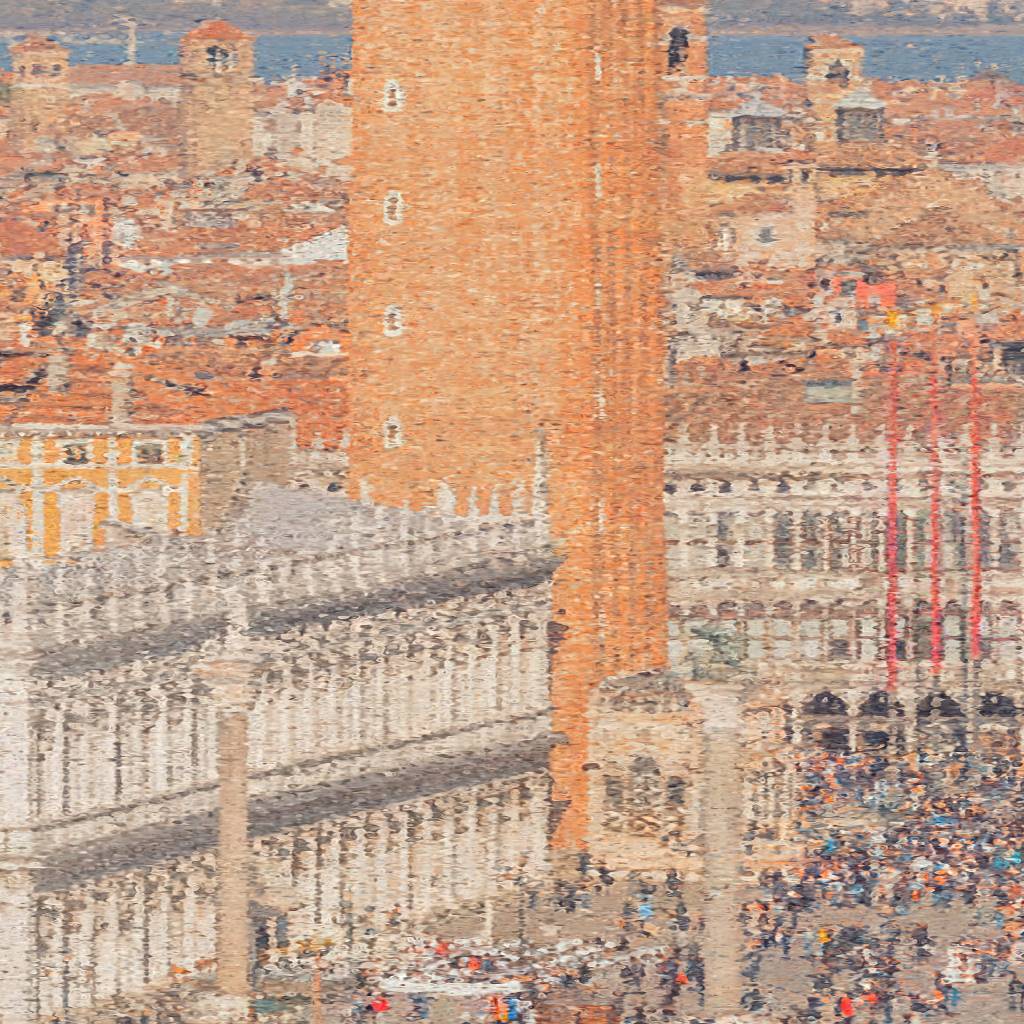} & 
        \includegraphics[width=0.1\linewidth]{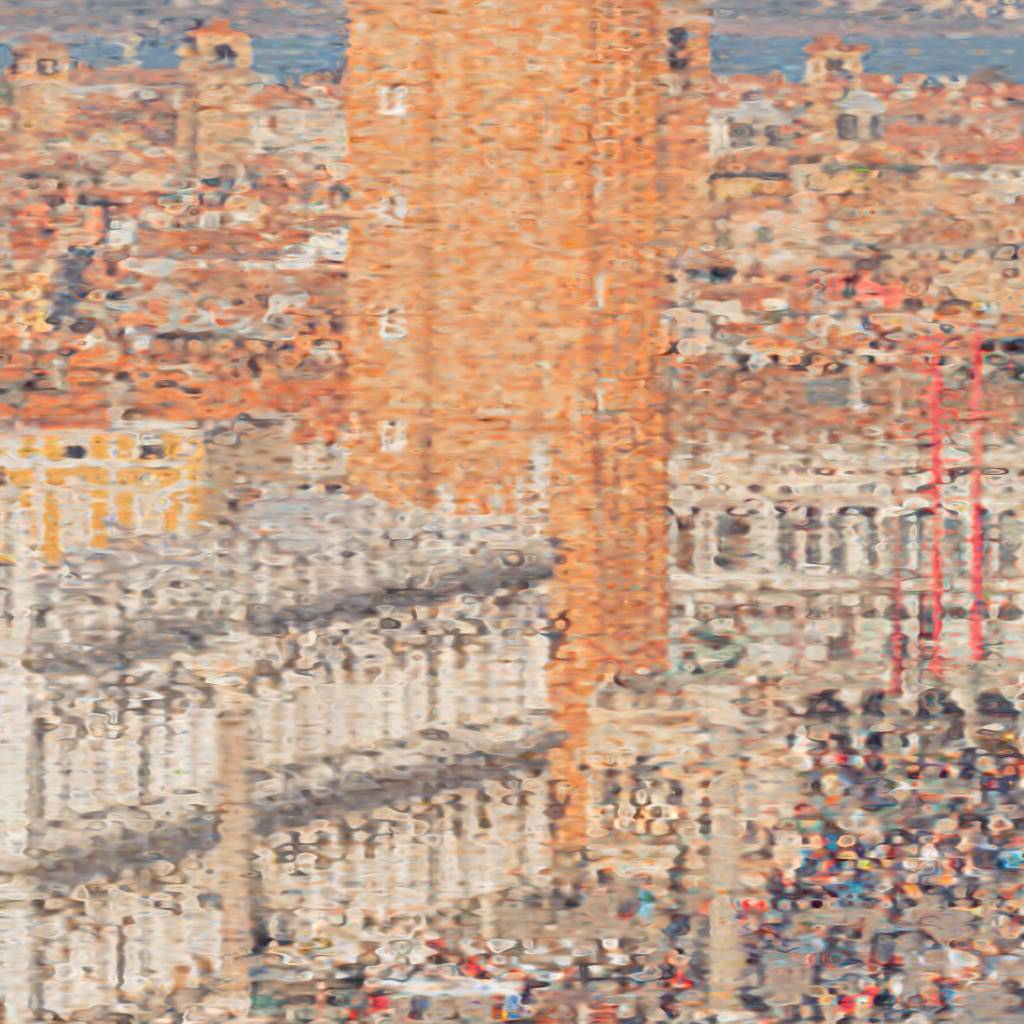} & 
        \includegraphics[width=0.1\linewidth]{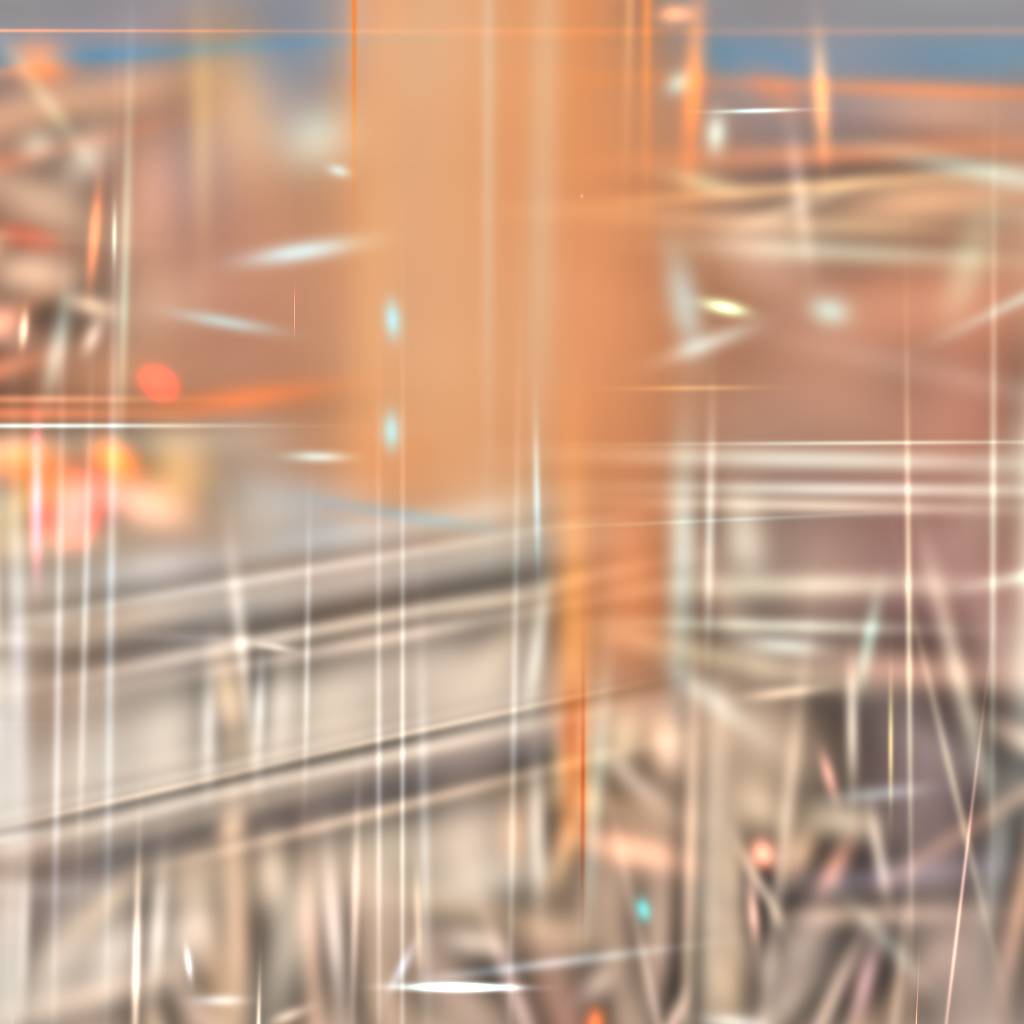} & 
        \includegraphics[width=0.1\linewidth]{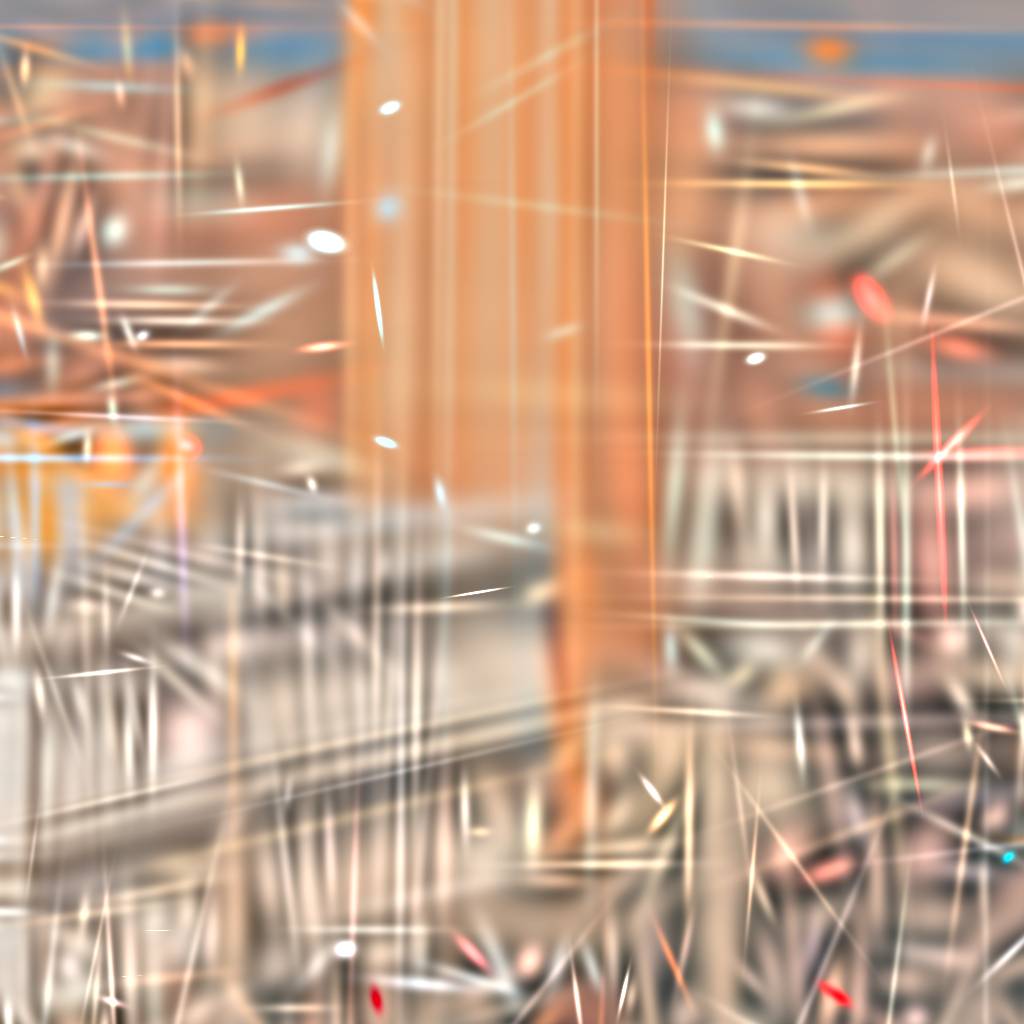} &
        \includegraphics[width=0.1\linewidth]{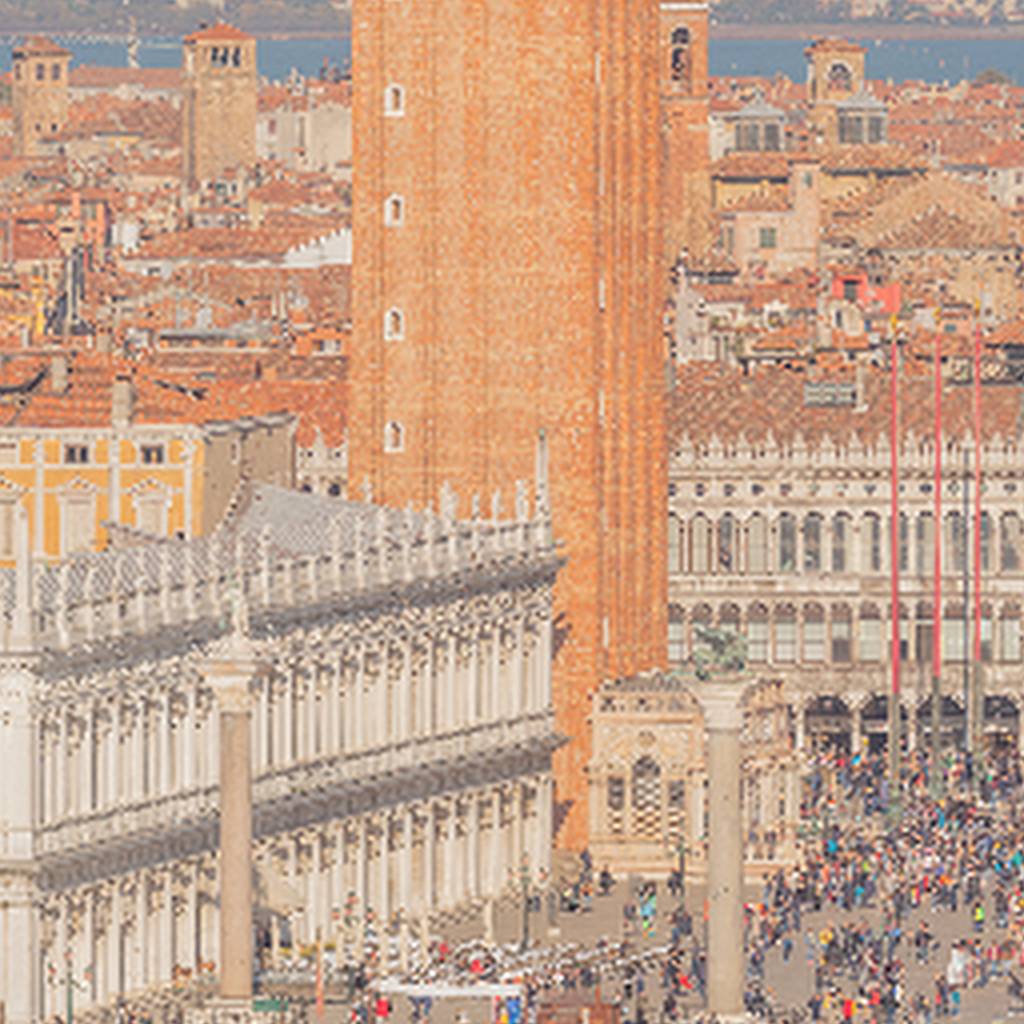} & 
        \includegraphics[width=0.1\linewidth]{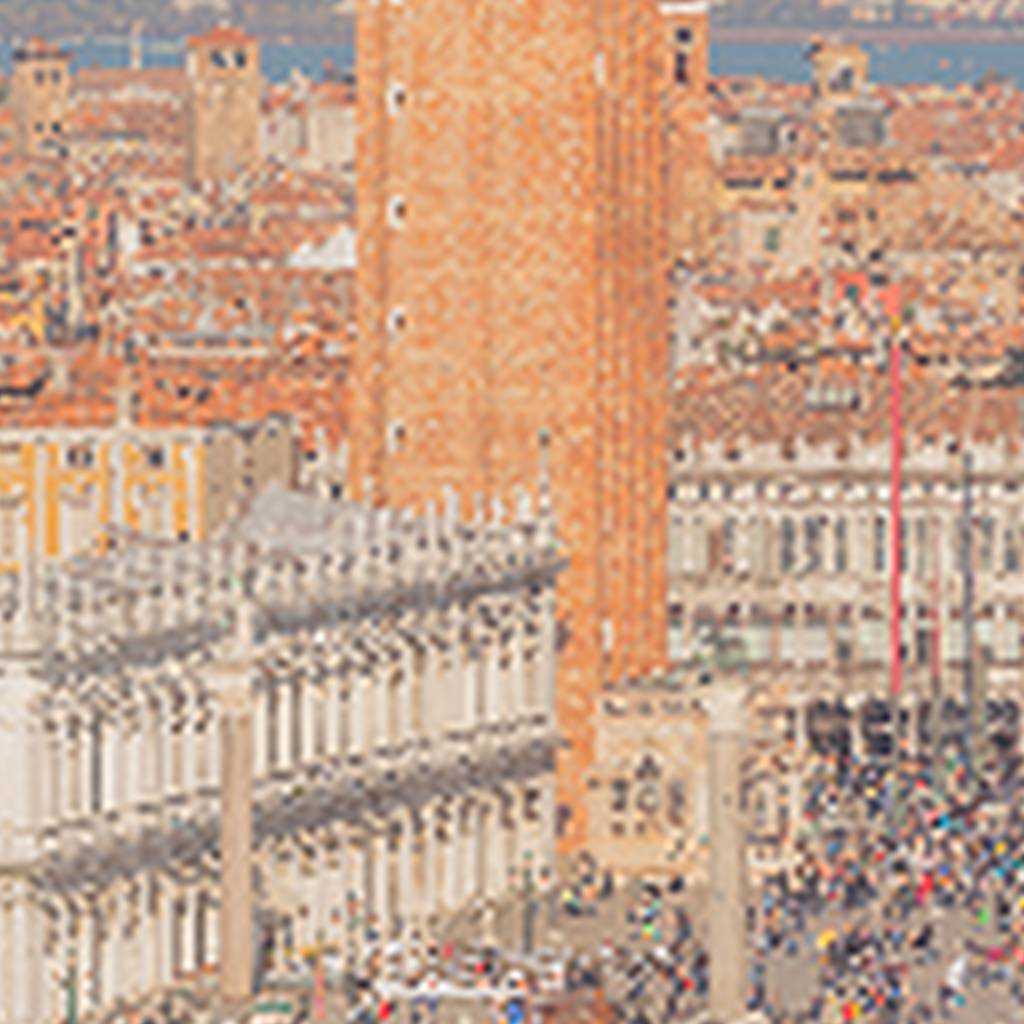} \\

        \rotatebox{90}{ \small \hspace{0.5em} \textit{Munich}} &
        \includegraphics[width=0.1\linewidth]{images/qualitatives/Munich-gt.jpg} & 
        \includegraphics[width=0.1\linewidth]{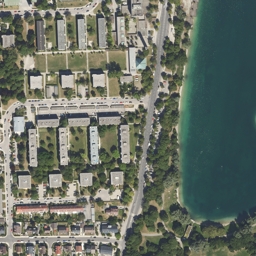} & 
        \includegraphics[width=0.1\linewidth]{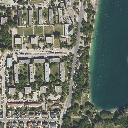} & 
        \includegraphics[width=0.1\linewidth]{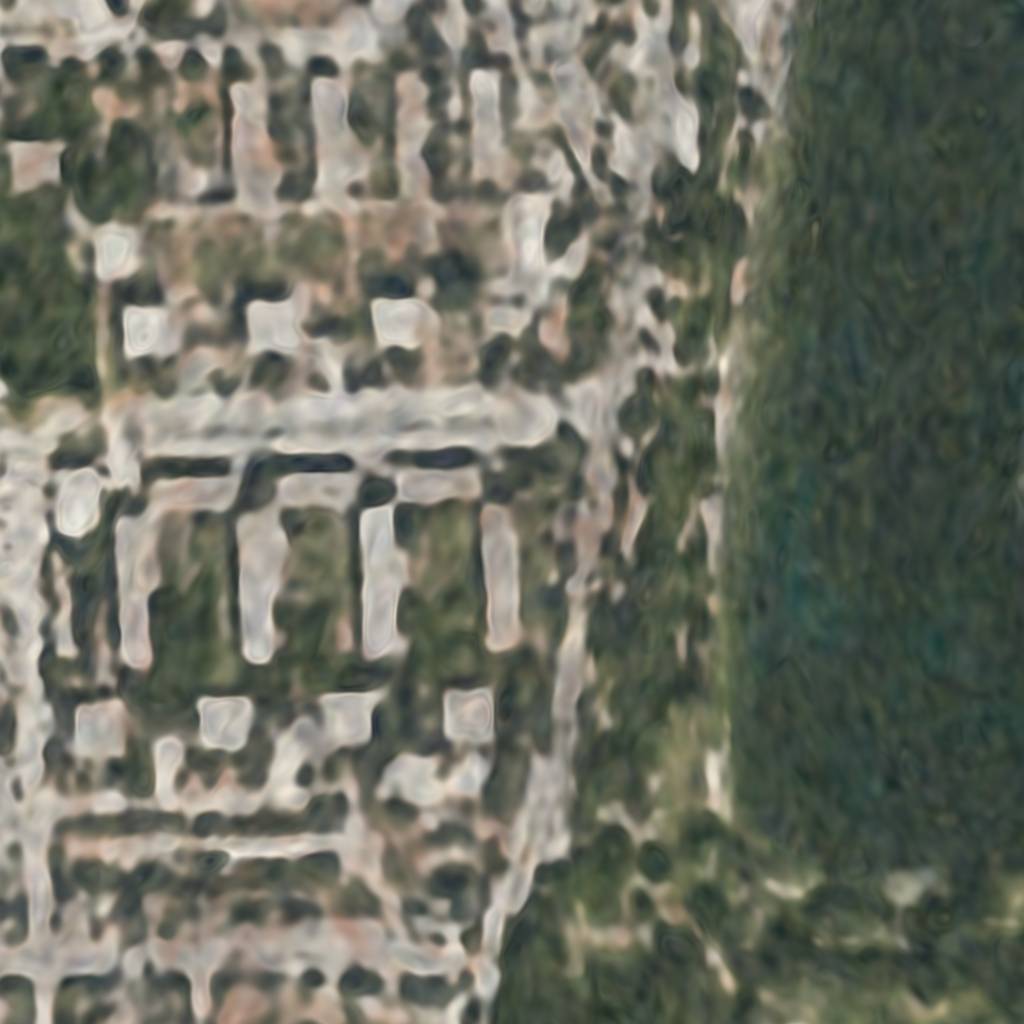} & 
        \includegraphics[width=0.1\linewidth]{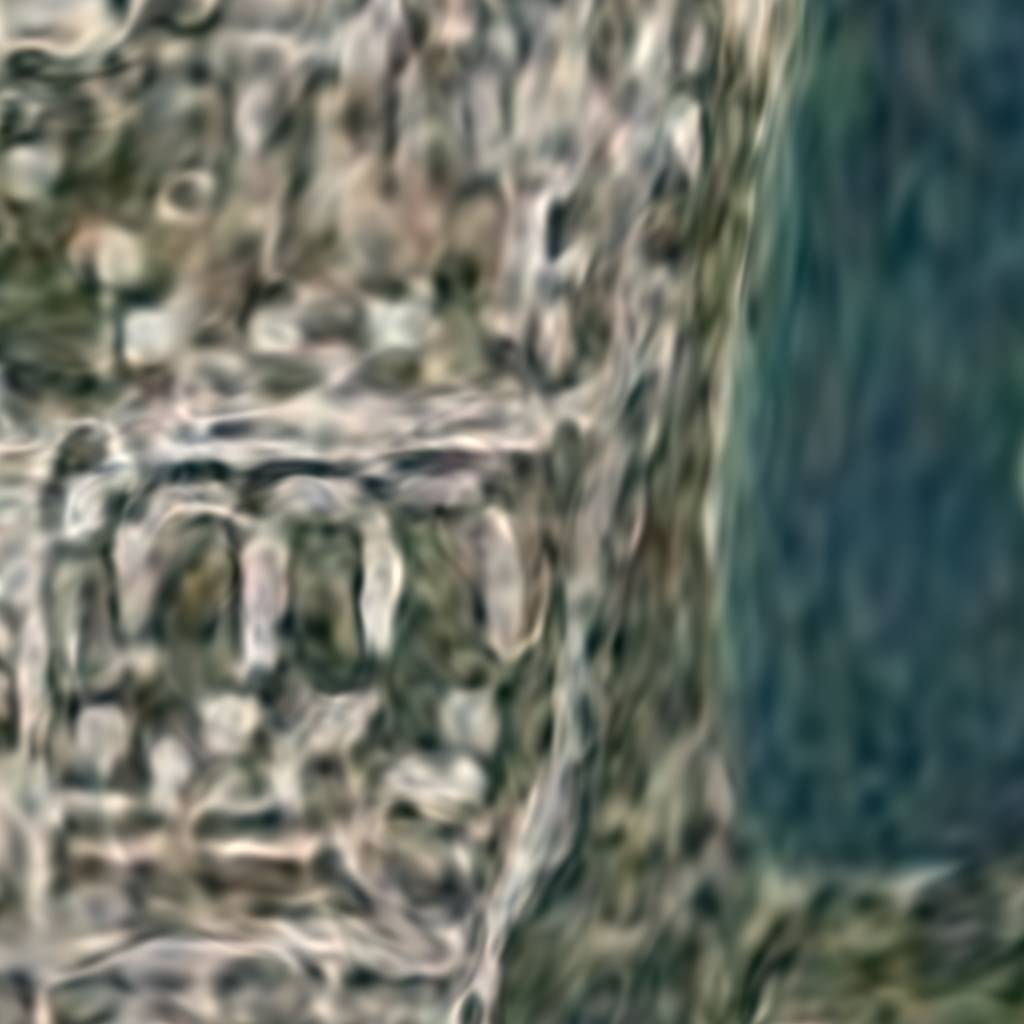} & 
        \includegraphics[width=0.1\linewidth]{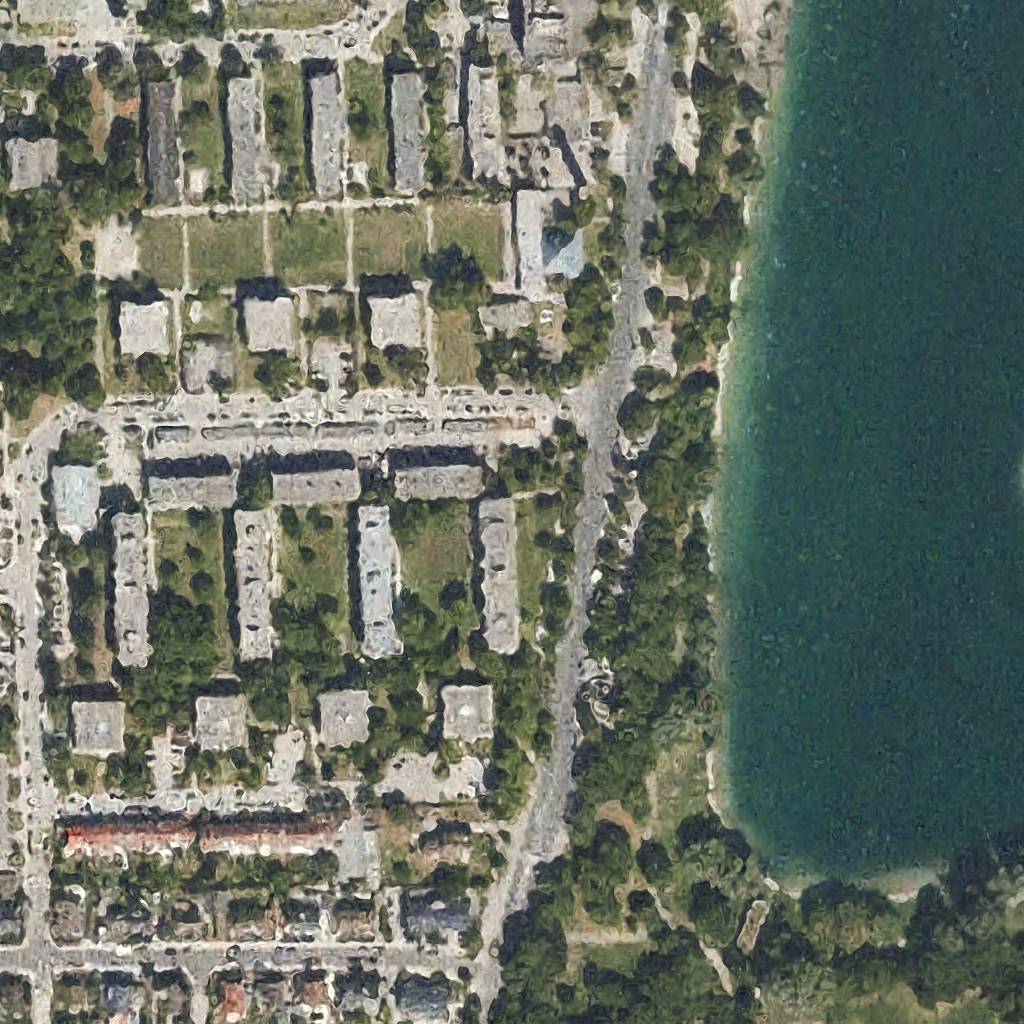} & 
        \includegraphics[width=0.1\linewidth]{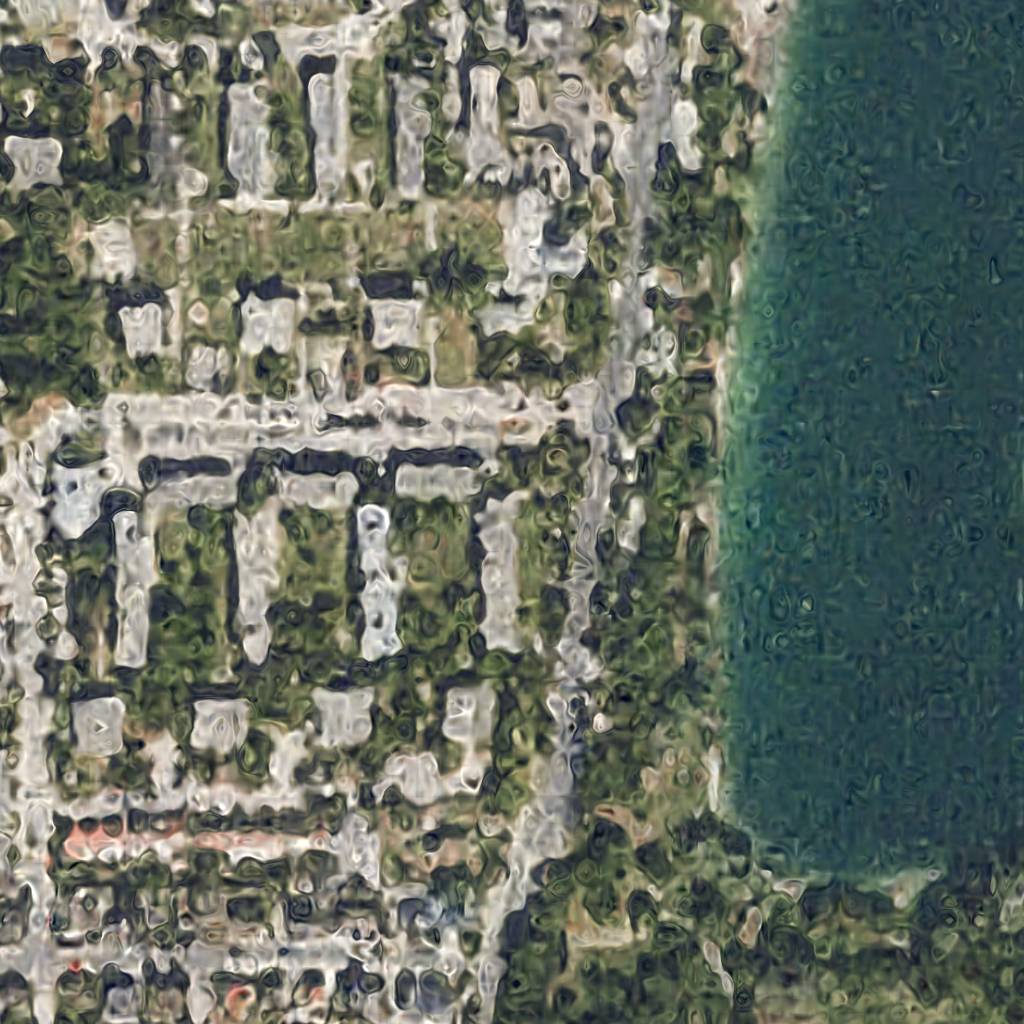} & 
        \includegraphics[width=0.1\linewidth]{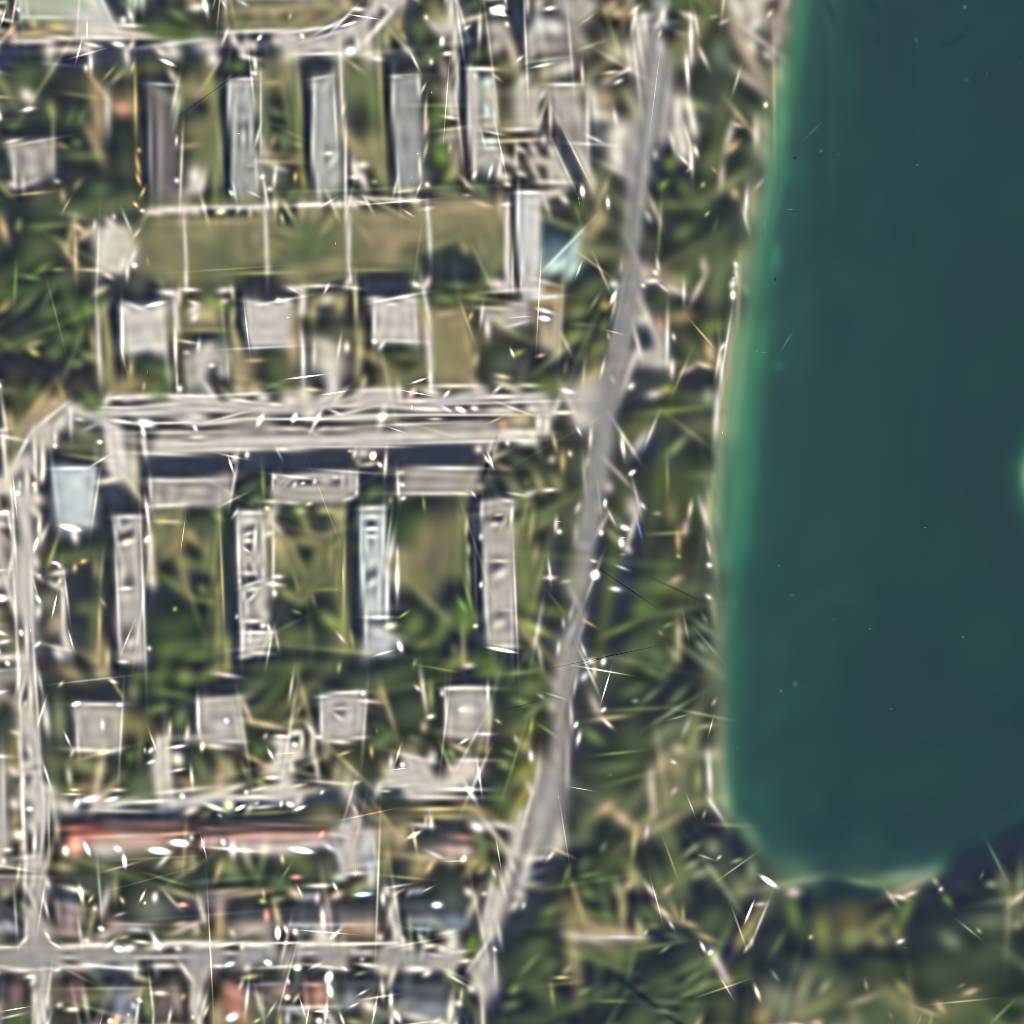} & 
        \includegraphics[width=0.1\linewidth]{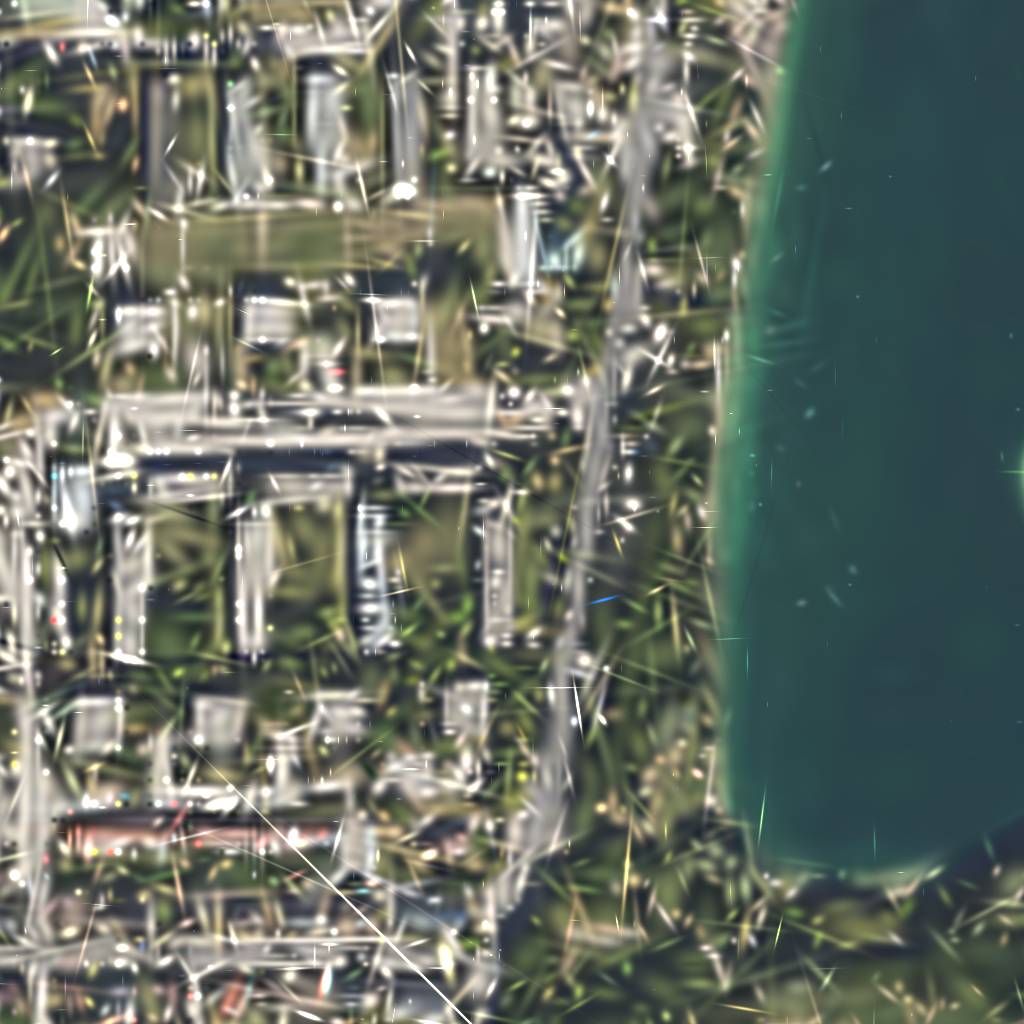} &
        \includegraphics[width=0.1\linewidth]{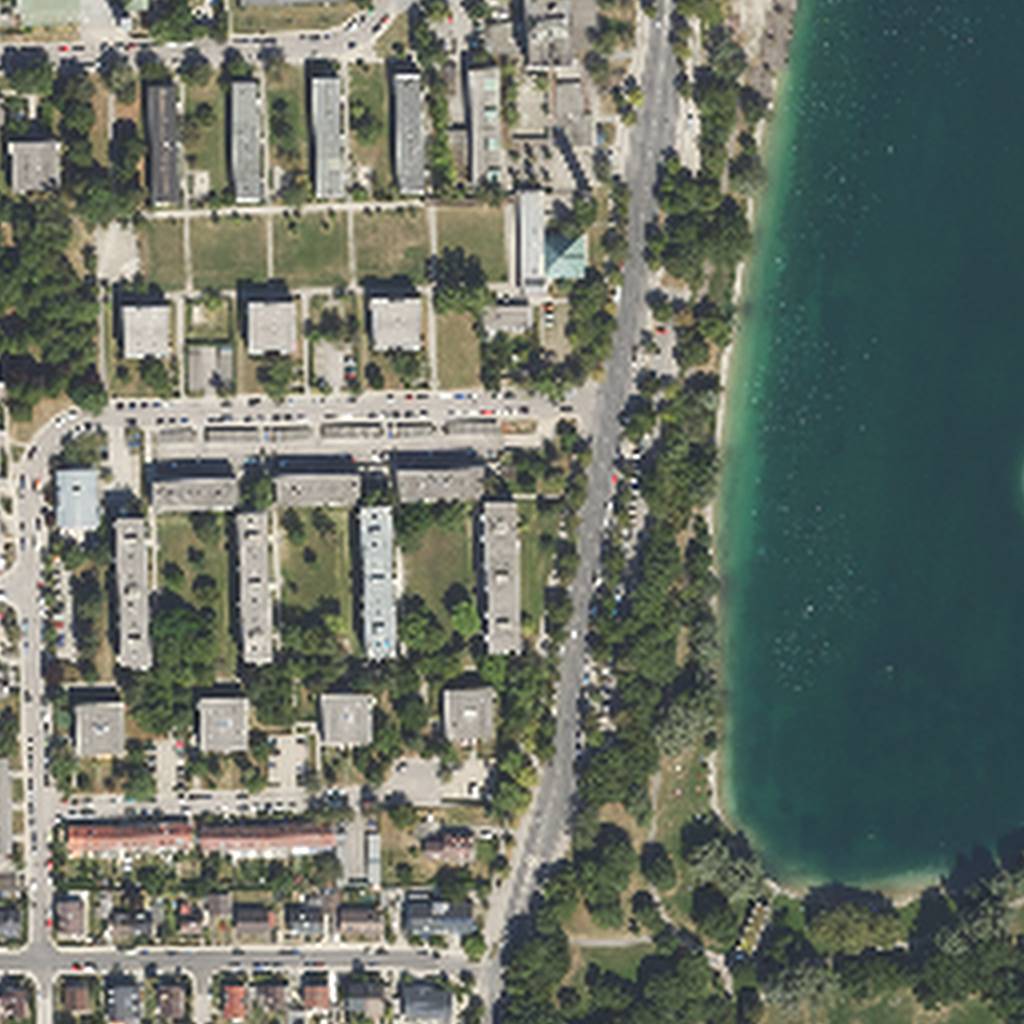} & 
        \includegraphics[width=0.1\linewidth]{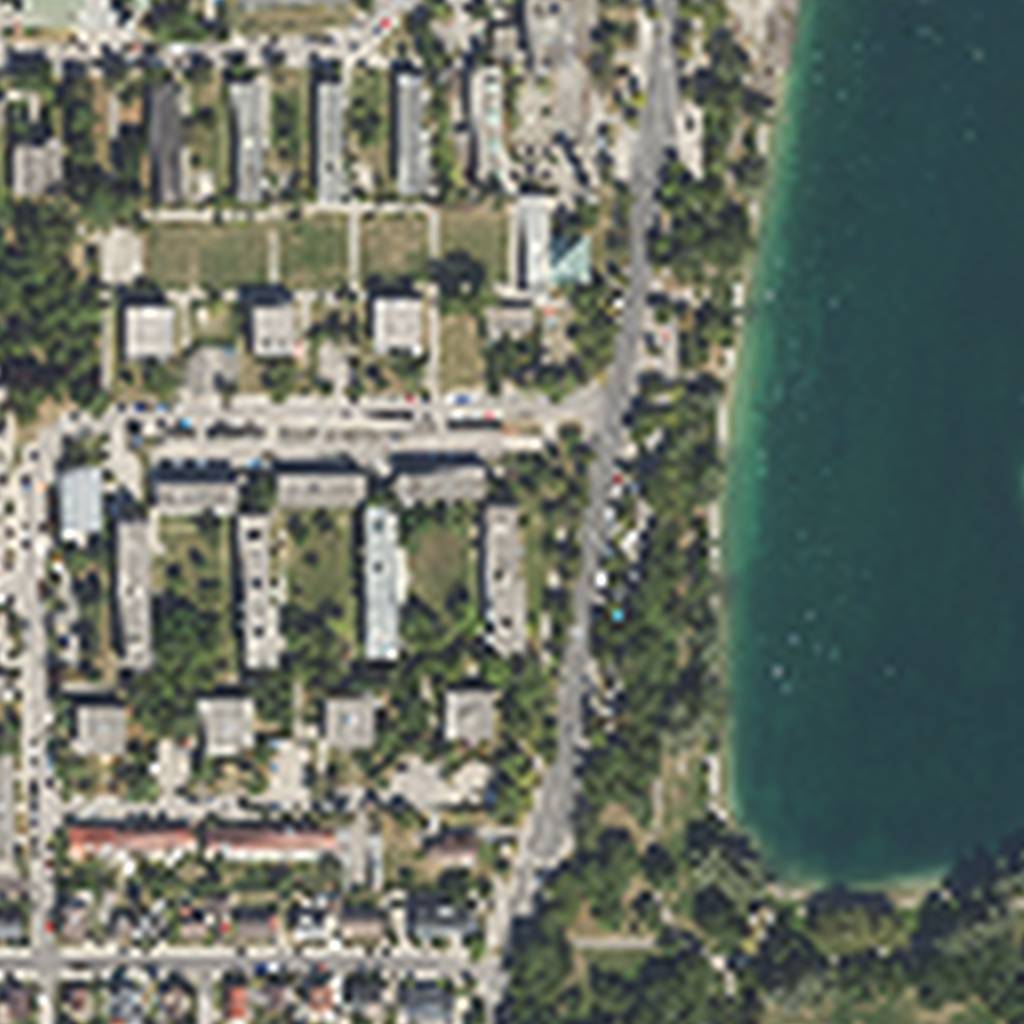} \\

        \rotatebox{90}{ \small \hspace{0.4em} \textit{Br. Can.}} &
        \includegraphics[width=0.1\linewidth]{images/qualitatives/BreastCancer-gt.jpg} & 
        \includegraphics[width=0.1\linewidth]{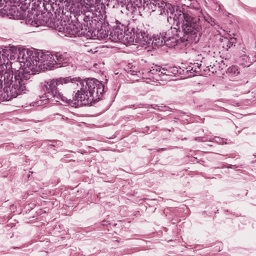} & 
        \includegraphics[width=0.1\linewidth]{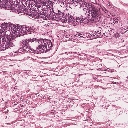} & 
        \includegraphics[width=0.1\linewidth]{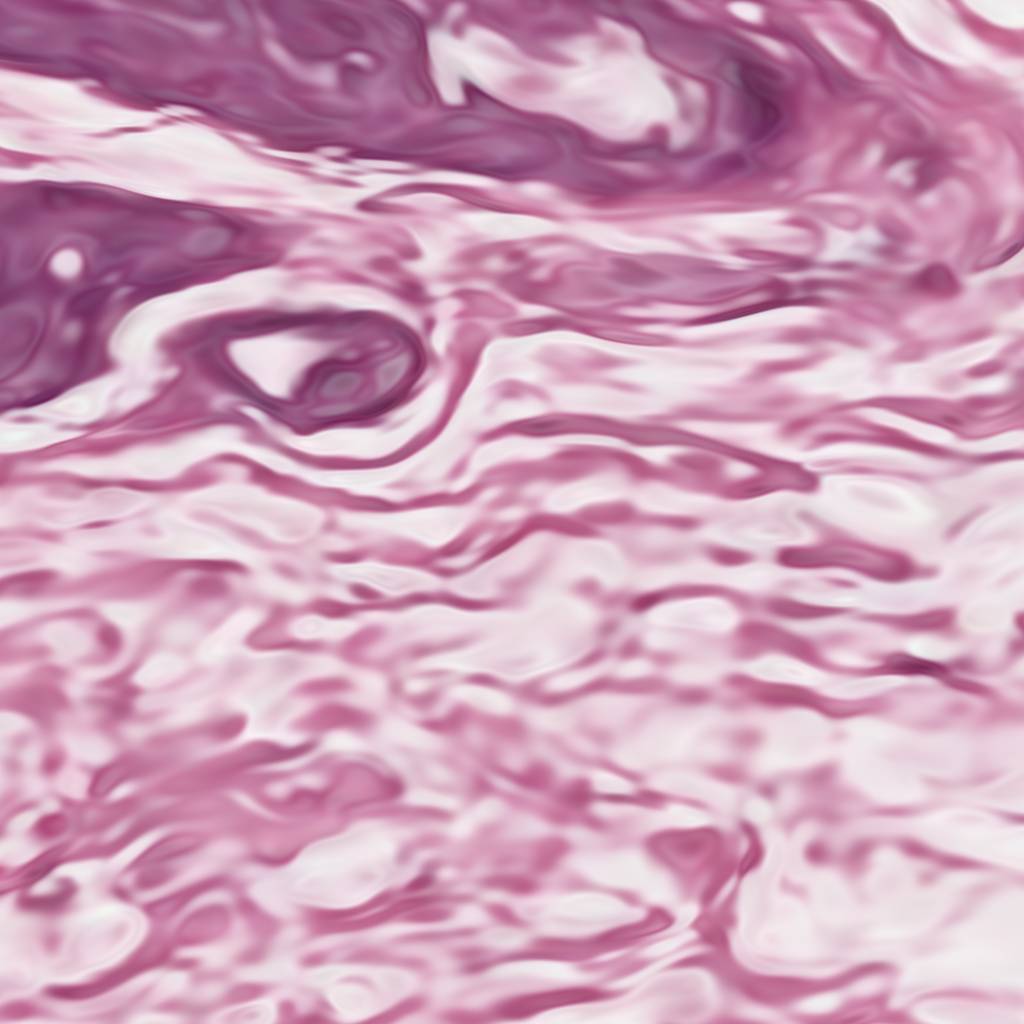} & 
        \includegraphics[width=0.1\linewidth]{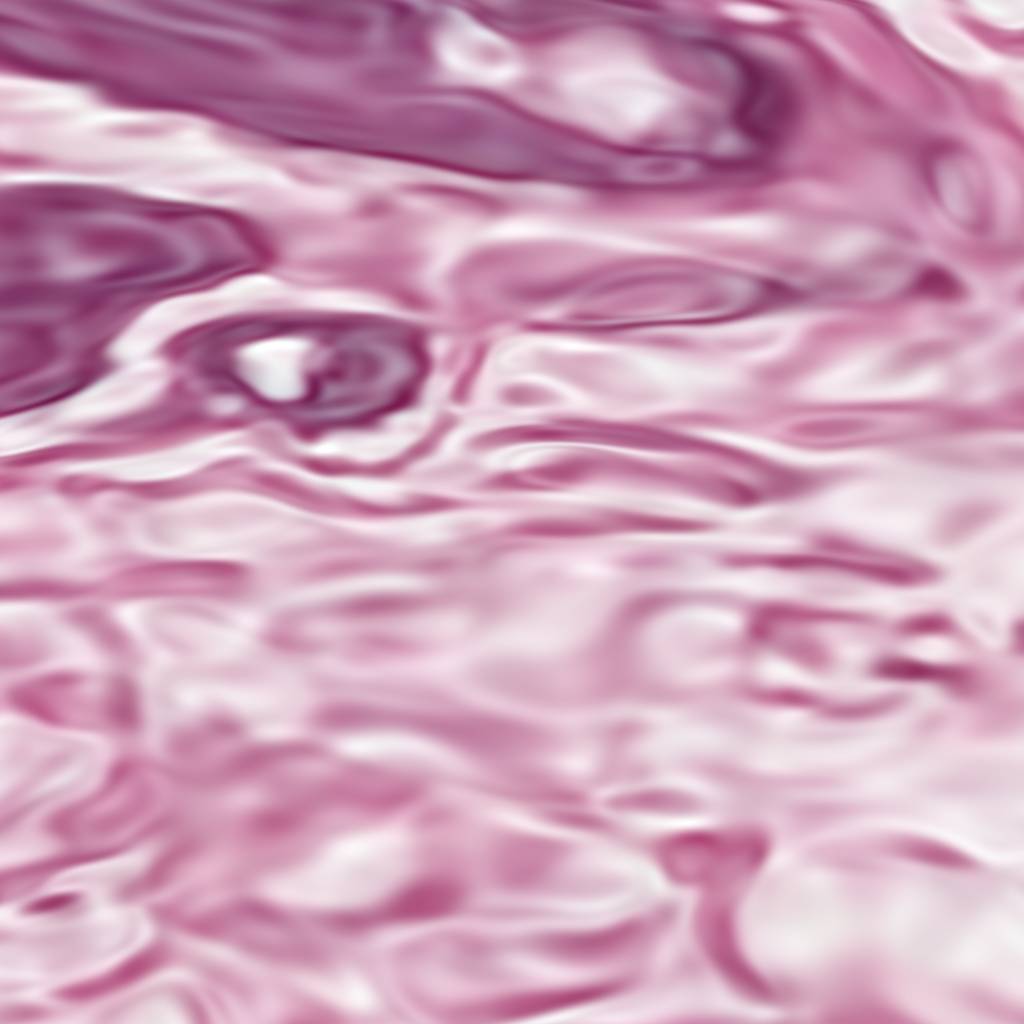} & 
        \includegraphics[width=0.1\linewidth]{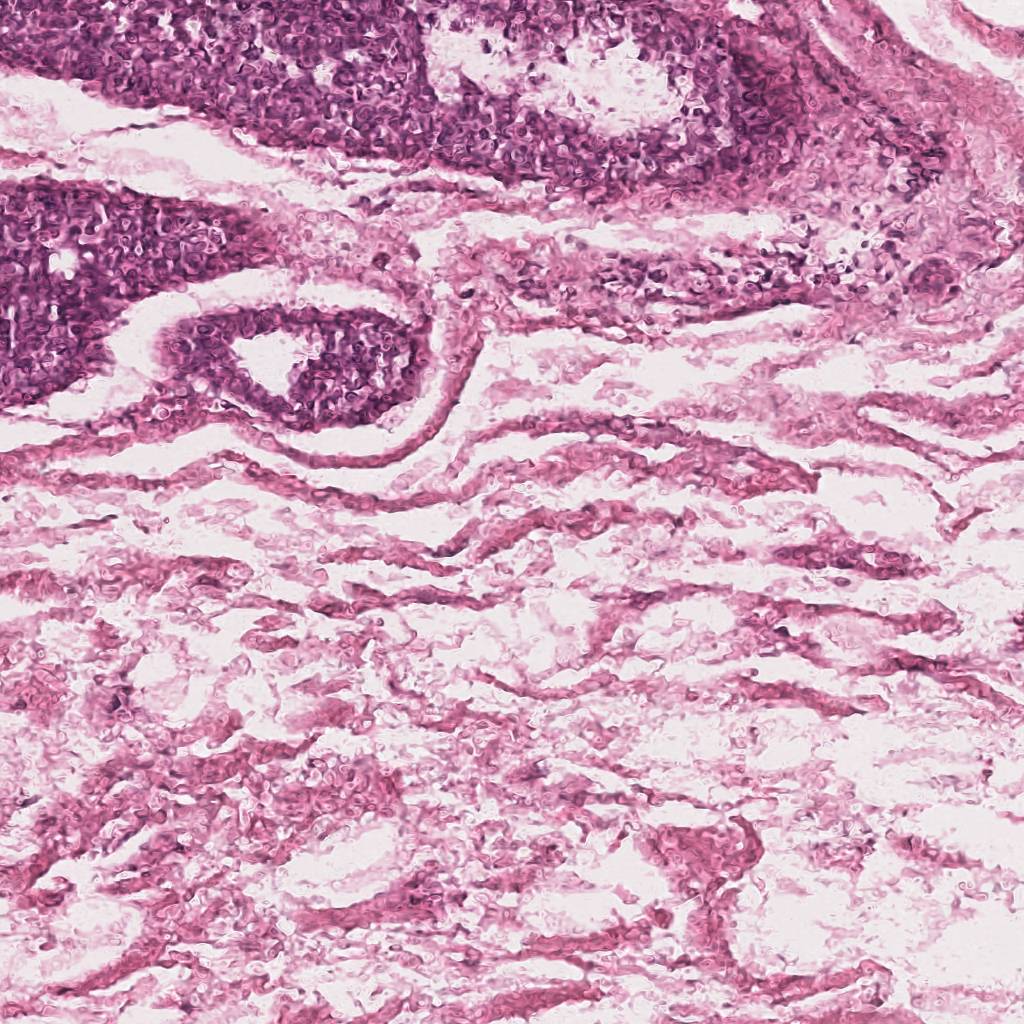} & 
        \includegraphics[width=0.1\linewidth]{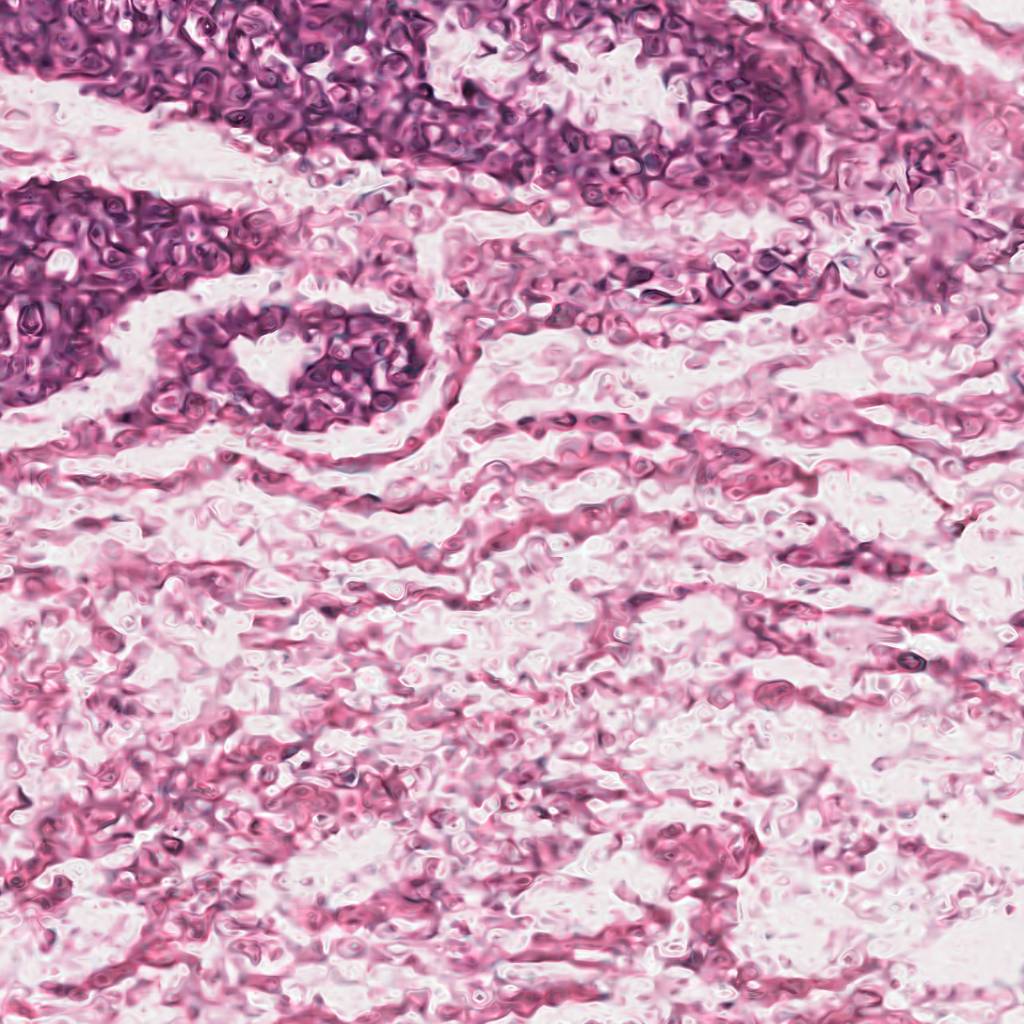} & 
        \includegraphics[width=0.1\linewidth]{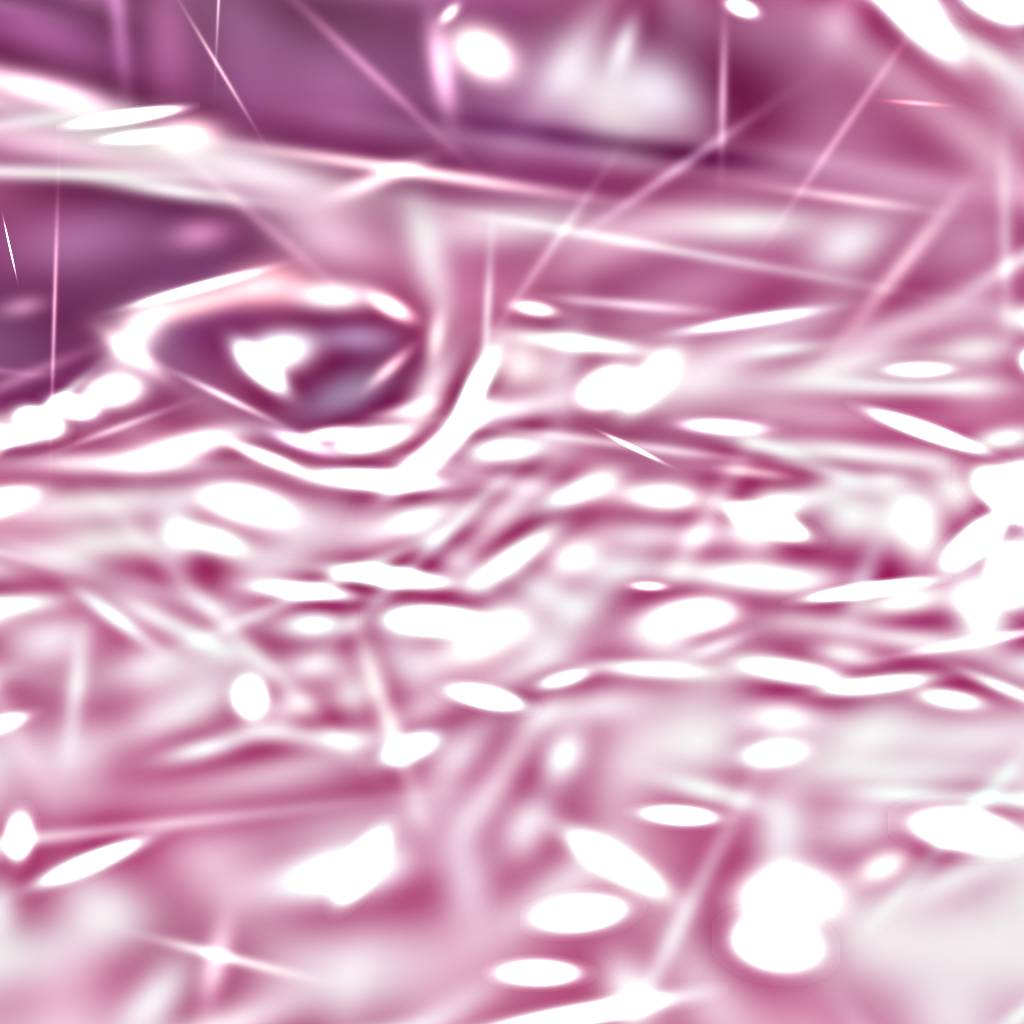} & 
        \includegraphics[width=0.1\linewidth]{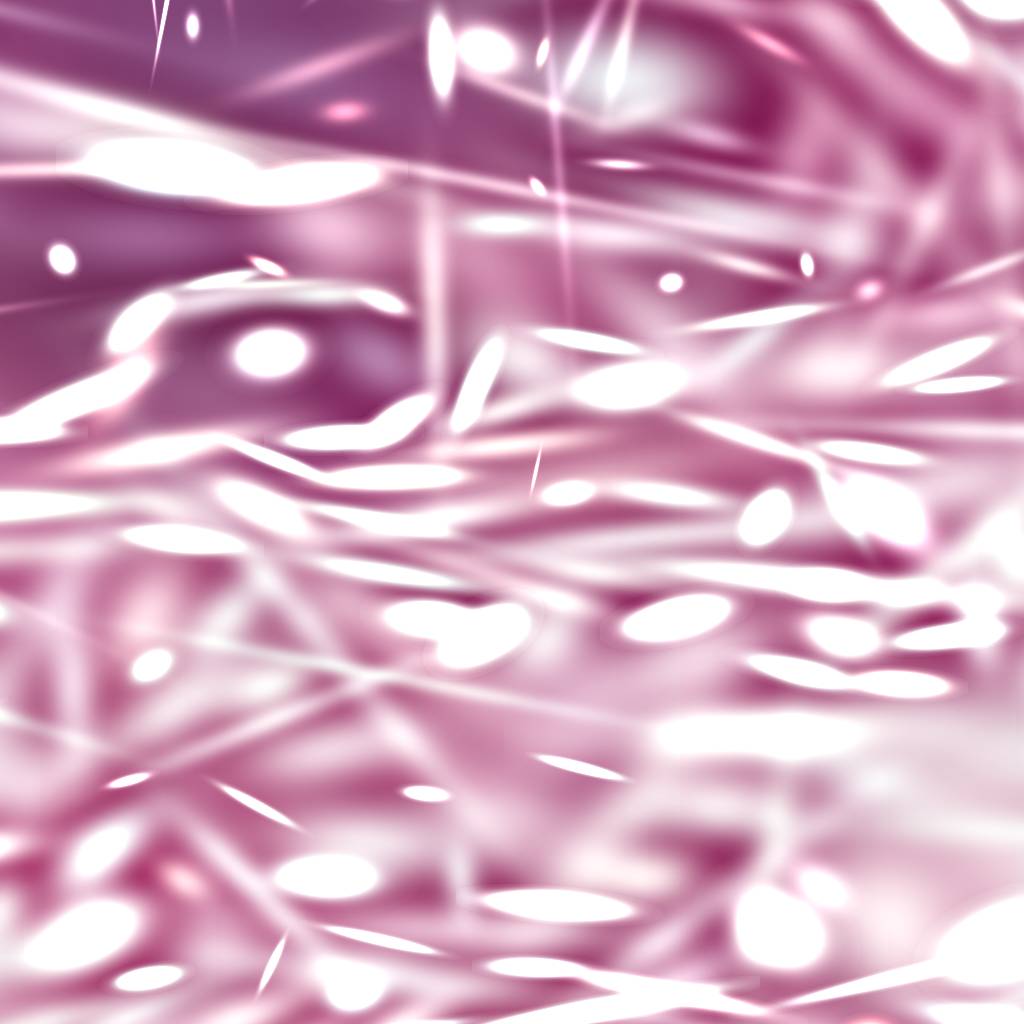} &
        \includegraphics[width=0.1\linewidth]{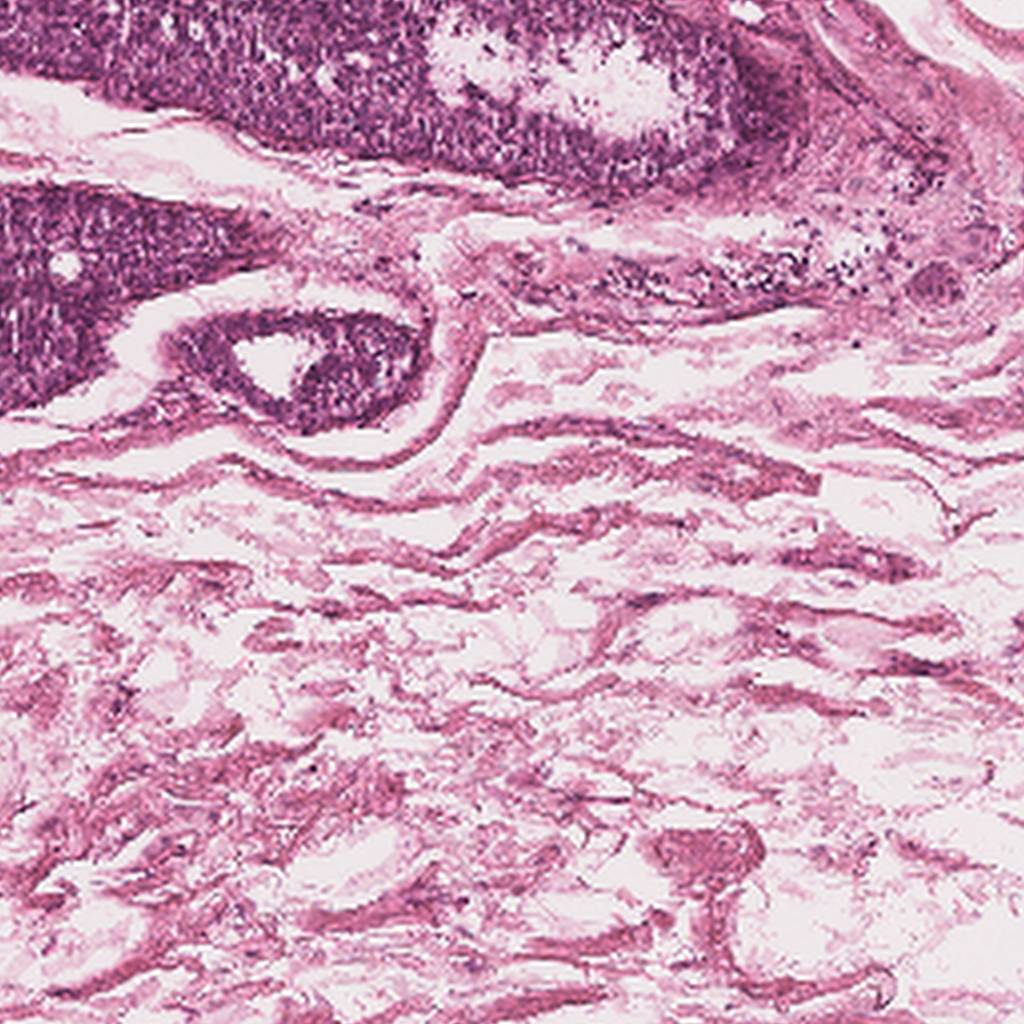} & 
        \includegraphics[width=0.1\linewidth]{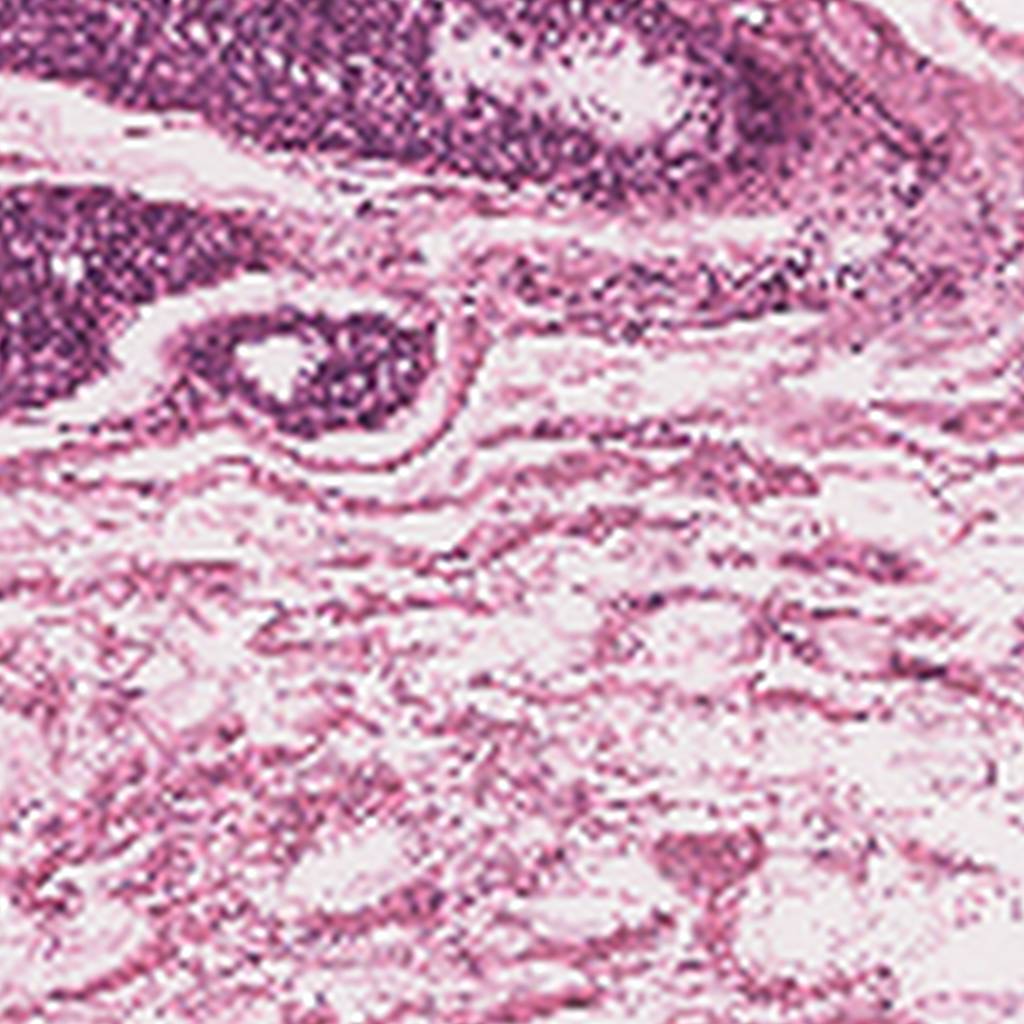} \\

    \end{tabular}}
    \caption{
    \textbf{Super-Resolution on \dataname{}: qualitative results.}
    We show $1024 \times 1024$ crops from full-resolution images and, for each family, we consider the same method as in the Human Evaluation. 
    }
    \label{fig:qualitatives_superresolution}
\end{figure*}

        \paragraph{Human Evaluation.}
            As PSNR is a known imperfect proxy for perceptual quality \cite{blau2018perception}, we complement the quantitative results obtained in the super-resolution experiments with a human evaluation study, conducted with \Nparticipants{} participants. 
            We excluded \textit{Andromeda} from the study after observing that its crops did not provide sufficient context for participants to reliably assess the quality of the upsampled image.
            For each family, we selected as representative the method yielding the highest average PSNR, with the exception of Hybrid Representations, where we selected MetricGrids (T=24) for both the $\times4$ and $\times8$ settings. 
            Although I-NGP attains a marginally higher PSNR at $\times4$, we observed that its super-resolved outputs are qualitatively poor, as high-resolution queries within the same low-resolution cell collapse to near-identical values, producing blocky, piecewise-constant reconstructions that lack genuine high-frequency detail. 
            MetricGrids, by contrast, recovers more faithful structure.
            Additional analysis can be found in \cref{sec:ingp_exclusion} of the Supplementary.
            \cref{fig:elo_boxplot} reports the Elo scores, bootstrapped 2000 times, at $\times4$ (Top) and $\times8$ (Bottom).
            The results corroborate the quantitative findings of \cref{tab:superres}, but highlight nuances that PSNR alone does not capture. 
            At $\times4$, Classic interpolations are preferred overall, followed closely by Hybrid Representations, with Implicit and Explicit methods ranking substantially lower. 
            Among the two, the Implicit family is clearly preferred over the Explicit one.
            The tight bootstrap confidence intervals confirm that this ranking is statistically meaningful.
            At $\times8$, Hybrid representations close the gap to classical algorithms, reaching essentially equivalent perceptual quality at the top of the overall ranking.
            At the bottom of the ranking, Explicit methods gain ground on Implicit ones. 
            However, both remain far below the Hybrid and Classical methods.
            While the reconstructions provided by Implicit and Explicit methods may be competitive in terms of PSNR, they are perceptually penalised for smooth outputs that suppress fine textures rather than plausibly hallucinate them.
            The per-image breakdown shows that the family-level ranking is broadly consistent across domains, with Hybrid and Classical methods leading in nearly every scene, indicating that the findings are not driven by any single image type.
            The one notable outlier is \textit{Venice} at $\times4$, where Classic methods dominate by an unusually large margin, likely reflecting the sensitivity of fine architectural detail in the panoramic image to the smoothing artifacts provided by LRs at moderate upsampling factors.

    \subsection{Qualitative Analysis}
        In \cref{fig:qualitatives_compression} and \cref{fig:qualitatives_superresolution}, we show qualitative results for the Compression and Super-Resolution tasks. 
        These figures highlight characteristic failure modes for each family. 
        Implicit representations produce blurry reconstructions, a direct consequence of the spectral bias of MLP-based approximators \cite{tancik2020fourfeat}. 
        This is most severe in texture-rich domains. 
        Indeed, in the Breast Cancer histopathology crops, fine tissue structures are smoothed into an unintelligible wash at both upsampling factors, rendering Implicit methods unsuitable for scientific imaging applications where fine-grained detail is diagnostically relevant.
        Explicit Gaussian-based representations, while achieving good results in the compression task, exhibit the most visually striking failure modes in super-resolution.
        At $\times8$, ringing and ghosting artifacts are clearly visible around high-contrast edges, as seen in the Venice pillar crop, and individual Gaussian primitives become perceptually distinct in the Breast Cancer crop, producing a characteristic blob-like texture that bears no structural resemblance to the underlying tissue.
        Hybrid representations are the most perceptually robust family overall, providing reconstructions that remain close to the ground truth across all tested images and in both compression and super-resolution experiments.
        Classical methods, while producing no artifacts, introduce the characteristic smoothing of interpolation-based methods, more clearly visible as a slight loss of sharpness in fine structural detail at $\times8$ relative to the ground-truth.

    \section{Discussion}
    We introduced \dataname{}, a curated benchmark of gigapixel images spanning ten scenes across diverse domains, and used it to systematically compare Learnable Representations on both Compression and Super-Resolution.
    This lets us revisit the two questions that motivated our study: \emph{whether and how} LRs can be effectively deployed at gigapixel scale, and \emph{which} family is most promising.
    On the first, LRs can indeed be brought to the gigapixel regime, but only with dedicated adaptation, such as tiling and gradient accumulation, and at a computational cost that becomes a decisive selection criterion rather than a secondary concern.
    On the second, no single family dominates, and the right choice depends on the task, the target compression ratio, and the application constraints.
    For Compression, Hybrid Representations are the clear recommendation at moderate ratios, with I-NGP offering a particularly favourable quality vs. compression time trade-off thanks to its CUDA-optimized implementation.
    When extreme compression is required, Implicit representations and SmartSplat are viable options. 
    The latter remains competitive across the full compression spectrum and yields the fastest inference among the implementations we benchmarked, making it the natural choice for latency-sensitive downstream tasks, provided fitting cost is not a bottleneck.
    For super-resolution, the results are more nuanced.
    Classical methods retain a robust perceptual advantage at $\times4$, while at $\times8$ Hybrid representations match them in human preference.
    Finally, the divergence between PSNR and human-preference rankings highlights that quantitative metrics are necessary but not sufficient in perceptually demanding regimes.
    
    \dataname{} provides the infrastructure to conduct both forms of evaluation at gigapixel scale, and we encourage future work to treat human assessment as a primary rather than supplementary signal.
    Although our findings reveal properties of the considered LR families in the giga-pixel regime, they stem from a ten-scene benchmark and thus should not be interpreted as broad, general conclusions.

    \section*{Acknowledgements}

This work was partially supported by DAIS – Ca' Foscari University of Venice within the IRIDE-B program. We are grateful to the entrepreneur and gigapixel photographer \href{https://www.jeffrey-martin.com/}{Jeffrey Martin}, for kindly providing the
stunning Venice panorama used in this work.
    {
        \small
        \bibliographystyle{ieeenat_fullname}
        \bibliography{main}
    }

    \clearpage
\maketitlesupplementary
\setcounter{section}{0}
\renewcommand{\thesection}{\Alph{section}}

In this document, we propose complementary results to the main paper.

\section{Limitations and Intended Scope} 
    \dataname{} comprises ten scenes, one or two per domain, and evaluates eight methods in sixteen configurations. 
    Every result in this paper should be read under that scope. 
    Our claims are comparative statements about the specific methods and configurations we evaluate, measured on these ten images, and they characterise the trade-offs a practitioner faces when choosing among them at gigapixel resolution. 
    They are not estimates of the expected behaviour of a representation family over the population of gigapixel images, and the averages we report are correspondingly sensitive to individual scenes: with one or two images per domain, a single scene can move a family average by more than the margins that separate families. 
    The benchmark therefore does not support domain-level conclusions, and we draw none. 
    Nor does it support claims that a family is intrinsically better or worse than another, as opposed to the evaluated instances of that family, since intra-family variance is in several cases larger than the inter-family gap. 
    Whether the trends we observe persist over a broader distribution of gigapixel imagery is an open question. Within this scope, however, we intend the benchmark to serve as an empirical guide and a starting point for practitioners choosing a representation for their own gigapixel workloads, to be refined as the evidence base grows.

\section{Provenance and Attribution of the \dataname{} images}
    
    \paragraph{Venice} is a gigapixel panorama ($64{,}000 \times 20{,}000$ px) shot by the photographer Jeffrey Martin (\url{https://www.jeffrey-martin.com/}). 
    Upon our request, the author provided the original full-resolution image and granted permission for its inclusion in \dataname{}. 
    We redistribute it under the CC BY-NC 4.0 license, with attribution to the author.
    
    \paragraph{Statuario and Calacatta Viola} are industrial marble scans, with resolutions of ($34015 \times 33750$) and ($22677 \times 50040$), respectively. 
    These images have been given to us by a professional who prefers to remain anonymous. 
    Therefore, we release the images under the CC0 license.
    
    \paragraph{Breast Cancer} originates from BRACS (BReAst Carcinoma Subtyping)~\cite{brancati2022bracs}, a public dataset of hematoxylin-and-eosin–stained breast tissue whole-slide images (WSIs) annotated for the subtyping of breast lesions into seven categories spanning the benign-to-malignant spectrum. 
    The slides are digitized at $40\times$ magnification and stored as multi-resolution Aperio SVS files, in which level~0 contains the full-resolution scan and coarser pyramid levels are retained for efficient access. 
    We obtain our image ($29{,}474 \times 36{,}430$ px) from a single BRACS WSI. 
    Using \texttt{pyvips}, we stream the level-0 plane, apply an isotropic Lanczos-3 downsampling that brings the total pixel count to ${\sim}1$\, gigapixel while preserving the native aspect ratio, flatten any alpha channel against a white background, and write a tiled, pyramidal, LZW-compressed BigTIFF. BRACS is released under the CC BY-NC 4.0 license~\cite{brancati2022bracs}. 
    We redistribute our derived image under the same license, with attribution to the originating work.
    
    \paragraph{Kidney} comes from the Kidney Pathology Image Segmentation (KPIs) challenge~\cite{deng2025kpis2024challengeadvancing}, held at MICCAI~2024, which targets pixel-level segmentation of glomeruli across normal and chronic kidney disease (CKD) models. 
    Its data consists of Periodic acid--Schiff (PAS) stained whole-slide images (WSIs) of whole kidney sections from preclinical rodent models, distributed as multi-resolution TIFF files in which level~0 holds the full-resolution scan. 
    We obtain our image as a $55{,}967 \times 20{,}800$ px spatial crop from the level-0 plane of a single WSI, without resampling. 
    The KPIs data is openly available on \href{https://www.google.com/url?q=https%3A%2F%2Fwww.synapse.org%2Fkpis24&sa=D&sntz=1&usg=AOvVaw0ISUz2RQwIKb-8y6XK9bVF}{Synapse} as public content with no access or redistribution restrictions, and is therefore covered by the CC BY 4.0 license under Synapse's Terms of Service. We redistribute our derived crop under the same license, citing the originating works~\cite{deng2025kpis2024challengeadvancing, tang2026holohisto} as requested by the data providers.
    
    \paragraph{Munich} is an aerial orthophoto crop of the city of Munich, taken from the gigapixel digital orthophoto (DOP) hosted on Wikimedia Commons and produced by the Bayerische Vermessungsverwaltung~\cite{bvv} (Bavarian state surveying authority). 
    Digital orthophotos are fully rectified, true-to-scale aerial images derived from the periodic Bavaria aerial survey. 
    We obtain our image as a $32{,}765 \times 37{,}454$ px spatial crop of the full-resolution orthophoto, without resampling. The source is released under the CC BY 4.0 license, and we redistribute our derived crop 
    % with more restrictive license CC BY-NC-SA 4.0, 
    with attribution to the Bayerische Vermessungsverwaltung~\cite{bvv}.
    
    \paragraph{Andromeda} is a crop of the sharpest and largest view of the Andromeda galaxy (M31) ever taken by the NASA/ESA Hubble Space Telescope~\cite{esahubble2026}, released through the ESA/Hubble outreach archive. 
    The released mosaic is itself a cropped section of the galaxy's disc, spanning $ 22{,}230\times  69{,}536$ px ($\sim$1.5 gigapixels). 
    We obtain our image as a $ 22{,}230\times 55{,}000$ px spatial crop, without resampling, yielding ${\sim}1.14$\,gigapixels. 
    The source is released under the CC BY 4.0 license; we redistribute our derived crop as part of \dataname{} 
    % under the more restrictive CC BY-NC-SA 4.0 license, 
    retaining the full original credit: NASA, ESA, J.~Dalcanton, B.~F.~Williams, L.~C.~Johnson (University of Washington), the PHAT team, and R.~Gendler.
    
    \paragraph{Italy} is a cloud-free composite of Italy that we generate from Copernicus Sentinel-2 Level-2A (atmospherically corrected) data~\cite{copernicus_browser2026}, retrieved through the Copernicus Data Space Ecosystem Process API. 
    Over the acquisition window June--September 2025, a least-cloud mosaic is composited server-side, with the Scene Classification Layer masking clouds, shadows, and cirrus, and the result is returned as a georeferenced (EPSG:4326), lossless DEFLATE-compressed GeoTIFF. 
    The full composite spans $35{,}040 \times 30{,}654$ px ($\sim$1.07 gigapixels) at an effective ground resolution of ${\sim}28$\,m/px. 
    Copernicus Sentinel data is provided under the European Union's free, full, and open data policy~\cite{copernicus_browser2026}, which permits redistribution and the creation of derivative works. 
    We redistribute our composite as part of \dataname{} 
    % under CC BY-NC-SA 4.0, 
    with the required acknowledgement ``Contains modified Copernicus Sentinel data 2025''.
    
    \paragraph{Lynxsdesign} is a gigapixel rendering of the ``Lynxsdesign'' demo scene from the official Blender demo-files collection~\cite{lynxsdesign2026}. We render the scene with the Cycles path-tracing engine, producing a $24588 \times 43712$ px image. 
    Per-tile .tif outputs are then stitched losslessly with \texttt{pyvips} into a tiled, pyramidal, DEFLATE-compressed BigTIFF. 
    The source scene is released under the CC BY license; we redistribute our derived rendering as part of \dataname{} 
    % under the more restrictive CC BY-NC-SA 4.0 license, 
    with attribution to the original author (Lynxsdesign).
    
    \paragraph{House Of Chores} is a gigapixel rendering of the ``House of Chores'' demo scene, formerly distributed in the official Blender demo-files collection (version~4.3)~\cite{houseofchores2024} (the file has since been removed from the collection). 
    We render the scene with its native EEVEE engine, tiling the virtual $  24{,}588\times43{,}712$ px frame into a $4{\times}4$ grid via camera-shift offsets, with each tile rendered at a $10\%$ overlap margin to suppress seams from EEVEE's view-dependent screen-space effects. 
    The overlap is cropped from each tile, and the $16$ tiles are stitched losslessly with \texttt{pyvips} into a tiled, pyramidal, LZW-compressed BigTIFF. 
    The source scene was released under the CC BY 4.0 license. 
    We redistribute our derived rendering as part of \dataname{} 
    % under the more restrictive CC BY-NC-SA 4.0 license, 
    with attribution to the original author (Blender Studio).
    
\section{Quantitative Characterization of Image Diversity}

    The domain labels described in \cref{tab:gigabench} alone are insufficient to attest to the \dataname{}'s diversity. 
    Thus, we complement this work with a set of low-level complexity statistics for every image. 
    We consider four complementary families:
    
    \begin{itemize}%
        \item \textbf{Spectral statistics.} 
        Natural images exhibit power-law spectra $P(f)\!\propto\!1/f^{\beta}$ with $\beta\!\approx\!2$~\cite{field1987relations,ruderman1994scaling}. 
        The slope $\beta$ measures how quickly energy decays with spatial frequency: a low $\beta$ means the image keeps substantial energy at fine scales (rich detail), a high $\beta$ means it is dominated by smooth, large-scale structure. 
        We fit $\mathbf{\beta}$ by least squares on the log--log radially averaged power spectral density (PSD) of the Hann-windowed, mean-removed full image, over log-spaced frequency bins in $[0.02, 0.80]\,f_{\mathrm{Ny}}$. 
        The fit $\mathbf{R^2}$ tells whether a power law is a good model in the first place. 
        Since a single slope cannot describe spectra that deviate from a power law, we also report $\mathbf{E_{\mathrm{HF}}}$, the fraction of (non-DC) spectral energy above $0.25\,f_{\mathrm{Ny}}$.
      
        \item \textbf{Edge statistics.} We report two measures of how many edges an image contains. 
        The mean Scharr gradient magnitude $\overline{\lVert\nabla I\rVert}$ needs no threshold and captures all local intensity variation, but it does not distinguish clean contours from noise. 
        The Canny edge-pixel density $\rho_{\mathrm{edge}}$ (thresholds $100/200$ on 8-bit luma) does the opposite: it depends on the chosen thresholds, but its non-maximum suppression and hysteresis keep only salient, connected contours.
        
        \item \textbf{Entropy statistics.} The Shannon entropy $H$ of the 256-bin luminance histogram gives the bits/pixel needed to encode each pixel on its own, ignoring \emph{where} values occur. 
        We pair it with the bits/pixel of a fixed lossless codec (JPEG~XL, effort~7) on the RGB image, which also exploits the similarity between nearby pixels and thus estimates the image's entropy \emph{rate} (codec-dependent, but comparable across images since the codec is fixed).
    
        \item \textbf{Texture statistics.} From the symmetric grey-level co-occurrence matrix (GLCM; 32 levels, distance 1, averaged over 4 orientations) we report \textbf{contrast}, $\sum_{i,j} p(i,j)(i-j)^2$, and \textbf{homogeneity}, $\sum_{i,j} p(i,j)/(1+(i-j)^2)$. 
        Both look at pairs of neighbouring pixels: contrast grows when neighbours differ (rough texture), homogeneity when they agree (uniform regions). Being local pair statistics, they complement the global, orientation-averaged view given by the spectrum.
    \end{itemize}
    
    \begin{table*}[t]
    \centering
    \small
    \setlength{\tabcolsep}{5pt}
\begin{tabular}{lccccccccc}
    \toprule
    & \multicolumn{3}{c}{Spectral} & \multicolumn{2}{c}{Edge}
    & \multicolumn{2}{c}{Entropy} & \multicolumn{2}{c}{Texture} \\
    \cmidrule(lr){2-4}\cmidrule(lr){5-6}\cmidrule(lr){7-8}\cmidrule(lr){9-10}
     & $\beta$ & $R^2$ & $E_{\mathrm{HF}}$
    & $\overline{\lVert\nabla I\rVert}$ & $\rho_{\mathrm{edge}}$
    & $H$ [bits/px] & JXL [bpp]
    & Contrast & Homog. \\
    \midrule
    Venice          & 2.03 & 1.00 & 0.085 & 0.022 & 0.059 & 7.51 & 7.25  & 2.34  & 0.73 \\
    Statuario       & 2.66 & 0.99 & 0.007 & 0.002 & 0.001 & 3.78 & 2.46  & 0.10  & 0.96 \\
    Calacatta Viola & 2.69 & 0.99 & 0.007 & 0.006 & 0.008 & 6.15 & 6.02  & 0.27  & 0.92 \\
    Breast Cancer   & 2.30 & 1.00 & 0.082 & 0.019 & 0.071 & 5.34 & 9.28  & 2.62  & 0.81 \\
    Kidney$^{\ddagger}$ & 4.28 & 0.98 & 0.003 & 0.015 & 0.010 & 7.07 & 5.74 & 0.51 & 0.82 \\
    Munich          & 2.54 & 0.99 & 0.091 & 0.043 & 0.144 & 7.61 & 9.47  & 5.51  & 0.60 \\
    Andromeda$^{\dagger}$ & 1.73 & 0.74 & 0.268 & 0.087 & 0.240 & 7.75 & 15.28 & 13.93 & 0.42 \\
    Italy           & 1.86 & 1.00 & 0.104 & 0.021 & 0.075 & 6.64 & 9.87  & 4.11  & 0.75 \\
    Lynxsdesign     & 2.24 & 0.99 & 0.022 & 0.025 & 0.057 & 7.49 & 10.82 & 2.95  & 0.56 \\
    House of Chores & 3.15 & 1.00 & 0.010 & 0.008 & 0.008 & 7.34 & 12.00 & 0.90  & 0.84 \\
    \midrule
    min / max ratio & 2.5$\times$ & -- & 85$\times$ & 35$\times$ & 235$\times$
    & 2.1$\times$ & 6.2$\times$ & 144$\times$ & 2.3$\times$ \\
    \bottomrule
    \end{tabular}
    \caption{\textbf{Complexity statistics of the GIGABENCH images}. $\beta$, $R^2$: power-law fit of the radial PSD; $E_{\mathrm{HF}}$: fraction of spectral energy above $0.25\,f_{\mathrm{Ny}}$; $H$: Shannon entropy of the luminance histogram; $\overline{\lVert\nabla I\rVert}$: mean Scharr gradient; $\rho_{\mathrm{edge}}$: Canny edge density; Contrast, Homog.: GLCM statistics; JXL: lossless JPEG~XL rate (RGB). $^{\dagger}$Non--power-law spectrum ($R^2{=}0.74$), see \cref{fig:psd}. $^{\ddagger}$Band-limited by the scanner optics: refitting below/above $0.2\,f_{\mathrm{Ny}}$ yields $\beta{=}3.47/5.68$, so the single slope overstates the decay.}
    \label{tab:diversity_stats}
\end{table*}

    \begin{figure}[t]
    \centering
    \includegraphics[width=\linewidth]{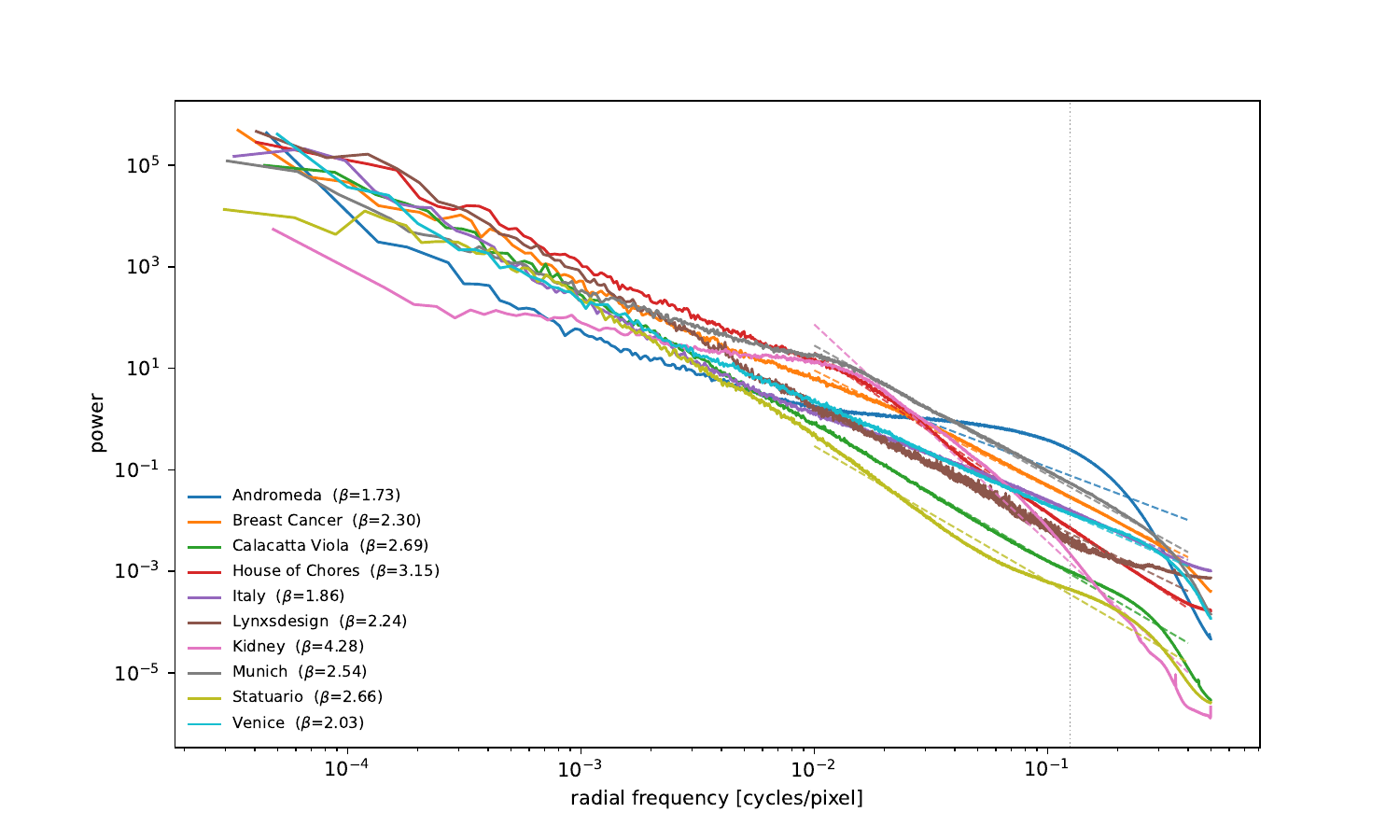}
    \caption{\textbf{Radially averaged power spectra} of the ten images (log--log; Hann-windowed, mean-removed full-resolution luminance).
    Dashed lines: power-law fits over $[0.02,0.80]\,f_{\mathrm{Ny}}$;     dotted vertical line: the $E_{\mathrm{HF}}$ cutoff $0.25\,f_{\mathrm{Ny}}$. The spectra differ not only in slope but in shape: most images follow a power law closely, Andromeda departs from it with a broad excess of mid-to-high-frequency energy, and Kidney rolls off sharply toward Nyquist, consistent with a band-limited acquisition.}
    \label{fig:psd}
\end{figure}

    \cref{tab:diversity_stats} reports the statistics and \cref{fig:psd} displays the radially averaged spectra.    
    The images span a wide range on every axis: $\beta$ covers $1.73$--$4.28$, straddling the natural-image regime $\beta\!\approx\!2$ (Italy, Venice) from both sides; $E_{\mathrm{HF}}$ spans two orders of magnitude, entropy ranges from $3.78$ to $7.75$ bits/px, and edge density and GLCM contrast vary by factors of ${\sim}235$ and ${\sim}144$, respectively. 
    Importantly, these axes are not redundant: the per-image orderings disagree across statistics. 
    House of Chores pairs high entropy ($7.34$ bits/px) with near-minimal edge density (its smooth synthetic gradients populate the histogram without creating contours).
    Andromeda combines the shallowest spectrum and the largest high-frequency energy with only moderate homogeneity. 
    Statuario is extremal on almost every measure, providing a near-degenerate stress case. 
    The spectra of \cref{fig:psd} additionally differ in \emph{shape}, not only in slope, reflecting genuinely different acquisition pipelines (photographic, scanned, composited, rendered). 
    We note the scope of this evidence: it establishes that the ten images occupy widely separated, partially decorrelated points in low-level statistics space, complementing the semantic domain labels.
    However, with ten images, it does not constitute a claim of exhaustive coverage of the gigapixel regime.

\section{Structural Similarity Analysis}
\label{sec:ssim}

\begin{figure}
    \centering
    \resizebox{\linewidth}{!}{
        \includegraphics[width=\linewidth]{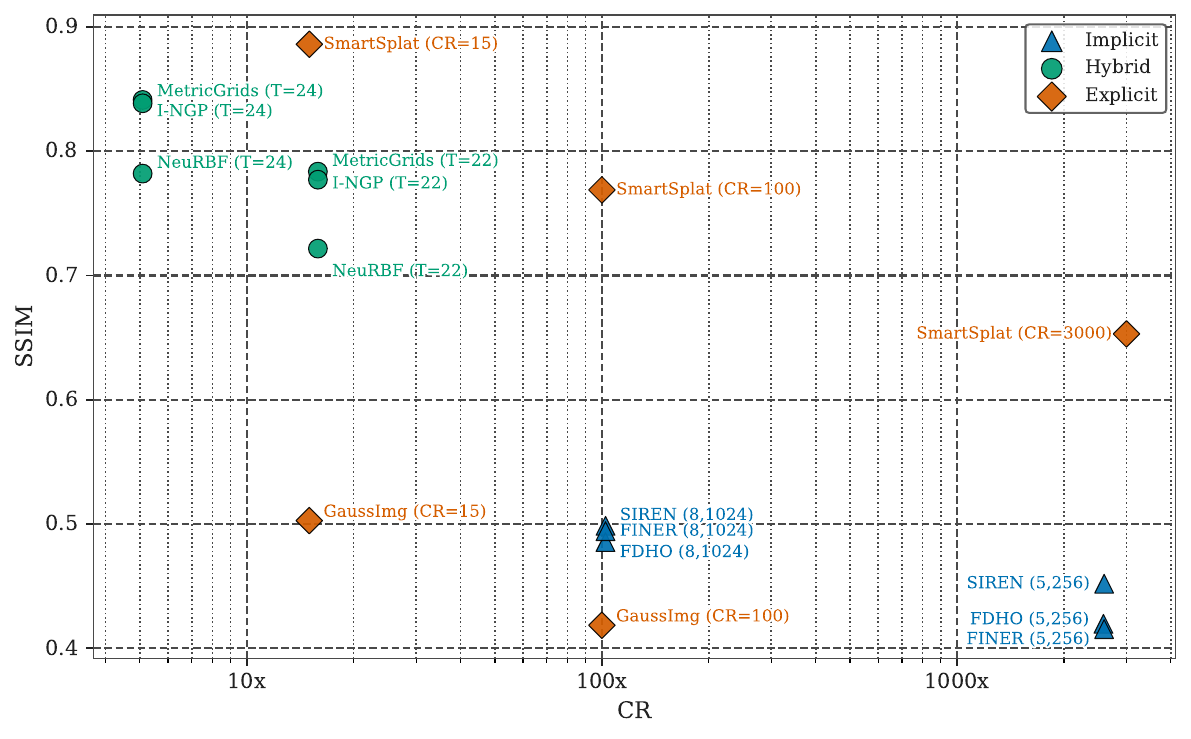}
        }
    \caption{
        \textbf{SSIM vs. Compression Ratio }in image compression.
        }
    \label{fig:fitting_ssim_cr}
\end{figure}

\begin{figure*}[ht]
    \centering
    \resizebox{\linewidth}{!}{
        \includegraphics[width=0.49\linewidth]{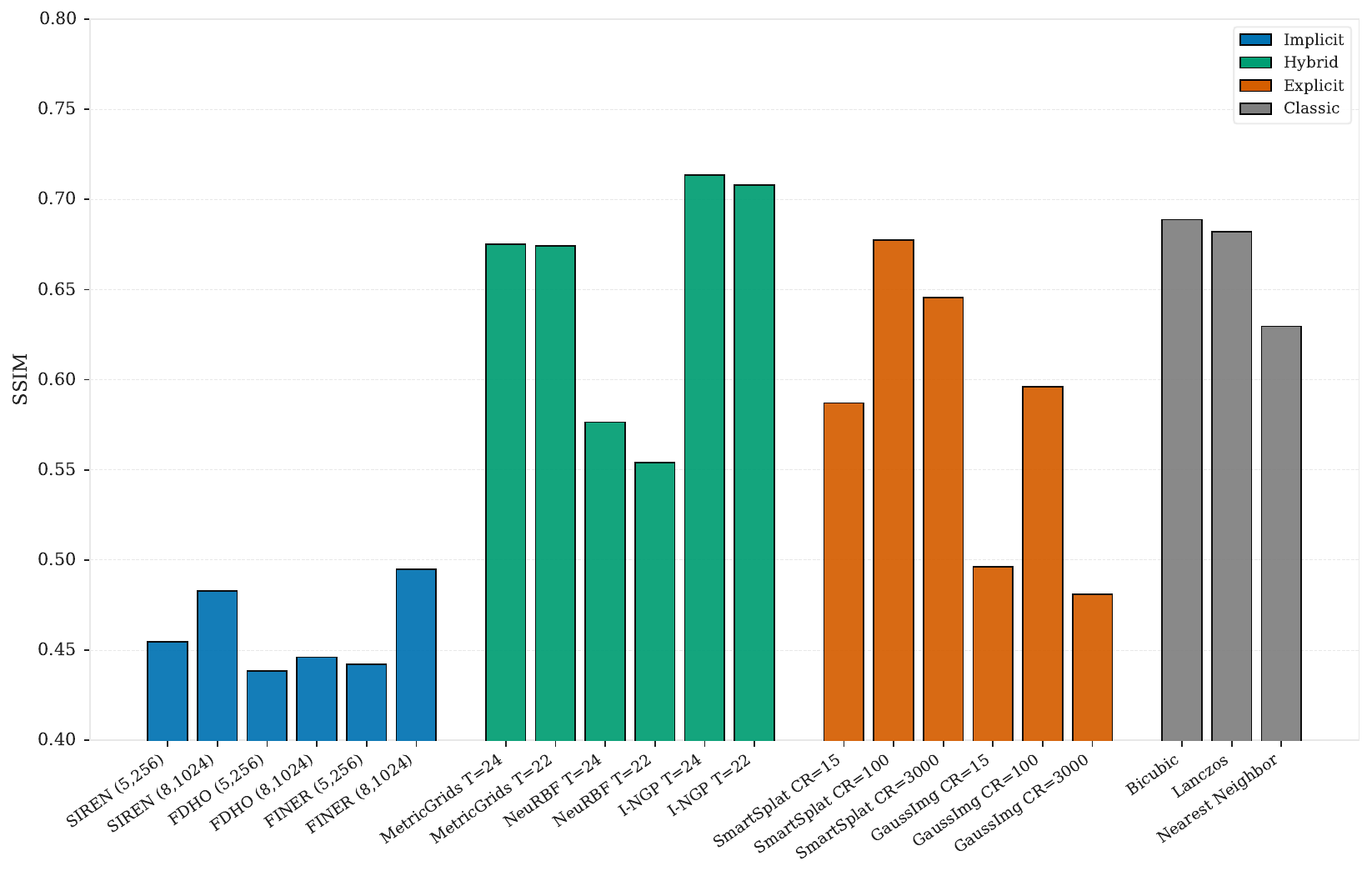}
        \includegraphics[width=0.49\linewidth]{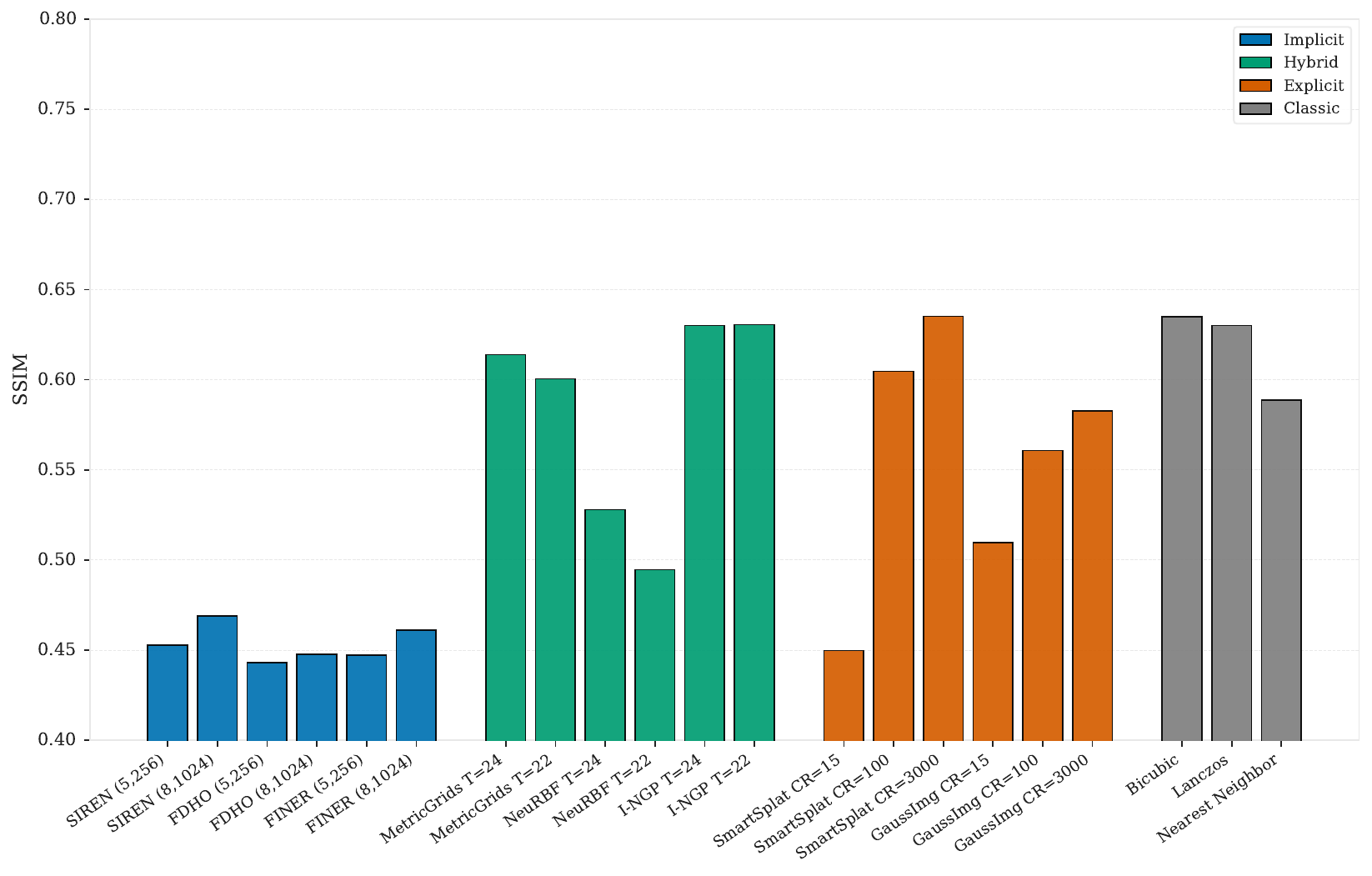}
        }
    \caption{
        \textbf{Super-Resolution SSIM} performance on \dataname{} (left: SR$\times4$, right: SR$\times8$).
        }
    \label{fig:superres_ssim}
\end{figure*}

    We complement the PSNR-based evaluation of the main paper with a parallel analysis under SSIM~\cite{wang2004ssim}.
    \cref{tab:ssim_fitting} and \cref{tab:ssim_sr} report per-image SSIM for every method on the Compression and $\times 4$/$\times 8$ Super-Resolution tasks, mirroring the layout of \cref{tab:superres} of the main paper.
    \cref{fig:fitting_ssim_cr} is a drop-in replica of the PSNR--CR plot of the main paper with PSNR replaced by SSIM, and \cref{fig:superres_ssim} reports the average SSIM for SR$\times 4$ and SR$\times 8$.

    \begin{table}[ht]
    \centering
    \resizebox{\linewidth}{!}{
       \begin{tabular}{ccc cccccccccc c}
    
            \toprule
            &
            &
            &
            \rothead{\imageone}
            & \rothead{\imagetwo}
            & \rothead{\imagethree}
            & \rothead{\imagefour}
            & \rothead{\imagefive}
            & \rothead{\imagesix}
            & \rothead{\imageseven}
            & \rothead{\imageeight}
            & \rothead{\imagenine}
            & \rothead{\imageten}
            & \textbf{Avg} \\

            \midrule
            \multirow{26}{*}{\rotatebox{90}{Compression}}
		& \multirow{8}{*}{\rotatebox{90}{\textbf{Implicit}}}
			& {\rotatebox{0}{\shortstack{SIREN (5,256)}}} & 0.450 & 0.926 & 0.823 & 0.645 & 0.353 & 0.143 & 0.048 & 0.560 & 0.218 & 0.354 & 0.452 \\
			\cmidrule(lr){3-14}

			& & {\rotatebox{0}{\shortstack{FINER (5,256)}}} & 0.181 & 0.927 & 0.825 & 0.646 & 0.301 & 0.114 & 0.048 & 0.558 & 0.220 & 0.335 & 0.415 \\
			\cmidrule(lr){3-14}

			& & {\rotatebox{0}{\shortstack{FDHO (5,256)}}} & 0.253 & 0.923 & 0.838 & 0.643 & 0.296 & 0.099 & 0.047 & 0.558 & 0.207 & 0.333 & 0.420 \\
			\cmidrule(lr){3-14}

			& & {\rotatebox{0}{\shortstack{SIREN (8,1024)}}} & 0.492 & \cellcolor{implicit!80}0.927 & 0.839 & 0.649 & 0.593 & \cellcolor{implicit!80}0.227 & 0.049 & 0.561 & \cellcolor{implicit!80}0.230 & \cellcolor{implicit!80}0.419 & \cellcolor{implicit!80}0.499 \\
			\cmidrule(lr){3-14}

			& & {\rotatebox{0}{\shortstack{FINER (8,1024)}}} & \cellcolor{implicit!80}0.495 & 0.926 & \cellcolor{implicit!80}0.843 & \cellcolor{implicit!80}0.657 & \cellcolor{implicit!80}0.608 & 0.190 & \cellcolor{implicit!80}0.051 & \cellcolor{implicit!80}0.565 & 0.222 & 0.385 & 0.494 \\
			\cmidrule(lr){3-14}

			& & {\rotatebox{0}{\shortstack{FDHO (8,1024)}}} & 0.491 & 0.925 & 0.834 & 0.646 & 0.596 & 0.160 & 0.049 & 0.561 & 0.218 & 0.375 & 0.486 \\
			\cmidrule(lr){2-14}

		& \multirow{8}{*}{\rotatebox{90}{\textbf{Hybrid}}}
			& {\rotatebox{0}{\shortstack{I-NGP (T=22)}}} & 0.782 & 0.983 & 0.964 & 0.876 & 0.919 & 0.650 & 0.438 & 0.793 & 0.617 & 0.749 & 0.777 \\
			\cmidrule(lr){3-14}

			& & {\rotatebox{0}{\shortstack{I-NGP (T=24)}}} & 0.854 & 0.986 & \underline{0.971} & 0.916 & 0.960 & \cellcolor{hybrid!80}\underline{0.769} & 0.657 & \cellcolor{hybrid!80}\underline{0.852} & 0.667 & 0.753 & 0.839 \\
			\cmidrule(lr){3-14}

			& & {\rotatebox{0}{\shortstack{NeuRBF (T=22)}}} & 0.715 & 0.980 & 0.950 & 0.824 & 0.868 & 0.540 & 0.321 & 0.743 & 0.551 & 0.724 & 0.722 \\
			\cmidrule(lr){3-14}

			& & {\rotatebox{0}{\shortstack{NeuRBF (T=24)}}} & 0.762 & 0.986 & 0.965 & 0.876 & 0.919 & 0.657 & 0.510 & 0.801 & 0.608 & 0.736 & 0.782 \\
			\cmidrule(lr){3-14}

			& & {\rotatebox{0}{\shortstack{MetricGrids (T=22)}}} & 0.791 & \underline{0.988} & 0.968 & 0.878 & 0.934 & 0.642 & 0.464 & 0.797 & 0.624 & 0.750 & 0.784 \\
			\cmidrule(lr){3-14}

			& & {\rotatebox{0}{\shortstack{MetricGrids (T=24)}}} & \cellcolor{hybrid!80}\underline{0.858} & \cellcolor{hybrid!80}\textbf{0.991} & \cellcolor{hybrid!80}\textbf{0.980} & \cellcolor{hybrid!80}\underline{0.919} & \cellcolor{hybrid!80}\underline{0.966} & 0.762 & \cellcolor{hybrid!80}\underline{0.664} & 0.844 & \cellcolor{hybrid!80}\underline{0.671} & \cellcolor{hybrid!80}0.757 & \cellcolor{hybrid!80}\underline{0.841} \\
			\cmidrule(lr){2-14}

		& \multirow{8}{*}{\rotatebox{90}{\textbf{Explicit}}}
			& {\rotatebox{0}{\shortstack{GaussianImage (CR=3000)}}} & 0.035 & 0.068 & 0.065 & 0.065 & 0.061 & 0.051 & 0.018 & 0.067 & 0.038 & 0.074 & 0.054 \\
			\cmidrule(lr){3-14}

			& & {\rotatebox{0}{\shortstack{GaussianImage (CR=100)}}} & 0.294 & 0.791 & 0.629 & 0.649 & 0.335 & 0.366 & 0.288 & 0.226 & 0.288 & 0.320 & 0.419 \\
			\cmidrule(lr){3-14}

			& & {\rotatebox{0}{\shortstack{GaussianImage (CR=15)}}} & 0.378 & 0.811 & 0.643 & 0.652 & 0.354 & 0.503 & 0.494 & 0.426 & 0.376 & 0.392 & 0.503 \\
			\cmidrule(lr){3-14}

			& & {\rotatebox{0}{\shortstack{SmartSplat (CR=3000)}}} & 0.671 & 0.969 & 0.920 & 0.759 & 0.751 & 0.412 & 0.139 & 0.687 & 0.519 & 0.704 & 0.653 \\
			\cmidrule(lr){3-14}

			& & {\rotatebox{0}{\shortstack{SmartSplat (CR=100)}}} & 0.788 & 0.974 & 0.944 & 0.845 & 0.932 & 0.661 & 0.392 & 0.776 & 0.614 & \underline{0.763} & 0.769 \\
			\cmidrule(lr){3-14}

			& & {\rotatebox{0}{\shortstack{SmartSplat (CR=15)}}} & \cellcolor{explicit!80}\textbf{0.909} & \cellcolor{explicit!80}0.982 & \cellcolor{explicit!80}0.965 & \cellcolor{explicit!80}\textbf{0.926} & \cellcolor{explicit!80}\textbf{0.978} & \cellcolor{explicit!80}\textbf{0.882} & \cellcolor{explicit!80}\textbf{0.836} & \cellcolor{explicit!80}\textbf{0.887} & \cellcolor{explicit!80}\textbf{0.721} & \cellcolor{explicit!80}\textbf{0.776} & \cellcolor{explicit!80}\textbf{0.886} \\

			\bottomrule

        \end{tabular}
}

    \caption{
        \textbf{SSIM Performance on \dataname{}} for Compression.
        Best SSIMs within each family are highlighted by different colours; the best overall is in \textbf{bold}, and runner-ups are \underline{underlined}.
    }
    \label{tab:ssim_fitting}
\end{table}

    \paragraph{In Image Compression}, the SSIM evaluation broadly corroborates the PSNR-based conclusions of the main paper.
    At the category level, the ordering is largely preserved, with a single exception. 
    Indeed, the Hybrid and Explicit groups swap, so that the ranking changes from Hybrid--Explicit--Implicit under PSNR, to Explicit--Hybrid--Implicit under SSIM, while the Implicit group retain its position.
    Intra-category orderings are almost entirely preserved, with only two swaps, one within the Implicit and one within the Hybrid family, both confined to lower-ranked methods rather than the category leaders, and therefore of limited consequence.
    The dataset-induced spread is likewise preserved: the high-frequency images that prove difficult under PSNR (e.g., \textit{Italy}, \textit{Munich}, \textit{Andromeda}) remain the hardest under SSIM across all categories, confirming that the proposed benchmark stresses image regression methods consistently across fidelity measures.
    The category-level swap is attributable entirely to SmartSplat.
    Under SSIM, SmartSplat (CR$=15$) attains the best compression score on $8$ of the $10$ images and the highest dataset average ($0.886$), a margin that is not present in the PSNR table, where the category sits $2$--$3$\,dB below the Hybrid frontier.
    We attribute this to SmartSplat's training objective, which combines an $L_1$ term with an SSIM term and therefore directly optimizes the metric reported here.
    This is consistent with the broader observation that PSNR and SSIM rankings can disagree when a competing method is trained against the metric in question, and motivates reporting both when benchmarking image regression methods on the proposed dataset.

    \begin{table}[ht]
    \centering
    \resizebox{\linewidth}{!}{
       \begin{tabular}{ccc cccccccccc c}
    
            \toprule
            &
            &
            &
            \rothead{\imageone}
            & \rothead{\imagetwo}
            & \rothead{\imagethree}
            & \rothead{\imagefour}
            & \rothead{\imagefive}
            & \rothead{\imagesix}
            & \rothead{\imageseven}
            & \rothead{\imageeight}
            & \rothead{\imagenine}
            & \rothead{\imageten}
            & \textbf{Avg} \\

            \midrule

		\multirow{30}{*}{\rotatebox{90}{Upsample $\mathbf{\times4}$}}
		& \multirow{8}{*}{\rotatebox{90}{\textbf{Implicit}}}
			& {\rotatebox{0}{\shortstack{SIREN (5,256)}}} & 0.450 & 0.927 & 0.838 & 0.643 & 0.350 & 0.142 & 0.048 & 0.560 & 0.221 & 0.365 & 0.455 \\
			\cmidrule(lr){3-14}

			& & {\rotatebox{0}{\shortstack{FINER (5,256)}}} & 0.429 & 0.923 & 0.828 & 0.643 & 0.329 & 0.129 & 0.048 & 0.558 & 0.210 & 0.327 & 0.442 \\
			\cmidrule(lr){3-14}

			& & {\rotatebox{0}{\shortstack{FDHO (5,256)}}} & 0.427 & 0.923 & 0.824 & 0.642 & 0.311 & 0.114 & 0.047 & 0.558 & 0.208 & 0.330 & 0.438 \\
			\cmidrule(lr){3-14}

			& & {\rotatebox{0}{\shortstack{SIREN (8,1024)}}} & 0.471 & 0.927 & 0.840 & 0.648 & 0.490 & 0.198 & 0.049 & 0.561 & 0.227 & 0.417 & 0.483 \\
			\cmidrule(lr){3-14}

			& & {\rotatebox{0}{\shortstack{FINER (8,1024)}}} & \cellcolor{implicit!80}0.480 & \cellcolor{implicit!80}0.929 & \cellcolor{implicit!80}0.844 & \cellcolor{implicit!80}0.659 & \cellcolor{implicit!80}0.547 & \cellcolor{implicit!80}0.215 & \cellcolor{implicit!80}0.049 & \cellcolor{implicit!80}0.565 & \cellcolor{implicit!80}0.235 & \cellcolor{implicit!80}0.428 & \cellcolor{implicit!80}0.495 \\
			\cmidrule(lr){3-14}

			& & {\rotatebox{0}{\shortstack{FDHO (8,1024)}}} & 0.428 & 0.922 & 0.828 & 0.642 & 0.348 & 0.123 & 0.048 & 0.558 & 0.213 & 0.351 & 0.446 \\
			\cmidrule(lr){2-14}

		& \multirow{8}{*}{\rotatebox{90}{\textbf{Hybrid}}}
			& {\rotatebox{0}{\shortstack{I-NGP (T=22)}}} & \underline{0.711} & \underline{0.971} & \underline{0.938} & \underline{0.805} & 0.850 & \underline{0.540} & \underline{0.332} & \cellcolor{hybrid!80}\textbf{0.720} & \cellcolor{hybrid!80}\textbf{0.524} & \cellcolor{hybrid!80}\underline{0.689} & \underline{0.708} \\
			\cmidrule(lr){3-14}

			& & {\rotatebox{0}{\shortstack{I-NGP (T=24)}}} & \cellcolor{hybrid!80}\textbf{0.715} & \cellcolor{hybrid!80}\textbf{0.971} & \cellcolor{hybrid!80}\textbf{0.938} & \cellcolor{hybrid!80}\textbf{0.806} & 0.866 & \cellcolor{hybrid!80}\textbf{0.573} & \cellcolor{hybrid!80}\textbf{0.364} & \underline{0.719} & \underline{0.510} & 0.672 & \cellcolor{hybrid!80}\textbf{0.713} \\
			\cmidrule(lr){3-14}

			& & {\rotatebox{0}{\shortstack{NeuRBF (T=22)}}} & 0.530 & 0.947 & 0.888 & 0.688 & 0.599 & 0.245 & 0.081 & 0.609 & 0.365 & 0.588 & 0.554 \\
			\cmidrule(lr){3-14}

			& & {\rotatebox{0}{\shortstack{NeuRBF (T=24)}}} & 0.573 & 0.950 & 0.903 & 0.714 & 0.666 & 0.296 & 0.119 & 0.624 & 0.357 & 0.562 & 0.576 \\
			\cmidrule(lr){3-14}

			& & {\rotatebox{0}{\shortstack{MetricGrids (T=22)}}} & 0.630 & 0.968 & 0.928 & 0.790 & \cellcolor{hybrid!80}0.884 & 0.477 & 0.227 & 0.691 & 0.479 & 0.668 & 0.674 \\
			\cmidrule(lr){3-14}

			& & {\rotatebox{0}{\shortstack{MetricGrids (T=24)}}} & 0.640 & 0.969 & 0.929 & 0.797 & 0.872 & 0.496 & 0.256 & 0.699 & 0.453 & 0.641 & 0.675 \\
			\cmidrule(lr){2-14}

		& \multirow{8}{*}{\rotatebox{90}{\textbf{Explicit}}}
			& {\rotatebox{0}{\shortstack{GaussianImage (CR=3000)}}} & 0.452 & 0.905 & 0.793 & 0.693 & 0.543 & 0.285 & 0.099 & 0.176 & 0.372 & 0.491 & 0.481 \\
			\cmidrule(lr){3-14}

			& & {\rotatebox{0}{\shortstack{GaussianImage (CR=100)}}} & 0.590 & 0.913 & 0.791 & 0.706 & 0.753 & 0.464 & 0.197 & 0.472 & 0.460 & 0.613 & 0.596 \\
			\cmidrule(lr){3-14}

			& & {\rotatebox{0}{\shortstack{GaussianImage (CR=15)}}} & 0.379 & 0.914 & 0.812 & 0.722 & 0.586 & 0.345 & 0.198 & 0.264 & 0.354 & 0.388 & 0.496 \\
			\cmidrule(lr){3-14}

			& & {\rotatebox{0}{\shortstack{SmartSplat (CR=3000)}}} & \cellcolor{explicit!80}0.665 & \cellcolor{explicit!80}0.969 & \cellcolor{explicit!80}0.923 & 0.752 & 0.748 & 0.401 & 0.129 & 0.682 & \cellcolor{explicit!80}0.497 & 0.689 & 0.646 \\
			\cmidrule(lr){3-14}

			& & {\rotatebox{0}{\shortstack{SmartSplat (CR=100)}}} & 0.664 & 0.963 & 0.914 & \cellcolor{explicit!80}0.772 & \cellcolor{explicit!80}0.874 & \cellcolor{explicit!80}0.500 & \cellcolor{explicit!80}0.203 & \cellcolor{explicit!80}0.694 & 0.493 & \cellcolor{explicit!80}\textbf{0.696} & \cellcolor{explicit!80}0.677 \\
			\cmidrule(lr){3-14}

			& & {\rotatebox{0}{\shortstack{SmartSplat (CR=15)}}} & 0.512 & 0.937 & 0.857 & 0.729 & 0.600 & 0.404 & 0.195 & 0.672 & 0.393 & 0.572 & 0.587 \\
			\cmidrule(lr){2-14}

		& \multirow{4}{*}{\rotatebox{90}{\textbf{Classic}}}
			& {\rotatebox{0}{\shortstack{Nearest Neighbor}}} & 0.663 & 0.955 & 0.906 & 0.717 & 0.791 & 0.348 & 0.157 & 0.648 & 0.470 & 0.642 & 0.630 \\
			\cmidrule(lr){3-14}

			& & {\rotatebox{0}{\shortstack{Bicubic}}} & \cellcolor{classic!80}0.697 & \cellcolor{classic!80}0.967 & \cellcolor{classic!80}0.929 & \cellcolor{classic!80}0.763 & \cellcolor{classic!80}\textbf{0.891} & \cellcolor{classic!80}0.475 & \cellcolor{classic!80}0.286 & \cellcolor{classic!80}0.700 & \cellcolor{classic!80}0.507 & \cellcolor{classic!80}0.673 & \cellcolor{classic!80}0.689 \\
			\cmidrule(lr){3-14}

			& & {\rotatebox{0}{\shortstack{Lanczos}}} & 0.690 & 0.965 & 0.927 & 0.757 & \underline{0.889} & 0.462 & 0.281 & 0.694 & 0.498 & 0.658 & 0.682 \\
			\midrule

		\multirow{30}{*}{\rotatebox{90}{Upsample $\mathbf{\times8}$}}
		& \multirow{8}{*}{\rotatebox{90}{\textbf{Implicit}}}
			& {\rotatebox{0}{\shortstack{SIREN (5,256)}}} & 0.448 & 0.927 & 0.836 & 0.641 & 0.347 & 0.140 & 0.048 & 0.559 & 0.219 & 0.361 & 0.453 \\
			\cmidrule(lr){3-14}

			& & {\rotatebox{0}{\shortstack{FINER (5,256)}}} & 0.433 & 0.924 & 0.832 & 0.643 & 0.336 & 0.134 & 0.048 & 0.559 & 0.214 & 0.350 & 0.447 \\
			\cmidrule(lr){3-14}

			& & {\rotatebox{0}{\shortstack{FDHO (5,256)}}} & 0.426 & 0.925 & 0.830 & \cellcolor{implicit!80}0.644 & 0.316 & 0.127 & 0.048 & 0.559 & 0.212 & 0.346 & 0.443 \\
			\cmidrule(lr){3-14}

			& & {\rotatebox{0}{\shortstack{SIREN (8,1024)}}} & \cellcolor{implicit!80}0.455 & \cellcolor{implicit!80}0.928 & \cellcolor{implicit!80}0.841 & 0.641 & 0.441 & \cellcolor{implicit!80}0.155 & \cellcolor{implicit!80}0.048 & \cellcolor{implicit!80}0.561 & \cellcolor{implicit!80}0.225 & \cellcolor{implicit!80}0.395 & \cellcolor{implicit!80}0.469 \\
			\cmidrule(lr){3-14}

			& & {\rotatebox{0}{\shortstack{FINER (8,1024)}}} & 0.433 & 0.925 & 0.834 & 0.641 & \cellcolor{implicit!80}0.441 & 0.145 & 0.046 & 0.557 & 0.216 & 0.372 & 0.461 \\
			\cmidrule(lr){3-14}

			& & {\rotatebox{0}{\shortstack{FDHO (8,1024)}}} & 0.425 & 0.925 & 0.832 & 0.642 & 0.335 & 0.147 & 0.047 & 0.560 & 0.203 & 0.360 & 0.448 \\
			\cmidrule(lr){2-14}

		& \multirow{8}{*}{\rotatebox{90}{\textbf{Hybrid}}}
			& {\rotatebox{0}{\shortstack{I-NGP (T=22)}}} & \cellcolor{hybrid!80}0.639 & 0.962 & \cellcolor{hybrid!80}0.911 & \cellcolor{hybrid!80}\textbf{0.747} & 0.738 & \underline{0.409} & 0.126 & 0.665 & \cellcolor{hybrid!80}0.456 & \cellcolor{hybrid!80}0.651 & \cellcolor{hybrid!80}0.631 \\
			\cmidrule(lr){3-14}

			& & {\rotatebox{0}{\shortstack{I-NGP (T=24)}}} & 0.637 & 0.962 & 0.910 & \underline{0.747} & \cellcolor{hybrid!80}0.738 & \cellcolor{hybrid!80}\textbf{0.411} & \underline{0.128} & \cellcolor{hybrid!80}\underline{0.665} & 0.455 & 0.647 & 0.630 \\
			\cmidrule(lr){3-14}

			& & {\rotatebox{0}{\shortstack{NeuRBF (T=22)}}} & 0.488 & 0.944 & 0.869 & 0.645 & 0.400 & 0.156 & 0.058 & 0.576 & 0.306 & 0.504 & 0.495 \\
			\cmidrule(lr){3-14}

			& & {\rotatebox{0}{\shortstack{NeuRBF (T=24)}}} & 0.556 & 0.946 & 0.877 & 0.652 & 0.575 & 0.174 & 0.069 & 0.576 & 0.342 & 0.512 & 0.528 \\
			\cmidrule(lr){3-14}

			& & {\rotatebox{0}{\shortstack{MetricGrids (T=22)}}} & 0.572 & 0.965 & 0.906 & 0.736 & 0.675 & 0.339 & 0.126 & 0.648 & 0.429 & 0.608 & 0.600 \\
			\cmidrule(lr){3-14}

			& & {\rotatebox{0}{\shortstack{MetricGrids (T=24)}}} & 0.619 & \cellcolor{hybrid!80}0.965 & 0.910 & 0.739 & 0.734 & 0.349 & \cellcolor{hybrid!80}\textbf{0.133} & 0.647 & 0.434 & 0.608 & 0.614 \\
			\cmidrule(lr){2-14}

		& \multirow{8}{*}{\rotatebox{90}{\textbf{Explicit}}}
			& {\rotatebox{0}{\shortstack{GaussianImage (CR=3000)}}} & 0.624 & 0.942 & 0.830 & 0.714 & 0.695 & 0.371 & 0.109 & 0.455 & 0.456 & 0.631 & 0.583 \\
			\cmidrule(lr){3-14}

			& & {\rotatebox{0}{\shortstack{GaussianImage (CR=100)}}} & 0.545 & 0.944 & 0.847 & 0.708 & 0.689 & 0.342 & 0.105 & 0.439 & 0.434 & 0.553 & 0.561 \\
			\cmidrule(lr){3-14}

			& & {\rotatebox{0}{\shortstack{GaussianImage (CR=15)}}} & 0.492 & 0.962 & 0.889 & 0.741 & 0.618 & 0.263 & 0.080 & 0.149 & 0.430 & 0.471 & 0.510 \\
			\cmidrule(lr){3-14}

			& & {\rotatebox{0}{\shortstack{SmartSplat (CR=3000)}}} & \cellcolor{explicit!80}\textbf{0.651} & \cellcolor{explicit!80}\textbf{0.968} & \cellcolor{explicit!80}\textbf{0.919} & \cellcolor{explicit!80}0.742 & \cellcolor{explicit!80}0.736 & \cellcolor{explicit!80}0.380 & \cellcolor{explicit!80}0.120 & \cellcolor{explicit!80}\textbf{0.673} & \cellcolor{explicit!80}\textbf{0.488} & \cellcolor{explicit!80}\textbf{0.675} & \cellcolor{explicit!80}\textbf{0.635} \\
			\cmidrule(lr){3-14}

			& & {\rotatebox{0}{\shortstack{SmartSplat (CR=100)}}} & 0.598 & 0.953 & 0.875 & 0.724 & 0.712 & 0.349 & 0.113 & 0.653 & 0.429 & 0.640 & 0.605 \\
			\cmidrule(lr){3-14}

			& & {\rotatebox{0}{\shortstack{SmartSplat (CR=15)}}} & 0.227 & 0.854 & 0.791 & 0.636 & 0.354 & 0.190 & 0.082 & 0.527 & 0.380 & 0.457 & 0.450 \\
			\cmidrule(lr){2-14}

		& \multirow{4}{*}{\rotatebox{90}{\textbf{Classic}}}
			& {\rotatebox{0}{\shortstack{Nearest Neighbor}}} & 0.615 & 0.956 & 0.895 & 0.705 & 0.649 & 0.295 & 0.088 & 0.638 & 0.426 & 0.619 & 0.589 \\
			\cmidrule(lr){3-14}

			& & {\rotatebox{0}{\shortstack{Bicubic}}} & \cellcolor{classic!80}\underline{0.647} & \cellcolor{classic!80}\underline{0.966} & \cellcolor{classic!80}\underline{0.918} & \cellcolor{classic!80}0.733 & \cellcolor{classic!80}\textbf{0.784} & \cellcolor{classic!80}0.372 & \cellcolor{classic!80}0.123 & \cellcolor{classic!80}0.665 & \cellcolor{classic!80}\underline{0.474} & \cellcolor{classic!80}\underline{0.665} & \cellcolor{classic!80}\underline{0.635} \\
			\cmidrule(lr){3-14}

			& & {\rotatebox{0}{\shortstack{Lanczos}}} & 0.643 & 0.965 & 0.916 & 0.728 & \underline{0.779} & 0.363 & 0.120 & 0.661 & 0.469 & 0.656 & 0.630 \\
			\bottomrule

        \end{tabular}
}
    \caption{
        \textbf{SSIM Performance on \dataname{} for SR$\times4$ and $\times8$.} 
        Best SSIMs within each family are highlighted by different colours; the best overall is in \textbf{bold}, and runner-ups are \underline{underlined}.
    }
    \label{tab:ssim_sr}
\end{table}

    \paragraph{In Image Super-Resolution}, 
    SSIM tracks PSNR $\times4$, reproducing the identical Hybrid--Classic--Explicit--Implicit ordering, and diverges only at $\times8$, where Explicit and Hybrid swap (Explicit--Classic--Hybrid--Implicit). 
    In all cases, the category margins are narrow except for the Implicit family, which ranks last with large gaps.
    Notably, neither metric agrees with the human evaluation, which yields Classic--Hybrid--Implicit--Explicit, with a pronounced gap separating the top two families from the bottom two.
    
\section{Quantitative Perceptual Similarity on Super Resolution}
\label{sec:lpips}

    As discussed in the main paper, PSNR and SSIM are distortion metrics that quantify agreement with a reference and are known to correlate poorly with perceived quality in Super-Resolution, which motivated our human evaluation.
    To complement both perspectives with a quantitative perceptual measure, we additionally report LPIPS~\cite{zhang2018perceptual}, which computes a learned, human-calibrated distance between deep features of the reconstruction and the reference, and is the de facto standard perceptual metric in the super-resolution literature.
    Evaluating LPIPS at gigapixel scale, however, is non-trivial: the memory footprint of a single forward pass grows linearly with the number of pixels, making full-image evaluation infeasible, and no established protocol exists in this regime.
    We therefore adopt a tiled protocol: each reconstruction is partitioned into non-overlapping $8192 \times 8192$ tiles, and then averaged over them.
    \cref{tab:lpips_sr} reports per-image LPIPS for every method at $\times 4$ and $\times 8$ upsampling, mirroring the layout of the PSNR and SSIM tables.
    Despite not being mathematically equivalent to a single forward pass, this metric remains informative since LPIPS is inherently local: it spatially averages feature distances whose receptive field spans only a few hundred pixels, so the discrepancy is confined to tile boundaries, a negligible fraction of the area at this tile size.
    
    \begin{table}[ht]
    \centering
    \resizebox{\linewidth}{!}{
       \begin{tabular}{ccc cccccccccc c}
    
            \toprule
            &
            &
            &
            \rothead{\imageone}
            & \rothead{\imagetwo}
            & \rothead{\imagethree}
            & \rothead{\imagefour}
            & \rothead{\imagefive}
            & \rothead{\imagesix}
            & \rothead{\imageseven}
            & \rothead{\imageeight}
            & \rothead{\imagenine}
            & \rothead{\imageten}
            & \textbf{Avg} \\

            \midrule

\multirow{30}{*}{\rotatebox{90}{Upsample $\mathbf{\times4}$}}
		& \multirow{8}{*}{\rotatebox{90}{\textbf{Implicit}}}
			& {\rotatebox{0}{\shortstack{SIREN (5,256)}}} & 0.621 & \cellcolor{implicit!80}0.119 & 0.265 & 0.509 & 0.775 & 0.975 & 0.937 & 0.576 & 0.787 & 0.434 & 0.600 \\
			\cmidrule(lr){3-14}

			& & {\rotatebox{0}{\shortstack{FINER (5,256)}}} & 0.643 & 0.121 & 0.274 & 0.492 & 0.801 & 0.986 & \cellcolor{implicit!80}0.927 & 0.586 & 0.801 & 0.549 & 0.618 \\
			\cmidrule(lr){3-14}

			& & {\rotatebox{0}{\shortstack{FDHO (5,256)}}} & 0.617 & 0.121 & 0.265 & 0.479 & 0.756 & 0.902 & 0.931 & 0.570 & 0.789 & 0.486 & 0.592 \\
			\cmidrule(lr){3-14}

			& & {\rotatebox{0}{\shortstack{SIREN (8,1024)}}} & 0.589 & 0.120 & 0.262 & 0.490 & 0.519 & 0.892 & 0.971 & 0.566 & \cellcolor{implicit!80}0.773 & 0.306 & 0.549 \\
			\cmidrule(lr){3-14}

			& & {\rotatebox{0}{\shortstack{FINER (8,1024)}}} & \cellcolor{implicit!80}0.567 & 0.120 & \cellcolor{implicit!80}0.259 & \cellcolor{implicit!80}0.458 & \cellcolor{implicit!80}0.362 & \cellcolor{implicit!80}0.783 & 1.056 & \cellcolor{implicit!80}0.560 & 0.779 & \cellcolor{implicit!80}0.253 & \cellcolor{implicit!80}0.520 \\
			\cmidrule(lr){3-14}

			& & {\rotatebox{0}{\shortstack{FDHO (8,1024)}}} & 0.656 & 0.122 & 0.267 & 0.500 & 0.738 & 0.960 & 0.965 & 0.593 & 0.882 & 0.479 & 0.616 \\
			\cmidrule(lr){2-14}

		& \multirow{8}{*}{\rotatebox{90}{\textbf{Hybrid}}}
			& {\rotatebox{0}{\shortstack{I-NGP (T=22)}}} & 0.423 & 0.081 & 0.156 & 0.249 & 0.453 & 0.515 & 0.700 & 0.316 & 0.672 & \underline{0.239} & 0.381 \\
			\cmidrule(lr){3-14}

			& & {\rotatebox{0}{\shortstack{I-NGP (T=24)}}} & 0.408 & 0.080 & 0.149 & 0.240 & 0.444 & 0.473 & 0.643 & 0.306 & 0.647 & 0.286 & 0.368 \\
			\cmidrule(lr){3-14}

			& & {\rotatebox{0}{\shortstack{NeuRBF (T=22)}}} & 0.489 & 0.072 & 0.145 & 0.377 & 0.599 & 0.762 & 1.063 & 0.371 & \cellcolor{hybrid!80}\textbf{0.443} & 0.285 & 0.461 \\
			\cmidrule(lr){3-14}

			& & {\rotatebox{0}{\shortstack{NeuRBF (T=24)}}} & 0.439 & 0.071 & 0.138 & 0.352 & 0.512 & 0.672 & 0.990 & 0.339 & nan & 0.295 & 0.423 \\
			\cmidrule(lr){3-14}

			& & {\rotatebox{0}{\shortstack{MetricGrids (T=22)}}} & \underline{0.381} & \underline{0.049} & \cellcolor{hybrid!80}\textbf{0.112} & \underline{0.209} & \cellcolor{hybrid!80}0.221 & \underline{0.414} & 0.610 & \cellcolor{hybrid!80}\underline{0.299} & 0.625 & \cellcolor{hybrid!80}\textbf{0.196} & \cellcolor{hybrid!80}\textbf{0.312} \\
			\cmidrule(lr){3-14}

			& & {\rotatebox{0}{\shortstack{MetricGrids (T=24)}}} & \cellcolor{hybrid!80}\textbf{0.363} & \cellcolor{hybrid!80}\textbf{0.048} & 0.128 & \cellcolor{hybrid!80}\textbf{0.203} & 0.275 & \cellcolor{hybrid!80}\textbf{0.381} & \cellcolor{hybrid!80}0.543 & 0.300 & \underline{0.597} & 0.288 & \underline{0.313} \\
			\cmidrule(lr){2-14}

		& \multirow{8}{*}{\rotatebox{90}{\textbf{Explicit}}}
			& {\rotatebox{0}{\shortstack{GaussImg (CR=3000)}}} & 0.868 & 0.565 & 0.656 & 0.637 & 0.600 & 0.876 & 0.964 & 0.688 & 1.025 & 0.715 & 0.759 \\
			\cmidrule(lr){3-14}

			& & {\rotatebox{0}{\shortstack{GaussImg (CR=100)}}} & 0.776 & 0.385 & 0.681 & 0.619 & 0.469 & 0.746 & 0.775 & 0.832 & 0.943 & 0.746 & 0.697 \\
			\cmidrule(lr){3-14}

			& & {\rotatebox{0}{\shortstack{GaussImg (CR=15)}}} & 0.768 & 0.305 & 0.528 & 0.533 & 0.511 & 0.686 & 0.604 & 0.792 & 0.934 & 0.785 & 0.645 \\
			\cmidrule(lr){3-14}

			& & {\rotatebox{0}{\shortstack{SmartSplat (CR=3000)}}} & 0.555 & 0.121 & 0.266 & 0.447 & 0.606 & 0.843 & 0.868 & 0.505 & \cellcolor{explicit!80}0.773 & \cellcolor{explicit!80}0.319 & 0.530 \\
			\cmidrule(lr){3-14}

			& & {\rotatebox{0}{\shortstack{SmartSplat (CR=100)}}} & \cellcolor{explicit!80}0.463 & \cellcolor{explicit!80}0.109 & \cellcolor{explicit!80}0.258 & \cellcolor{explicit!80}0.316 & \cellcolor{explicit!80}0.201 & 0.554 & 0.634 & 0.394 & 0.793 & 0.324 & \cellcolor{explicit!80}0.405 \\
			\cmidrule(lr){3-14}

			& & {\rotatebox{0}{\shortstack{SmartSplat (CR=15)}}} & 0.598 & 0.125 & 0.270 & 0.342 & 0.567 & \cellcolor{explicit!80}0.533 & \cellcolor{explicit!80}0.537 & \cellcolor{explicit!80}0.357 & 0.844 & 0.578 & 0.475 \\
			\cmidrule(lr){2-14}

		& \multirow{4}{*}{\rotatebox{90}{\textbf{Classic}}}
			& {\rotatebox{0}{\shortstack{Nearest Neighbor}}} & \cellcolor{classic!80}0.390 & \cellcolor{classic!80}0.061 & 0.127 & \cellcolor{classic!80}0.240 & 0.455 & \cellcolor{classic!80}0.446 & 0.649 & \cellcolor{classic!80}\textbf{0.297} & \cellcolor{classic!80}0.636 & \cellcolor{classic!80}0.344 & 0.365 \\
			\cmidrule(lr){3-14}

			& & {\rotatebox{0}{\shortstack{Bicubic}}} & 0.431 & 0.068 & \cellcolor{classic!80}\underline{0.120} & 0.324 & \cellcolor{classic!80}\textbf{0.070} & 0.533 & \cellcolor{classic!80}\textbf{0.327} & 0.415 & 0.733 & 0.374 & \cellcolor{classic!80}0.339 \\
			\cmidrule(lr){3-14}

			& & {\rotatebox{0}{\shortstack{Lanczos}}} & 0.439 & 0.072 & 0.127 & 0.332 & \underline{0.077} & 0.548 & \underline{0.333} & 0.427 & 0.737 & 0.402 & 0.349 \\
			\midrule

		\multirow{30}{*}{\rotatebox{90}{Upsample $\mathbf{\times8}$}}
		& \multirow{8}{*}{\rotatebox{90}{\textbf{Implicit}}}
			& {\rotatebox{0}{\shortstack{SIREN (5,256)}}} & 0.624 & \cellcolor{implicit!80}0.120 & 0.267 & 0.513 & 0.778 & 0.977 & 0.979 & 0.573 & 0.792 & 0.438 & 0.606 \\
			\cmidrule(lr){3-14}

			& & {\rotatebox{0}{\shortstack{FINER (5,256)}}} & 0.643 & 0.120 & 0.269 & 0.495 & 0.799 & 1.013 & 0.958 & 0.579 & 0.816 & 0.475 & 0.617 \\
			\cmidrule(lr){3-14}

			& & {\rotatebox{0}{\shortstack{FDHO (5,256)}}} & 0.619 & 0.120 & 0.265 & \cellcolor{implicit!80}0.486 & 0.766 & 0.937 & \cellcolor{implicit!80}0.927 & 0.570 & 0.789 & 0.473 & 0.595 \\
			\cmidrule(lr){3-14}

			& & {\rotatebox{0}{\shortstack{SIREN (8,1024)}}} & \cellcolor{implicit!80}0.585 & 0.120 & \cellcolor{implicit!80}0.257 & 0.500 & 0.390 & 0.897 & 0.987 & \cellcolor{implicit!80}0.569 & \cellcolor{implicit!80}0.782 & \cellcolor{implicit!80}\textbf{0.325} & \cellcolor{implicit!80}0.541 \\
			\cmidrule(lr){3-14}

			& & {\rotatebox{0}{\shortstack{FINER (8,1024)}}} & 0.593 & 0.122 & 0.271 & 0.518 & \cellcolor{implicit!80}0.355 & 0.895 & 1.060 & 0.618 & 0.938 & 0.465 & 0.584 \\
			\cmidrule(lr){3-14}

			& & {\rotatebox{0}{\shortstack{FDHO (8,1024)}}} & 0.689 & 0.120 & 0.265 & 0.497 & 0.701 & \cellcolor{implicit!80}0.887 & 1.072 & 0.585 & 0.991 & 0.467 & 0.627 \\
			\cmidrule(lr){2-14}

		& \multirow{8}{*}{\rotatebox{90}{\textbf{Hybrid}}}
			& {\rotatebox{0}{\shortstack{I-NGP (T=22)}}} & \underline{0.530} & 0.108 & 0.222 & \underline{0.356} & 0.653 & 0.619 & 0.817 & \cellcolor{hybrid!80}\textbf{0.413} & 0.810 & \cellcolor{hybrid!80}0.399 & 0.493 \\
			\cmidrule(lr){3-14}

			& & {\rotatebox{0}{\shortstack{I-NGP (T=24)}}} & 0.532 & 0.108 & 0.223 & \cellcolor{hybrid!80}\textbf{0.356} & 0.652 & 0.615 & 0.813 & \underline{0.414} & 0.815 & 0.410 & 0.494 \\
			\cmidrule(lr){3-14}

			& & {\rotatebox{0}{\shortstack{NeuRBF (T=22)}}} & 0.606 & \underline{0.091} & 0.221 & 0.504 & 0.708 & 0.932 & 1.105 & 0.505 & \cellcolor{hybrid!80}\textbf{0.622} & 0.435 & 0.573 \\
			\cmidrule(lr){3-14}

			& & {\rotatebox{0}{\shortstack{NeuRBF (T=24)}}} & \cellcolor{hybrid!80}\textbf{0.520} & \cellcolor{hybrid!80}\textbf{0.090} & \cellcolor{hybrid!80}\underline{0.212} & 0.461 & 0.603 & 0.870 & 1.073 & 0.502 & \underline{0.638} & 0.412 & 0.538 \\
			\cmidrule(lr){3-14}

			& & {\rotatebox{0}{\shortstack{MetricGrids (T=22)}}} & 0.533 & 0.099 & 0.226 & 0.396 & 0.419 & \underline{0.564} & \underline{0.570} & 0.454 & 0.842 & 0.453 & \underline{0.456} \\
			\cmidrule(lr){3-14}

			& & {\rotatebox{0}{\shortstack{MetricGrids (T=24)}}} & 0.539 & 0.102 & 0.237 & 0.382 & \cellcolor{hybrid!80}0.347 & \cellcolor{hybrid!80}\textbf{0.563} & \cellcolor{hybrid!80}\textbf{0.569} & 0.469 & 0.849 & 0.477 & \cellcolor{hybrid!80}\textbf{0.453} \\
			\cmidrule(lr){2-14}

		& \multirow{8}{*}{\rotatebox{90}{\textbf{Explicit}}}
			& {\rotatebox{0}{\shortstack{GaussImg (CR=3000)}}} & 0.834 & 0.310 & 0.636 & 0.604 & 0.780 & 0.978 & 1.032 & 0.725 & 1.049 & 0.554 & 0.750 \\
			\cmidrule(lr){3-14}

			& & {\rotatebox{0}{\shortstack{GaussImg (CR=100)}}} & 0.763 & 0.336 & 0.599 & 0.604 & 0.465 & 0.751 & 0.849 & 0.831 & 0.964 & 0.701 & 0.686 \\
			\cmidrule(lr){3-14}

			& & {\rotatebox{0}{\shortstack{GaussImg (CR=15)}}} & 0.778 & 0.134 & 0.335 & 0.511 & 0.842 & 0.905 & 0.863 & 0.930 & 0.860 & 0.767 & 0.692 \\
			\cmidrule(lr){3-14}

			& & {\rotatebox{0}{\shortstack{SmartSplat (CR=3000)}}} & \cellcolor{explicit!80}0.570 & 0.125 & \cellcolor{explicit!80}0.275 & 0.438 & 0.602 & 0.840 & 0.863 & 0.495 & \cellcolor{explicit!80}0.805 & \cellcolor{explicit!80}\underline{0.330} & 0.534 \\
			\cmidrule(lr){3-14}

			& & {\rotatebox{0}{\shortstack{SmartSplat (CR=100)}}} & 0.633 & \cellcolor{explicit!80}0.120 & 0.281 & \cellcolor{explicit!80}0.398 & \cellcolor{explicit!80}0.388 & \cellcolor{explicit!80}0.664 & 0.723 & \cellcolor{explicit!80}0.484 & 0.881 & 0.531 & \cellcolor{explicit!80}0.510 \\
			\cmidrule(lr){3-14}

			& & {\rotatebox{0}{\shortstack{SmartSplat (CR=15)}}} & 0.772 & 0.237 & 0.322 & 0.457 & 0.650 & 0.705 & \cellcolor{explicit!80}0.581 & 0.627 & 0.839 & 0.657 & 0.585 \\
			\cmidrule(lr){2-14}

		& \multirow{4}{*}{\rotatebox{90}{\textbf{Classic}}}
			& {\rotatebox{0}{\shortstack{Nearest Neighbor}}} & \cellcolor{classic!80}0.536 & \cellcolor{classic!80}0.108 & 0.218 & \cellcolor{classic!80}0.368 & 0.669 & \cellcolor{classic!80}0.618 & \cellcolor{classic!80}0.860 & \cellcolor{classic!80}0.420 & \cellcolor{classic!80}0.833 & 0.431 & \cellcolor{classic!80}0.506 \\
			\cmidrule(lr){3-14}

			& & {\rotatebox{0}{\shortstack{Bicubic}}} & 0.600 & 0.124 & \cellcolor{classic!80}\textbf{0.207} & 0.466 & \cellcolor{classic!80}\textbf{0.232} & 0.798 & 0.894 & 0.524 & 0.936 & \cellcolor{classic!80}0.383 & 0.516 \\
			\cmidrule(lr){3-14}

			& & {\rotatebox{0}{\shortstack{Lanczos}}} & 0.622 & 0.127 & 0.214 & 0.478 & \underline{0.242} & 0.820 & 0.925 & 0.543 & 0.950 & 0.420 & 0.534 \\
			\bottomrule

        \end{tabular}
}
    \caption{
        \textbf{LPIPS Performance on \dataname{} for SR$\times4$ and $\times8$ (lower is better).} 
        Best LPIPS within each family are highlighted by different colours; the best overall is in \textbf{bold}, and runner-ups are \underline{underlined}.
    }
    \label{tab:lpips_sr}
\end{table}

    \cref{tab:lpips_sr} reveals three consistent patterns.
    Although LPIPS is calibrated on human judgments, and one might therefore expect it to align more closely with our human evaluation than with distortion metrics, at the family level it preserves the PSNR ordering, Hybrid--Classic--Explicit--Implicit, at both scales; in particular, it does not reproduce the Classic-first ranking of the human study.
    The agreement with PSNR, however, breaks down within families and at the top: MetricGrids overtakes I-NGP by a clear margin ($0.312$ vs.\ $0.368$ at $\times 4$), providing quantitative support for the analysis of \cref{sec:ingp_exclusion}, as LPIPS penalizes the nearest-neighbour-like blockiness of I-NGP; similarly, SmartSplat (CR$=$100) surpasses its CR$=$3000 counterpart, and Nearest Neighbour overtakes Bicubic at $\times 8$.
    These inversions are consistent with the known tendency of LPIPS to penalize over-smoothing more than aliasing: at $\times 8$, Nearest Neighbour scores better than Bicubic on 7 out of 10 images (e.g., $0.536$ vs.\ $0.600$ on Venice, $0.833$ vs.\ $0.936$ on Lynx), despite trailing it by roughly $1$ dB in PSNR.
    The perceptual penalty assigned to a given artifact is, however, strongly content-dependent: on Kidney, the same blocking artifacts cost Nearest Neighbour almost a threefold LPIPS increase over Bicubic ($0.669$ vs.\ $0.232$ at $\times 8$), whereas PSNR reports only a $1.5$ dB gap, showing that per-image LPIPS scores cannot be compared across images and that averages may hide large content-specific effects.
    Overall, LPIPS corroborates the central claim of our evaluation: quantitative metrics, including perceptual ones, are informative but not sufficient at gigapixel scale, and rankings among closely matched methods should be interpreted jointly with human judgment.

\section{Additional motivation for the exclusion of I-NGP in the Human Evaluation}
\label{sec:ingp_exclusion}
    
    \begin{figure}[ht]
    \centering
    \resizebox{\linewidth}{!}{
    \setlength{\tabcolsep}{1px}
    \begin{tabular}{cccc}
    
     \textbf{GT} & \textbf{I-NGP} & \textbf{MetricGrids} & \textbf{Nearest} \\

        \includegraphics[width=0.24\linewidth]{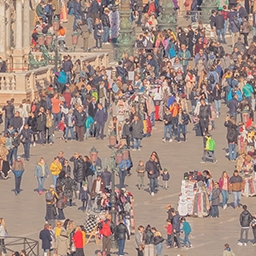} & 
        \includegraphics[width=0.24\linewidth]{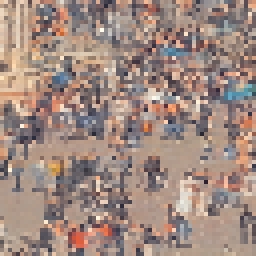} & 
        \includegraphics[width=0.24\linewidth]{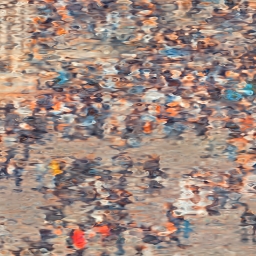} & 
        \includegraphics[width=0.24\linewidth]{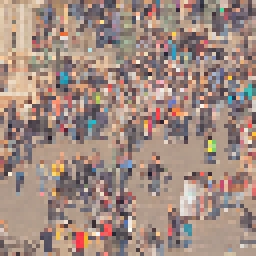} \\

    \end{tabular}}
    \caption{\textbf{Motivation for the exclusion of Instant-NGP from the human evaluation, in favor of the MetricGrids model.} We show the $\times4$ super-resolution qualitatives on a crop of Venice, performed by I-NGP (T=24), MetricGrids (T=24) and Nearest Neighbor.}
    \label{fig:ingp_human_eval}
\end{figure}

    In the main paper, we selected MetricGrids~(T{=}24) as the representative of the Hybrid family despite the marginal PSNR advantage of I-NGP at $\times4$. We detail here why this PSNR gap is misleading and why I-NGP is unsuitable for the human evaluation under our super-resolution protocol. %
    
    \paragraph{Observed effect.}
        As shown in Fig.~\ref{fig:ingp_human_eval}, I-NGP super-resolved outputs are visually indistinguishable from a nearest-neighbor upsampling of the low-resolution reconstruction. 
        Each $\times s$ block of high-resolution pixels (with $s\!\in\!\{4,8\}$ the scale factor) is rendered as a near-constant patch, yielding hard block boundaries and no synthesized high-frequency content. 
        MetricGrids, by contrast, produces continuous transitions between low-resolution sample positions and recovers a more plausible fine structure, yet with wavy artifacts. 
        
    \paragraph{Cause.}
        The difference lies in how I-NGP and MetricsGrids evaluate a query that falls \emph{between} the low-resolution samples seen during fitting.
        
        In I-NGP, every hash level returns a bilinear interpolation of its cell-corner features, so within each low-resolution cell the encoded feature varies only linearly with sub-pixel position, with the sole non-linearity being a small ReLU MLP applied pointwise. 
        Since this MLP is supervised only at the low-resolution sample positions, it has no incentive to shape the output between them, and the in-cell reconstruction collapses to a nearly flat plateau that snaps at cell boundaries. 
        The result is an effectively nearest-neighbor upsampling, hence a shallow bilinear ramp, visually a block of constant colour. 
        In other words, I-NGP fits the low-resolution samples faithfully but interpolates them with a degenerate, almost-linear function, so it cannot synthesize any detail at the new high-resolution coordinates.
    
        MetricGrids avoids this degeneracy by construction. 
        Beyond the linear grid, it adds elementary metric grids whose features depend non-linearly on the sub-pixel displacement ($\|\mathbf{x}-\mathbf{x}_i\|_p^p$, acting as Taylor high-order terms), and a high-order extrapolation decoder that compounds still higher-order terms. 
        The reconstruction between low-resolution samples is therefore a genuine non-linear function of position rather than an affine ramp, which lets MetricGrids vary the signal inside each cell and recover plausible high-frequency structure. 
        The same non-linearity is what occasionally overshoots, producing the mild wavy artifacts noted above.
    
    \paragraph{Implication.}
        High-resolution PSNR rewards I-NGP for exactly reproducing the low-resolution sample values it was fit on, but does not penalize the absence of any synthesized detail between them. Therefore, it fails to capture the perceptual quality that the human evaluation targets. 
        Accordingly, we exclude I-NGP from the human study and adopt MetricGrids~(T{=}24) as the Hybrid representative for both the $\times4$ and $\times8$ settings.

\section{Additional Aggregate Results}
\label{sec:add_agg_res}

    All the aggregate values in the main paper are means over the $N=10$ images of \dataname{}. 
    Since the benchmark spans heterogeneous content, a mean alone does not convey how concentrated the underlying per-image scores are, nor how much of the observed ranking is attributable to a few scenes.
    We therefore report, for every method and metric, compounding dispersion statistics across the benchmark images, including the Standard Deviation of the population (\textit{Std}), and ranking statistics such as the \textit{Median}, the Interquartile Range (\textit{IQR}), the \textit{minima} and the \textit{maxima}.
    \cref{tab:dispersion_psnr_ssim_compression} reports the statistics for compression (PSNR, SSIM, CT, IT), and \cref{tab:dispersion_psnr_ssim_superres} for super-resolution (PSNR and SSIM at $\times 4$ and $\times 8$).

\begin{table*}
\resizebox{\linewidth}{!}{%
\begin{tabular}{lcccccccccccc}
\toprule
 & \multicolumn{3}{c}{PSNR (dB)} & \multicolumn{3}{c}{SSIM} & \multicolumn{3}{c}{CT (min)} & \multicolumn{3}{c}{IT (s)} \\
 & Mean ± Std & Median ± IQR & Min-Max & Mean ± Std & Median ± IQR & Min-Max & Mean ± Std & Median ± IQR & Min-Max & Mean ± Std & Median ± IQR & Min-Max \\
\midrule
SIREN (5,256) & 20.27 ± 5.01 & 21.10 ± 5.58 & 13.16 - 31.26 & 0.452 ± 0.273 & 0.402 ± 0.372 & 0.048 - 0.926 & 18.28 ± 1.51 & 18.08 ± 1.48 & 15.72 - 20.83 & 39.66 ± 2.77 & 39.51 ± 5.52 & 36.40 - 44.06 \\
FINER (5,256) & 19.18 ± 5.62 & 19.01 ± 7.67 & 13.00 - 32.01 & 0.415 ± 0.290 & 0.318 ± 0.433 & 0.048 - 0.927 & 24.02 ± 1.76 & 23.69 ± 1.28 & 21.43 - 27.82 & 71.67 ± 5.42 & 71.48 ± 8.16 & 65.03 - 80.69 \\
FDHO (5,256) & 18.81 ± 5.14 & 17.62 ± 7.23 & 12.60 - 27.16 & 0.420 ± 0.289 & 0.315 ± 0.403 & 0.047 - 0.923 & 24.53 ± 1.47 & 23.80 ± 1.28 & 22.78 - 27.26 & 30.24 ± 1.77 & 30.31 ± 2.81 & 28.11 - 33.15 \\
SIREN (8,1024) & 22.71 ± 5.32 & 23.33 ± 7.67 & 13.21 - 32.29 & 0.499 ± 0.263 & 0.527 ± 0.358 & 0.049 - 0.927 & 175.57 ± 11.11 & 173.55 ± 19.98 & 163.28 - 192.88 & 431.35 ± 27.46 & 428.27 ± 46.33 & 401.50 - 482.18 \\
FINER (8,1024) & 22.67 ± 5.25 & 23.67 ± 6.12 & 13.28 - 31.66 & 0.494 ± 0.270 & 0.530 ± 0.382 & 0.051 - 0.926 & 203.61 ± 12.81 & 203.23 ± 20.54 & 189.15 - 229.20 & 620.27 ± 40.29 & 614.79 ± 69.78 & 574.99 - 698.06 \\
FDHO (8,1024) & 21.79 ± 5.06 & 23.35 ± 6.63 & 13.21 - 30.45 & 0.486 ± 0.272 & 0.526 ± 0.376 & 0.049 - 0.925 & 206.22 ± 13.01 & 203.94 ± 21.12 & 191.57 - 229.19 & 390.93 ± 26.46 & 386.44 ± 46.52 & 362.90 - 440.56 \\
I-NGP (T=22) & 31.13 ± 7.63 & 29.54 ± 6.83 & 16.77 - 45.24 & 0.777 ± 0.163 & 0.787 ± 0.234 & 0.438 - 0.983 & 1.01 ± 0.08 & 1.02 ± 0.12 & 0.85 - 1.11 & 11.40 ± 0.69 & 11.25 ± 0.98 & 10.62 - 12.73 \\
I-NGP (T=24) & 33.19 ± 7.34 & 31.81 ± 8.66 & 19.46 - 46.27 & 0.839 ± 0.116 & 0.853 ± 0.192 & 0.657 - 0.986 & 1.68 ± 0.22 & 1.68 ± 0.32 & 1.26 - 2.02 & 11.50 ± 0.68 & 11.39 ± 1.08 & 10.73 - 12.79 \\
NeuRBF (T=22) & 29.10 ± 7.60 & 27.40 ± 5.88 & 15.56 - 44.21 & 0.722 ± 0.193 & 0.734 ± 0.265 & 0.321 - 0.980 & 11.20 ± 0.47 & 11.13 ± 0.77 & 10.67 - 12.06 & 44.96 ± 2.40 & 45.48 ± 3.46 & 41.24 - 49.37 \\
NeuRBF (T=24) & 31.08 ± 7.74 & 29.24 ± 7.33 & 17.38 - 46.22 & 0.782 ± 0.150 & 0.782 ± 0.232 & 0.510 - 0.986 & 14.76 ± 0.50 & 14.58 ± 0.90 & 14.09 - 15.51 & 44.97 ± 2.34 & 45.67 ± 3.08 & 39.86 - 47.75 \\
MetricGrids (T=22) & 31.58 ± 8.07 & 29.66 ± 7.96 & 17.06 - 47.06 & 0.784 ± 0.160 & 0.794 ± 0.251 & 0.464 - 0.988 & 12.61 ± 0.40 & 12.52 ± 0.56 & 12.14 - 13.46 & 40.17 ± 3.60 & 40.03 ± 4.12 & 33.29 - 46.66 \\
MetricGrids (T=24) & 33.59 ± 8.10 & 31.71 ± 10.03 & 19.54 - 48.58 & 0.841 ± 0.117 & 0.851 ± 0.196 & 0.664 - 0.991 & 18.98 ± 0.90 & 18.55 ± 1.41 & 18.08 - 20.80 & 40.62 ± 4.00 & 40.47 ± 5.84 & 34.07 - 47.49 \\
GaussianImage (CR=3000) & 5.58 ± 3.48 & 5.75 ± 4.05 & 1.24 - 13.92 & 0.054 ± 0.017 & 0.063 ± 0.025 & 0.018 - 0.074 & 32.56 ± 1.90 & 32.27 ± 3.18 & 30.03 - 36.01 & 0.55 ± 1.34 & 0.10 ± 0.01 & 0.09 - 4.57 \\
GaussianImage (CR=100) & 17.18 ± 1.98 & 16.92 ± 2.31 & 13.70 - 20.68 & 0.419 ± 0.185 & 0.327 ± 0.273 & 0.226 - 0.791 & 36.90 ± 2.46 & 36.97 ± 3.54 & 32.90 - 41.70 & 1.27 ± 1.78 & 0.11 ± 2.57 & 0.10 - 4.43 \\
GaussianImage (CR=15) & 17.74 ± 2.11 & 17.11 ± 2.59 & 15.28 - 22.55 & 0.503 ± 0.144 & 0.460 ± 0.226 & 0.354 - 0.811 & 66.54 ± 5.05 & 67.19 ± 4.02 & 55.10 - 74.88 & 5.85 ± 0.98 & 5.50 ± 1.81 & 4.87 - 7.49 \\
SmartSplat (CR=3000) & 24.05 ± 6.57 & 23.35 ± 3.02 & 13.16 - 38.39 & 0.653 ± 0.232 & 0.695 ± 0.200 & 0.139 - 0.969 & 69.26 ± 3.90 & 69.31 ± 5.27 & 63.20 - 77.03 & 0.62 ± 1.43 & 0.14 ± 0.01 & 0.13 - 4.90 \\
SmartSplat (CR=100) & 28.74 ± 6.53 & 27.64 ± 8.93 & 15.05 - 38.07 & 0.769 ± 0.168 & 0.782 ± 0.224 & 0.392 - 0.974 & 76.97 ± 5.57 & 76.97 ± 8.85 & 68.32 - 85.50 & 1.31 ± 1.76 & 0.17 ± 2.57 & 0.15 - 4.51 \\
SmartSplat (CR=15) & 30.40 ± 4.69 & 29.98 ± 4.97 & 21.73 - 40.43 & 0.886 ± 0.083 & 0.898 ± 0.108 & 0.721 - 0.982 & 98.48 ± 6.43 & 97.02 ± 8.38 & 90.42 - 110.57 & 3.36 ± 2.54 & 5.14 ± 5.08 & 0.25 - 5.94 \\
\bottomrule
\end{tabular}
}
\caption{\textbf{Dispersion statistics of PSNR, SSIM, Compression Time (CT) and Inference Time (IT) across the benchmark images for the task of Image Compression.}}
\label{tab:dispersion_psnr_ssim_compression}
\end{table*}

\begin{table*}
\resizebox{\linewidth}{!}{%
\begin{tabular}{lcccccccccccc}
 & \multicolumn{6}{c}{$\mathbf{\times4}$} & \multicolumn{6}{c}{$\mathbf{\times8}$} \\
 \cmidrule(lr){2-7}\cmidrule(lr){8-13}
 & \multicolumn{3}{c}{PSNR (dB)} & \multicolumn{3}{c}{SSIM} & \multicolumn{3}{c}{PSNR (dB)} & \multicolumn{3}{c}{SSIM} \\
 & Mean ± Std & Median ± IQR & Min-Max & Mean ± Std & Median ± IQR & Min-Max & Mean ± Std & Median ± IQR & Min-Max & Mean ± Std & Median ± IQR & Min-Max \\
\midrule
SIREN (5,256) & 20.99 ± 5.57 & 21.28 ± 6.71 & 13.19 - 32.12 & 0.455 ± 0.274 & 0.408 ± 0.369 & 0.048 - 0.927 & 20.85 ± 5.53 & 21.10 ± 6.62 & 13.19 - 32.07 & 0.453 ± 0.275 & 0.404 ± 0.370 & 0.048 - 0.927 \\
FINER (5,256) & 19.44 ± 4.42 & 20.08 ± 5.76 & 13.13 - 27.68 & 0.442 ± 0.277 & 0.379 ± 0.383 & 0.048 - 0.923 & 20.07 ± 4.79 & 20.48 ± 6.15 & 13.17 - 28.95 & 0.447 ± 0.276 & 0.392 ± 0.378 & 0.048 - 0.924 \\
FDHO (5,256) & 18.92 ± 4.40 & 19.76 ± 6.34 & 12.98 - 27.30 & 0.438 ± 0.279 & 0.378 ± 0.388 & 0.047 - 0.923 & 19.73 ± 5.01 & 20.06 ± 6.68 & 13.12 - 29.69 & 0.443 ± 0.278 & 0.386 ± 0.385 & 0.048 - 0.925 \\
SIREN (8,1024) & 22.31 ± 5.38 & 21.98 ± 7.25 & 13.22 - 32.27 & 0.483 ± 0.264 & 0.481 ± 0.352 & 0.049 - 0.927 & 21.76 ± 5.55 & 21.51 ± 7.21 & 13.21 - 32.43 & 0.469 ± 0.270 & 0.448 ± 0.354 & 0.048 - 0.928 \\
FINER (8,1024) & 22.89 ± 5.56 & 22.82 ± 7.22 & 13.24 - 33.26 & 0.495 ± 0.263 & 0.514 ± 0.353 & 0.049 - 0.929 & 21.20 ± 4.90 & 21.01 ± 6.34 & 13.20 - 29.71 & 0.461 ± 0.272 & 0.437 ± 0.365 & 0.046 - 0.925 \\
FDHO (8,1024) & 19.34 ± 4.06 & 20.28 ± 5.86 & 13.14 - 25.56 & 0.446 ± 0.276 & 0.389 ± 0.375 & 0.048 - 0.922 & 20.42 ± 4.97 & 20.48 ± 7.72 & 13.18 - 29.63 & 0.448 ± 0.275 & 0.393 ± 0.385 & 0.047 - 0.925 \\
I-NGP (T=22) & 27.09 ± 7.14 & 25.38 ± 6.92 & 14.43 - 41.13 & 0.708 ± 0.188 & 0.715 ± 0.262 & 0.332 - 0.971 & 24.27 ± 6.89 & 23.13 ± 4.90 & 11.70 - 38.81 & 0.631 ± 0.234 & 0.658 ± 0.243 & 0.126 - 0.962 \\
I-NGP (T=24) & 27.11 ± 7.18 & 25.27 ± 7.38 & 14.43 - 41.15 & 0.713 ± 0.183 & 0.717 ± 0.254 & 0.364 - 0.971 & 24.25 ± 6.88 & 23.12 ± 4.91 & 11.62 - 38.71 & 0.630 ± 0.234 & 0.656 ± 0.244 & 0.128 - 0.962 \\
NeuRBF (T=22) & 24.24 ± 7.29 & 22.81 ± 4.52 & 11.33 - 38.89 & 0.554 ± 0.254 & 0.594 ± 0.262 & 0.081 - 0.947 & 22.95 ± 7.20 & 22.01 ± 2.71 & 10.87 - 38.04 & 0.495 ± 0.268 & 0.496 ± 0.298 & 0.058 - 0.944 \\
NeuRBF (T=24) & 25.11 ± 7.38 & 23.89 ± 5.39 & 11.74 - 39.33 & 0.576 ± 0.247 & 0.599 ± 0.294 & 0.119 - 0.950 & 23.63 ± 7.15 & 22.85 ± 4.40 & 11.21 - 38.06 & 0.528 ± 0.262 & 0.566 ± 0.249 & 0.069 - 0.946 \\
MetricGrids (T=22) & 26.64 ± 7.56 & 24.48 ± 8.85 & 12.89 - 40.75 & 0.674 ± 0.221 & 0.680 ± 0.344 & 0.227 - 0.968 & 24.21 ± 7.22 & 22.81 ± 4.10 & 11.42 - 39.09 & 0.600 ± 0.240 & 0.628 ± 0.256 & 0.126 - 0.965 \\
MetricGrids (T=24) & 26.86 ± 7.54 & 24.85 ± 8.29 & 13.09 - 41.00 & 0.675 ± 0.216 & 0.670 ± 0.322 & 0.256 - 0.969 & 24.51 ± 7.22 & 23.22 ± 4.98 & 11.68 - 39.16 & 0.614 ± 0.240 & 0.633 ± 0.260 & 0.133 - 0.965 \\
GaussianImage (CR=3000) & 15.65 ± 3.16 & 14.57 ± 4.36 & 11.88 - 22.87 & 0.481 ± 0.248 & 0.471 ± 0.349 & 0.099 - 0.905 & 19.72 ± 2.88 & 20.06 ± 1.90 & 13.29 - 24.63 & 0.583 ± 0.229 & 0.627 ± 0.254 & 0.109 - 0.942 \\
GaussianImage (CR=100) & 21.93 ± 3.21 & 22.45 ± 3.02 & 13.97 - 25.96 & 0.596 ± 0.196 & 0.602 ± 0.275 & 0.197 - 0.913 & 20.64 ± 4.01 & 21.14 ± 1.99 & 12.00 - 28.29 & 0.561 ± 0.235 & 0.549 ± 0.268 & 0.105 - 0.944 \\
GaussianImage (CR=15) & 17.07 ± 2.77 & 16.57 ± 2.85 & 12.55 - 23.10 & 0.496 ± 0.233 & 0.384 ± 0.340 & 0.198 - 0.914 & 9.23 ± 6.02 & 7.03 ± 8.02 & 2.65 - 22.35 & 0.510 ± 0.282 & 0.482 ± 0.406 & 0.080 - 0.962 \\
SmartSplat (CR=3000) & 24.04 ± 6.75 & 23.09 ± 3.35 & 13.03 - 38.83 & 0.646 ± 0.236 & 0.685 ± 0.212 & 0.129 - 0.969 & 22.83 ± 6.15 & 22.05 ± 2.59 & 12.33 - 35.32 & 0.635 ± 0.238 & 0.674 ± 0.211 & 0.120 - 0.968 \\
SmartSplat (CR=100) & 23.68 ± 4.86 & 23.46 ± 5.34 & 12.83 - 30.45 & 0.677 ± 0.218 & 0.695 ± 0.308 & 0.203 - 0.963 & 19.69 ± 3.34 & 20.42 ± 1.88 & 11.43 - 23.92 & 0.605 ± 0.237 & 0.647 ± 0.250 & 0.113 - 0.953 \\
SmartSplat (CR=15) & 18.75 ± 2.67 & 18.69 ± 1.79 & 12.91 - 23.67 & 0.587 ± 0.213 & 0.586 ± 0.284 & 0.195 - 0.937 & 15.68 ± 2.48 & 15.57 ± 3.00 & 11.30 - 19.48 & 0.450 ± 0.242 & 0.418 ± 0.351 & 0.082 - 0.854 \\
Nearest Neighbor & 24.58 ± 6.96 & 23.98 ± 5.62 & 12.14 - 38.36 & 0.630 ± 0.234 & 0.656 ± 0.260 & 0.157 - 0.955 & 22.45 ± 6.71 & 22.01 ± 2.60 & 11.04 - 36.92 & 0.589 ± 0.248 & 0.629 ± 0.217 & 0.088 - 0.956 \\
Bicubic & 26.00 ± 6.96 & 24.87 ± 6.02 & 13.75 - 39.89 & 0.689 ± 0.206 & 0.699 ± 0.310 & 0.286 - 0.967 & 23.57 ± 6.77 & 22.90 ± 2.96 & 11.99 - 38.13 & 0.635 ± 0.241 & 0.665 ± 0.254 & 0.123 - 0.966 \\
Lanczos & 25.82 ± 6.97 & 24.65 ± 6.17 & 13.54 - 39.67 & 0.682 ± 0.208 & 0.692 ± 0.318 & 0.281 - 0.965 & 23.39 ± 6.77 & 22.67 ± 3.06 & 11.80 - 37.93 & 0.630 ± 0.242 & 0.658 ± 0.254 & 0.120 - 0.965 \\

\bottomrule
\end{tabular}
}
\caption{\textbf{Dispersion statistics of PSNR and SSIM across the benchmark images for the task of Image Super Resolution $\times4$ and $\times8$.}}
\label{tab:dispersion_psnr_ssim_superres}
\end{table*}

    The dispersion statistics confirm the conclusions of the main paper. 
    The family-level orderings do not depend on the aggregator: Hybrid--Explicit--Implicit under PSNR and Explicit--Hybrid--Implicit under SSIM hold on the median exactly as on the mean, on both tasks. 
    The same is true of the competitiveness of classical interpolation, with bicubic staying above every Implicit configuration on the median at both factors ($24.87$ vs. $22.82$~dB at $\times 4$, $22.90$ vs. $21.51$~dB at $\times 8$). 
    Only adjacent entries change places, always by less than $\sigma/\sqrt{N}$: MetricGrids ($T{=}24$) and I-NGP ($T{=}24$) swap the first position in compression, as do the $T{=}22$ and $T{=}24$ variants at $\times 4$. We treat these pairs as indistinguishable at $N=10$. 
    The one real change is SmartSplat (CR$=$15), which on the median moves above every $T{=}22$ Hybrid configuration it trailed on the mean, in line with the strong SSIM results of Explicit representations.
    
    The per-image scores are also skewed, and the direction depends on the family. 
    For all six Hybrid configurations the median sits $1.4$--$1.9$~dB below the mean; for the three $(8,1024)$ Implicit architectures it sits $0.6$--$1.6$~dB above it. 
    The mean thus favours the former and penalises the latter, and the gap between the best method of each family shrinks from $10.9$ to $8.1$~dB when measured on the median. 
    The cause is the content, not the capacity: every Implicit variant bottoms out at about $13$~dB ($0.05$ SSIM) in compression and at both super-resolution factors, whatever its architecture and parameter budget, so a few high-frequency scenes fix their lower tail. 
    Hybrid representations lose quality on those same scenes, but do not collapse.

\section{Additional Human Evaluation scores}
\begin{figure*}
    \centering
    \includegraphics[width=\linewidth]{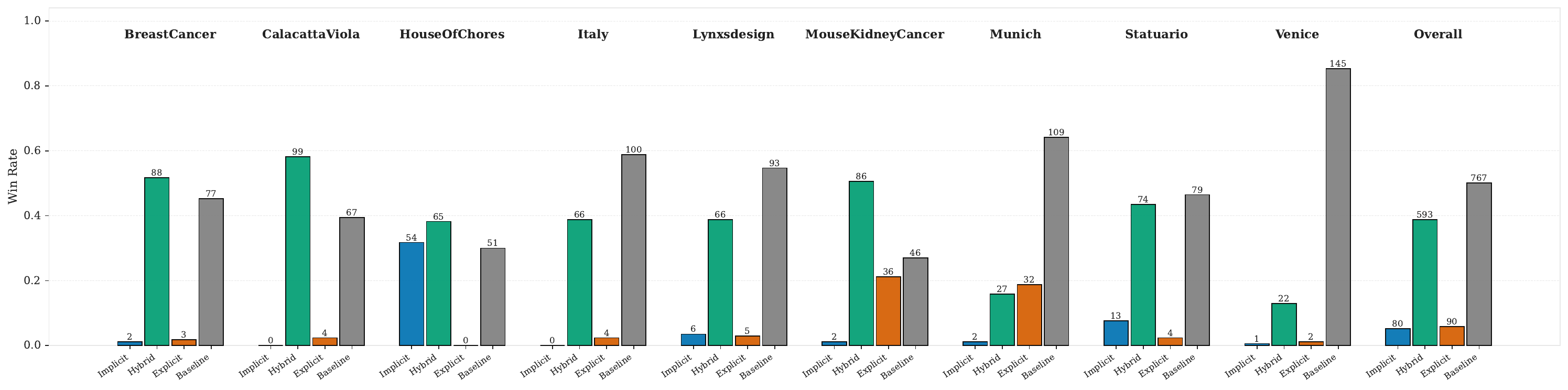}
    \includegraphics[width=\linewidth]{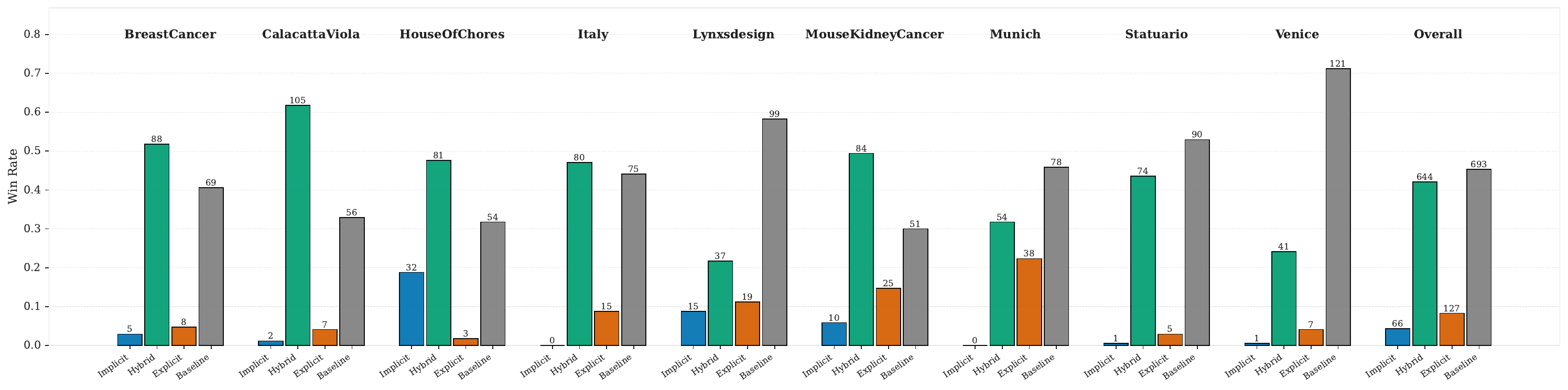}
    \caption{\textbf{Win Rates of the Human Evaluation Study} (i.e. how many participants voted a certain model as the first preference). The label displays the absolute number of voters.}
    \label{fig:win_rate_hist}
\end{figure*}

    This section reports the complete human-evaluation data summarised by the Elo analysis of the main paper.
    For every image and method family, \cref{tab:human_eval} lists both the win rate and the bootstrapped Elo score (mean and standard deviation over $2000$ resamples).
    The two are complementary, as the win rate is an unweighted count of how many times the method family has been chosen as the best, and is therefore directly interpretable, whereas Elo additionally rewards victories over stronger opponents and penalises losses to weaker ones, yielding a schedule-aware ranking.
    \cref{fig:win_rate_hist} visualises the per-image win rates and \cref{fig:elo_boxplot} of the main paper reports the corresponding Elo distributions; both are views of the same judgements tabulated in \cref{tab:human_eval}.
    
    The two metrics agree at the top of the ranking. 
    Across images, classical and hybrid methods accumulate the large majority of wins and the highest Elo scores, consistent with the family-level conclusions of \cref{tab:superres}.
    They diverge, however, at the bottom.
    
    \begin{table*}[t]
    \centering

    \resizebox{\linewidth}{!}{
        \begin{tabular}{ccc ccccccccc c}
            \toprule
            & & & \textit{\imageone} & \textit{\imagetwo} & \textit{\imagethree} & \textit{\imagefour} & \textit{\imagefive} & \textit{\imagesix}  & \textit{\imageeight} & \textit{\imagenine} & \textit{\imageten} & \textbf{Overall} \\
            \midrule

\multirow{8}{*}{\rotatebox{90}{\textbf{Win Rate}}} 
		& \multirow{4}{*}{\rotatebox{0}{$\times 4$}}
			& Implicit & 1 & 13 & 0 & 2 & 2 & 2 & 0 & 6 & 54 & 80 \\
			& & Hybrid & 22 & 74 & 99 & 88 & 86 & 27 & 66 & 66 & 65 & 593 \\
			& & Explicit & 2 & 4 & 4 & 3 & 36 & 32 & 4 & 5 & 0 & 90 \\
			& & Classic & 145 & 79 & 67 & 77 & 46 & 109 & 100 & 93 & 51 & 767 \\

			\cmidrule(lr){3-13}
		& \multirow{4}{*}{\rotatebox{0}{$\times 8$}}
			& Implicit & 1 & 1 & 2 & 5 & 10 & 0 & 0 & 15 & 32 & 66 \\
			& & Hybrid & 41 & 74 & 105 & 88 & 84 & 54 & 80 & 37 & 81 & 644 \\
			& & Explicit & 7 & 5 & 7 & 8 & 25 & 38 & 15 & 19 & 3 & 127 \\
			& & Classic & 121 & 90 & 56 & 69 & 51 & 78 & 75 & 99 & 54 & 693 \\

			\cmidrule(lr){2-13}
		\multirow{8}{*}{\rotatebox{90}{\textbf{ELO}}}
		& \multirow{4}{*}{\rotatebox{0}{$\times 4$}}
			& Implicit & $716 \pm 62$ & $811 \pm 23$ & $683 \pm 37$ & $868 \pm 23$ & $704 \pm 23$ & $660 \pm 24$ & $716 \pm 45$ & $893 \pm 19$ & $1134 \pm 22$ & $843 \pm 6$ \\
			& & Hybrid & $1372 \pm 65$ & $1214 \pm 23$ & $1390 \pm 43$ & $1344 \pm 34$ & $1174 \pm 18$ & $1120 \pm 14$ & $1336 \pm 45$ & $1251 \pm 27$ & $1214 \pm 25$ & $1201 \pm 6$ \\
			& & Explicit & $243 \pm 91$ & $763 \pm 25$ & $597 \pm 45$ & $464 \pm 50$ & $1008 \pm 19$ & $942 \pm 21$ & $542 \pm 51$ & $562 \pm 42$ & $492 \pm 62$ & $728 \pm 8$ \\
			& & Classic & $1669 \pm 84$ & $1212 \pm 24$ & $1329 \pm 38$ & $1325 \pm 34$ & $1114 \pm 14$ & $1277 \pm 21$ & $1406 \pm 50$ & $1295 \pm 32$ & $1159 \pm 27$ & $1228 \pm 7$ \\

			\cmidrule(lr){3-13}
		& \multirow{4}{*}{\rotatebox{0}{$\times 8$}}
			& Implicit & $705 \pm 29$ & $630 \pm 33$ & $736 \pm 25$ & $841 \pm 24$ & $915 \pm 13$ & $705 \pm 22$ & $588 \pm 28$ & $856 \pm 19$ & $1018 \pm 15$ & $816 \pm 6$ \\
			& & Hybrid & $1237 \pm 27$ & $1294 \pm 29$ & $1326 \pm 33$ & $1281 \pm 30$ & $1103 \pm 19$ & $1132 \pm 15$ & $1306 \pm 24$ & $1101 \pm 16$ & $1193 \pm 23$ & $1180 \pm 6$ \\
			& & Explicit & $644 \pm 32$ & $755 \pm 29$ & $688 \pm 31$ & $626 \pm 36$ & $879 \pm 19$ & $999 \pm 20$ & $819 \pm 26$ & $822 \pm 21$ & $608 \pm 35$ & $802 \pm 7$ \\
			& & Classic & $1414 \pm 34$ & $1322 \pm 32$ & $1249 \pm 22$ & $1252 \pm 27$ & $1102 \pm 13$ & $1164 \pm 16$ & $1287 \pm 24$ & $1221 \pm 22$ & $1181 \pm 19$ & $1202 \pm 6$ \\
            \bottomrule
        \end{tabular}
    }
    
    \caption{\textbf{Win Rates and Elo scores of the human evaluation}}
       
    \label{tab:human_eval}
\end{table*}

\section{Additional Implementation Details}
\label{sec:impl}
    All the experiments are performed on a machine having a single 32 GB NVidia RTX 5090 and 128 GB of RAM, requiring approximately 418 GPUh, divided into 182 GPUh for Image Compression and 236GPUh for Super-Resolution at both scales.

    \subsection{Training Details}
    For each family, we defined standard training protocols to guarantee convergence. For instance, in compression, Implicit methods are trained for 8 passes over the full image. 
    As each training step consists of processing a single pixel, and scenes differ in resolution, 8 full image passes map to a different number of steps per scene. Thus, the number of training steps ranges from 30000 to 35000. On the other hand, to achieve convergence, we trained Explicit methods for 10000 training steps, each performed over a full image due to Gaussian Splatting requirements. 
    In the following paragraphs, we describe the exact training details for each family of methods.

    \paragraph{Implicit} models are trained with a PyTorch-based custom implementation, on an MSE loss with a batch size of $2^{18}$ pixels and the Adam optimizer, with $\beta_1=0.9$, $\beta_2=0.95$. 
    The number of training steps corresponds to processing every pixel of the image dataset 8 times (about $30000-35000$ steps) for Compression, while we fix $35000$ steps for Super-Resolution. 
    For each image pass, we shuffle the pixel set with a fixed seed and sample batches according to the new ordering, so a step is a batch of pixels. Only for the Big variants $(8,1024)$, the Adam hyperparameter $\beta_2$ has been scaled according to \cite{marek2026smallbstraining}. 
    This procedure slows down the training slightly, but prevents the model from falling into local minima, which would result in a completely failed fit.
    The alternative would be to increase the batch size, which is too costly for gigapixel images.

    \paragraph{Hybrid} models have been trained according to the original implementations, with custom code for Super-Resolution inference.
    The hyperparameter choice is the same as the original papers suggest for Gigapixel fitting: an MSE loss with a batch size of $2^{18}$, $35000$ steps, hash table hyperparameters $F=2, L=16, \,\log T=24 $, Adam optimizer with $\beta_1 = 0.9, \beta_2 = 0.99, \varepsilon = 10^{-15}$.

    \paragraph{Explicit} models have been trained from the public implementation of SmartSplat, which also contained the code for training GaussianImage, in particular with the RS factorization of the matrix $\Sigma$, which we used for this work. 
    Then, we implemented the custom and mathematically equivalent tiled training with gradient accumulation to enable gigapixel image fitting, detailed in \cref{alg:tiled-gs-training}, and custom inference code for Super-Resolution.    
    More specifically, the tiles are processed as a sliding window with no overlap, and they are created as square crops of size $8192^2$ when possible, and shorter near the borders of the images. 
    The optimizer step is taken once per full-image pass, thus making the update mathematically equivalent to processing the entire image at once. The training has been conducted with a budget of $10000$ full-image optimizer steps.

    \begin{algorithm}
    \begin{algorithmic}
        \Require $I\in\mathbb{R}^{H\times W}$ \Comment{Gigapixel Image}
        \Statex \hspace{2.3em}$\theta=\{\boldsymbol{\mu},\mathbf{s},\mathbf{r},\mathbf{c}\}$ \Comment{Gaussian Parameters}
        \Statex \hspace{2.3em}$T_k=\{(x_0,y_0)\}_{k}$   \Comment{Tile Centers}
        \Statex \hspace{2.3em}$(w,h)$ \Comment{Tile Sizes}
        \Statex \hspace{-1.0em}\textbf{Notation:} subscript $\mathcal{M}$ = parameter subset kept for a tile
\Statex \hrulefill
    \ForAll{tiles $T_k$ with origin $(x_0,y_0)$ and size $w{\times}h$}
        \State $\mathcal{M} \gets \{\, i : \text{Gaussian } i \text{ overlaps } T_k \, \text{by at least } 3\sigma\}$
        
        \State $\boldsymbol{\mu}' \gets \dfrac{2\big(\,\tfrac{1}{2}(\boldsymbol{\mu}_{\mathcal{M}}{+}1)\,(W,H) - (x_0,y_0)\,\big)}{(w,h)} - 1$
        \Statex\Comment{Re-center selected means for current tile}
        
        \State $\mathcal{G} \gets \textsc{Project}(\boldsymbol{\mu}',\, \mathbf{s}_{\mathcal{M}},\, \mathbf{r}_{\mathcal{M}})$
        \State $\hat{I}_k \gets \textsc{Rasterize}(\mathcal{G},\, \mathbf{c}_{\mathcal{M}})$ \Comment{Forward step}
        \State $\ell \gets \mathcal{L}\big(\hat{I}_k,\; I[\,T_k\,]\big)$
        \State \textsc{Backward}($\ell$) \Comment{$\nabla\ell$ \emph{adds} into the grad buffers}
    \EndFor
    \State \textsc{OptimizerStep} \Comment{one update from all tiles}
    \State \textsc{ZeroGrad}
        
    \end{algorithmic}
    \caption{Tiled Gaussian Splatting Training Step.}
    \label{alg:tiled-gs-training}
\end{algorithm}

    \subsection{Timing Procedure}
    All runtimes reported in the main paper are obtained with a single shared protocol, executed on the same machine (RTX~5090, 32GB VRAM, 128GB RAM, CUDA~\texttt{12.8}, PyTorch~\texttt{2.9}) with exclusive access to the GPU and no concurrent workloads. 
    Timestamps are recorded with \texttt{time.perf\_counter()} from the Python standard library; every timestamp is immediately preceded by a \texttt{torch.cuda.synchronize()} call, so that the asynchronous execution of CUDA kernels is never attributed to the wrong phase.
    
    \textbf{Compression Time (CT)} spans the fitting loop only: it starts after the image has been loaded and the representation has been instantiated, and ends after the last parameter update, including all tile scheduling, host-to-device transfers and gradient accumulation performed within the loop. 
    
    \textbf{Inference Time (IT)} spans the decoding of the full-resolution raster from the fitted representation, including tile scheduling and host-side assembly of the output, and excluding serialisation to disk and metric computation.
        
    \paragraph{Scope of the comparison.} CT and IT are end-to-end wall-clock measurements of the reference implementation of each method. They therefore reflect both the algorithm and its engineering: I-NGP builds on fused CUDA kernels, whereas most of the remaining methods are implemented in PyTorch. 
    We deliberately report the cost a practitioner incurs with the publicly available code, rather than an idealised measure of algorithmic complexity, and comparisons across families should be read accordingly.

\section{Additional qualitative results}
\begin{figure*}[ht]
    \centering
    \resizebox{\linewidth}{!}{
    \setlength{\tabcolsep}{1px}
    \begin{tabular}{c ccccc}
        & \textit{Venice} & \textit{Statuario}& \textit{Calacatta} & \textit{Breast Cancer} & \textit{Kidney} \\
        \rotatebox{90}{ \small \hspace{2em} \textbf{GT}} &
        \includegraphics[width=0.2\linewidth, height=0.12\linewidth]{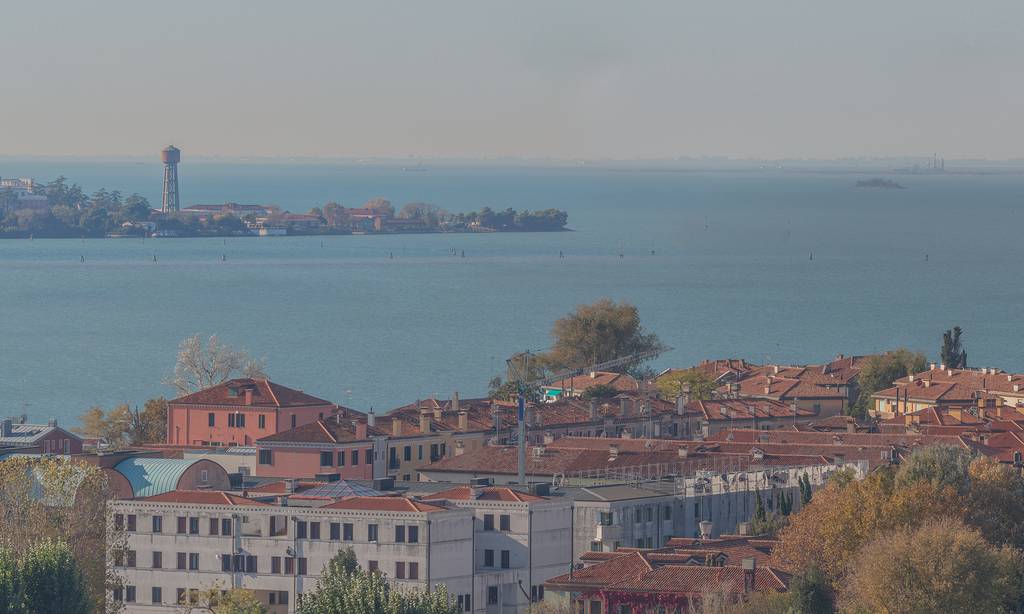} & 
        \includegraphics[width=0.2\linewidth, height=0.12\linewidth]{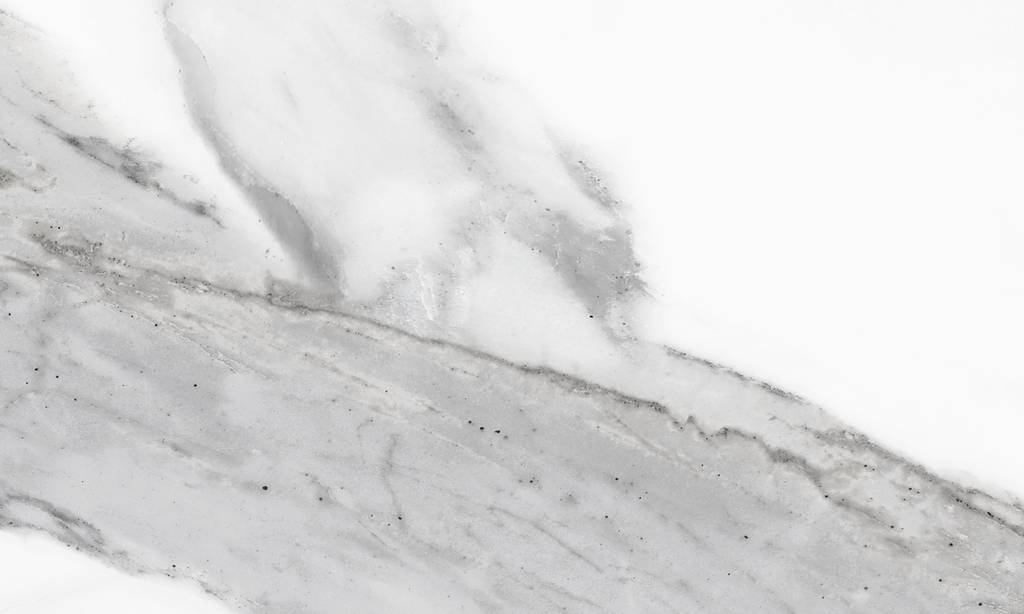} & 
        \includegraphics[width=0.2\linewidth, height=0.12\linewidth]{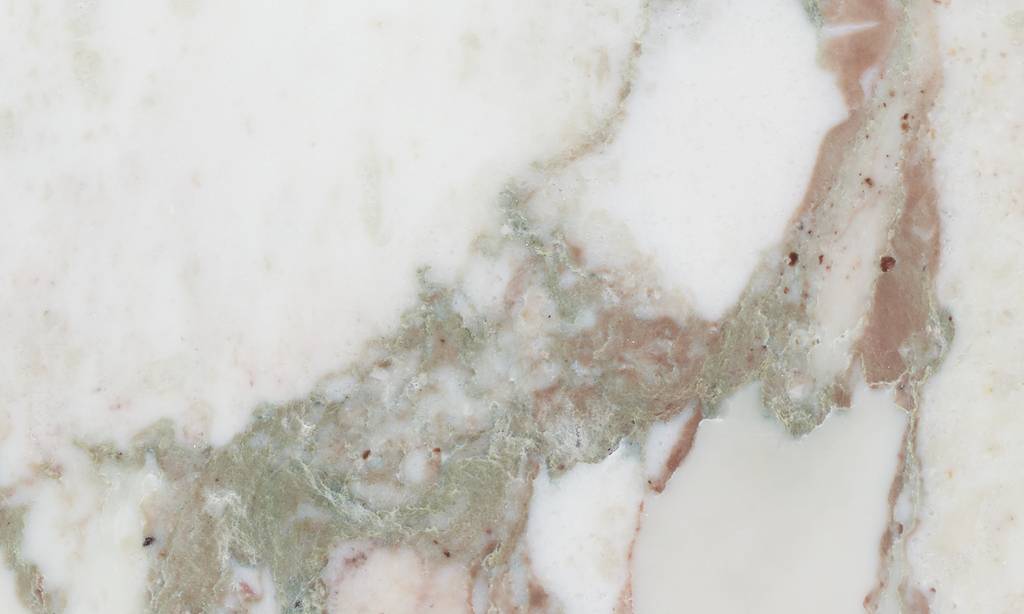} & 
        \includegraphics[width=0.2\linewidth, height=0.12\linewidth]{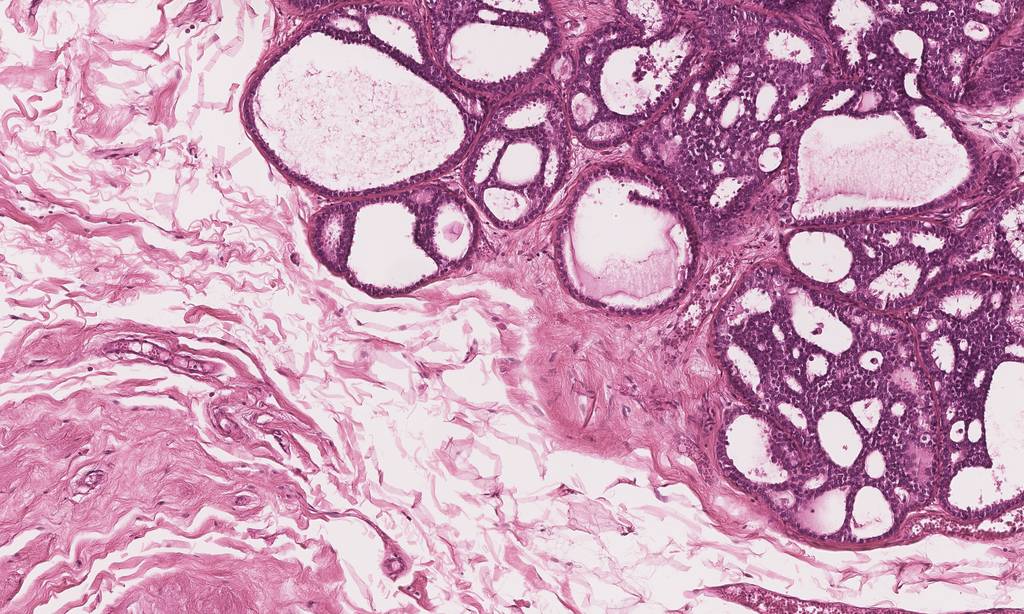} & 
        \includegraphics[width=0.2\linewidth, height=0.12\linewidth]{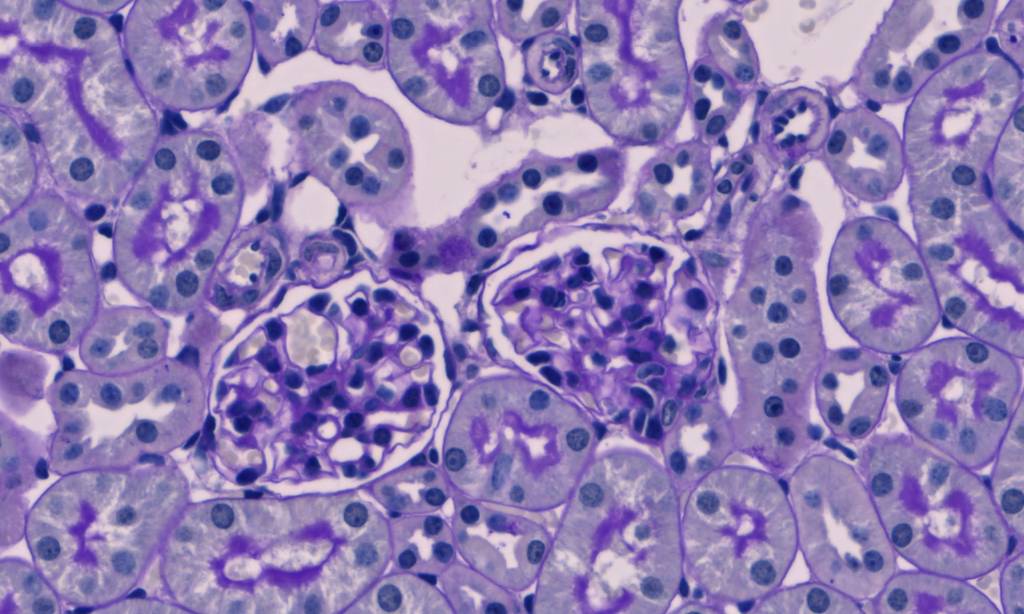} \\

        \rotatebox{90}{ \small \hspace{0.75em} \textbf{SIREN}} &
        \includegraphics[width=0.2\linewidth, height=0.12\linewidth]{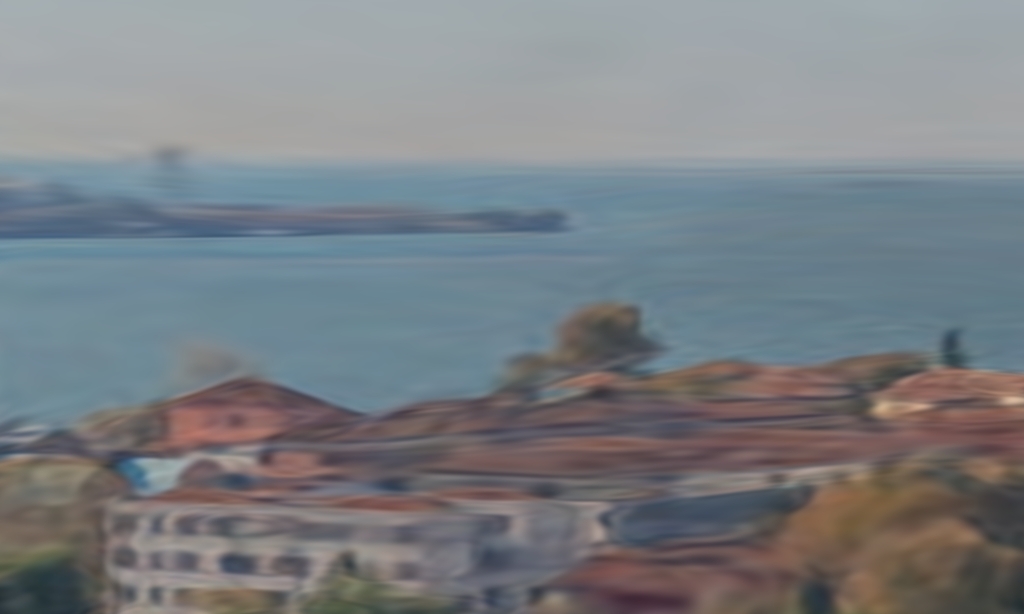} & 
        \includegraphics[width=0.2\linewidth, height=0.12\linewidth]{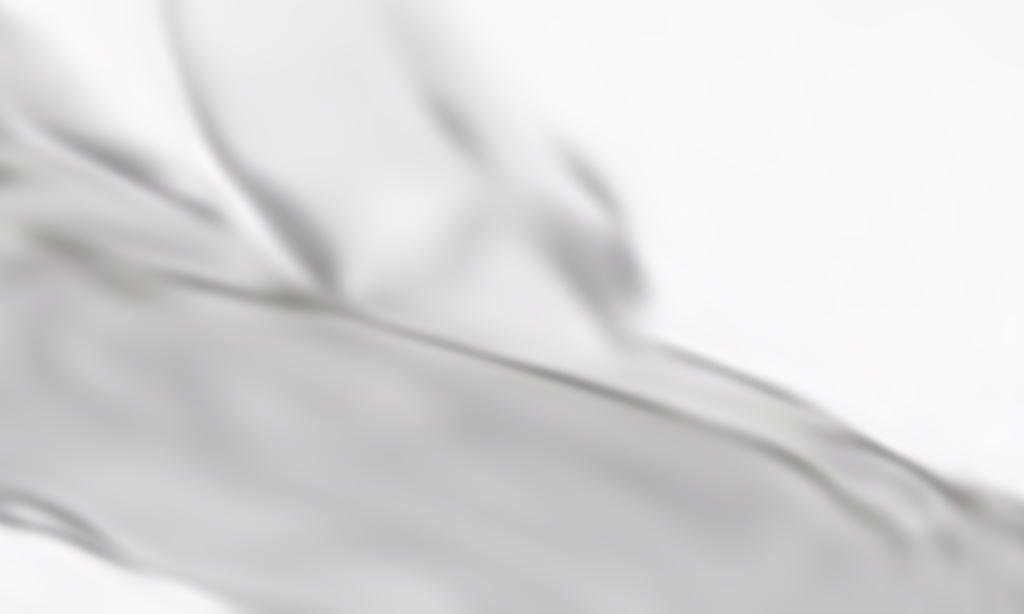} & 
        \includegraphics[width=0.2\linewidth, height=0.12\linewidth]{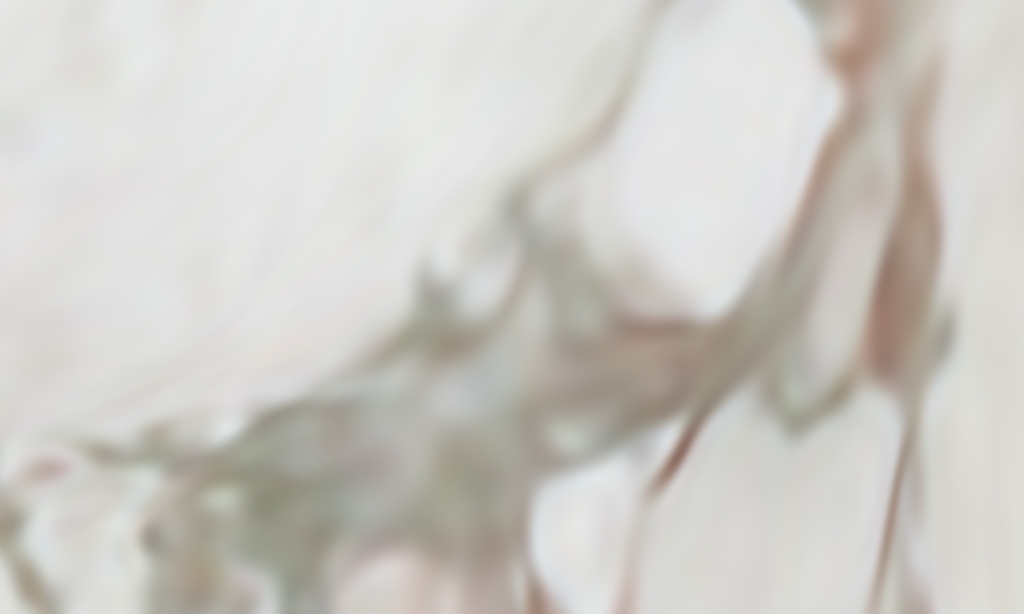} & 
        \includegraphics[width=0.2\linewidth, height=0.12\linewidth]{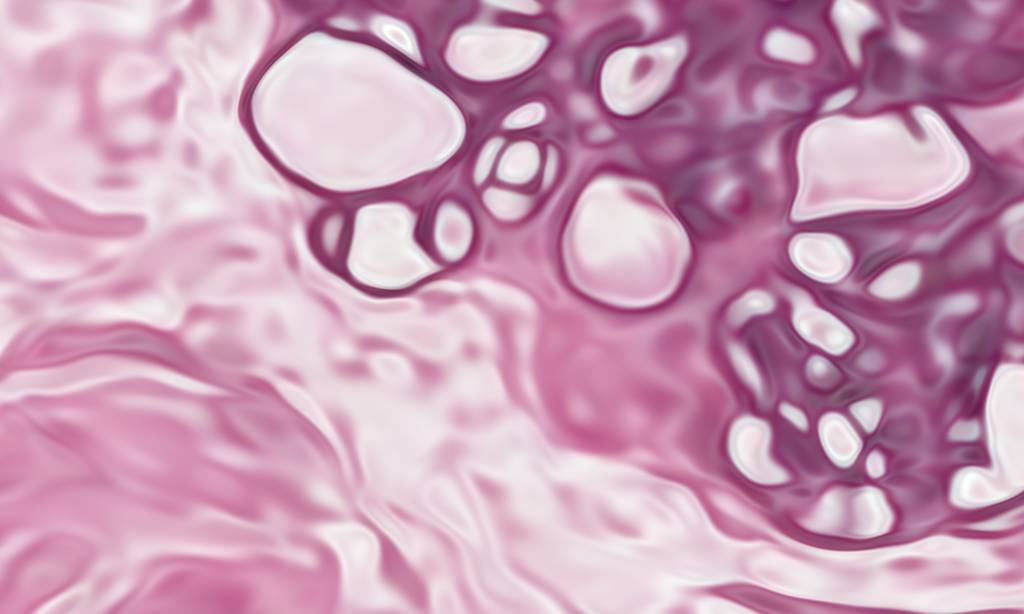} & 
        \includegraphics[width=0.2\linewidth, height=0.12\linewidth]{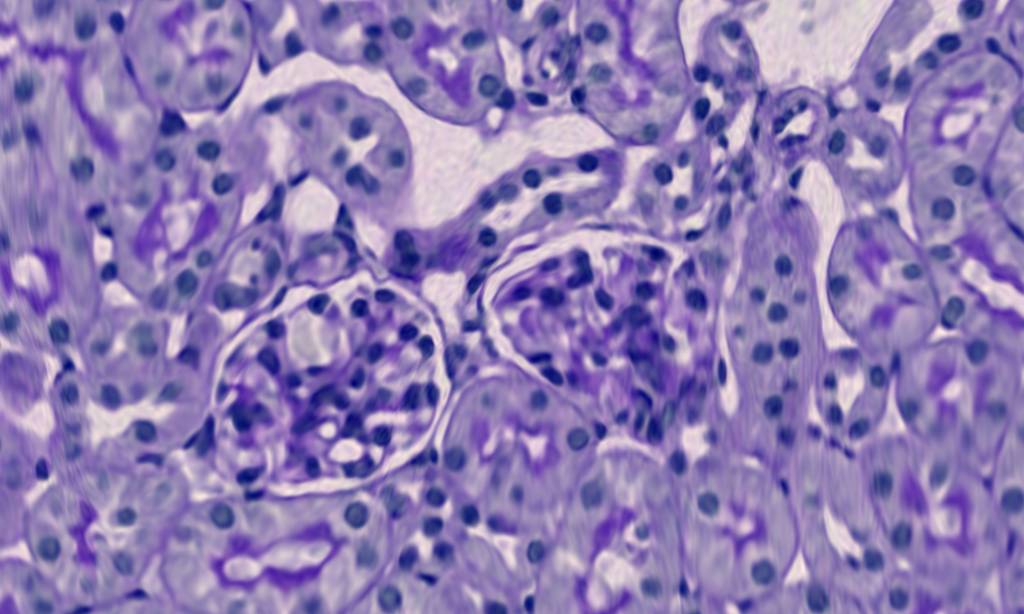} \\

        \rotatebox{90}{ \small \hspace{0.75em} \textbf{FDHO}} &
        \includegraphics[width=0.2\linewidth, height=0.12\linewidth]{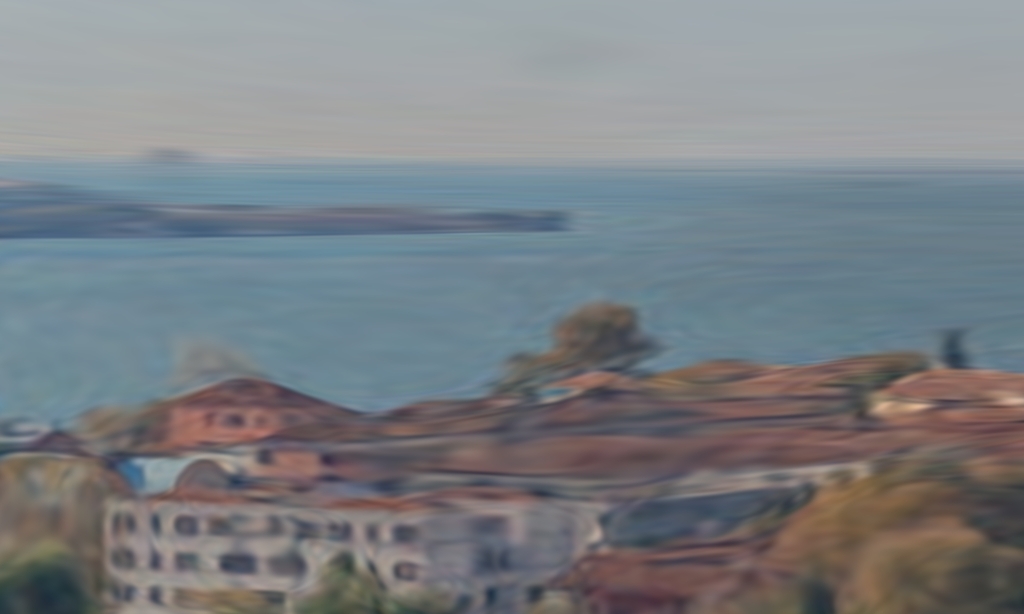} & 
        \includegraphics[width=0.2\linewidth, height=0.12\linewidth]{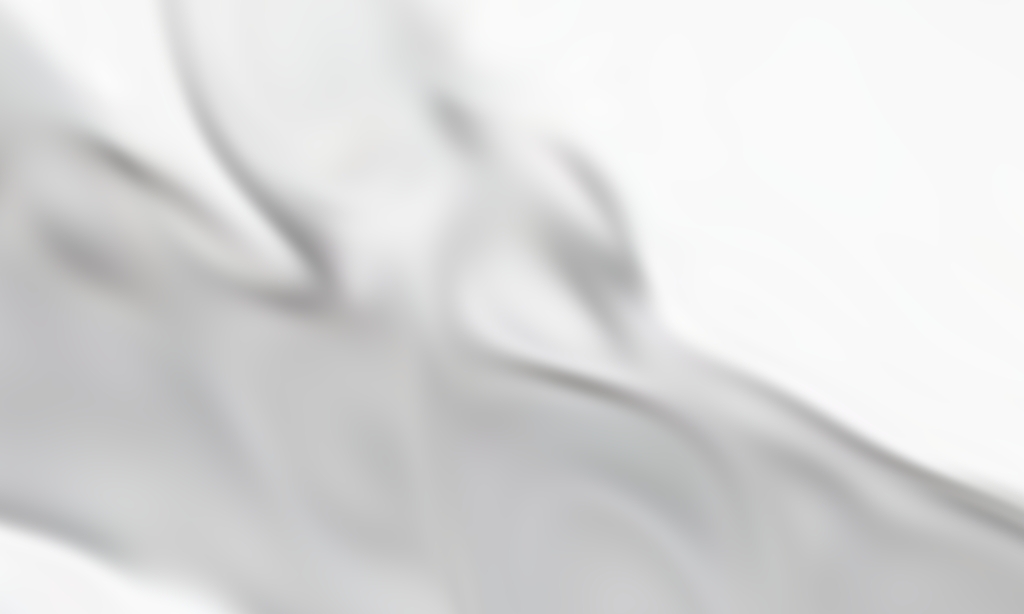} & 
        \includegraphics[width=0.2\linewidth, height=0.12\linewidth]{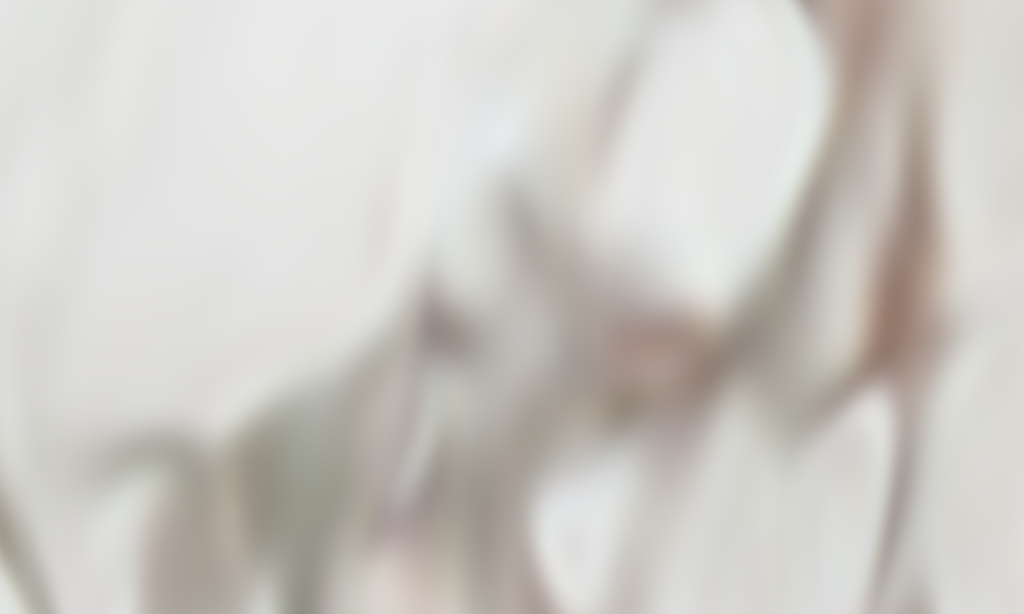} & 
        \includegraphics[width=0.2\linewidth, height=0.12\linewidth]{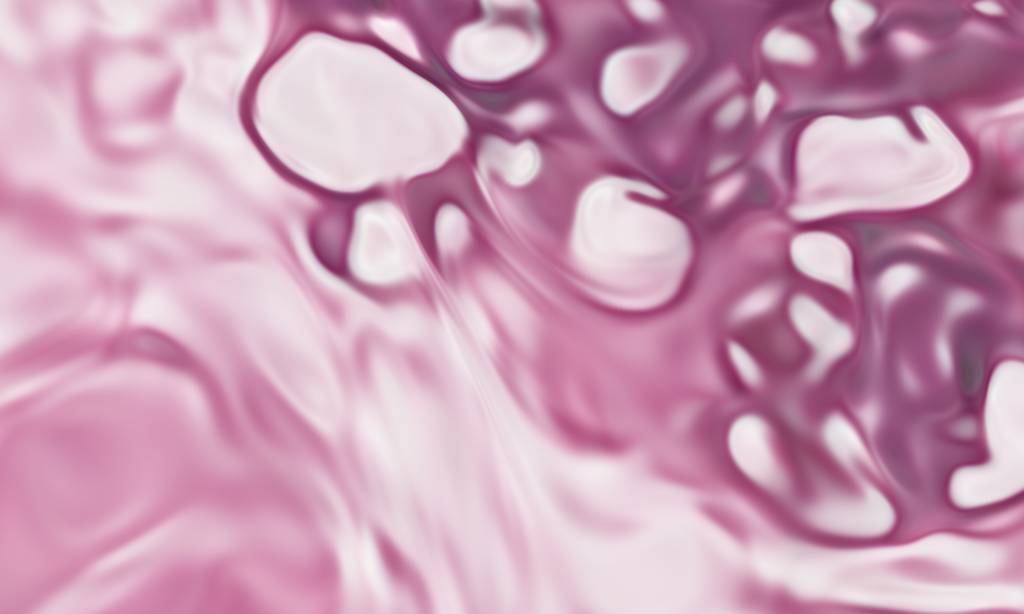} & 
        \includegraphics[width=0.2\linewidth, height=0.12\linewidth]{images/qualitatives_fitting_suppl/fit_MouseKidneyCancer_gt.jpg} \\

        \rotatebox{90}{ \small \hspace{0.75em} \textbf{FINER}} &
        \includegraphics[width=0.2\linewidth, height=0.12\linewidth]{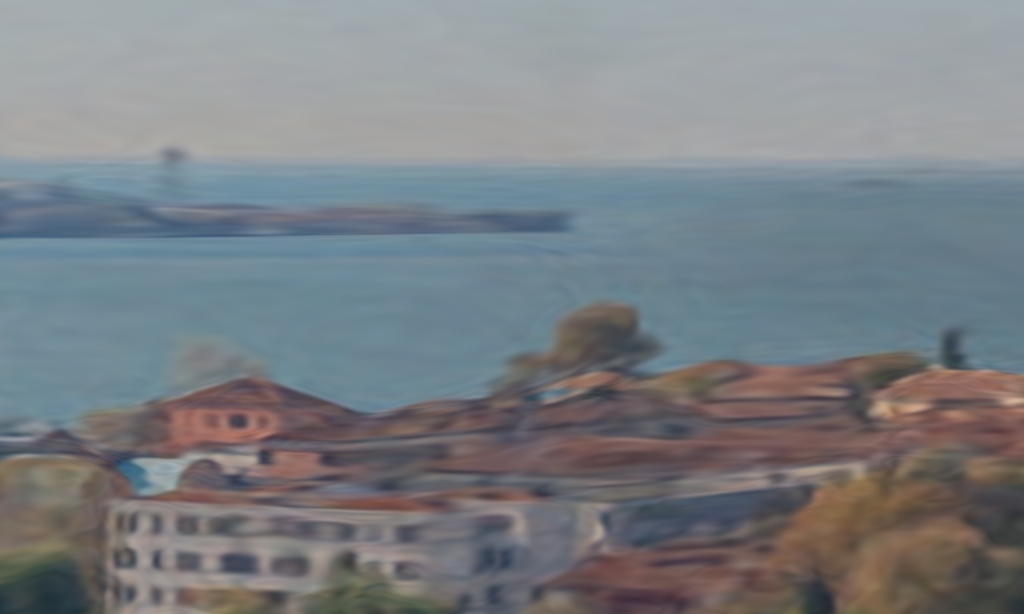} & 
        \includegraphics[width=0.2\linewidth, height=0.12\linewidth]{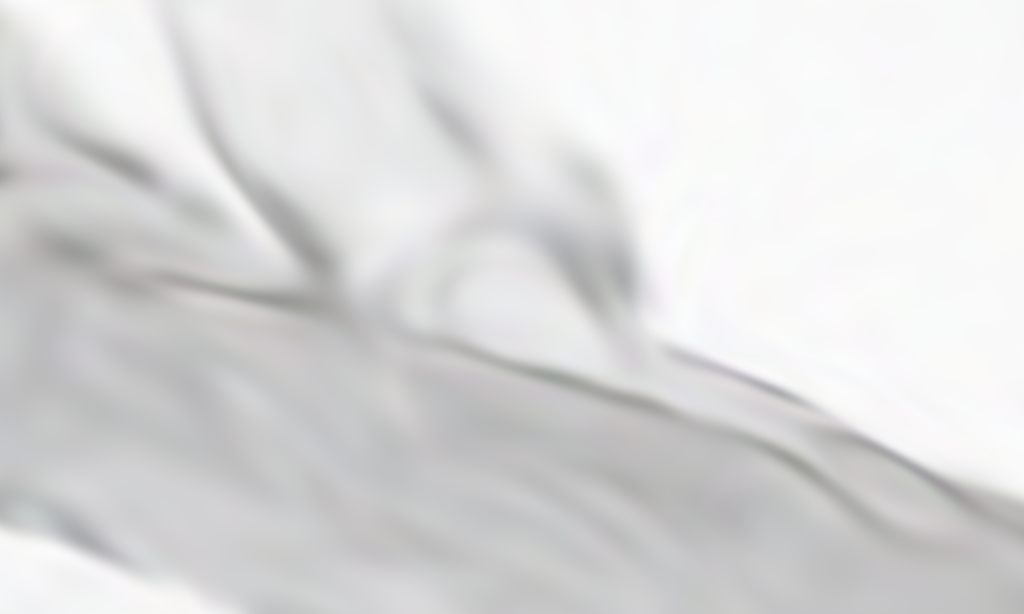} & 
        \includegraphics[width=0.2\linewidth, height=0.12\linewidth]{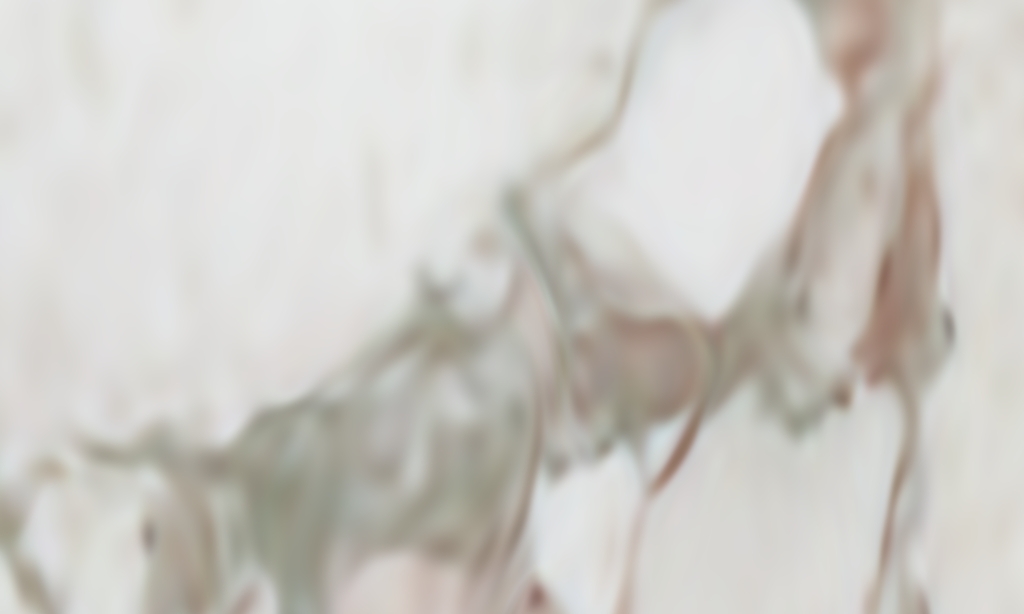} & 
        \includegraphics[width=0.2\linewidth, height=0.12\linewidth]{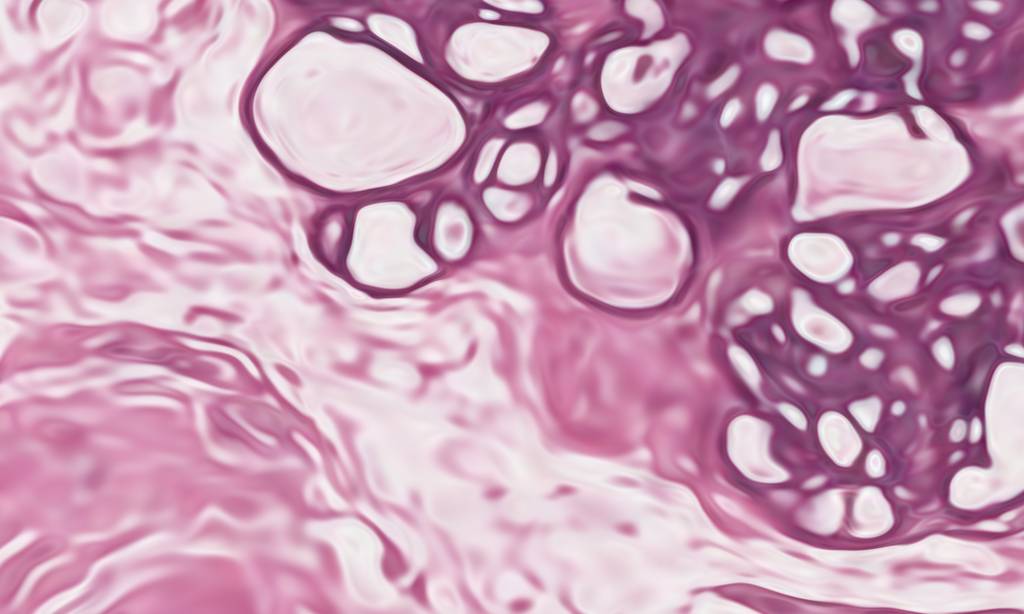} & 
        \includegraphics[width=0.2\linewidth, height=0.12\linewidth]{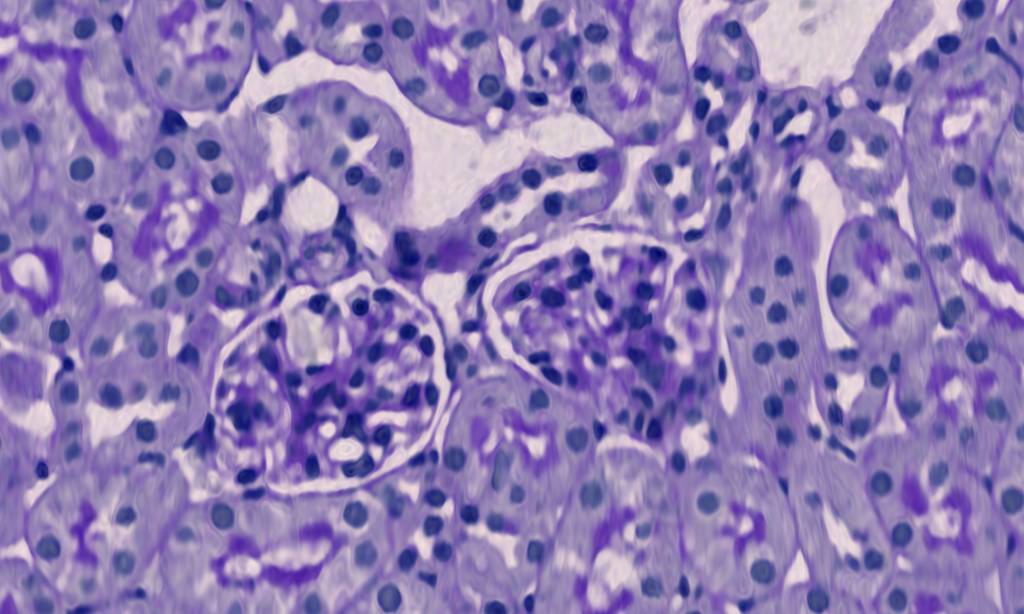} \\

        \rotatebox{90}{ \small \hspace{0.75em} \textbf{I-NGP}} &
        \includegraphics[width=0.2\linewidth, height=0.12\linewidth]{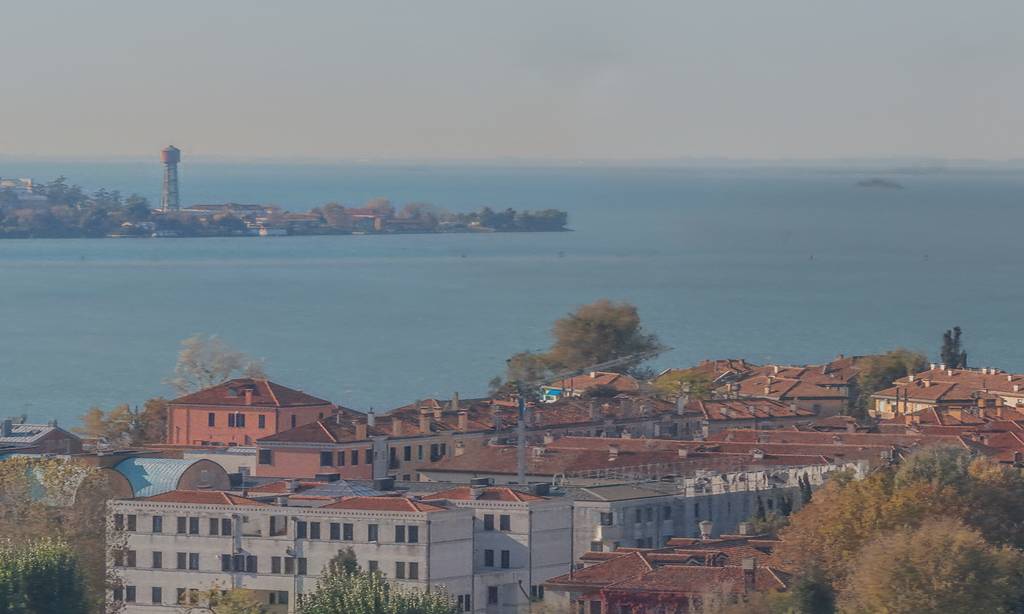} & 
        \includegraphics[width=0.2\linewidth, height=0.12\linewidth]{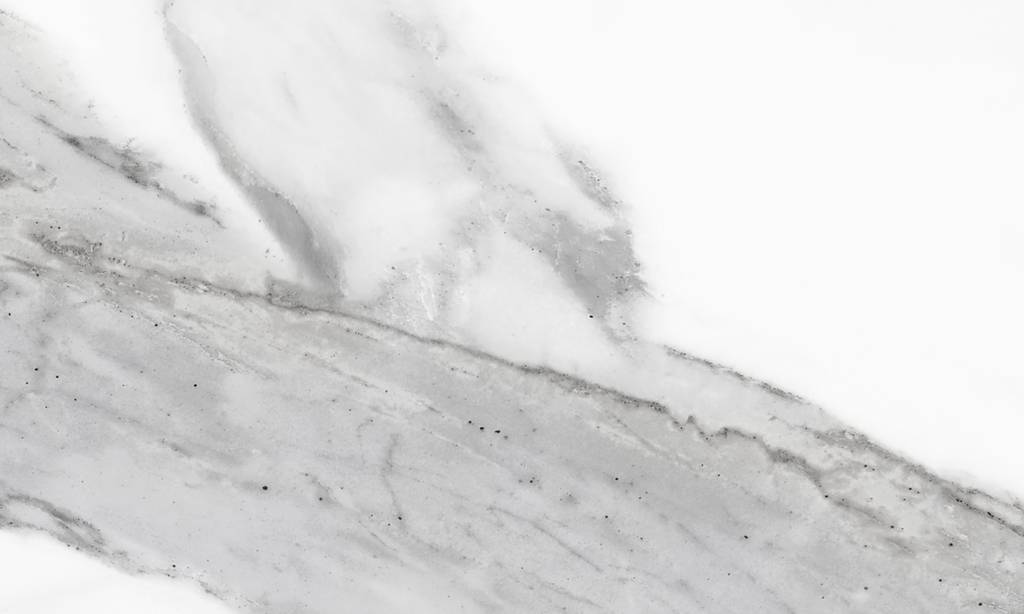} & 
        \includegraphics[width=0.2\linewidth, height=0.12\linewidth]{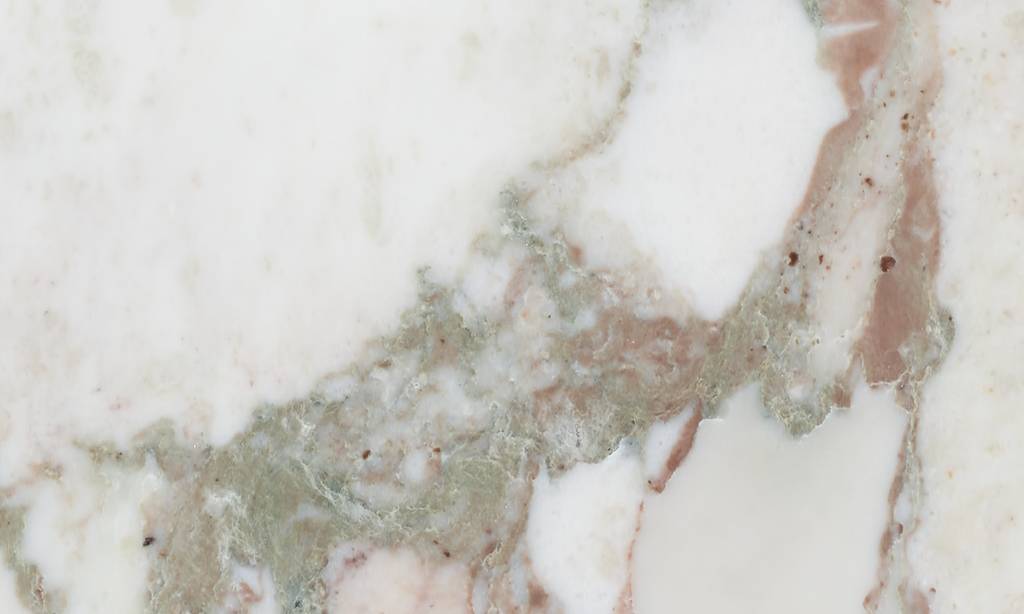} & 
        \includegraphics[width=0.2\linewidth, height=0.12\linewidth]{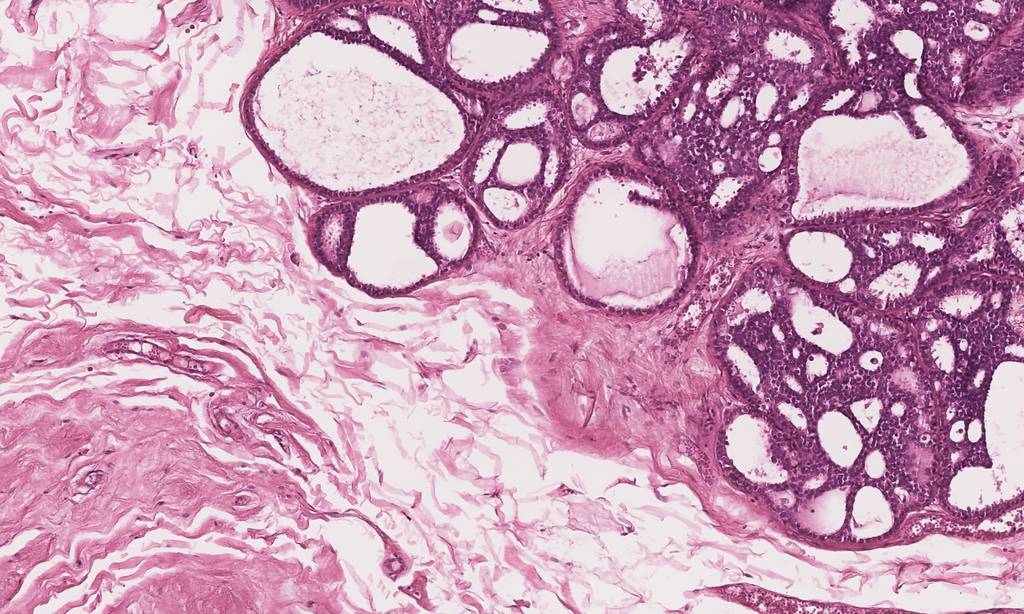} & 
        \includegraphics[width=0.2\linewidth, height=0.12\linewidth]{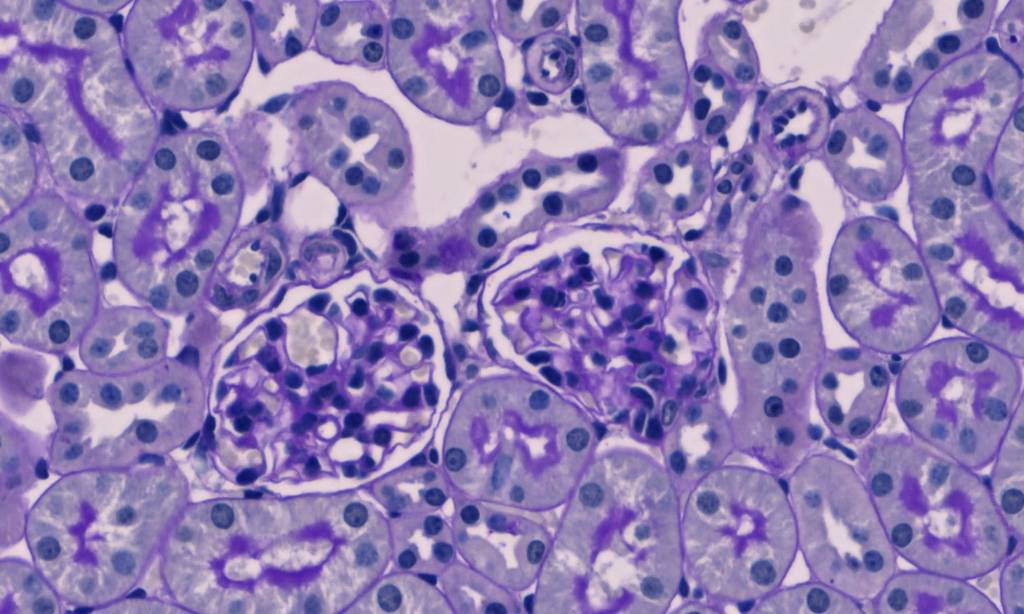} \\

        \rotatebox{90}{ \small \hspace{0.75em} \textbf{NeuRBF}} &
        \includegraphics[width=0.2\linewidth, height=0.12\linewidth]{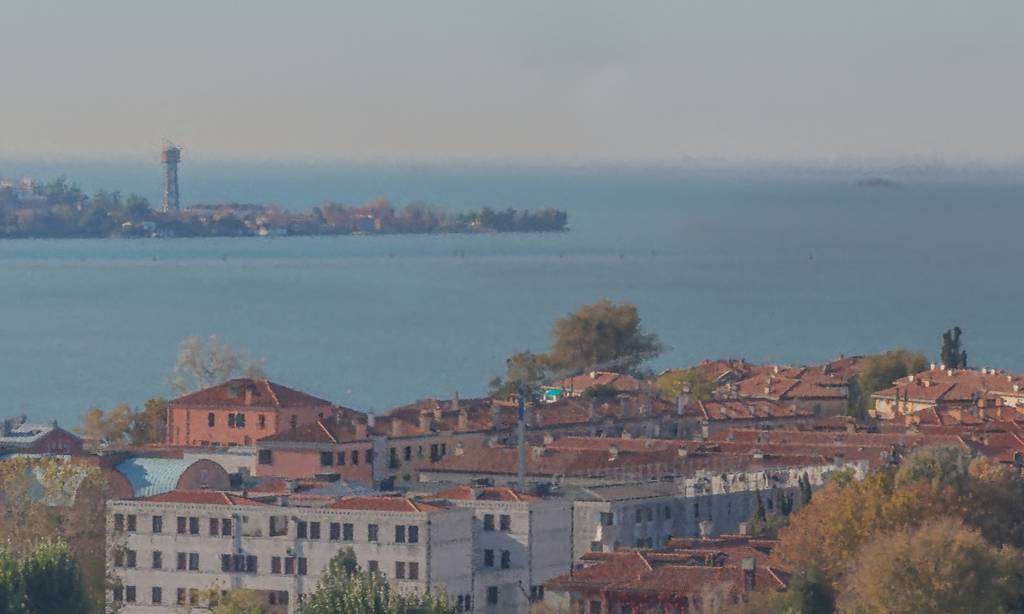} & 
        \includegraphics[width=0.2\linewidth, height=0.12\linewidth]{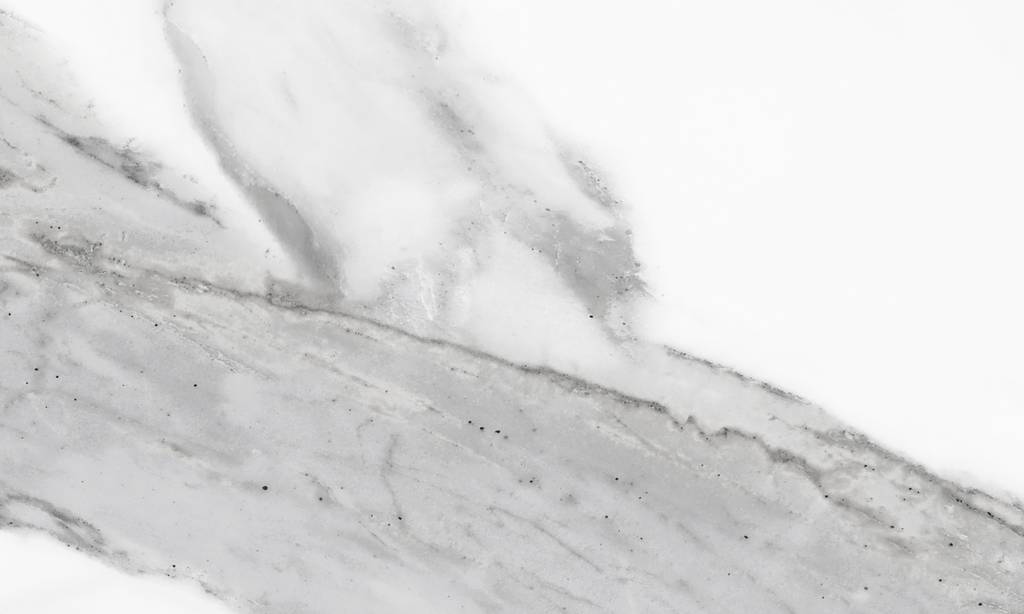} & 
        \includegraphics[width=0.2\linewidth, height=0.12\linewidth]{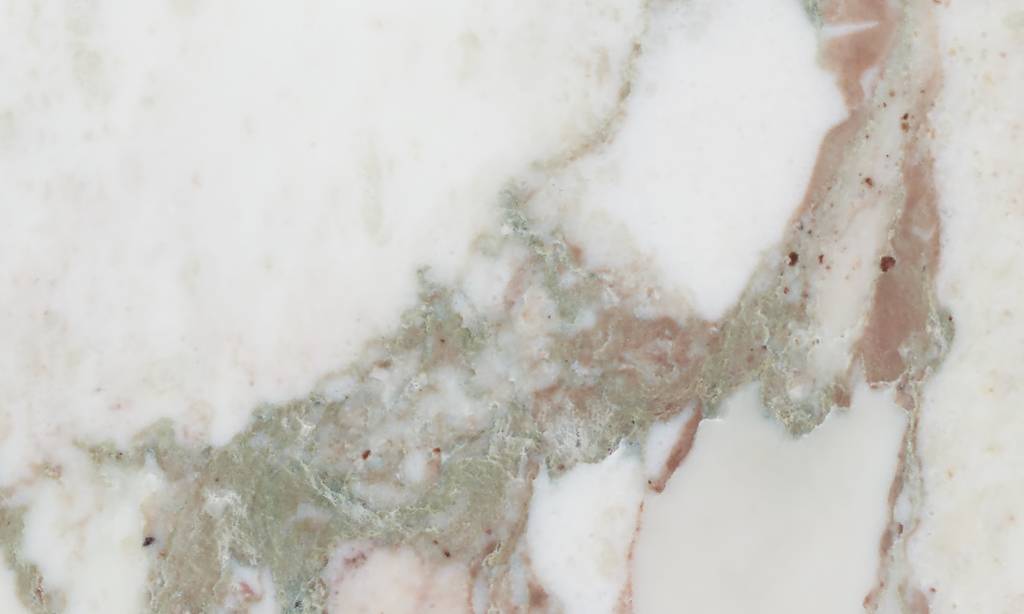} & 
        \includegraphics[width=0.2\linewidth, height=0.12\linewidth]{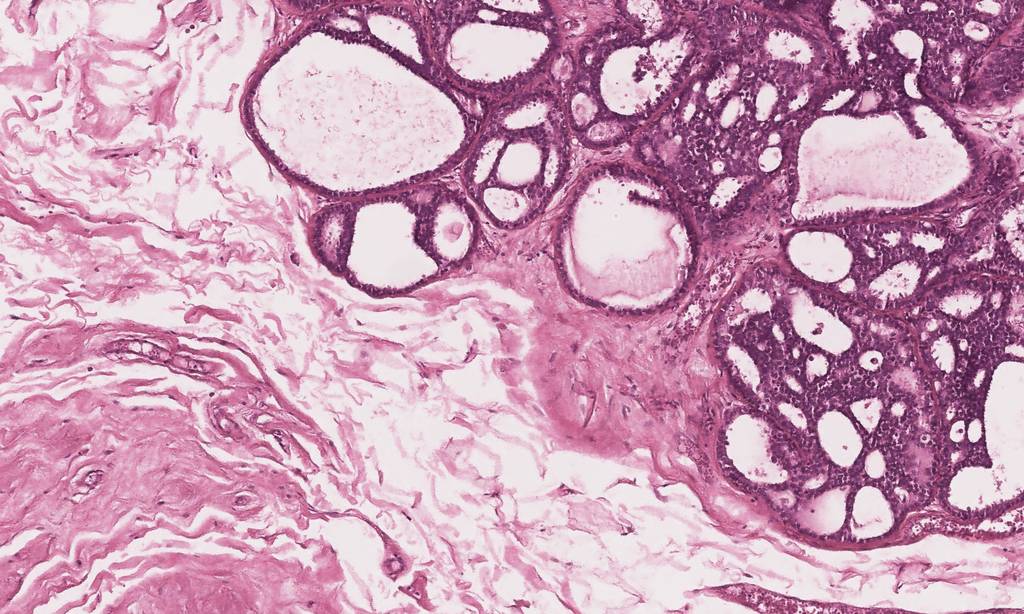} & 
        \includegraphics[width=0.2\linewidth, height=0.12\linewidth]{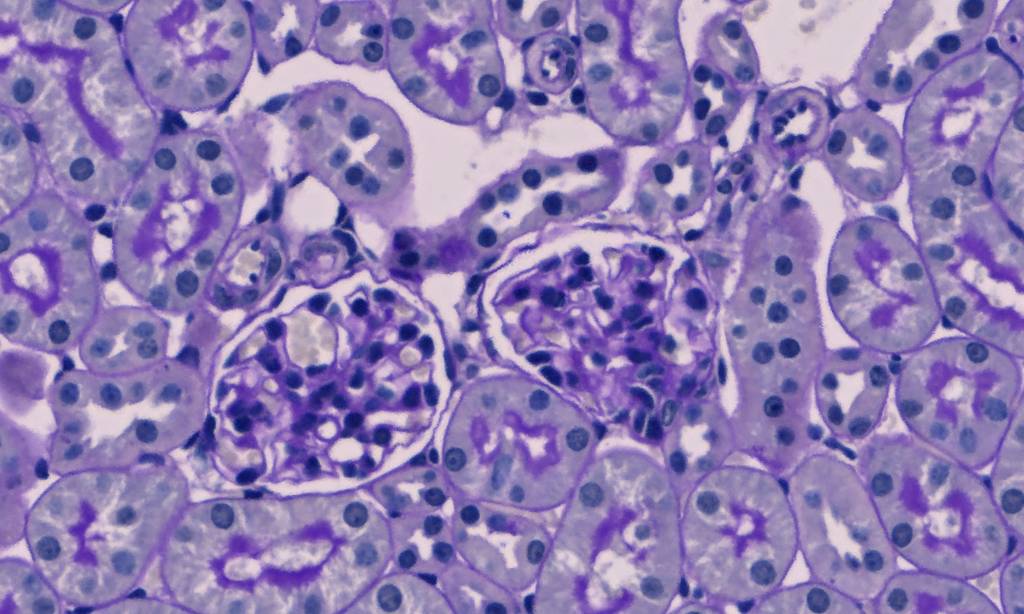} \\

        \rotatebox{90}{ \small \hspace{0.75em} \textbf{MetricGrids}} &
        \includegraphics[width=0.2\linewidth, height=0.12\linewidth]{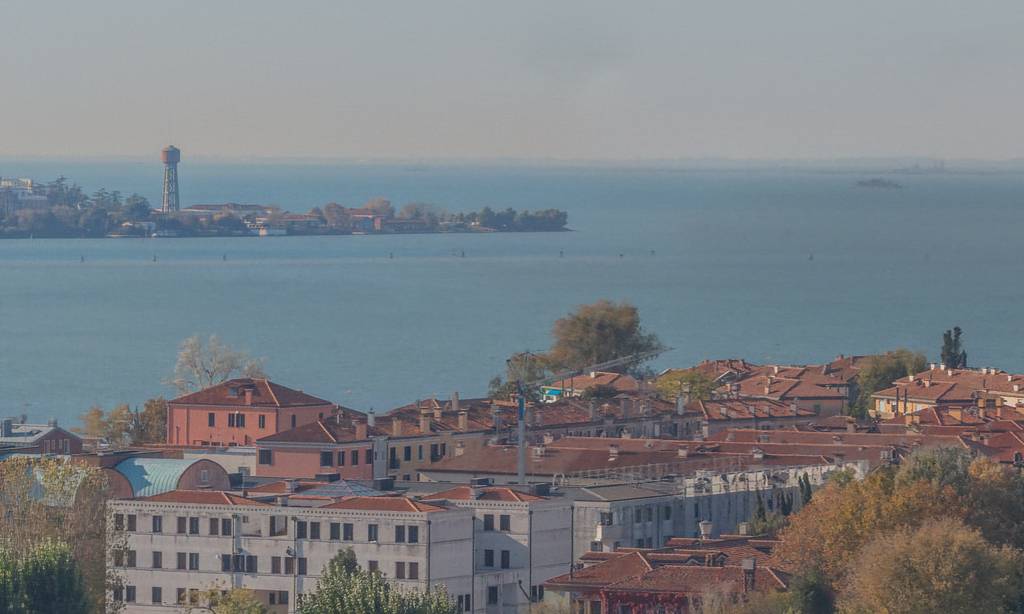} & 
        \includegraphics[width=0.2\linewidth, height=0.12\linewidth]{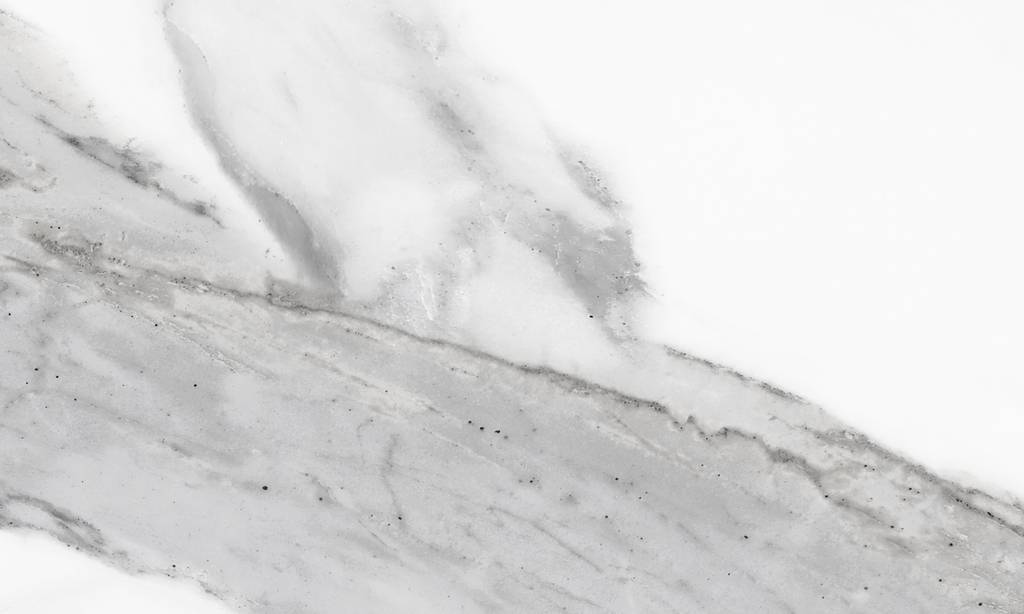} & 
        \includegraphics[width=0.2\linewidth, height=0.12\linewidth]{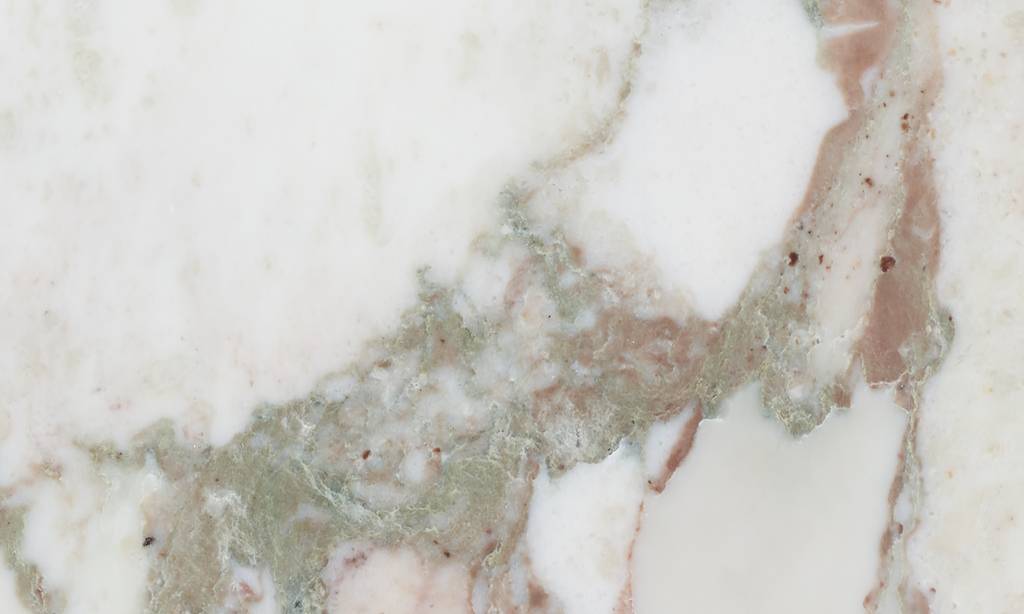} & 
        \includegraphics[width=0.2\linewidth, height=0.12\linewidth]{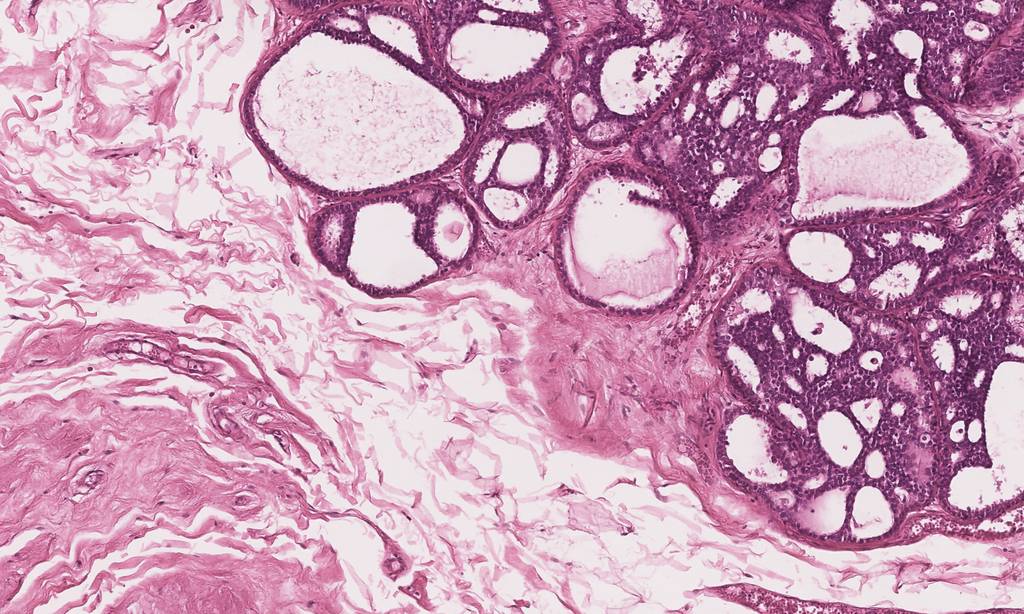} & 
        \includegraphics[width=0.2\linewidth, height=0.12\linewidth]{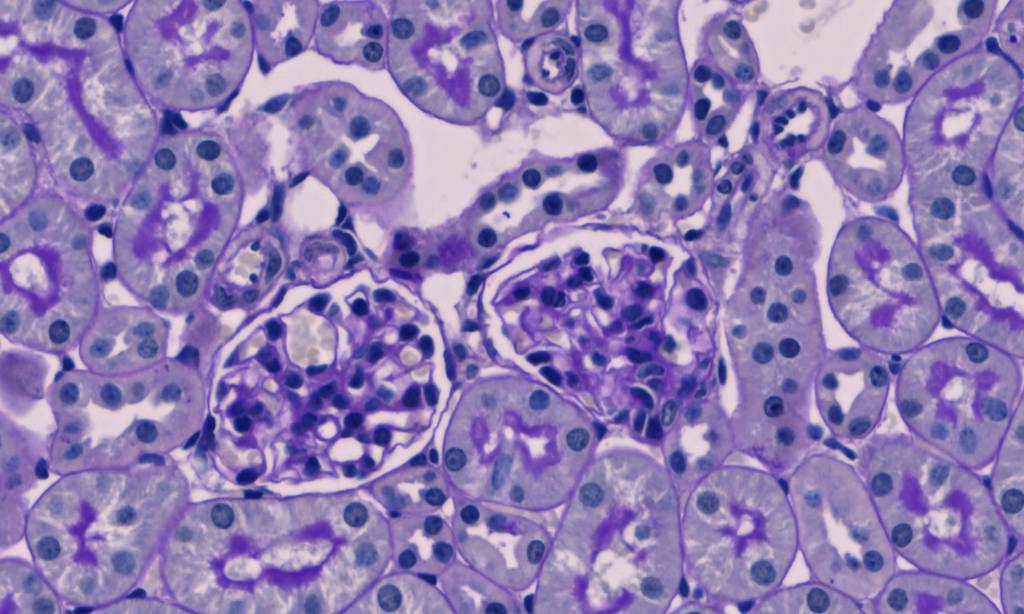} \\

        \rotatebox{90}{ \small \hspace{0.75em} \textbf{GaussImg}} &
        \includegraphics[width=0.2\linewidth, height=0.12\linewidth]{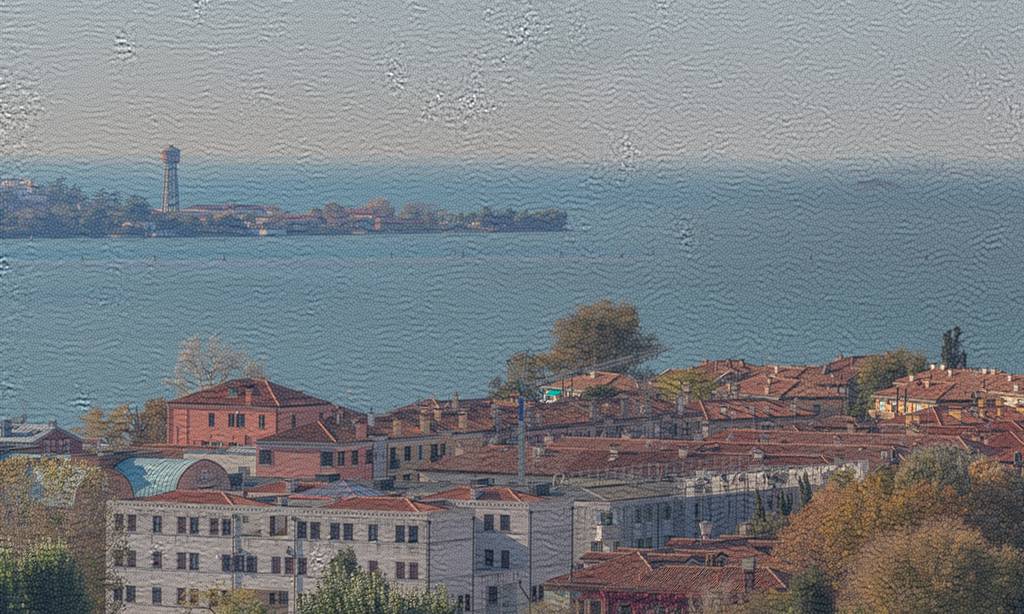} & 
        \includegraphics[width=0.2\linewidth, height=0.12\linewidth]{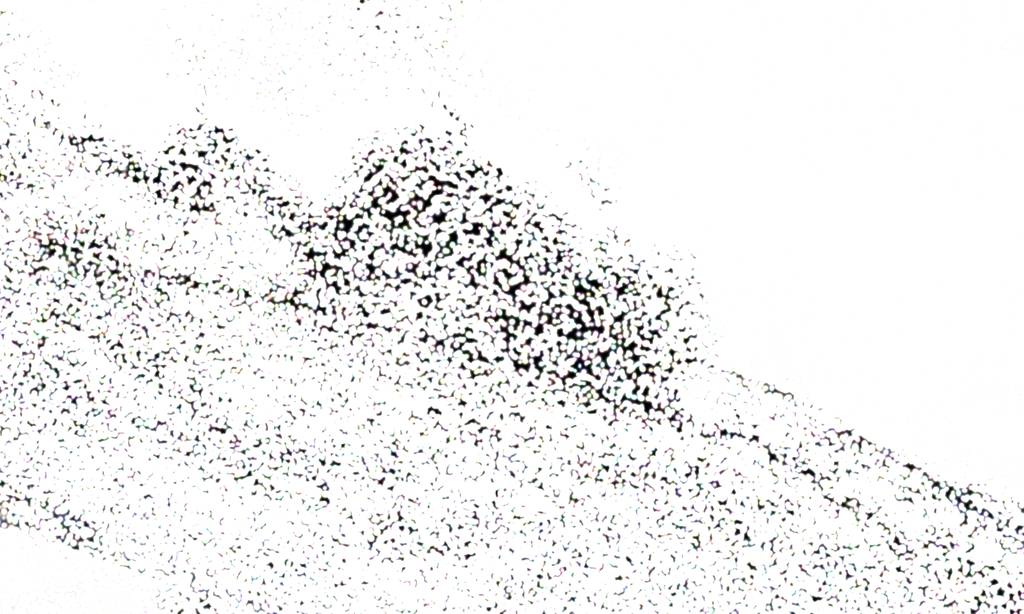} & 
        \includegraphics[width=0.2\linewidth, height=0.12\linewidth]{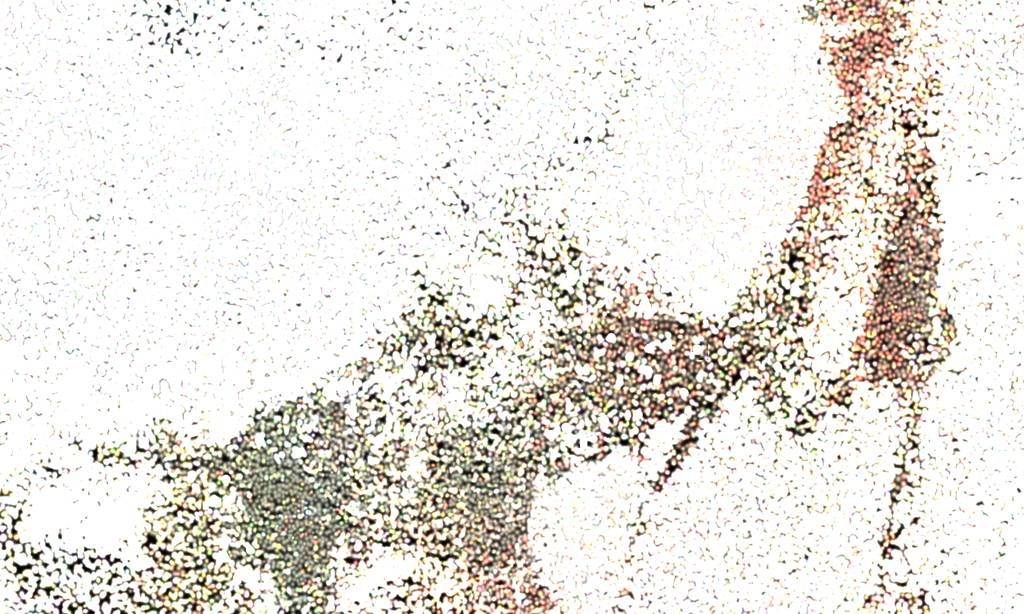} & 
        \includegraphics[width=0.2\linewidth, height=0.12\linewidth]{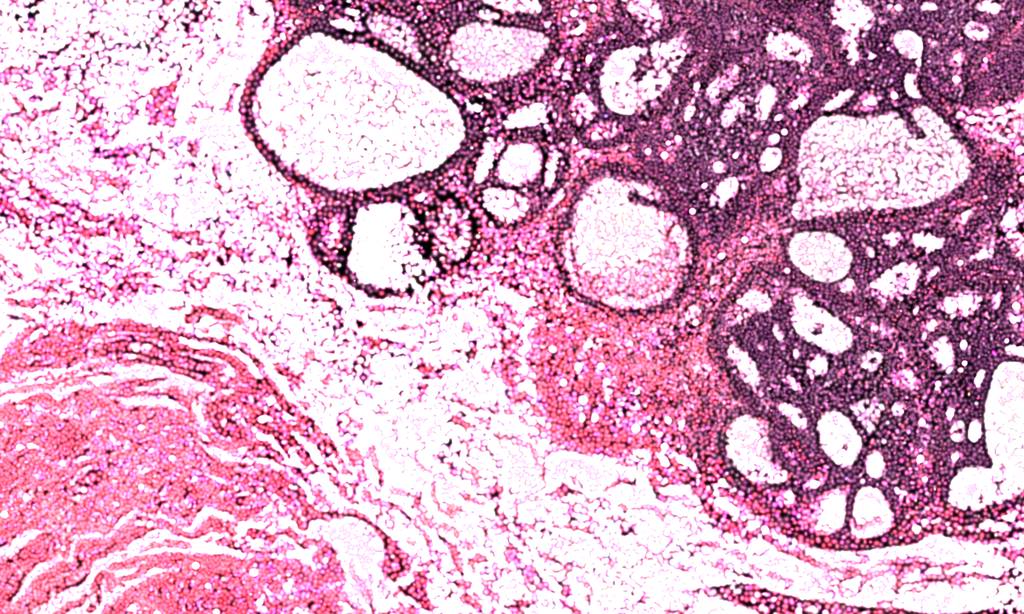} & 
        \includegraphics[width=0.2\linewidth, height=0.12\linewidth]{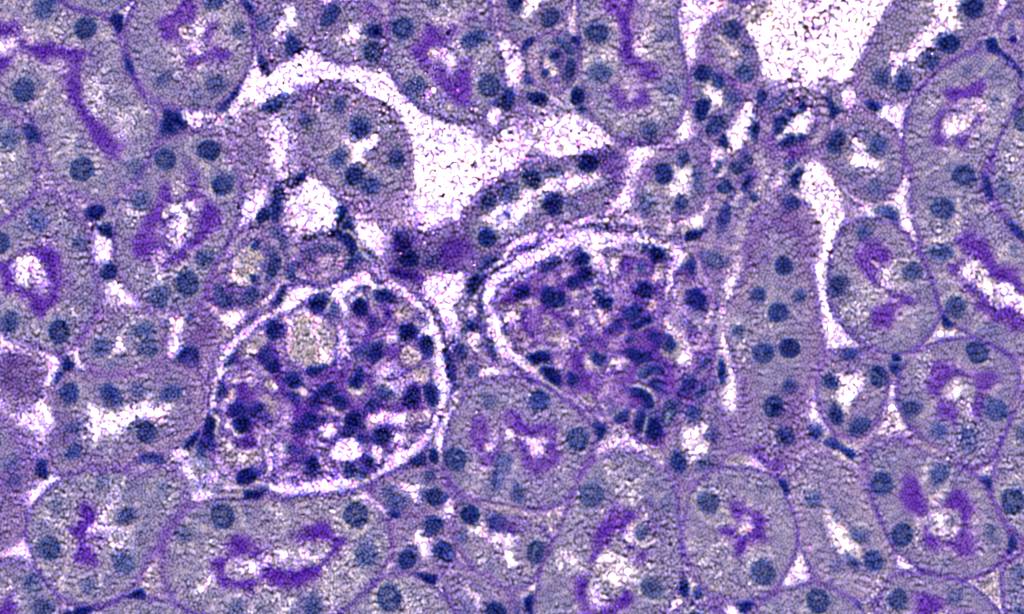} \\

        \rotatebox{90}{ \small \hspace{0.75em} \textbf{SmartSplat}} &
        \includegraphics[width=0.2\linewidth, height=0.12\linewidth]{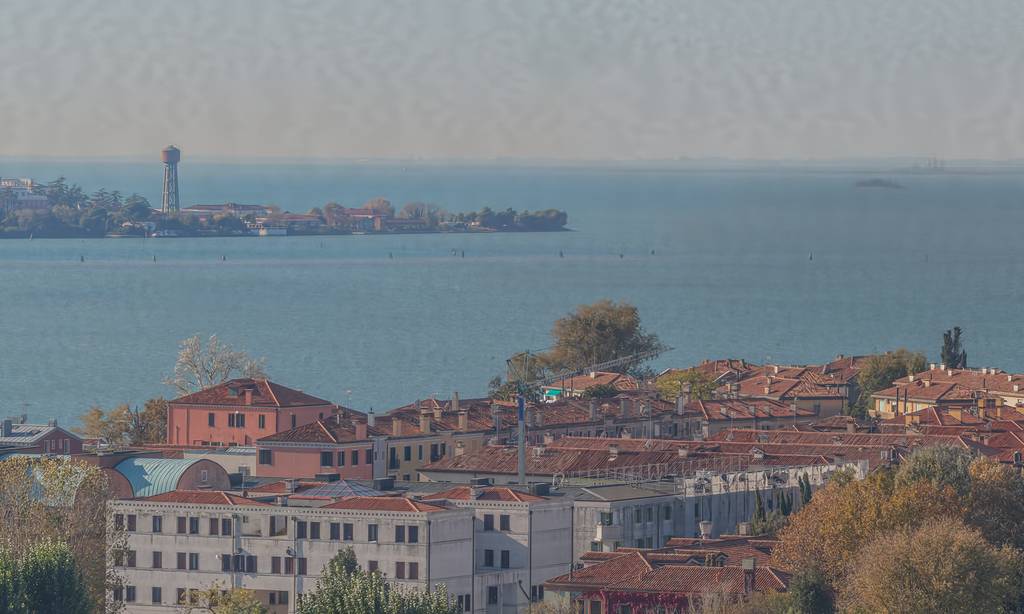} & 
        \includegraphics[width=0.2\linewidth, height=0.12\linewidth]{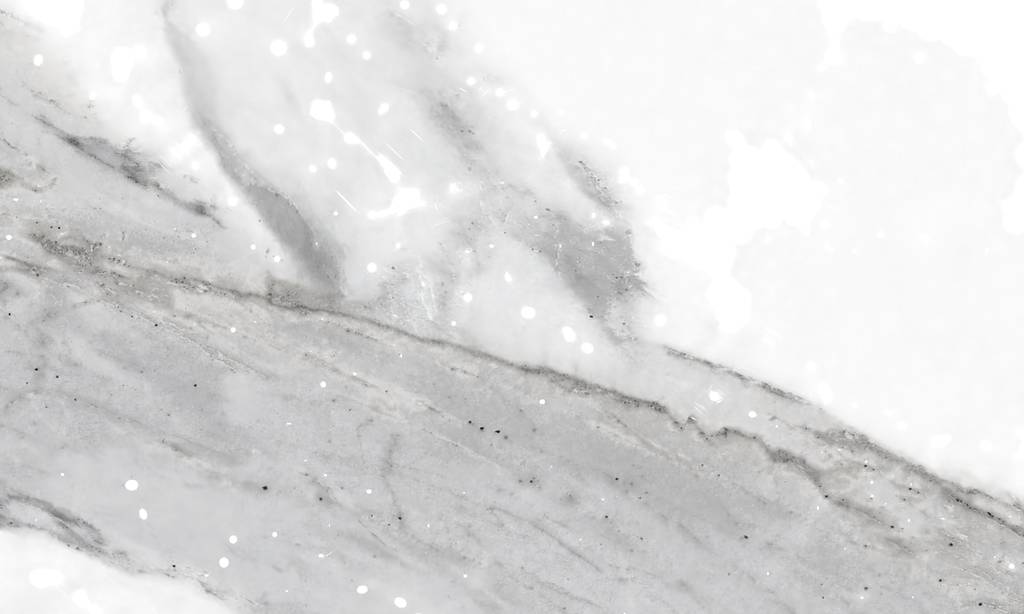} & 
        \includegraphics[width=0.2\linewidth, height=0.12\linewidth]{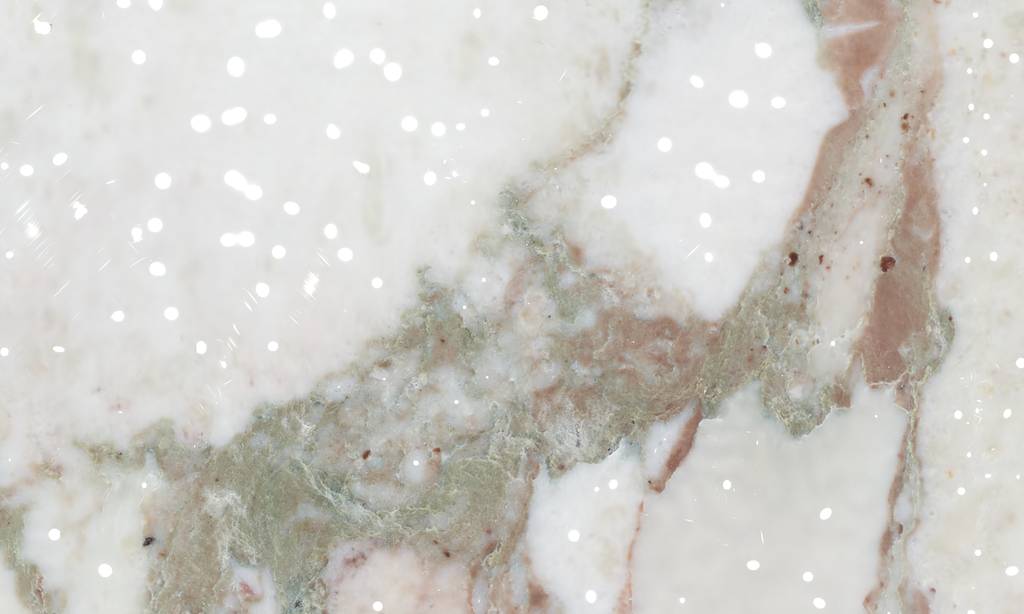} & 
        \includegraphics[width=0.2\linewidth, height=0.12\linewidth]{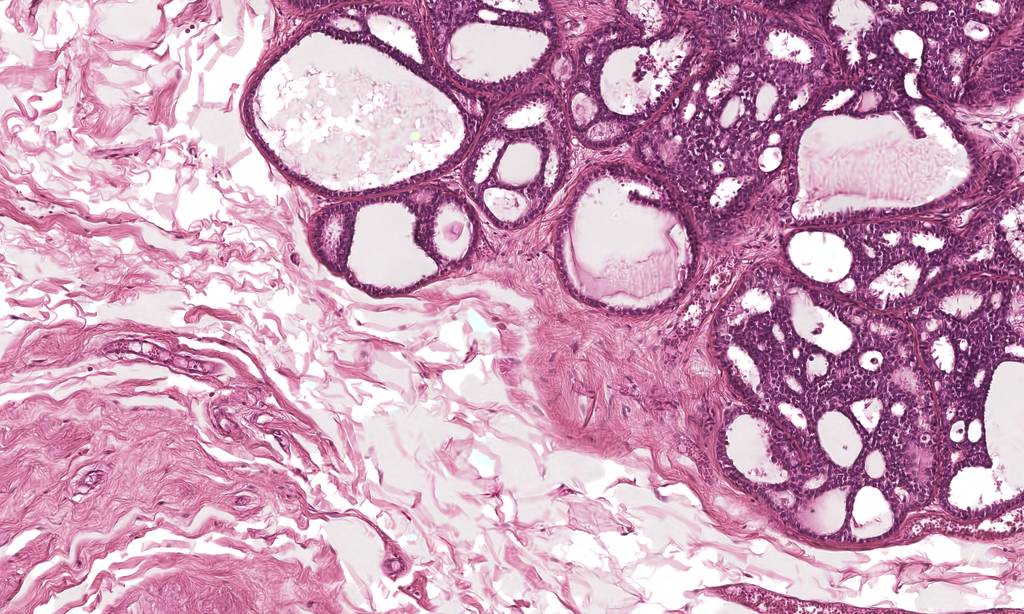} & 
        \includegraphics[width=0.2\linewidth, height=0.12\linewidth]{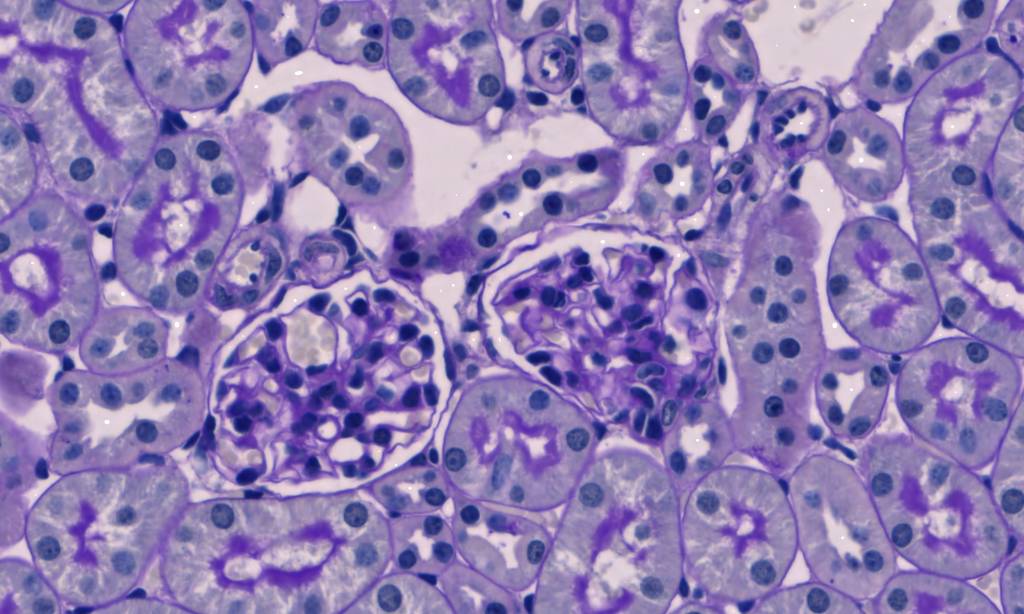} \\

    \end{tabular}}
    \caption{
    \textbf{Additional fitting qualitatives results on \dataname{}.}
    We show $1228 \times 2048$ crops from the full resolution images.
    }
    \label{fig:suppl_fit_qualitatives_1}

\end{figure*}

\begin{figure*}[ht]
    \centering
    \resizebox{\linewidth}{!}{
    \setlength{\tabcolsep}{1px}
    \begin{tabular}{l ccccc}
        & \textit{Munich} & \textit{Andromeda}& \textit{Italy} & \textit{Lynxsdesign} & \textit{House Of Chores} \\
        \rotatebox{90}{  \hspace{2em} \textbf{GT}} &
        \includegraphics[width=0.2\linewidth, height=0.12\linewidth]{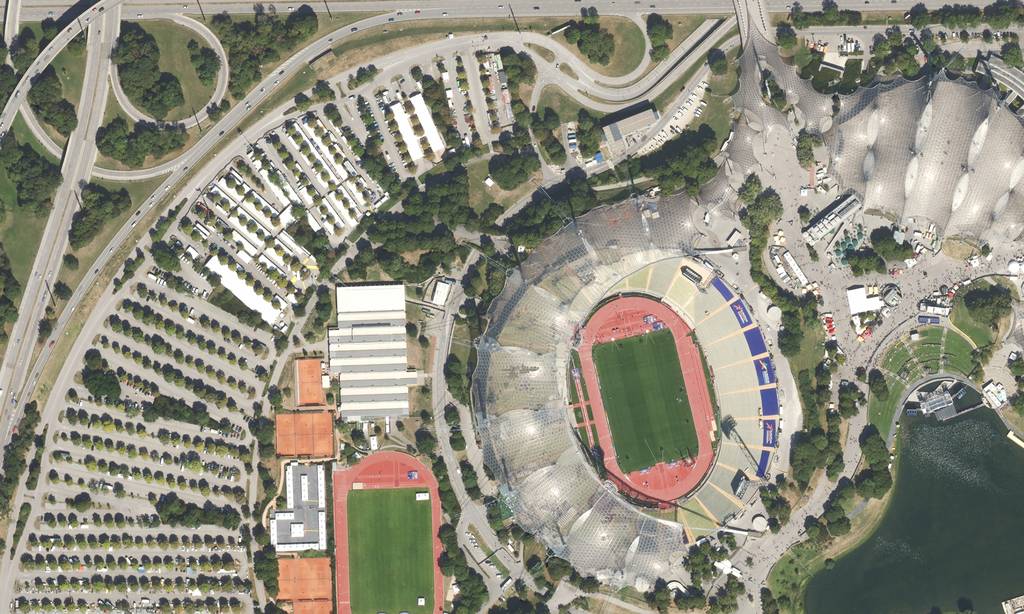} & 
        \includegraphics[width=0.2\linewidth, height=0.12\linewidth]{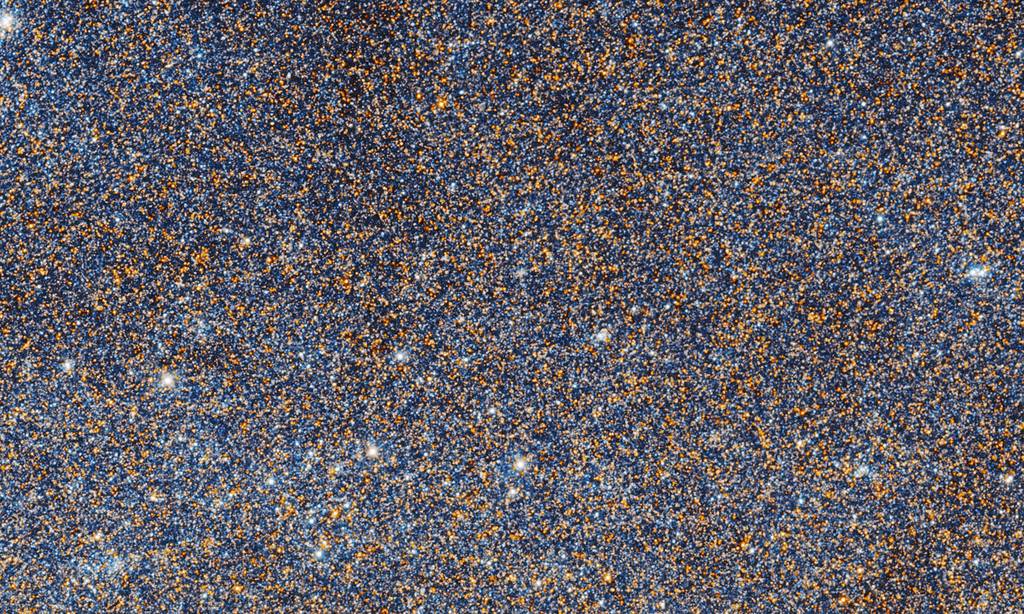} & 
        \includegraphics[width=0.2\linewidth, height=0.12\linewidth]{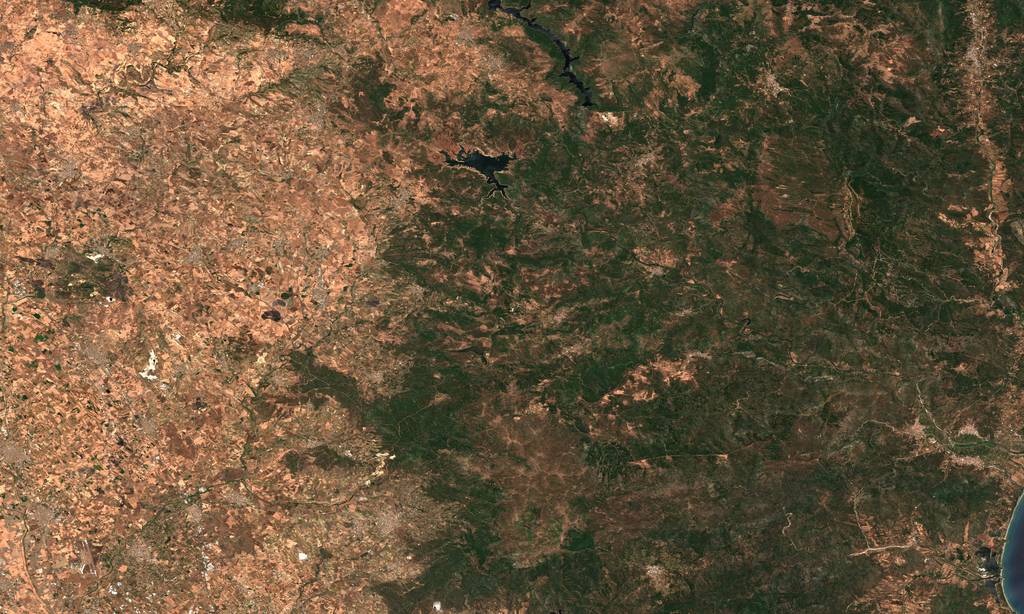} & 
        \includegraphics[width=0.2\linewidth, height=0.12\linewidth]{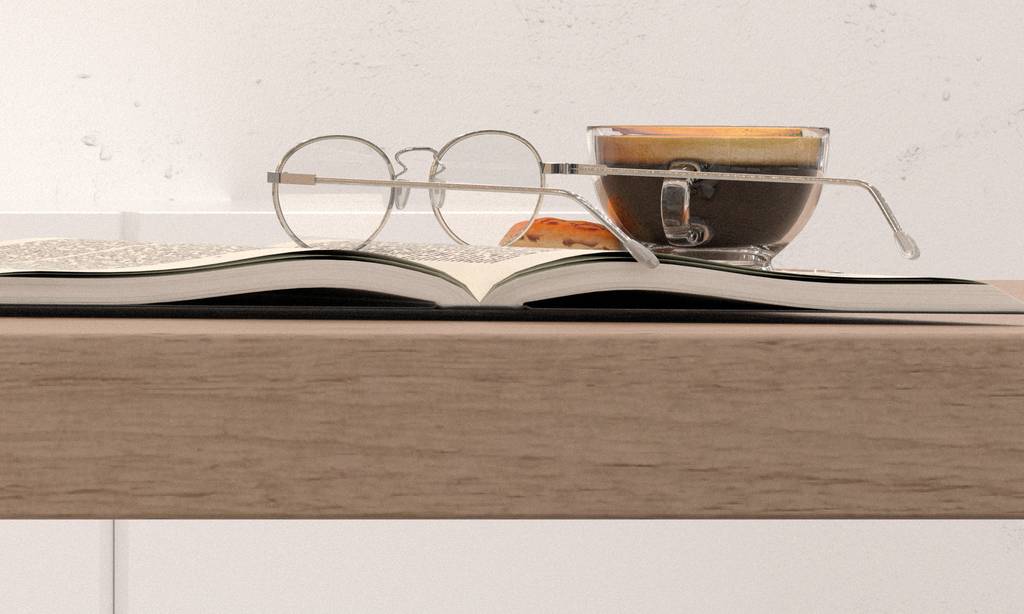} & 
        \includegraphics[width=0.2\linewidth, height=0.12\linewidth]{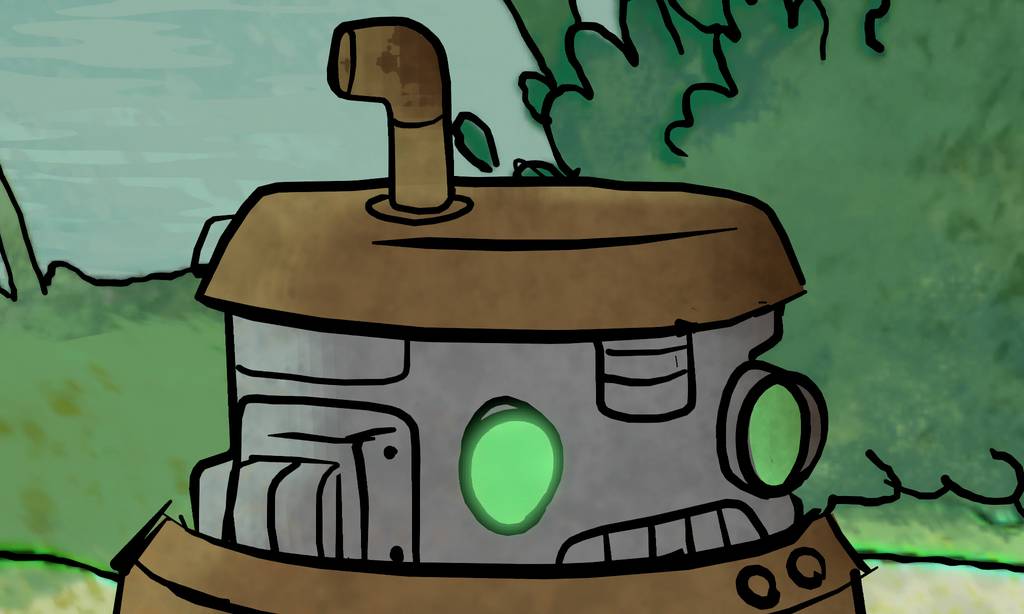} \\

        \rotatebox{90}{ \small \hspace{0.75em} \textbf{SIREN}} &
        \includegraphics[width=0.2\linewidth, height=0.12\linewidth]{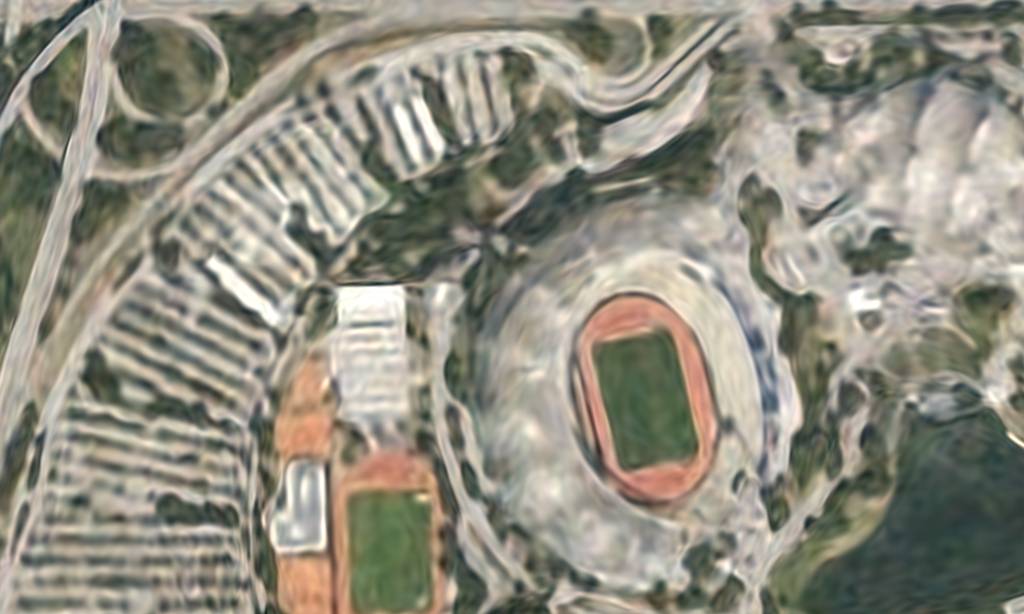} & 
        \includegraphics[width=0.2\linewidth, height=0.12\linewidth]{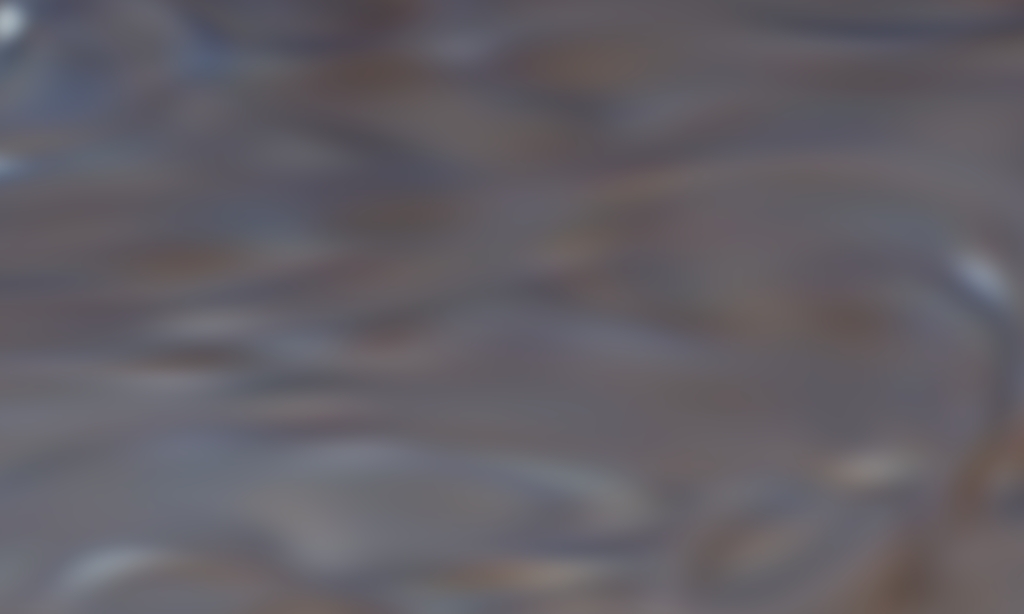} & 
        \includegraphics[width=0.2\linewidth, height=0.12\linewidth]{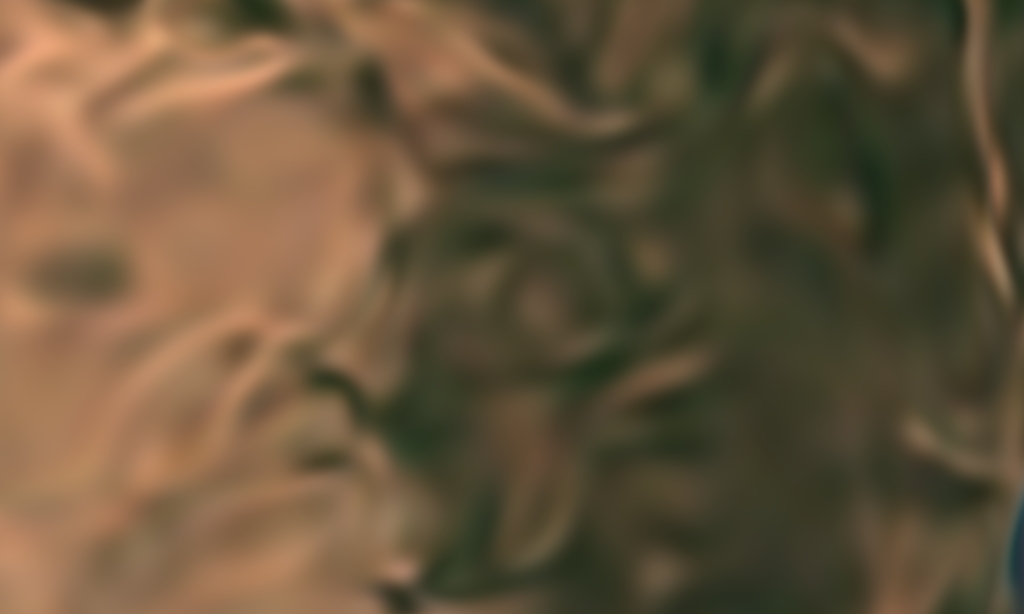} & 
        \includegraphics[width=0.2\linewidth, height=0.12\linewidth]{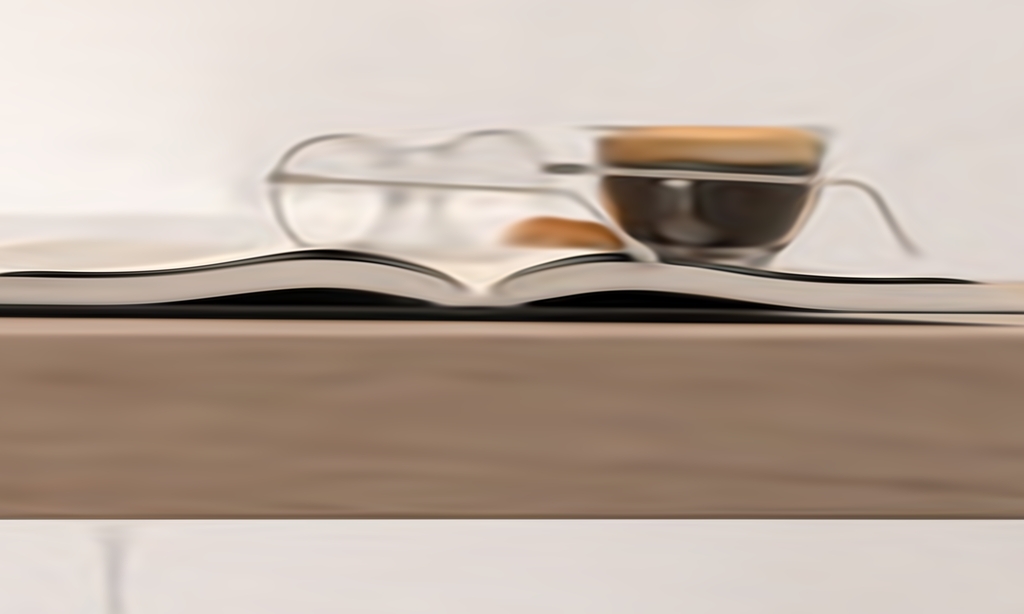} & 
        \includegraphics[width=0.2\linewidth, height=0.12\linewidth]{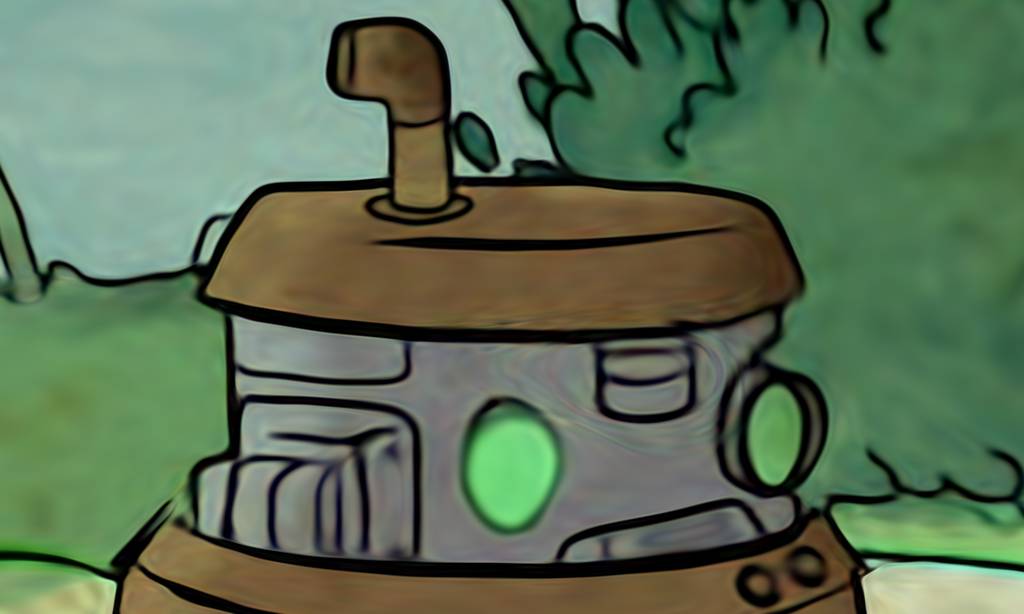} \\

        \rotatebox{90}{ \small \hspace{0.75em} \textbf{FDHO}} &
        \includegraphics[width=0.2\linewidth, height=0.12\linewidth]{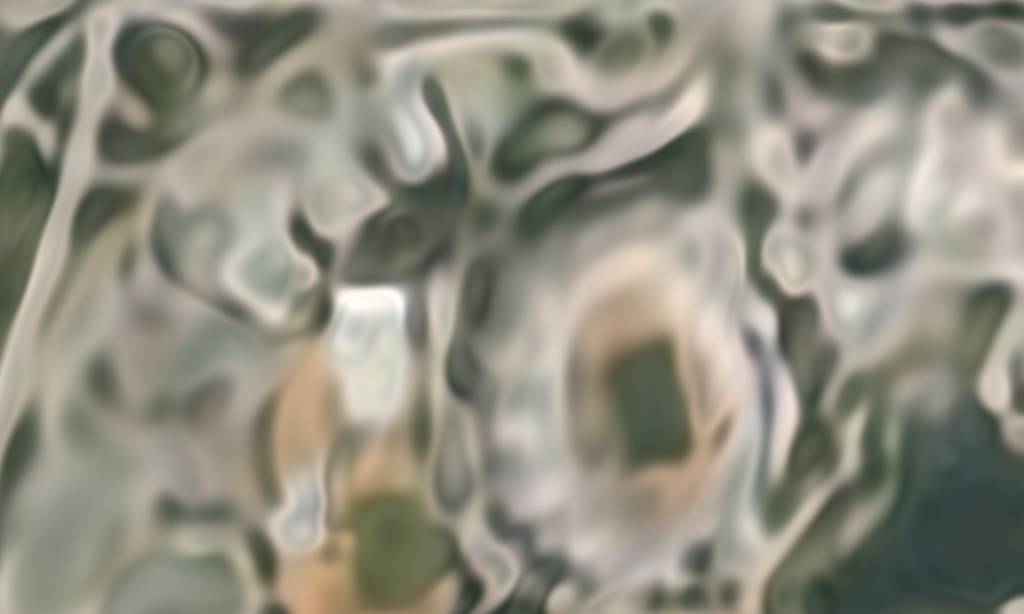} & 
        \includegraphics[width=0.2\linewidth, height=0.12\linewidth]{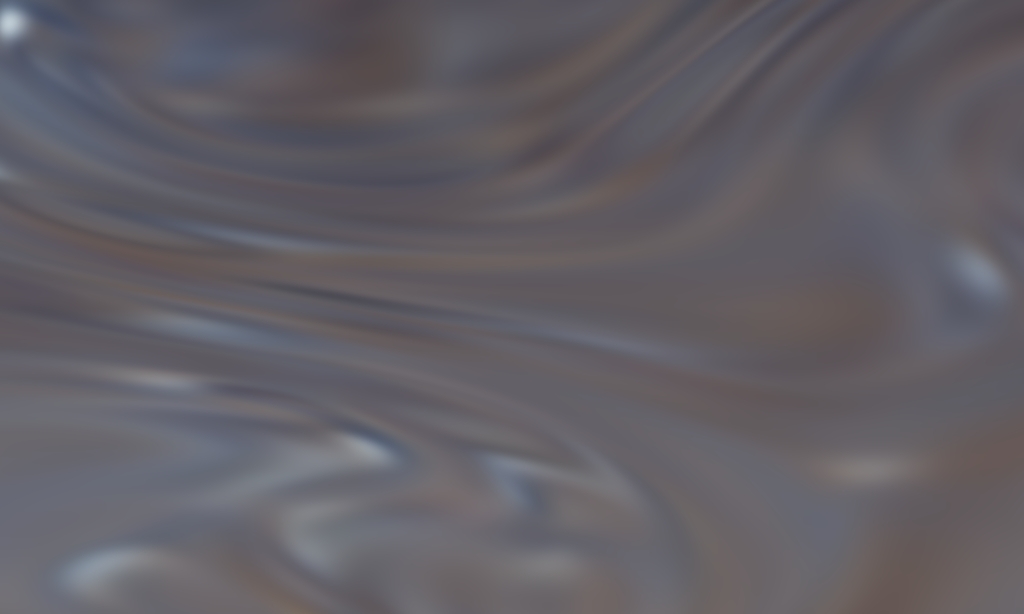} & 
        \includegraphics[width=0.2\linewidth, height=0.12\linewidth]{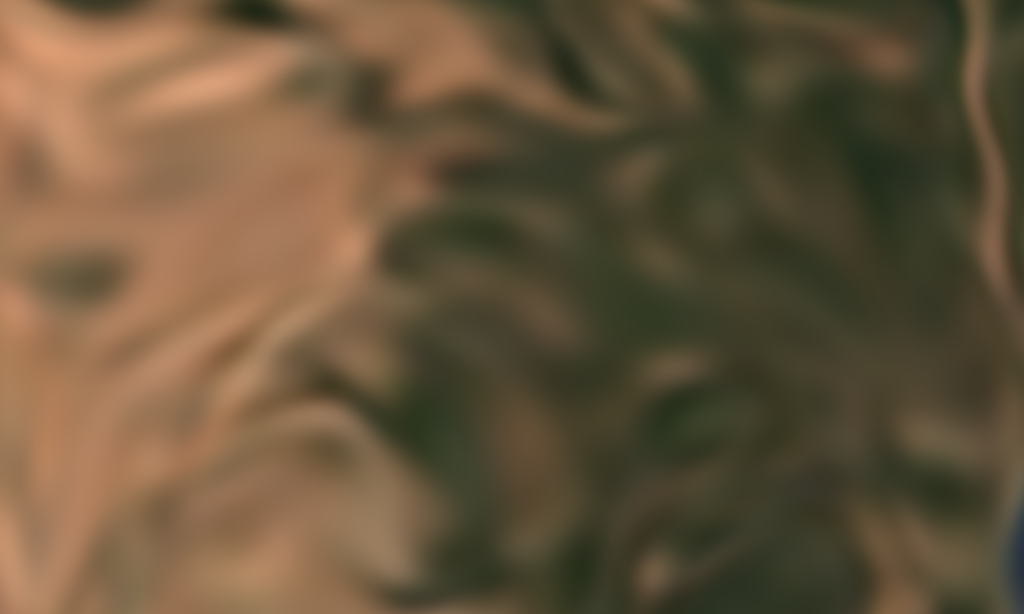} & 
        \includegraphics[width=0.2\linewidth, height=0.12\linewidth]{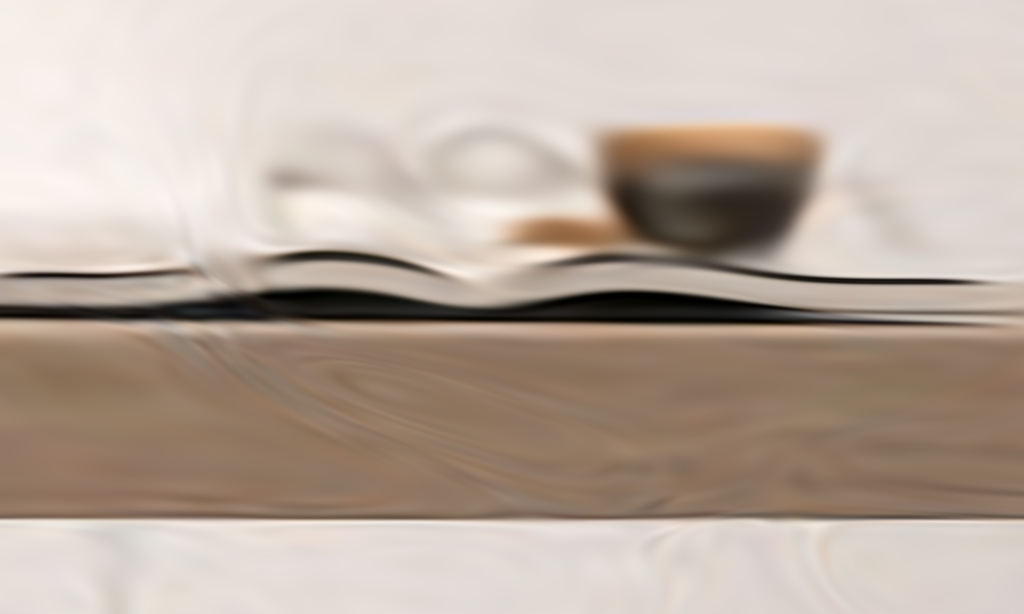} & 
        \includegraphics[width=0.2\linewidth, height=0.12\linewidth]{images/qualitatives_fitting_suppl/fit_HouseOfChores_gt.jpg} \\

        \rotatebox{90}{ \small \hspace{0.75em} \textbf{FINER}} &
        \includegraphics[width=0.2\linewidth, height=0.12\linewidth]{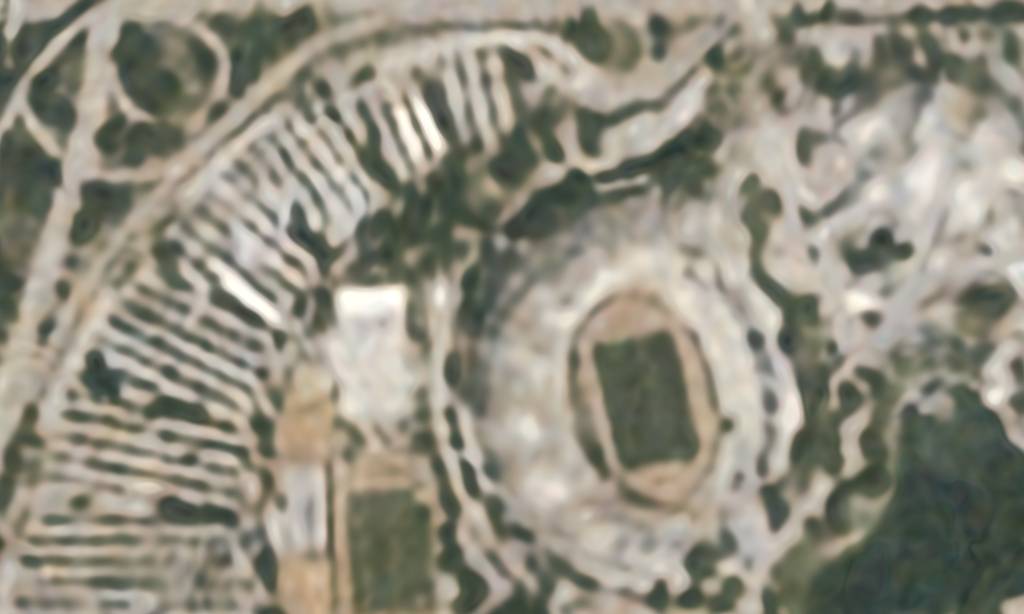} & 
        \includegraphics[width=0.2\linewidth, height=0.12\linewidth]{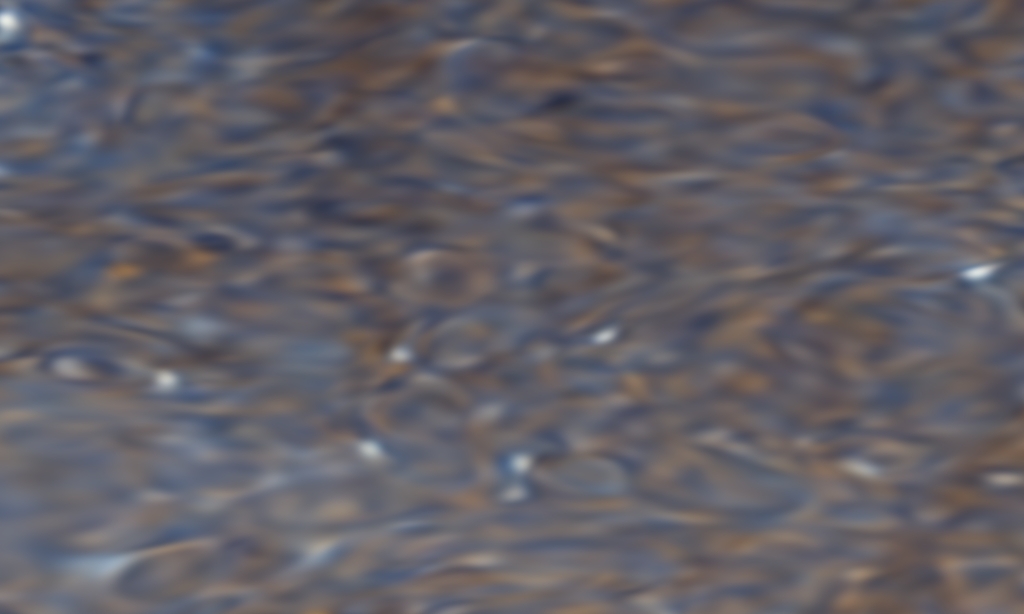} & 
        \includegraphics[width=0.2\linewidth, height=0.12\linewidth]{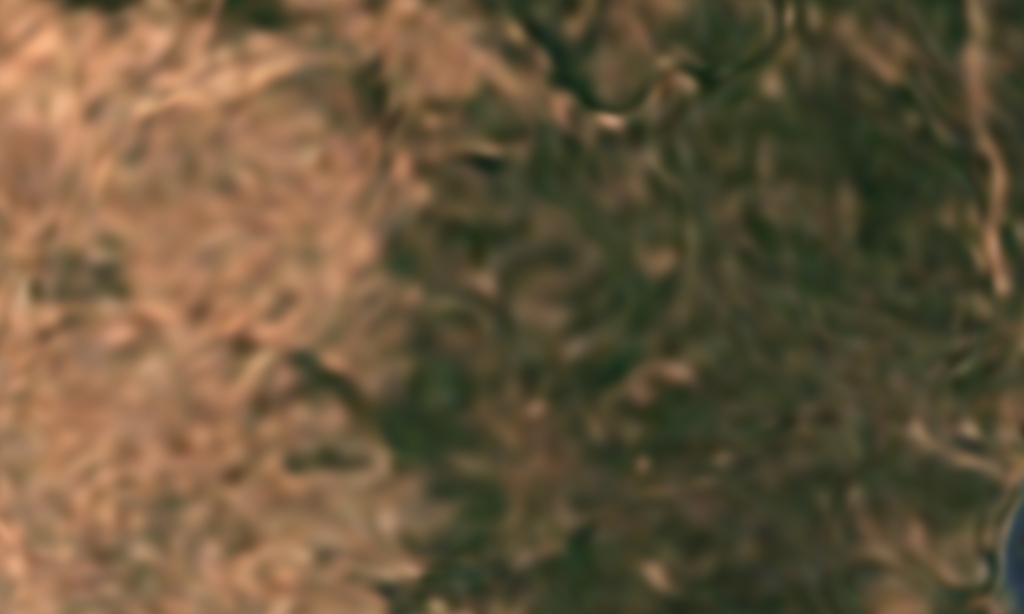} & 
        \includegraphics[width=0.2\linewidth, height=0.12\linewidth]{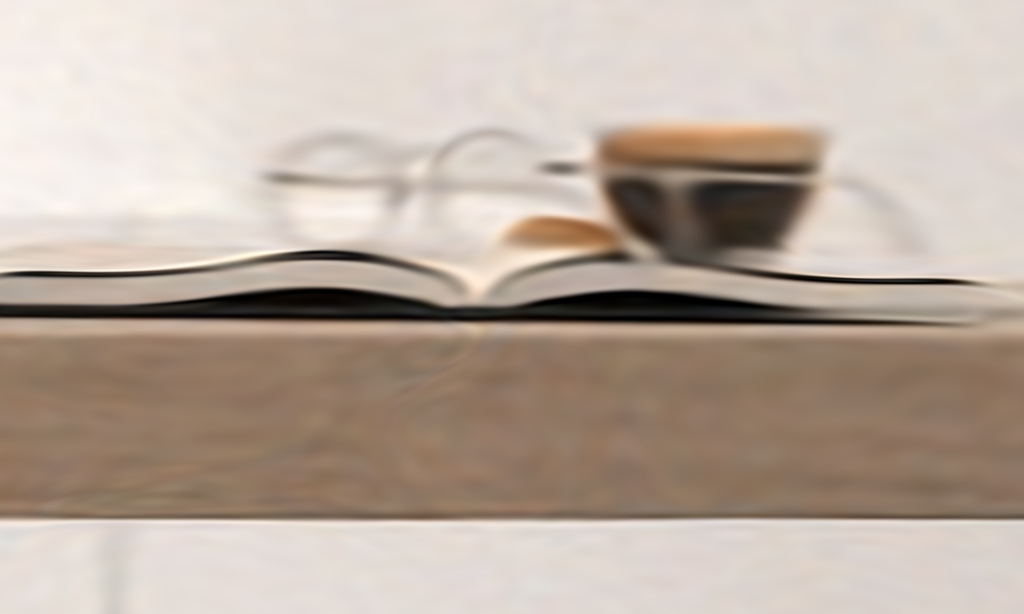} & 
        \includegraphics[width=0.2\linewidth, height=0.12\linewidth]{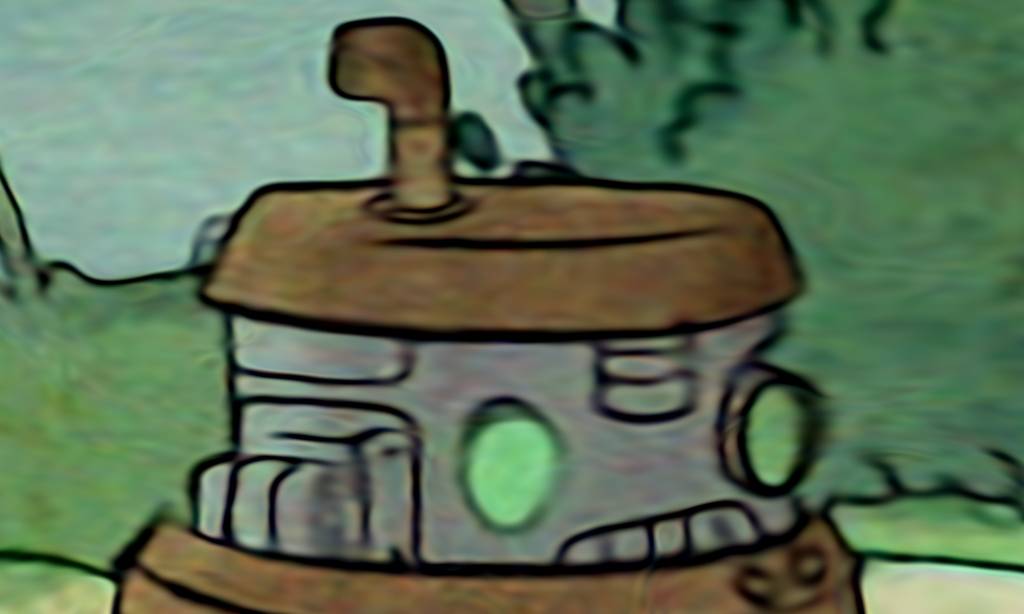} \\

        \rotatebox{90}{ \small \hspace{0.75em} \textbf{I-NGP}} &
        \includegraphics[width=0.2\linewidth, height=0.12\linewidth]{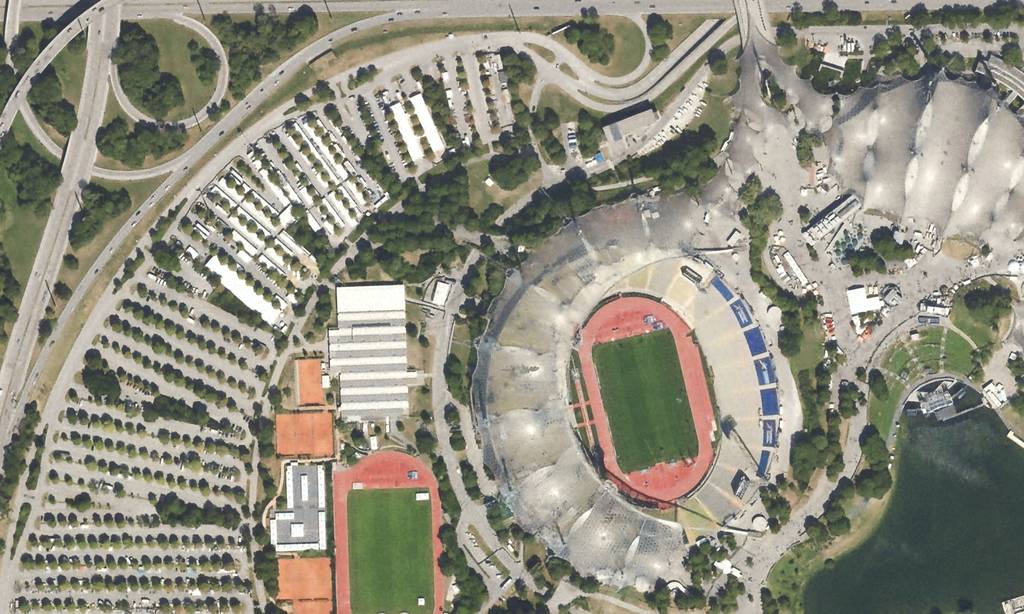} & 
        \includegraphics[width=0.2\linewidth, height=0.12\linewidth]{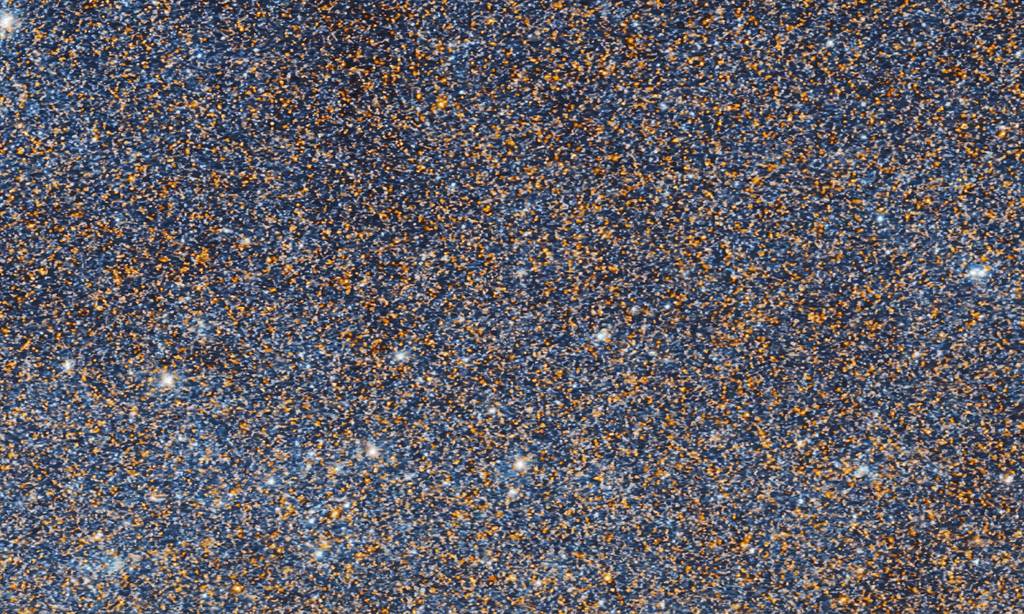} & 
        \includegraphics[width=0.2\linewidth, height=0.12\linewidth]{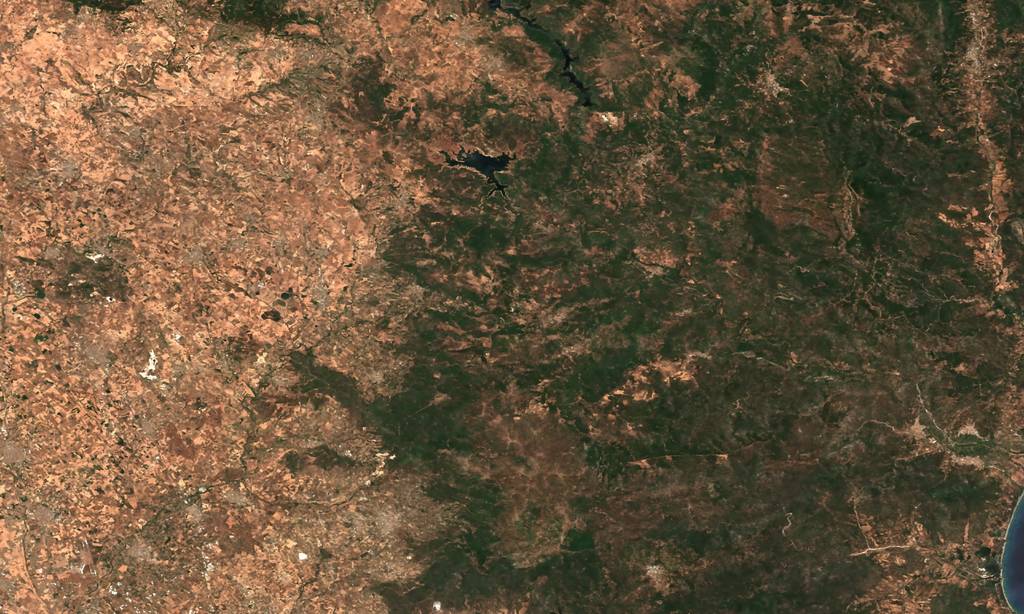} & 
        \includegraphics[width=0.2\linewidth, height=0.12\linewidth]{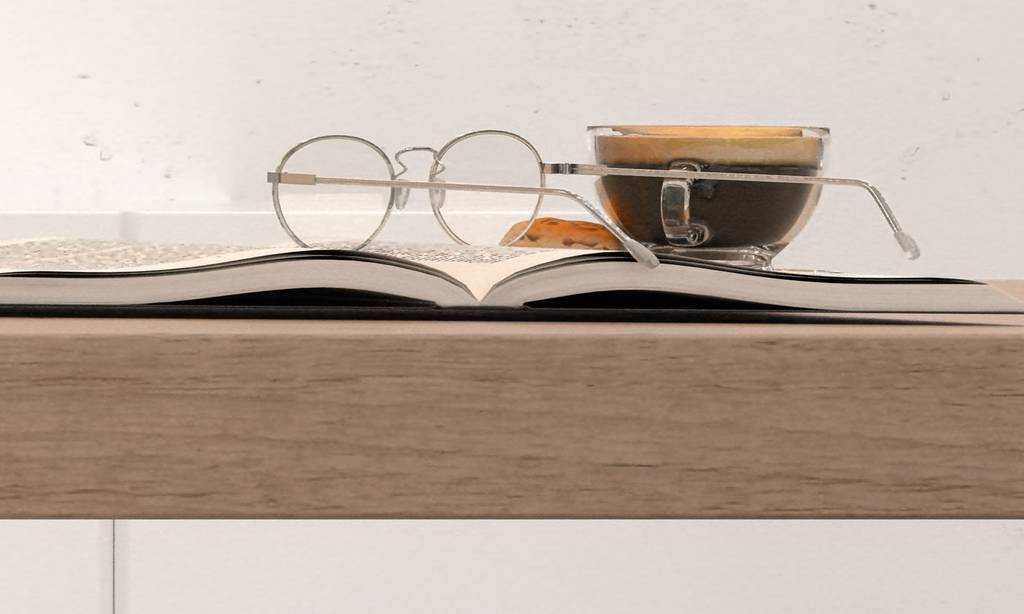} & 
        \includegraphics[width=0.2\linewidth, height=0.12\linewidth]{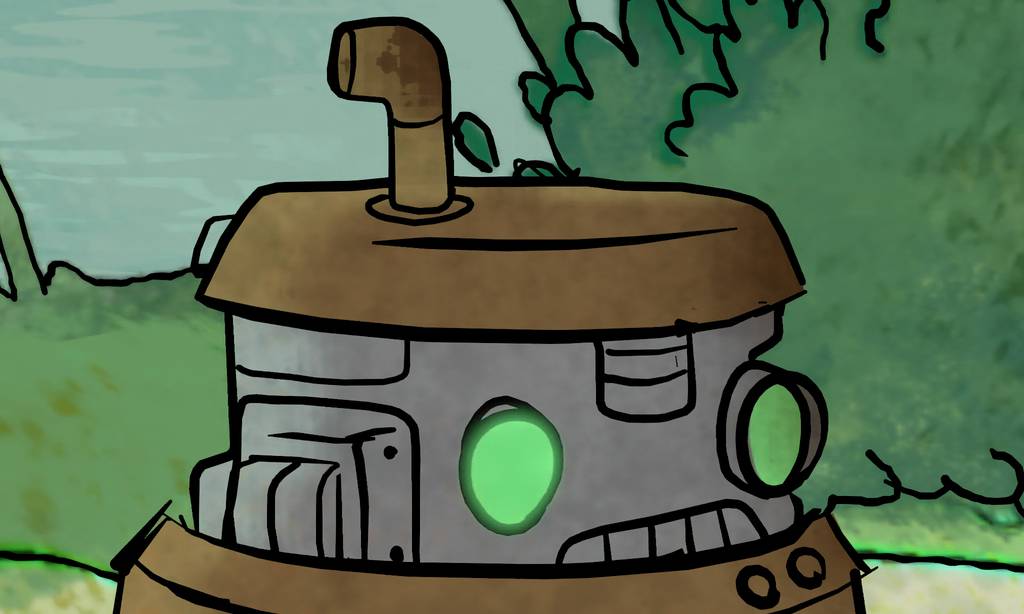} \\

        \rotatebox{90}{ \small \hspace{0.75em} \textbf{NeuRBF}} &
        \includegraphics[width=0.2\linewidth, height=0.12\linewidth]{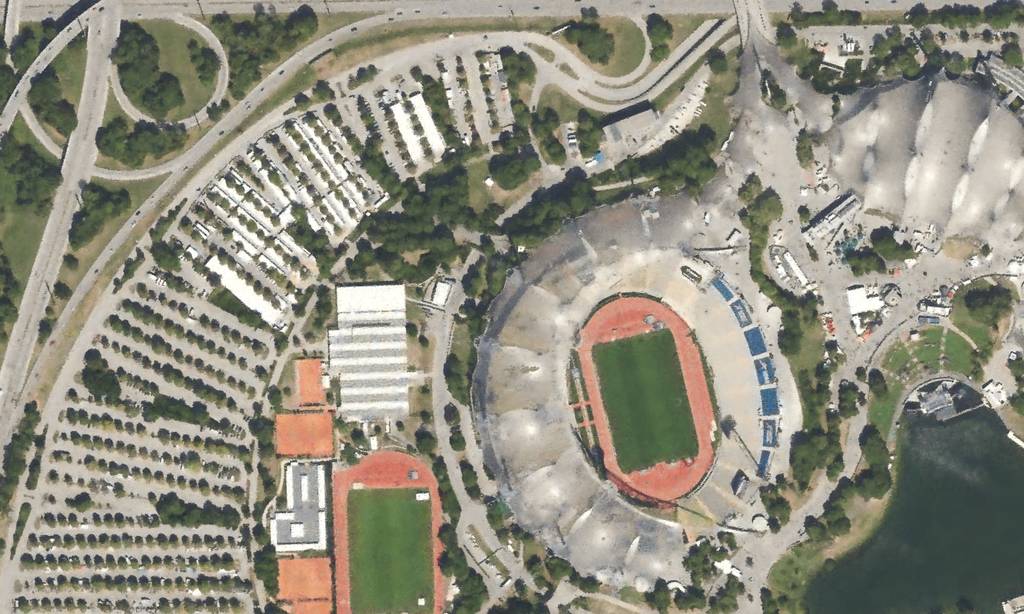} & 
        \includegraphics[width=0.2\linewidth, height=0.12\linewidth]{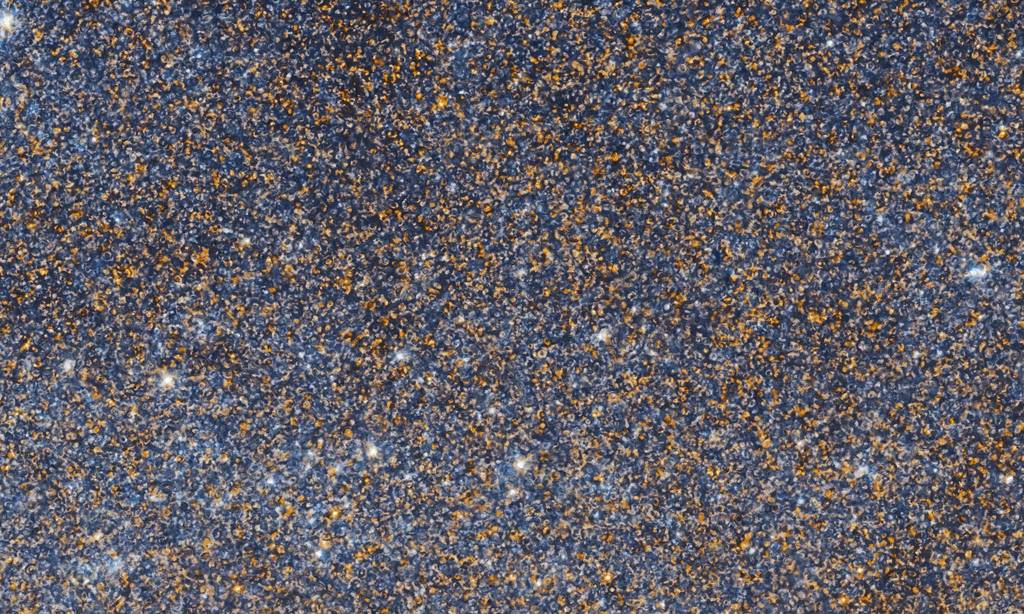} & 
        \includegraphics[width=0.2\linewidth, height=0.12\linewidth]{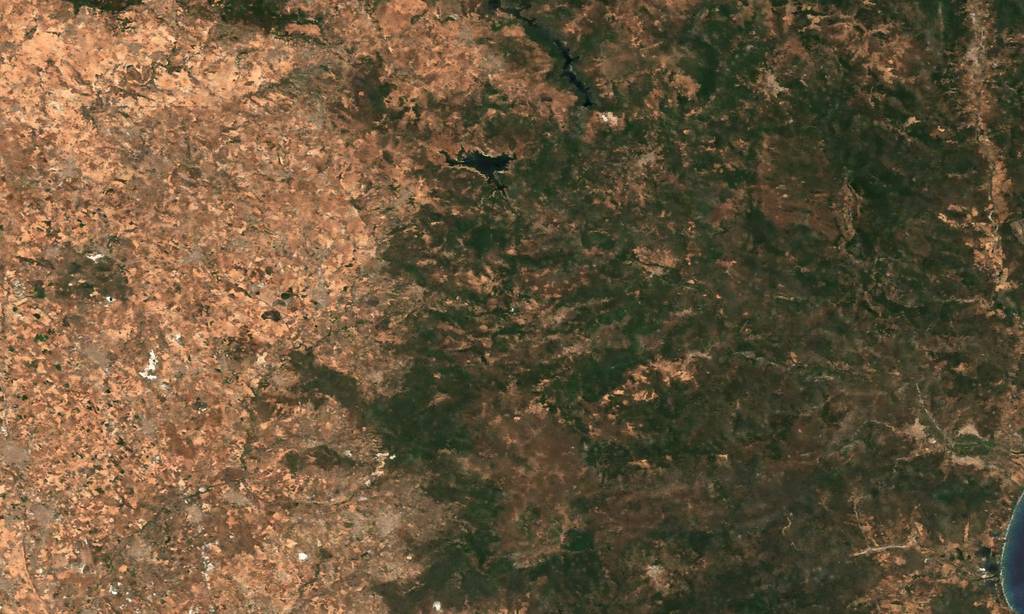} & 
        \includegraphics[width=0.2\linewidth, height=0.12\linewidth]{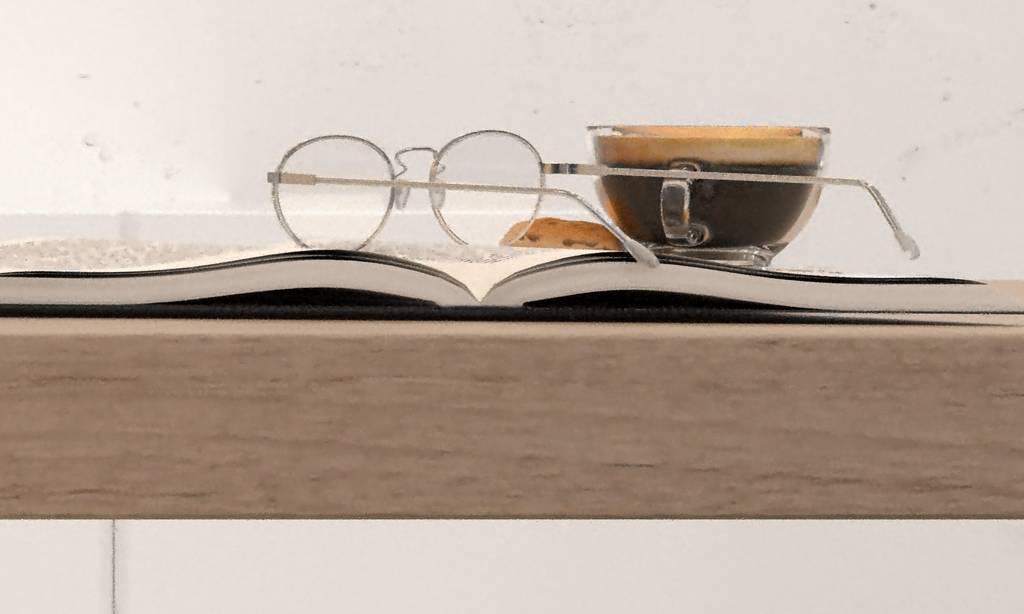} & 
        \includegraphics[width=0.2\linewidth, height=0.12\linewidth]{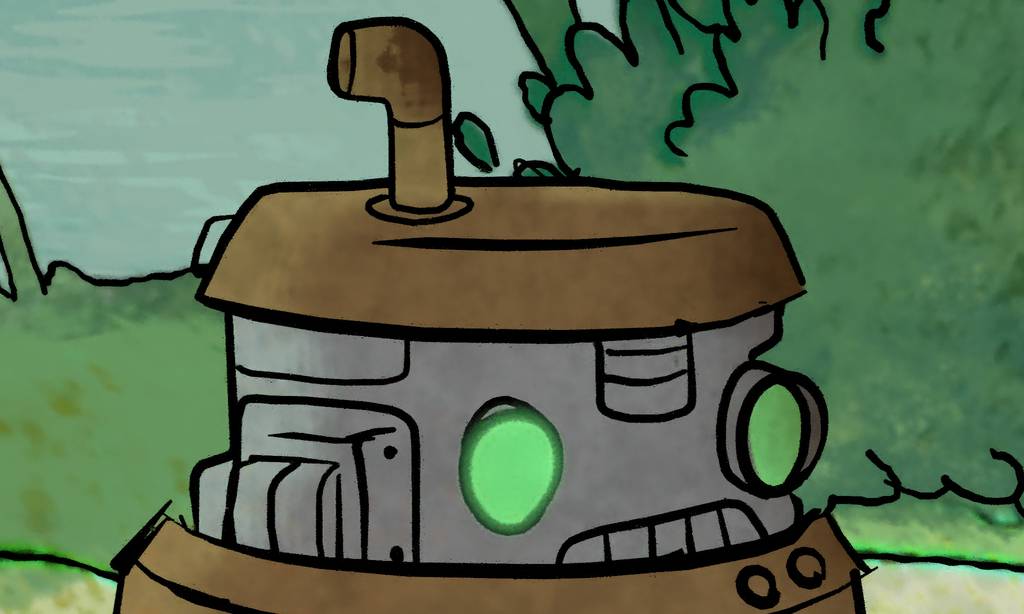} \\

        \rotatebox{90}{ \small \hspace{0.75em} \textbf{MetricGrids}} &
        \includegraphics[width=0.2\linewidth, height=0.12\linewidth]{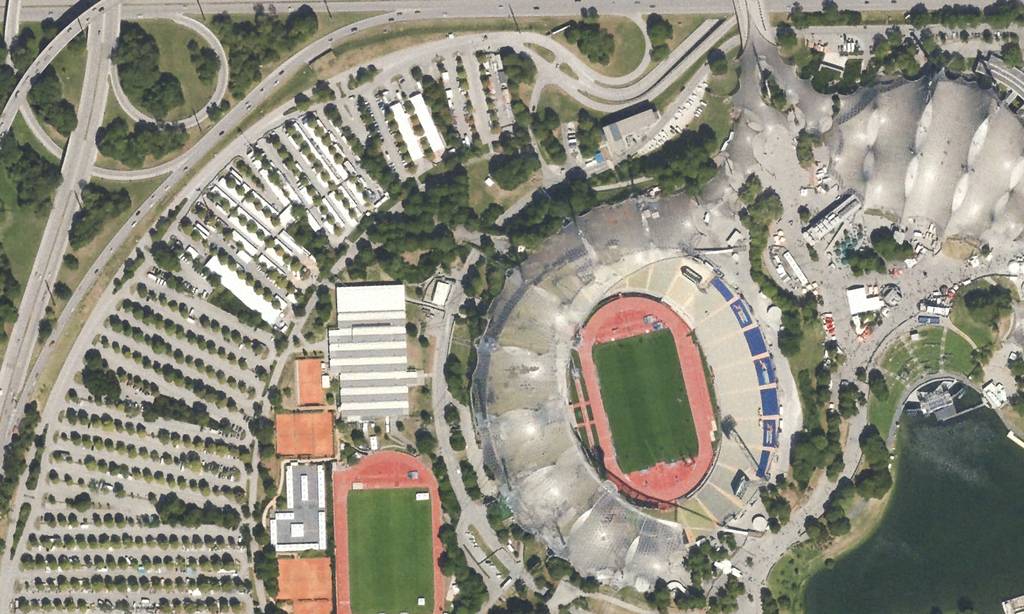} & 
        \includegraphics[width=0.2\linewidth, height=0.12\linewidth]{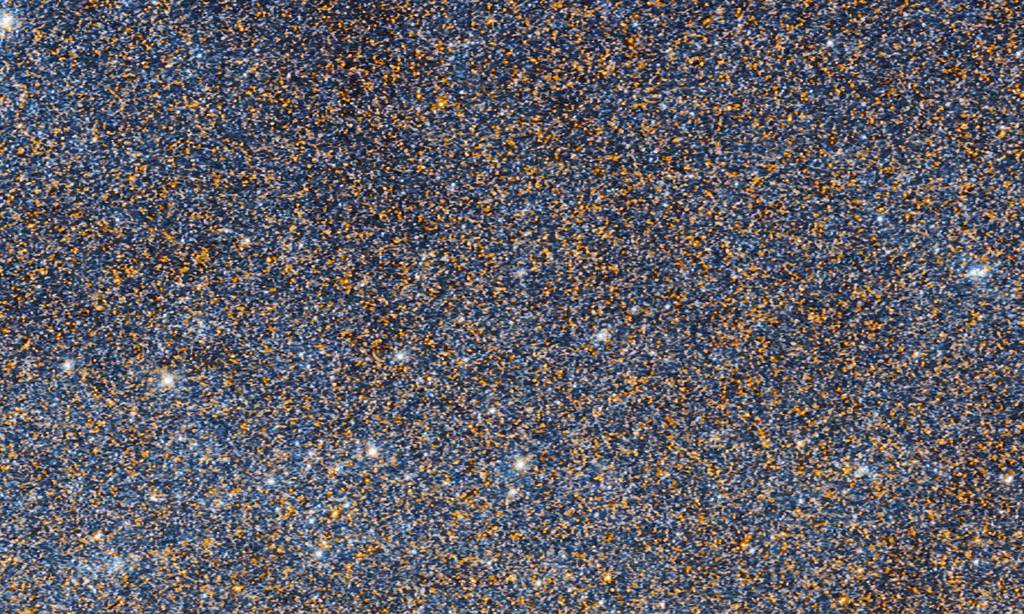} & 
        \includegraphics[width=0.2\linewidth, height=0.12\linewidth]{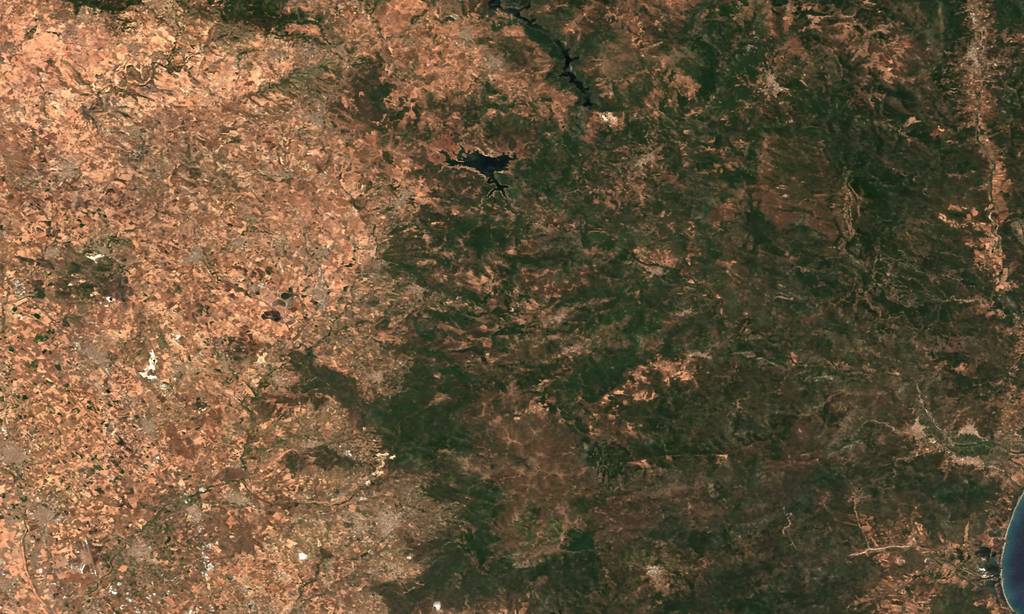} & 
        \includegraphics[width=0.2\linewidth, height=0.12\linewidth]{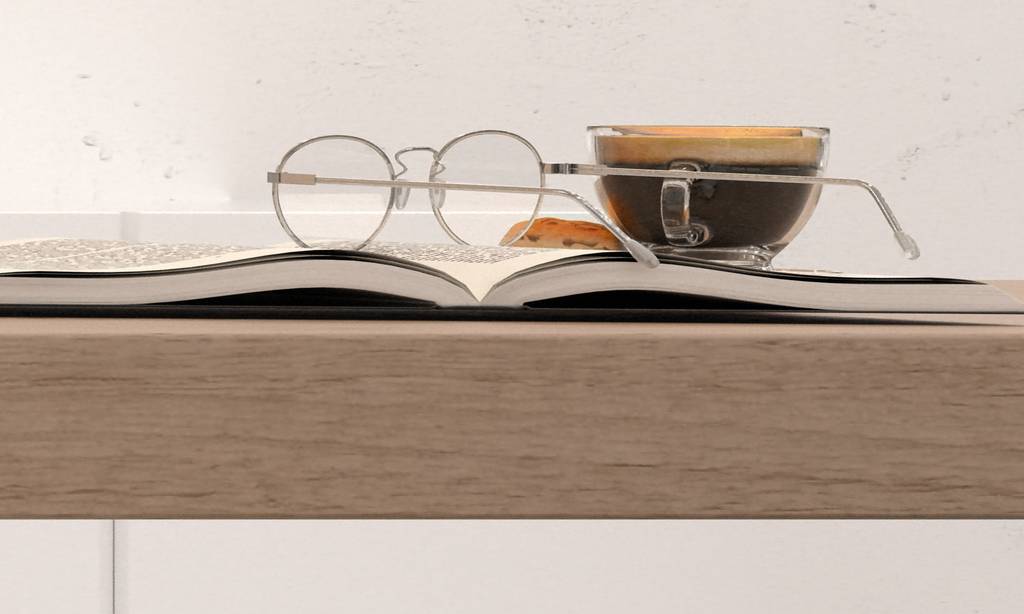} & 
        \includegraphics[width=0.2\linewidth, height=0.12\linewidth]{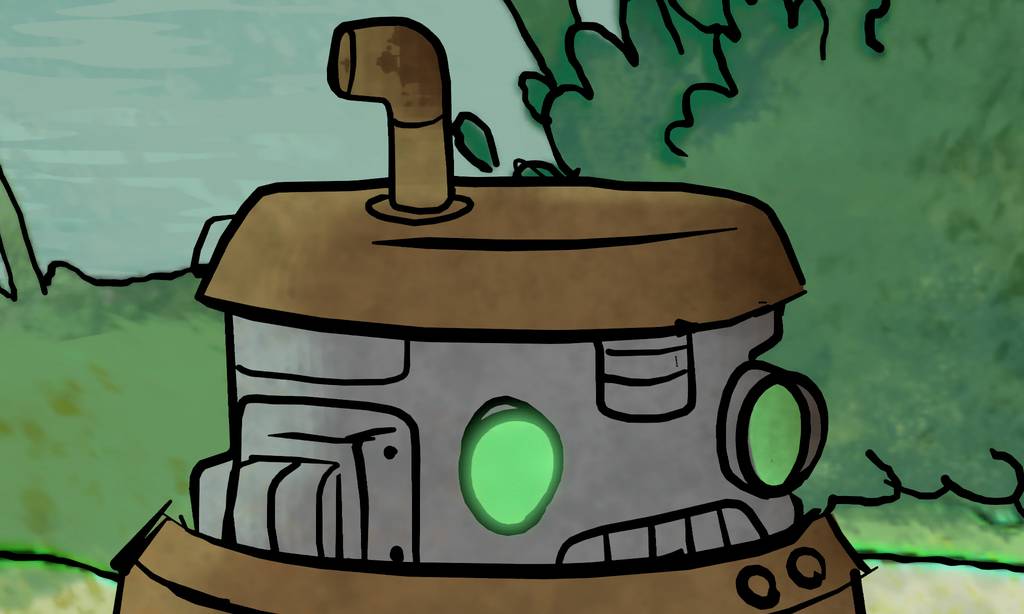} \\

        \rotatebox{90}{ \small \hspace{0.75em} \textbf{GaussImg}} &
        \includegraphics[width=0.2\linewidth, height=0.12\linewidth]{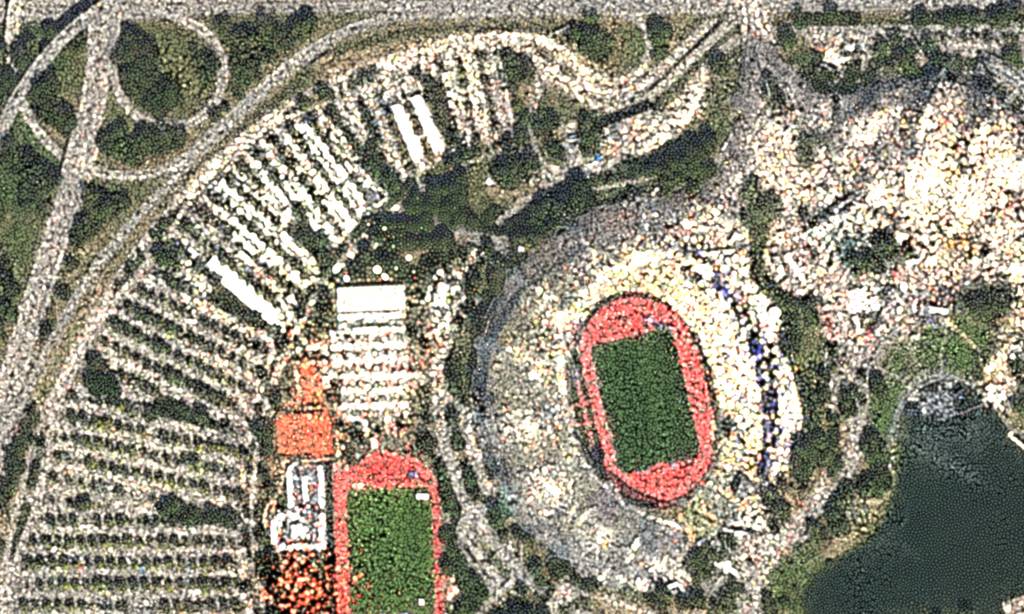} & 
        \includegraphics[width=0.2\linewidth, height=0.12\linewidth]{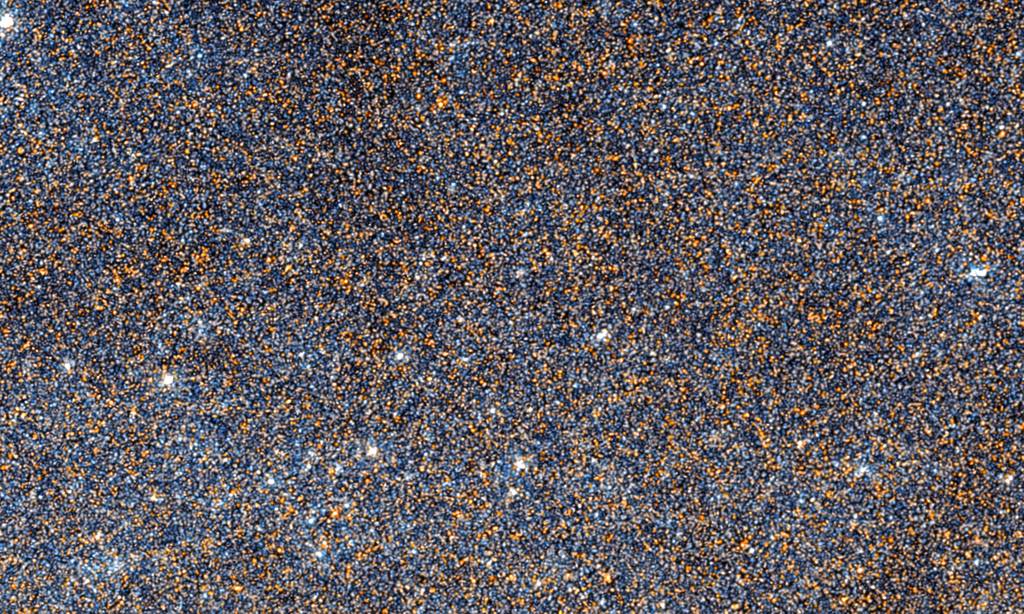} & 
        \includegraphics[width=0.2\linewidth, height=0.12\linewidth]{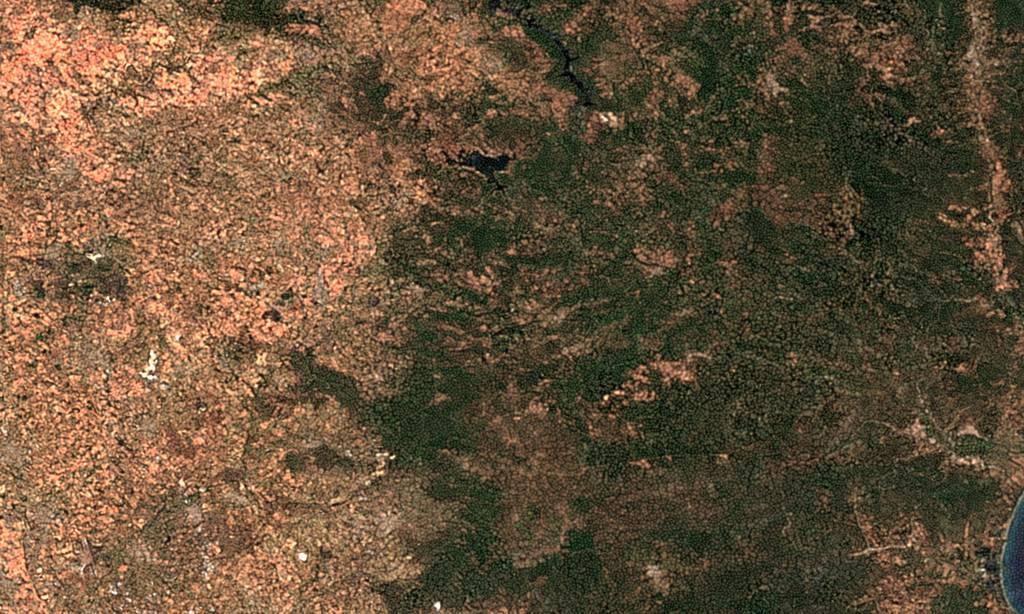} & 
        \includegraphics[width=0.2\linewidth, height=0.12\linewidth]{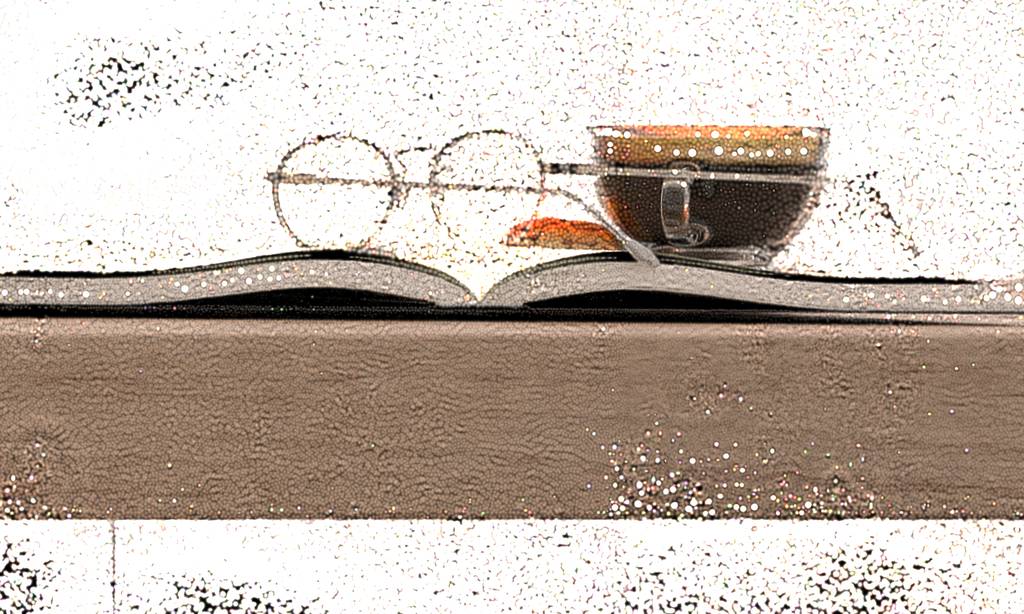} & 
        \includegraphics[width=0.2\linewidth, height=0.12\linewidth]{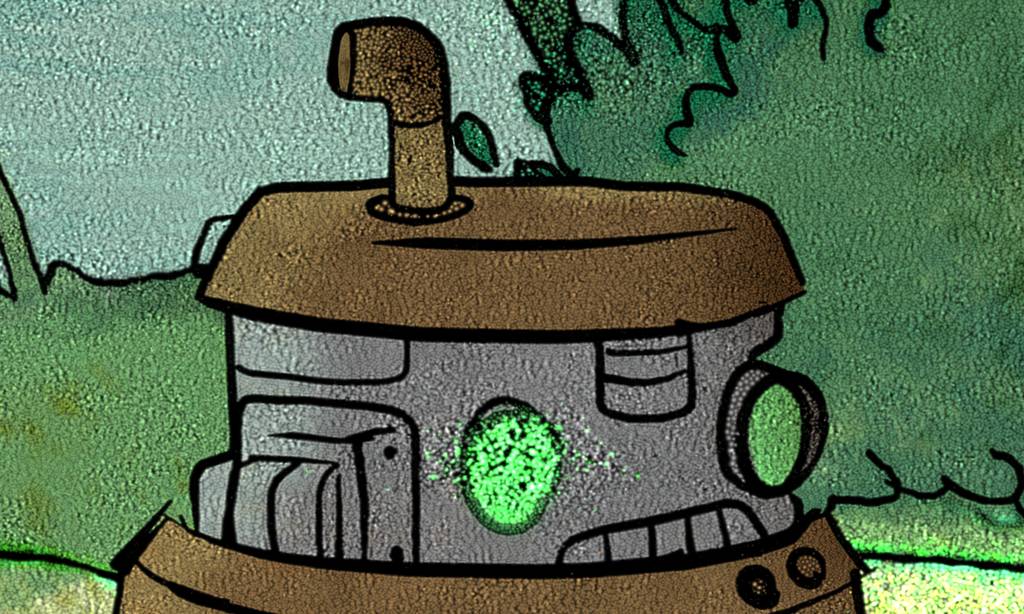} \\

        \rotatebox{90}{ \small \hspace{0.75em} \textbf{SmartSplat}} &
        \includegraphics[width=0.2\linewidth, height=0.12\linewidth]{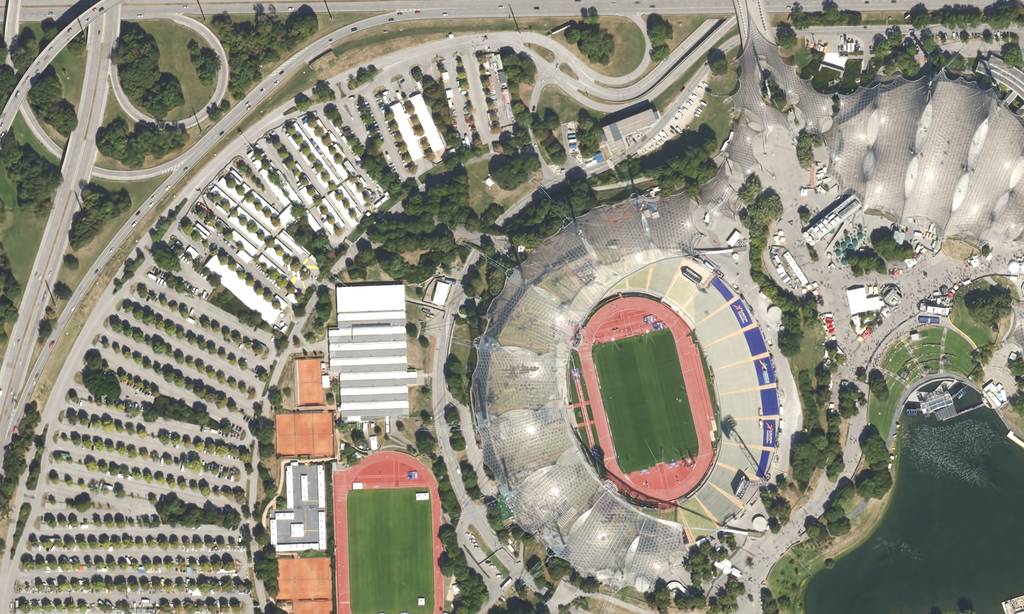} & 
        \includegraphics[width=0.2\linewidth, height=0.12\linewidth]{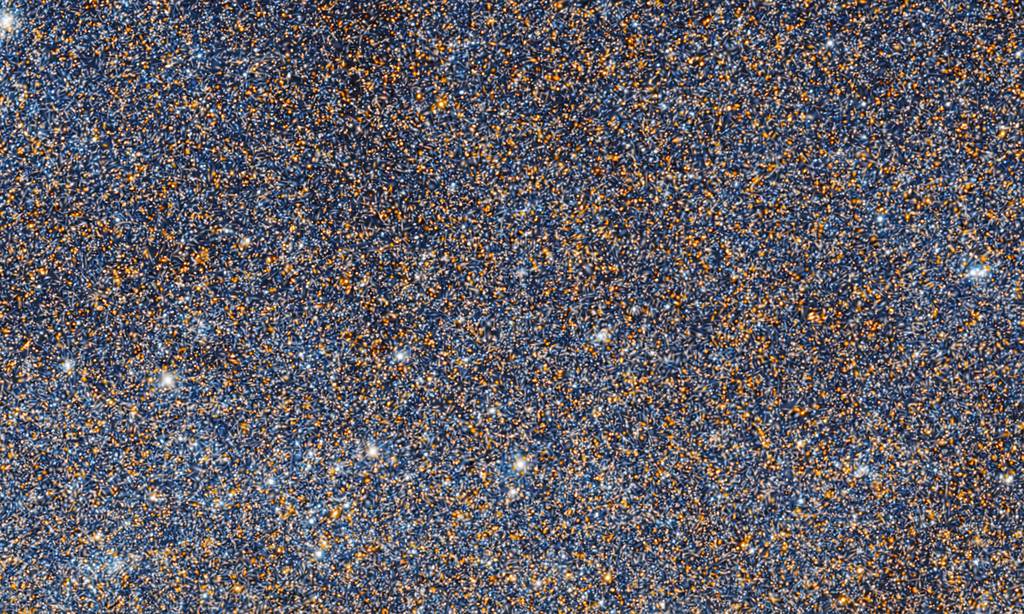} & 
        \includegraphics[width=0.2\linewidth, height=0.12\linewidth]{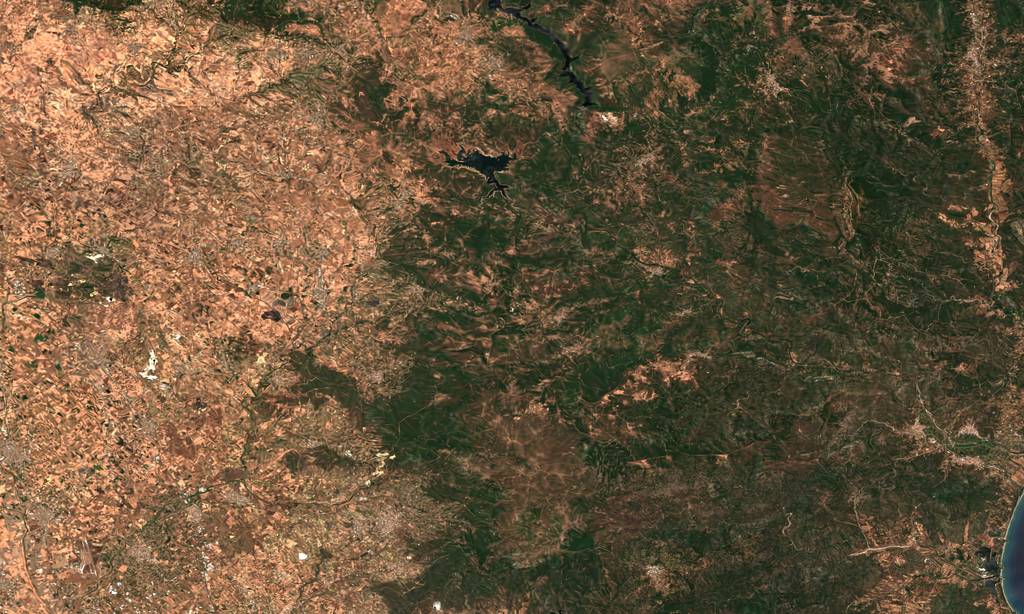} & 
        \includegraphics[width=0.2\linewidth, height=0.12\linewidth]{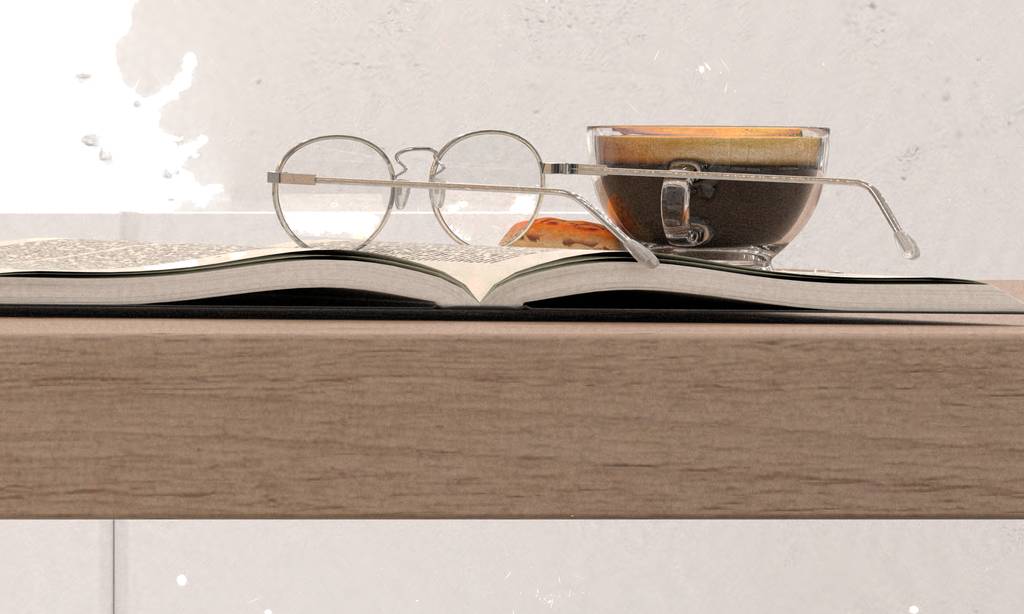} & 
        \includegraphics[width=0.2\linewidth, height=0.12\linewidth]{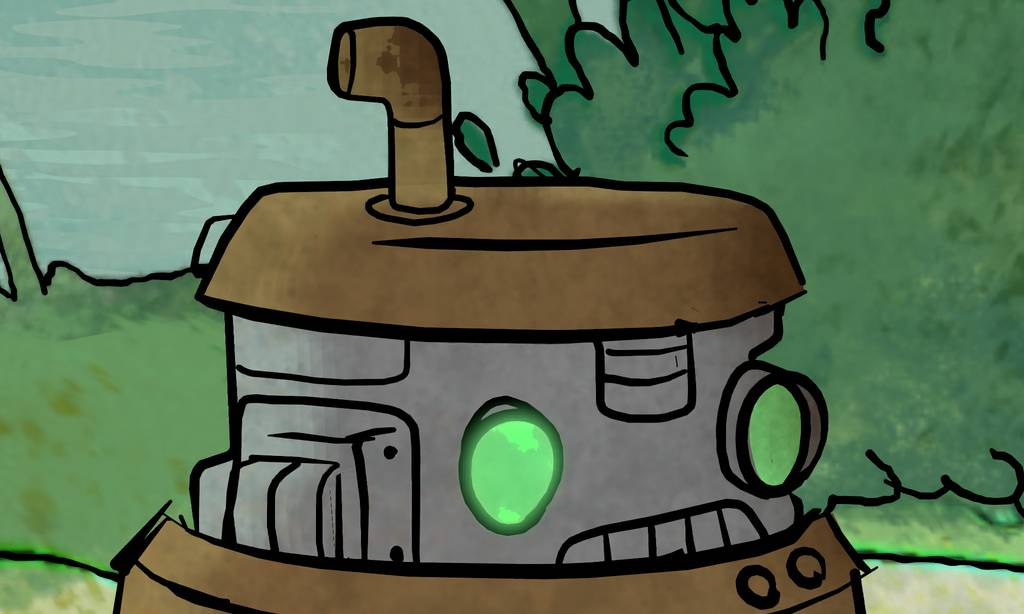} \\

    \end{tabular}}
    \caption{
    \textbf{Additional fitting qualitatives results on \dataname{}.}
    We show $1228 \times 2048$ crops from the full resolution images.
    }

    \label{fig:suppl_fit_qualitatives_2}

\end{figure*}

\begin{figure*}[ht]
    \centering
    \resizebox{\linewidth}{!}{
    \setlength{\tabcolsep}{1px}
    \begin{tabular}{c ccccc}
        & \textit{Venice} & \textit{Statuario}& \textit{Calacatta} & \textit{Breast Cancer} & \textit{Kidney} \\
        \rotatebox{90}{ \small \hspace{2em} \textbf{GT}} &
        \includegraphics[width=0.2\linewidth, height=0.12\linewidth]{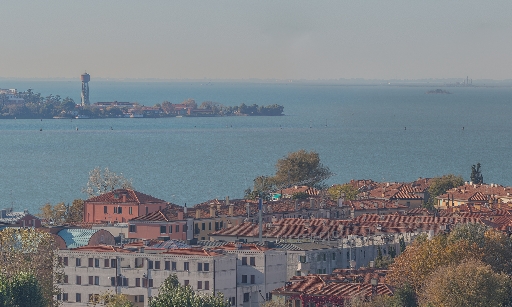} & 
        \includegraphics[width=0.2\linewidth, height=0.12\linewidth]{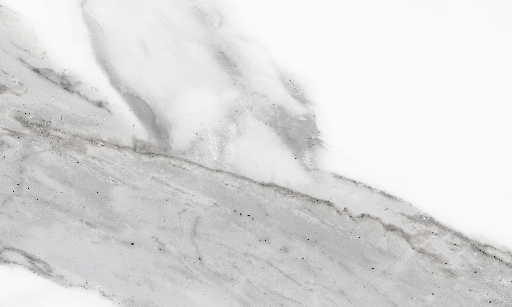} & 
        \includegraphics[width=0.2\linewidth, height=0.12\linewidth]{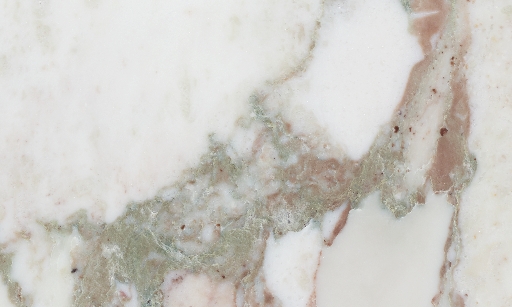} & 
        \includegraphics[width=0.2\linewidth, height=0.12\linewidth]{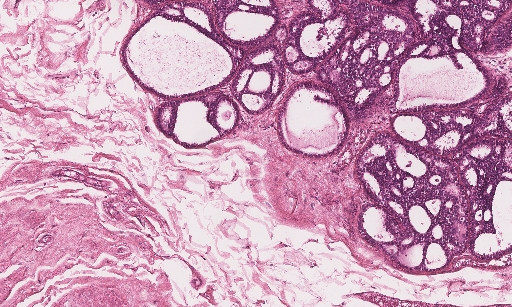} & 
        \includegraphics[width=0.2\linewidth, height=0.12\linewidth]{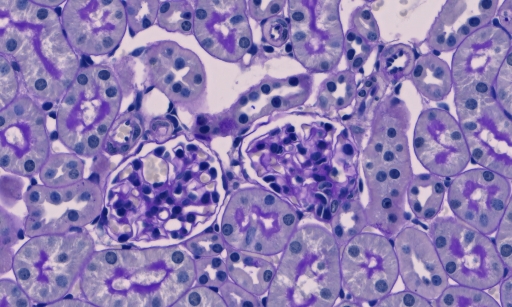} \\

        \rotatebox{90}{ \small \hspace{1.5em} \textbf{SIREN}} &
        \includegraphics[width=0.2\linewidth, height=0.12\linewidth]{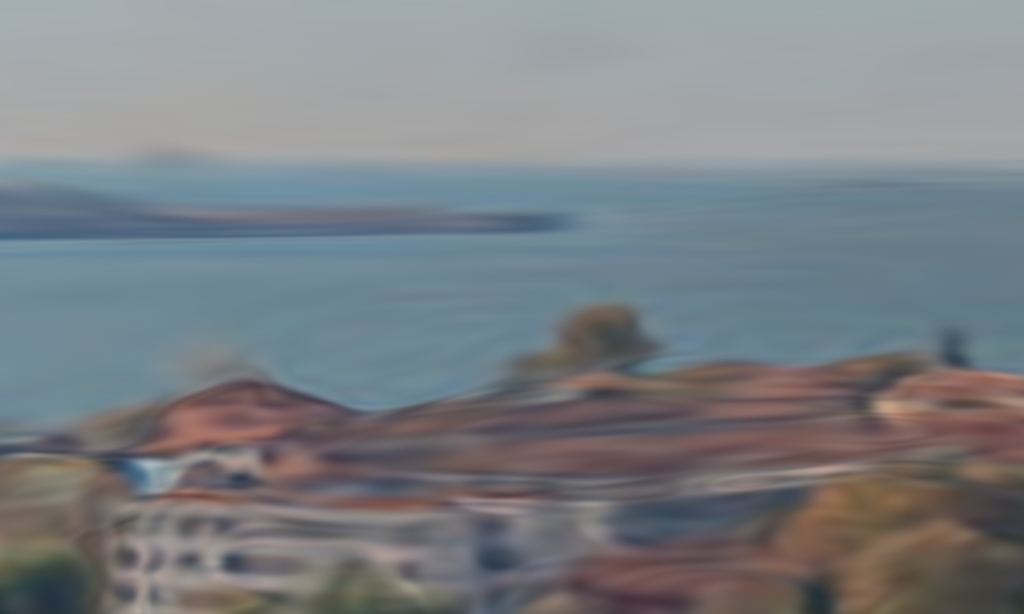} & 
        \includegraphics[width=0.2\linewidth, height=0.12\linewidth]{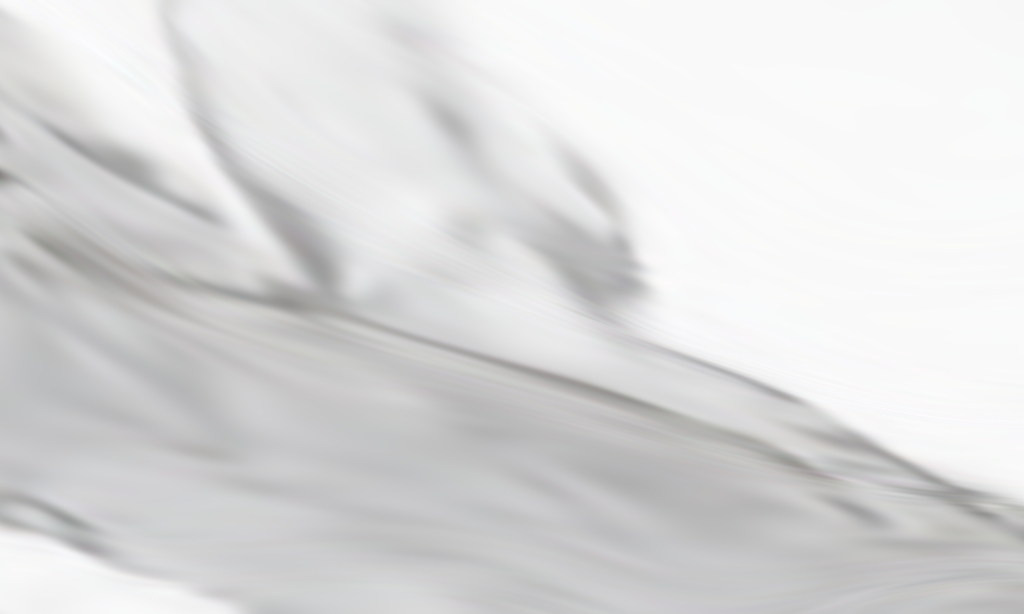} & 
        \includegraphics[width=0.2\linewidth, height=0.12\linewidth]{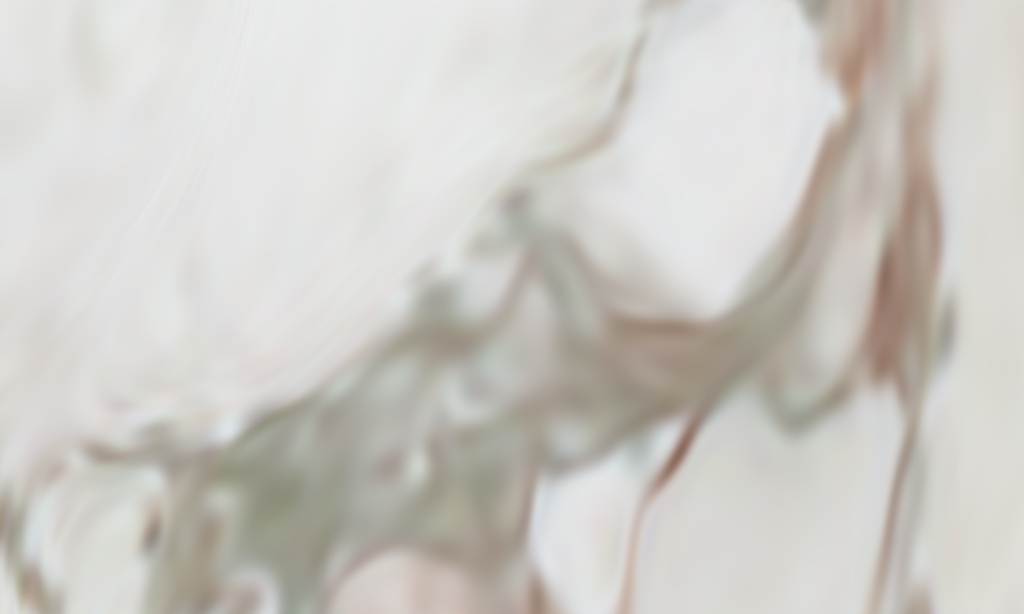} & 
        \includegraphics[width=0.2\linewidth, height=0.12\linewidth]{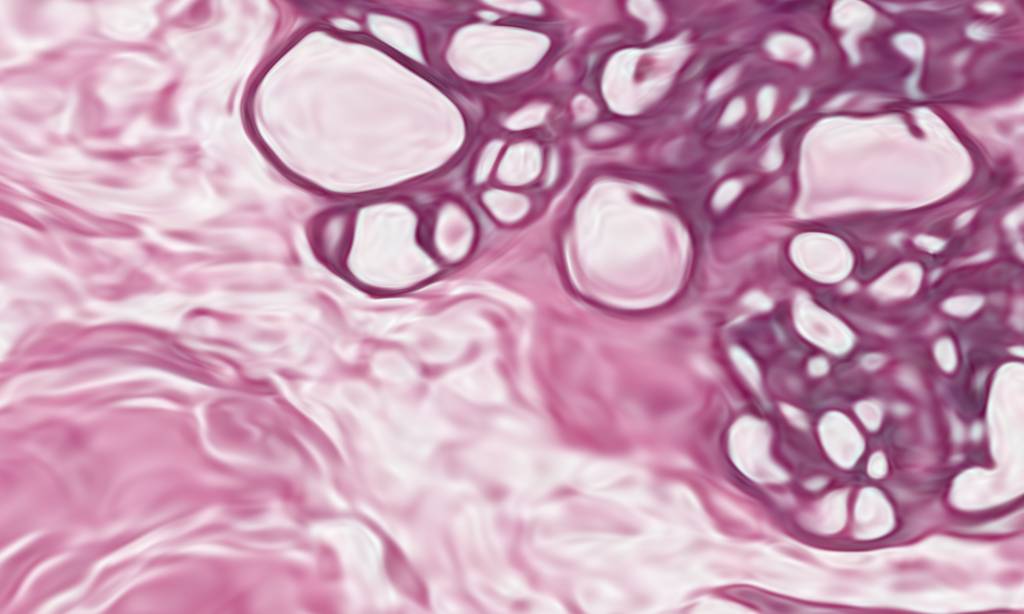} & 
        \includegraphics[width=0.2\linewidth, height=0.12\linewidth]{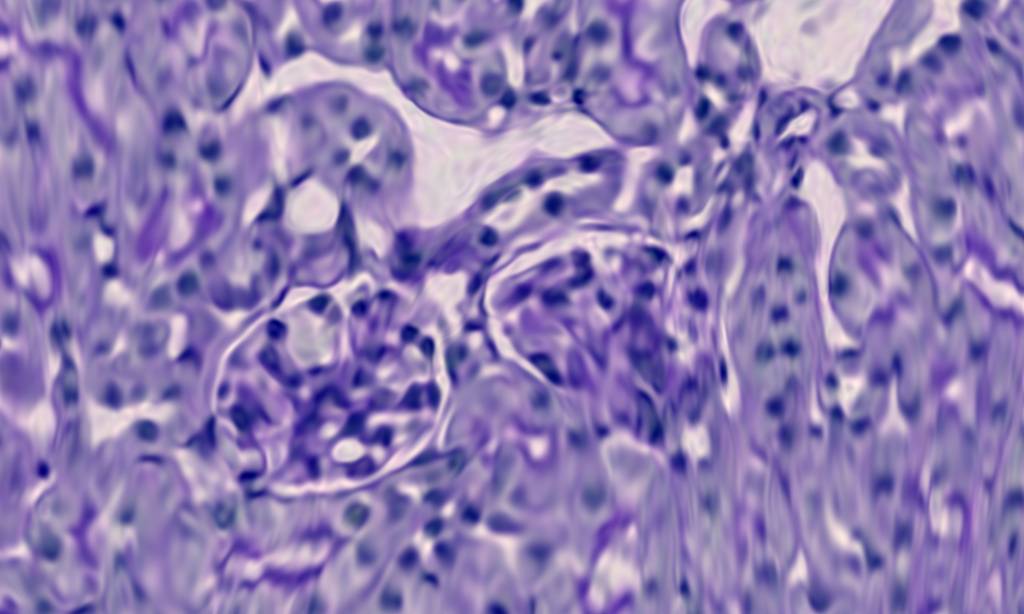} \\

        \rotatebox{90}{ \small \hspace{1em} \textbf{FDHO}} &
        \includegraphics[width=0.2\linewidth, height=0.12\linewidth]{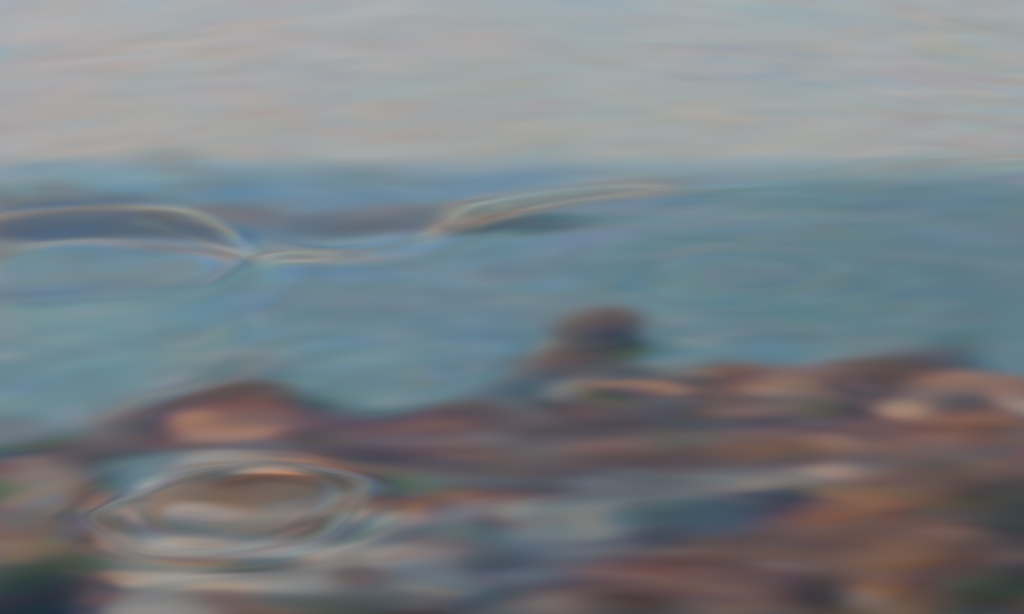} & 
        \includegraphics[width=0.2\linewidth, height=0.12\linewidth]{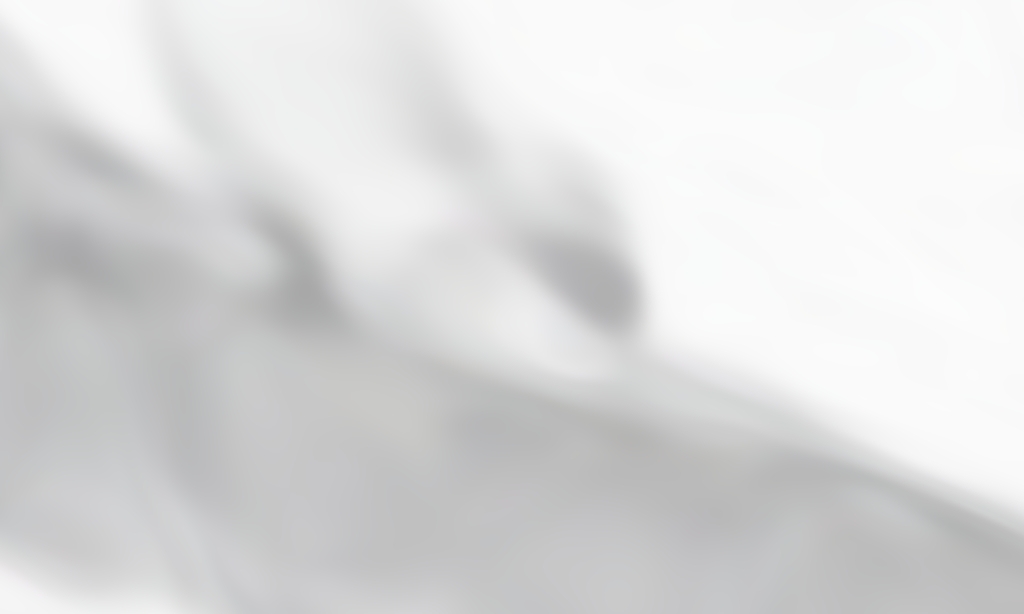} & 
        \includegraphics[width=0.2\linewidth, height=0.12\linewidth]{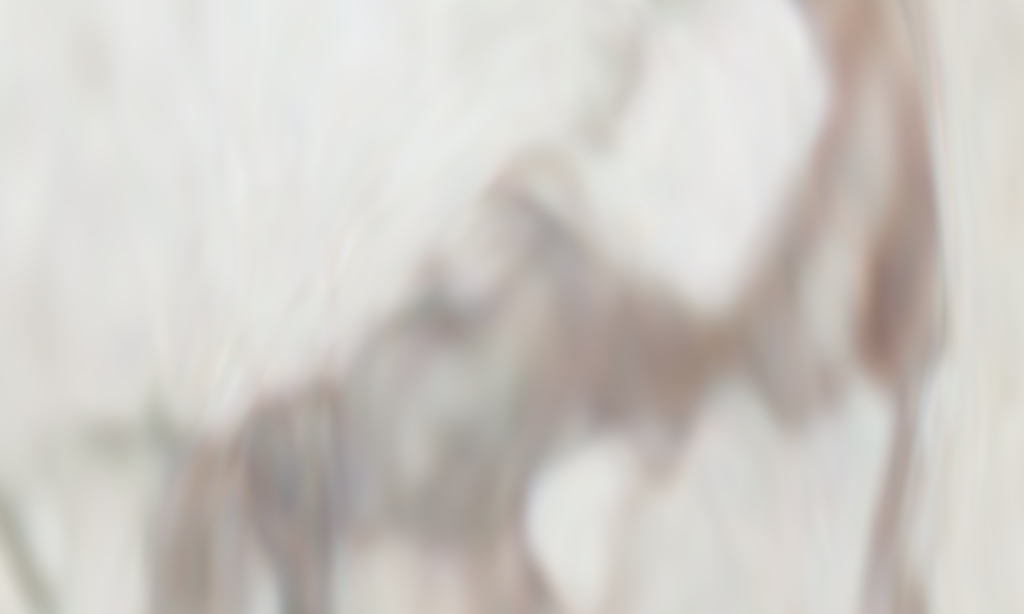} & 
        \includegraphics[width=0.2\linewidth, height=0.12\linewidth]{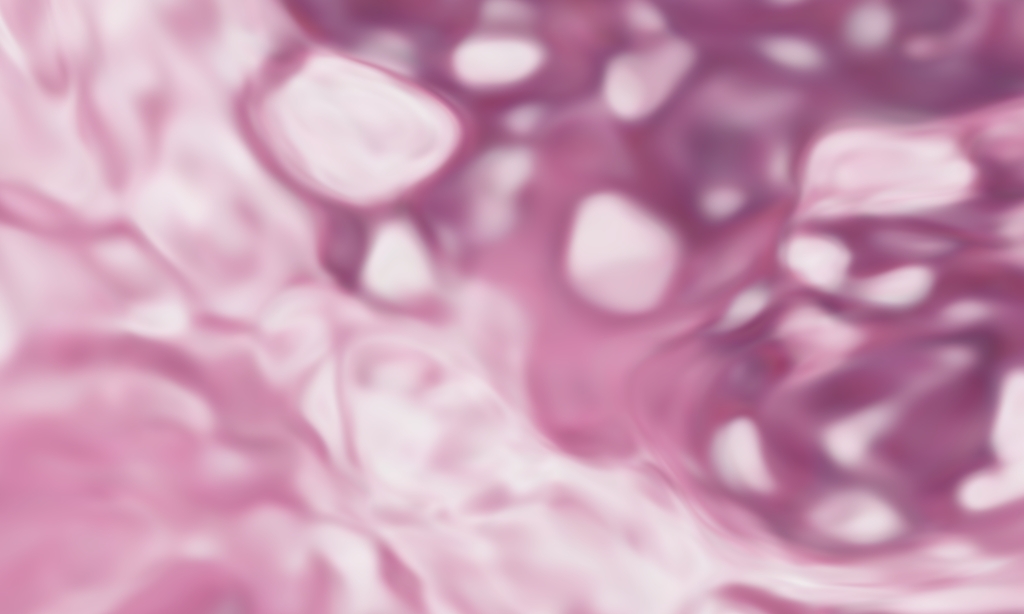} & 
        \includegraphics[width=0.2\linewidth, height=0.12\linewidth]{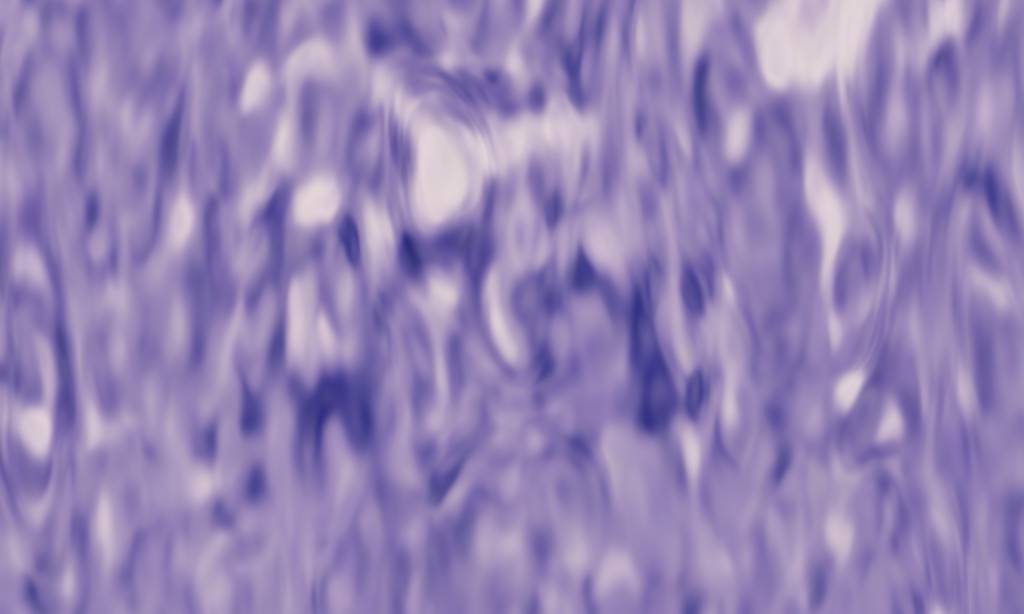} \\

        \rotatebox{90}{ \small \hspace{1em} \textbf{FINER}} &
        \includegraphics[width=0.2\linewidth, height=0.12\linewidth]{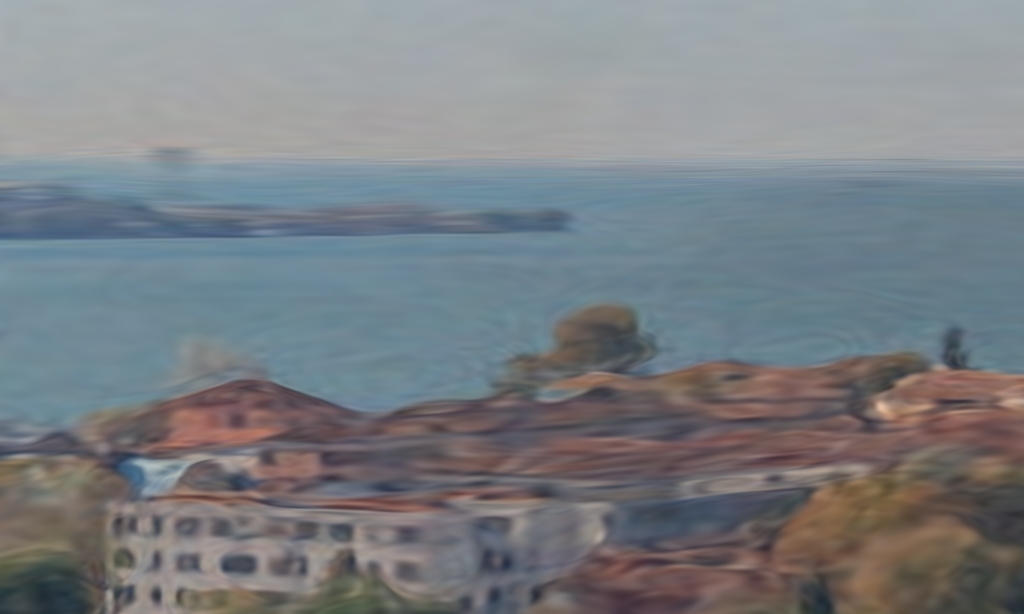} & 
        \includegraphics[width=0.2\linewidth, height=0.12\linewidth]{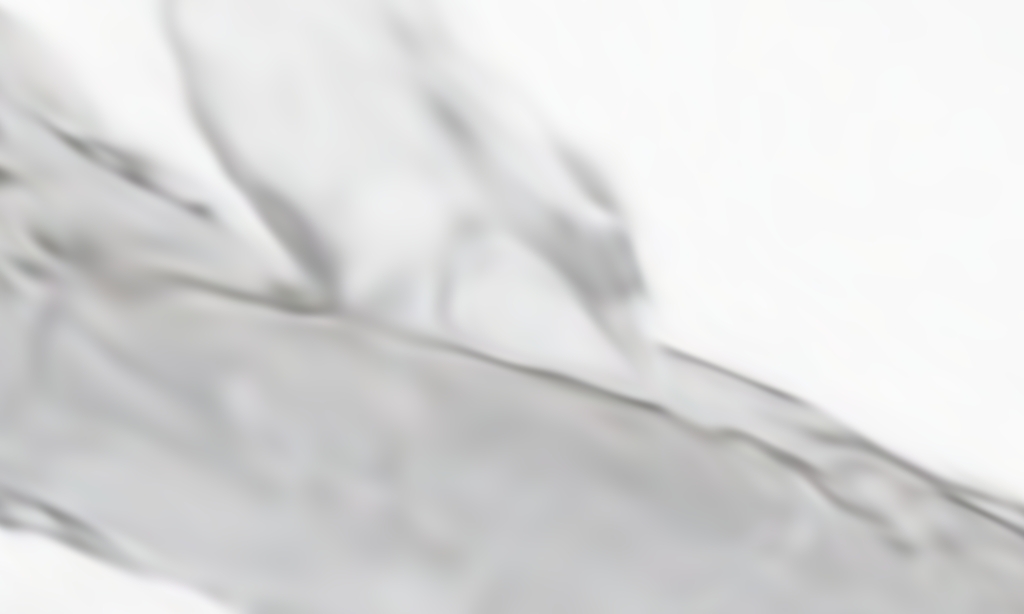} & 
        \includegraphics[width=0.2\linewidth, height=0.12\linewidth]{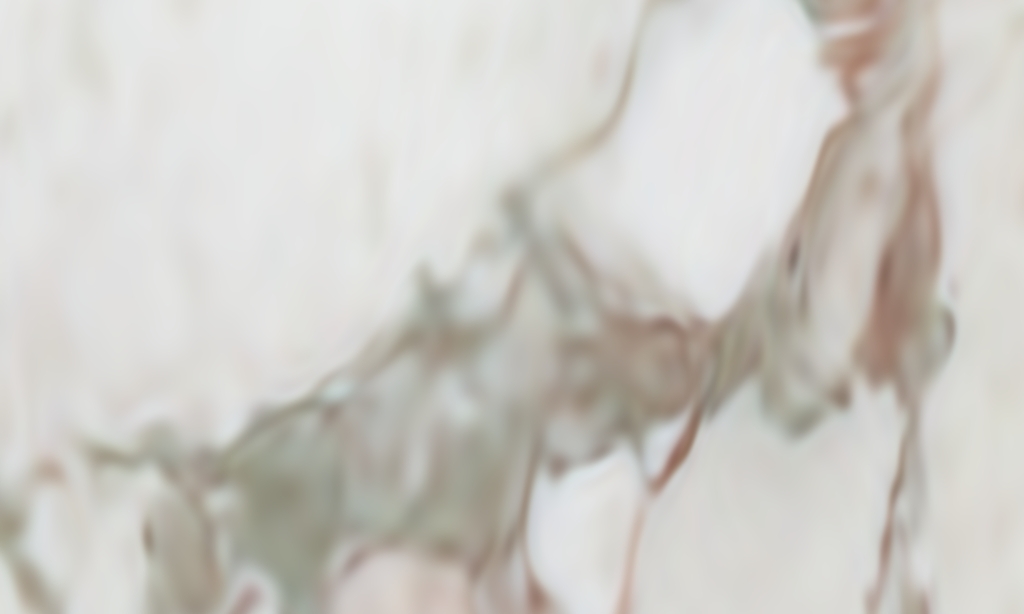} & 
        \includegraphics[width=0.2\linewidth, height=0.12\linewidth]{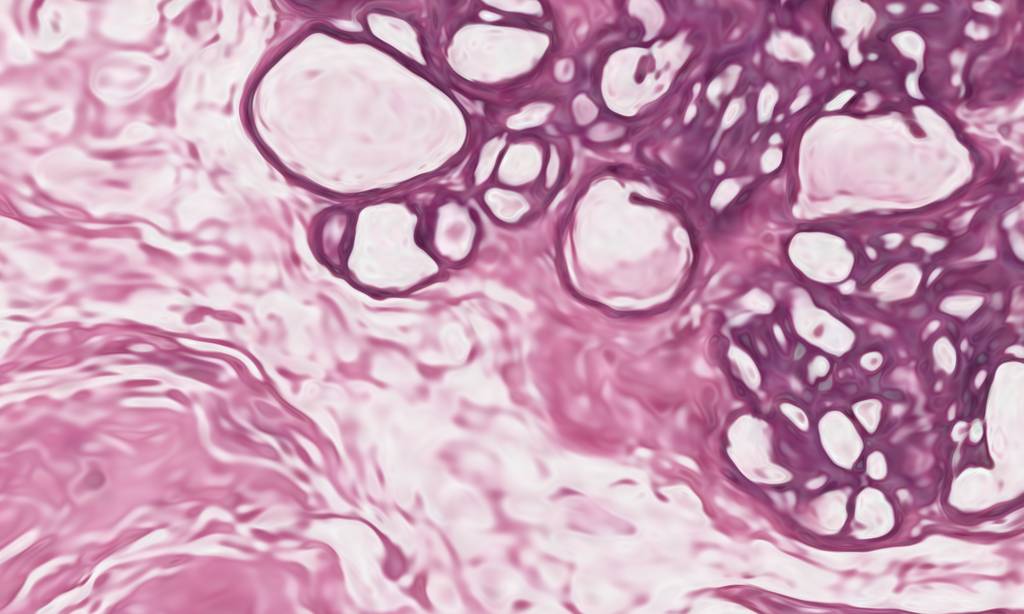} & 
        \includegraphics[width=0.2\linewidth, height=0.12\linewidth]{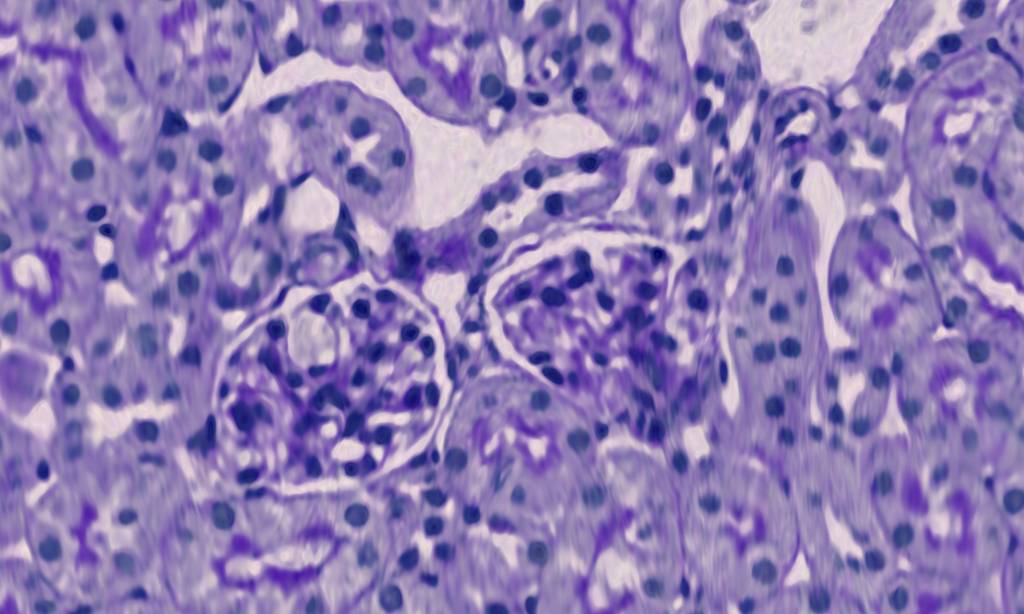} \\

        \rotatebox{90}{ \small \hspace{1.25em} \textbf{I-NGP}} &
        \includegraphics[width=0.2\linewidth, height=0.12\linewidth]{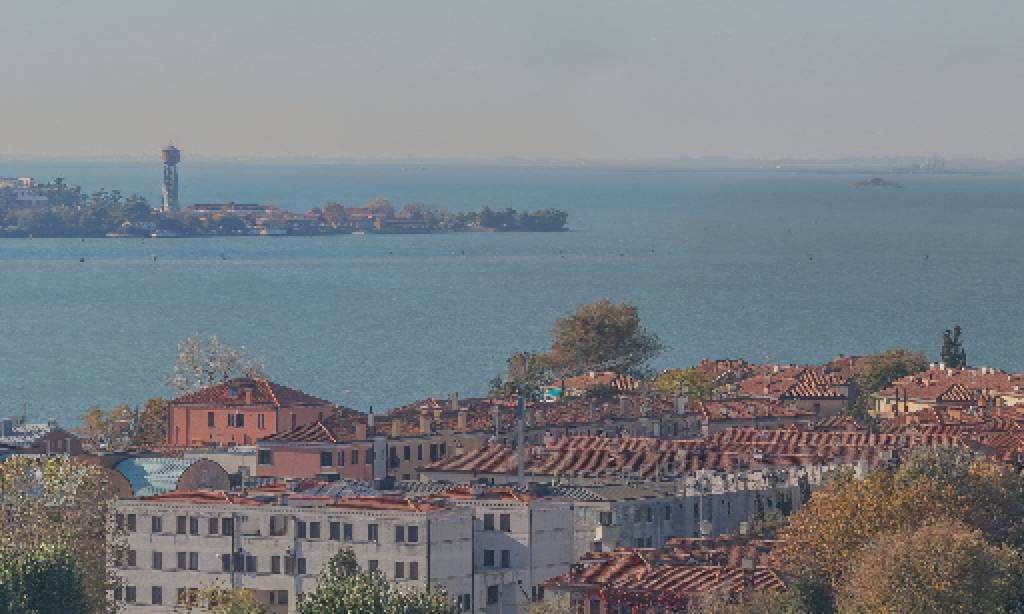} & 
        \includegraphics[width=0.2\linewidth, height=0.12\linewidth]{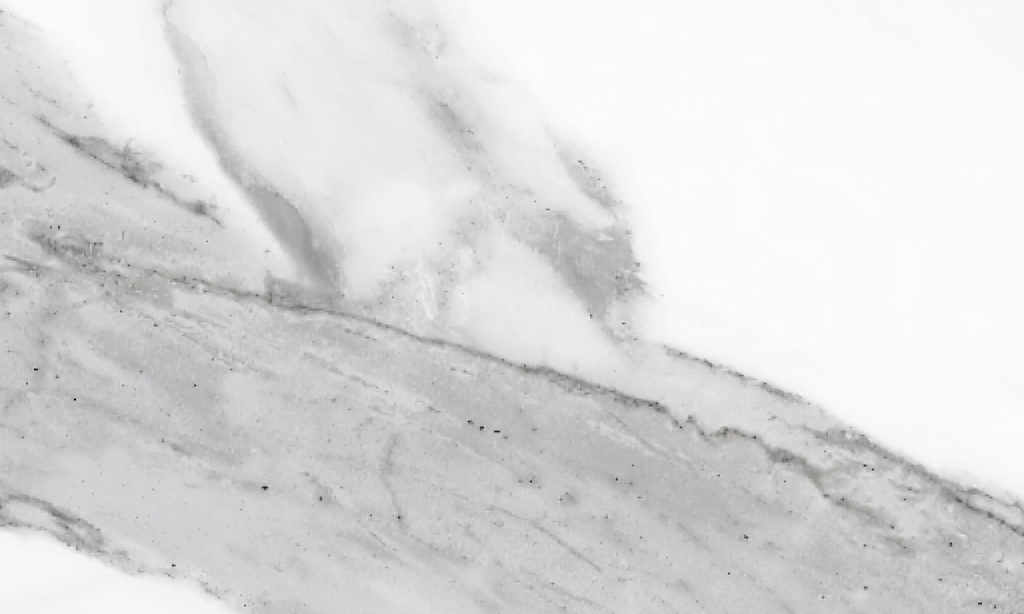} & 
        \includegraphics[width=0.2\linewidth, height=0.12\linewidth]{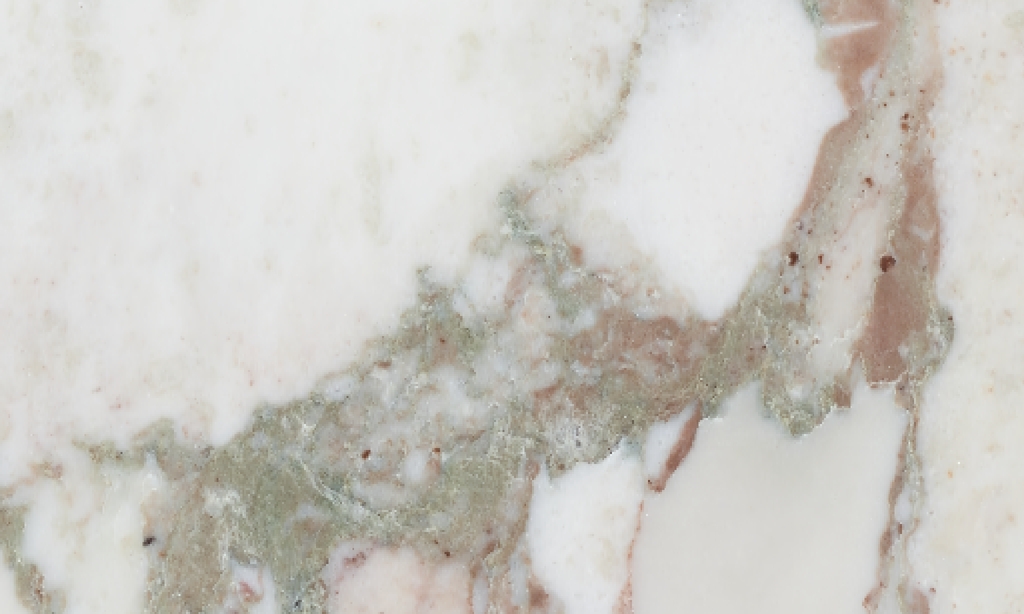} & 
        \includegraphics[width=0.2\linewidth, height=0.12\linewidth]{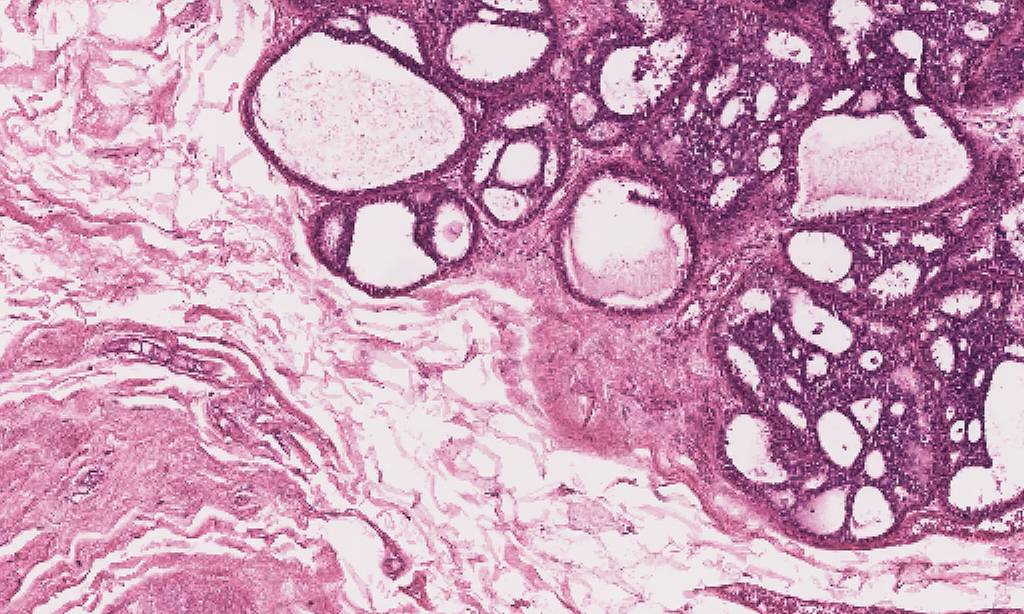} & 
        \includegraphics[width=0.2\linewidth, height=0.12\linewidth]{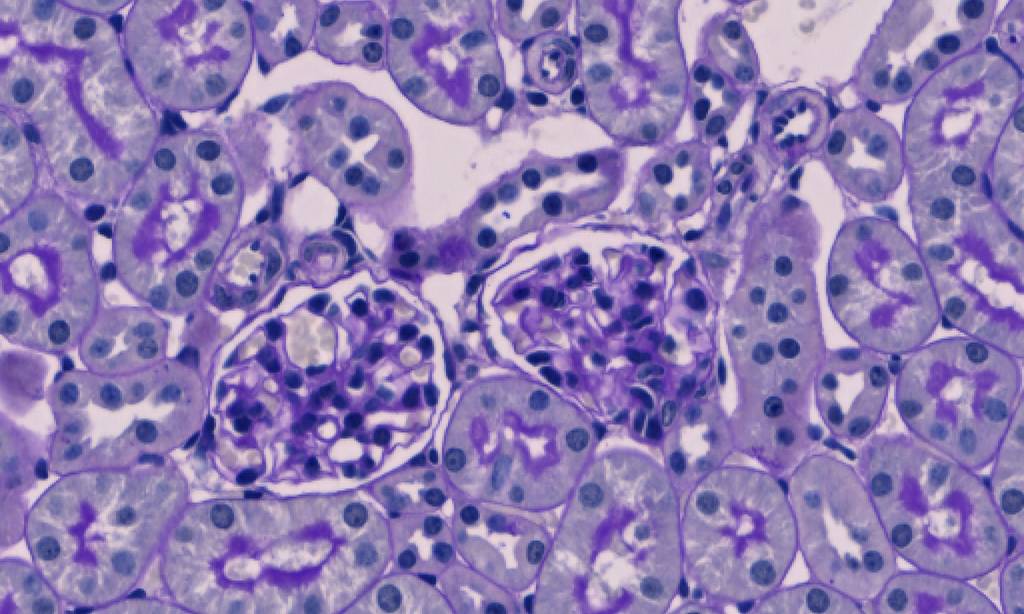} \\

        \rotatebox{90}{ \small \hspace{1em} \textbf{NeuRBF}} &
        \includegraphics[width=0.2\linewidth, height=0.12\linewidth]{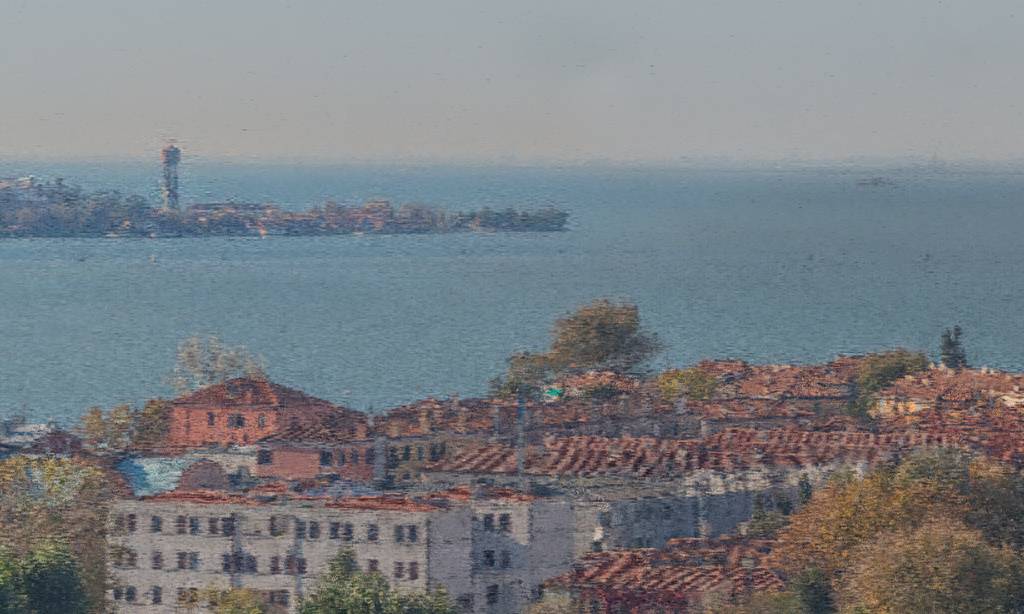} & 
        \includegraphics[width=0.2\linewidth, height=0.12\linewidth]{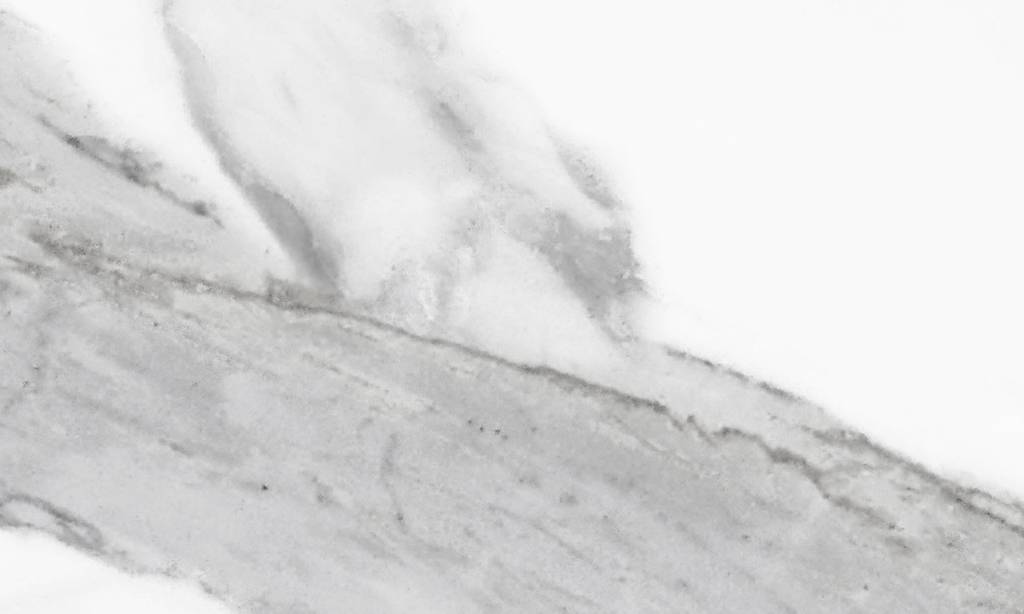} & 
        \includegraphics[width=0.2\linewidth, height=0.12\linewidth]{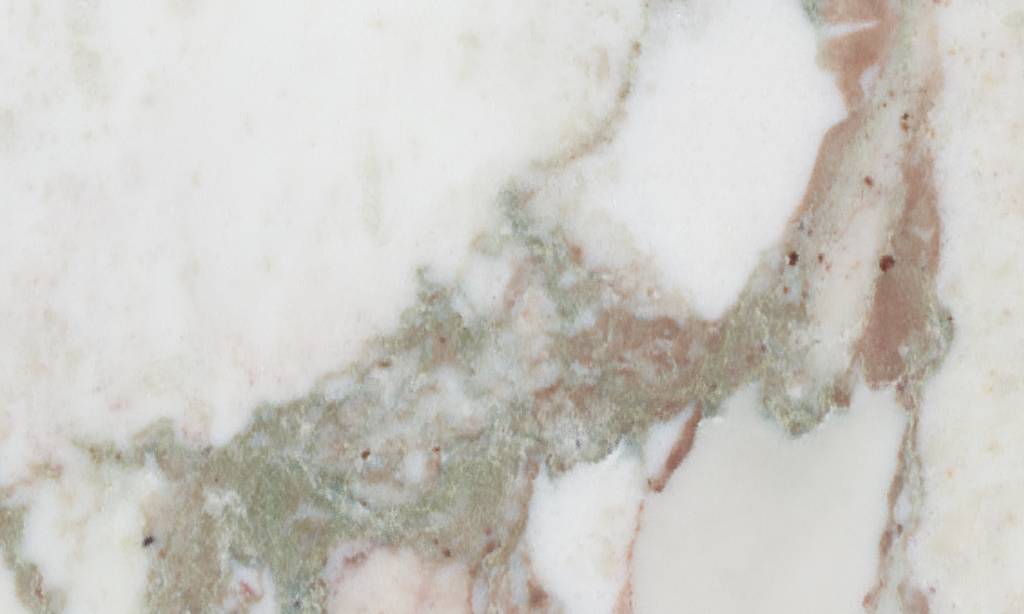} & 
        \includegraphics[width=0.2\linewidth, height=0.12\linewidth]{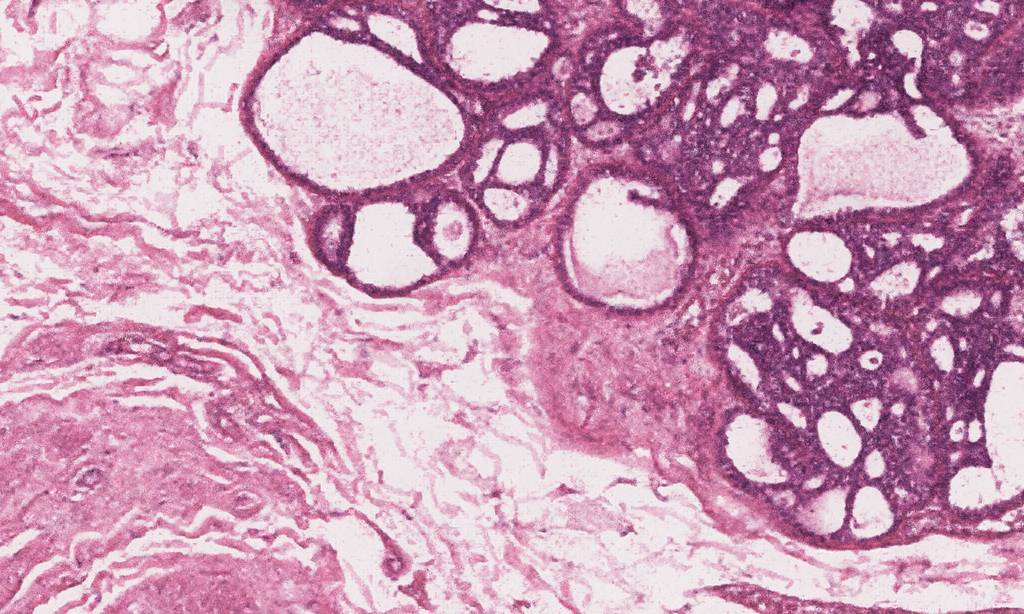} & 
        \includegraphics[width=0.2\linewidth, height=0.12\linewidth]{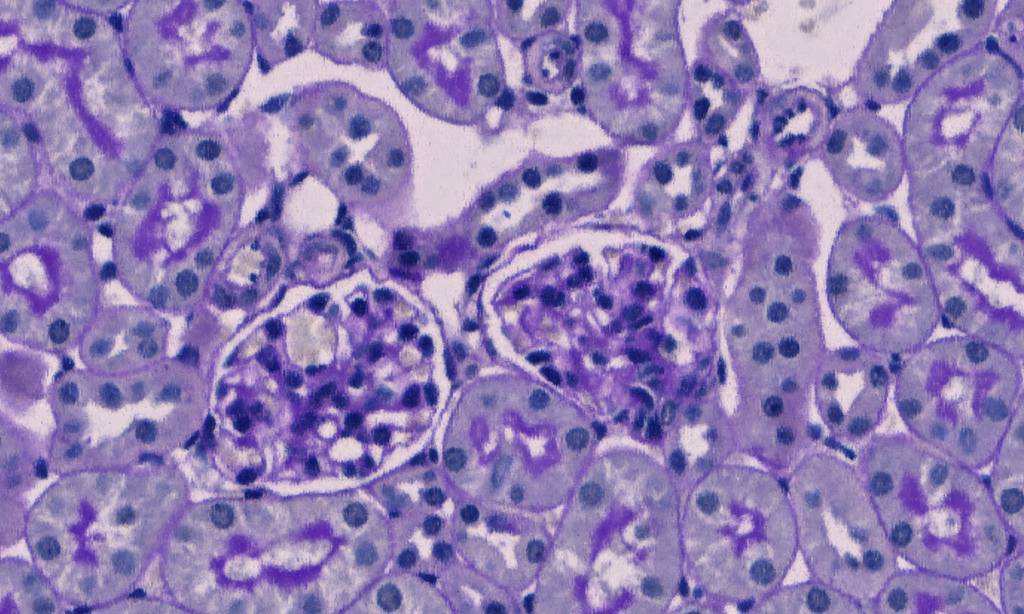} \\

        \rotatebox{90}{ \small \hspace{0.25em} \textbf{MetricGrids}} &
        \includegraphics[width=0.2\linewidth, height=0.12\linewidth]{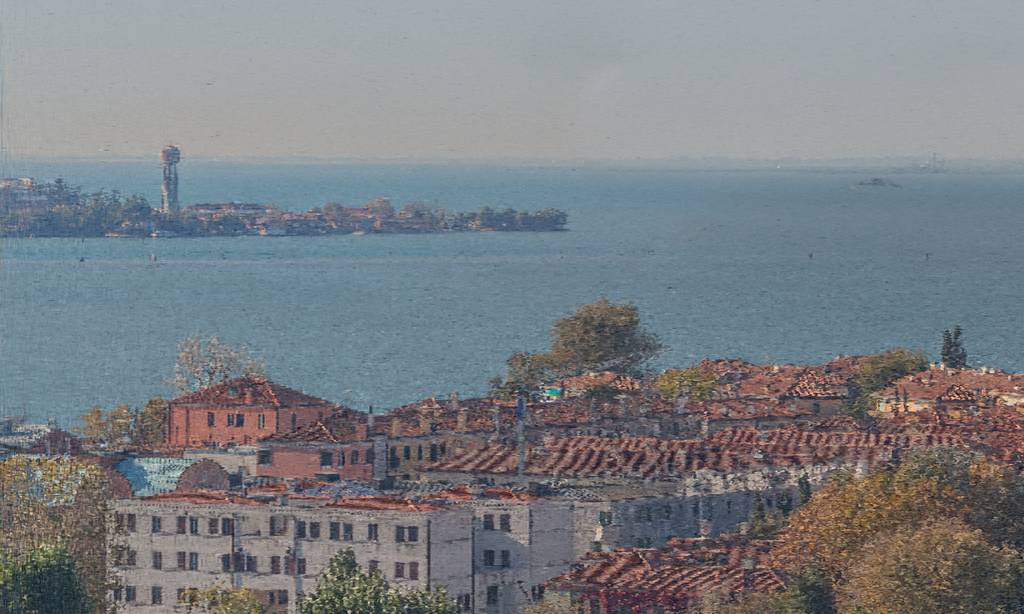} & 
        \includegraphics[width=0.2\linewidth, height=0.12\linewidth]{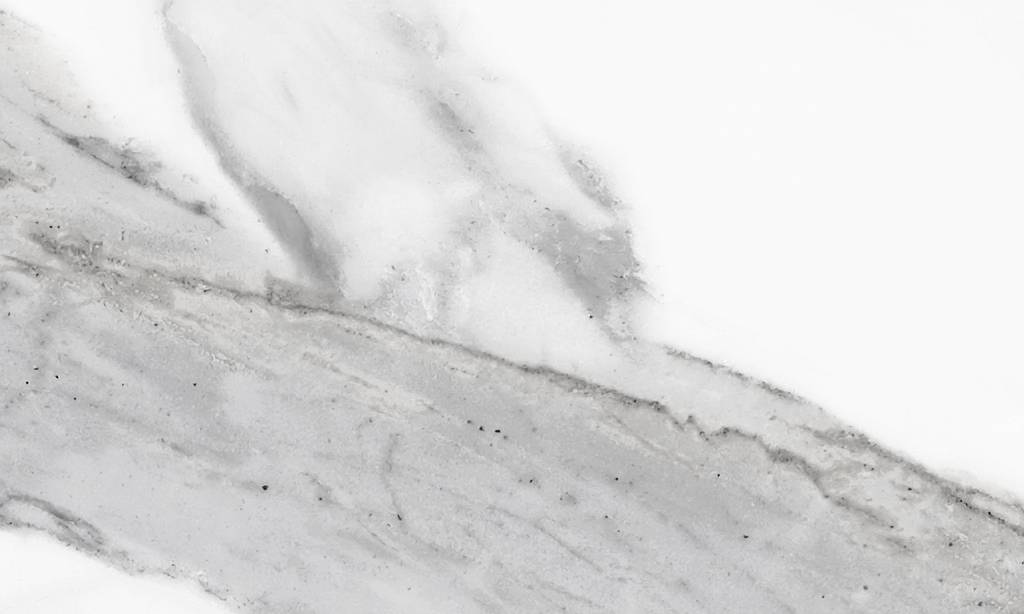} & 
        \includegraphics[width=0.2\linewidth, height=0.12\linewidth]{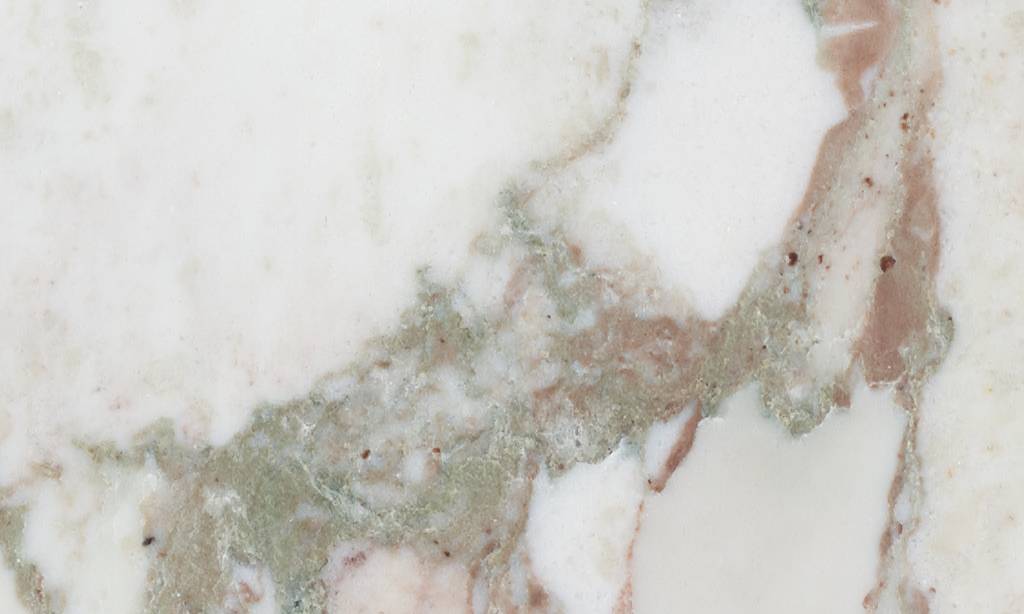} & 
        \includegraphics[width=0.2\linewidth, height=0.12\linewidth]{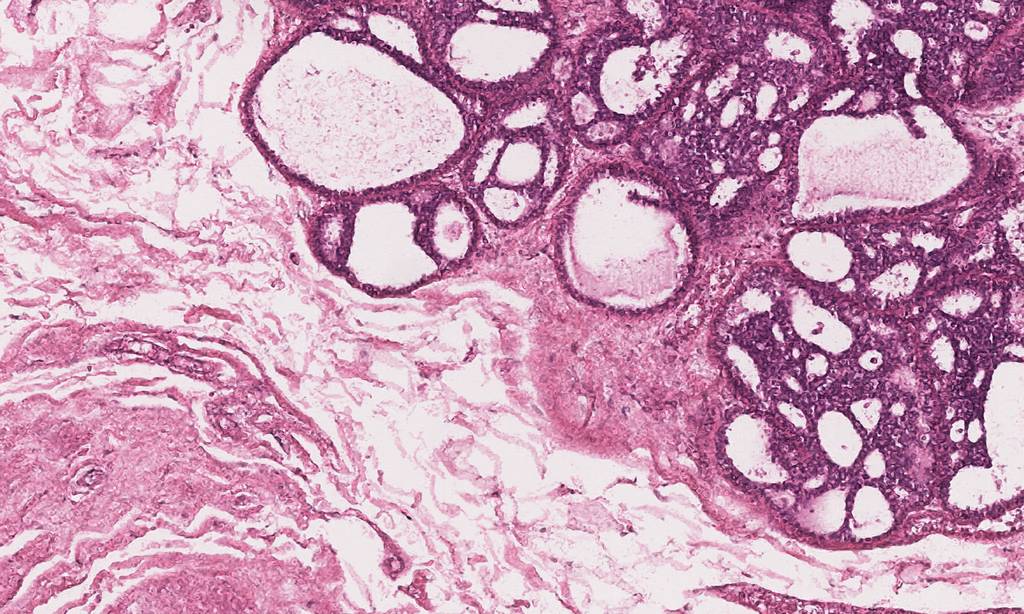} & 
        \includegraphics[width=0.2\linewidth, height=0.12\linewidth]{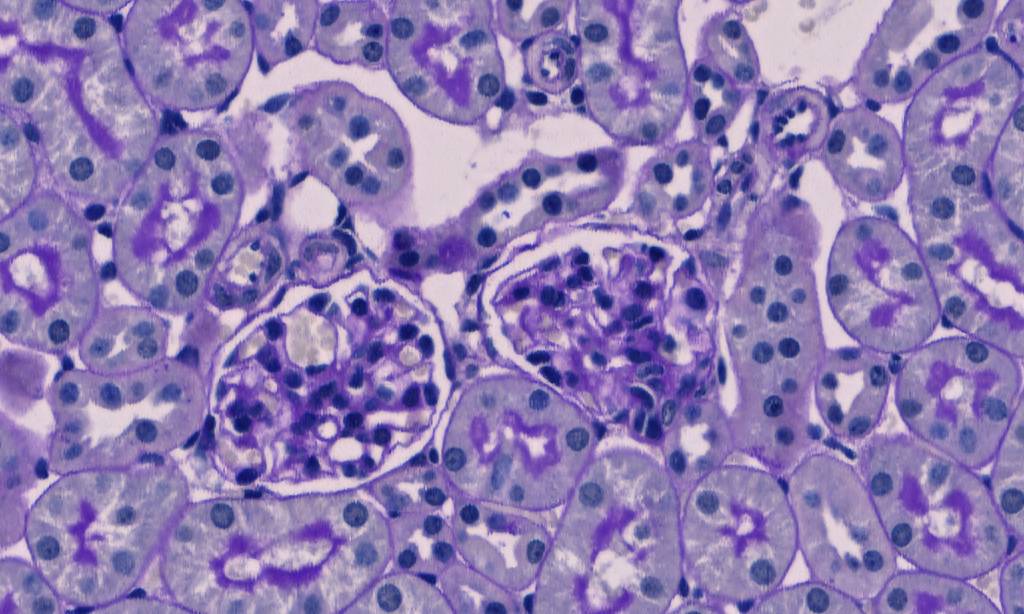} \\

        \rotatebox{90}{ \small \hspace{0.5em} \textbf{GaussImg}} &
        \includegraphics[width=0.2\linewidth, height=0.12\linewidth]{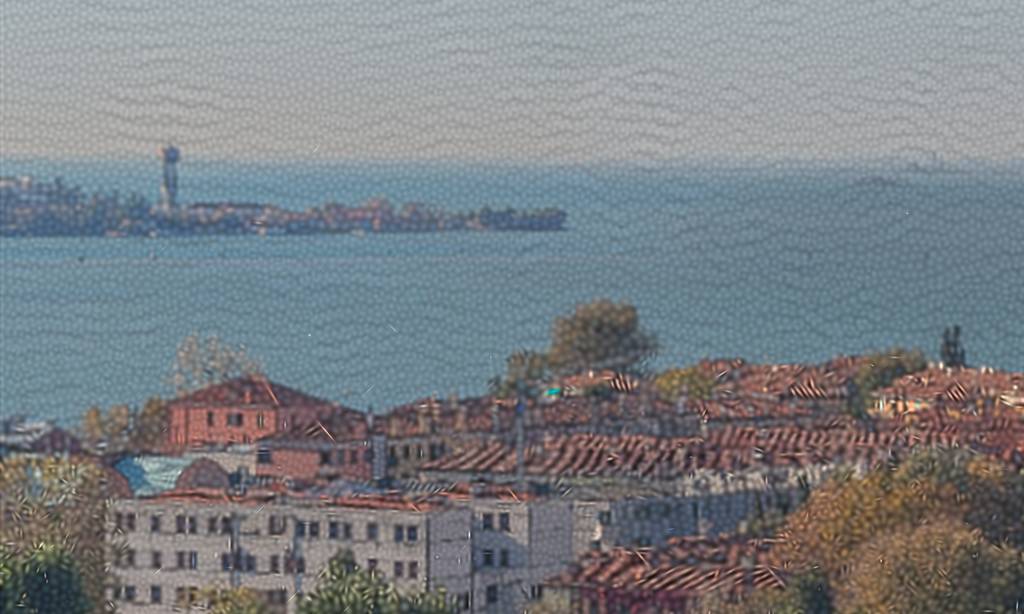} & 
        \includegraphics[width=0.2\linewidth, height=0.12\linewidth]{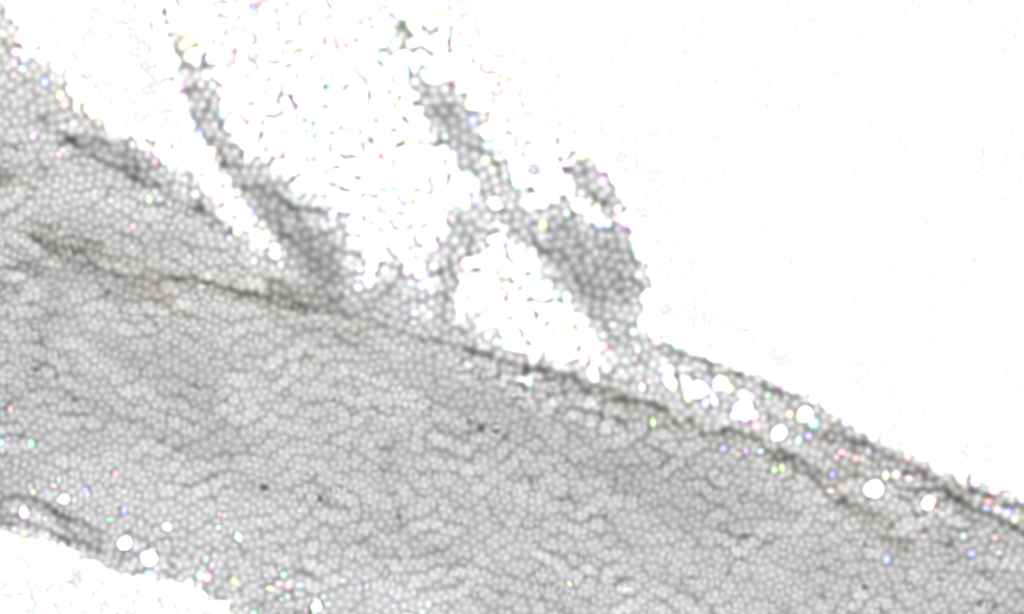} & 
        \includegraphics[width=0.2\linewidth, height=0.12\linewidth]{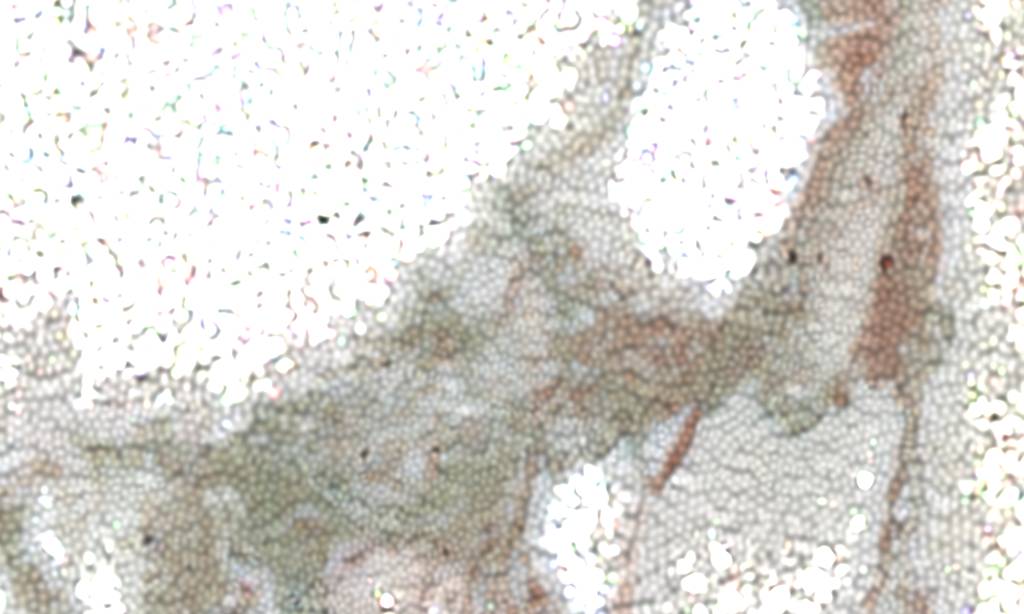} & 
        \includegraphics[width=0.2\linewidth, height=0.12\linewidth]{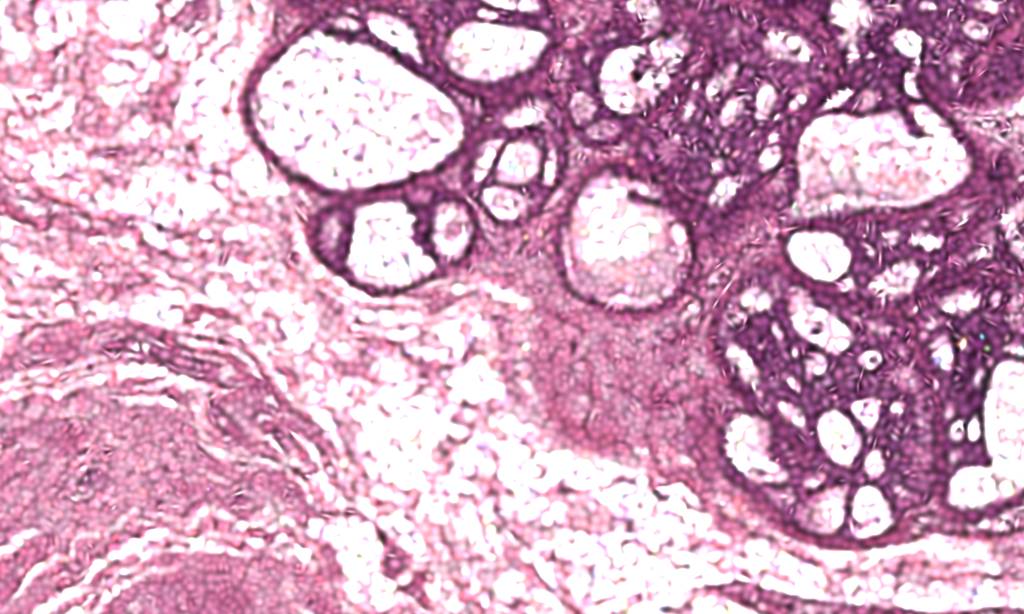} & 
        \includegraphics[width=0.2\linewidth, height=0.12\linewidth]{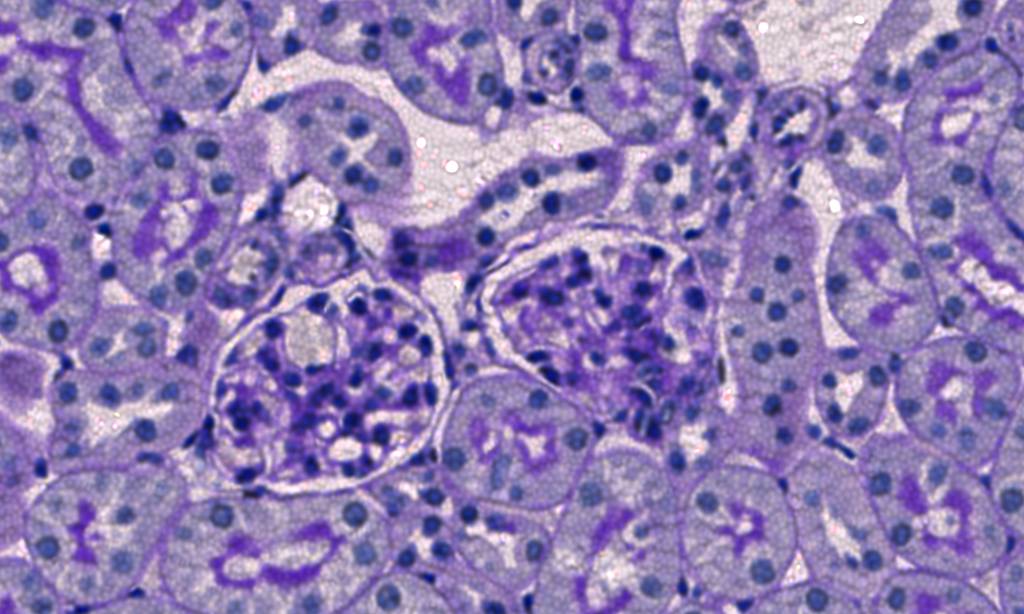} \\

        \rotatebox{90}{ \small \hspace{0.25em} \textbf{SmartSplat}} &
        \includegraphics[width=0.2\linewidth, height=0.12\linewidth]{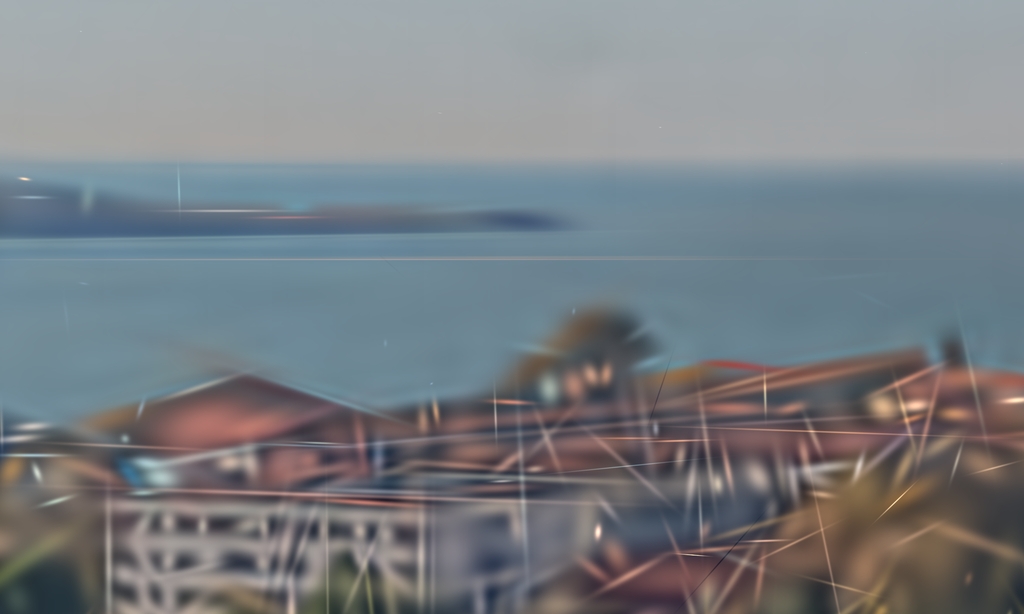} & 
        \includegraphics[width=0.2\linewidth, height=0.12\linewidth]{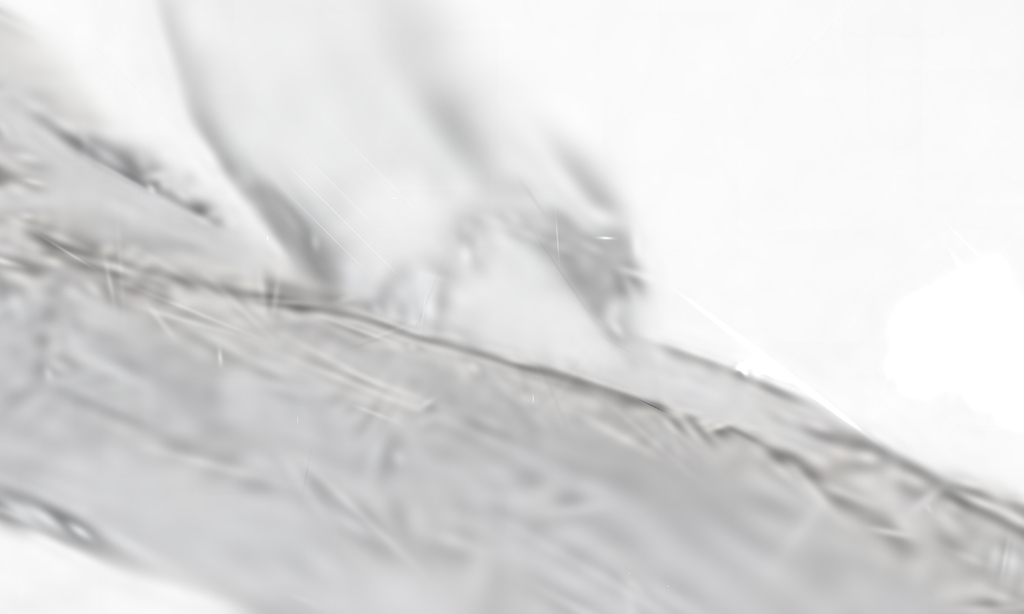} & 
        \includegraphics[width=0.2\linewidth, height=0.12\linewidth]{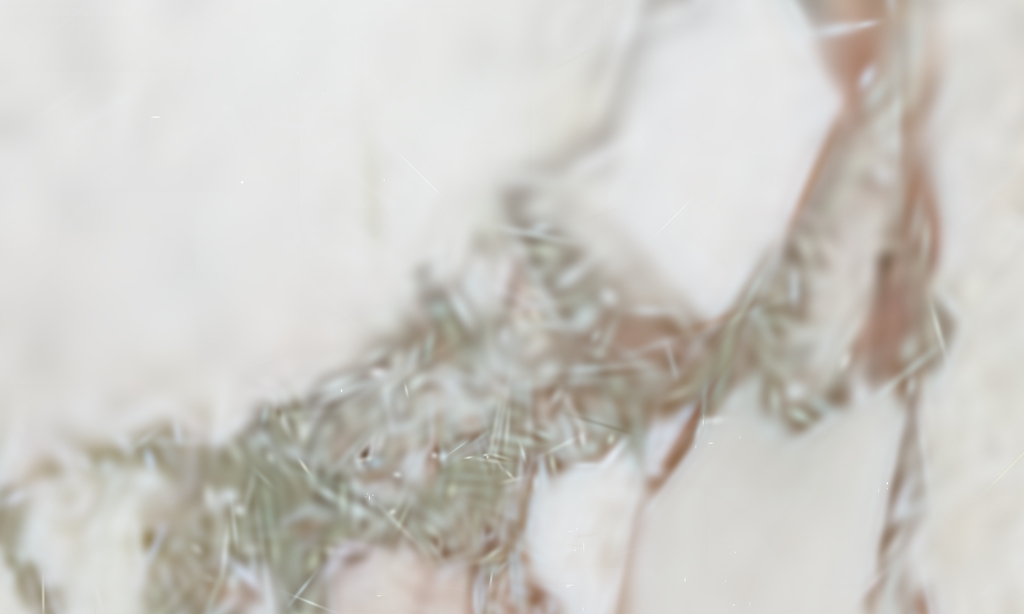} & 
        \includegraphics[width=0.2\linewidth, height=0.12\linewidth]{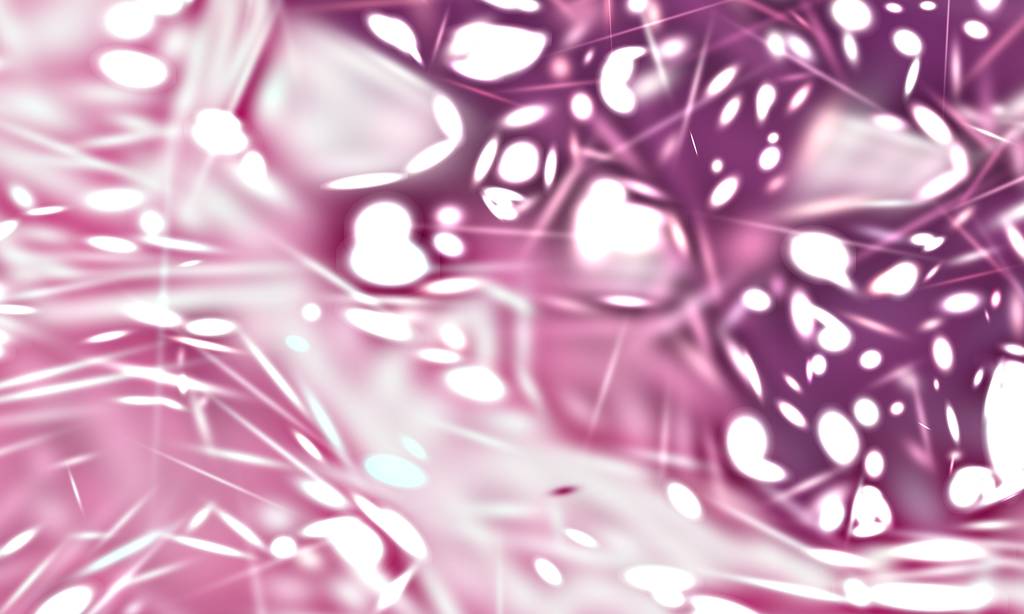} & 
        \includegraphics[width=0.2\linewidth, height=0.12\linewidth]{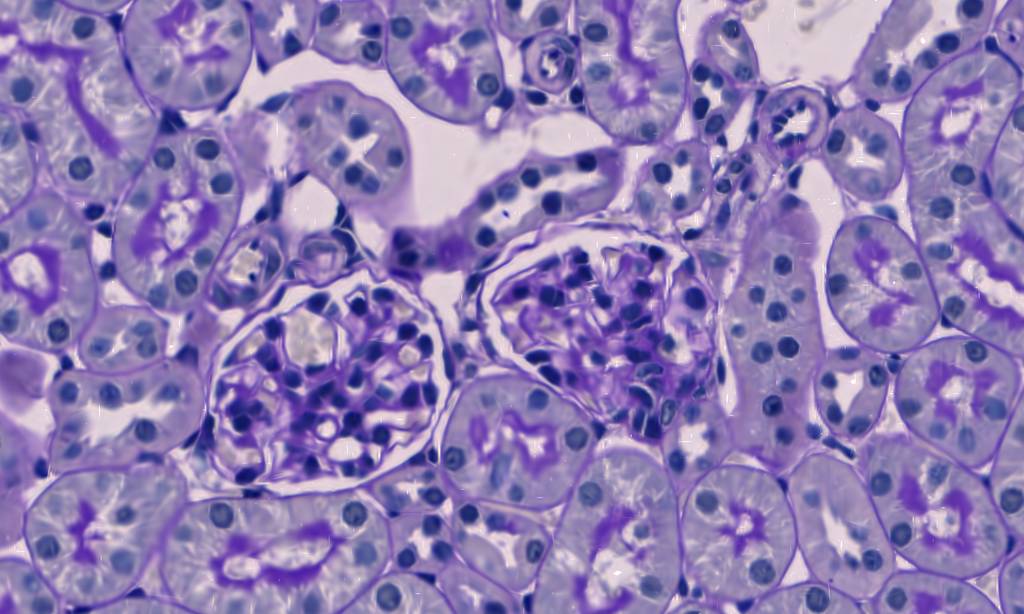} \\

        \rotatebox{90}{ \small \hspace{1em} \textbf{Bicubic}} &
        \includegraphics[width=0.2\linewidth, height=0.12\linewidth]{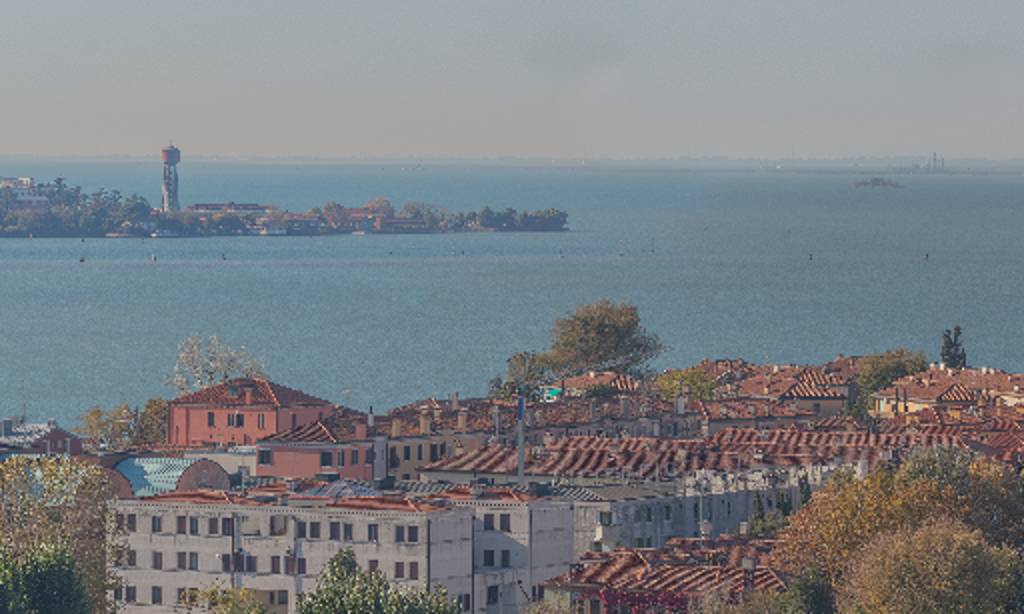} & 
        \includegraphics[width=0.2\linewidth, height=0.12\linewidth]{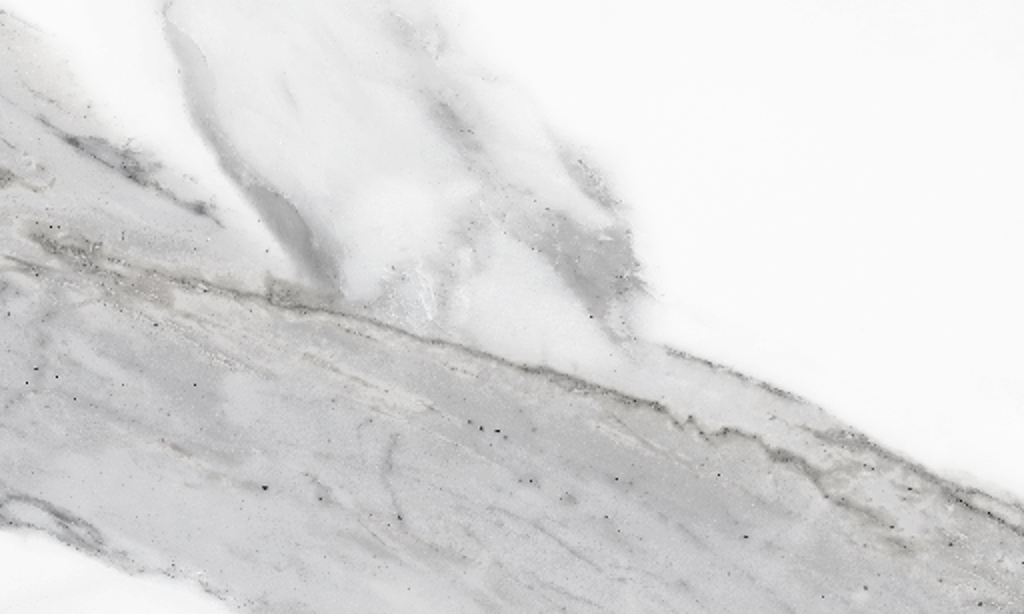} & 
        \includegraphics[width=0.2\linewidth, height=0.12\linewidth]{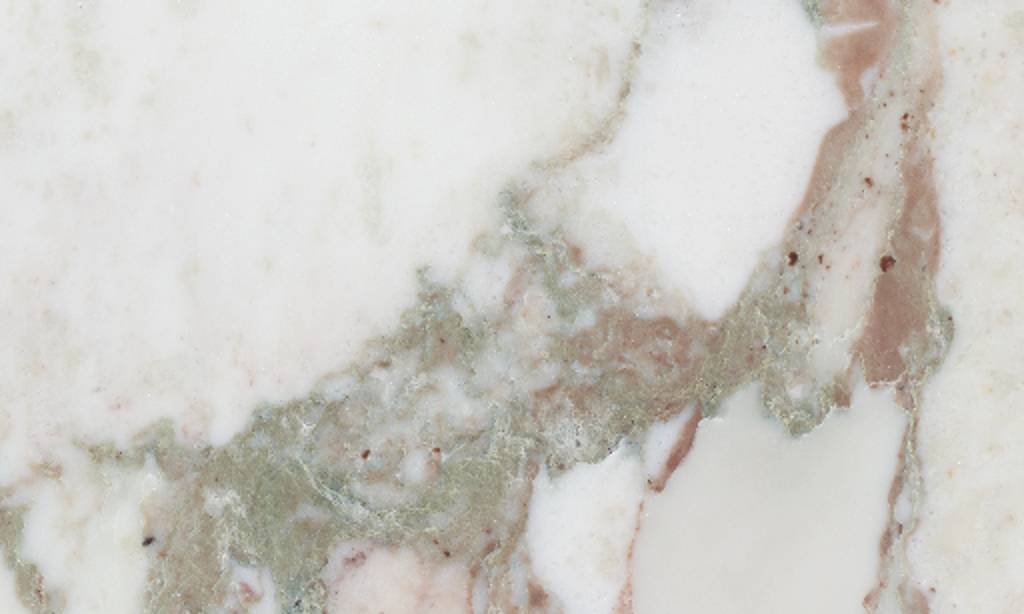} & 
        \includegraphics[width=0.2\linewidth, height=0.12\linewidth]{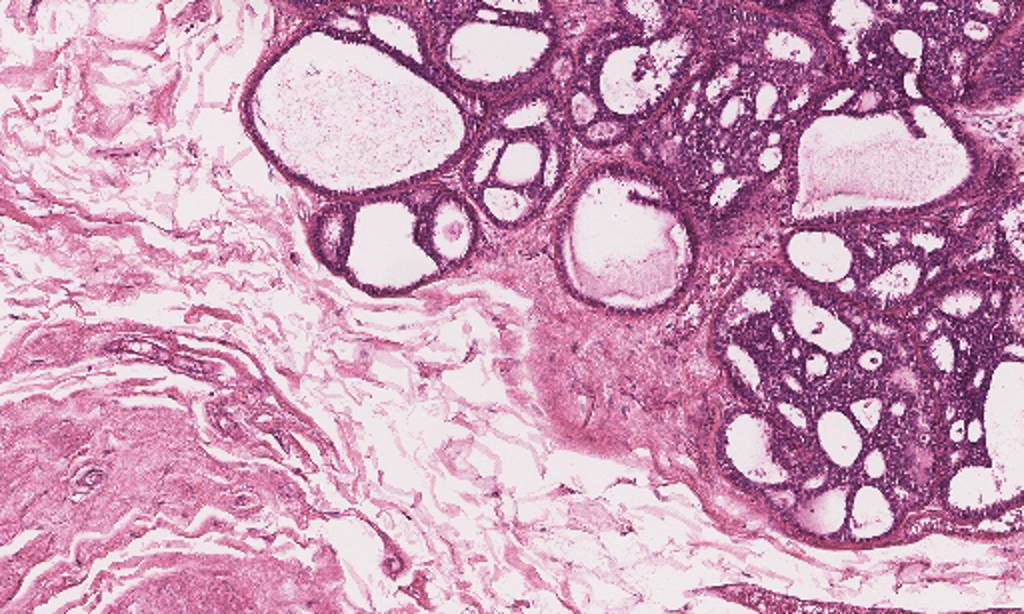} & 
        \includegraphics[width=0.2\linewidth, height=0.12\linewidth]{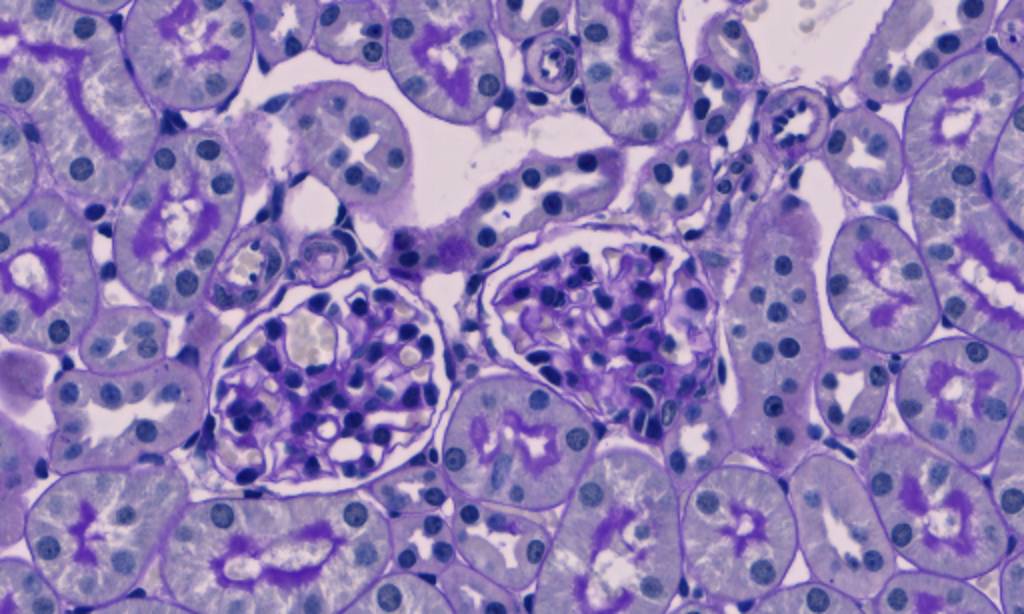} \\

    \end{tabular}}
    \caption{
    \textbf{Additional super resolution $\times4$ qualitatives results on \dataname{}.}
    We show $1228 \times 2048$ crops from the full resolution images.
    }
    \label{fig:suppl_sr_x4_qualitatives_1}

\end{figure*}

\begin{figure*}[ht]
    \centering
    \resizebox{\linewidth}{!}{
    \setlength{\tabcolsep}{1px}
    \begin{tabular}{l ccccc}
        & \textit{Munich} & \textit{Andromeda}& \textit{Italy} & \textit{Lynxsdesign} & \textit{House Of Chores} \\
        \rotatebox{90}{ \small \hspace{2em} \textbf{GT}} &
        \includegraphics[width=0.2\linewidth, height=0.12\linewidth]{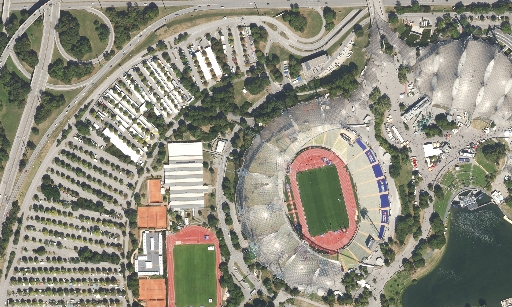} & 
        \includegraphics[width=0.2\linewidth, height=0.12\linewidth]{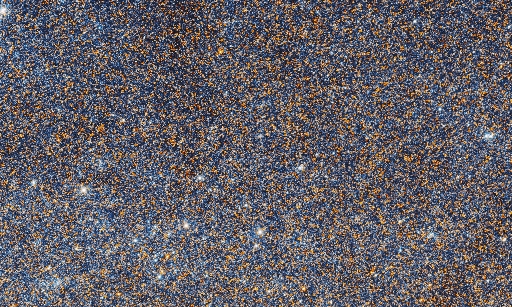} & 
        \includegraphics[width=0.2\linewidth, height=0.12\linewidth]{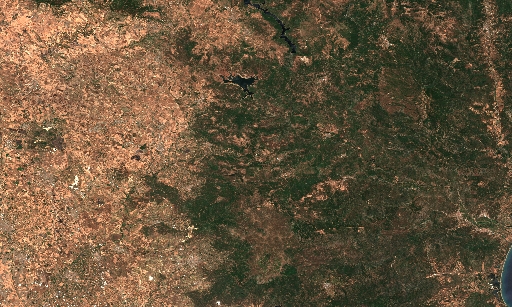} & 
        \includegraphics[width=0.2\linewidth, height=0.12\linewidth]{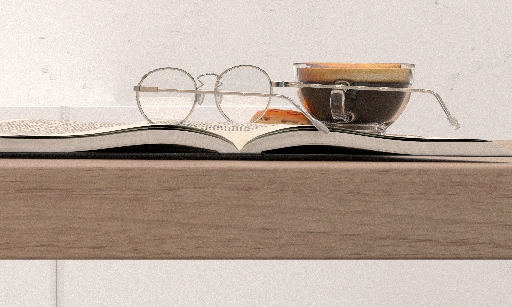} & 
        \includegraphics[width=0.2\linewidth, height=0.12\linewidth]{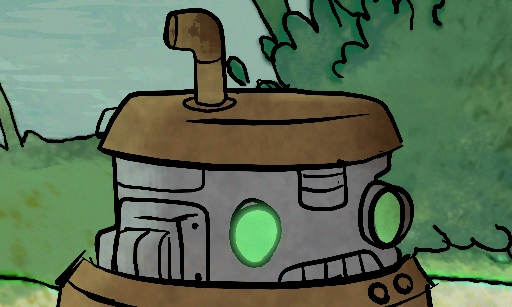} \\

        \rotatebox{90}{ \small \hspace{1.5em} \textbf{SIREN}} &
        \includegraphics[width=0.2\linewidth, height=0.12\linewidth]{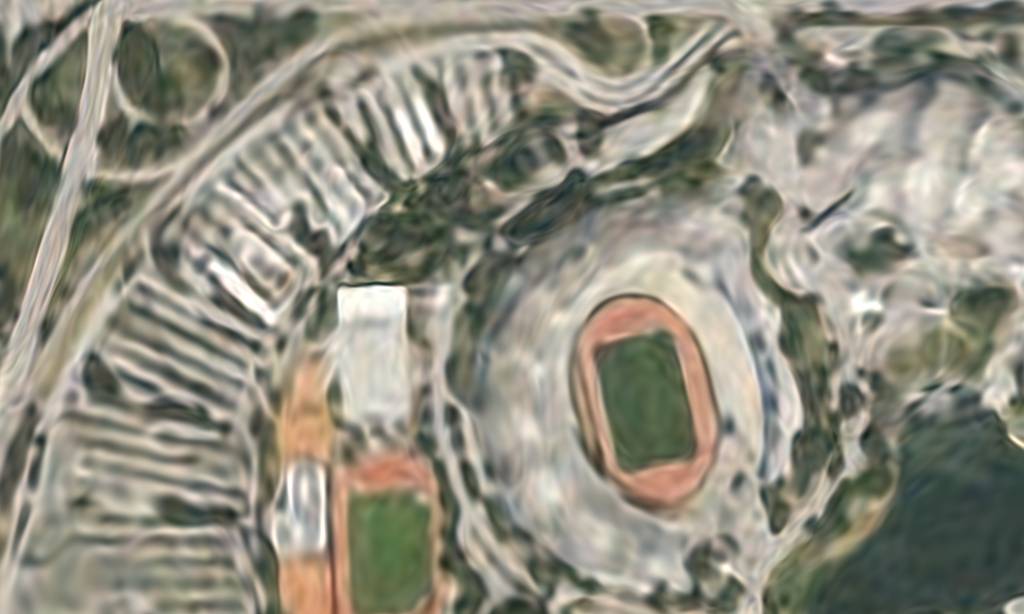} & 
        \includegraphics[width=0.2\linewidth, height=0.12\linewidth]{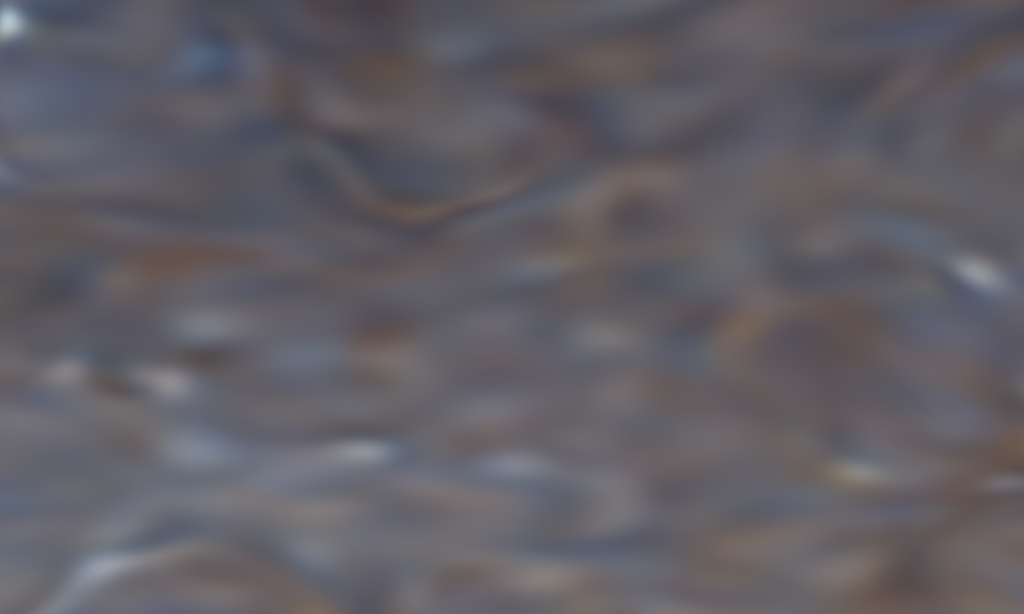} & 
        \includegraphics[width=0.2\linewidth, height=0.12\linewidth]{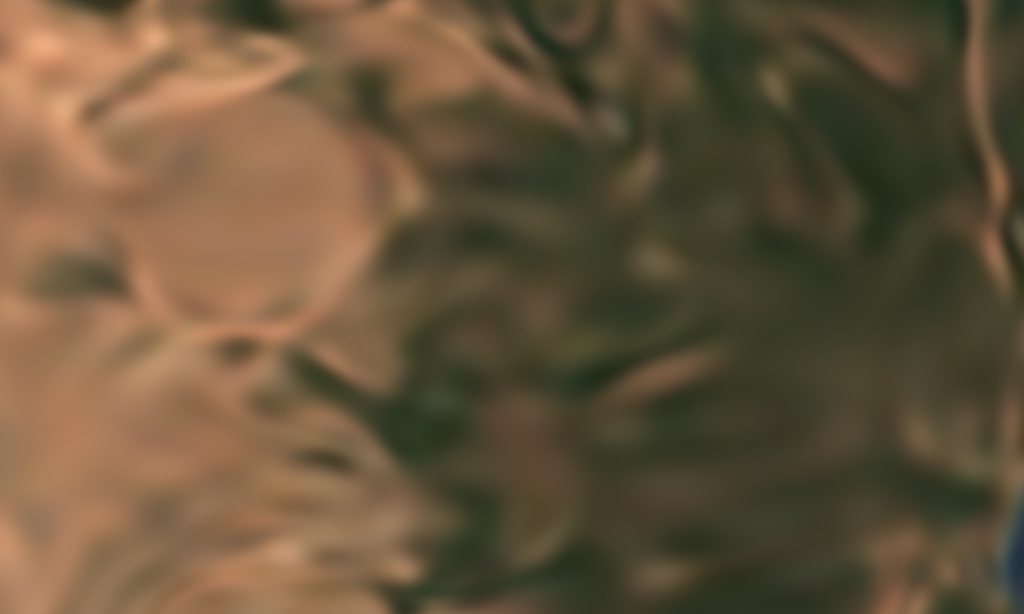} & 
        \includegraphics[width=0.2\linewidth, height=0.12\linewidth]{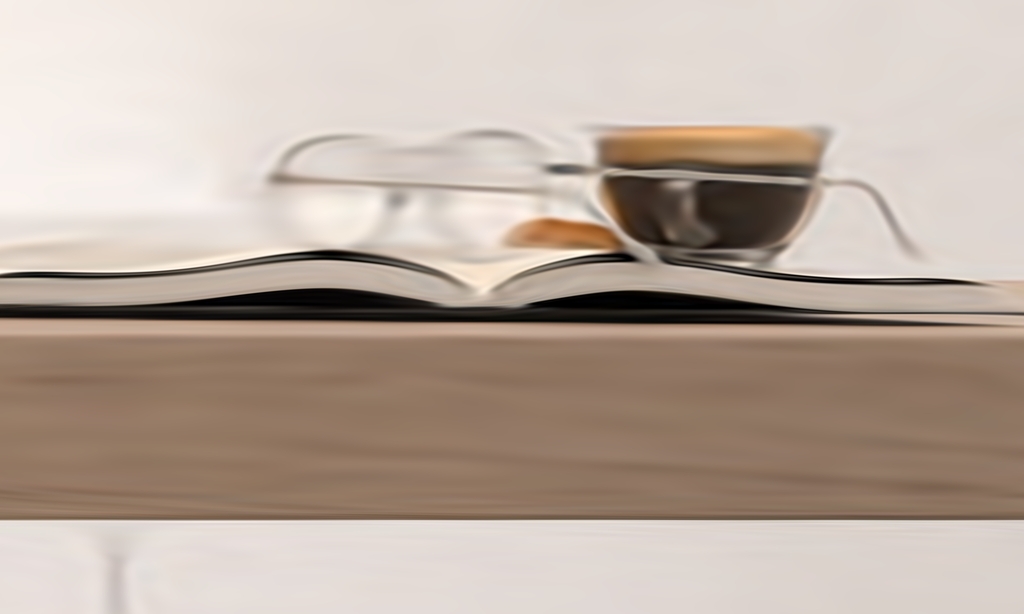} & 
        \includegraphics[width=0.2\linewidth, height=0.12\linewidth]{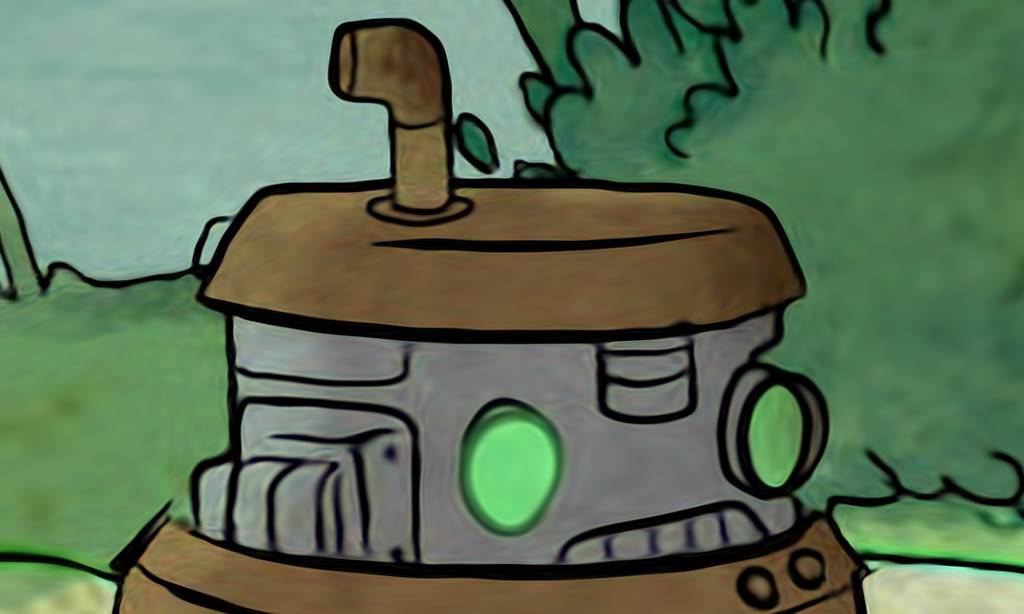} \\

        \rotatebox{90}{ \small \hspace{1em} \textbf{FDHO}} &
        \includegraphics[width=0.2\linewidth, height=0.12\linewidth]{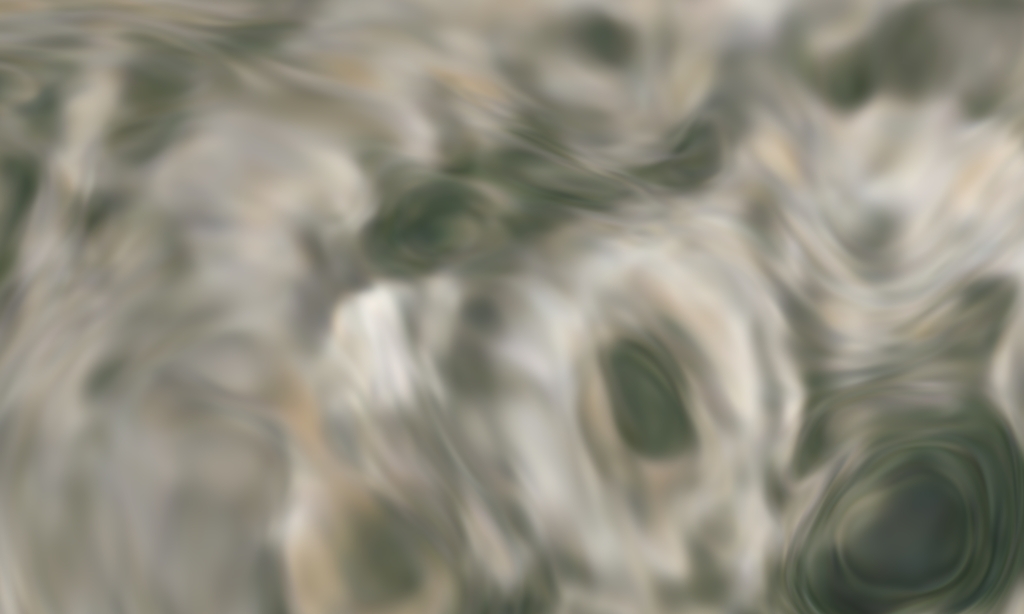} & 
        \includegraphics[width=0.2\linewidth, height=0.12\linewidth]{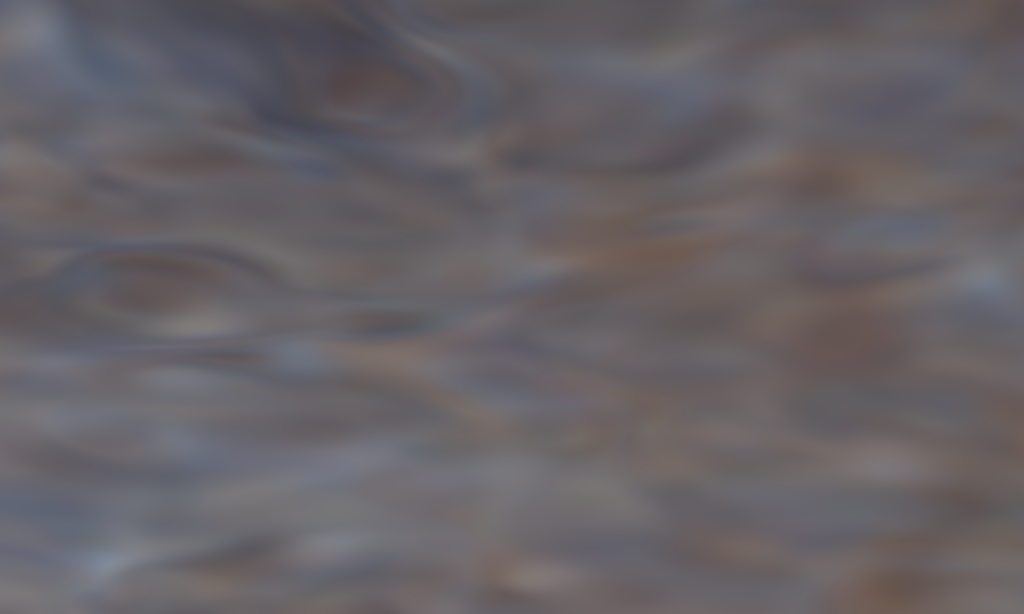} & 
        \includegraphics[width=0.2\linewidth, height=0.12\linewidth]{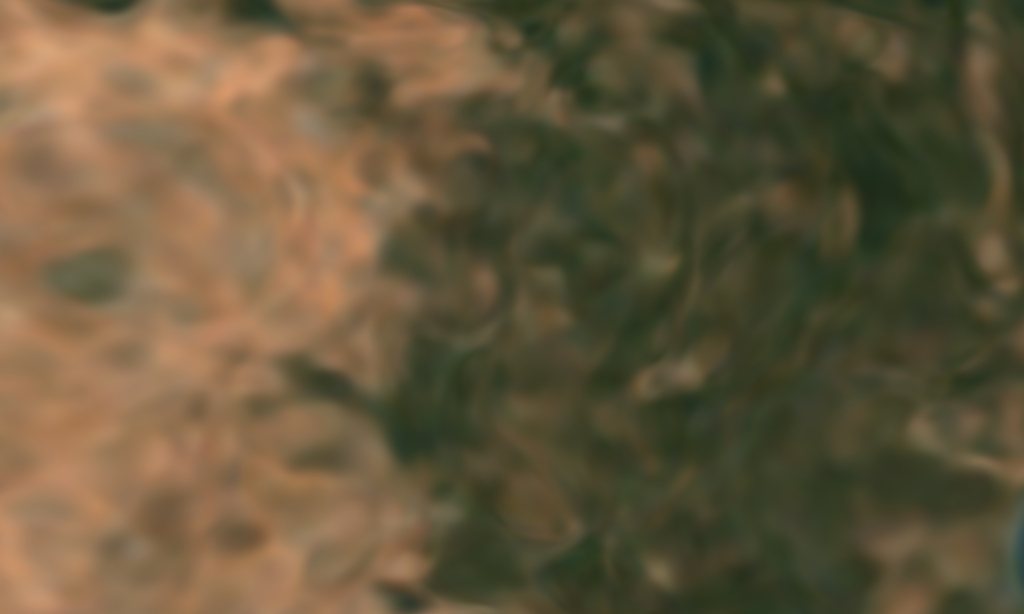} & 
        \includegraphics[width=0.2\linewidth, height=0.12\linewidth]{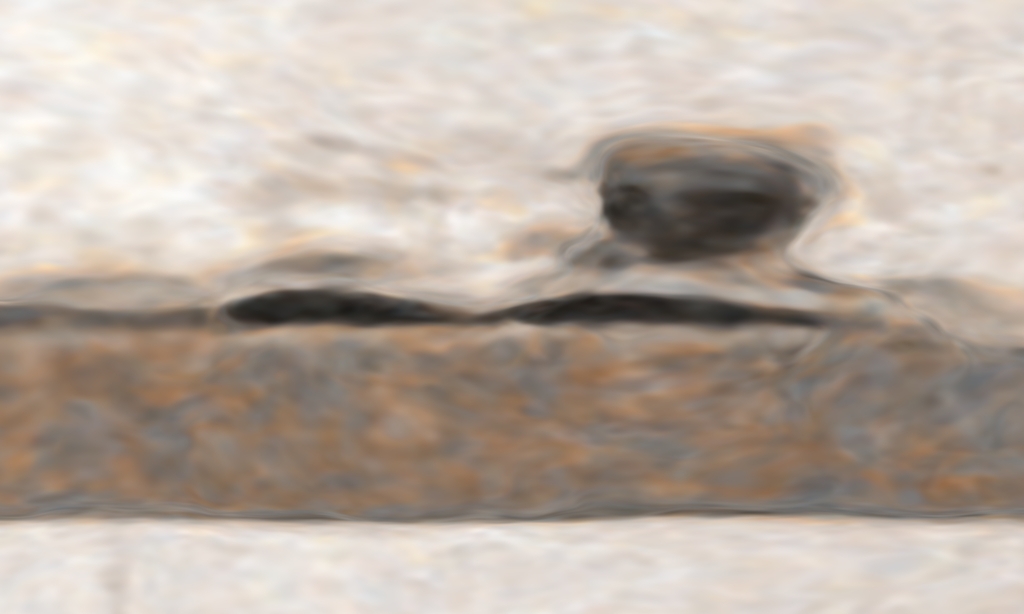} & 
        \includegraphics[width=0.2\linewidth, height=0.12\linewidth]{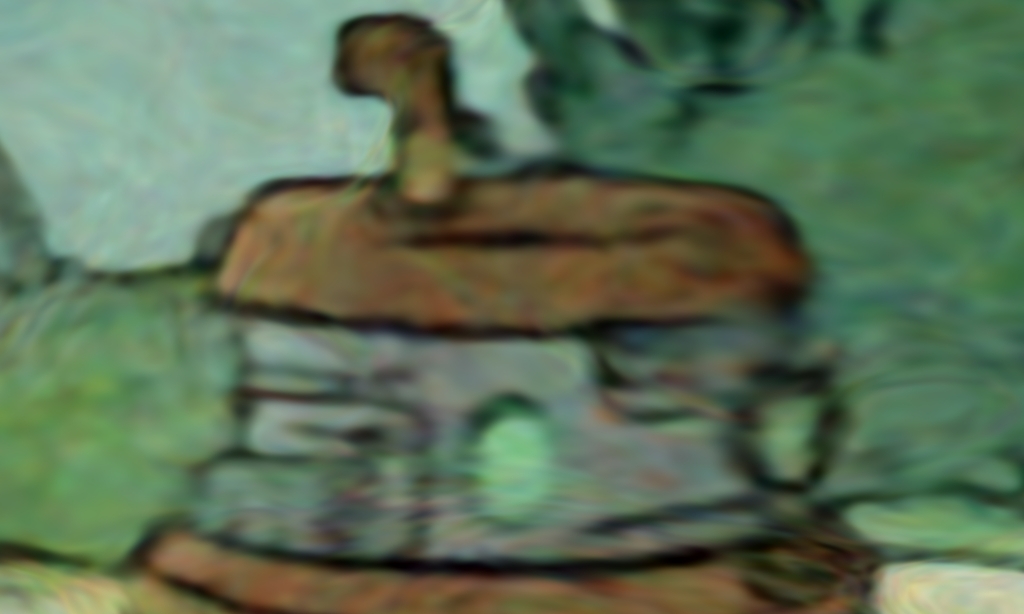} \\

        \rotatebox{90}{ \small \hspace{1em} \textbf{FINER}} &
        \includegraphics[width=0.2\linewidth, height=0.12\linewidth]{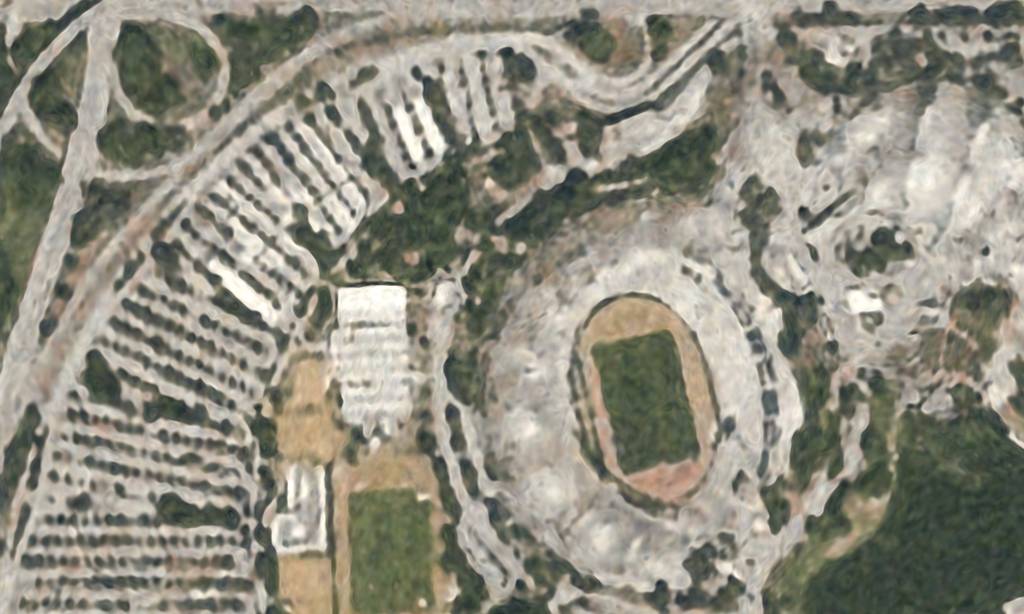} & 
        \includegraphics[width=0.2\linewidth, height=0.12\linewidth]{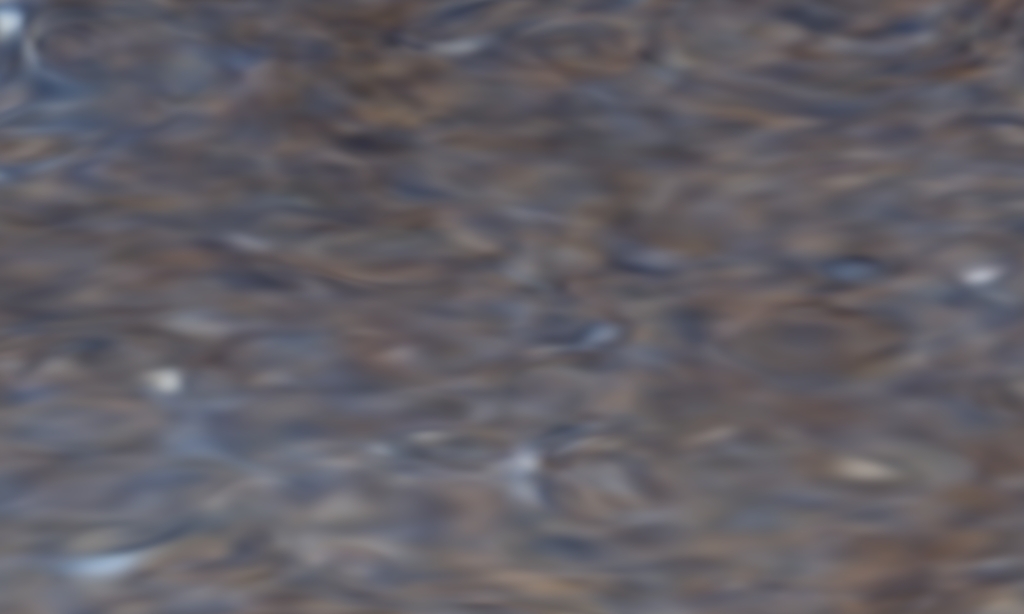} & 
        \includegraphics[width=0.2\linewidth, height=0.12\linewidth]{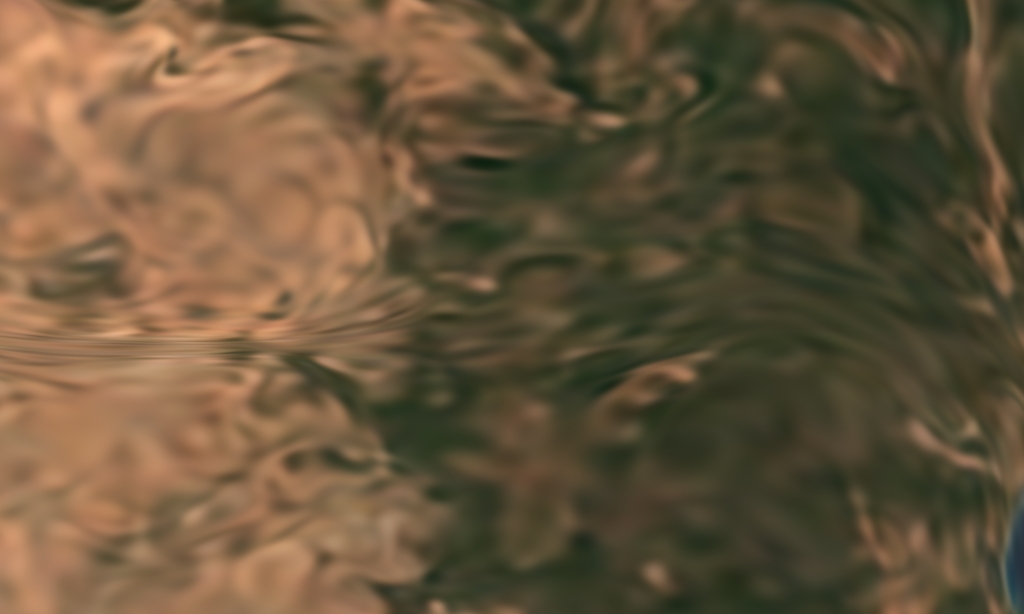} & 
        \includegraphics[width=0.2\linewidth, height=0.12\linewidth]{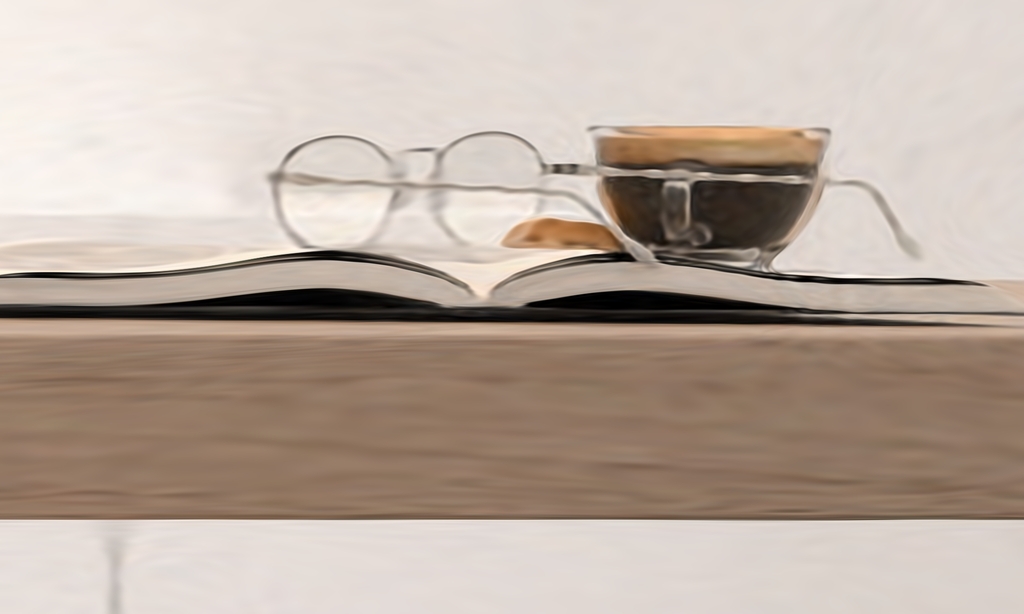} & 
        \includegraphics[width=0.2\linewidth, height=0.12\linewidth]{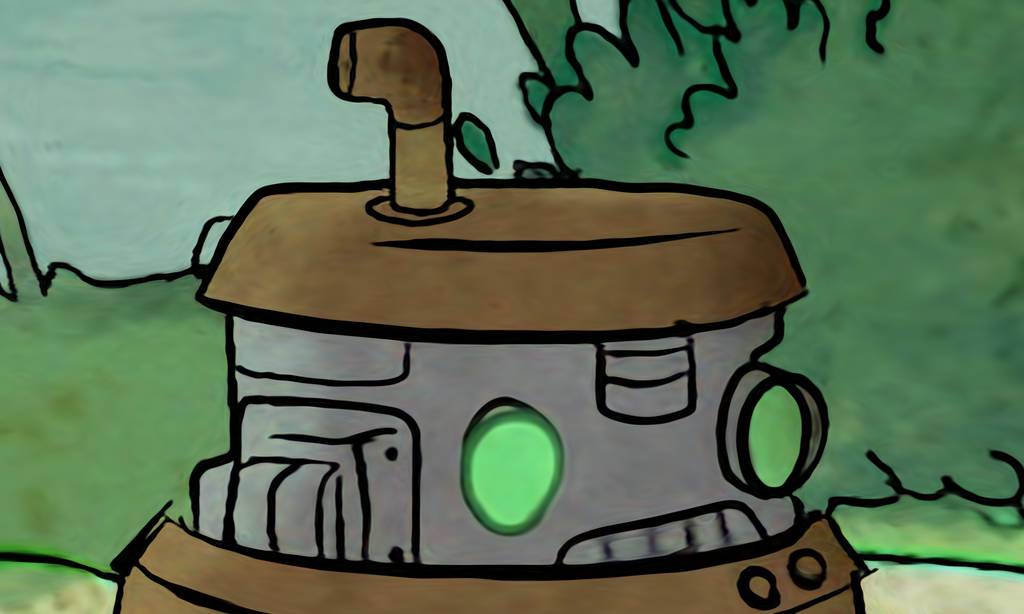} \\

        \rotatebox{90}{ \small \hspace{1.25em} \textbf{I-NGP}} &
        \includegraphics[width=0.2\linewidth, height=0.12\linewidth]{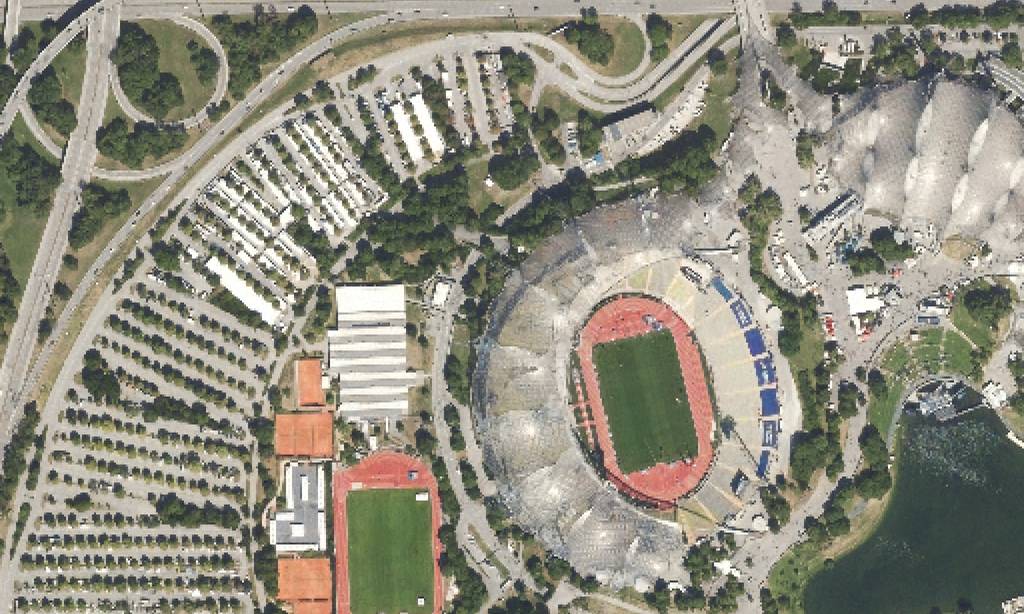} & 
        \includegraphics[width=0.2\linewidth, height=0.12\linewidth]{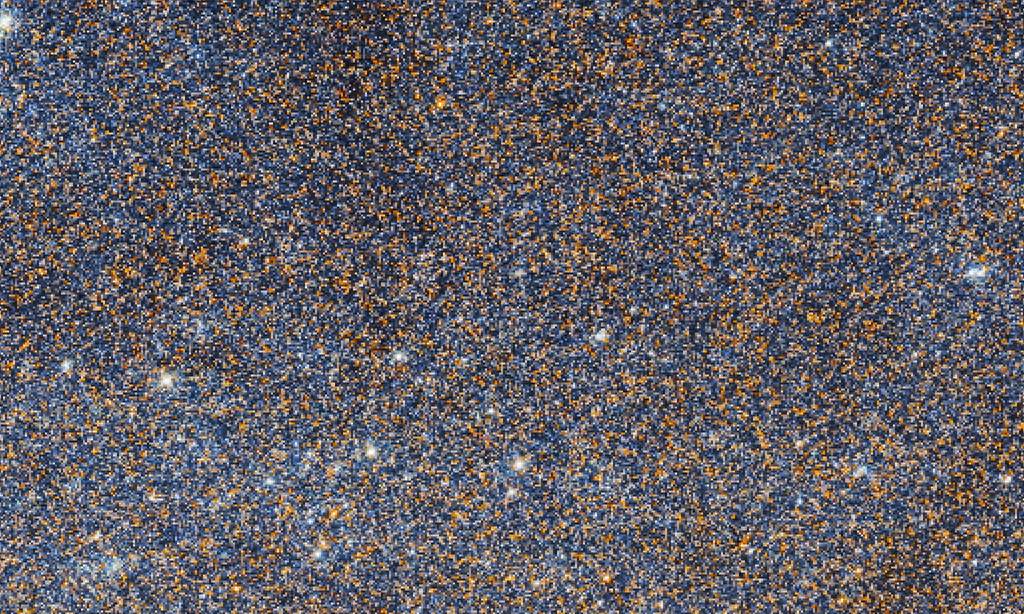} & 
        \includegraphics[width=0.2\linewidth, height=0.12\linewidth]{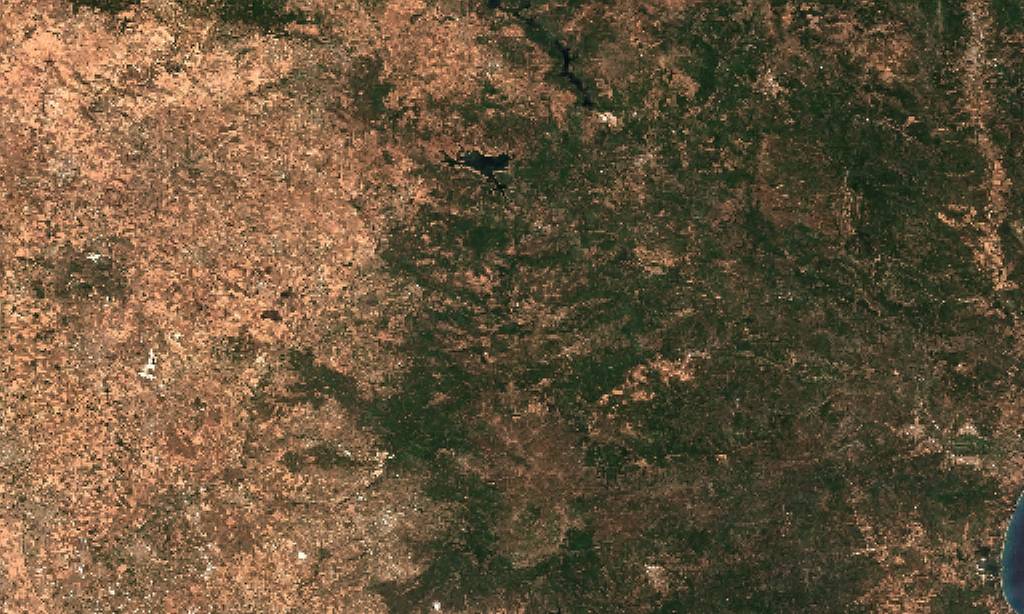} & 
        \includegraphics[width=0.2\linewidth, height=0.12\linewidth]{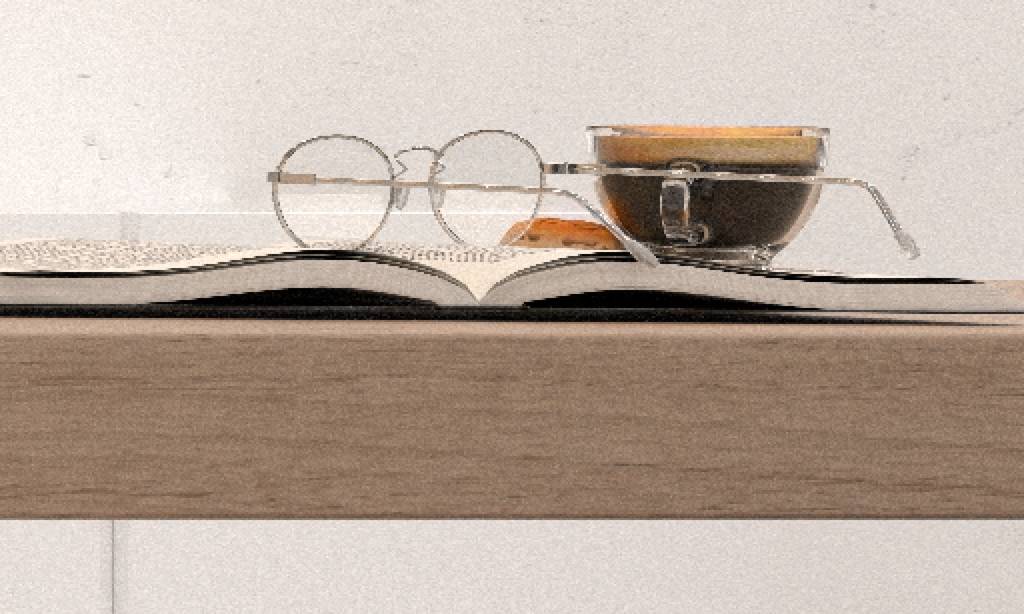} & 
        \includegraphics[width=0.2\linewidth, height=0.12\linewidth]{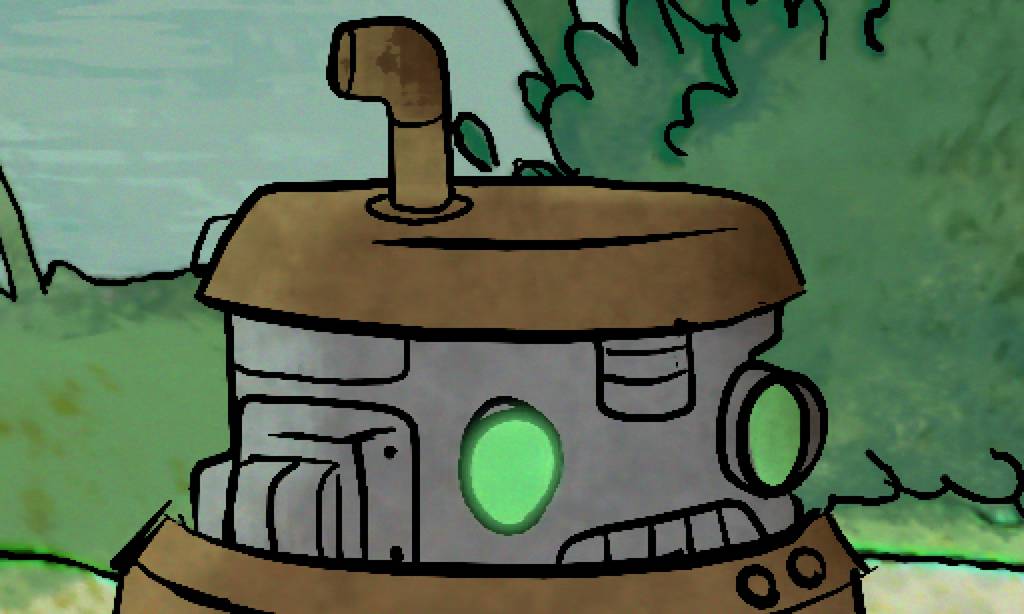} \\

        \rotatebox{90}{ \small \hspace{1em} \textbf{NeuRBF}} &
        \includegraphics[width=0.2\linewidth, height=0.12\linewidth]{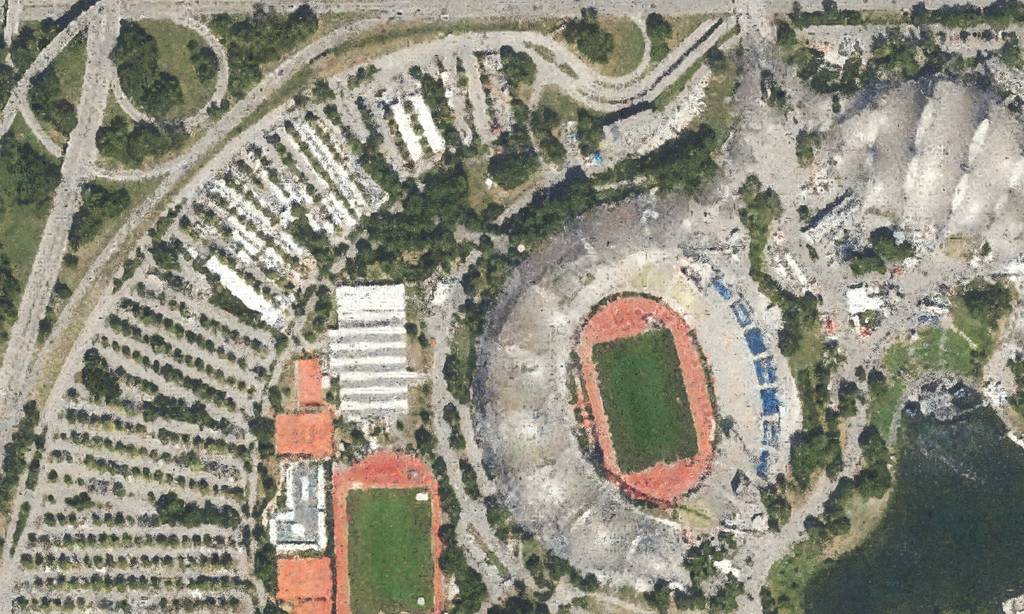} & 
        \includegraphics[width=0.2\linewidth, height=0.12\linewidth]{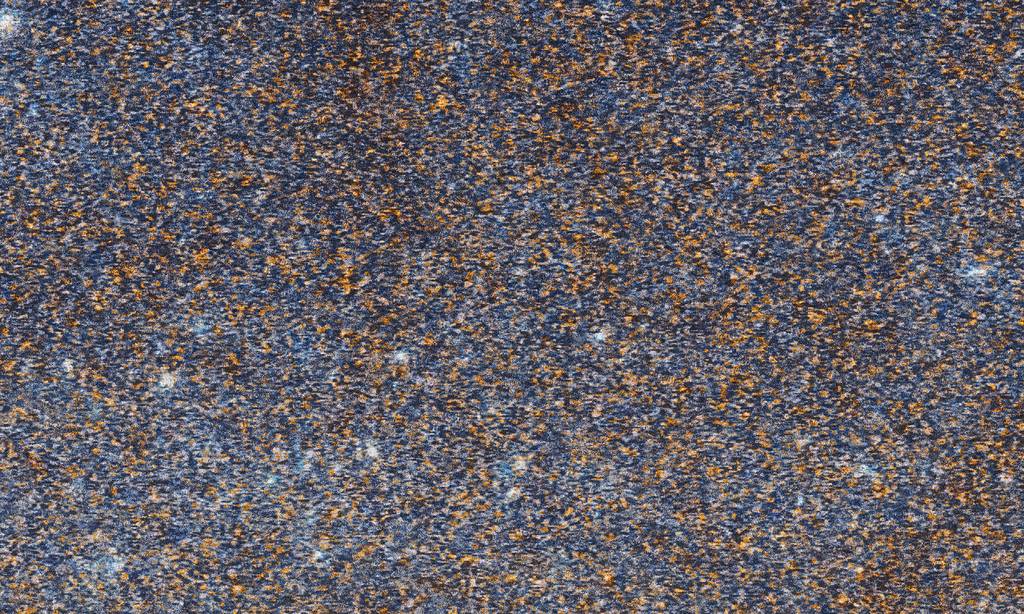} & 
        \includegraphics[width=0.2\linewidth, height=0.12\linewidth]{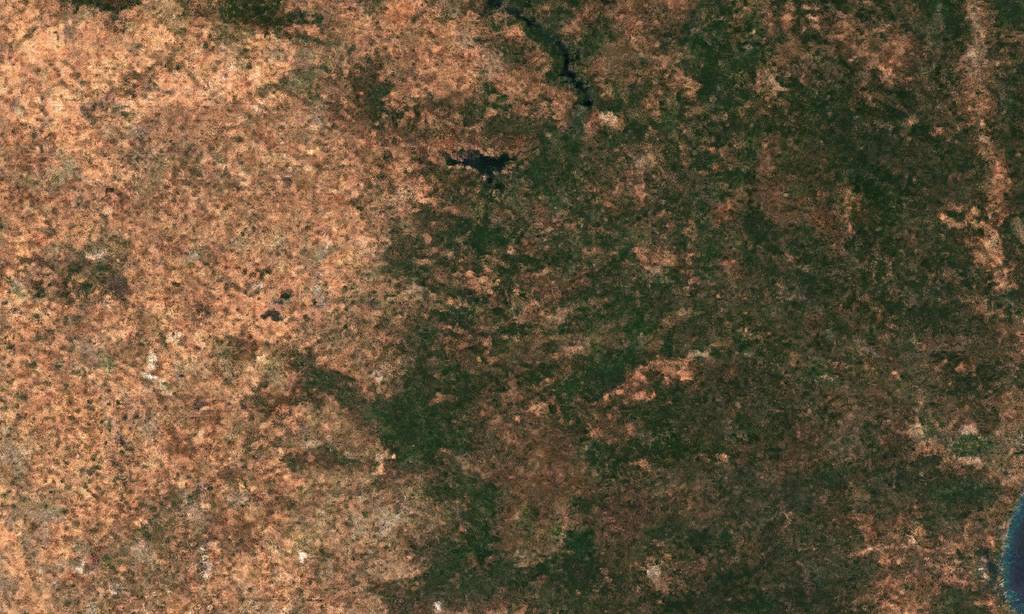} & 
        \includegraphics[width=0.2\linewidth, height=0.12\linewidth]{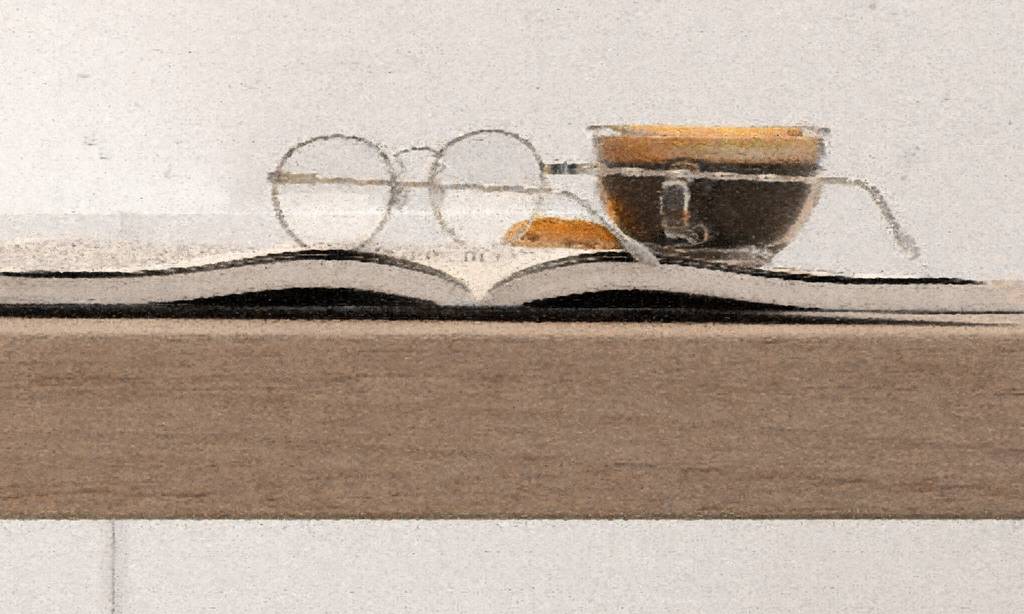} & 
        \includegraphics[width=0.2\linewidth, height=0.12\linewidth]{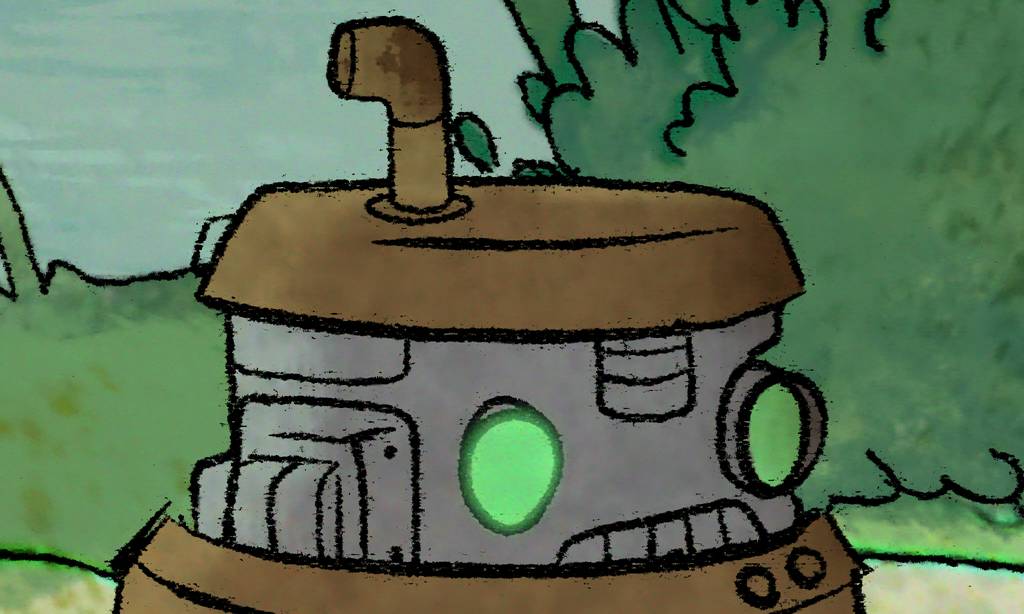} \\

        \rotatebox{90}{ \small \hspace{0.25em} \textbf{MetricGrids}} &
        \includegraphics[width=0.2\linewidth, height=0.12\linewidth]{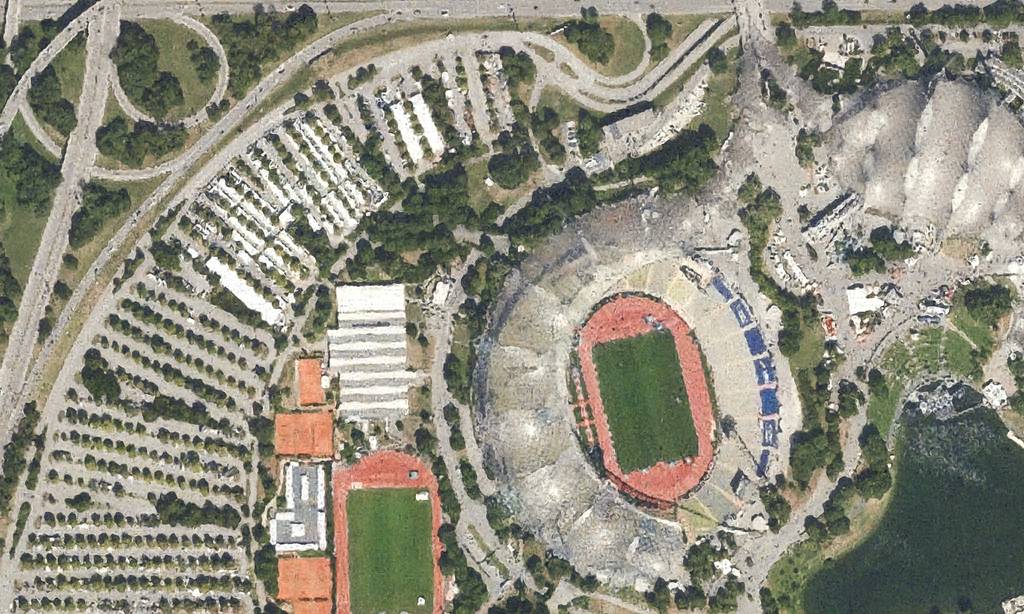} & 
        \includegraphics[width=0.2\linewidth, height=0.12\linewidth]{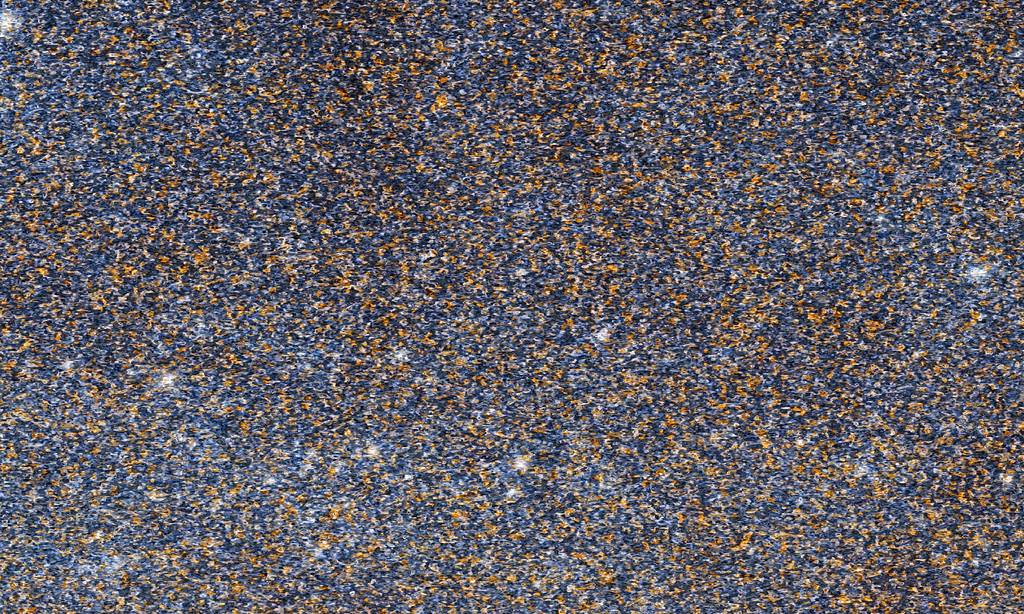} & 
        \includegraphics[width=0.2\linewidth, height=0.12\linewidth]{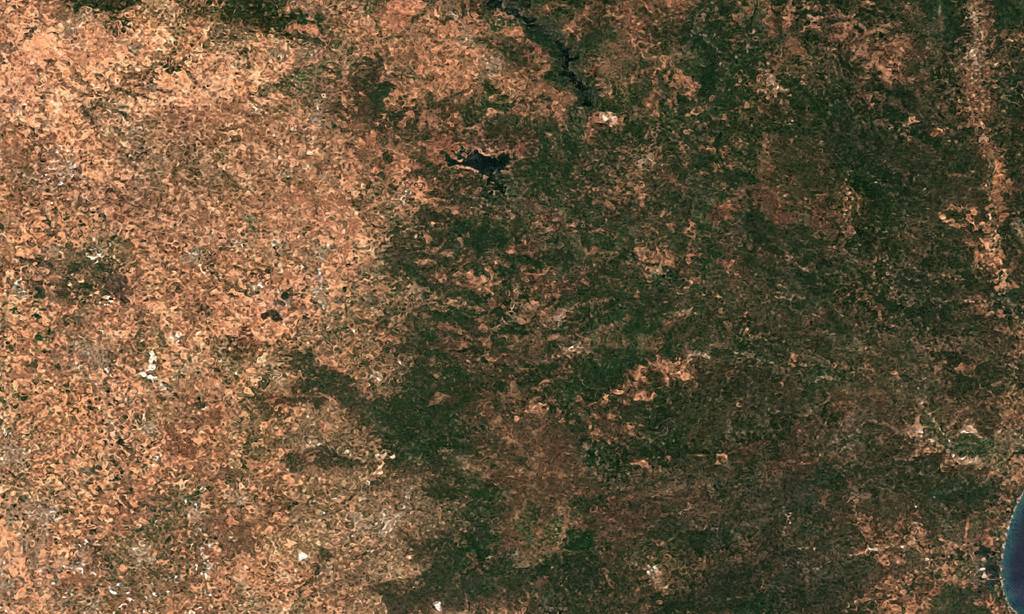} & 
        \includegraphics[width=0.2\linewidth, height=0.12\linewidth]{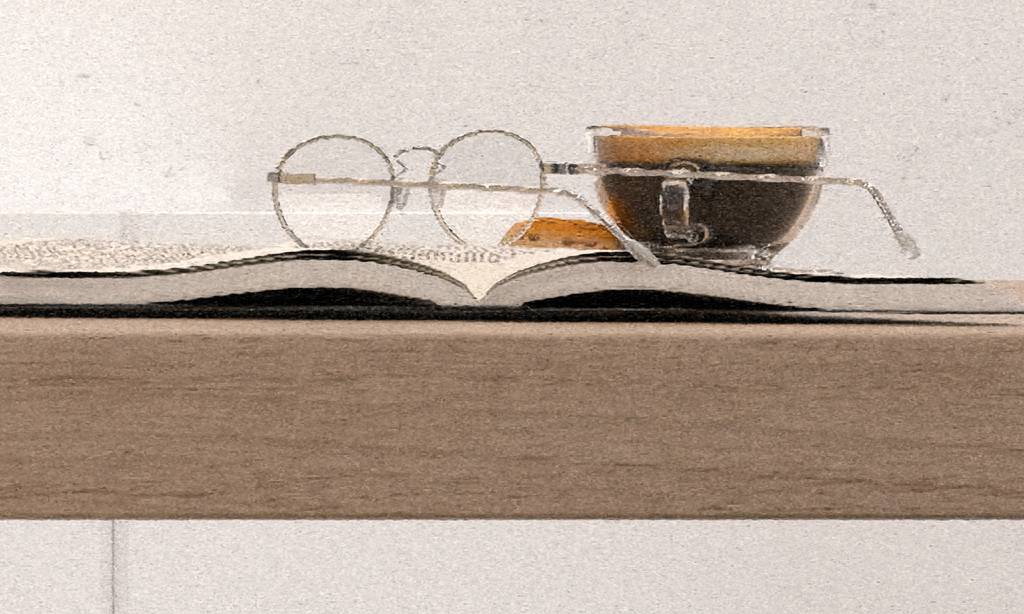} & 
        \includegraphics[width=0.2\linewidth, height=0.12\linewidth]{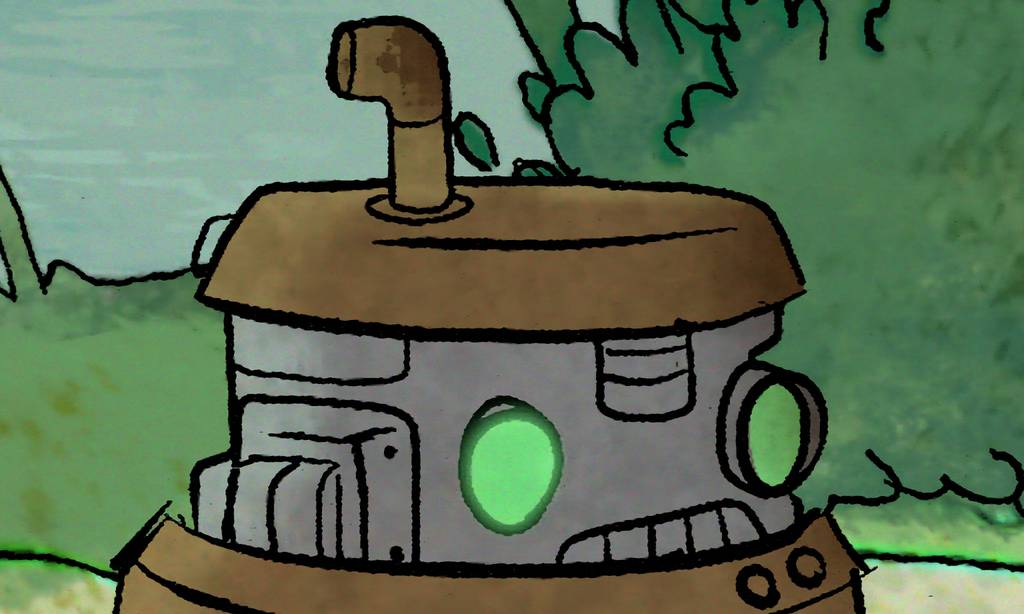} \\

        \rotatebox{90}{ \small \hspace{0.5em} \textbf{GaussImg}} &
        \includegraphics[width=0.2\linewidth, height=0.12\linewidth]{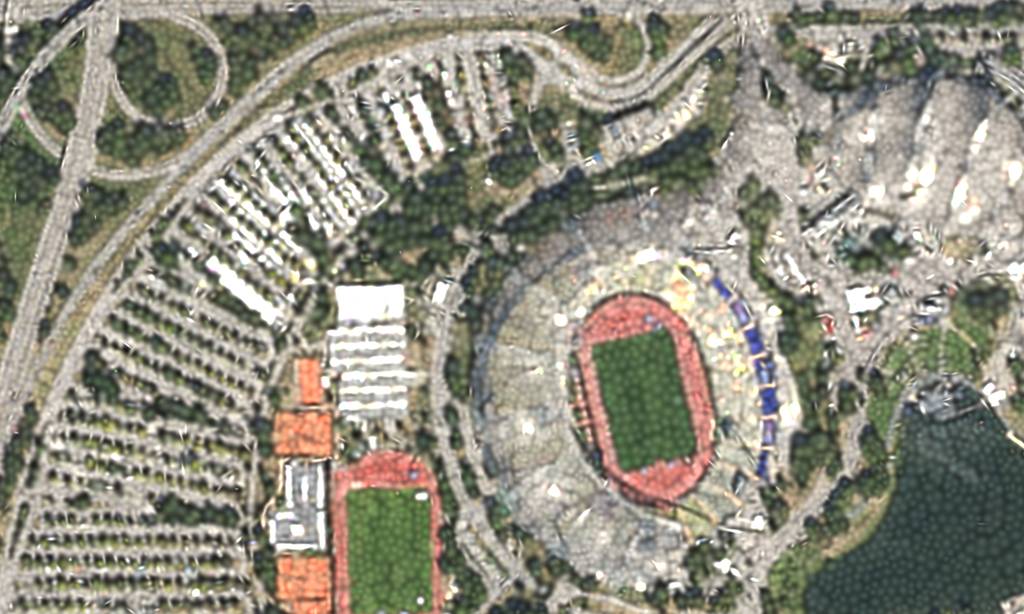} & 
        \includegraphics[width=0.2\linewidth, height=0.12\linewidth]{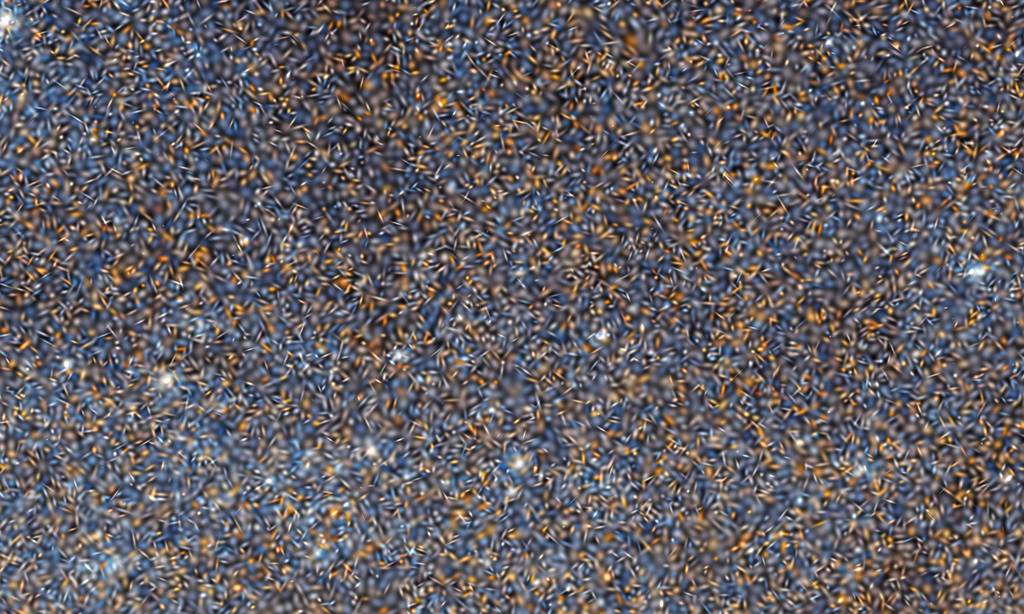} & 
        \includegraphics[width=0.2\linewidth, height=0.12\linewidth]{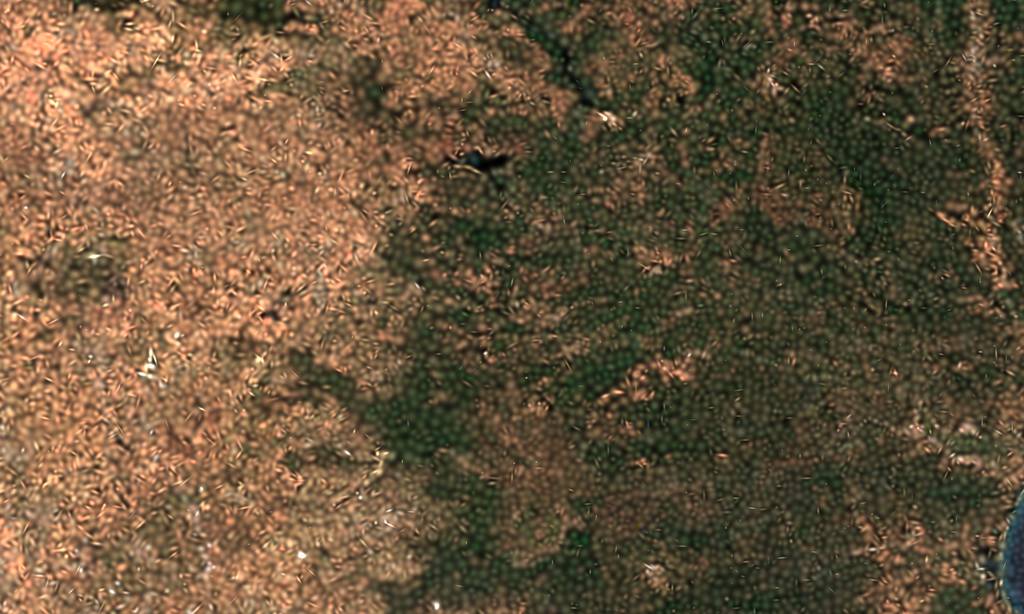} & 
        \includegraphics[width=0.2\linewidth, height=0.12\linewidth]{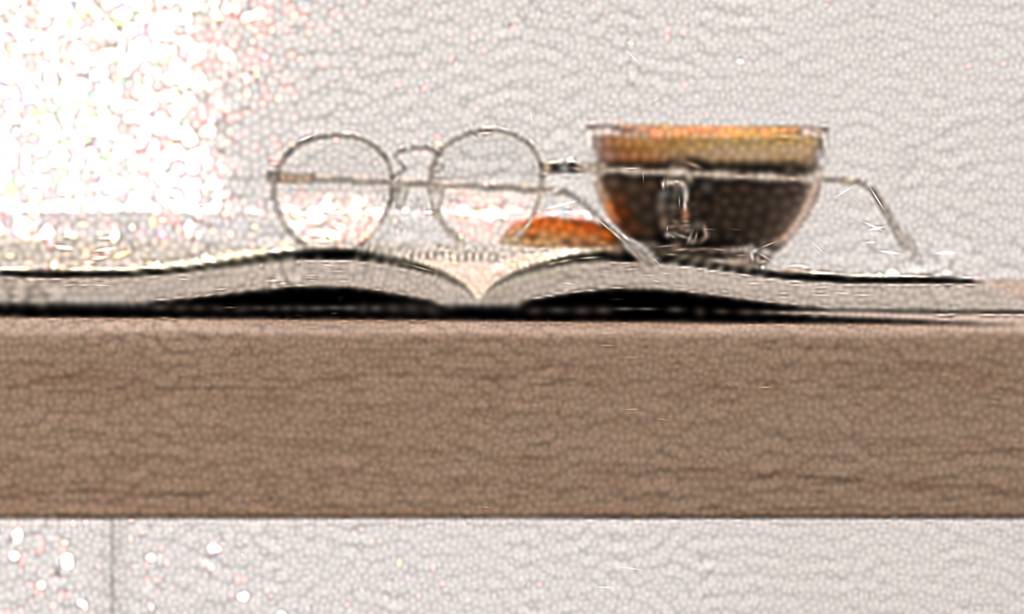} & 
        \includegraphics[width=0.2\linewidth, height=0.12\linewidth]{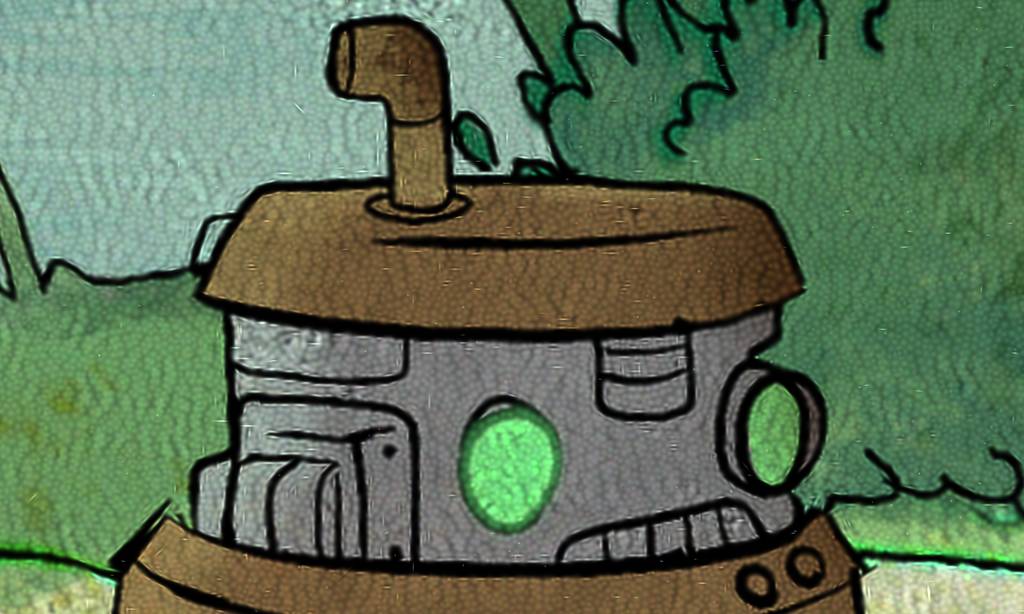} \\

        \rotatebox{90}{ \small \hspace{0.25em} \textbf{SmartSplat}} &
        \includegraphics[width=0.2\linewidth, height=0.12\linewidth]{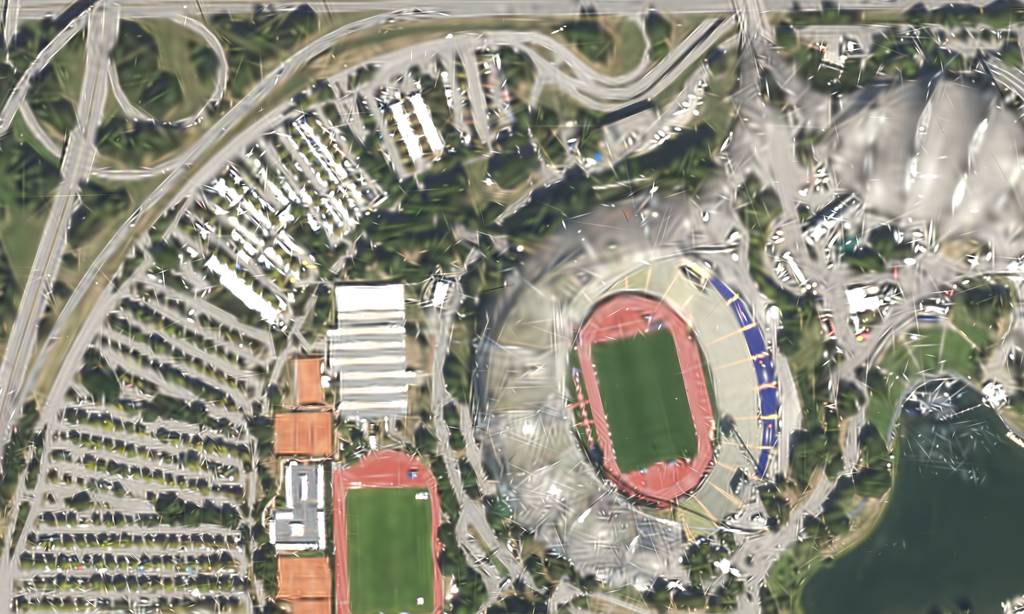} & 
        \includegraphics[width=0.2\linewidth, height=0.12\linewidth]{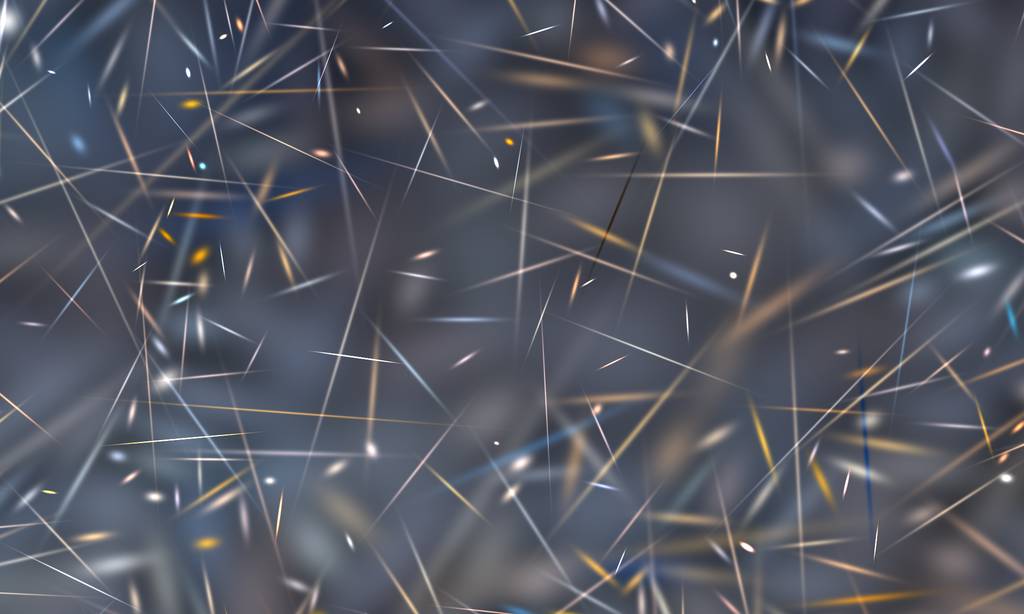} & 
        \includegraphics[width=0.2\linewidth, height=0.12\linewidth]{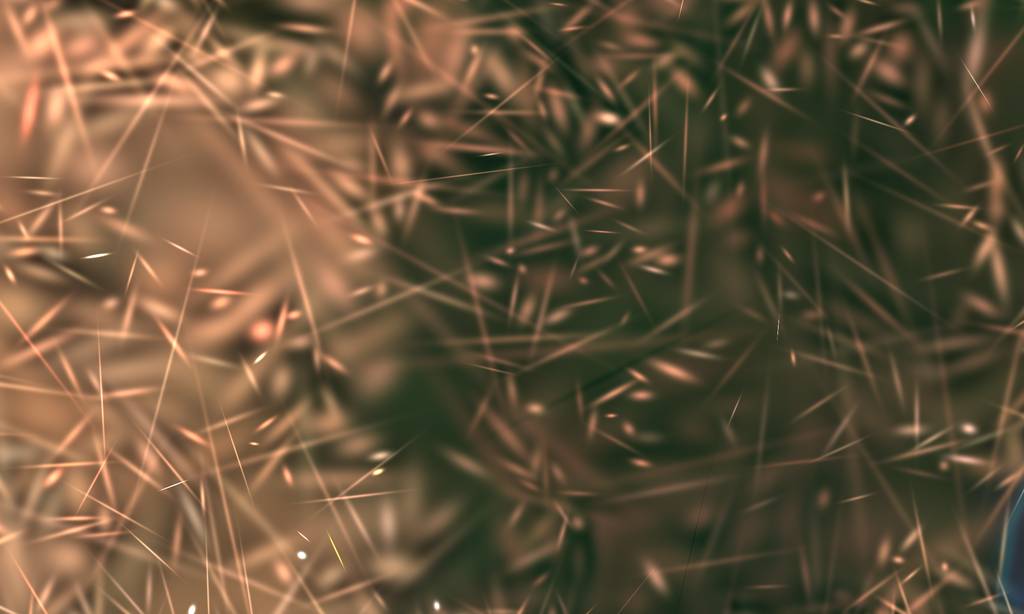} & 
        \includegraphics[width=0.2\linewidth, height=0.12\linewidth]{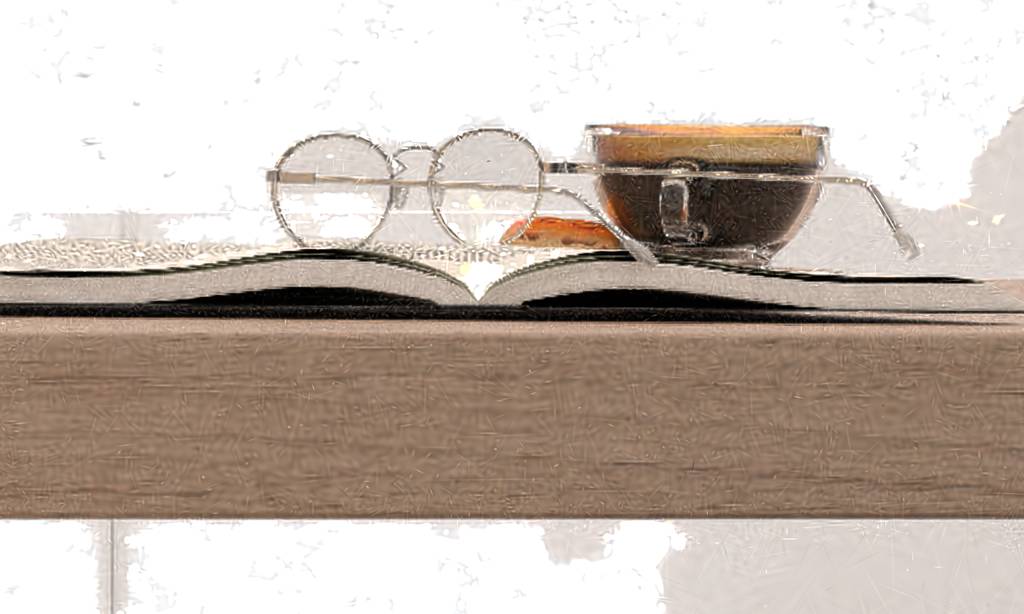} & 
        \includegraphics[width=0.2\linewidth, height=0.12\linewidth]{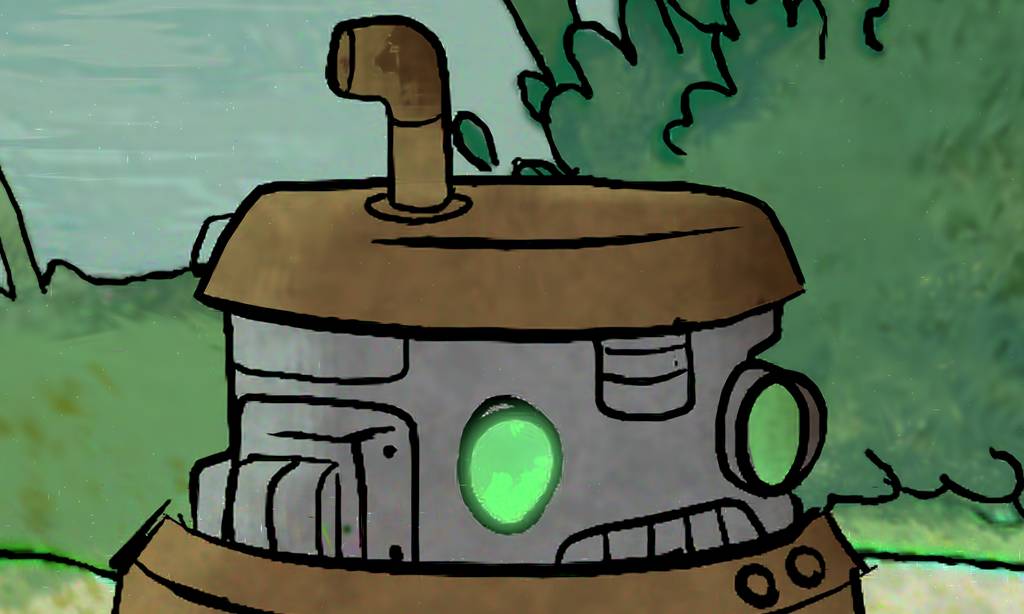} \\

        \rotatebox{90}{ \small \hspace{1em} \textbf{Bicubic}} &
        \includegraphics[width=0.2\linewidth, height=0.12\linewidth]{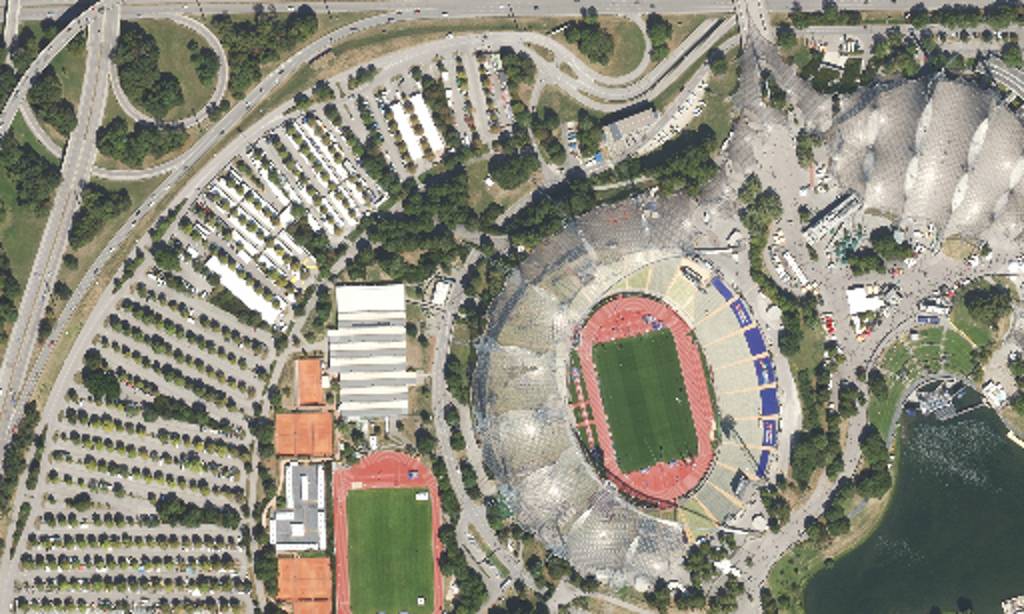} & 
        \includegraphics[width=0.2\linewidth, height=0.12\linewidth]{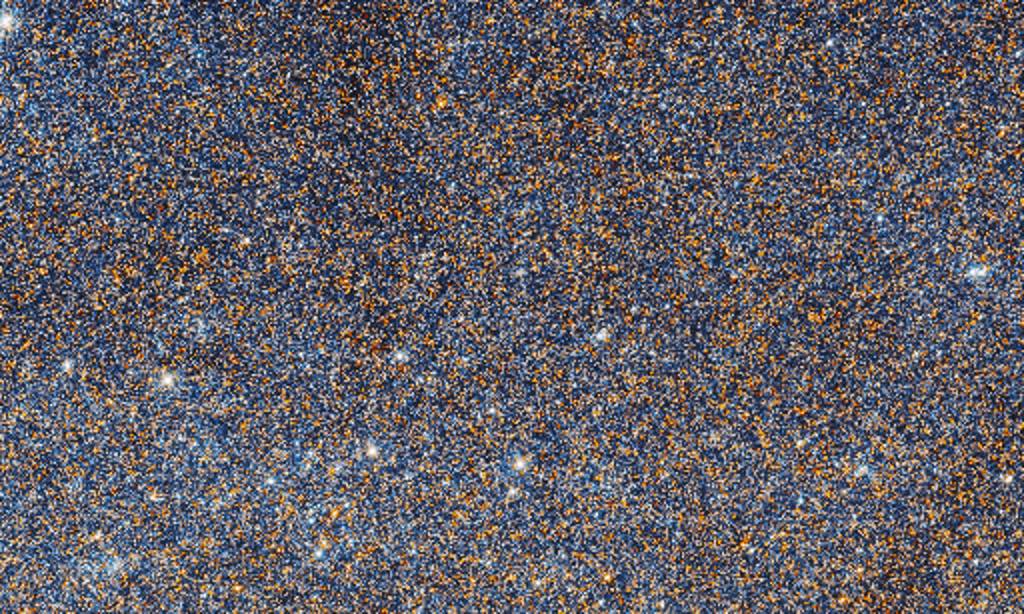} & 
        \includegraphics[width=0.2\linewidth, height=0.12\linewidth]{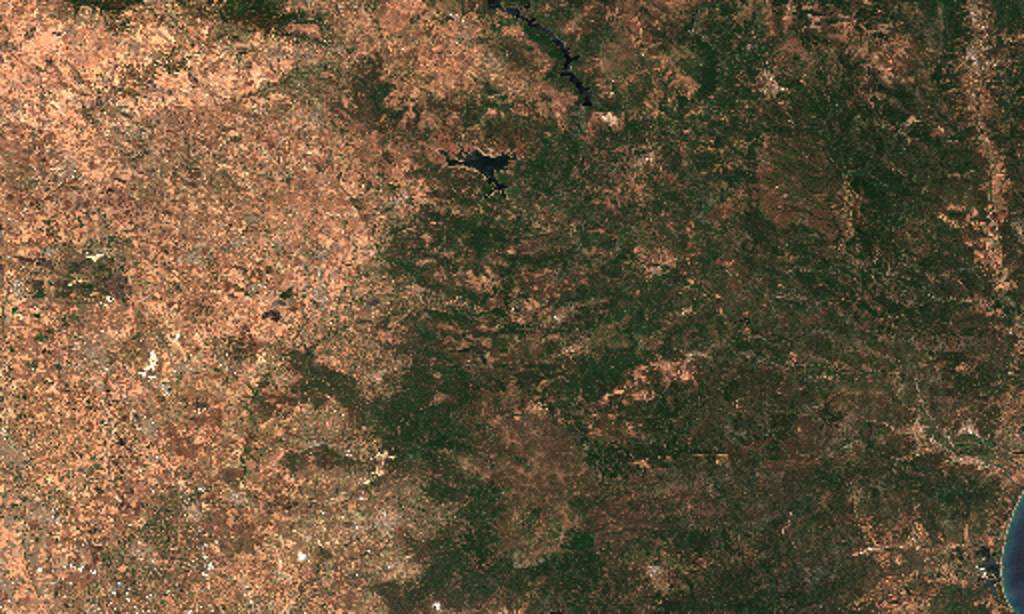} & 
        \includegraphics[width=0.2\linewidth, height=0.12\linewidth]{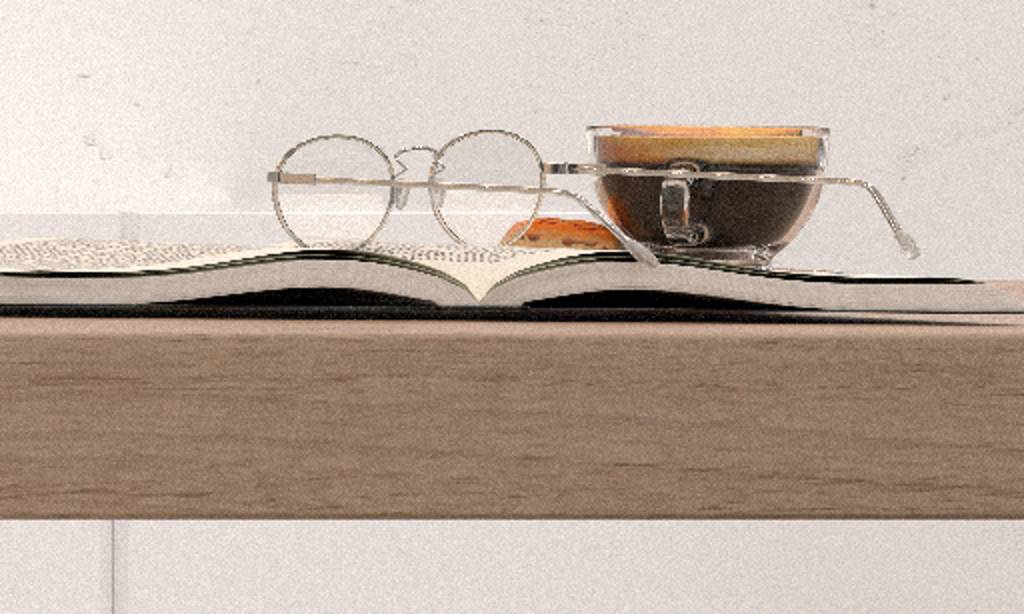} & 
        \includegraphics[width=0.2\linewidth, height=0.12\linewidth]{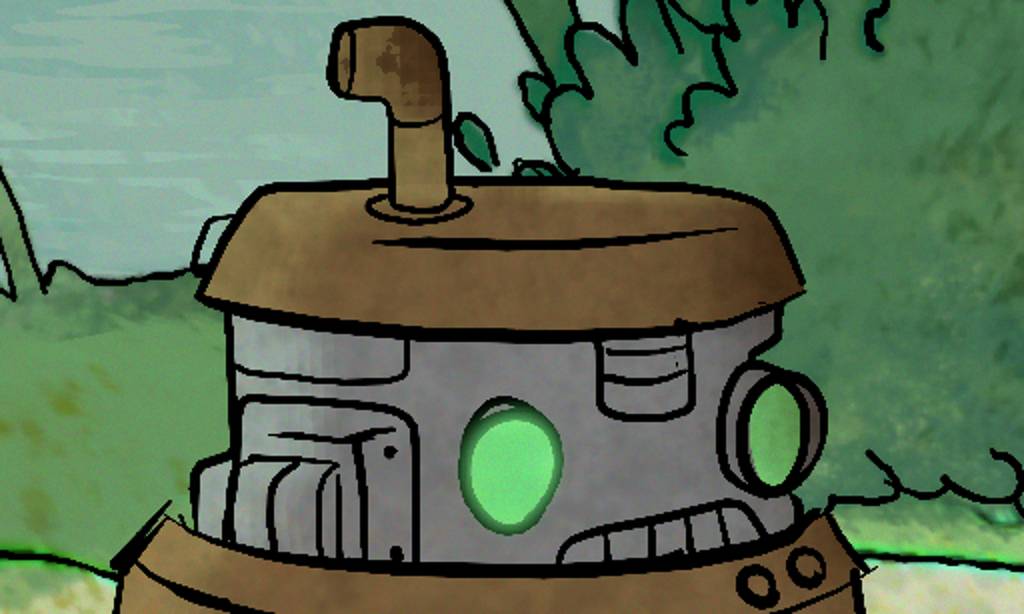} \\

    \end{tabular}}
    \caption{
    \textbf{Additional super resolution $\times4$ qualitatives results on \dataname{}.}
    We show $1228 \times 2048$ crops from the full resolution images.
    }

    \label{fig:suppl_sr_x4_qualitatives_2}

\end{figure*}

\begin{figure*}[ht]
    \centering
    \resizebox{\linewidth}{!}{
    \setlength{\tabcolsep}{1px}
    \begin{tabular}{l ccccc}
        & \textit{Venice} & \textit{Statuario}& \textit{Calacatta} & \textit{Breast Cancer} & \textit{Kidney} \\
        \rotatebox{90}{ \small \hspace{2em} \textbf{GT}} &
        \includegraphics[width=0.2\linewidth, height=0.12\linewidth]{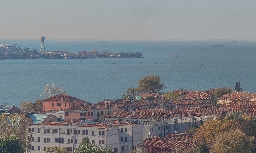} & 
        \includegraphics[width=0.2\linewidth, height=0.12\linewidth]{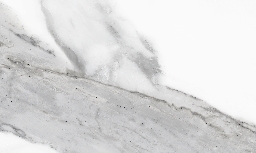} & 
        \includegraphics[width=0.2\linewidth, height=0.12\linewidth]{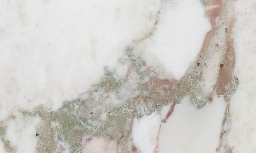} & 
        \includegraphics[width=0.2\linewidth, height=0.12\linewidth]{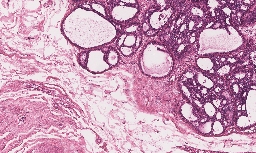} & 
        \includegraphics[width=0.2\linewidth, height=0.12\linewidth]{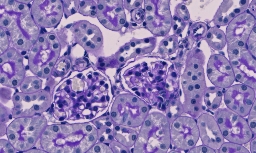} \\

        \rotatebox{90}{\small \hspace{1.5em} \textbf{SIREN}} &
        \includegraphics[width=0.2\linewidth, height=0.12\linewidth]{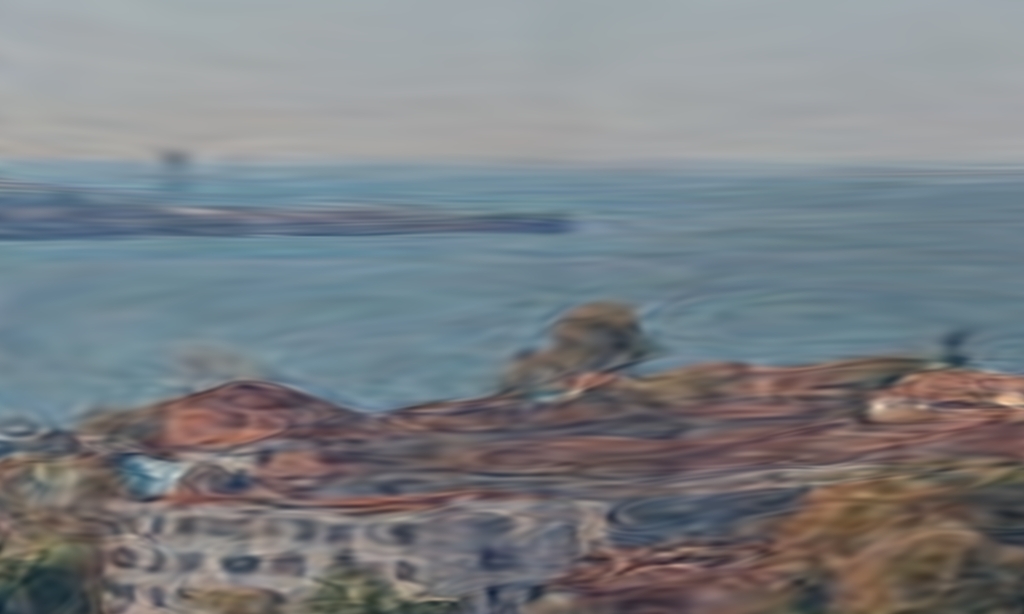} & 
        \includegraphics[width=0.2\linewidth, height=0.12\linewidth]{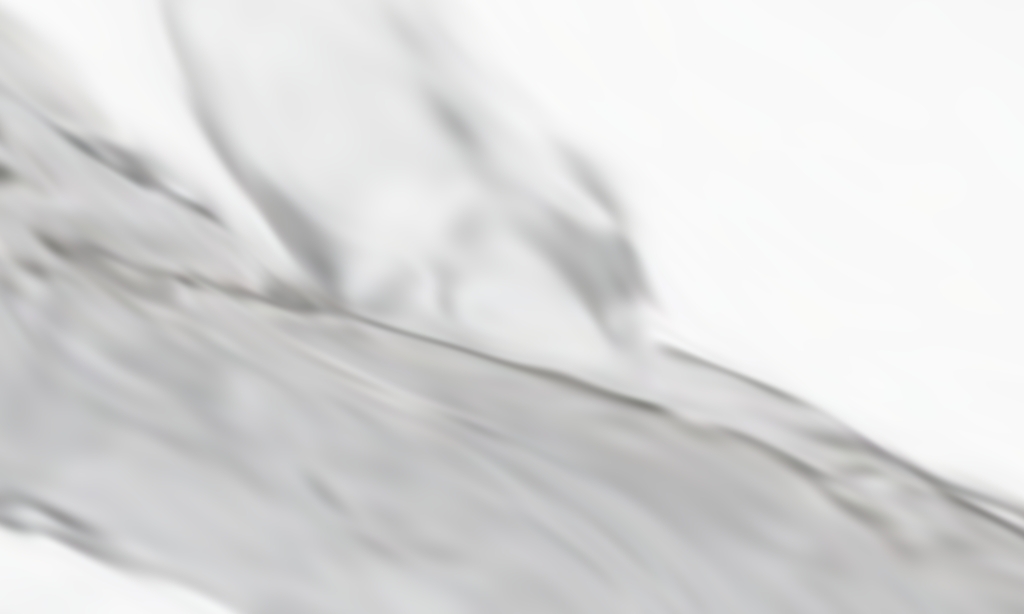} & 
        \includegraphics[width=0.2\linewidth, height=0.12\linewidth]{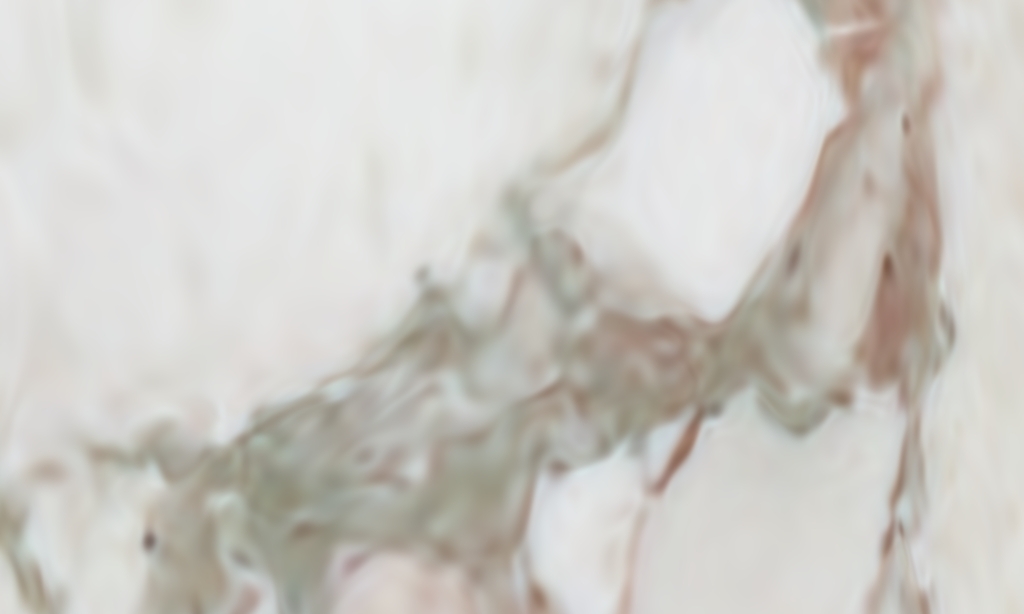} & 
        \includegraphics[width=0.2\linewidth, height=0.12\linewidth]{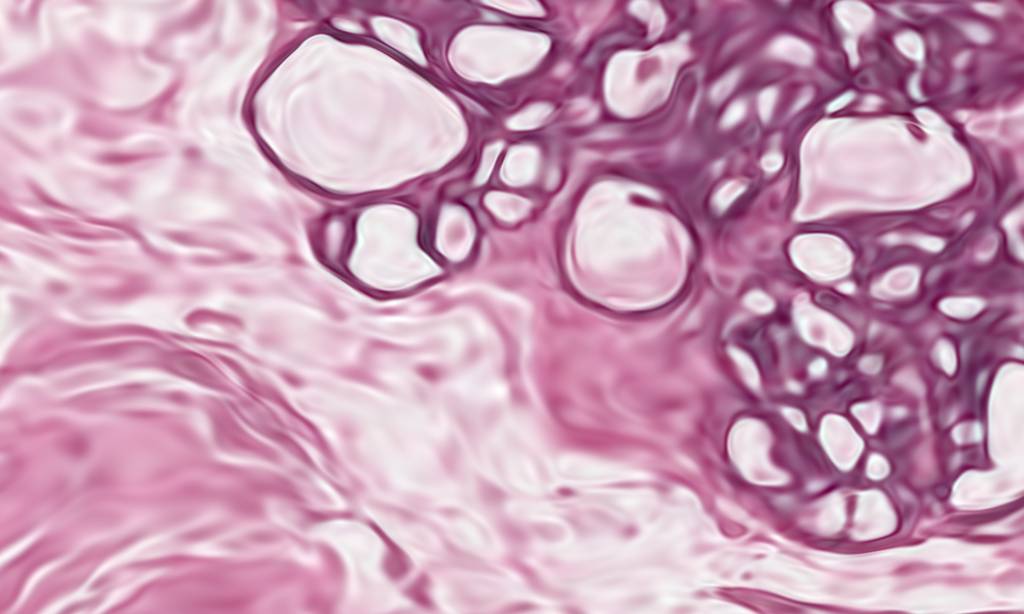} & 
        \includegraphics[width=0.2\linewidth, height=0.12\linewidth]{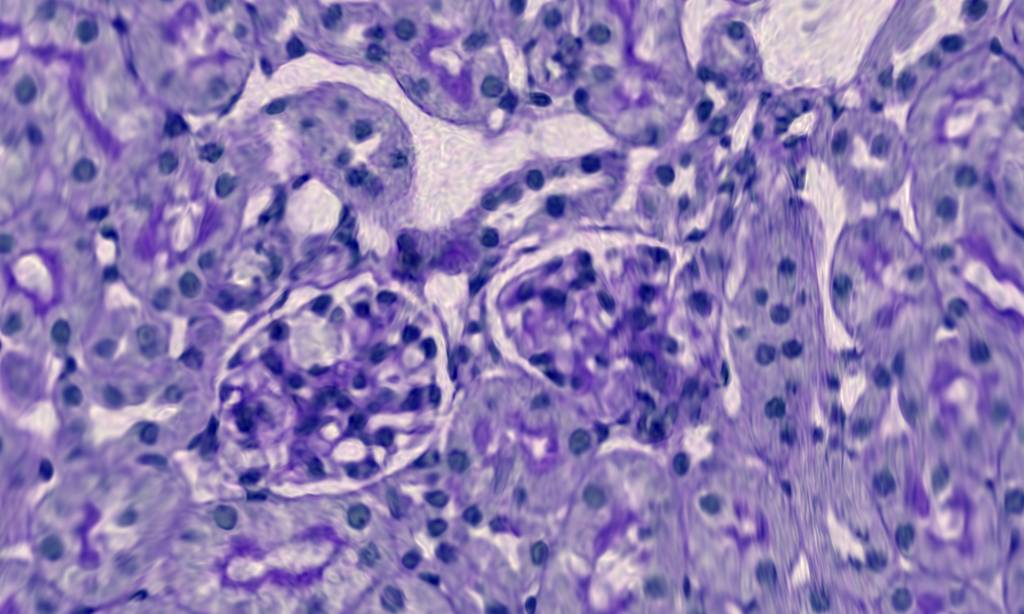} \\

        \rotatebox{90}{ \small \hspace{1em} \textbf{FDHO}} &
        \includegraphics[width=0.2\linewidth, height=0.12\linewidth]{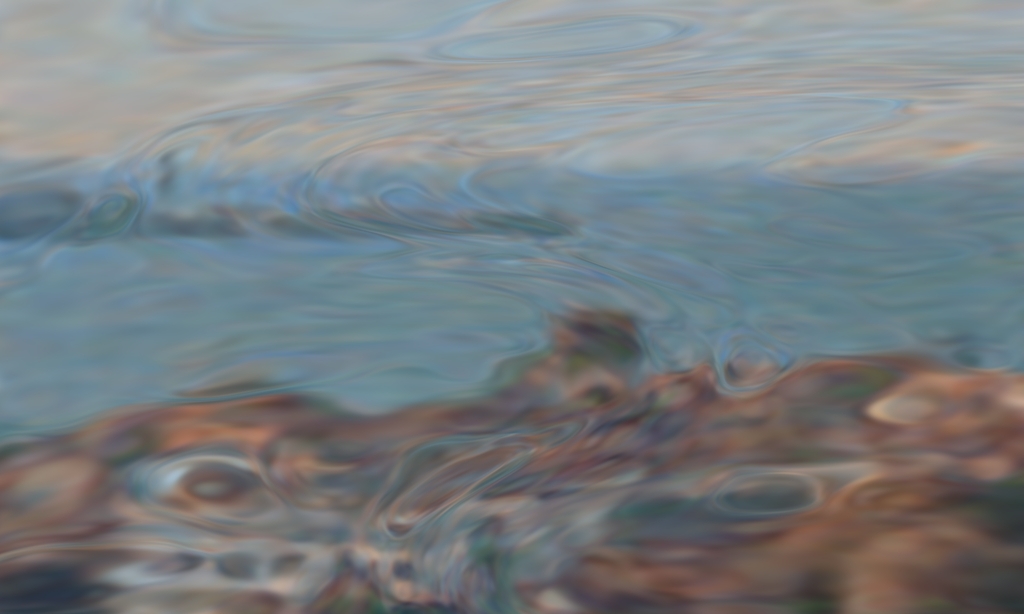} & 
        \includegraphics[width=0.2\linewidth, height=0.12\linewidth]{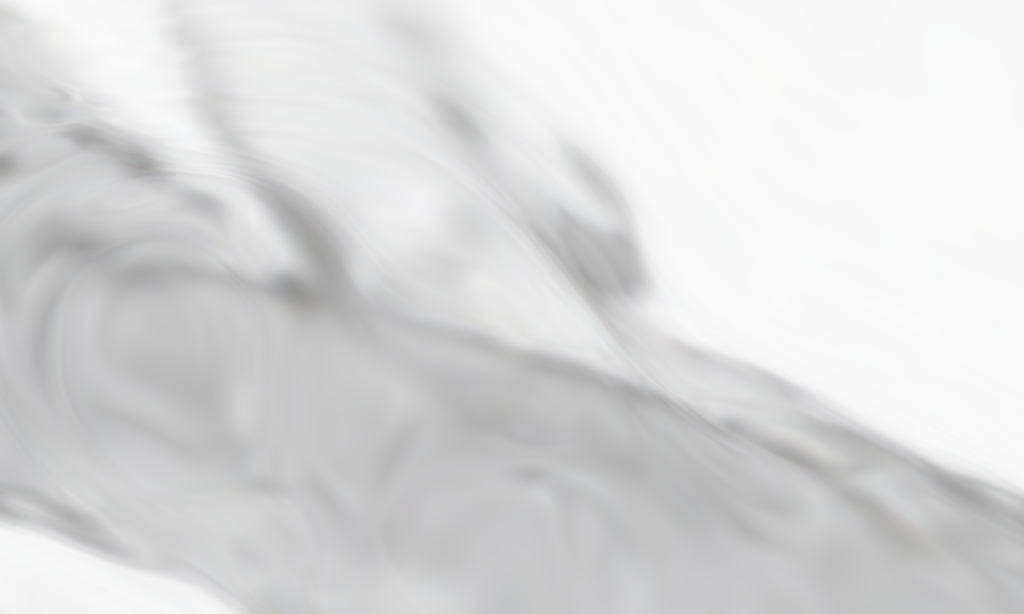} & 
        \includegraphics[width=0.2\linewidth, height=0.12\linewidth]{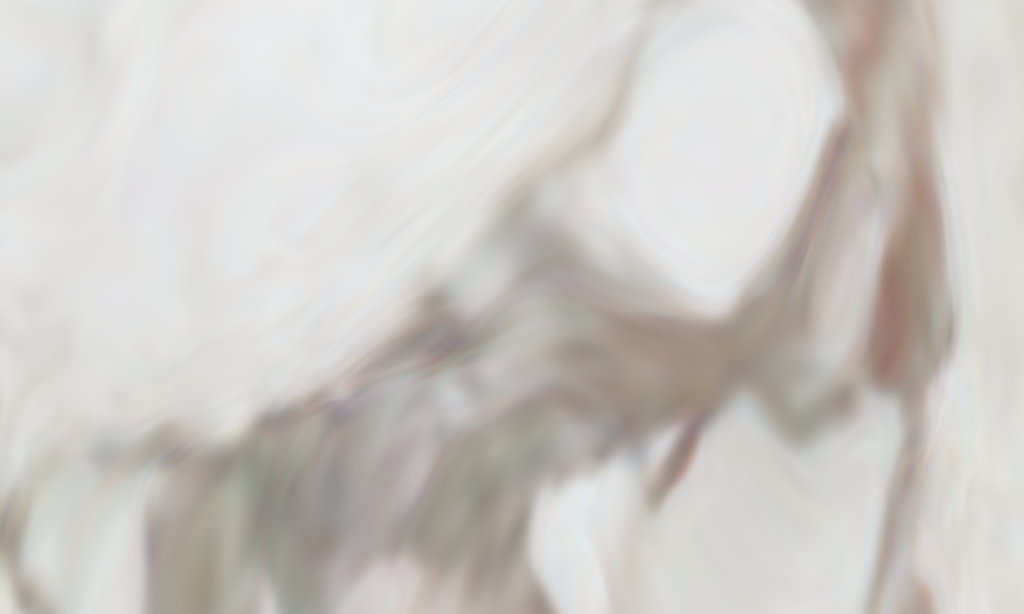} & 
        \includegraphics[width=0.2\linewidth, height=0.12\linewidth]{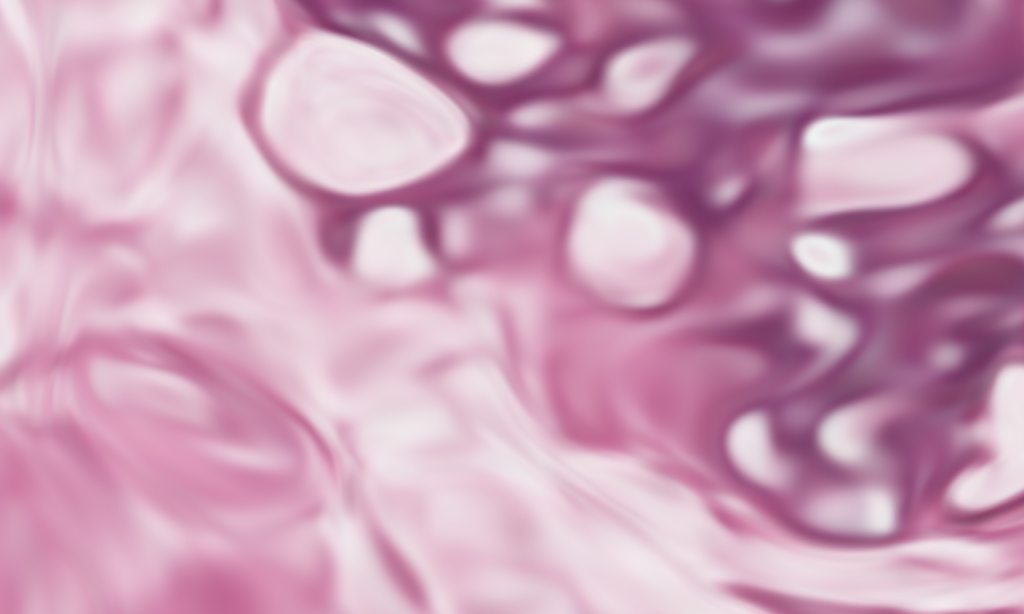} & 
        \includegraphics[width=0.2\linewidth, height=0.12\linewidth]{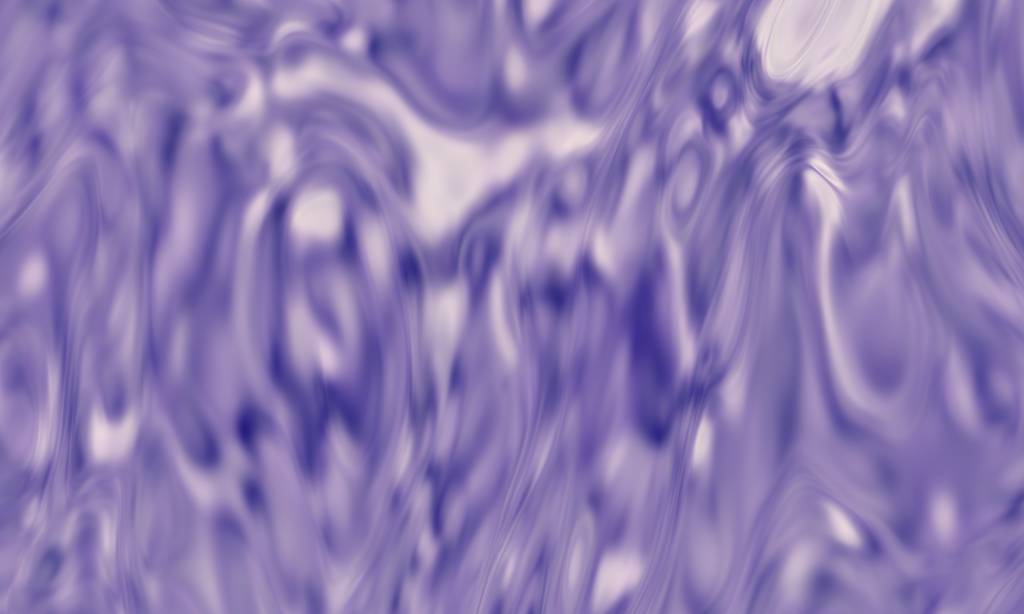} \\

        \rotatebox{90}{ \small \hspace{1em} \textbf{FINER}} &
        \includegraphics[width=0.2\linewidth, height=0.12\linewidth]{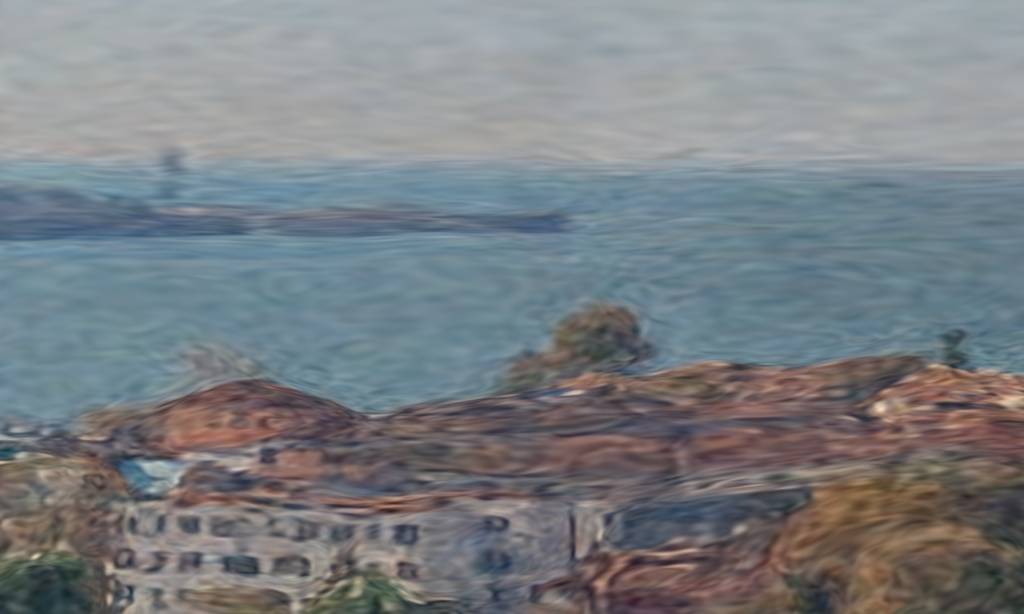} & 
        \includegraphics[width=0.2\linewidth, height=0.12\linewidth]{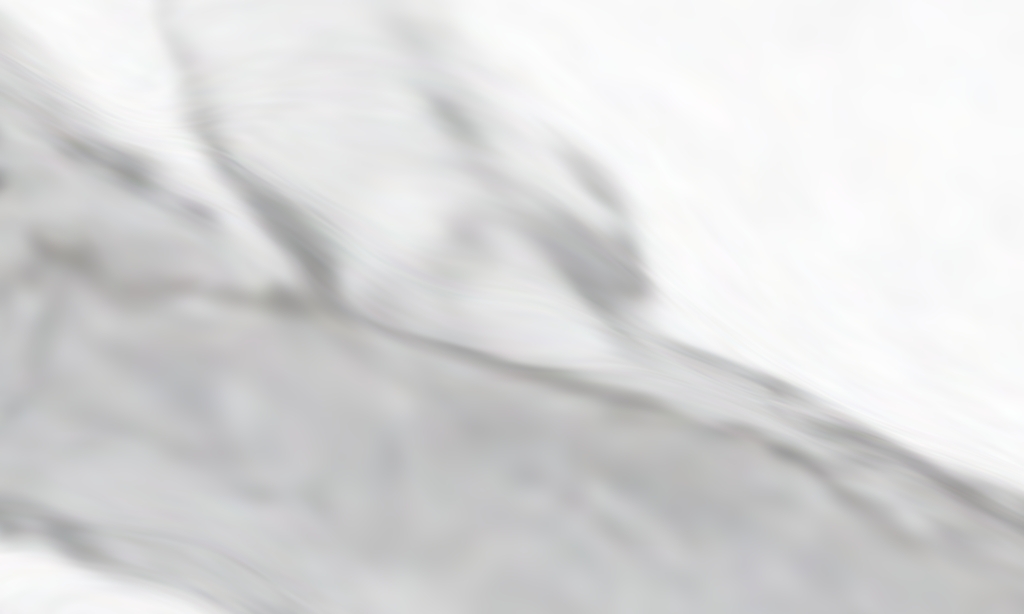} & 
        \includegraphics[width=0.2\linewidth, height=0.12\linewidth]{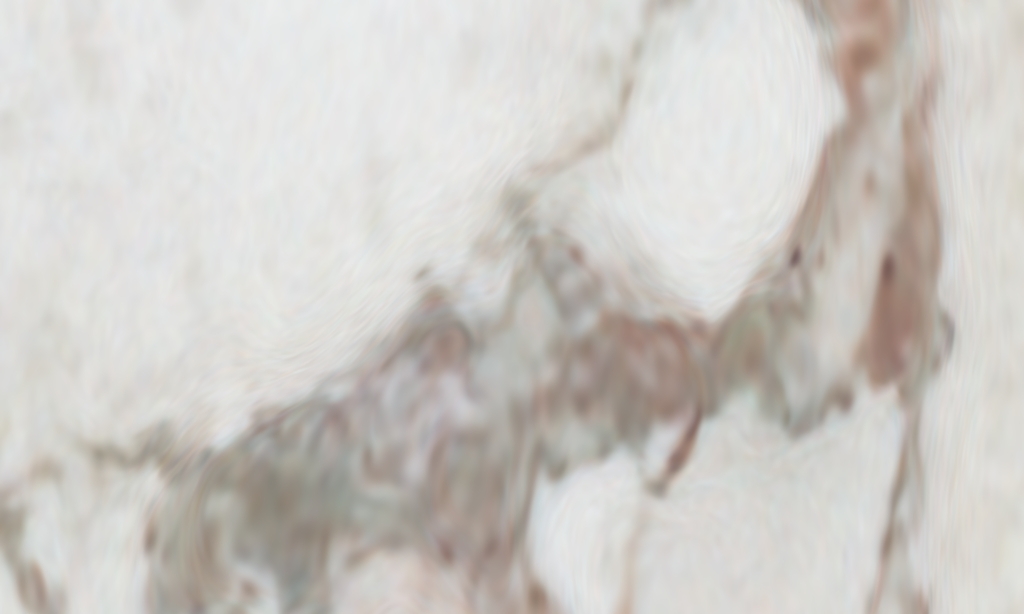} & 
        \includegraphics[width=0.2\linewidth, height=0.12\linewidth]{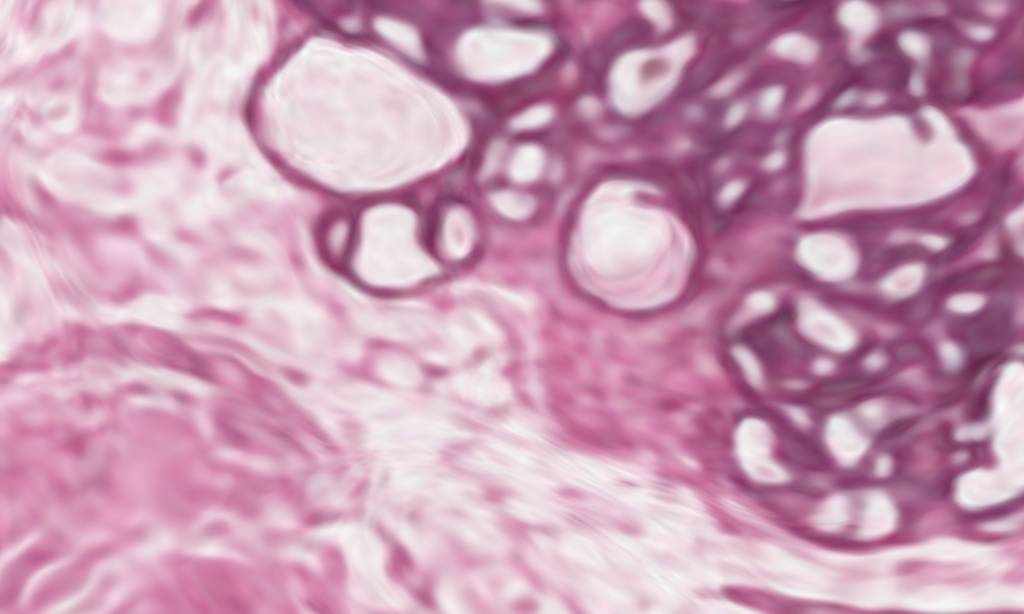} & 
        \includegraphics[width=0.2\linewidth, height=0.12\linewidth]{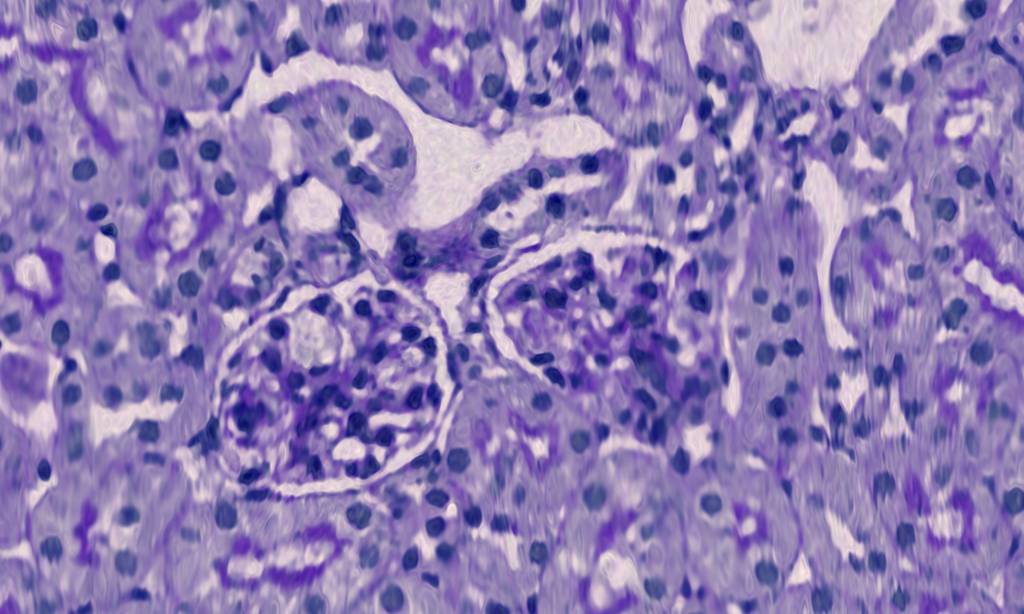} \\

        \rotatebox{90}{ \small \hspace{1.25em} \textbf{I-NGP}} &
        \includegraphics[width=0.2\linewidth, height=0.12\linewidth]{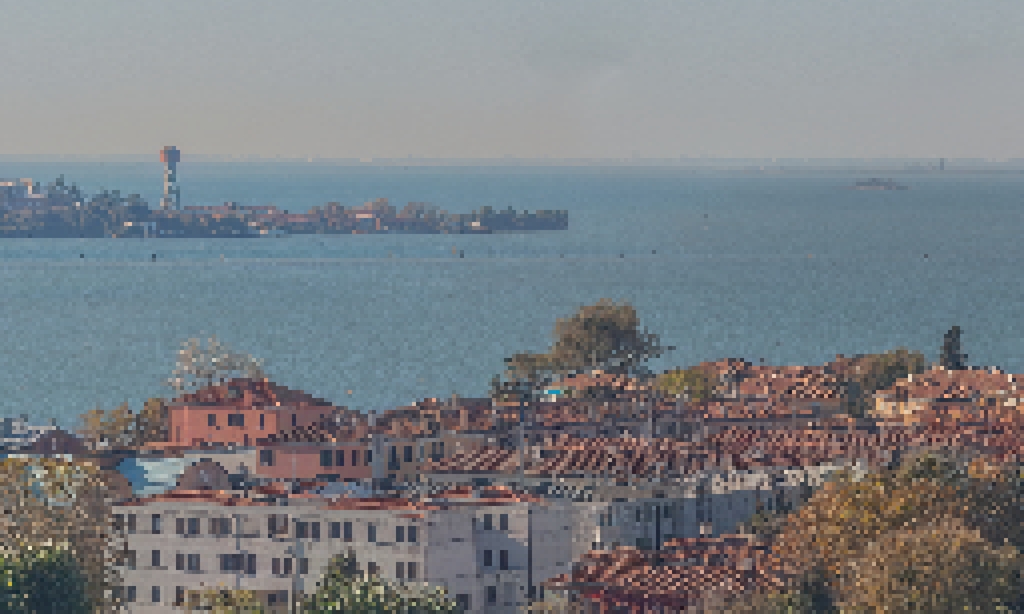} & 
        \includegraphics[width=0.2\linewidth, height=0.12\linewidth]{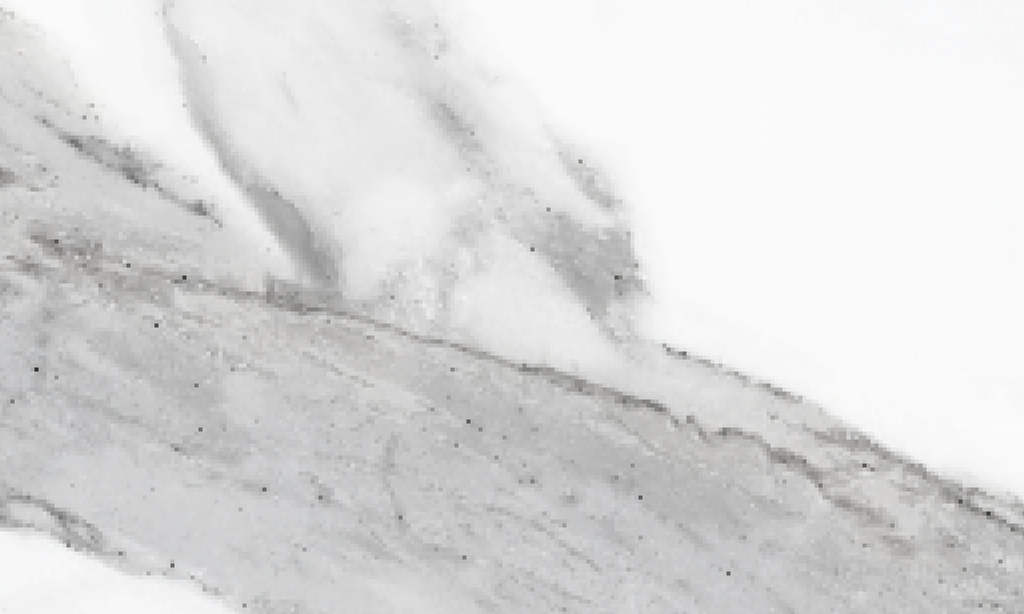} & 
        \includegraphics[width=0.2\linewidth, height=0.12\linewidth]{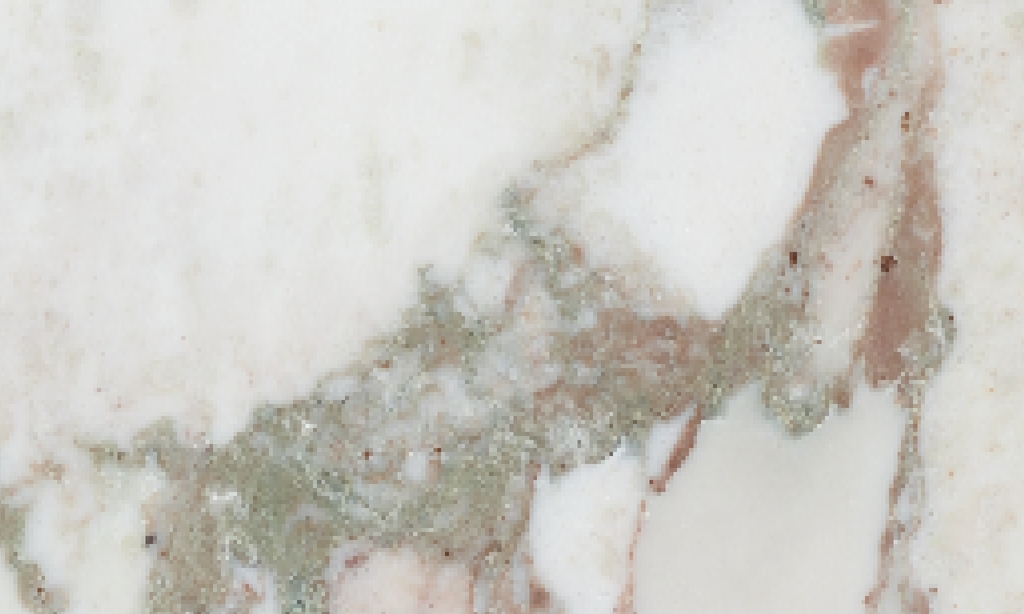} & 
        \includegraphics[width=0.2\linewidth, height=0.12\linewidth]{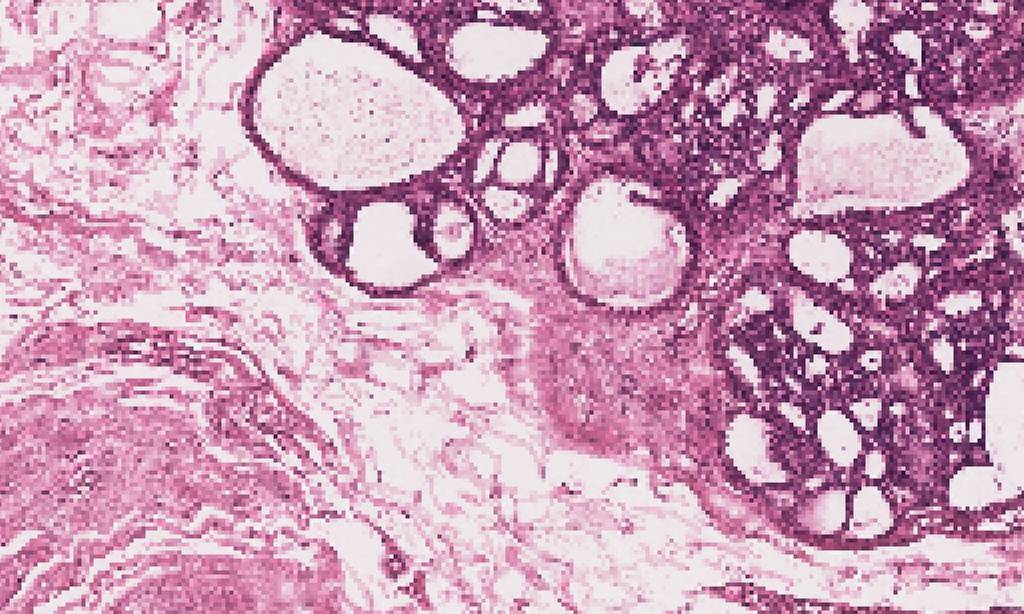} & 
        \includegraphics[width=0.2\linewidth, height=0.12\linewidth]{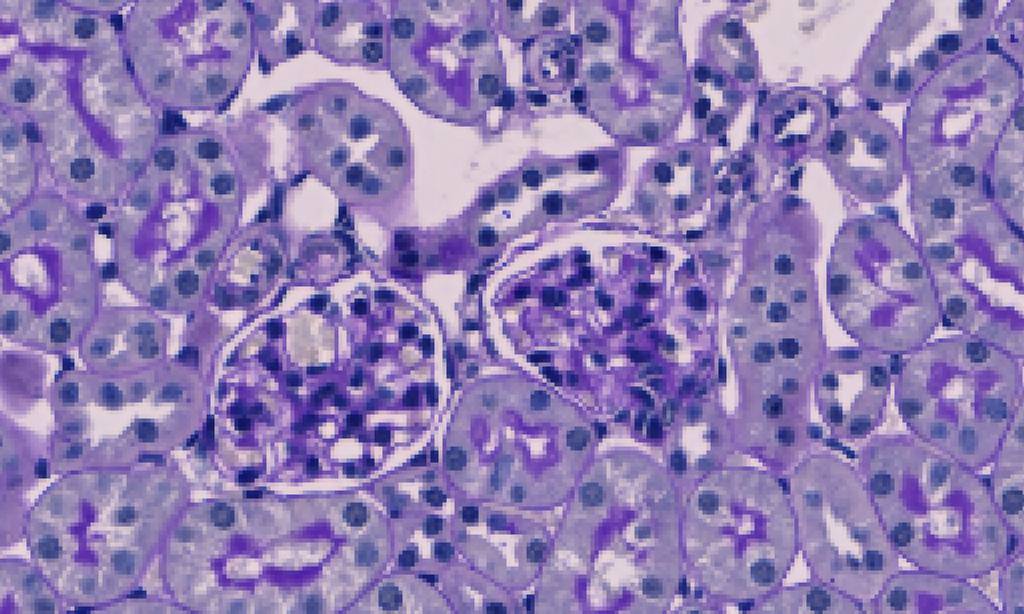} \\

        \rotatebox{90}{ \small \hspace{1em} \textbf{NeuRBF}} &
        \includegraphics[width=0.2\linewidth, height=0.12\linewidth]{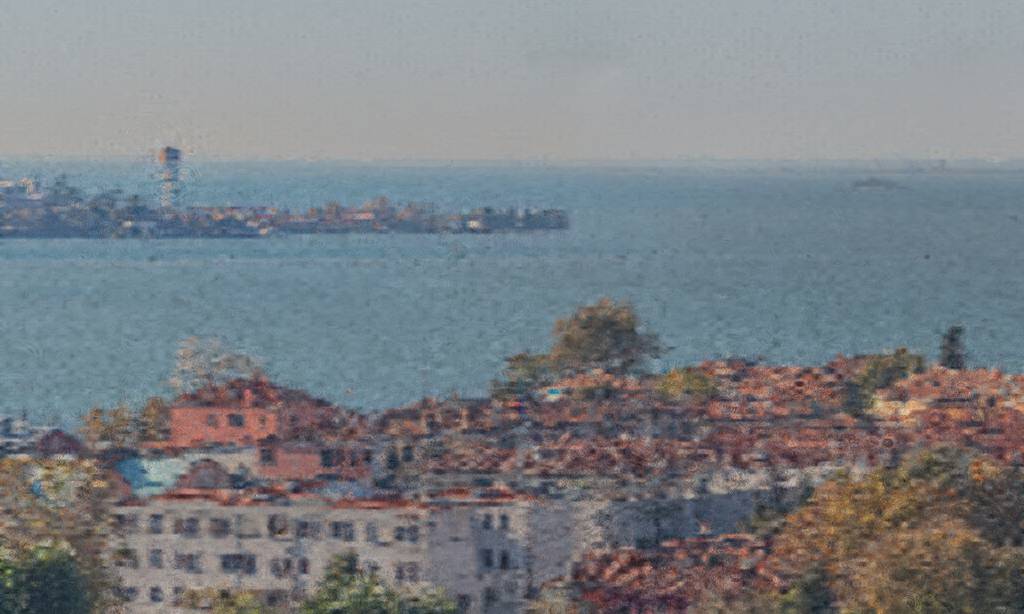} & 
        \includegraphics[width=0.2\linewidth, height=0.12\linewidth]{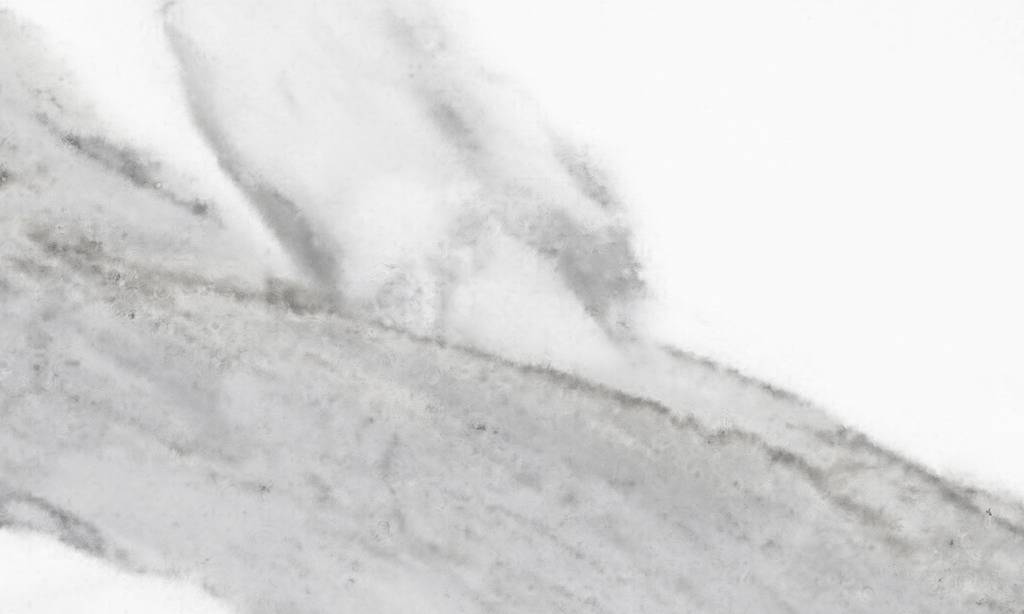} & 
        \includegraphics[width=0.2\linewidth, height=0.12\linewidth]{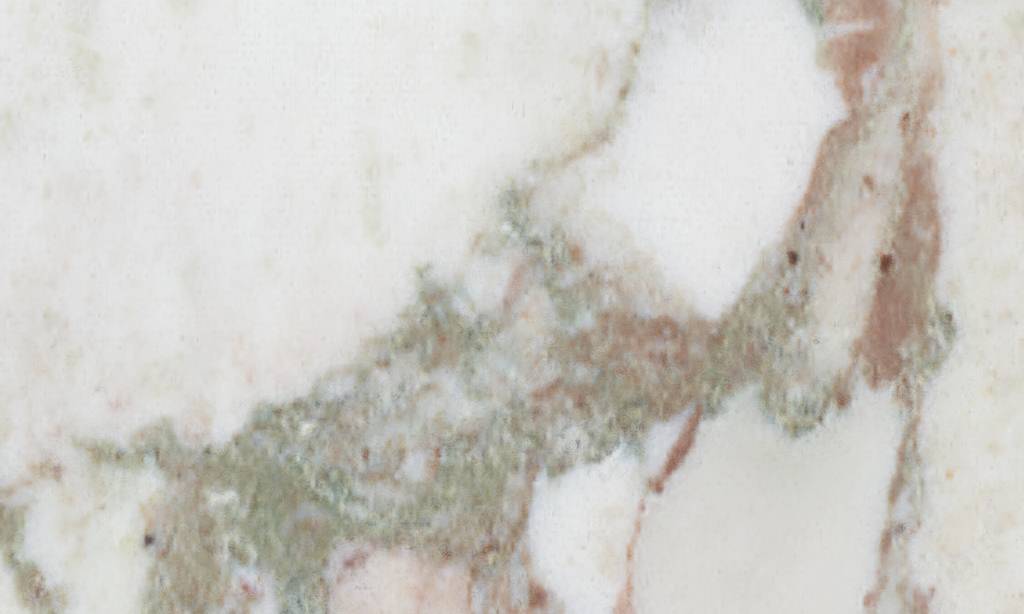} & 
        \includegraphics[width=0.2\linewidth, height=0.12\linewidth]{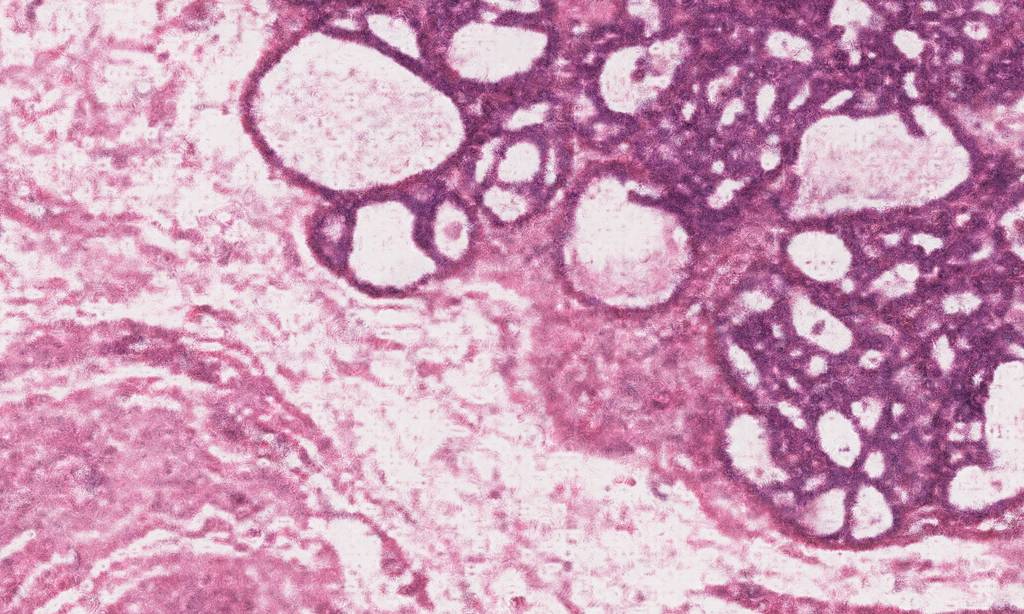} & 
        \includegraphics[width=0.2\linewidth, height=0.12\linewidth]{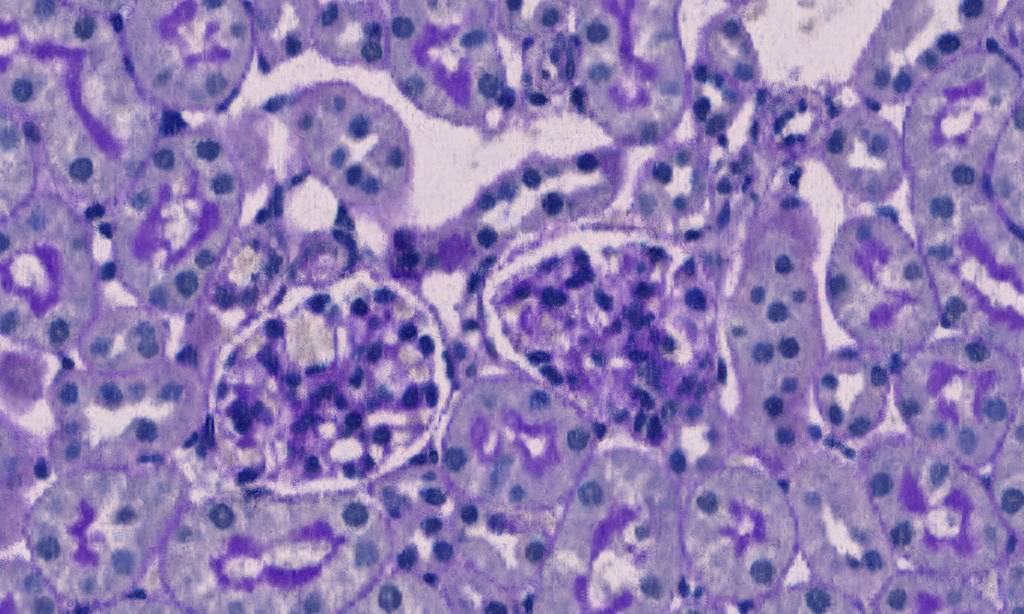} \\

        \rotatebox{90}{ \small \hspace{0.25em} \textbf{MetricGrids}} &
        \includegraphics[width=0.2\linewidth, height=0.12\linewidth]{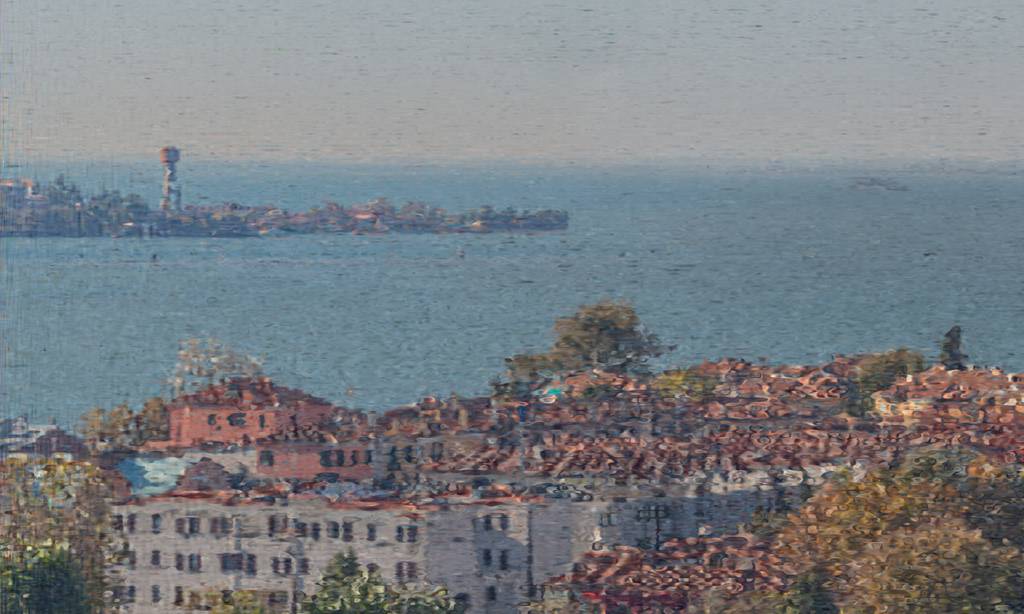} & 
        \includegraphics[width=0.2\linewidth, height=0.12\linewidth]{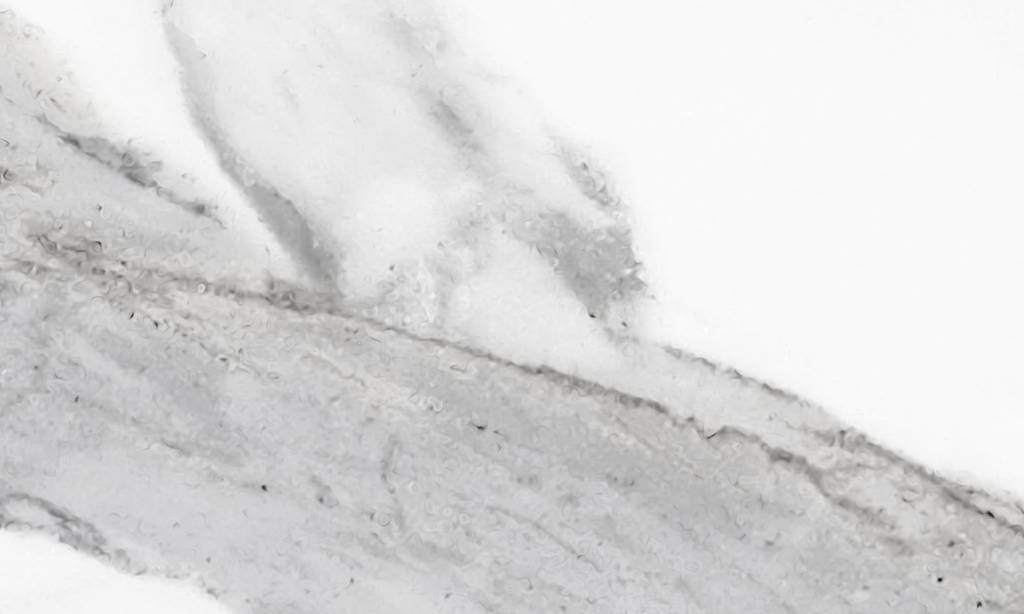} & 
        \includegraphics[width=0.2\linewidth, height=0.12\linewidth]{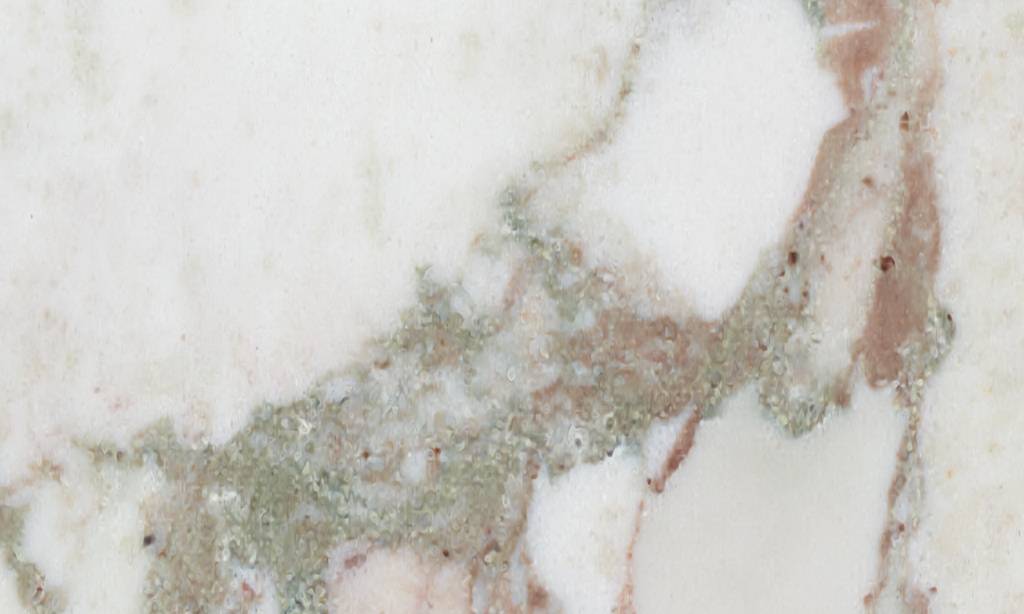} & 
        \includegraphics[width=0.2\linewidth, height=0.12\linewidth]{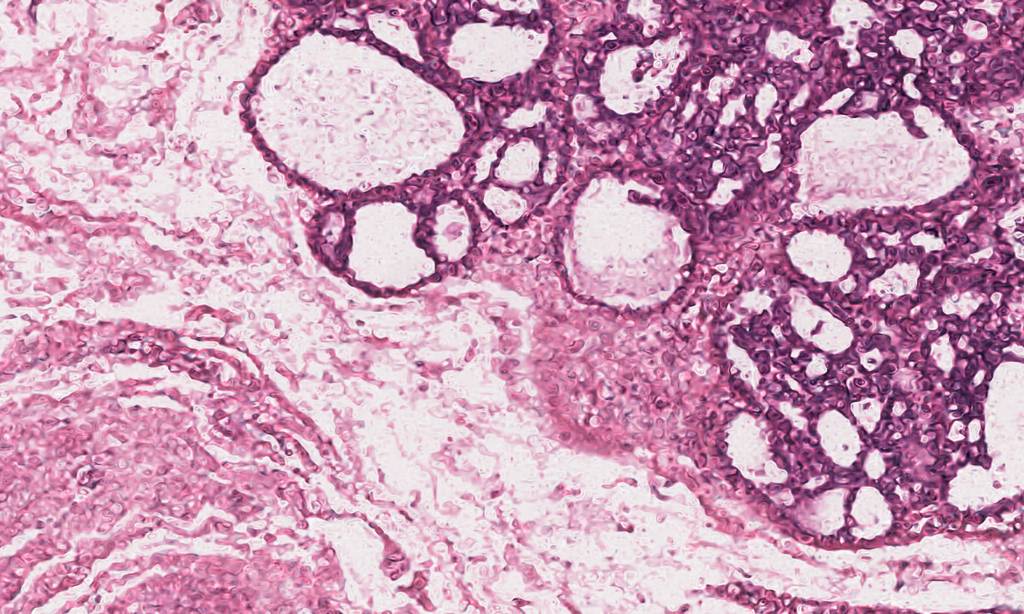} & 
        \includegraphics[width=0.2\linewidth, height=0.12\linewidth]{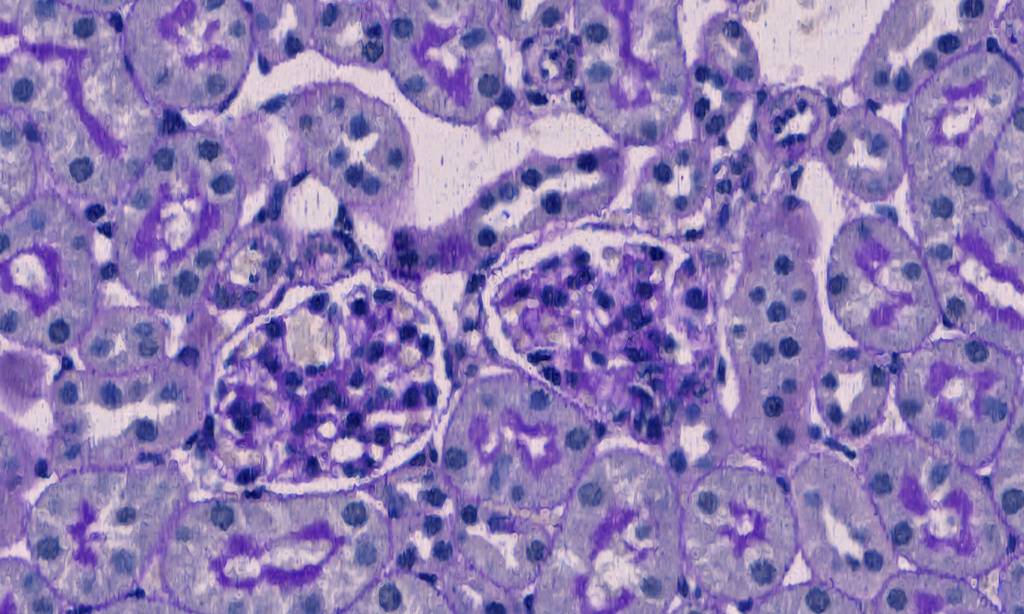} \\

        \rotatebox{90}{ \small \hspace{0.5em} \textbf{GaussImg}} &
        \includegraphics[width=0.2\linewidth, height=0.12\linewidth]{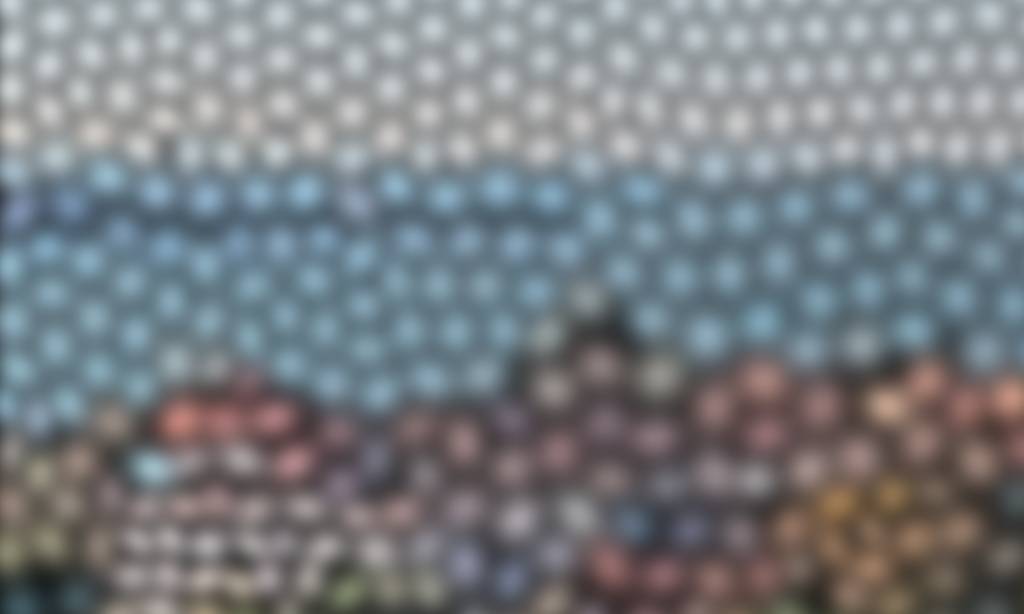} & 
        \includegraphics[width=0.2\linewidth, height=0.12\linewidth]{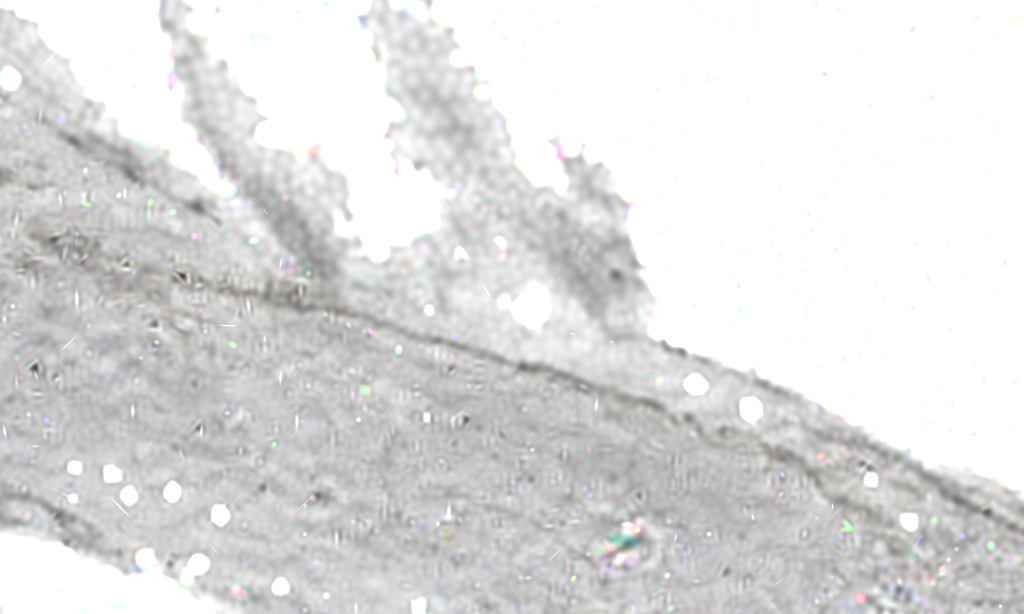} & 
        \includegraphics[width=0.2\linewidth, height=0.12\linewidth]{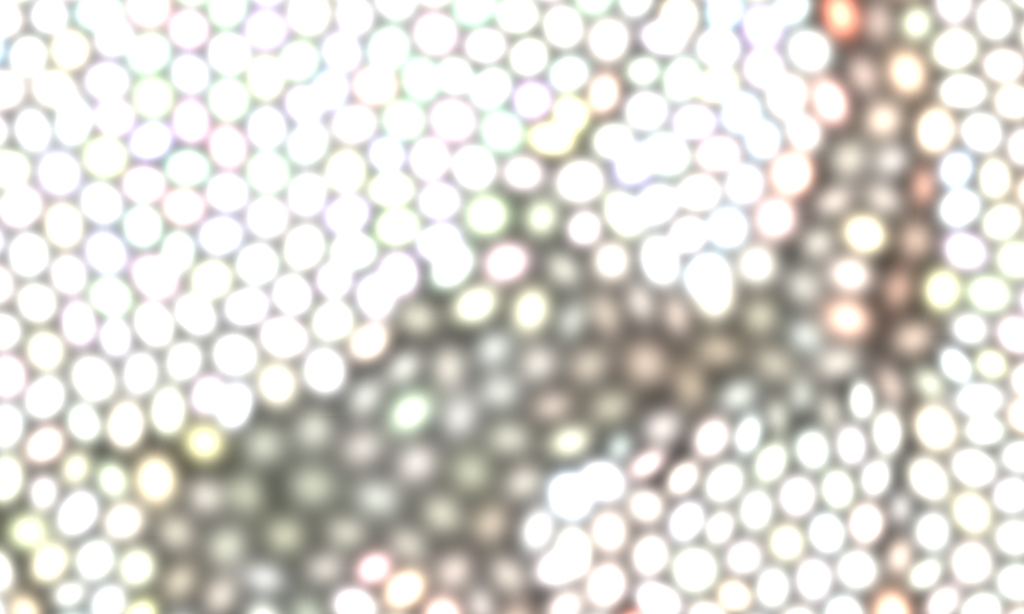} & 
        \includegraphics[width=0.2\linewidth, height=0.12\linewidth]{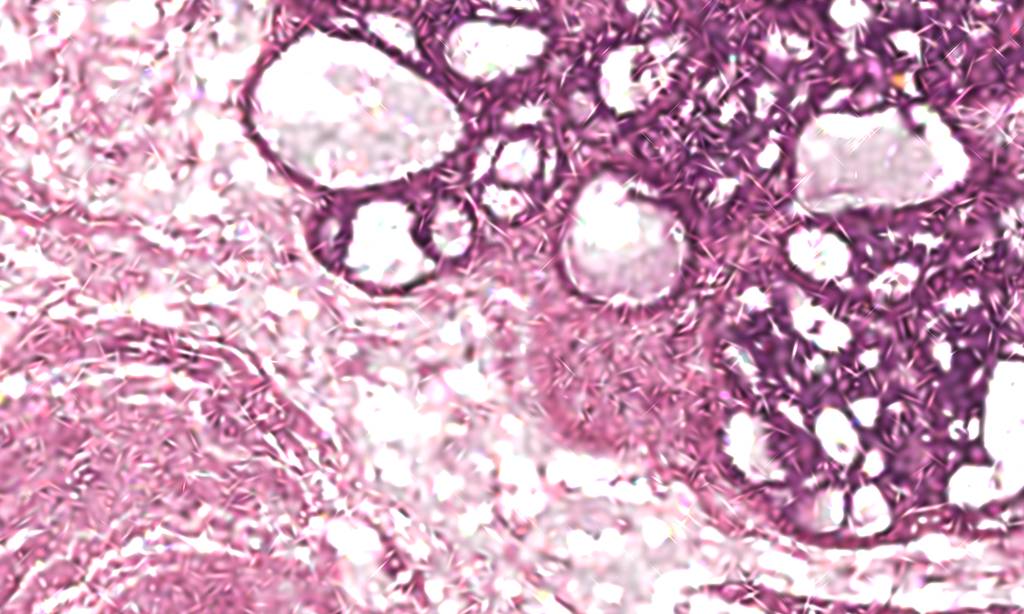} & 
        \includegraphics[width=0.2\linewidth, height=0.12\linewidth]{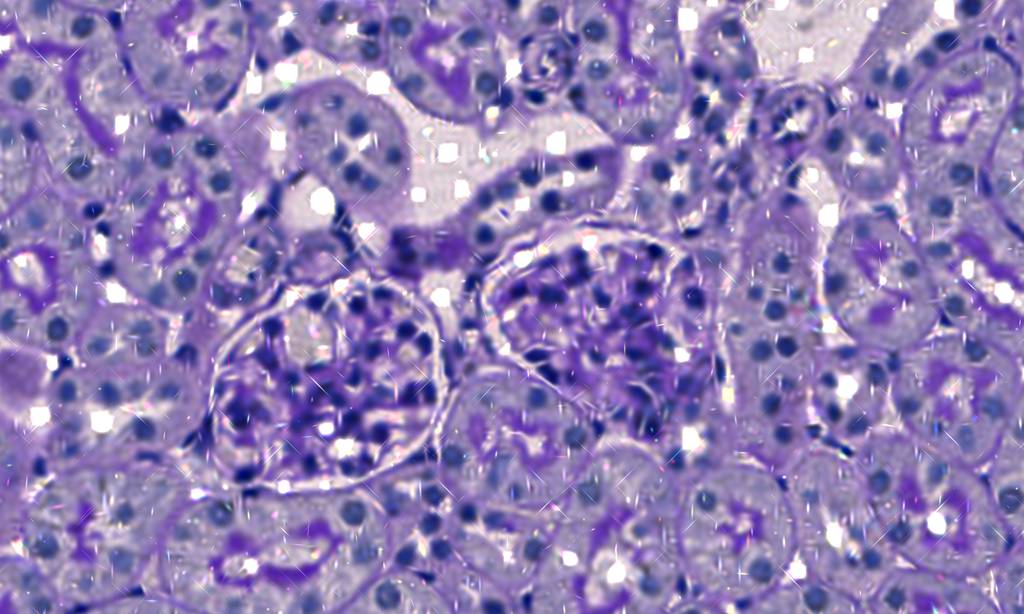} \\

        \rotatebox{90}{ \small \hspace{0.25em} \textbf{SmartSplat}} &
        \includegraphics[width=0.2\linewidth, height=0.12\linewidth]{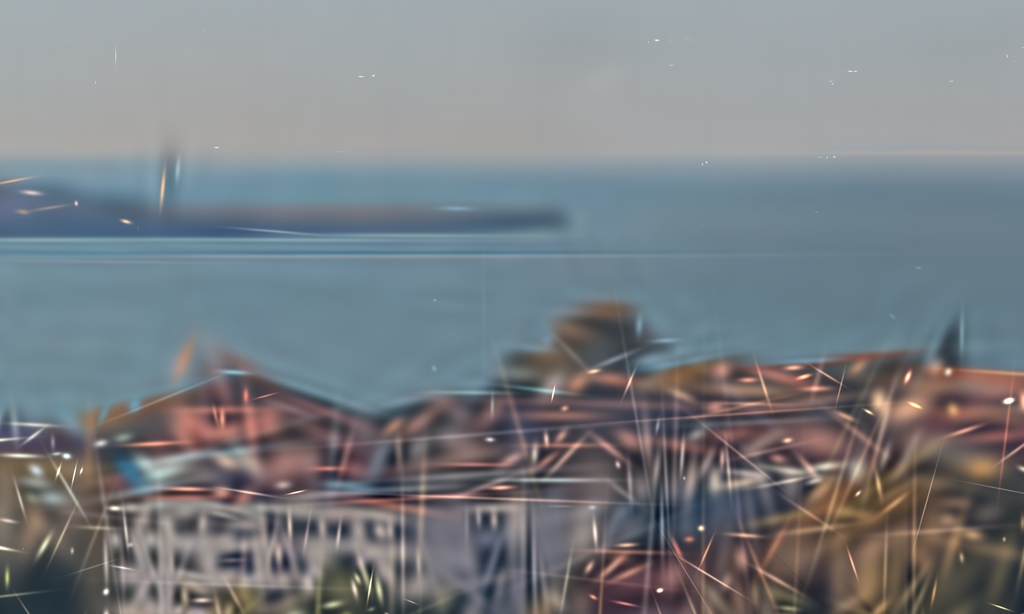} & 
        \includegraphics[width=0.2\linewidth, height=0.12\linewidth]{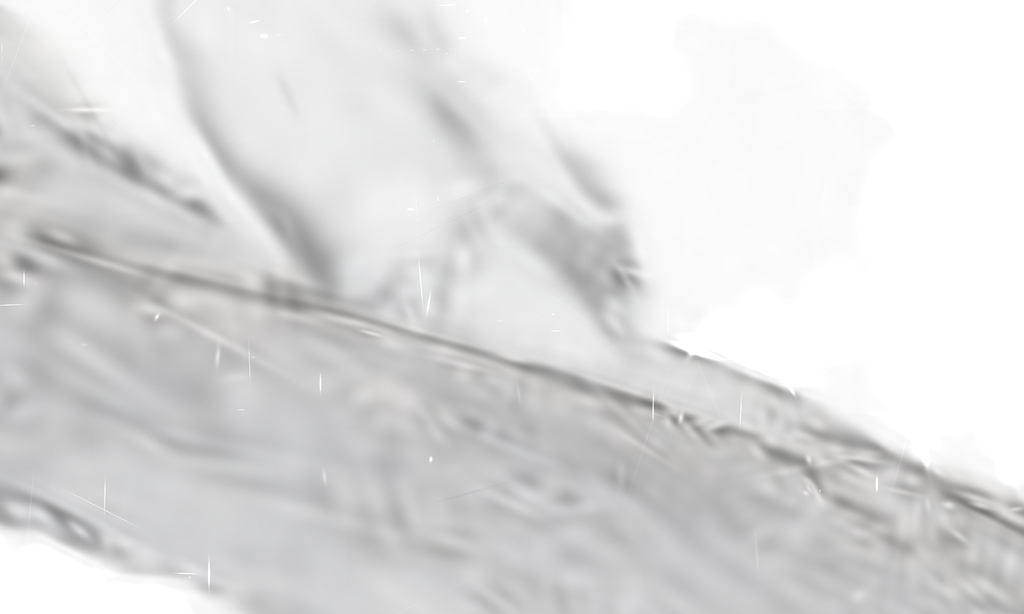} & 
        \includegraphics[width=0.2\linewidth, height=0.12\linewidth]{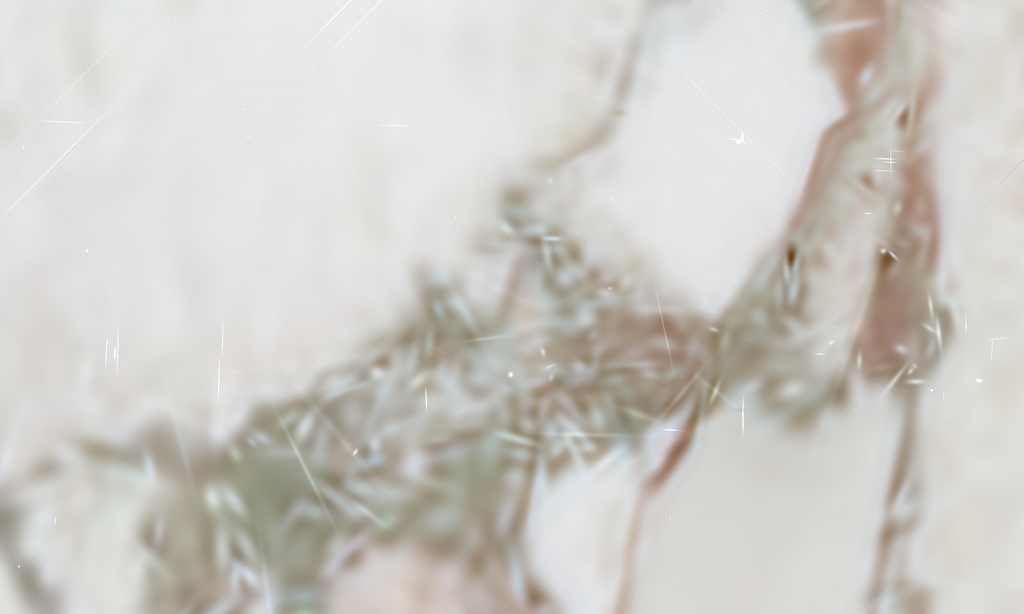} & 
        \includegraphics[width=0.2\linewidth, height=0.12\linewidth]{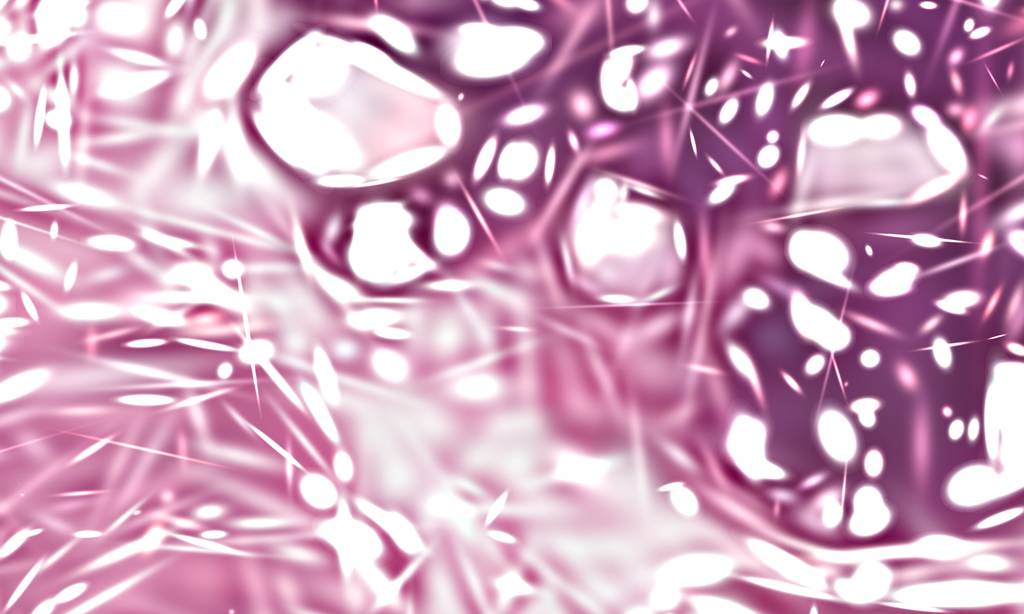} & 
        \includegraphics[width=0.2\linewidth, height=0.12\linewidth]{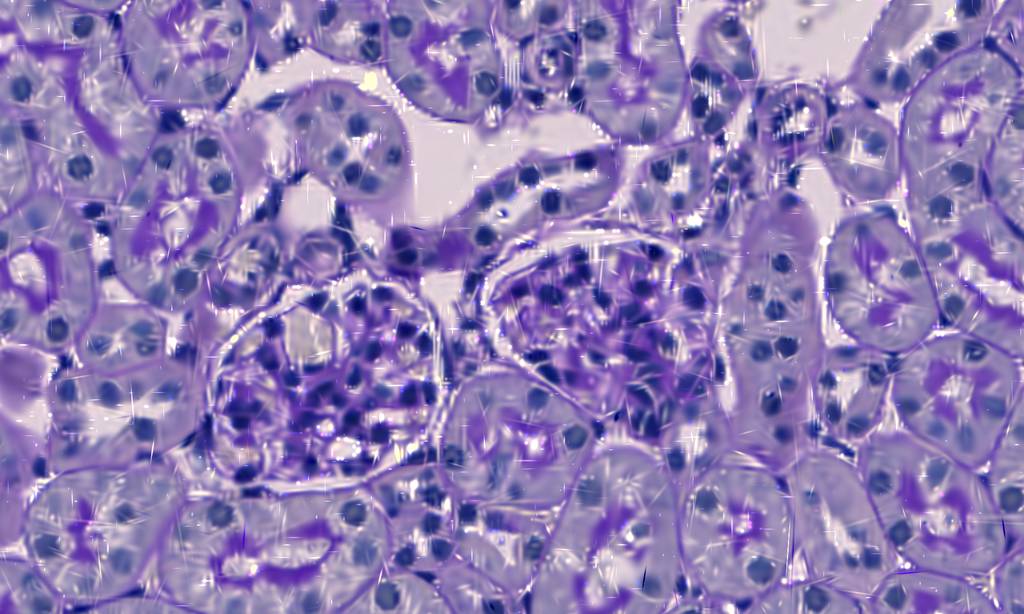} \\

        \rotatebox{90}{ \small \hspace{1em} \textbf{Bicubic}} &
        \includegraphics[width=0.2\linewidth, height=0.12\linewidth]{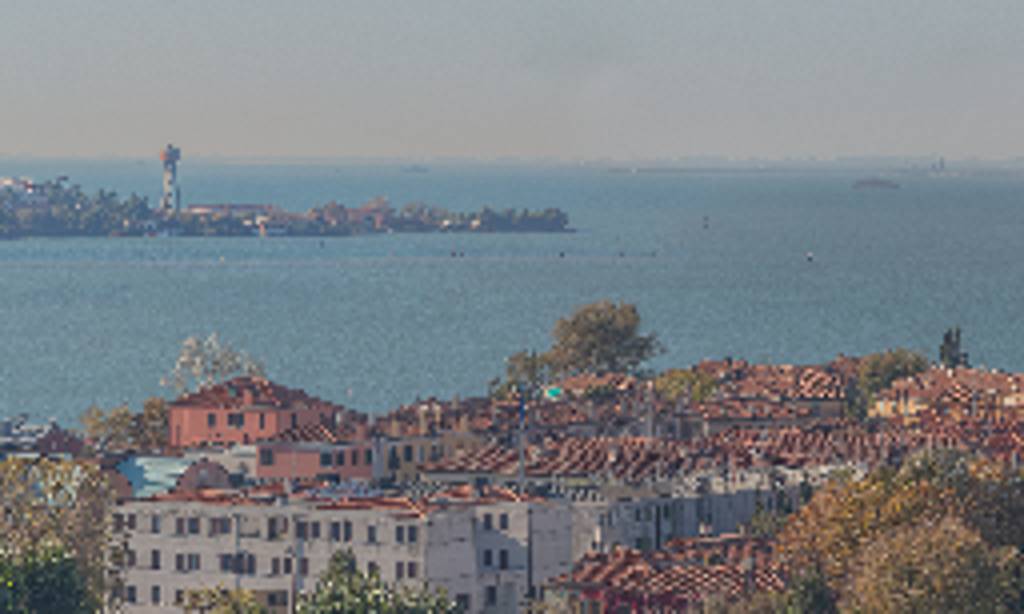} & 
        \includegraphics[width=0.2\linewidth, height=0.12\linewidth]{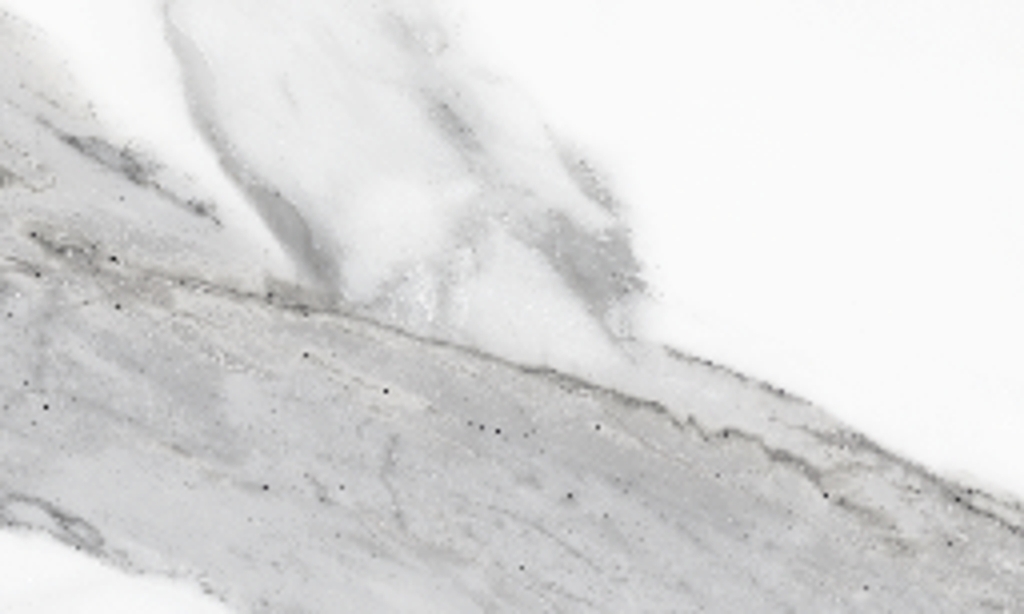} & 
        \includegraphics[width=0.2\linewidth, height=0.12\linewidth]{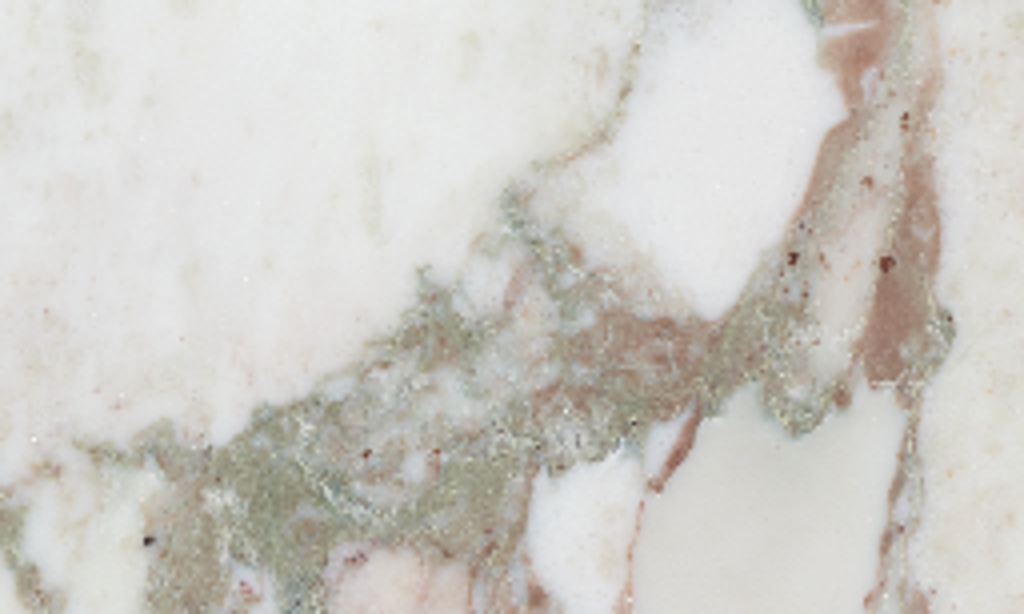} & 
        \includegraphics[width=0.2\linewidth, height=0.12\linewidth]{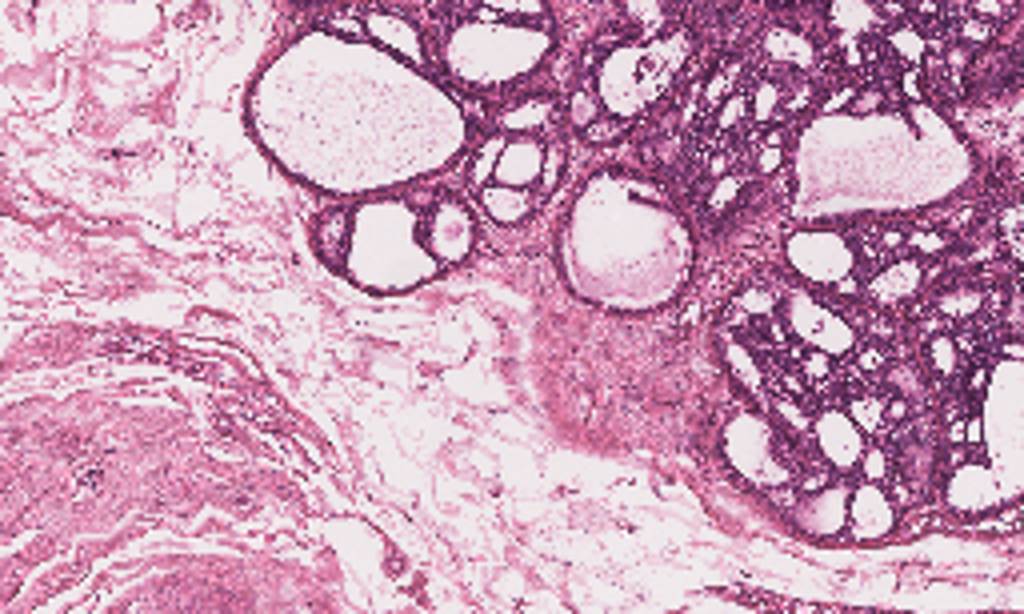} & 
        \includegraphics[width=0.2\linewidth, height=0.12\linewidth]{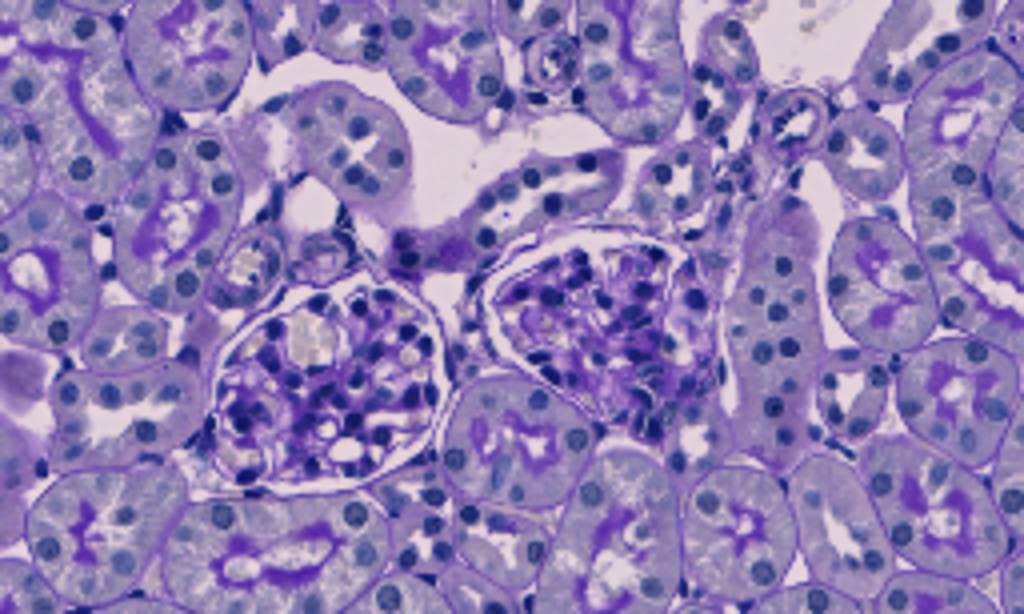} \\
    \end{tabular}}
    \caption{
    \textbf{Additional super resolution $\times8$ qualitatives results on \dataname{}.}
    We show $1228 \times 2048$ crops from the full resolution images.
    }
    \label{fig:suppl_sr_x8_qualitatives_1}

\end{figure*}

\begin{figure*}[ht]
    \centering
    \resizebox{\linewidth}{!}{
    \setlength{\tabcolsep}{1px}
    \begin{tabular}{l ccccc}
        & \textit{Munich} & \textit{Andromeda}& \textit{Italy} & \textit{Lynxsdesign} & \textit{House Of Chores} \\
        \rotatebox{90}{ \small \hspace{2em} \textbf{GT}} &
        \includegraphics[width=0.2\linewidth, height=0.12\linewidth]{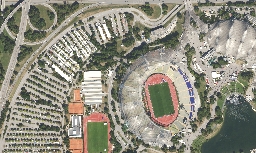} & 
        \includegraphics[width=0.2\linewidth, height=0.12\linewidth]{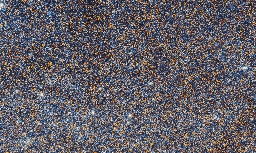} & 
        \includegraphics[width=0.2\linewidth, height=0.12\linewidth]{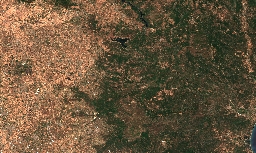} & 
        \includegraphics[width=0.2\linewidth, height=0.12\linewidth]{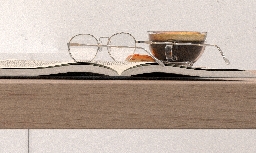} & 
        \includegraphics[width=0.2\linewidth, height=0.12\linewidth]{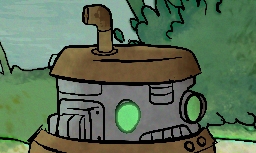} \\

        \rotatebox{90}{ \small \hspace{1.5em} \textbf{SIREN}} &
        \includegraphics[width=0.2\linewidth, height=0.12\linewidth]{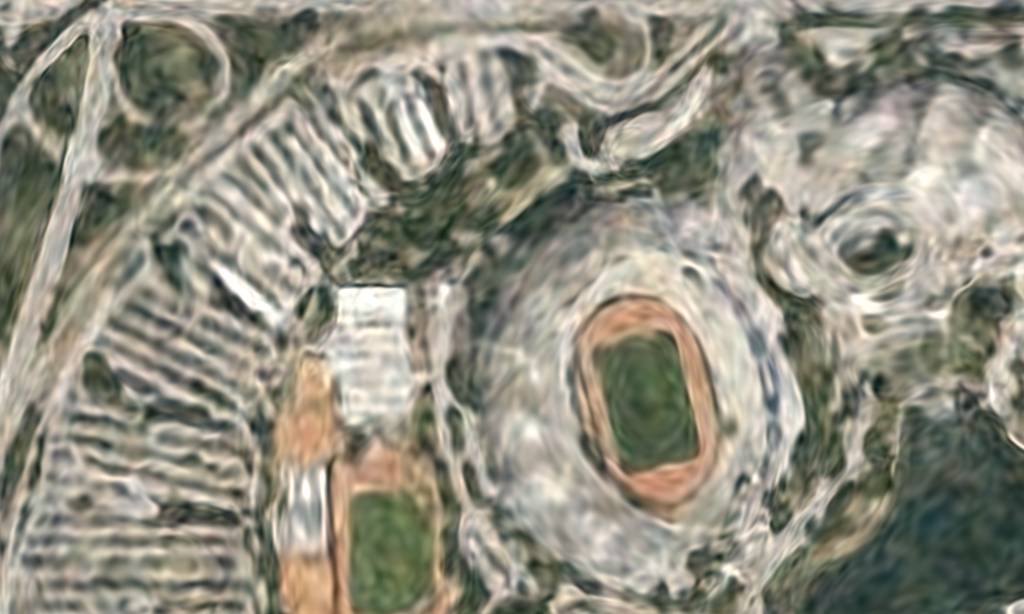} & 
        \includegraphics[width=0.2\linewidth, height=0.12\linewidth]{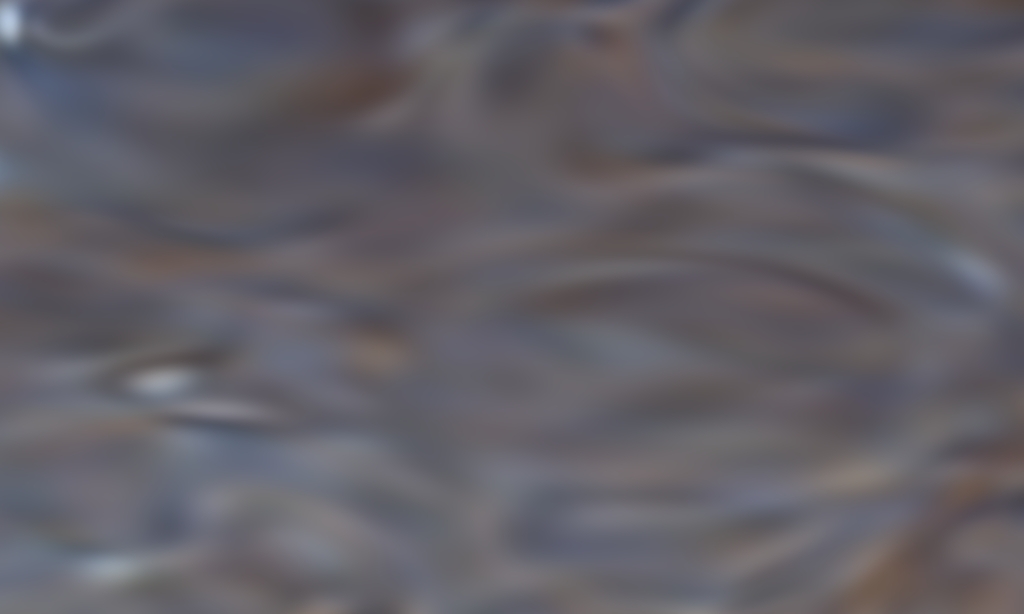} & 
        \includegraphics[width=0.2\linewidth, height=0.12\linewidth]{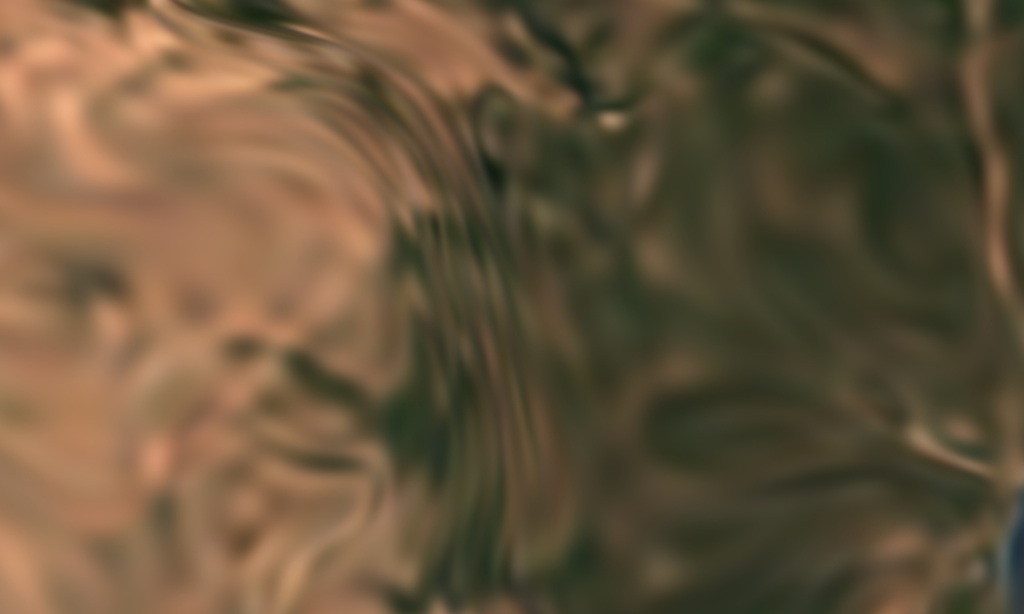} & 
        \includegraphics[width=0.2\linewidth, height=0.12\linewidth]{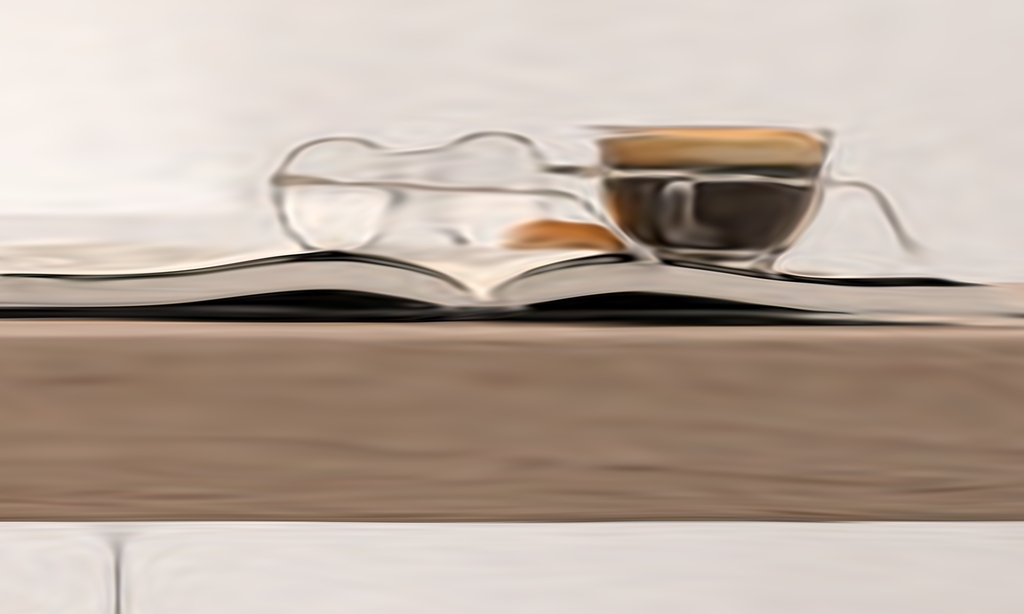} & 
        \includegraphics[width=0.2\linewidth, height=0.12\linewidth]{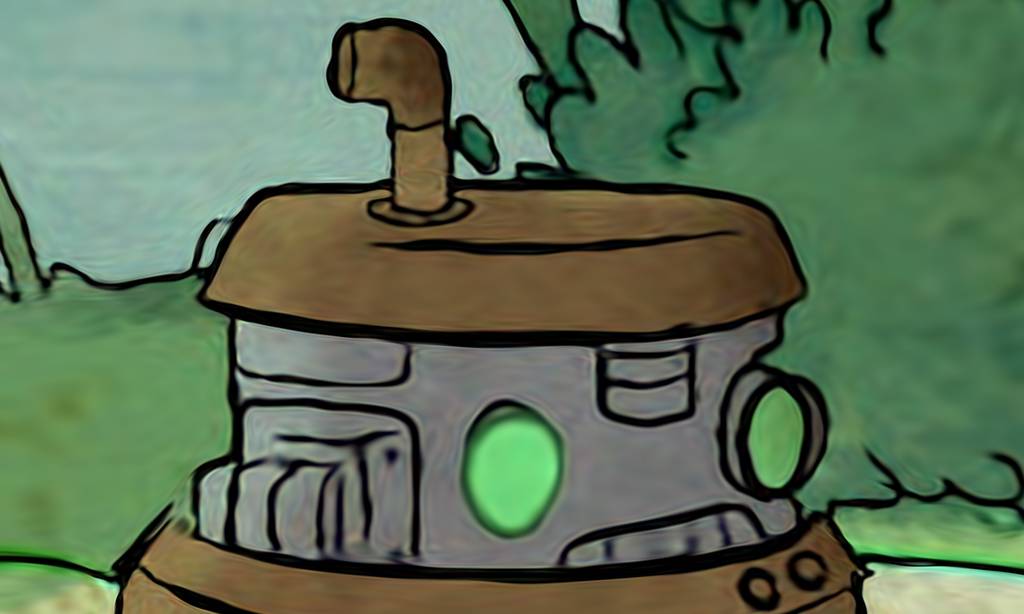} \\

        \rotatebox{90}{ \small \hspace{1em} \textbf{FDHO}} &
        \includegraphics[width=0.2\linewidth, height=0.12\linewidth]{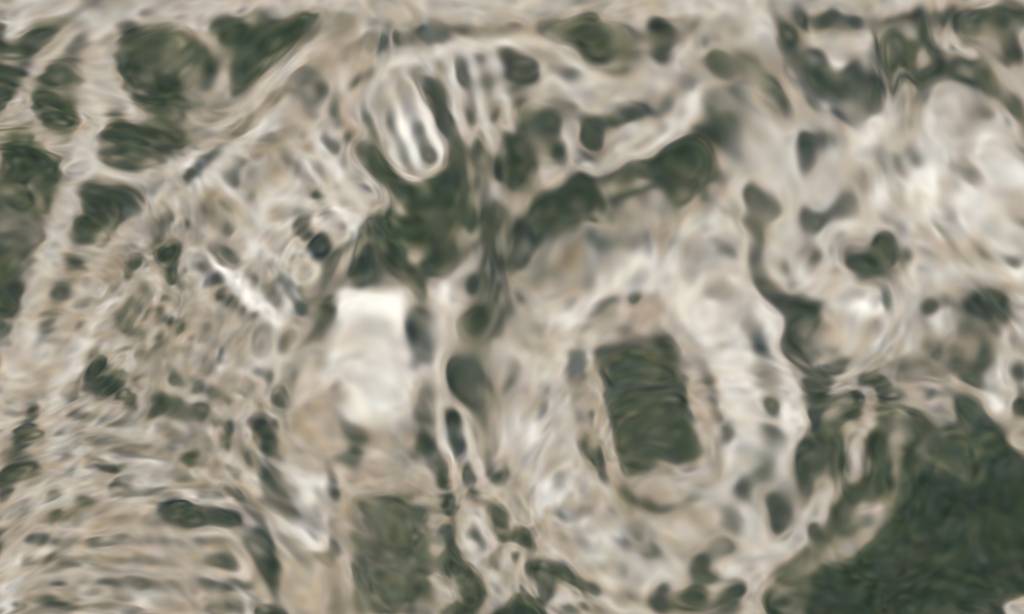} & 
        \includegraphics[width=0.2\linewidth, height=0.12\linewidth]{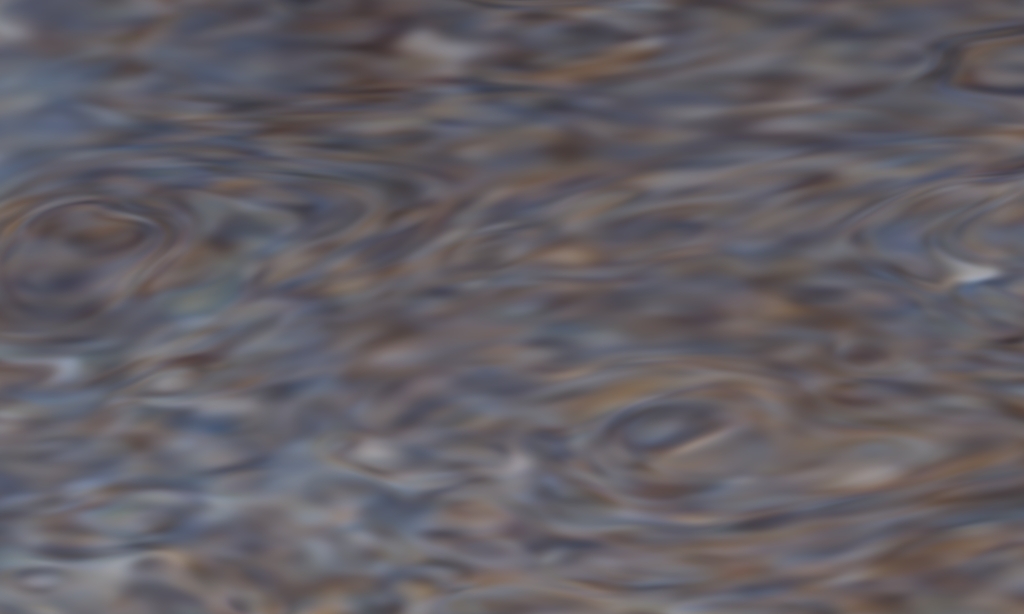} & 
        \includegraphics[width=0.2\linewidth, height=0.12\linewidth]{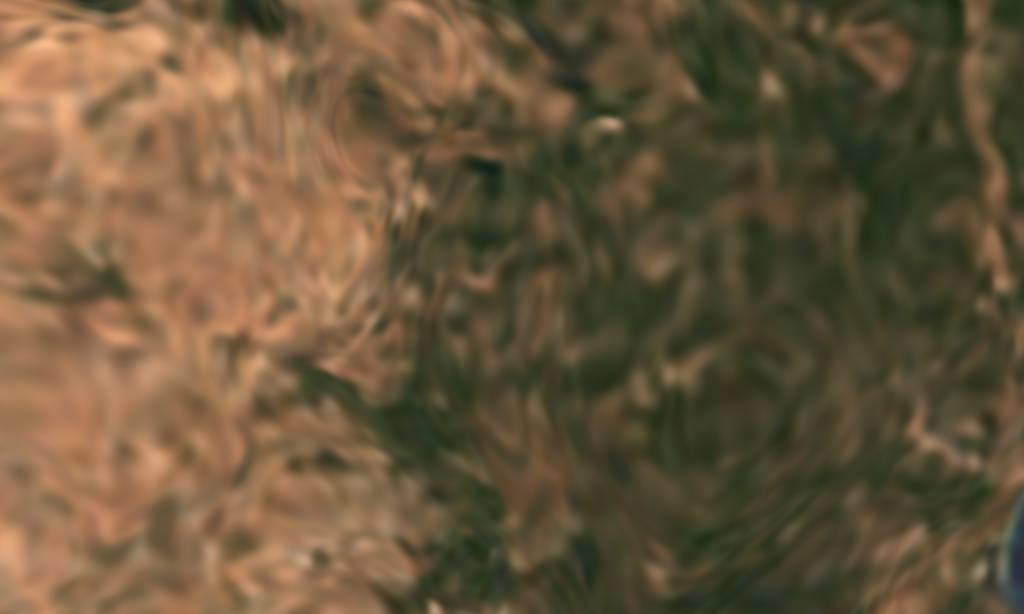} & 
        \includegraphics[width=0.2\linewidth, height=0.12\linewidth]{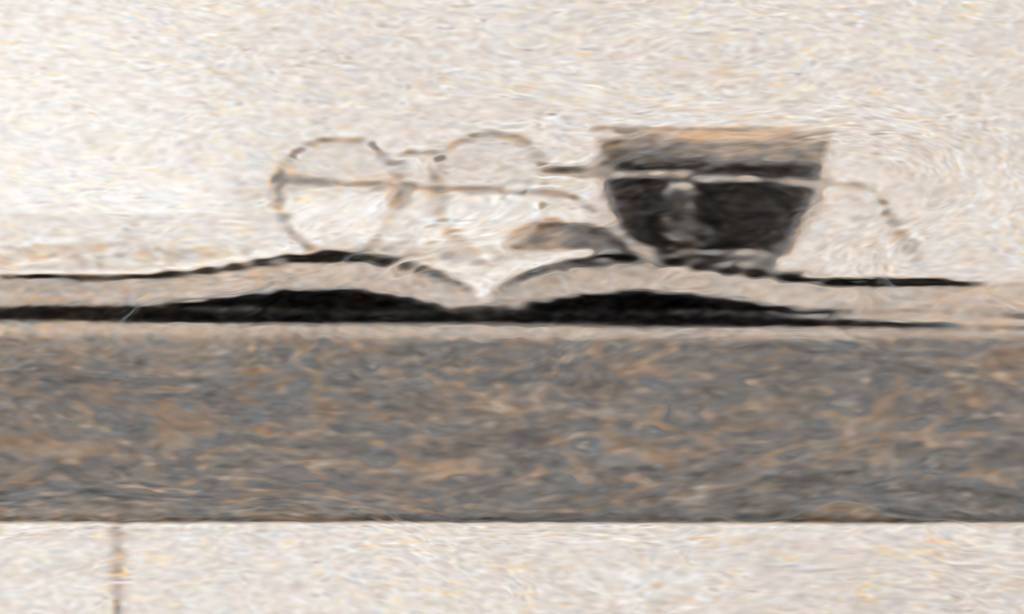} & 
        \includegraphics[width=0.2\linewidth, height=0.12\linewidth]{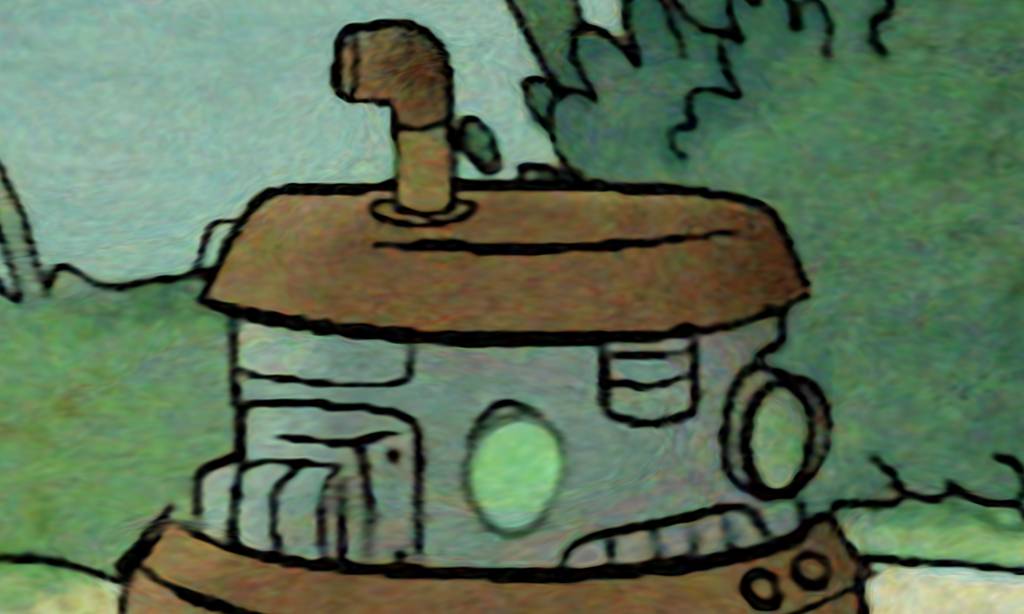} \\

        \rotatebox{90}{ \small \hspace{1em} \textbf{FINER}} &
        \includegraphics[width=0.2\linewidth, height=0.12\linewidth]{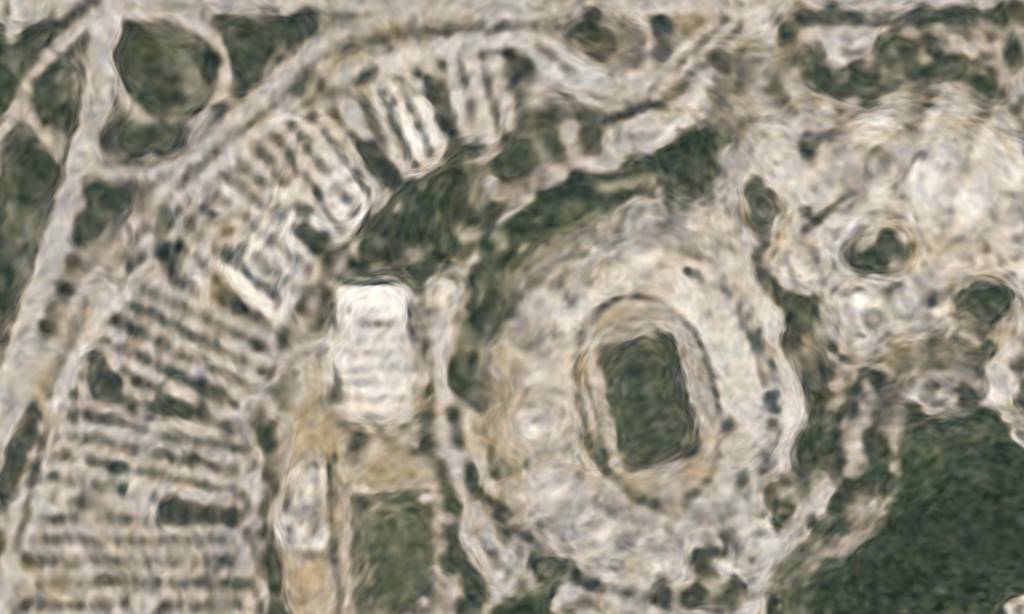} & 
        \includegraphics[width=0.2\linewidth, height=0.12\linewidth]{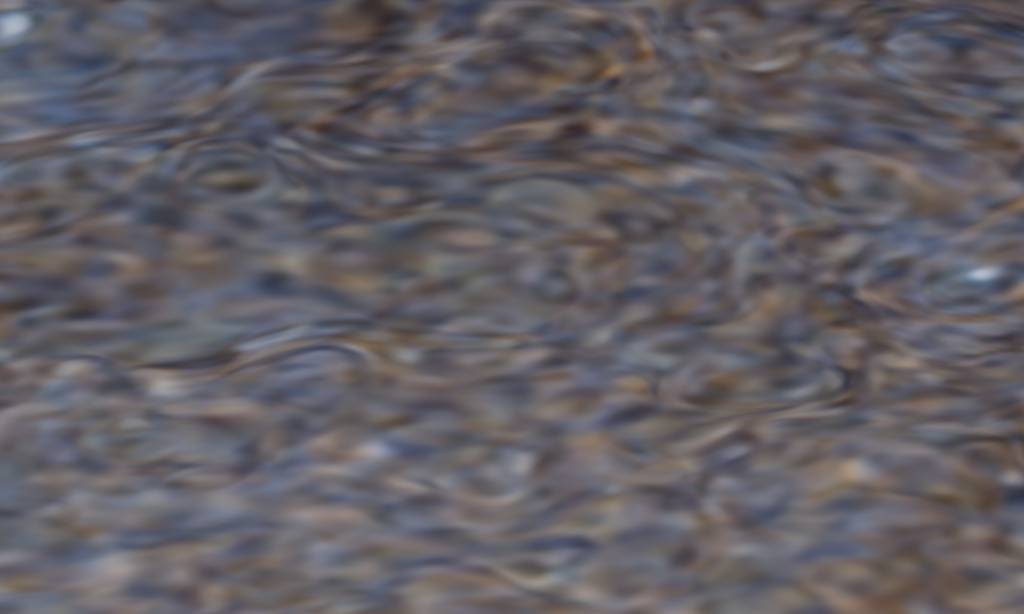} & 
        \includegraphics[width=0.2\linewidth, height=0.12\linewidth]{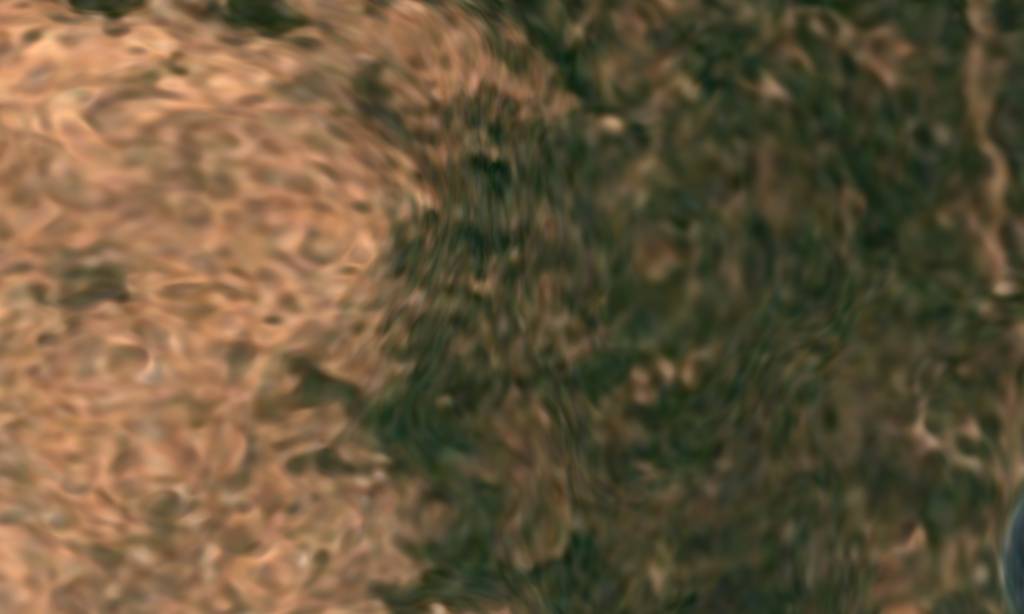} & 
        \includegraphics[width=0.2\linewidth, height=0.12\linewidth]{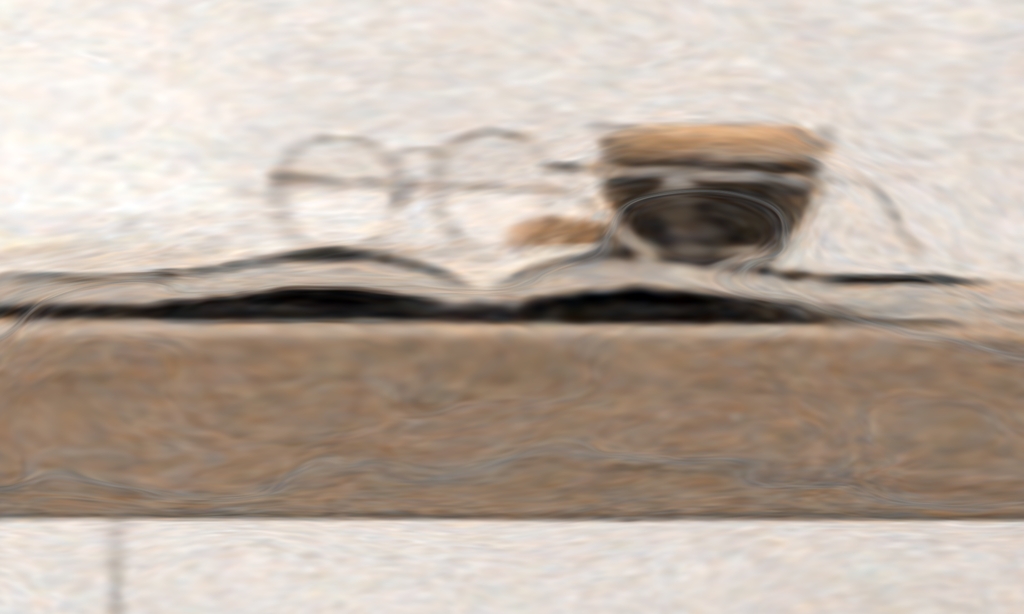} & 
        \includegraphics[width=0.2\linewidth, height=0.12\linewidth]{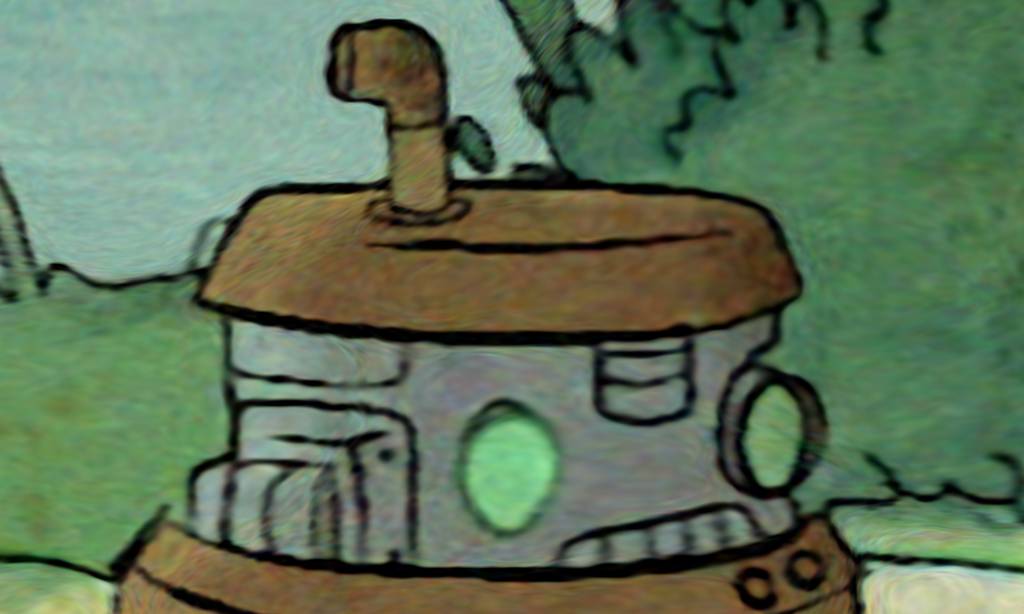} \\

        \rotatebox{90}{ \small \hspace{1.25em} \textbf{I-NGP}} &
        \includegraphics[width=0.2\linewidth, height=0.12\linewidth]{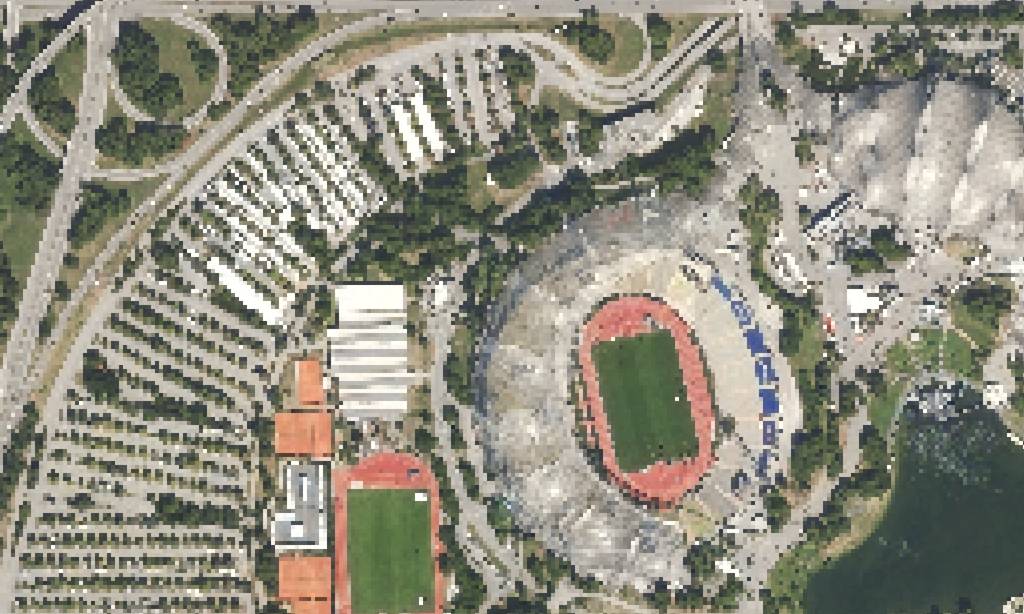} & 
        \includegraphics[width=0.2\linewidth, height=0.12\linewidth]{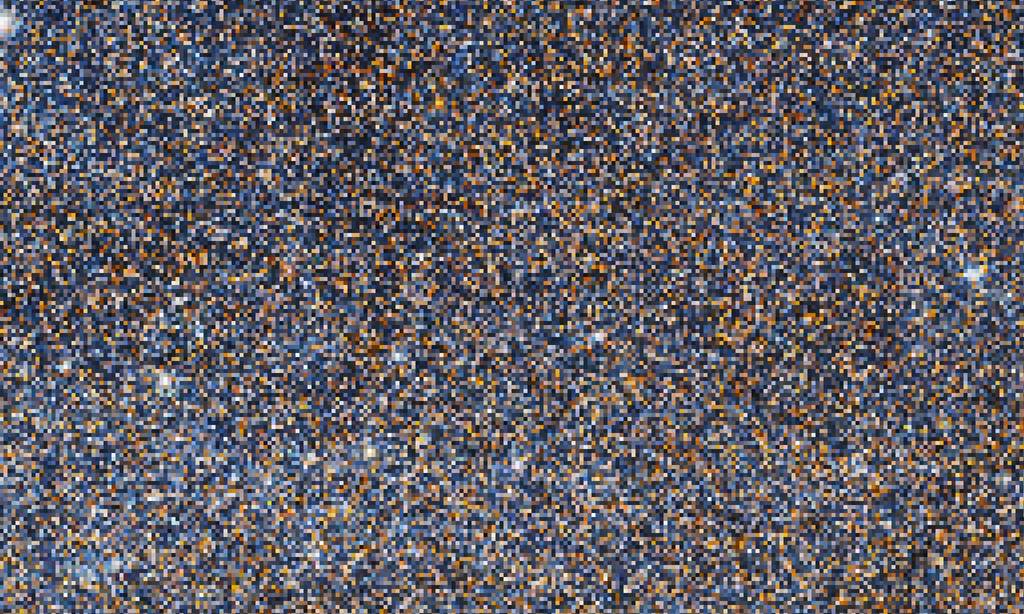} & 
        \includegraphics[width=0.2\linewidth, height=0.12\linewidth]{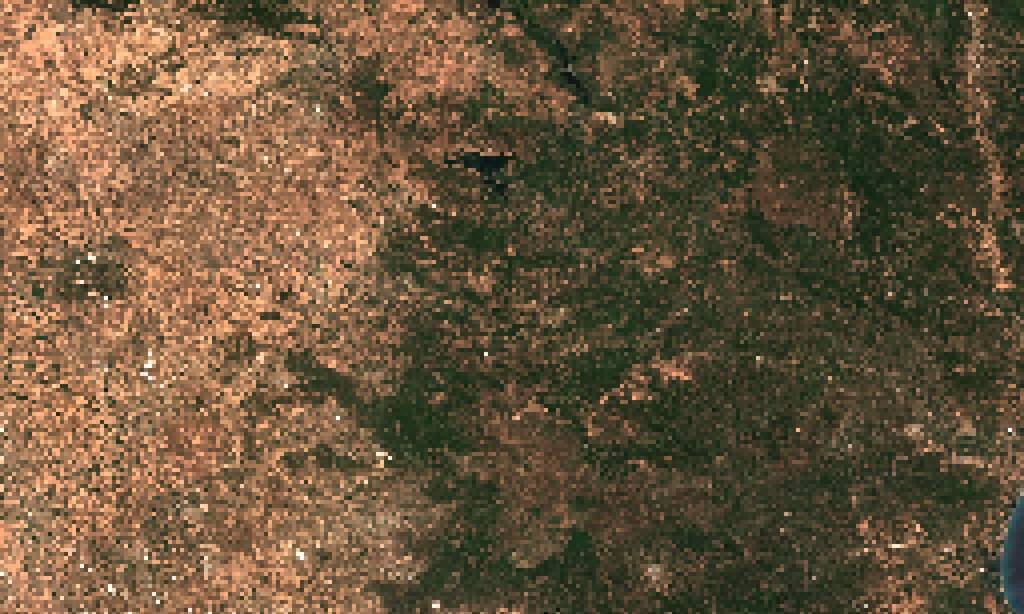} & 
        \includegraphics[width=0.2\linewidth, height=0.12\linewidth]{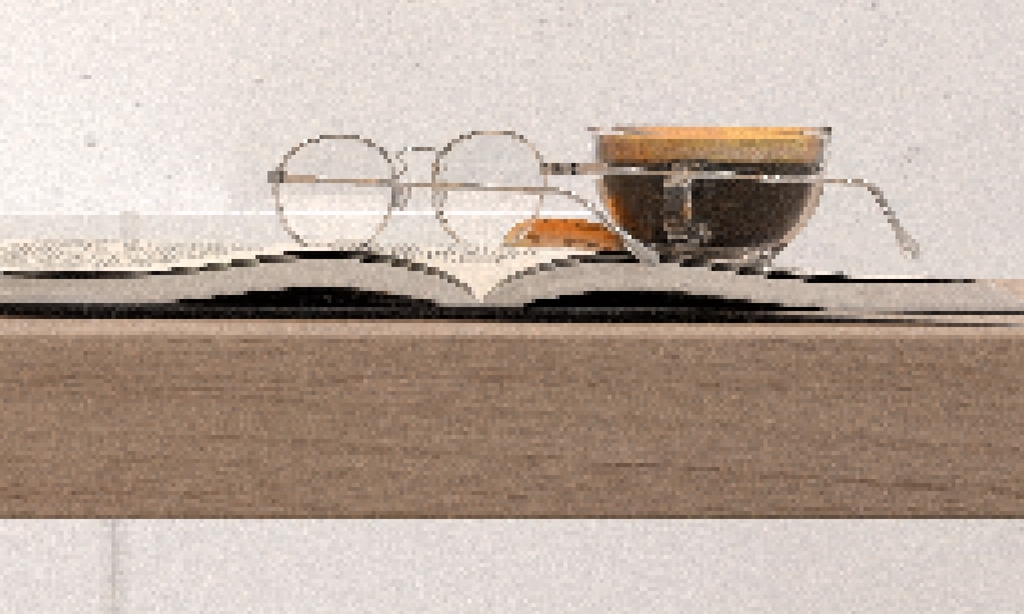} & 
        \includegraphics[width=0.2\linewidth, height=0.12\linewidth]{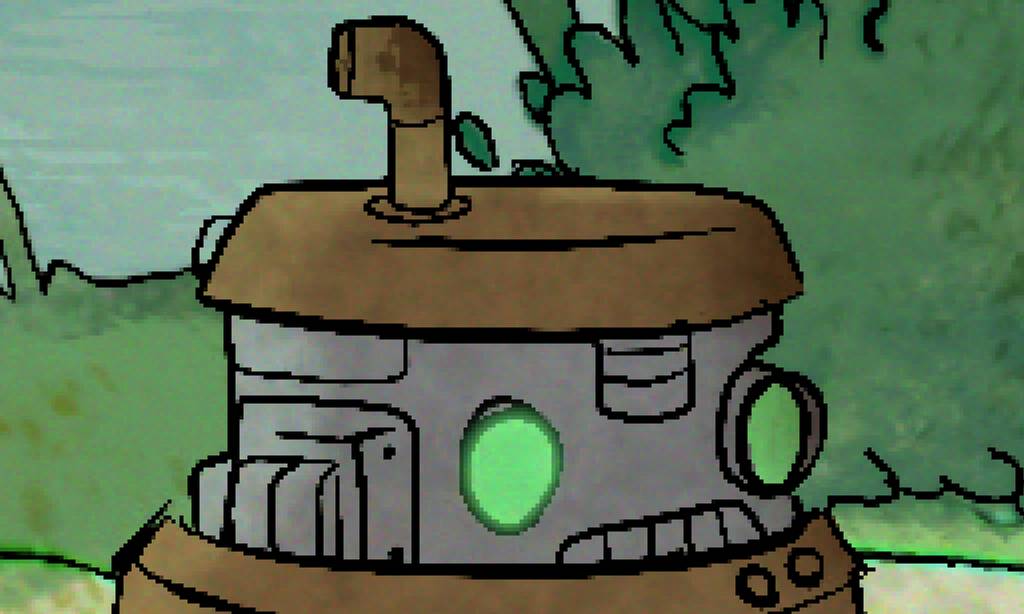} \\

        \rotatebox{90}{ \small \hspace{1em} \textbf{NeuRBF}} &
        \includegraphics[width=0.2\linewidth, height=0.12\linewidth]{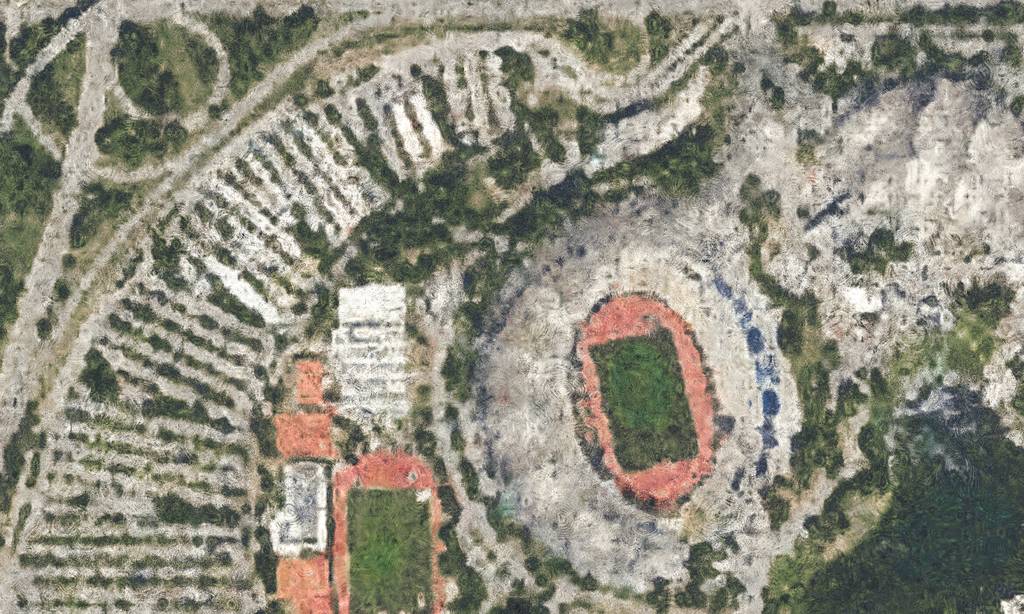} & 
        \includegraphics[width=0.2\linewidth, height=0.12\linewidth]{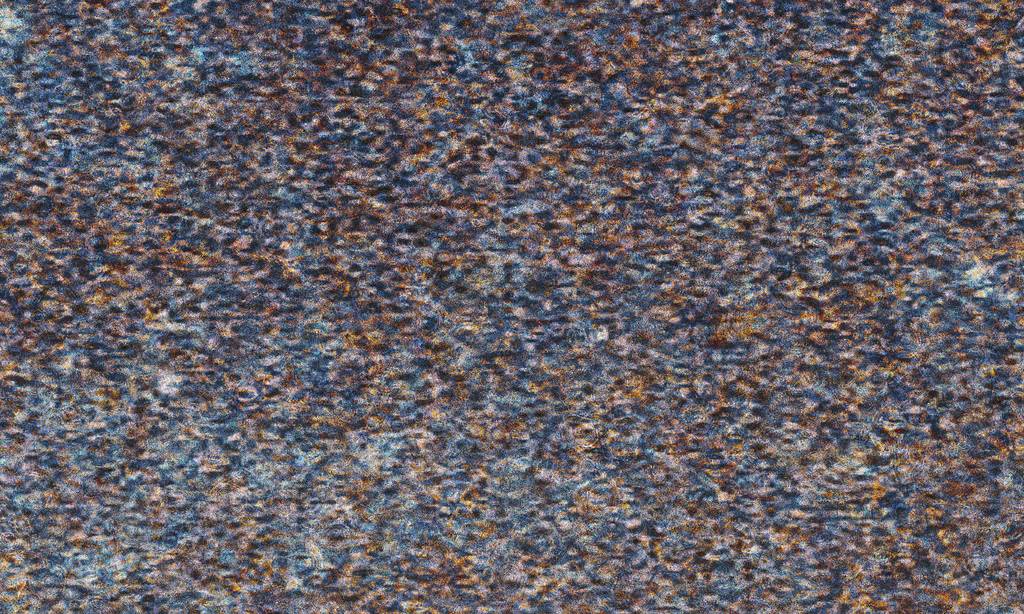} & 
        \includegraphics[width=0.2\linewidth, height=0.12\linewidth]{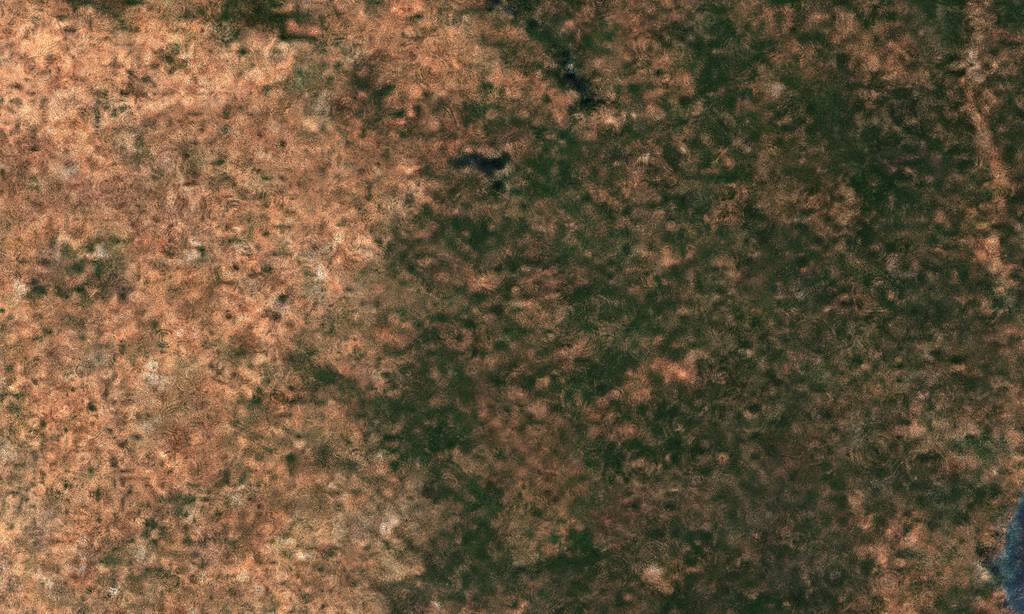} & 
        \includegraphics[width=0.2\linewidth, height=0.12\linewidth]{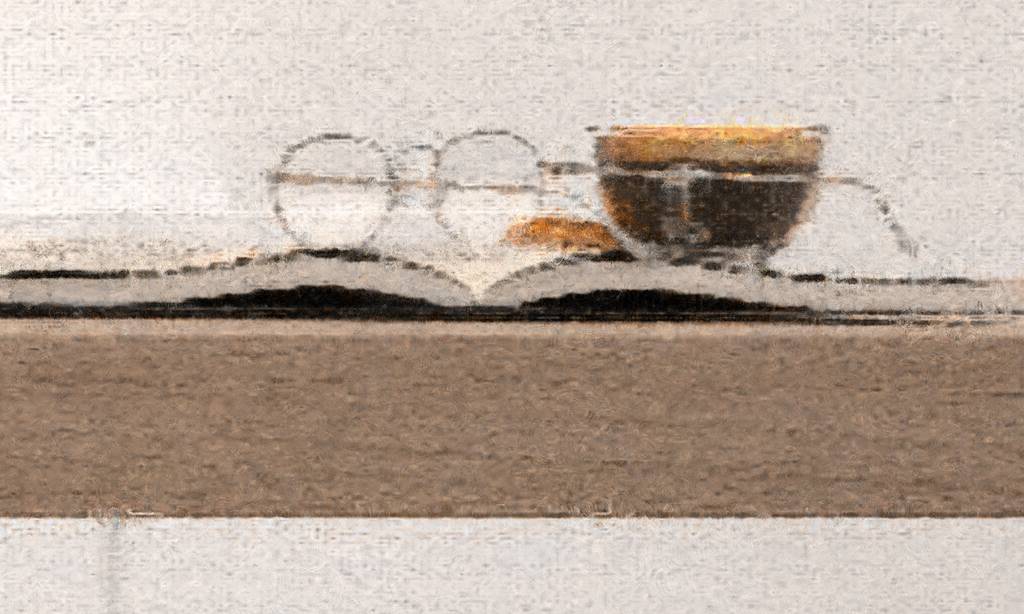} & 
        \includegraphics[width=0.2\linewidth, height=0.12\linewidth]{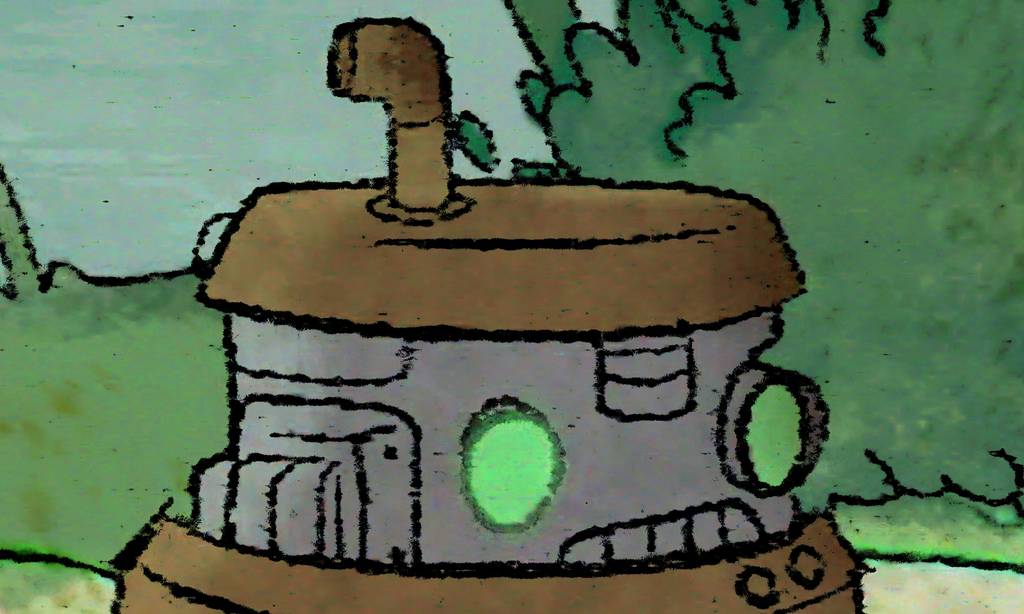} \\

        \rotatebox{90}{ \small \hspace{0.25em} \textbf{MetricGrids}} &
        \includegraphics[width=0.2\linewidth, height=0.12\linewidth]{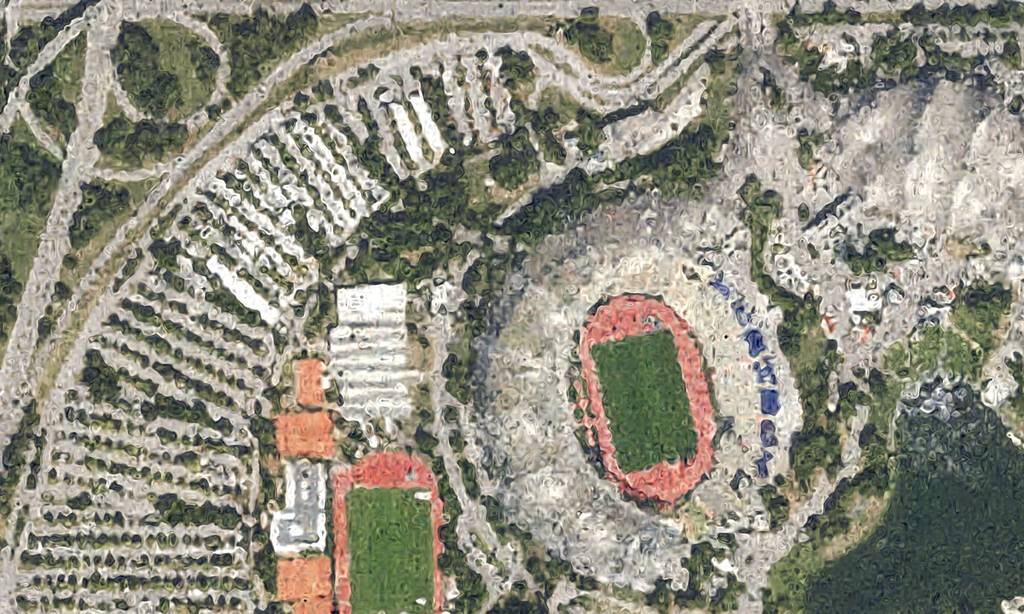} & 
        \includegraphics[width=0.2\linewidth, height=0.12\linewidth]{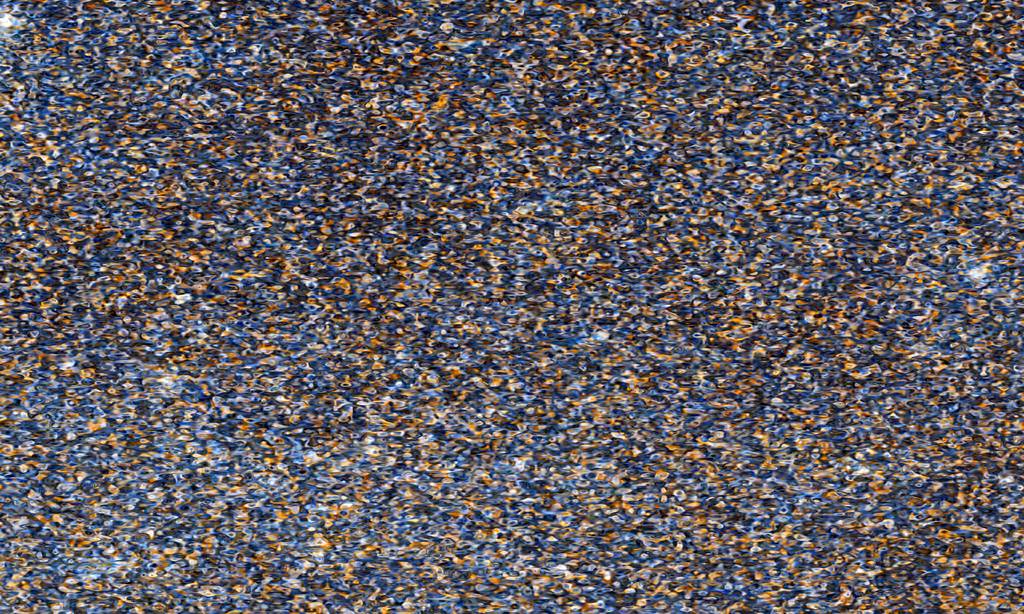} & 
        \includegraphics[width=0.2\linewidth, height=0.12\linewidth]{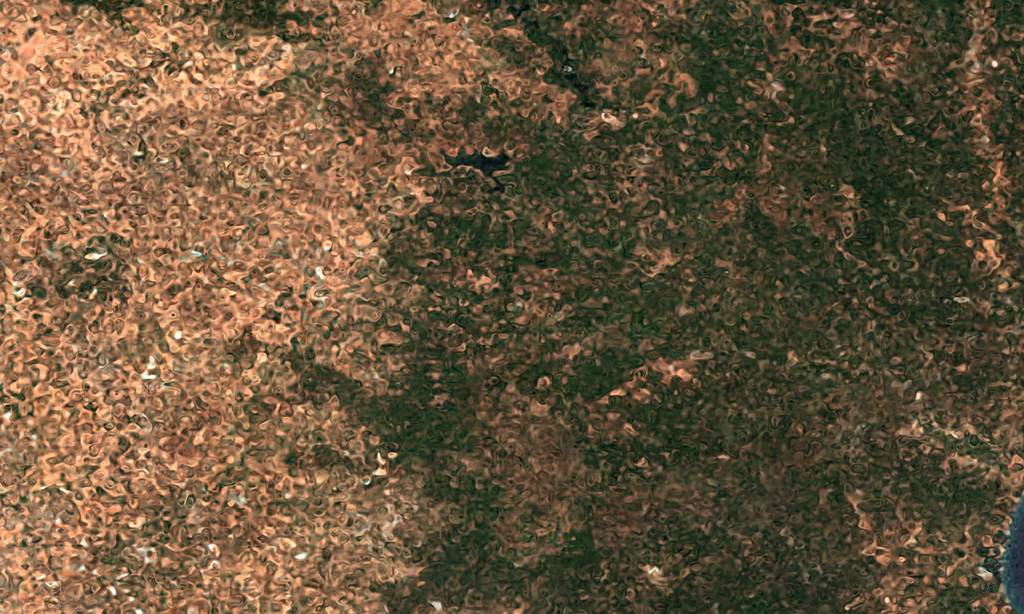} & 
        \includegraphics[width=0.2\linewidth, height=0.12\linewidth]{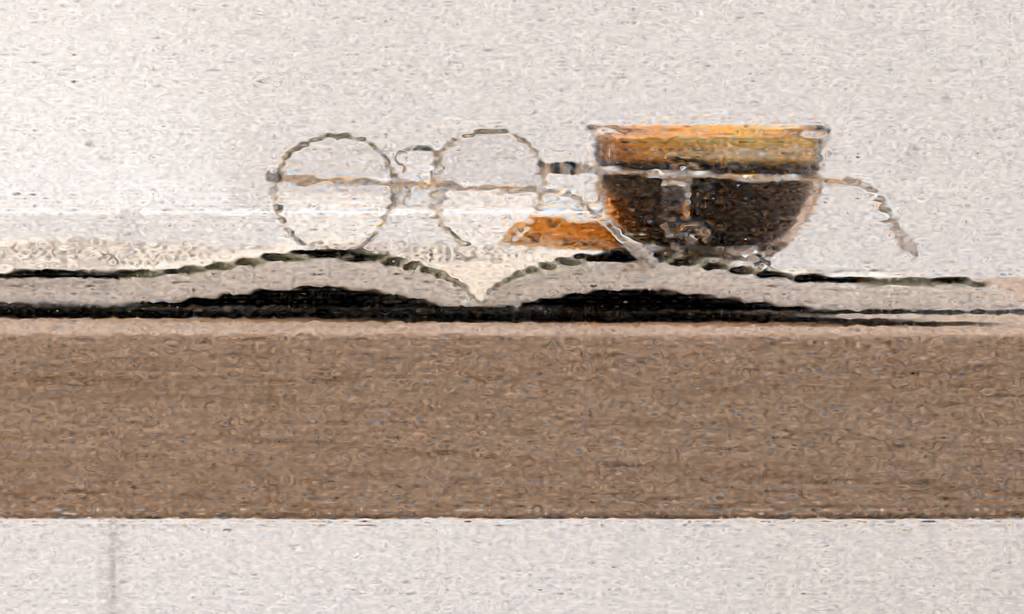} & 
        \includegraphics[width=0.2\linewidth, height=0.12\linewidth]{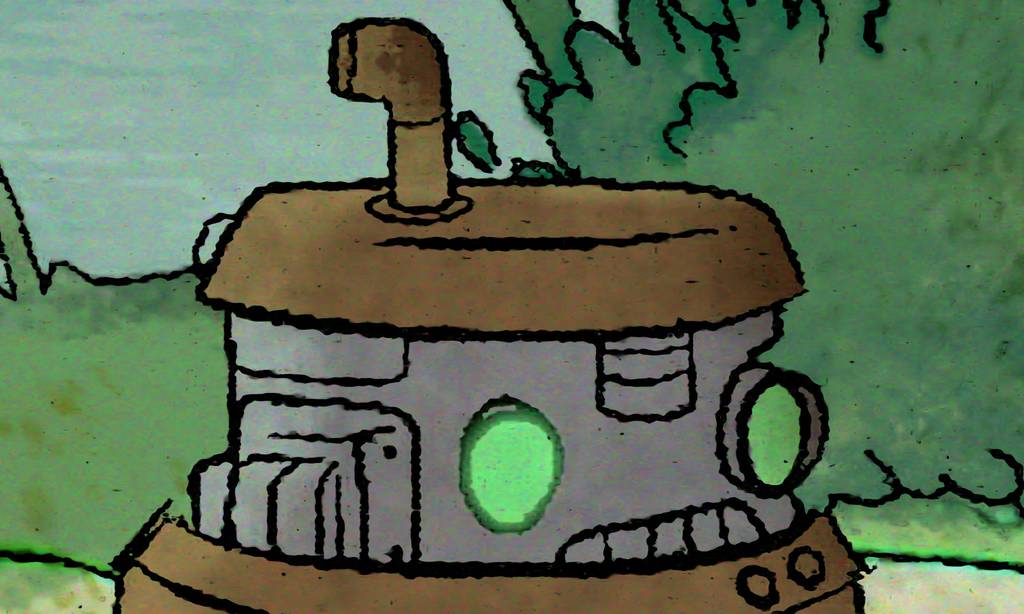} \\

        \rotatebox{90}{ \small \hspace{0.5em} \textbf{GaussImg}} &
        \includegraphics[width=0.2\linewidth, height=0.12\linewidth]{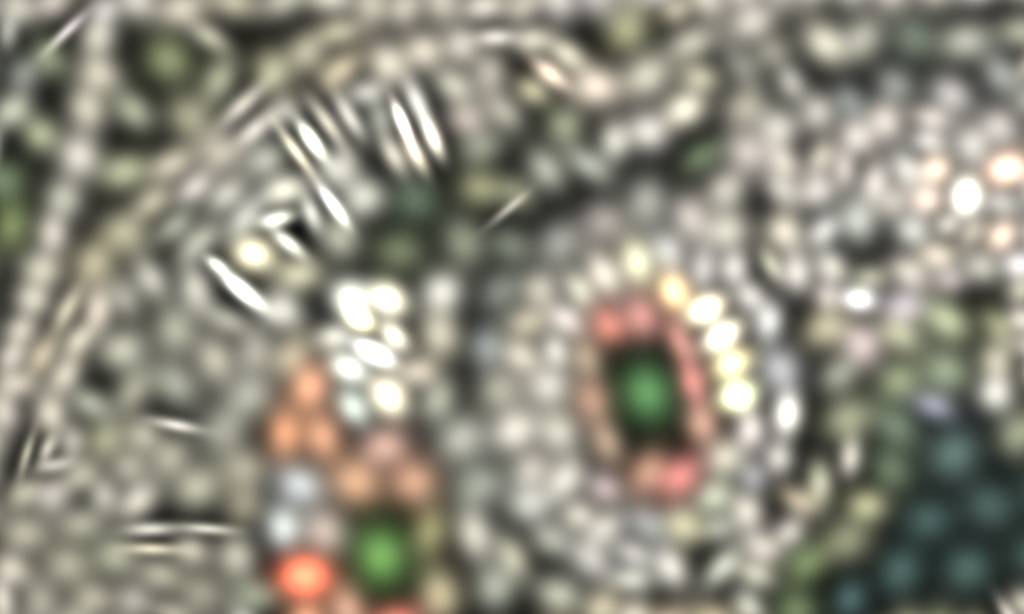} & 
        \includegraphics[width=0.2\linewidth, height=0.12\linewidth]{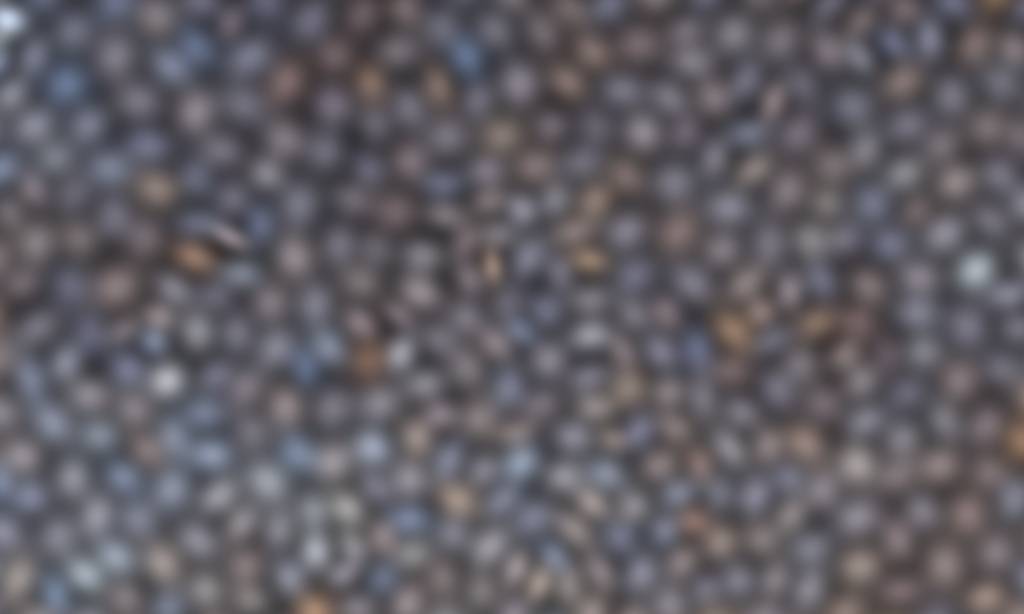} & 
        \includegraphics[width=0.2\linewidth, height=0.12\linewidth]{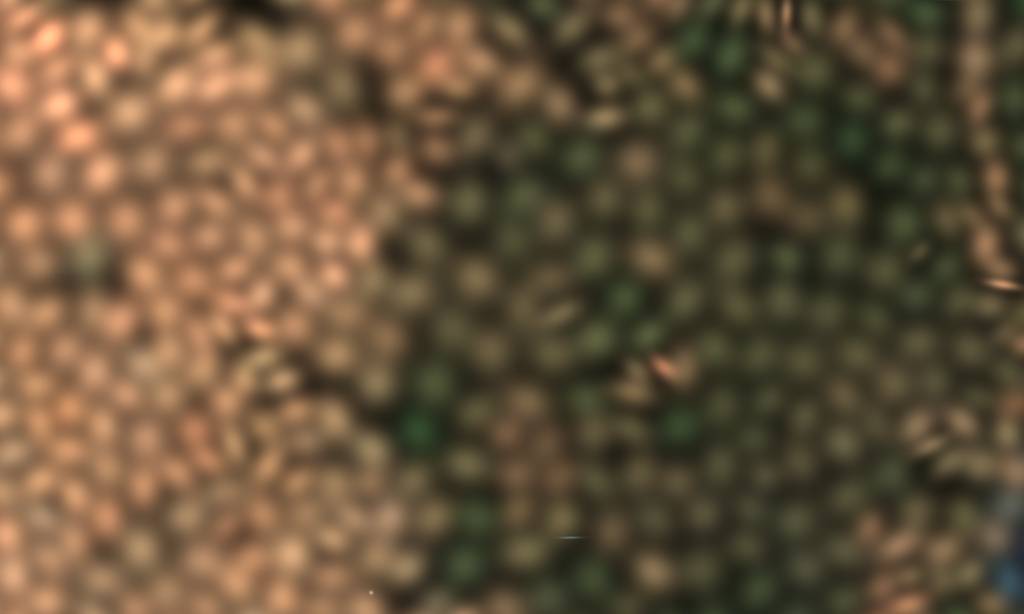} & 
        \includegraphics[width=0.2\linewidth, height=0.12\linewidth]{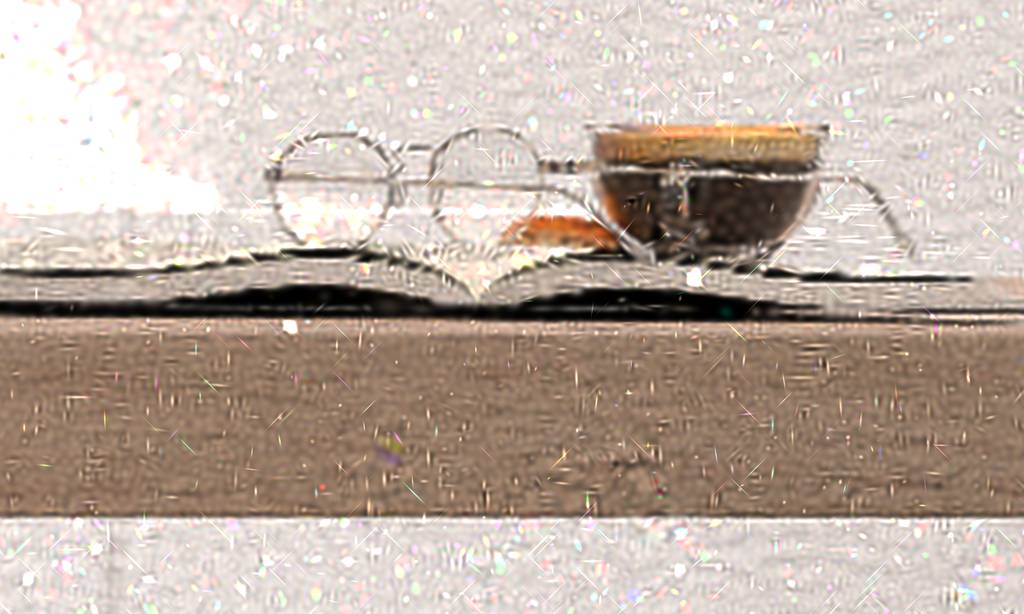} & 
        \includegraphics[width=0.2\linewidth, height=0.12\linewidth]{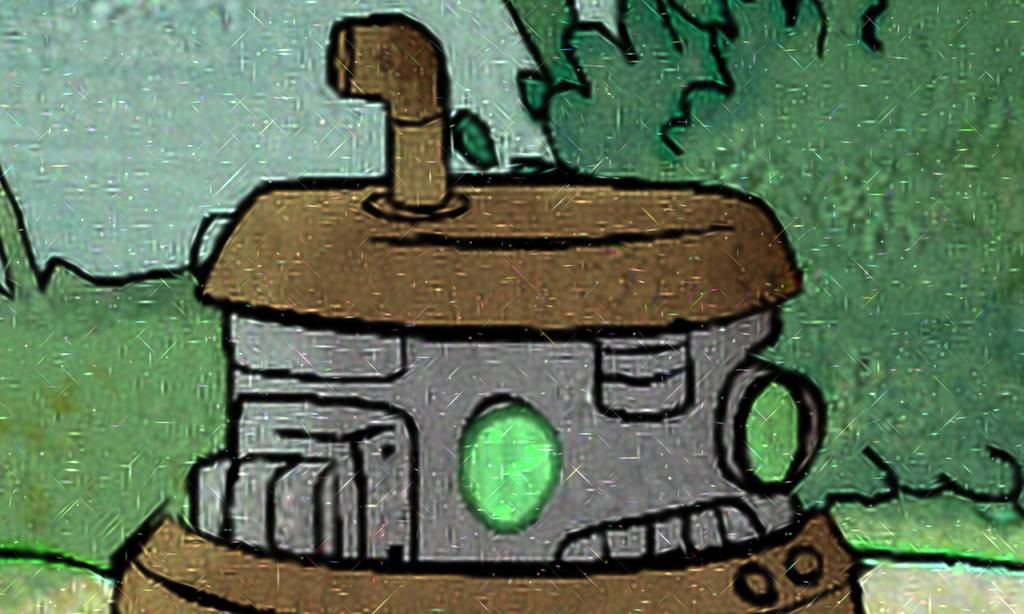} \\

        \rotatebox{90}{ \small \hspace{0.25em} \textbf{SmartSplat}} &
        \includegraphics[width=0.2\linewidth, height=0.12\linewidth]{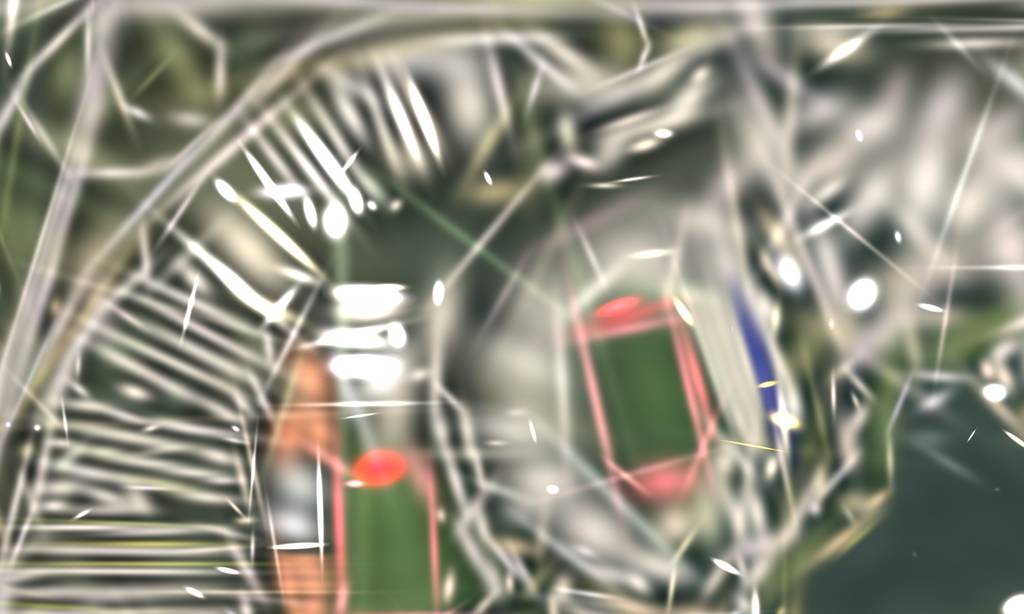} & 
        \includegraphics[width=0.2\linewidth, height=0.12\linewidth]{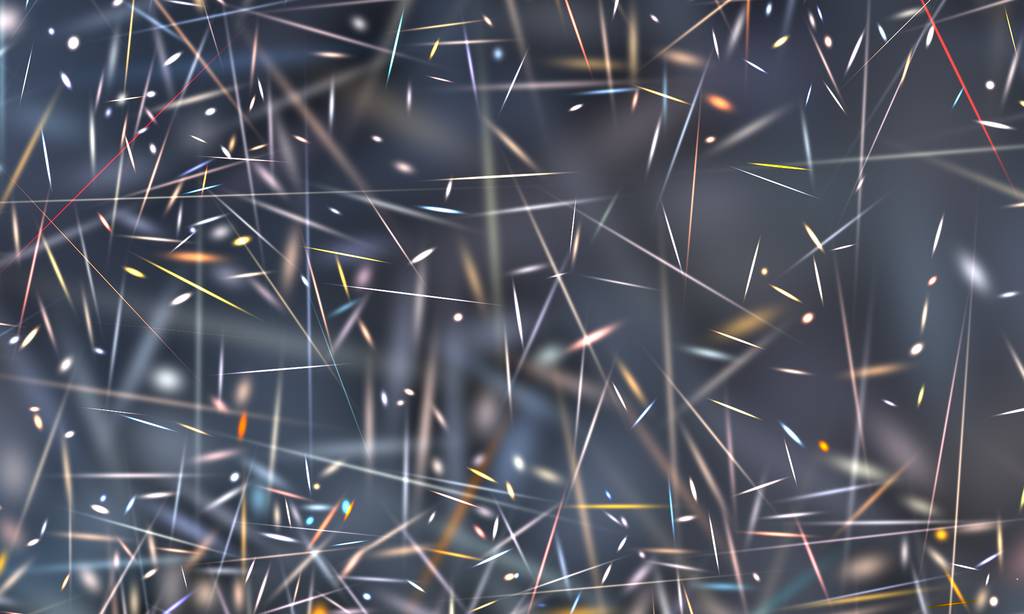} & 
        \includegraphics[width=0.2\linewidth, height=0.12\linewidth]{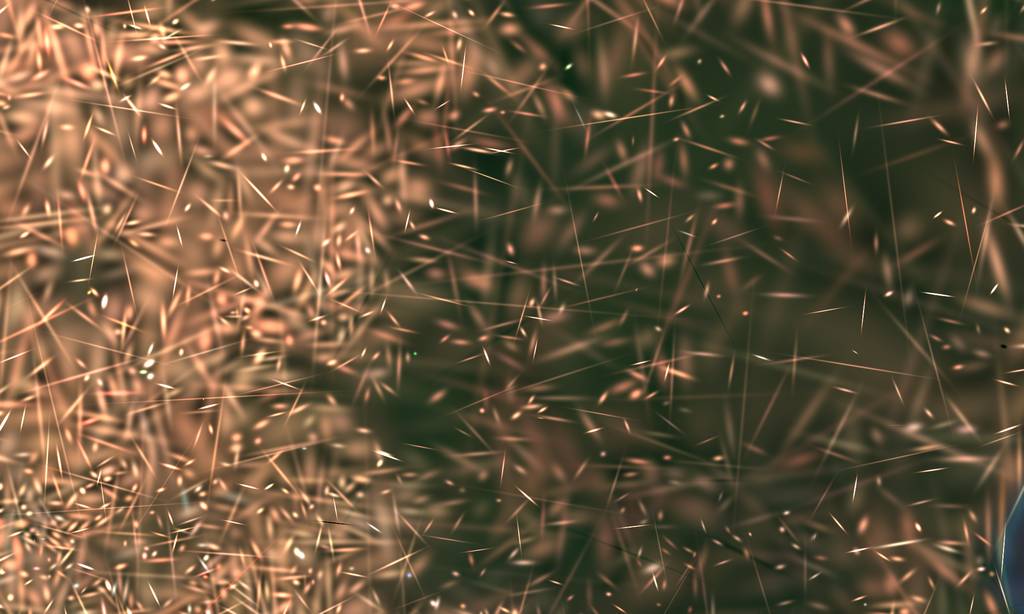} & 
        \includegraphics[width=0.2\linewidth, height=0.12\linewidth]{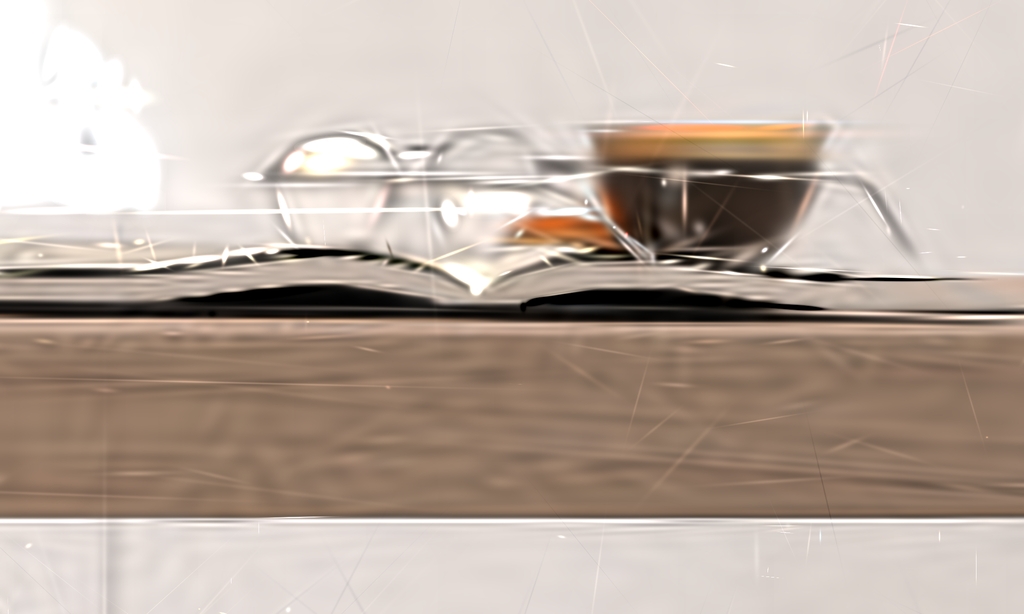} & 
        \includegraphics[width=0.2\linewidth, height=0.12\linewidth]{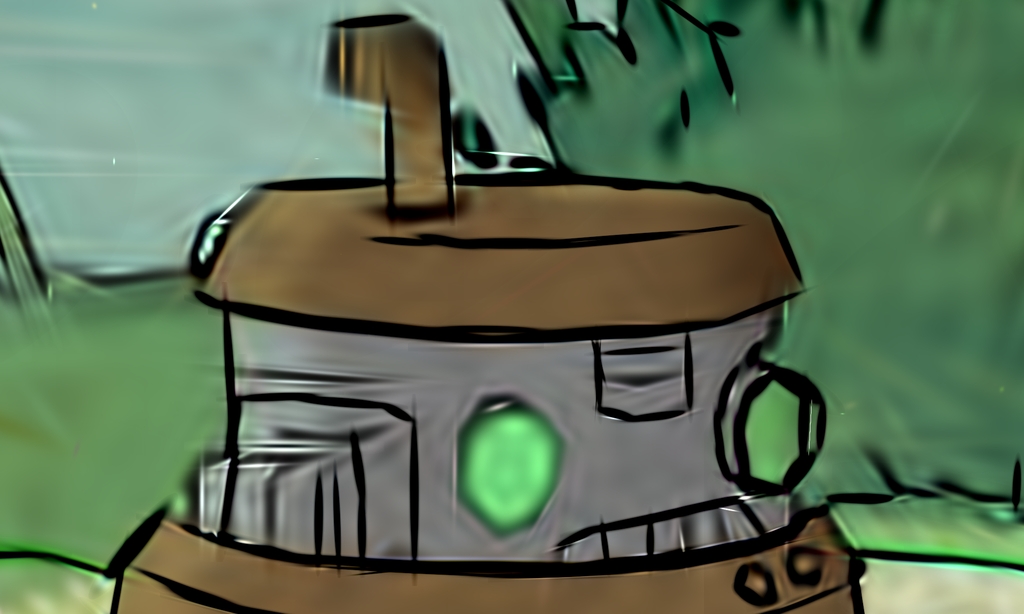} \\

        \rotatebox{90}{ \small \hspace{1em} \textbf{Bicubic}} &
        \includegraphics[width=0.2\linewidth, height=0.12\linewidth]{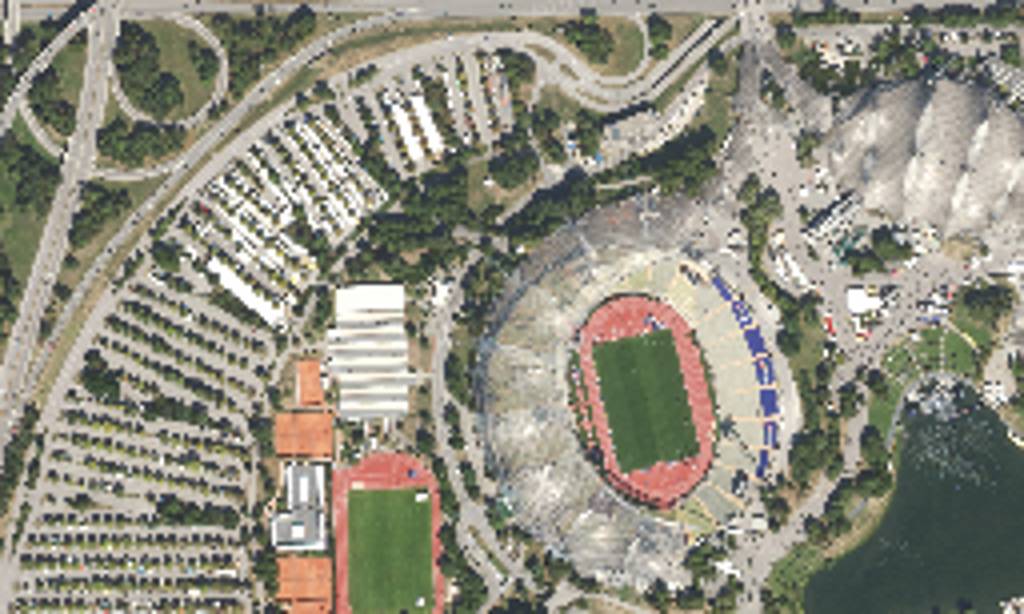} & 
        \includegraphics[width=0.2\linewidth, height=0.12\linewidth]{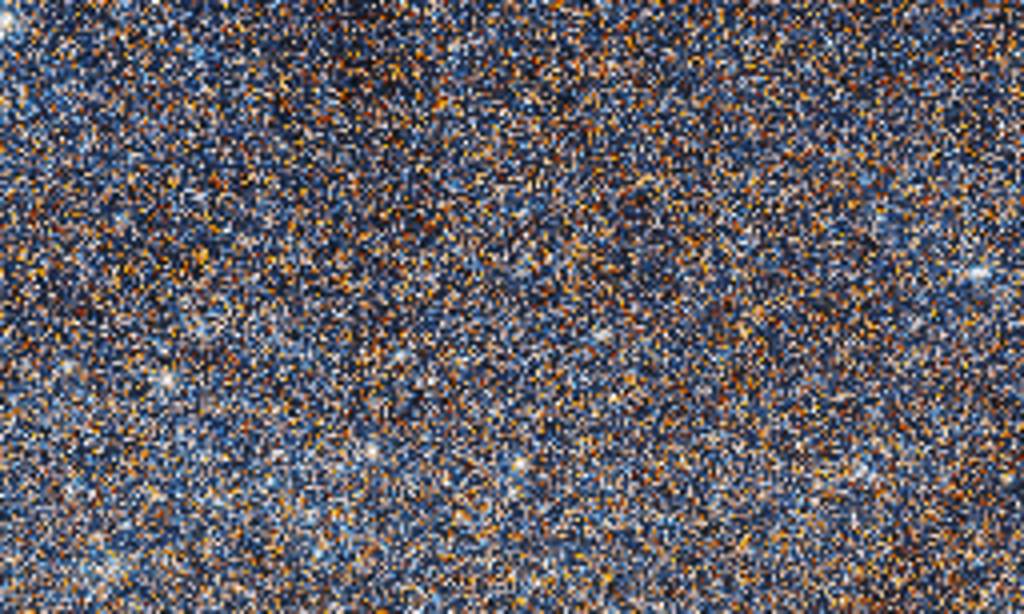} & 
        \includegraphics[width=0.2\linewidth, height=0.12\linewidth]{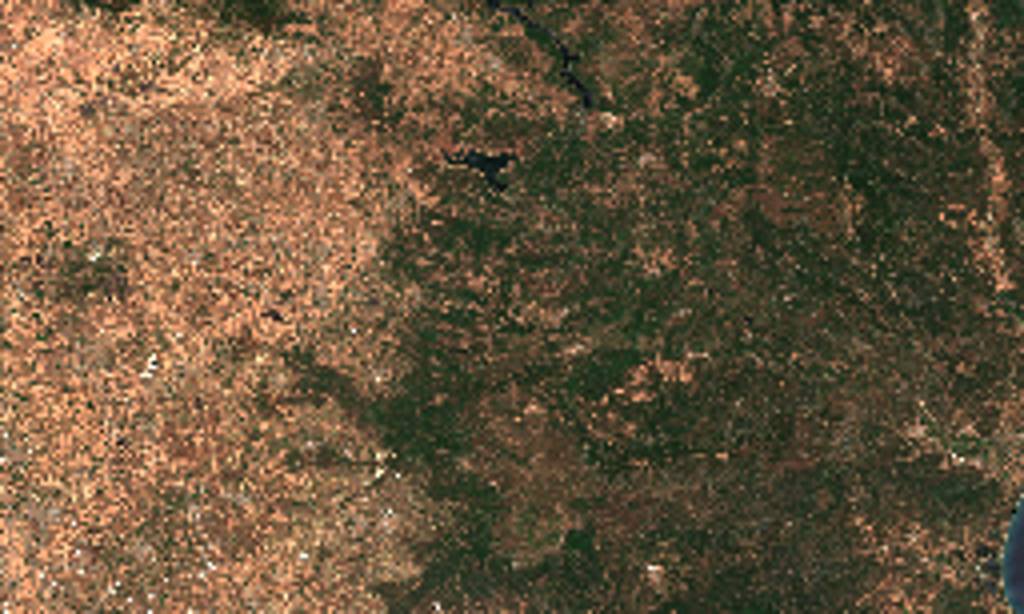} & 
        \includegraphics[width=0.2\linewidth, height=0.12\linewidth]{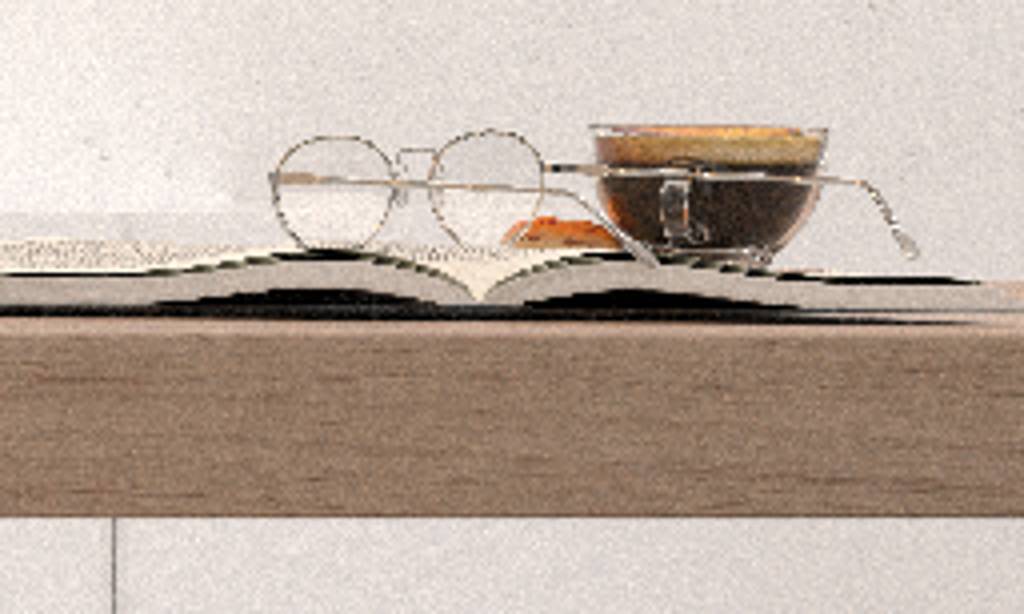} & 
        \includegraphics[width=0.2\linewidth, height=0.12\linewidth]{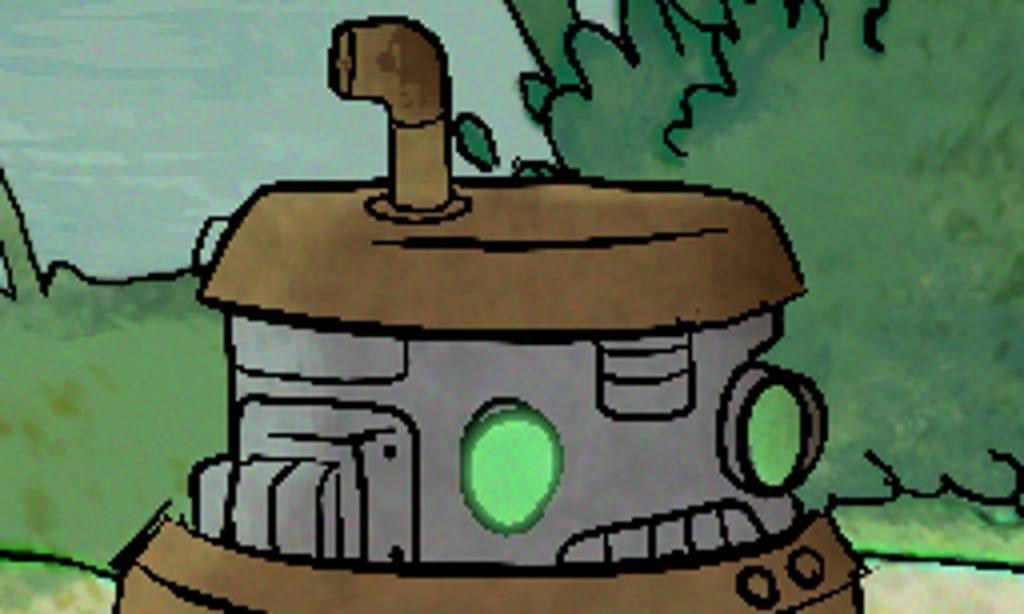} \\
    \end{tabular}}
    \caption{
    \textbf{Additional super resolution $\times8$ qualitatives results on \dataname{}.}
    We show $1228 \times 2048$ crops from the full resolution images.
    }

    \label{fig:suppl_sr_x8_qualitatives_2}

\end{figure*}

    We report additional qualitative results on \dataname{}.
    In particular, we display in \cref{fig:suppl_fit_qualitatives_1} and \cref{fig:suppl_fit_qualitatives_2} additional qualitative results on Image Compression, and in \cref{fig:suppl_sr_x4_qualitatives_1} \cref{fig:suppl_sr_x4_qualitatives_2}, \cref{fig:suppl_sr_x8_qualitatives_1} and \cref{fig:suppl_sr_x8_qualitatives_2} additional qualitative results for Image Super-Resolution.
    
    For each model, we have included the result relative to \textbf{the most performant version}, in terms of PSNR.

\begin{figure*}[ht]
    \centering
    \resizebox{0.6\linewidth}{!}{
    \setlength{\tabcolsep}{1px}
    \begin{tabular}{c ccc}
        & \textbf{SIREN} & \textbf{FDHO} & \textbf{FINER} \\
        \rotatebox{90}{\hspace{3.2em}\textit{(5, 256)}} & 
            \includegraphics[width=0.2\linewidth]{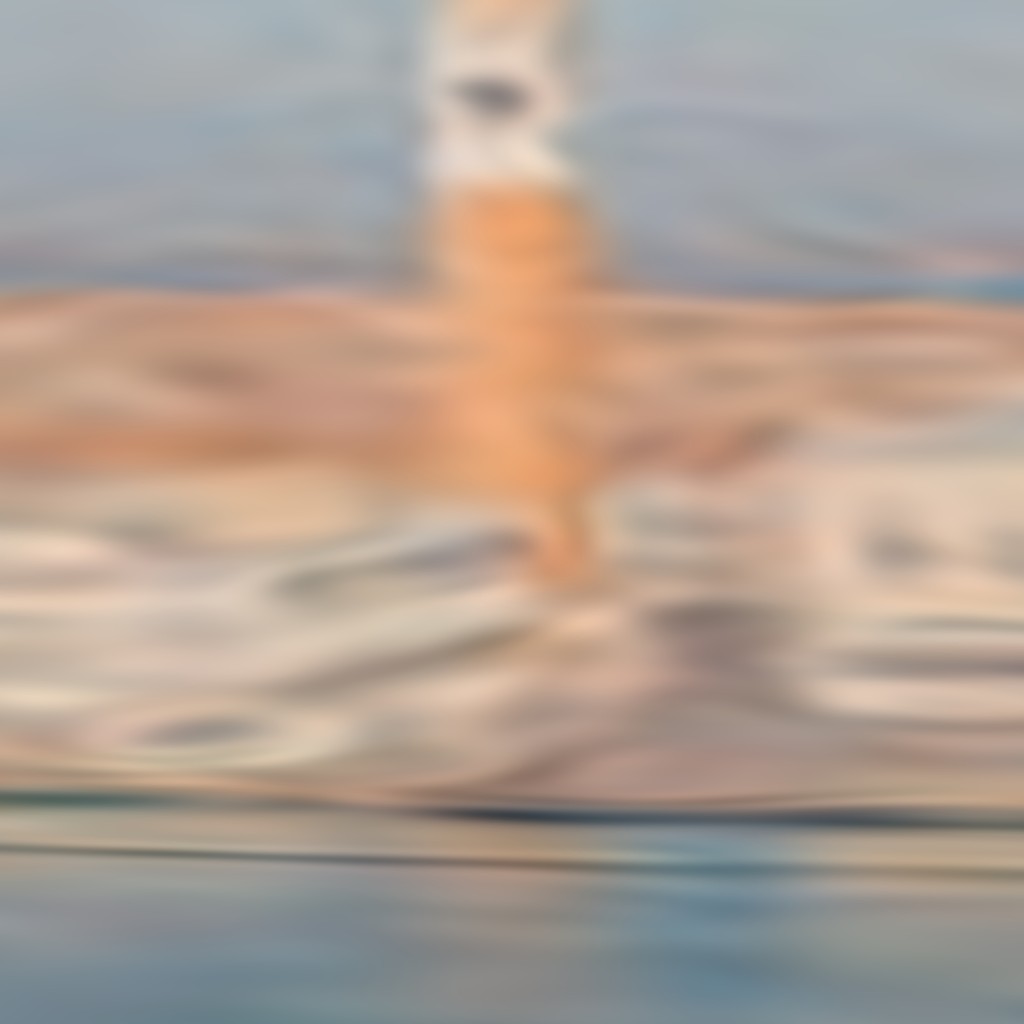} & 
            \includegraphics[width=0.2\linewidth]{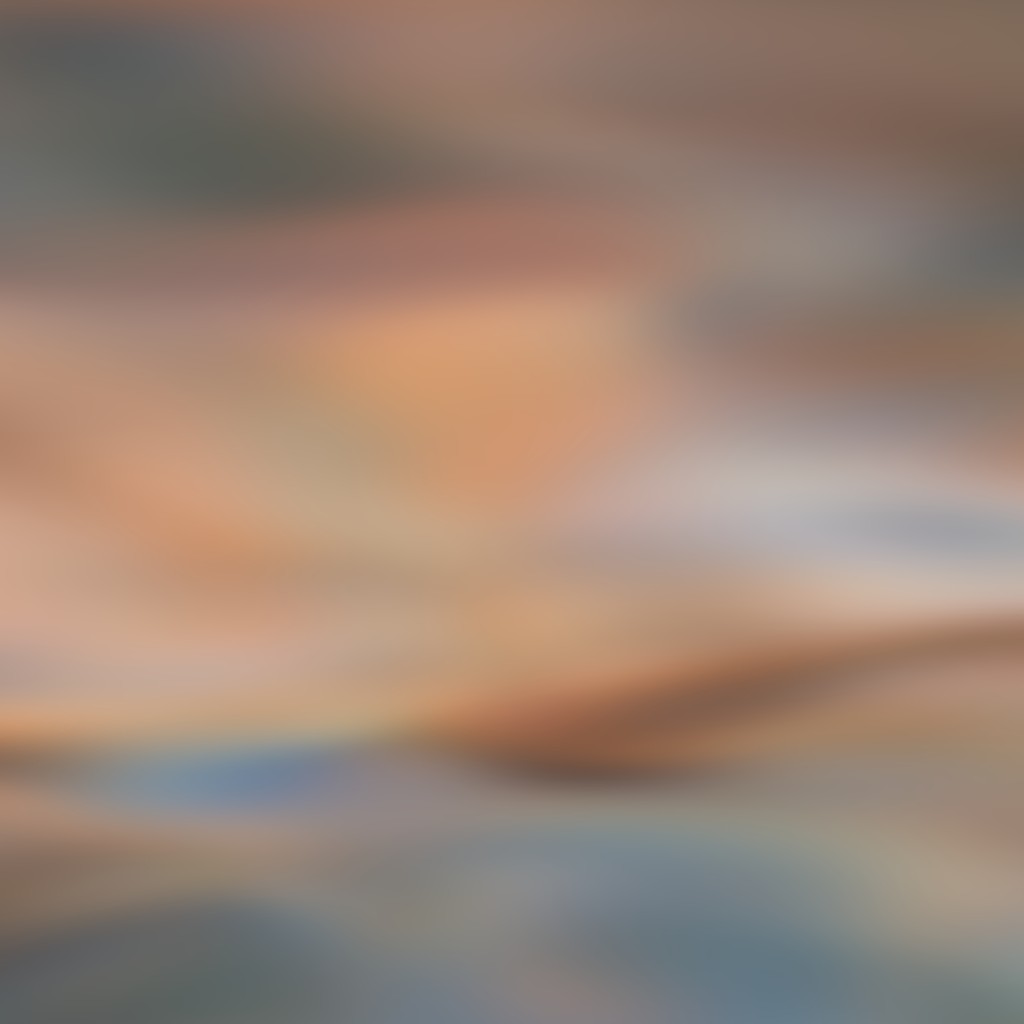} & 
            \includegraphics[width=0.2\linewidth]{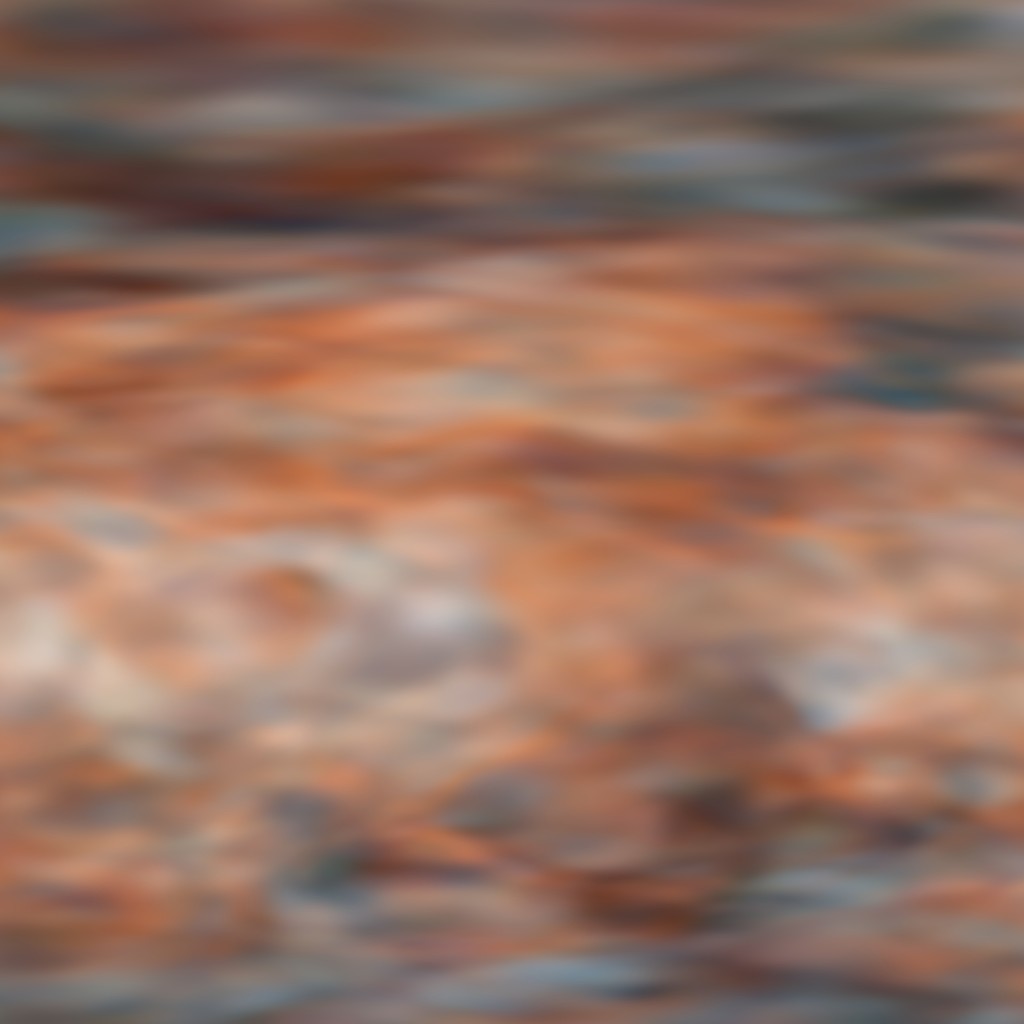} \\
        \rotatebox{90}{\hspace{3.2em}\textit{(8, 1024)}} & 
            \includegraphics[width=0.2\linewidth]{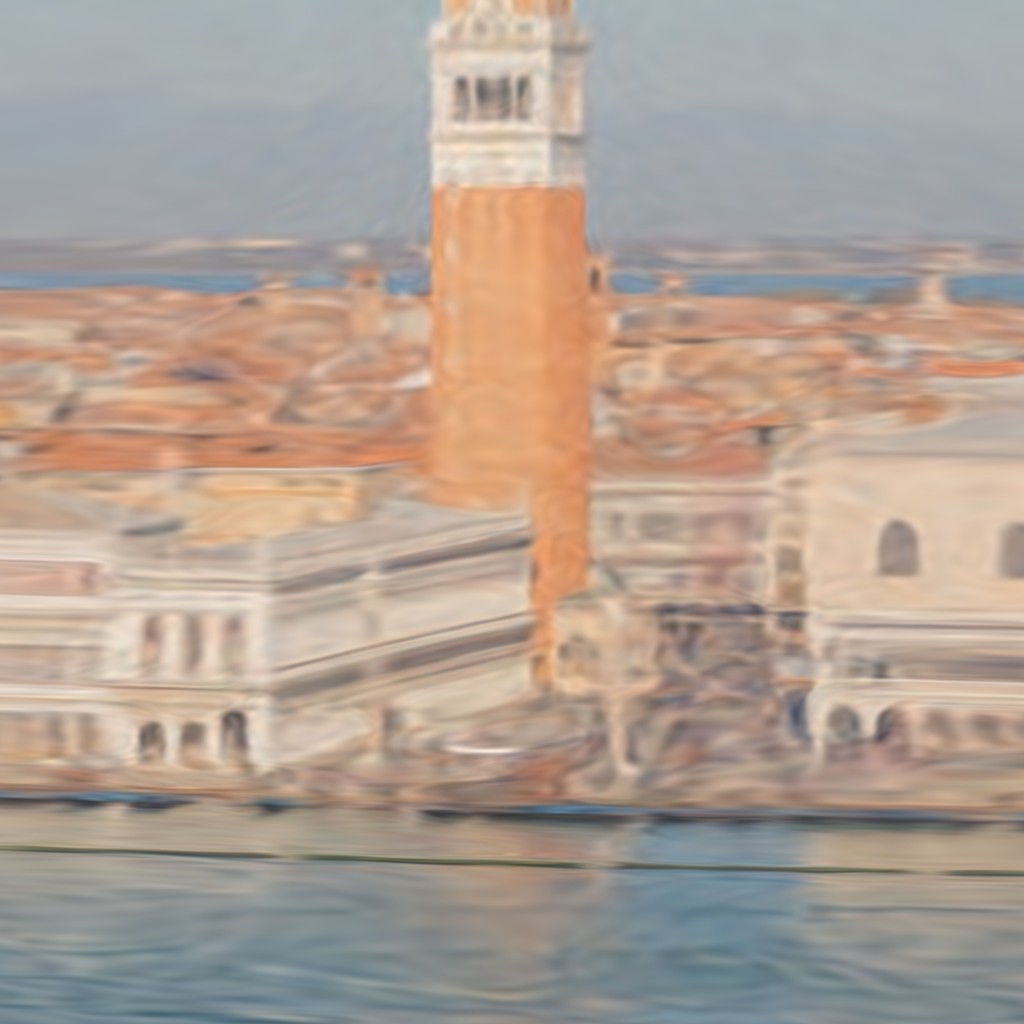} & 
            \includegraphics[width=0.2\linewidth]{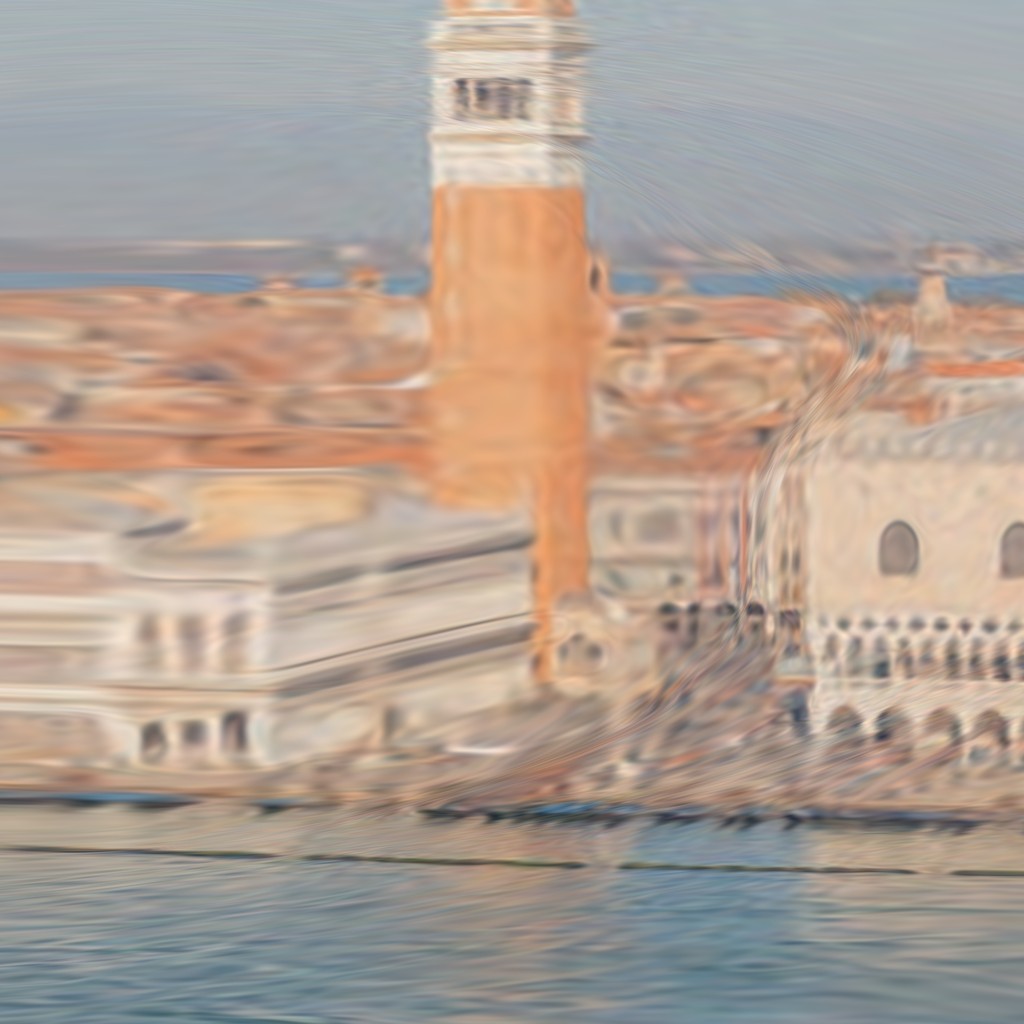} & 
            \includegraphics[width=0.2\linewidth]{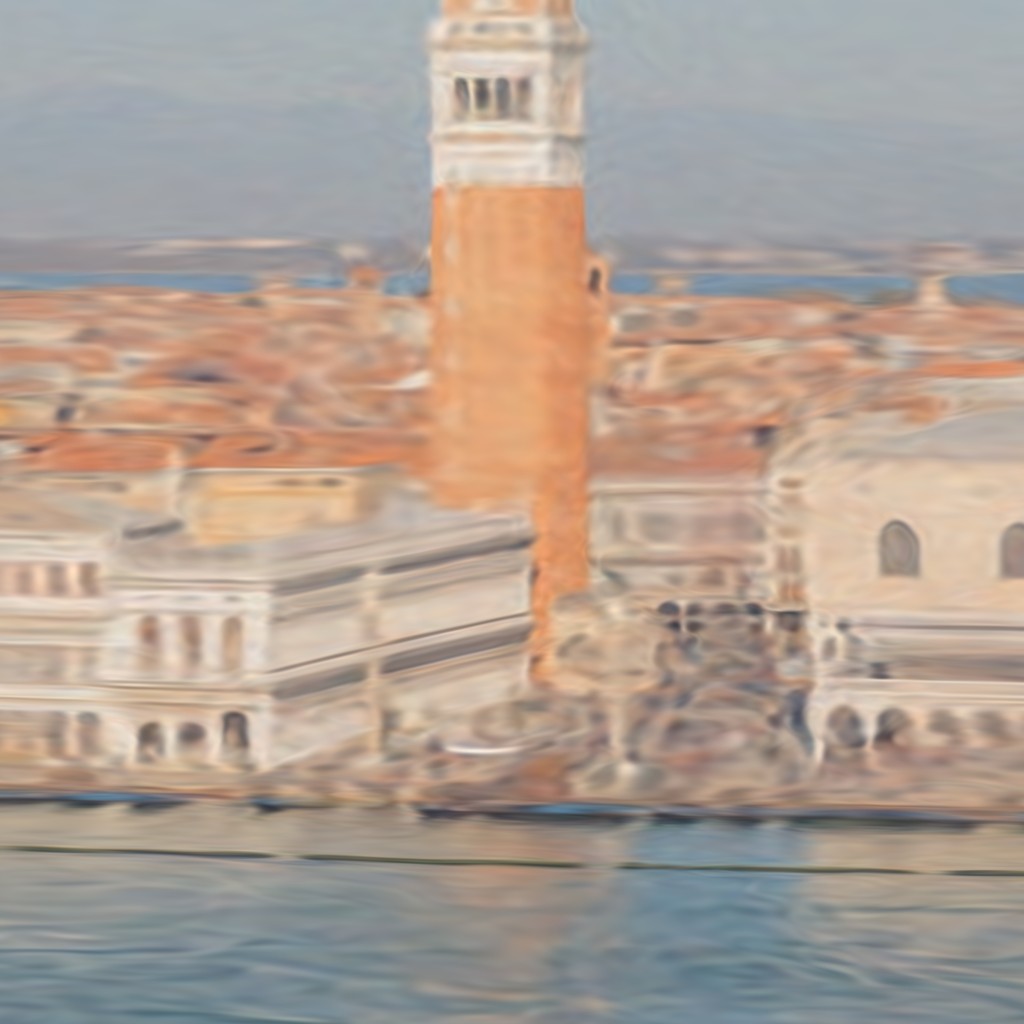} \\

        & \textbf{I-NGP} & \textbf{NeuRBF} & \textbf{MetricGrids} \\
        \rotatebox{90}{\hspace{3.8em}\textit{T=22}} & 
            \includegraphics[width=0.2\linewidth]{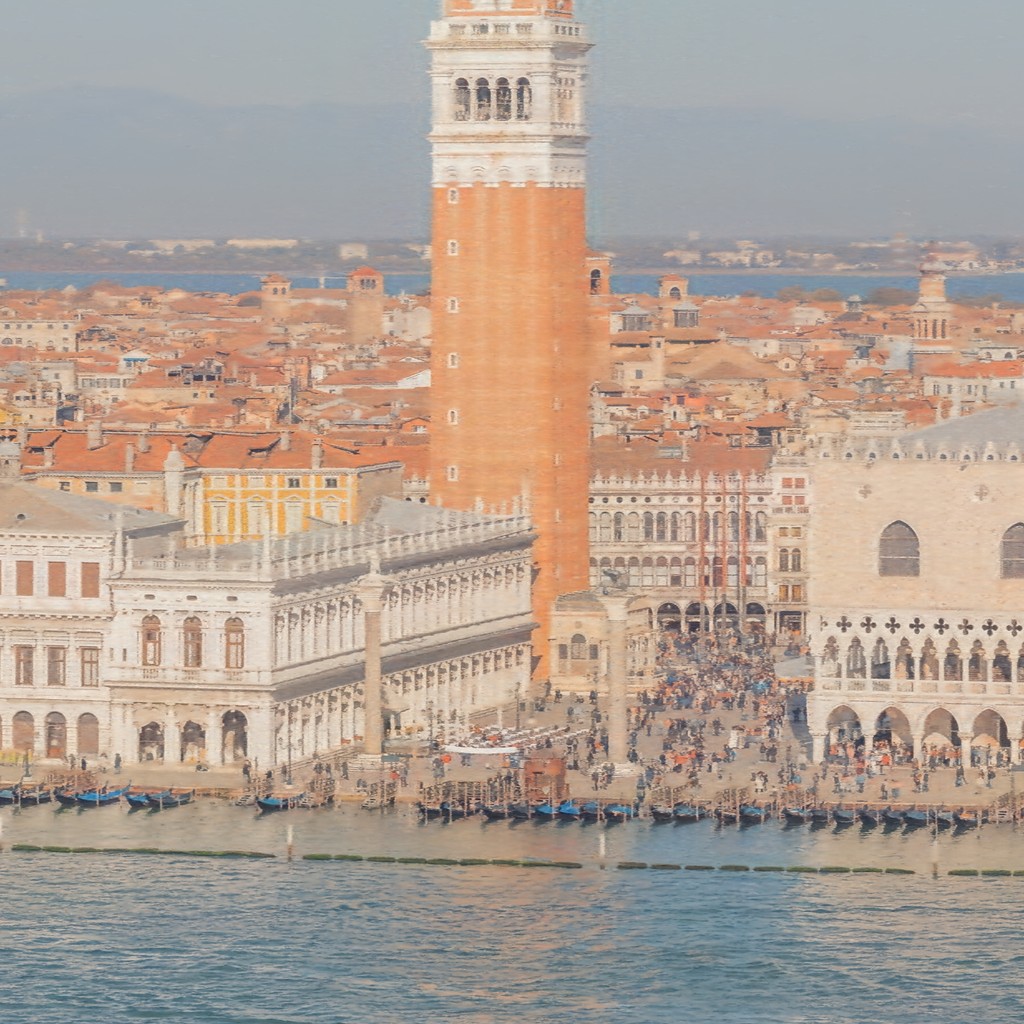} & 
            \includegraphics[width=0.2\linewidth]{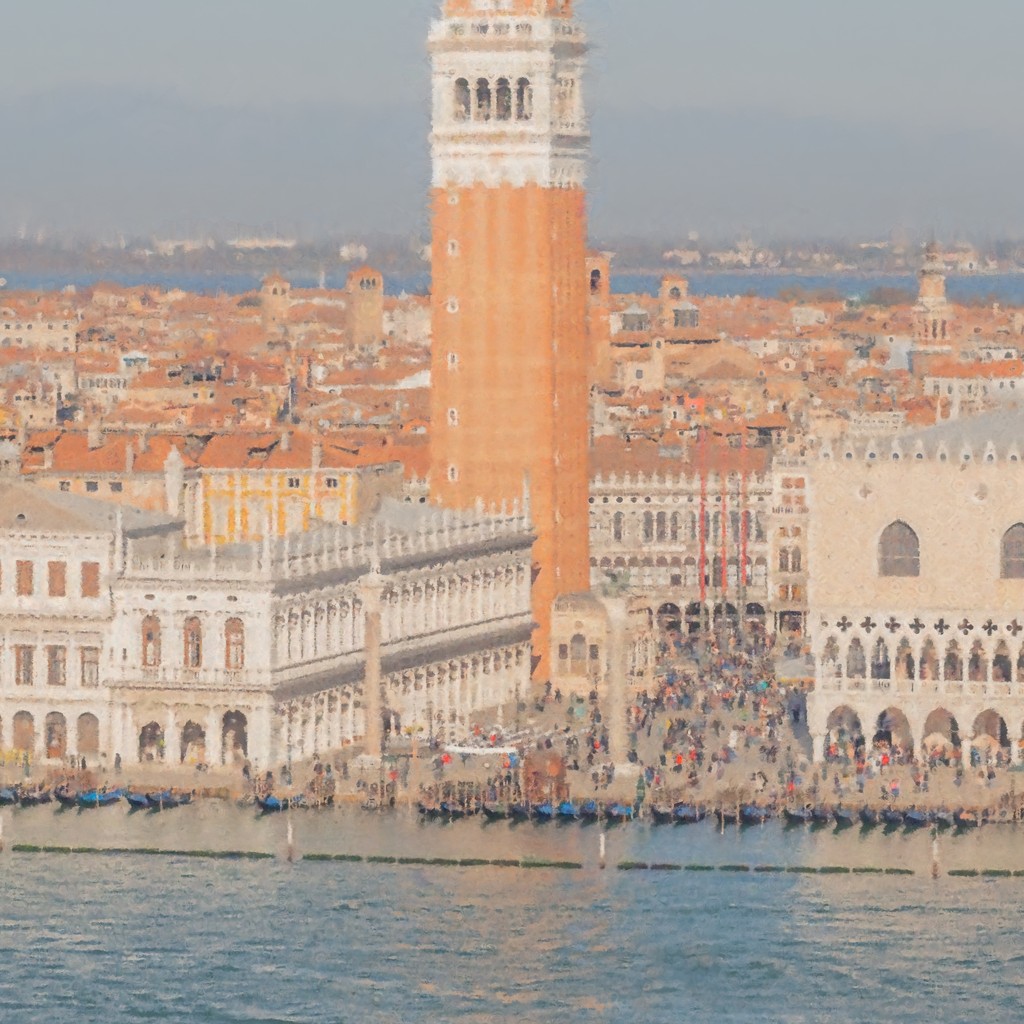} & 
            \includegraphics[width=0.2\linewidth]{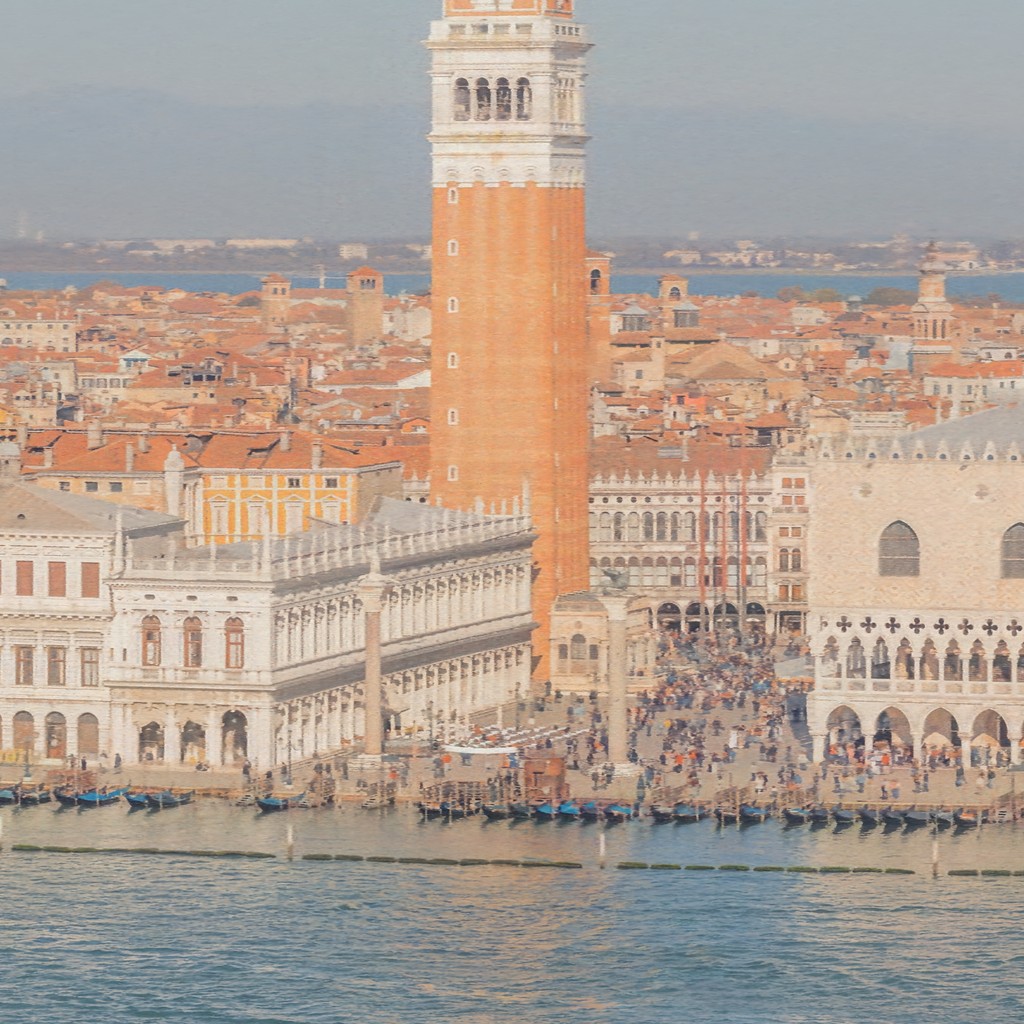} \\
        \rotatebox{90}{\hspace{3.8em}\textit{T=24}} & 
            \includegraphics[width=0.2\linewidth]{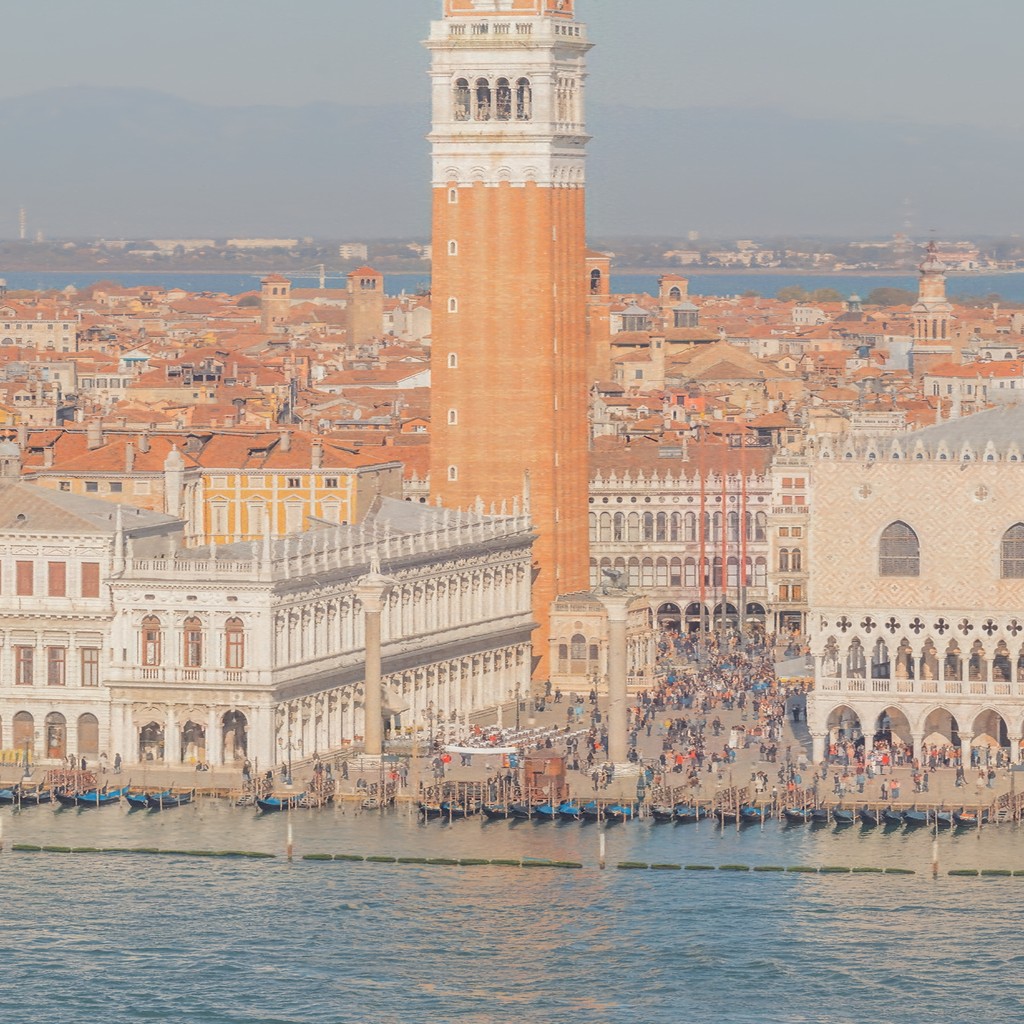} & 
            \includegraphics[width=0.2\linewidth]{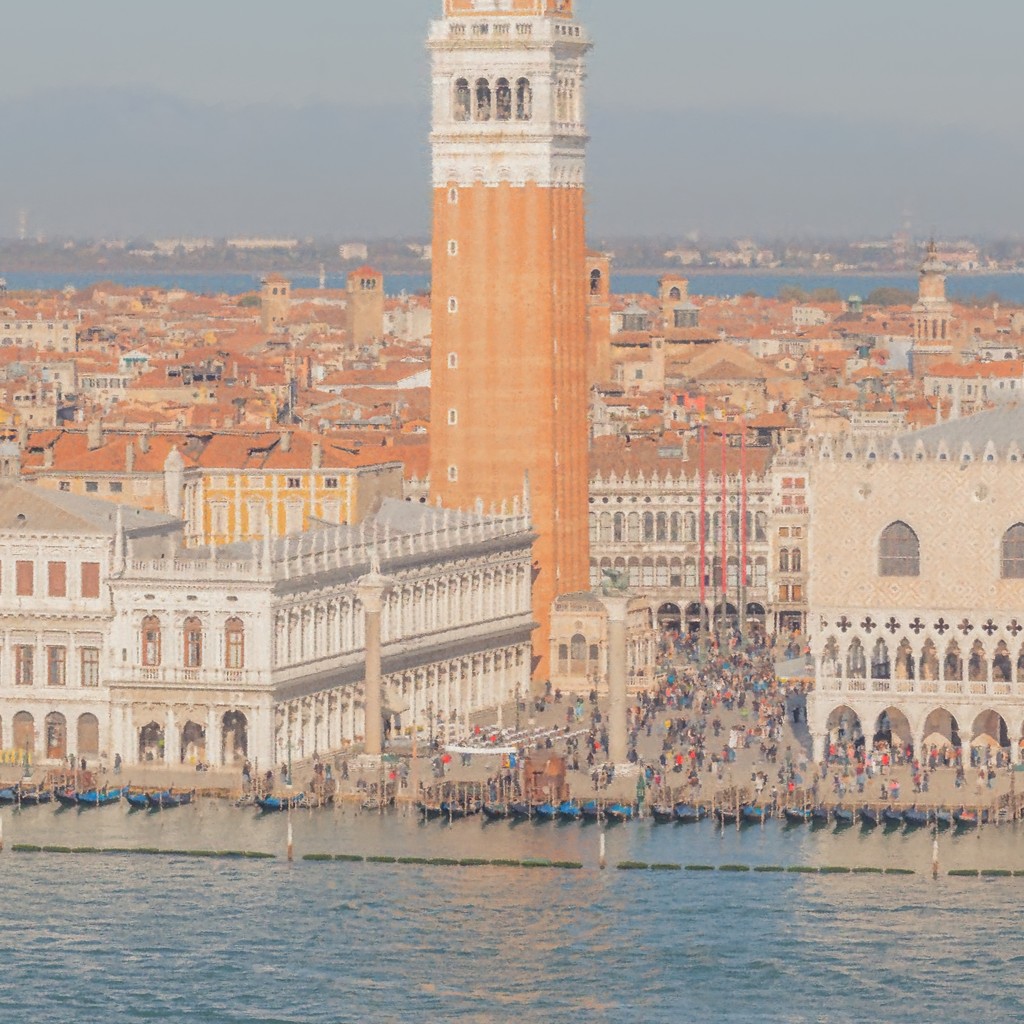} & 
            \includegraphics[width=0.2\linewidth]{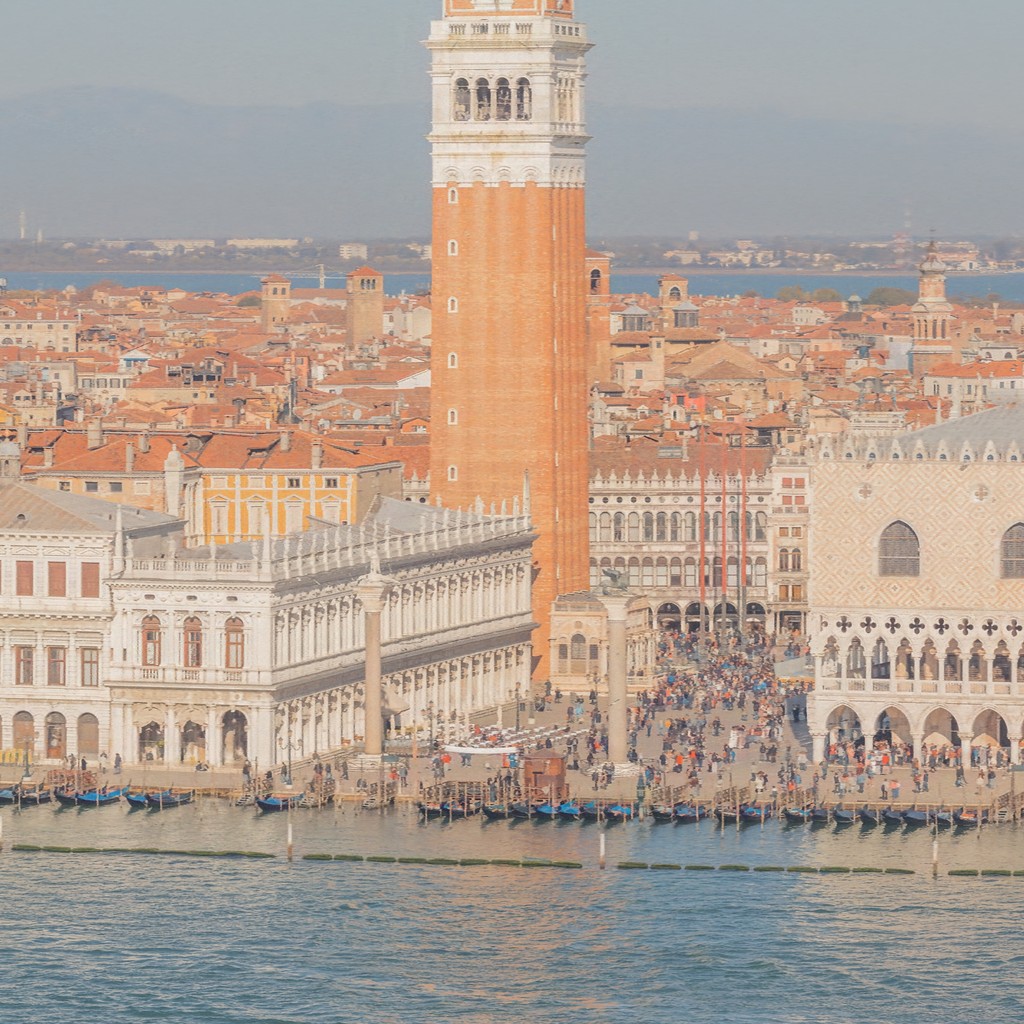} \\

        & \textit{CR=3000} & \textit{CR=100} & \textit{CR=15} \\
        \rotatebox{90}{\hspace{2.5em}\textbf{GaussImg}} & 
            \includegraphics[width=0.2\linewidth]{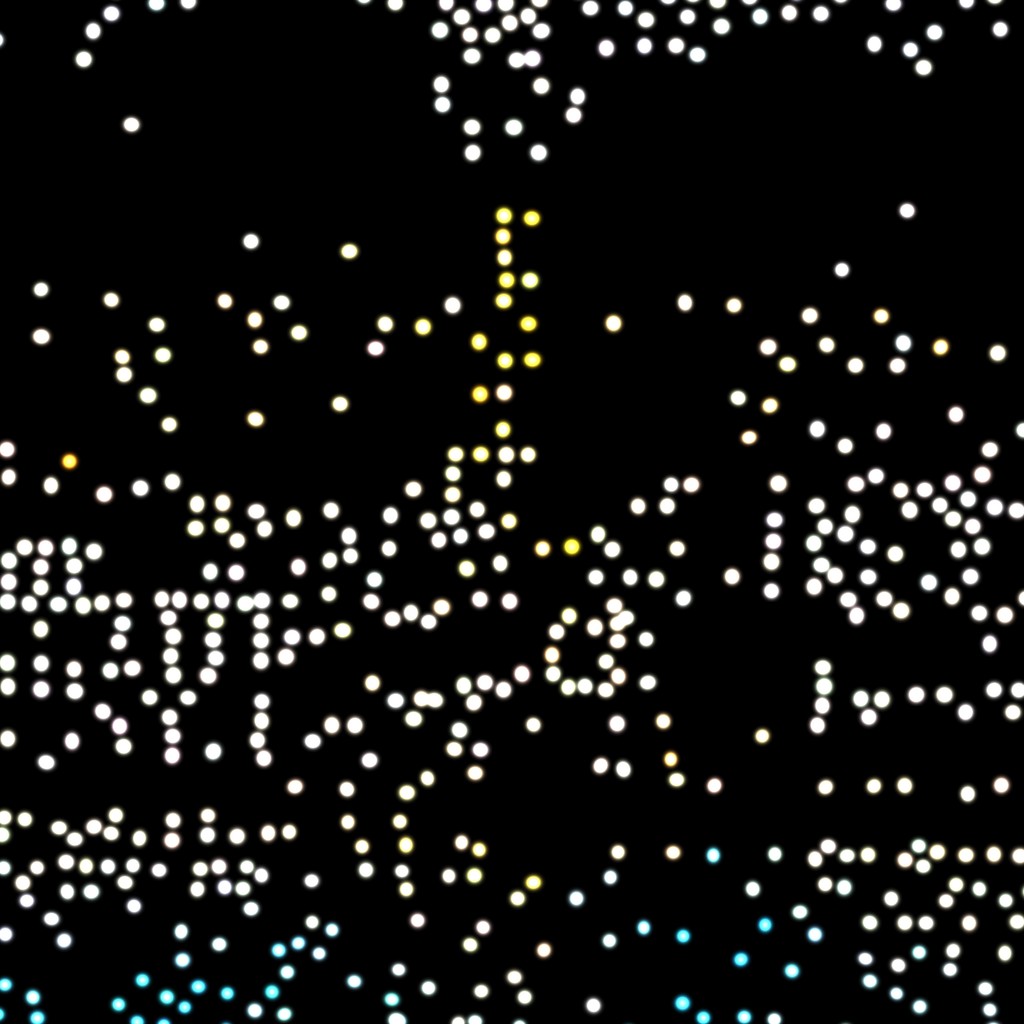} &
            \includegraphics[width=0.2\linewidth]{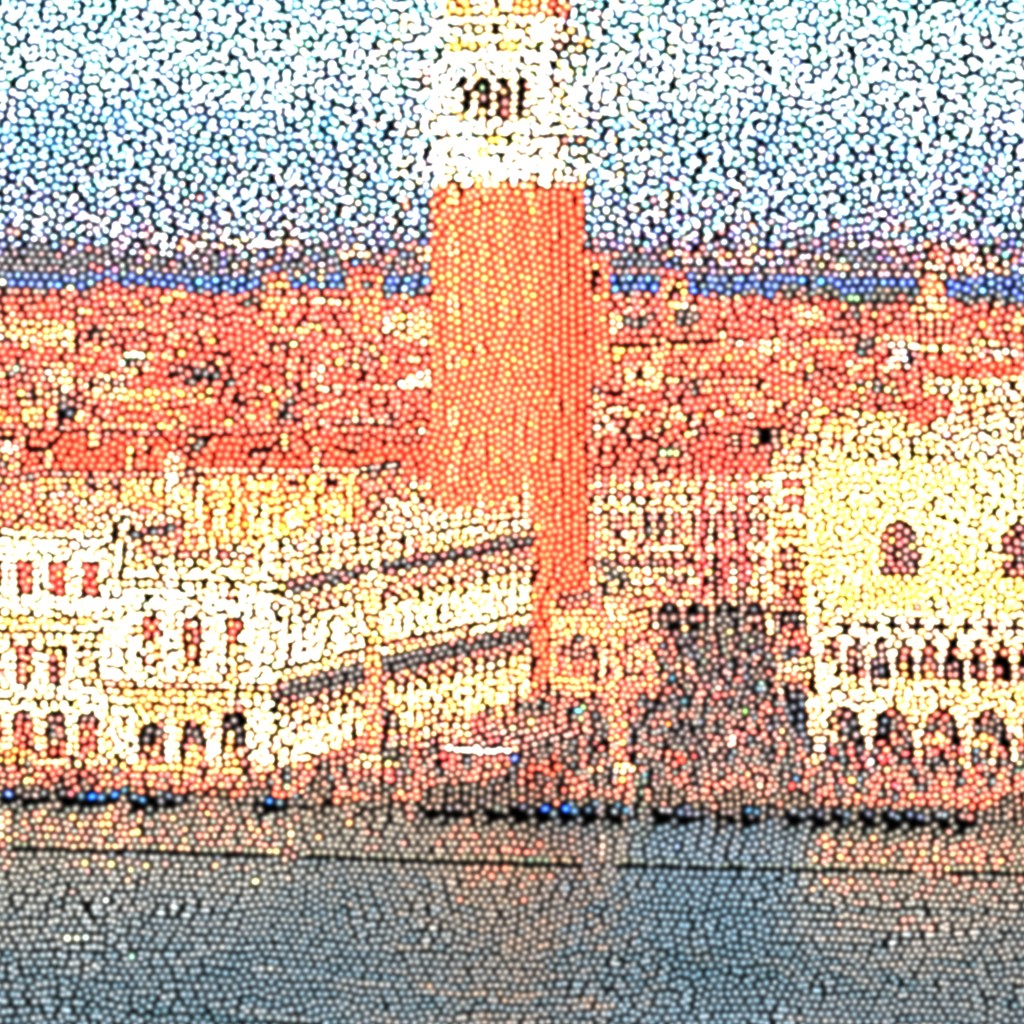} &
            \includegraphics[width=0.2\linewidth]{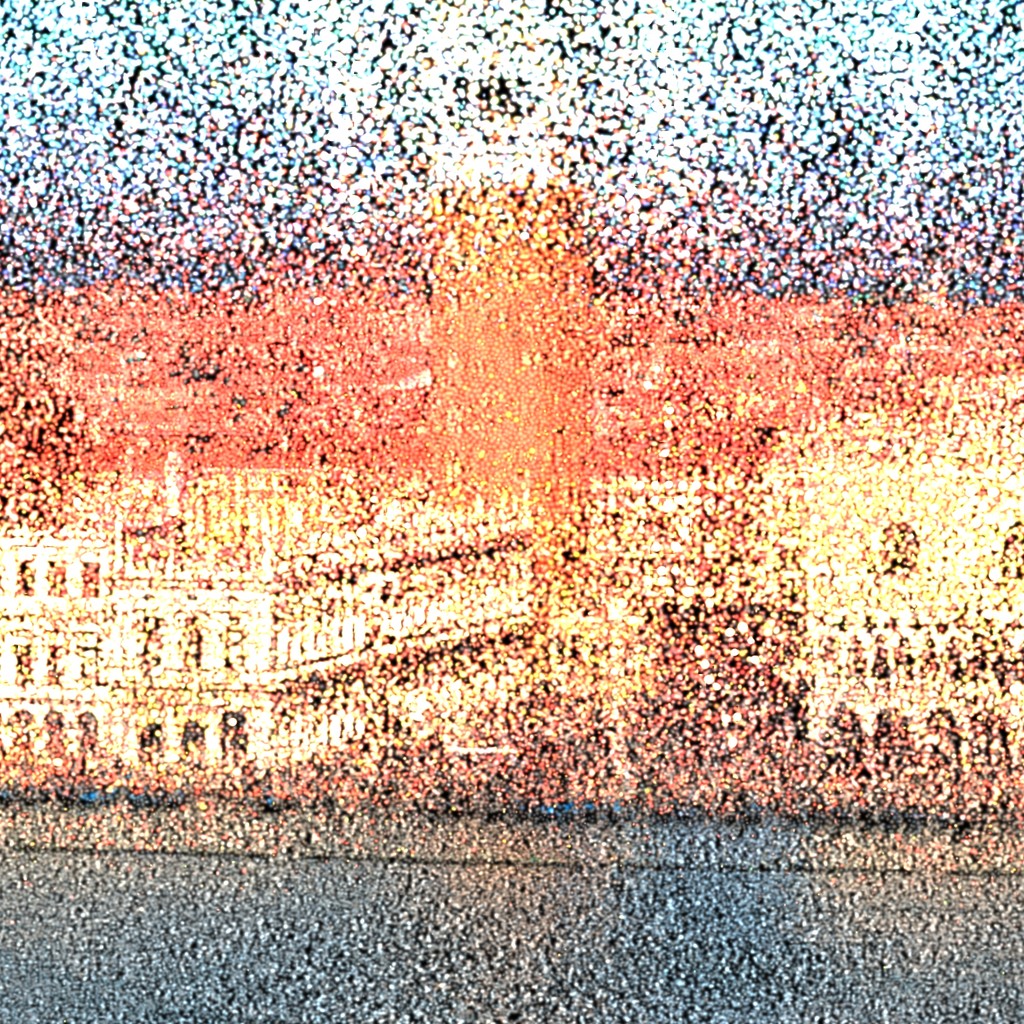} \\
        \rotatebox{90}{\hspace{2.4em}\textbf{SmartSplat}} & 
            \includegraphics[width=0.2\linewidth]{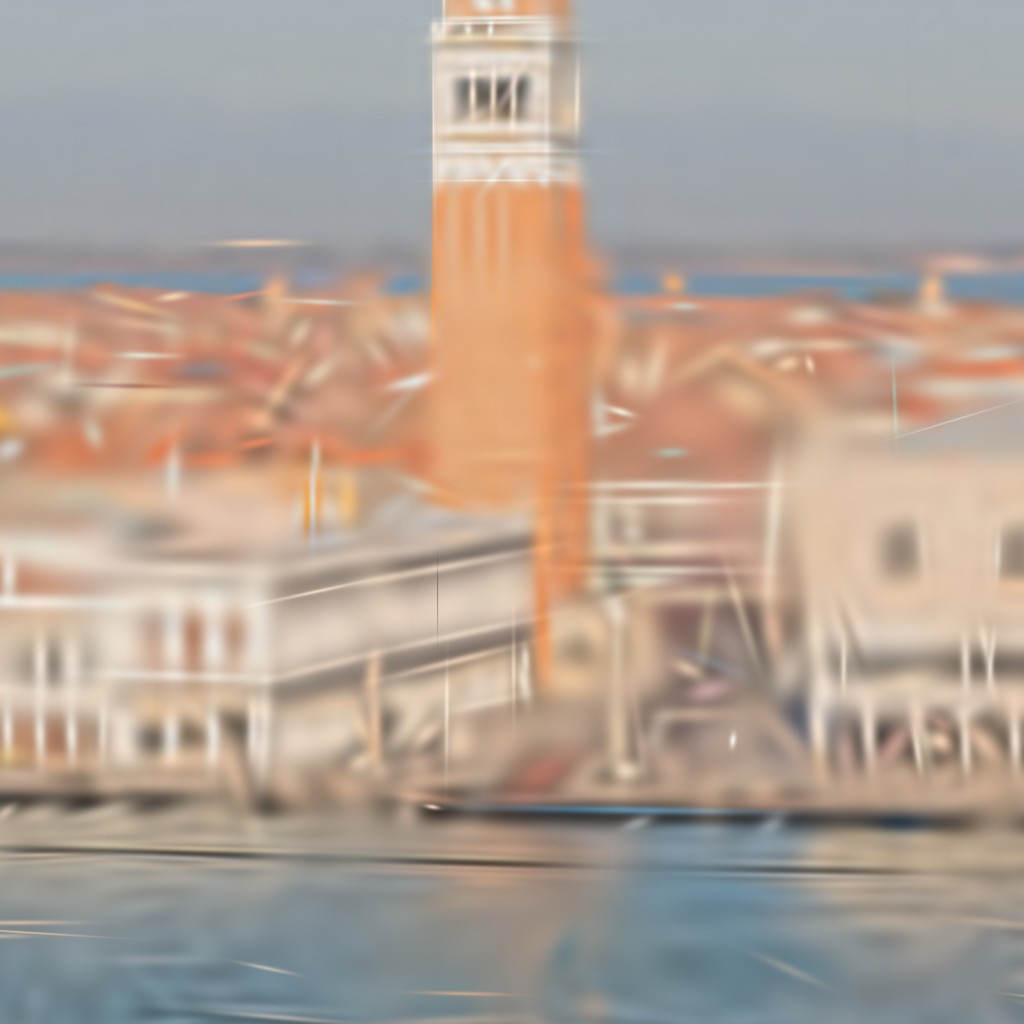} & 
            \includegraphics[width=0.2\linewidth]{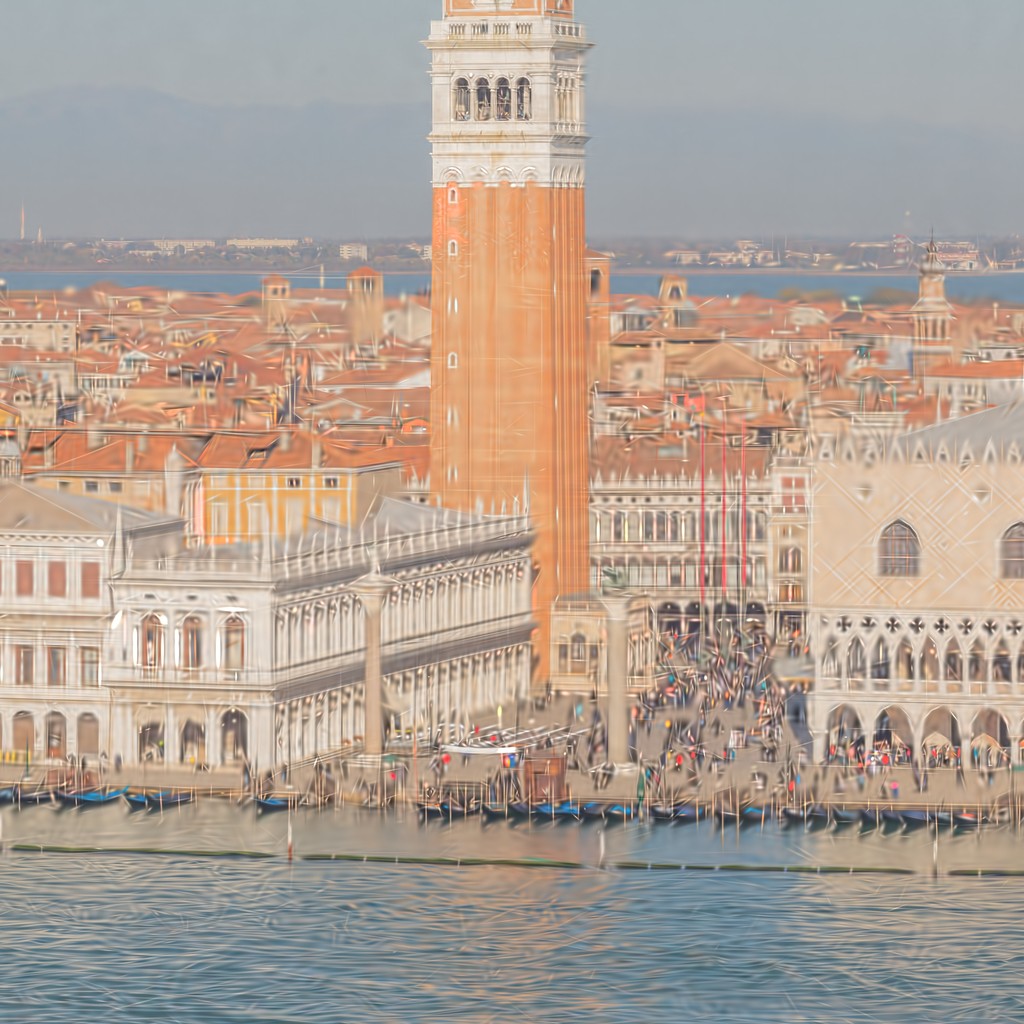} & 
            \includegraphics[width=0.2\linewidth]{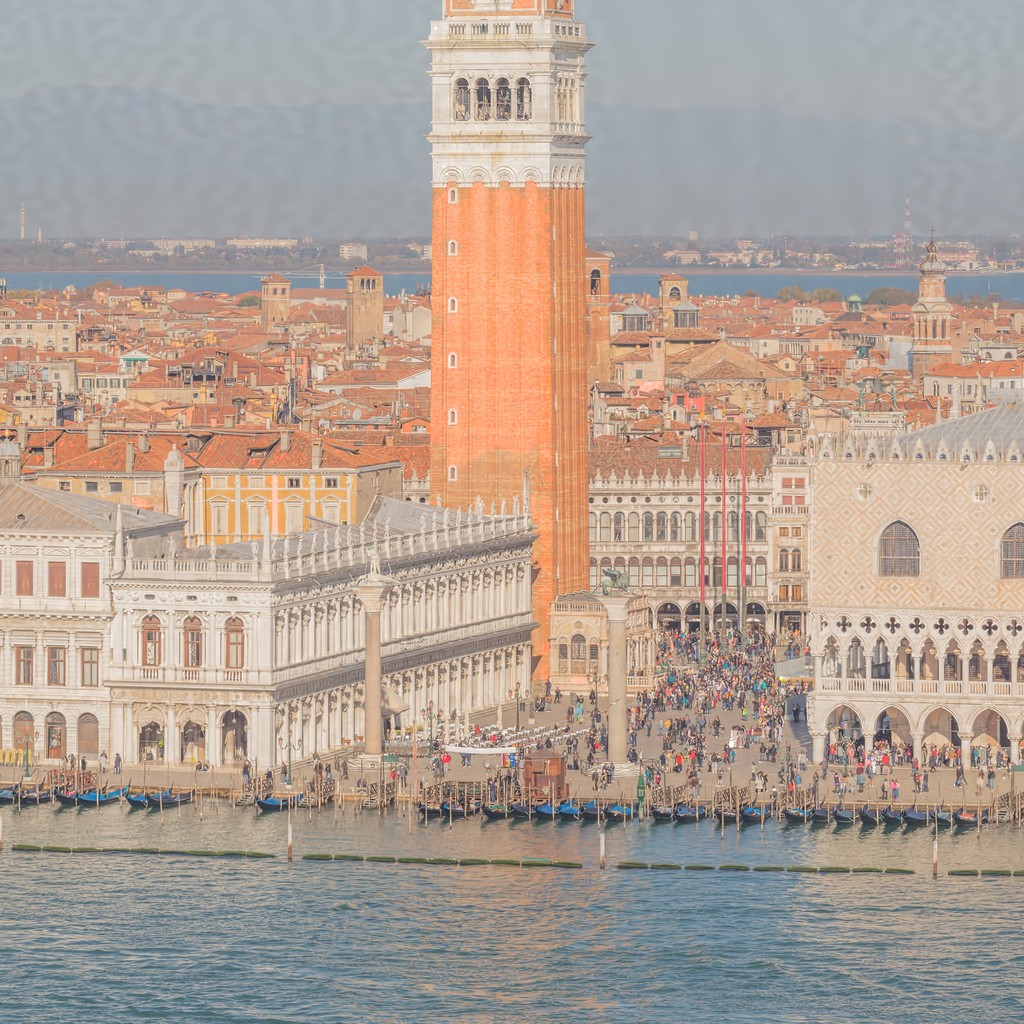} \\
    \end{tabular}
    }
    \label{fig:supp_all_venice}
    \caption{\textbf{Additional fitting qualitative results on \dataname{}}. We show $2048\times2048$ crops from the reconstruction of Venice, for all the method variants.}
\end{figure*}

    Finally, \cref{fig:supp_all_venice} depicts the reconstructions of a $2048\times 2048$ crop from Venice, using the different variants of the considered LRs.
    
\end{document}